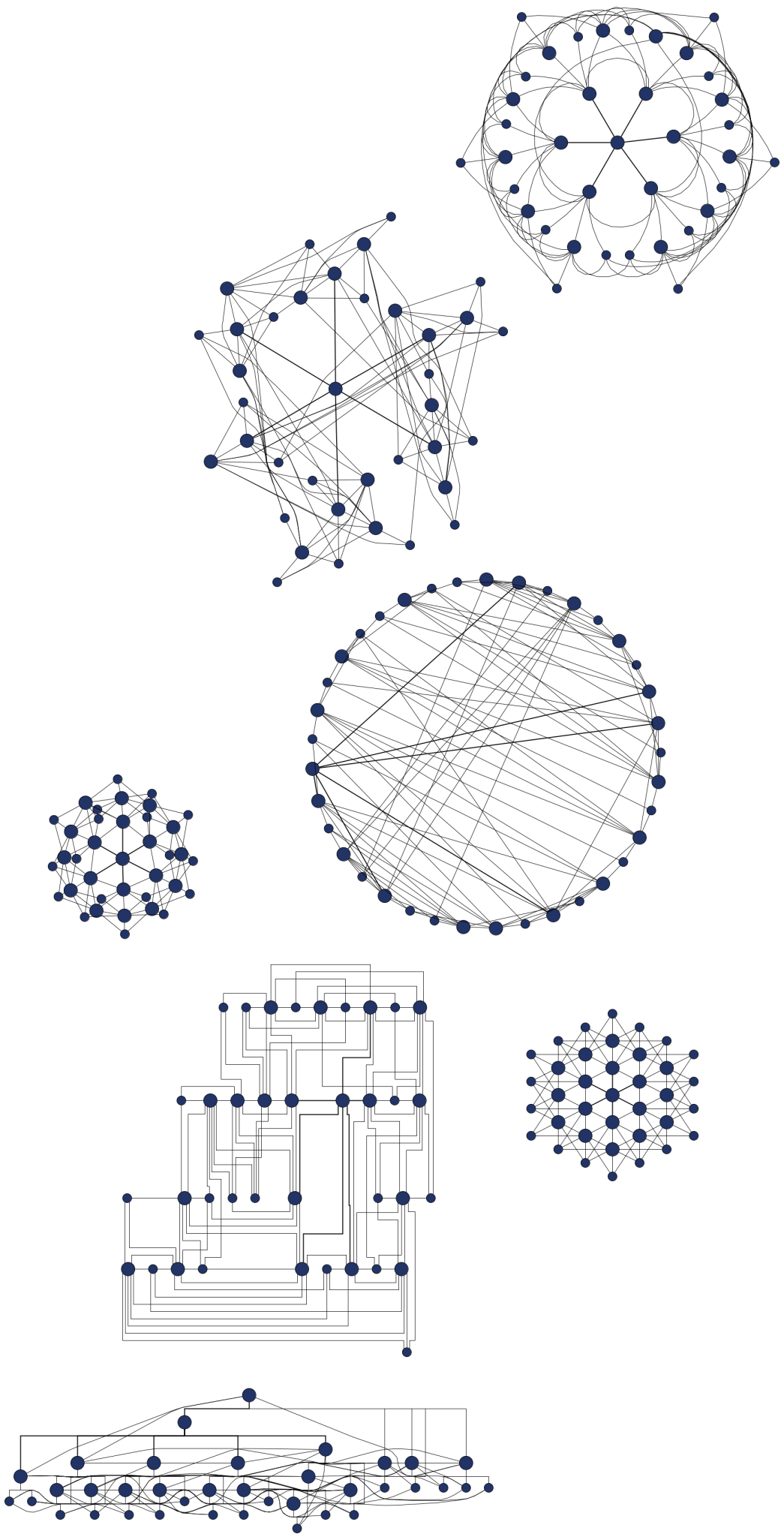

# STRUCTURE OF ARTIFICIAL NEURAL NETWORKS

## Empirical Investigations

JULIAN J. STIER

April 2024

A dissertation submitted to the
Faculty of Informatics and Mathematics
University of Passau

Advisor: Prof. Dr. Michael Granitzer
Chair of Data Science

2nd Examiner: Prof. Dr. Mathilde Mougeot

Rigorosum: 26.09.2024

JULIAN J. STIER

# STRUCTURE OF ARTIFICIAL NEURAL NETWORKS

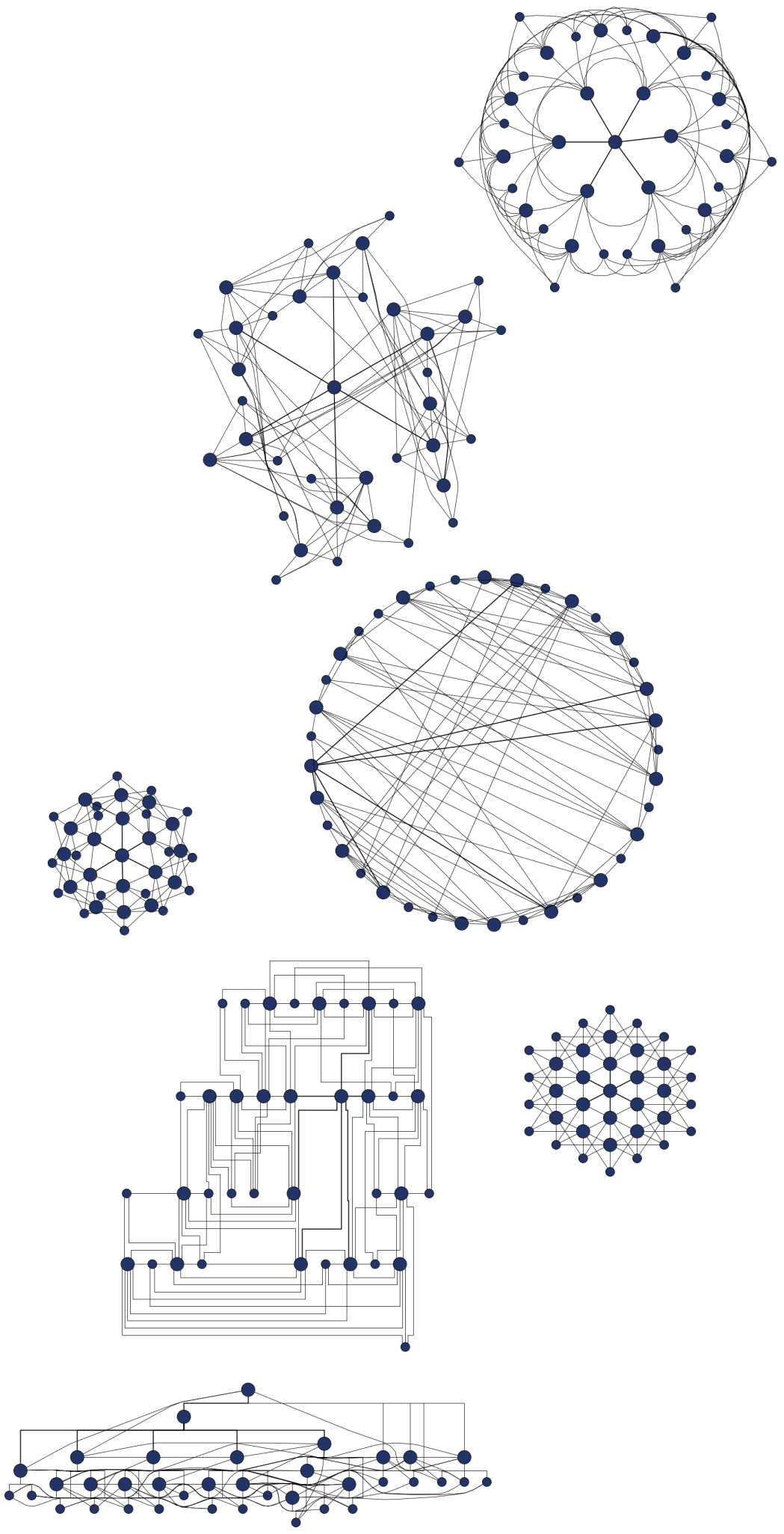

# STRUCTURE OF ARTIFICIAL NEURAL NETWORKS

## Empirical Investigations

JULIAN J. STIER

April 2024

A dissertation submitted to the
Faculty of Informatics and Mathematics
University of Passau

Advisor: Prof. Dr. Michael Granitzer
Chair of Data Science

2nd Examiner: Prof. Dr. Mathilde Mougeot

Rigorosum: 26.09.2024

Modernity's scary,
antiquity's worse,
can't go forward,
can't reverse.
So here's a plan
to ease your unkowing,
ride the horse
in the direction it's going.

— Chuck Lorre

Dedicated to my parents Manfred Stier & Maria Dietz-Stier for their unconditional love and support.

# ABSTRACT

Within one decade, Deep Learning overtook the dominating solution methods of countless problems of artificial intelligence. "Deep" refers to the deep architectures with operations in manifolds of which there are no immediate observations. For these deep architectures some kind of structure is pre-defined – but what is this structure? With a formal definition for structures of neural networks, neural architecture search problems and solution methods can be formulated under a common framework. Both practical and theoretical questions arise from closing the gap between applied neural architecture search and learning theory. Does structure make a difference or can it be chosen arbitrarily?

This work is concerned with deep structures of artificial neural networks and examines automatic construction methods under empirical principles to shed light on to the so called "black-box models".

Our contributions include a formulation of graph-induced neural networks that is used to pose optimisation problems for neural architecture. We analyse structural properties for different neural network objectives such as correctness, robustness or energy consumption and discuss how structure affects them. Selected automation methods for neural architecture optimisation problems are discussed and empirically analysed. With the insights gained from formalising graph-induced neural networks, analysing structural properties and comparing the applicability of neural architecture search methods qualitatively and quantitatively we advance these methods in two ways. First, new predictive models are presented for replacing computationally expensive evaluation schemes, and second, new generative models for informed sampling during neural architecture search are analysed and discussed.

# ZUSAMMENFASSUNG

Deep Learning hat innerhalb einer Dekade die dominierenden Lösungsansätze für zahlreiche Probleme in der künstlichen Intelligenz übernommen. "Deep" bezieht sich auf die tiefen Architekturen mit Operationen in Mannigfaltikeiten zu denen es oft keine unmittelbare Beobachtungen gibt. Für diese tiefen Architekturen wird *by design* eine Form von Struktur vorgegeben – aber was ist diese Struktur? Mit Hilfe einer formaleren Definition der Struktur neuronaler Netzwerke lassen sich Probleme und Lösungsmethoden aus dem Bereich der neuronalen Architektursuche in einem einheitlichen Framework zusammenführen. Sowohl praktische, als auch theoretische Fragen kommen beim Brückenschlag zwischen neuronaler Architektursuche und Lerntheorie auf. Macht Struktur einen Unterschied oder kann sie willkürlich gewählt werden?

Diese Arbeit beschäftigt sich mit den tiefen Strukturen künstlicher neuronaler Netzwerke und beleuchtet automatische Konstruktionsmethoden unter empirischen Gesichtspunkten um Licht in die so genannten “black-box-Modelle” zu bringen.

Unsere Beiträge beinhalten eine Konstruktion graph-induzierter neuronaler Netzwerke, die für die Formulierung von Optimierungsproblemen zur neuronalen Architektursuche genutzt werden. Wir analysieren strukturelle Eigenschaften für verschiedene erwünschte Richtwerte neuronaler Netzwerke wie beispielsweise Korrektheit, Robustheit oder Energieverbrauch und erörtern den Einfluss des Faktors Struktur auf diese Werte. Ausgewählte Automatisierungsmethoden zur neuronalen Architektursuche werden diskutiert und empirisch untersucht. Mit Erkenntnissen aus der Formalisierung graph-induzierter neuronaler Netzwerke, aus der Analyse struktureller Eigenschaften und aus den Vergleichen zur Anwendbarkeit von Methoden zur neuronalen Architektursuche entwickeln wir diese Methoden auf zweierlei Arten weiter. Erstens stellen wir neue Vorhersagemodelle vor, die rechenintensive Trainings- und Bewertungsschema teilweise während einer neuronalen Architektursuche ersetzen und zweitens untersuchen und diskutieren wir neue generative Modelle, um während einer neuronalen Architektursuche geschickt neue Architekturkandidaten zu generieren.

*I propose to consider the question,*
*"Can machines think?"*
*This should begin with definitions of the meaning of the terms "'machine"' and "'think"'.* [..] *Instead of attempting such a definition I shall replace the question by another, which is closely related to it and is expressed in relatively unambiguous words.*

— Alan M. Turing [286]

# ACKNOWLEDGMENTS

Taking a first step into the world of academic work by pursuing a doctorate is a tremendously exciting experience, both in its lows and in its highs. Never before have I been so active in the observation and reflection of my own developmental process as I have been over the past seven years. This has only been possible in large part because of a vibrant and fostering environment that allowed for academic freedom, support in times of doubt, honesty, and space for creativity.

Therefore, I want to first and foremost thank my doctoral advisor Prof. Dr. Michael Granitzer who not only motivationally leads with a curious mind, inspiring ideas and constructive criticism, but also goes to great lengths to make this development environment possible. Thank you for shaping this journey and for your continuous support. It is a pleasure learning from you.

With the self-commitment to put my individual contributions on a more formal basis, a larger work has emerged than I had imagined. Capturing the volume of work and providing feedback takes a lot of time and effort. I would like to express my appreciation to Prof. Dr. Mathilde Mougeot for agreeing to act as external examiner and to all the other members of the examination committee.

Such a long time, with its struggles and achievements, does not pass without good company, both in research and in private. I am very grateful to Jörg Schlötterer for his time & support during quiet academic times, to Christofer Fellicious & Lorenz Wendlinger for fantastic humor, to Sahib Julka for great taste in music, and to Thomas Weißgerber & Mehdi Ben Amor for diligent collaboration on infrastructure topics beyond borders. Many thanks to you and many colleagues for shaping office culture: Annemarie Gertler-Weber, Johannes Jurgovsky, Wolfgang Mages, Emanuel Berndl, Torben Schnuchel, Stefanie Urchs, Fatemeh Rizi, Armin Gerl, Max Bachl, Andreas Wölfl, and also Ramona Kühn, Dr. Jelena Mitrovic, Stefanie Riederer, Jan Wackerbauer, Kanishka Ghosh, Saber Zerhoudi, Stefan Becher, Ashish Ashutosh, Max Klabunde, Bianca Meier, Dr. Florian Lemmerich, Dr. Harald Kosch, Dr. Tomas Sauer, Sathish Purushothaman, Michael Dinzinger, Khouloud Saadi, Felix Bölz, Paul Lachat, Sascha Schiegg, Harshil Darji, Dr. Pierre-Edouard

Portier, Ousmane Touat, Simon Neumeyer, and many students I had the pleasure to collaborate with. Looking back on my early teaching days in theoretical computer science and databases, I also have good memories working with Christina Ehrlinger, Alex Stenzer, Matthias Schmid, and Dr. Rupert Hölzl. I'm very grateful to Stephanie Pauli for her constant support and positive spirit. And to Alizée Bertrand & Ophélie Coueffe and the IRIXYS network for their support on a research exchange to INSA Lyon and beyond.

An academic journey is only one of three pillars under a spirit of gōngfu 功夫. I am very grateful to Florian Völkl for accompanying my martial arts journey over the last decade and not just being my shīfu 师傅 but my shīfu 师父. This includes the Nam Wah Pai school in Passau with Alois Himsl, Reinhold Schneider, Patricia Bachmann, and many more with its members and larger network of 南華派.

Last but not least I am extremely grateful of having my family and friends. Many of you contributed with feedback or open ears to my writings. This work would not have been possible without you.

# CONTENTS

# LIST OF FIGURES

# LIST OF TABLES

# LISTINGS

# LIST OF ALGORITHMS

# ACRONYMS

**AI** Artificial Intelligence 9, *Glossary:* Artificial Intelligence

**BNN** Biological Neural Networks 65, *Glossary:* biological neural network

**CE-loss** The cross-entropy is a common loss function for multi-class classification problems. 76, 78, *Glossary:* cross-entropy loss

**CNN** The term *convolutional neural network* refers to either the convolutional layer or a neural architecture containing such convolutional layers. A convolutional layer uses a small kernel weight matrix which is slided across the input to compute its output. 79–81, 95, 122, 251, *Glossary:* convolutional neural network

**CT** A computational theme is a directed acyclic graph restricted with a vertex collapse condition. 131, 133, 135, 146, 147, 149, 150, 153, 156, 158, 163, 173, 175–182, 184–187, 189, 190, 192, 196, 197, 199, 217, 249, 283, 292, *Glossary:* computational theme

**CT-NAS** A benchmark dataset of 863 CTs (a subset of directed acyclic graphs) and over 287,265 trained models with evaluation measures on multiple datasets. 16, 22, 131, 146–150, 154, 179, 180, 190, 196–198, 268, 289, 293

**DAG** A directed acyclic graph is a digraph that contains no cycles. 3, 18, 38, 39, 74, 75, 102, 103, 107–111, 113, 114, 117, 118, 123, 124, 127, 129–132, 163, 173–175, 224–226, 238, 284–286, 291, 292, 310, *Glossary:* directed acyclic graph

**DARTS** DARTS stands for differentiable architecture search and can be considered as a type of neural architecture search (NAS) in which sub-modules are integrated into a hyper-architecture and learned jointly in a differentiable manner. 16, 121, 127, 219, 221, 236–239, 248–253, 256, 257, 283, 287, 288, 292, 295, 296, *Glossary:* differentiable architecture search

**DeepGG** DeepGG stands for Deep Graph Generator and is a state-machine based graph generator. 262–266, 271, 273–275, 288, 290

**DGMG** DGMG stands for Deep Generative Model for Graphs and is a state-machine based graph generator. 260, 262, 263, 265, 266, 271, 275, 288

**DNN** A DNN is a statistical model composed of multiple layers, inspired by biological neural nets. 3, 51, 55, 85, 91, 101, 105, 106, 158, 162, 163, 165, 172, 196, 197, 206, 212, 216, 218, 237, 248, 263, 271, 273, 274, 279, 283, 286, 289–294, 305, *Glossary:* deep neural network

**DSM** A deep state machine is a finite state automaton with learned transitions. 271, 273, 274, 278, *Glossary:* deep state machine

**ERM** Empirical risk minimisation 78, *Glossary:* empirical risk minimisation

**FaDE** FaDE stands for **F**ast **D**ifferentiable **E**stimation. The method emplos a differentiable hyper-architecture to combine architecture parameters of cells into FaDE-scores. These FaDE-scores are used to navigate a pseudo-gradient step in an open graph-based search space. 17, 250, 254–257, 288, 290, 296

**GCN** A graph convolutional network is a neural network with a layer applicable to graphs. 85, *Glossary:* graph convolutional network

**GED** The graph edit distance is a distance between graphs based on elementary operations. 38, 194, 195, 277, *Glossary:* graph edit distance

**GNN** A graph neural network refers to a neural network layer that operates on graph-based data through the usage of a graphs adjacency matrix or its Laplacian or it refers to a neural architecture containing such operational layers. Graph neural networks are part of the field of geometric deep learning and extends capabilities of DNNs to other data topologies such as graphs or sets. 85, 257, *Glossary:* graph neural network

**GRAN** GRAN [160] stands for Graph Recurrent Attention Network and is a graph generative model that employs recurrence, attention and message passing to sample blocks of constructive steps to learn a distribution of graphs. 38, 260–262, 265, 266, 268, 269, 275, 288

**GraphRNN** GraphRNN stands for Graph Recurrent Neural Network and is a graph generative model by You et al. [328]. 38, 260, 262, 265, 266, 275, 288

**HPO** Hyperparameter optimisation 217, *Glossary:* hyperparameter optimisation

**i.i.d.** independent and identically distributed 19, 49, 56, 60, 78, 229, 230

**KLD** Kullback-Leibler divergence is a common statistical divergence. 56, 58, 61, 62, 70, 78, *Glossary:* Kullback-Leibler divergence

**MAE** The mean absolute error (MAE) is an error based on the $l_1$-norm. It can be used for optimisation (training) or evaluation of statistical models. 87, *Glossary:* mean absolute error

**MLE** Maximum Likelihood Estimation 60, 61, 64, 70, 76, *Glossary:* maximum likelihood estimation

**MMD** Maximum Mean Discrepancy 59, 272, *Glossary:* maximum mean discrepancy

**MSE** The mean squared error (MSE) is an error based on the $l_2$-norm. MSE is used for training and also sometimes for evaluation of statistical models. 87, 90, *Glossary:* mean squared error

**NAS** Neural architecture search is a field concerned with the automated design of neural models, see neural architecture search. 3, 9, 15, 43, 63, 91, 94, 115–119, 121, 126, 128, 135, 150, 162, 167, 182, 185, 189, 217–223, 225, 227, 235, 237–239, 243–245, 247–250, 254–257, 259, 268, 269, 278, 283, 286–290, 292, 293, 295, 296, *Glossary:* neural architecture search

**ReLU** The rectified linear unit is a popular activation functions. 67–69, 158, 161, 164, 165, 267, *Glossary:* rectified linear unit

**RF** A random forest is an ensemble of trees from which a decision is drawn by bagging. 170, 249, *Glossary:* random forest

**SGD** Stochastic gradient descent is a family of gradient-based optimisation techniques. 70, 295, *Glossary:* stochastic gradient descent

**SV** Shapley Values are attributions to players in a game that reflect their individual contribution, obtained through the Shapley Value method, a solution concept for a game-theoretical setting based on marginal payoff. 43, 209–211, 216, *Glossary:* Shapley Value

**UAP** Universal Approximation Property 93, 94, 112, 114, 115, 120, 157, 164, 292, *Glossary:* universal approximation property

**w.r.t.** with respect to 19, 71, 73, 74, 76, 78, 86–88, 95, 97, 119, 123, 125, 126, 180, 187, 188, 190, 192, 195, 208, 255, 269, 275–277, 286, 288

**WS** Watts-Strogatz *Glossary:* Watts-Strogatz model

Part I

# FROM NEUROBIOLOGY TO MACHINE LEARNING

Neurobiology was and is an inspiring source for the development of machine and especially deep learning. The human brain consists of billions of neurons and trillions of connecting axons or dendrites and resembles a complex structure. Questions around the meaning of this structure and its properties motivated our research goals and questions in deep learning.

# 1 INTRODUCTION

*The following entails:*

The functional form of deep neural networks (DNNs) is usually introduced through successive applications of $\sigma(W\mathbf{x} + B)$ with layer-wise weights $W$ and biases $B$, an input feature vector $\mathbf{x}$ and a non-linearity $\sigma$. On the other hand, neural architecture search (NAS) as a field concerned with the automated discovery of neural architectures, considers significantly more complicated forms. For example, architectures in NAS are often treated as labelled directed acyclic graphs (DAGs). But what is an architecture or a DAG in the context of the above form?

The two fields open up a huge gap between how deep neural networks are properly defined on the theoretical side and what is actually meant by the architecture of a DNN in NAS on the applied side. The research goals and questions of this work are located in this field of tension and while our work is not closing this wide gap, it proposes ideas, formalisations and empirical observations that could contribute to a better understanding and future development of deep learning theories and NAS-methods.

## 1.1 THE GOAL OF THIS THESIS

This work is concerned with the relationship between **deep neural networks** and their **structure**. For this purpose, the work delves into the formal definition of deep neural networks and into how structure in the form of graphs can be made explicit. Making this formal underlying more explicit frames the term of an *architecture*. This formulation then serves to understand both the properties of different neural architecture search methods and the analysis of deep neural network structures. Experiments with pruning, growing, evolutionary or genetic searches are then repeatedly brought back to this formal underlyings as to sharpen the interpretation and arguments on their advantages and disadvantages.

### 1.1.1 *Deep Neural Networks*

Deep neural networks are the first object of our study. They are originally inspired from biological neural networks (BNNs) - hence their name [240]. There is a common theme for what deep neural networks (DNNs) are actually about but because neural networks originate less from a formal but more from a biologically inspired ground, there are variations in how to refer (rigorously) to neural networks. For this reason, it is difficult to arbitrarily look at structures of different neural networks and conclude from properties of the one properties of the other. For example, some formulations of neural networks are inherently more powerful in terms of expressiveness or learning capabilities than others. Therefore, to study and understand the relationship between deep neural networks and their structure, it is vitally important to clarify what both objects exactly are.

Superficially, a neural network can be seen as a function $f : D \to C$ with input and output spaces $D$ and $C$ in the reals. This neural network $f$ can be highly non-linear because of the repeated application of intermediate activation functions. Before and after the application of activations, affine transformations e.g. of the form $x \mapsto Wx + B, x \in \mathbb{R}^{d_1}, W \in \mathbb{R}^{d_2 \times d_1}, B \in \mathbb{R}^{d_2}$ take place with input dimension $d_1 \in \mathbb{N}$ and output dimension $d_2 \in \mathbb{N}$.

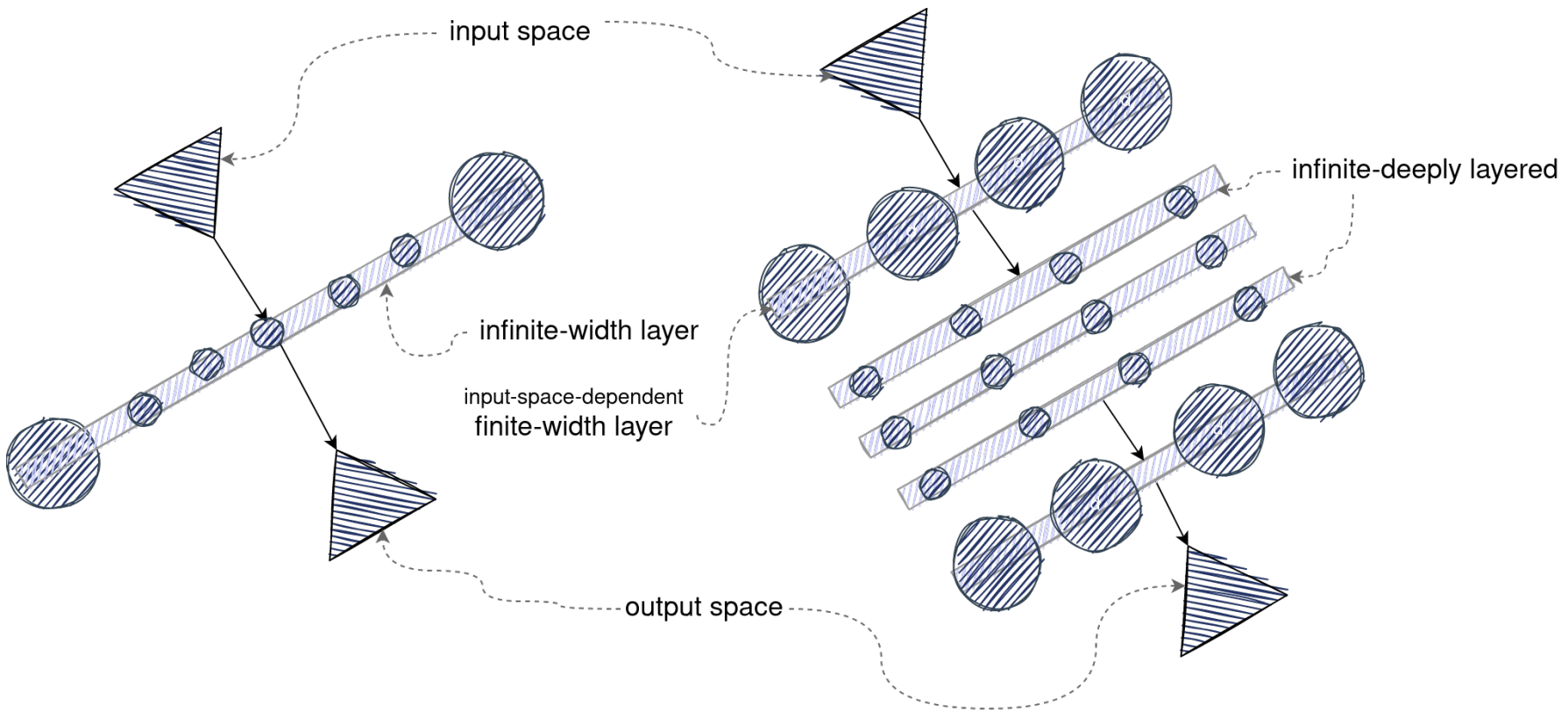

**Figure 1.1:** Sketch of two classes of neural networks capable of universal approximation: single-layered infinite-width neural networks and bounded-width infinite-depth neural networks. Both act over an input space of $\mathbb{R}^n$ with the first class having an arbitrary size of layers $\in \mathbb{N}$ hidden layers and the second class having an arbitrary number "width $\in \mathbb{N}$" of finite-sized layers with width $> n + 1$.

### 1.1.2 *Deep Learning Architectures*

The second object of study is the structure of neural networks. Structure is a reductionist view of a neural network and is closely related to the architecture of a neural network. We treat structure as a graph or sets of graphs and clarify what we understand of an architecture to frame our research questions under this formalism.

Consider this first example of an architectural choice for the width $d_2$ of a neural network: the inner working of the exemplary neural network $f$ carries a number of $d_1 \cdot d_2 + d_2$ parameters contained in weights $W$ and biases $B$. This number $d_1 \cdot d_2 + d_2$ can be e.g. set to $784 \cdot 1000 + 1000 = 785,000$ parameters for a single linear layer of a neural network for the MNIST dataset. While the size of the input space of 784 is defined by the representation of our problem at hand – MNIST digits of $28 \times 28$ pixels – the choice of the size of $d_2$ is not as obvious. Often, the choice of $d_2$ is based on expert knowledge. We will later elaborate from results of universal approximation theorems that there exist suggestions for such a number. The width $d_2$ is an architectural choice and implies a structure of the neural network having at least one hidden layer of such width.

But not only the layer size $d_2$ of the parameter matrix $W$ needs to be chosen: subsequent layers can vary in size and connectivity and employed activation functions can change in their functional form. Intermediate functions can even have their own parametric form. The combinatoric space of architectural choices then explodes drastically. Today's deep neural networks are of the size of up to hundreds of billions of parameters [29] – and that excludes architectural choices of how they are composed.

Graphs enable a natural description for the structural composition of neural networks. Directed acyclic graphs are the technical underlying mode of operation for inferencing, auto-differentiation and even parallelisation of many modern deep learning libraries. This work shapes the structure of neural networks as directed acyclic graphs.

### 1.1.3 *DNNs and Structure in Three Research Complexes*

With neural networks on the one hand and their structure on the other, this work is organised in three research complexes. The first complex tackles the theoretical foundation of deep neural networks induced by directed acyclic graphs. Based on this formalism, the second complex concerns the analysis of the structure of neural networks. The analytical results reveal the complexity and diverse dependencies between application domain, neural network structure and various other hyperparameters and motivate the investigation of non-linear automated methods. These automated methods are part of the third complex which illustrates different possibilities for assumptions and solution approaches to automatically construct neural networks jointly with distinct architectural structures.

### 1.1.4 *Research Complex I: Formalism of Neural Network Structure*

The first complex seeks to formalise neural networks and their structure. Fields such as neural architecture search progressed in unifying their terminology and introducing benchmarks to compare automated methods. The employed formalisms, however, differ heavily. Further, capabilities of neural networks are often not properly defined and with the enormous gap between practical neural architecture search and analyses in neural network theory comes a lot of ambiguity in terms of what the actual problems are about.

Learning theory still treats neural networks as consecutively layered affine transformations with non-linearities in between [30, 43, 79, 216, 223]. But neural architecture search seeks to construct complex architectures over directed acyclic graphs with millions to trillions of parameters. If formal theory were completed, there would be no need for structure beyond layers but if empirical observations with successful models are right and their architecture is more than a technical side effect, structure could have a significant influence, e.g. on expressivity or energy consumption.

The research complex on formalising a structure of neural networks aims to fill the gap between neural architecture search as an applied field and underlying theoretical considerations to bring more clarity into the influence and effect of neural architecture search methods. Naturally, the following question arises:

> ⋄ Research Complex I
>
> How can the structure of deep neural networks be defined such that structural optimisation problems can be formulated?

With Part iii, this work casts the relationship between neural networks and their structure as an constructive step from a graph to a function space[1] $\mathcal{A}(G)$ and can be seen as a forgetful functor from a neural network to its underlying graph structure. Many automated machine learning techniques for deep learning models can then be treated as a multi-level optimisation problem. This multi-level optimisation problem encompasses data for learning and generalisation on the one hand and efficient architectural search through different levels of continuous and combinatorial spaces on the other hand.

But there are also downsides involved. A typical analytical function space would provide mathematical benefical properties such as operations that have relationships between the elements of the function space and its codomain. Graphs are a much looser topological structure

1 The term function space for neural networks is very loosely used for a set of functions for which some operations such as concatenation can be easily defined. For a more rigorous terminology we refer to the technical background.

but are motivated from an applied perspective for their expressiveness.

> ◇ Research Complex I
>
> Are graphs a suitable representation for neural network structure?

The usage of graphs as partial representation of neural architectures stems from two conflicting perspectives: formal unification and expressivity. Automating neural design tries to **unify** very heterogenic parameteric choices in one formalism. Unified formalisms restrict **expressivity** and give rise to new generalisations which potentially render preceding formalisms obsolete or as a special case. Under this conflict, graphs are a valuable tradeoff between formalism and expressivity: graphs are combinatorial in nature which has weaknesses and strengths. An apparent weakness is the difficulty of enumerating and thus sampling or comparing them. Their strength lies in a definite expressivity that goes far beyond other representational objects. Graph theory is a well-established field with deep insights to support that choice, which we discuss in section 6.1.

The formalisation also leads to further detailled theoretical questions:

> ◇ Research Complex I
>
> What are the difficulties, advantages and disadvantages of the proposed formalism?

### 1.1.5 *Research Complex II: Analysis of Structures of Neural Networks*

Many insights on neural networks are stemming from empirics, especially because their size has grown to magnitudes in which emergent phenomena appear that have not been observed with smaller models. Examples for emergent phenomena in neural networks are grokking [228] and double descent [201]. Grokking [228] refers to the observation that a high degree of generalisation quite suddenly occurs after overfitting already for hundreds of epochs. Double descent [201] refers to the observation that larger parameterised models initially deteriorate in performance before entering another regime where they improve as parameters increase. When structure is empirically involved in the vast differences of models, can we observe certain commonalities within or across application domains?

Approaching structures of neural networks from a data scientific perspective, leads to an analysis of common large neural network models. Naturally, one can ask

◇◇ Research Complex II

Do neural network models differ in terms of structure or are they just different with respect to the choice of other hyperparameters such as activations, training epochs, data sampling or augmentation strategy, optimisation procedure, or the choice of loss function?

The rather young field of structure analysis searches for more general patterns or rules that govern the architecture of neural networks.

◇◇ Research Complex II

Are well studied network theoretic properties involved in the behaviour of neural networks? What are common structures or patterns that can be observed?

The analysis of structures of deep learning models not just requires the data scientific approach of finding patterns but also to understand how the formulation of structure leads to different analytical results. A more fundamental question is therefore concerned with how structural results present themselves and whether they change with different formulations of structure. Convolutional networks are a common example for a structural pattern and can be seen as an examplary answer to how questions on structure could be answered.

◇◇ Research Complex II

How could analytical results look like to guide further architecture development?

### 1.1.6 *Research Complex III: Automating Neural Networks and Neural Structure*

The analysis of deep learning structures is highly complex and in its infacies. Furthermore, a vast amount of other factors such as initial values, optimisation methods, data sampling strategy, learning rate, batch size and number of epochs and many more influence a deep neural networks' performance. An automated and synthetic approach is then usually easier to achieve certain goals than a manual one. Also, in contrast to the analytic aspect of research complex II, the synthetic aspect of automating neural network design is a highly active research field. The scientific purpose of automated design is to systemise, develop and compare methods for finding deep neural networks that optimise performance metrics or comply with other objectives such as low energy or memory consumption, fast inference time, high robust-ness or explainability. Automated neural network design has various names such as NAS or neuroevolution and overlaps with approaches like regularisation and pruning.

> ⋄ ⋄ ⋄ Research Complex III
>
> Which methods for neural architecture search exist, how can they be compared and what are their differences?

The underlying theoretic formalism, tackled in research complex I, sheds light on how to unify methods under a problem formulation with different modifications or assumptions. There exists a large gap on how methods for neural architecture search relate. Methods can relate formally by their assumptions or detailled techniques, or on a comparative level by their computatinal budget, energy cost, performance guarantees and so on.

Some of the techniques are improved upon, e.g. by incorporating knowledge on structural patterns through biases on the search space design or by extending loose evolutionary searches through surrogate models that provide sophisticated sampling of new architectures by learning from previous examples.

> ⋄ ⋄ ⋄ Research Complex III
>
> With knowledge on structure from research complex II, how can neural architecture search methods be improved or guided?

## 1.2 A MOTIVATION FOR DEEP LEARNING

Over the last decade, Deep Learning proved itself to be the most dominant solution for problems in the research field of Artificial Intelligence (AI) for which deterministic, rule- or knowledge-based approaches struggle [79]. AI seeks to tackle problems which are highly repetitive and intuitive for humans but complex for deterministic computer programs or even complex for intelligent humans. Such problems include the automatic categorisation or labelling of images based on their visualised content, the detection of objects or individuals in images and videos (**image processing**), the steering of robotic motor controls in manufacturing, logistics or autonomous driving (**robotics**), the transformation of speech into written language, the generation of synthetic speech (**audio processing**), the translation of written text into other languages (**n**atural **l**anguage **p**rocessing - **NLP**), the dynamic re-routing in complex navigation networks (**intelligent routing**), the search of information in databases (**text processing**) and various more problems within these and similar fields. Practical solutions in Artificial Intelligence nowadays consist of complex pipelines mixing well-studied algorithms from applied mathematics and computer science and probabilistic methods from **machine learning**, often based on data.

Machine Learning encompasses a wide range of methods for using data to learn parameters of models which aim to provide approximate solutions for the underlying problem. Such problems include **classification** in which an exemplary sample is associated with one or many categories, **regression** in which a value within a range is estimated for an exemplary sample, **clustering** in which exemplary samples are sorted into groups based on metric criteria, or **projections** or **generative** tasks. The variety of problems is tackled with an even highery variety of potential methods. Methods include linear regression, naïve bayes, decision trees, logistic regression, $k$-neirest-neighbors, support vector machines, $k$-means, dbscan or **deep neural networks** - and this present work is particularly interested in the latter model and its highly colorful parametric space.

To tackle above machine learning problems, algorithms require an appropriate form of **data** and a notion to measure how a found approximate solution **generalises** on unseen examples or changing environments.

A typical dataset used in our empirical studies is the MNIST[2] digit classification dataset. MNIST poses an image classification problem in which 70,000 images of pixel size 28 by 28 of handwritten digits need to be categorised into one of the ten digits zero, one, two, .., eight and nine. With each pixel having an 8-bit grayscale value between zero and 255, the solution to the posed problem can be seen as a function with signature[3] $\{0, \ldots, 255\}^{28\times 28} \to \{0, \ldots, 9\}$. This constitutes an input space of $1.148505 \cdot 10^{1888}$ possible grayscale images depicting a digit of which we have a tiny subset of 70,000 samples - in fact, this space is so huge that the Eddington number[4] with being roughly $3.15 \cdot 10^{79}$ and providing a well-established guess of the number of protons in the observable universe seems to be quite small in comparison. And still, this particular image classification problem over ten classes of digits can be considered to be solved with state-of-the-art models providing accuracies of 99.x% and high robustness.

*"Any two of the very large dimensionless numbers occuring in Nature are connected by a single mathematical relation, in which the coefficients are of the order of magnitude unity."* [53, *Dirac, Large Number Hypothesis, p. 201*]

*A man said to the universe: "Sir, I exist!" "However," replied the universe, "The fact has not created in me a sense of obligation."* [20, *Barrow citing Stephen Crane on his work on his historical background to the Anthropic Principle*]

A single method often can be used in different problem settings. In this sense, a deep neural network can be easily cast as a *classifier* with stochastic gradient descent optimizing a cross-entropy loss or can be cast a *regressor* with stochastic gradient descent optimizing a mean squared error loss - we clarify the necessary details of these terms and dynamics in part ii on page 27.

2 The **M**odified **NIST** dataset [148] were originally used to "show that convolutional neural networks outperform all other techniques" at that time and act as a baseline benchmark for machine learning techniques in the vision domain up to today.

3 You might notice at this point that we are using computer scientific terms such as *function signatures* where it is convenient instead of referring to domain and co-domain.

4 Wikipedia and John Barrow [20, eq 3.1 p. 393] provide an appropriate entry point into serious guesses on estimates of the number of protons in the observable universe.

The universality of deep neural networks and the rich variety of application domains make them a fascinating object of study.

## 1.3 ON BIOLOGICAL MOTIVATION

The human species is as of today the best proof for the existence of evolutionary systems that are capable of high adaptation to diverse environmental circumstances and capable of philosophical reasoning. A seeming paradox of self-reflection and curiosity, paired with many unsettled questions about the nature of intelligence, constantly inspires researchers and engineers to bring intelligent features into artificially or non-biologically created systems. Examples of this biological source of inspiration include

- **evolutionary principles** inspired by hypothetical forces in biological and other physical systems,
- **connectionist ideas** and **network emergent phenomenons** inspired by the human brain and other network-organised systems,
- and **learning principles** inspired by human and animal developmental observations and models.

Concerning deep neural networks, the human brain is probably the strongest source of inspiration. This explains historically many terms used in deep learning such as artificial neural networks, neurons, or *learning*. Dendrites and axons in the human brain as depicted in fig. 1.2 on the next page are not only an inspiration for the first perceptron theory [240] on how information amongst neurons might be exchanged, but also for the usage of network theoretic approaches to understand emergent properties of different neural networks.

The human brain also inspires researchers to raise information-theoretic questions, which then can also be posed in deep learning in similar ways. For example, Barabasi et al. observe the "presence of reproducible local and global architectural features of" the brain's "connectome" and leaves them "raising a fundamental question: where, and how, is the neuronal connectivity information encoded in an organism?" [16]. Or "how does a brain wire, to varying degree of reproducibility, a network of billions of nodes and trillions of links?" [16]. They "start from the hypothesis that the genetic identity of neurons guides synapse and gap-junction formation and show that such genetically driven wiring predicts the existence of specific biclique motifs in the connectome" [16]. This is in very strong analogy to our analysis of artificial neural network structure in chapter 9 on page 155. They further note that "to explore the mechanisms responsible for wiring reproducibility we must first decide how similar the wiring of two connectomes is. This is a graph isomorphism problem, one of the most challenging computational problems in graph theory" [16]. With this realisation, they identify a key problem

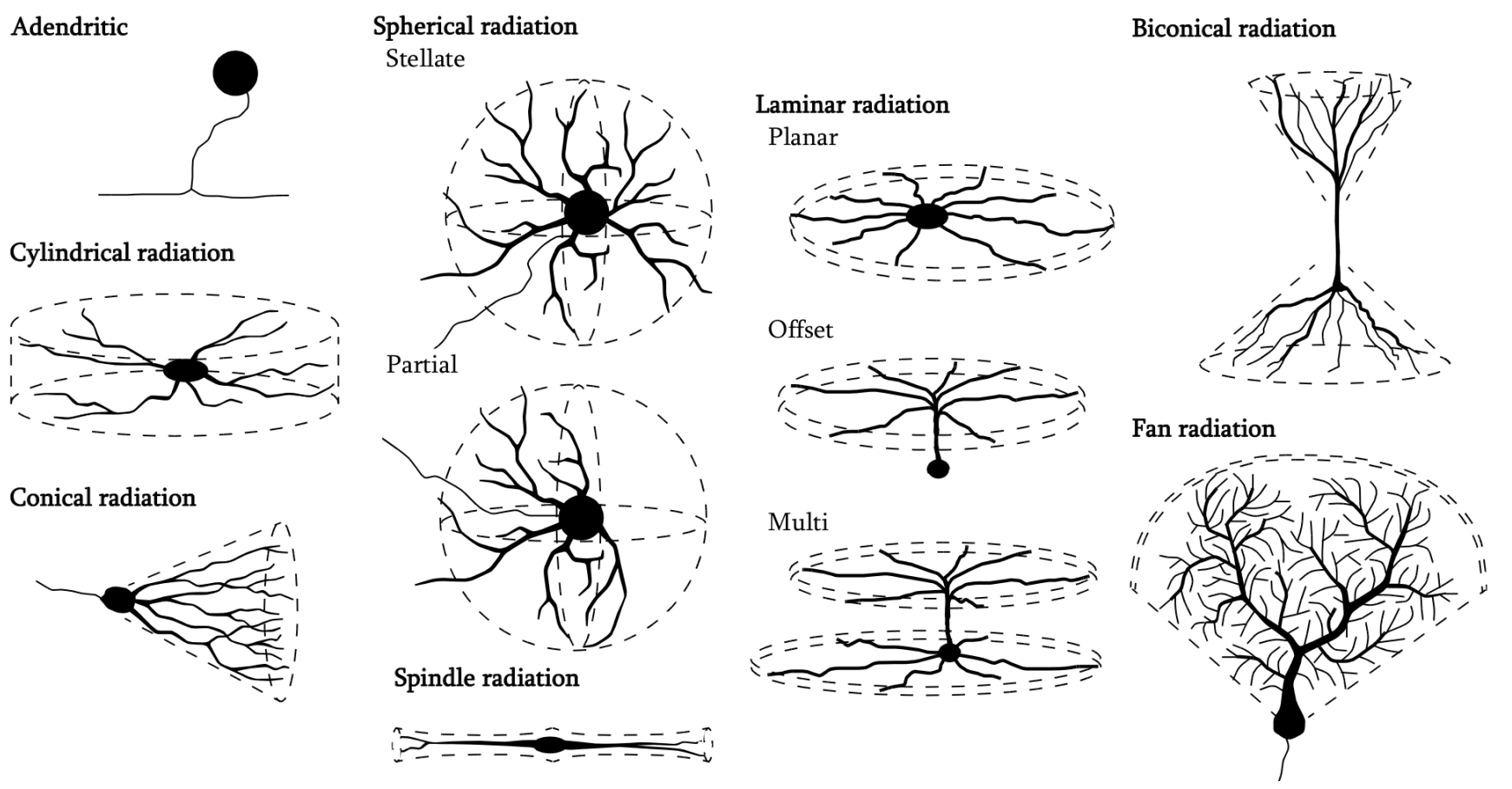

**Figure 1.2:** Dendrite structures of biological neurons taken from Fiala & Harris [67]. They note that "dendrites exhibit enormously diverse forms" and that "understanding the structural diversity [..] will be essential for understanding [..] the contribution dendrites make to mental processes".

– the *graph isomorphism problem* – that is also present in this work when tackling search space design in section 7.2 or generative models for neural structure in chapter 14.

Human brains are still far from understood in neuroscience. According to Winding et al., "to date, complete connectomes have been mapped for only three organisms, each with several hundred brain neurons: the nematode C. elegans, the larva of the sea squirt Ciona intestinalis, and of the marine annelid Platynereis dumerilii" [307].

We find further interesting recent information from neuroscience in [185]: Makarova et al. write, that "animal brains differ by more than 12 orders of magnitude in volume and more than 10 orders of magnitude in the number of neurons" [185]. "The largest known brains can reach up to 10 kg and 257 billion neurons" [185] "Insects occupy an intermediate position: their brains can contain from about 4600 neurons (7400 cells in the whole nervous system), as in the miniature parasitoid wasp Megaphragma amalphitanum, to 1.7 million neurons, as in the spider-hunting wasp Pepsis thisbe" [185]. "Full-body connectome is available for Caenorhabditis elegans [308]" [185]. "But even the mouse brain is too big for an exhaustive exploration on the cellular level: mapping its connectomes with existing methods will take many years" [185]. "It should be noted that the brain volume of the largest insects is not yet known" [185] "The smallest number of neurons in an insect brain, 4 400, has been found in the first instar nymph of the thrips Heliothrips haemorrhoidalis" [185] "The organisation of the brain is almost identical in large insects and in the smallest ones" [185]

These statements tell us about the current progress of mapping out biological neural systems and how their parametric capacities seem to relate. Many areas in the human brain have been cartographed, e.g. in Brodmann areas [237, Fig. 1.9, p.24]. E. T. Rolls provides an extensive in-troduction to *brain connectivity* [237]. Rolls focuses on *what* is computed by the human brain and *where* [237]. Roberts et al. further note that "a key goal of evolutionary neuroscience is to go beyond a description of what has changed in the brains of diverse animals over time to draw general principles and try to understand the mechanisms by which neural circuits and behaviours change" [235]. Biological motivation is therefore not only the origin, but also a plea for further inspiration on how artificial systems could learn and obtain capabilities of high adaption to unseen circumstances.

# 2 OVERVIEW OF THIS WORK

*The following entails:*

## 2.1 OVERVIEW

At the heart of our work, research questions and contributions is our formulation of **graph-induced neural networks**.

Part ii introduces our research goal around the structures of neural networks, for which we organise emerging research questions in three complexes as presented in section 1.1.4 on *formalism*, section 1.1.5 on *structure analysis*, and section 1.1.6 on *automation*.

Chapters 3 to 5 in part ii serve to draw connections between theoretical topics and our formalism of graph-induced neural networks, and also provide explanations of concepts used in the experiments such as evaluation methods. Readers familiar with the common terms in machine learning can skip these chapters and return occasionaly for detailled clarifications.

We then introduce graph-induced neural networks in sections 6.1 to 6.3. The **formalisation** provides a unified language for methods concerned with structure analysis, neural architecture search, and search space design. This unified language includes the definition of structural themes and universal architectures.

Our interest with structure analysis is to find answers to the initially posed questions and to deepen understanding of the proposed formulation. Graph-induced neural networks are further used to formulate optimisation problems on varying levels which align with what is commonly understood as *architecture search* [61] and *search space design* [230]. Neural architecture search is an umbrella term for methods which tackle these optimisation problems; and search space design is highly related to NAS in the sense of taking the optimisation to the next level of abstraction.

We formally compare the formulation of graph-induced neural networks with existing approaches in section 6.4, and while structure is a first-class citizen of our formulation, we will also critique the limita-tions of the formalisation based on theoretical and empirical findings throughout and in the conclusion of our work.

Part iv is then practically concerned with **structure analysis**. Chapter 9 gives an introduction into basics of analysing the structure of neural networks. We present our experiments on correlation analyses between structural properties and correctness, robustness and energy consumption measures. The first two measures correctness & robustness are analysed in experiments with graphs drawn from distributions of classical random graph generators. We find first insights from correlation analysis, but also realise advantages and disadvantages of using random graphs as structural priors in the analysis: Candidates of the search space may be biased and never fully sampled.

The insights lead us to re-think the search space design jointly with the theoretical considerations in section 7.2.3 on structural themes such that we conduct further analyses with computational themes on our created CT-NAS benchmark in chapter 10. While these analyses lead to further insights on different types of data-dependent search space behaviours, clustering large candidate search spaces again turns out to be difficult to analyse. Structure analyses not only can relate properties of single graphs with targeted measures (correctness, robustness, energy, ..) but also can include groupings of structural themes based on similarity measures or treating them more fuzzy as distributions of graphs – which could be learned.

In part v we turn our focus to existing **methods** for **neural architecture search** and look at them under the perspective of graph-induced neural networks. The first method we consider is **pruning** in chapter 11 with which we originally started our studies. Our contribution there is of methodological nature in section 11.3 by transferring Shapley Value, a game-theoretic solution concept, to neural pruning. We conducted a formal analysis of non-additive payoffs in such a game-theoretic setting in section 11.4, motivated by understanding the limitations of the Shapley Value based pruning method. For completeness, we also summarised approximation methods for Shapley Values in section 4.3 in our formal background.

Subsequently, we focus on evolutionary- and **genetic algorithms** for neural architecture search in chapter 12. A core study besides our publications contains results of experiments on variation operators (Var. Ops) in graph space. These operations produce new graphs and are motivated by our graph-induced neural network formulation but also align with genetic algorithms in the case of graphs as an underlying genetic encoding. Our main result is that, still, selection appears to be the strongest force for graph-based genetic architecture searches. But we also observe differences of variation operators and we suppose that successful operators might be domain-dependent – which could again be learned. We close the chapter with also contextualizing the well-established DARTS-type of methods for neural architecture search and observe that differentiable relaxation and finite restriction of the previously coarsely structured open space accelerates searches significantly.

The observations in part v & part v lead us to investigate in two promising directions of **advancing** neural architecture search: predictive and generative extensions.

Predictive models avoid estimating expensive target measures and accelerate neural architecture searches. We build upon the idea of differentiable architecture search in section 13.2 to obtain quickly evaluated scores which we use back in an open search space and find that such scores can also be used as a surrogate for performance estimations. Our approach is called Fast Differentiable Estimation (FaDE) and poses an alternative to other performance predictions.

We then consider generative models in chapter 14 which concentrate on the first-class citizens *graphs* in our formulation of graph-induced neural networks. Learning generative models of graphs based on data such as neural architecture benchmarks turns out to comprise many challenges on its own. With DeepGG we investigate on de-biasing the existing model DGMG in section 14.2 and experiments between the models GraphRNN and GRAN lead us to modify GRAN into a conditional generative model MCGRAN in section 14.4. MCGRAN turns out to be promising for neural architecture search on its own.

Our work in chapter 14 on generative models opens two new strong branches for future research: DeepGG and DGMG can be understood as deep state machines, a concept utilizing deep (recurrent) neural networks for learning transitions of finite state automatons. Deep state machines can and are viewed decoupled from neural architecture search. With results of structure analyses from part iv in mind, we further investigate on graph assembly sequences as to avoid costly learning of generative models and instead using quickly generated graphs from assembly sequences.

## 2.2 CONTRIBUTIONS

An overview of our contributions is summarised in table 2.1 on page 20. We classified our contributions into seven types:

**Formal** This contribution type encompasses definitions, formal alignments, qualitative comparisons, and arguments justifying included decisions.

**Method** A methodological contribution is a detailled description including formal aspects and an embedding into a problem formulation for which the method provides a solution approach.

**Survey** Related work of a research field which is systematically edited and brought into context is a surveying contribution.

Data — Contributions containing collected real-world data, data-generation processes, pre-processing steps, and basic analysis therein.

Emp. Analysis — Empirical analysis involving statistical tests, correlations, clusterings, visual analyses, and associated reasoning.

Form. Analysis — A formal analysis of a properly defined setting, i.e. proofs with induction, inclusion or of bounds.

Implementation — Technical contributions with code or related resources.

Further, we split our contributions into two categories **major** and **minor** as to provide an overview of how large of a space the contribution takes in light of the overall research goal or how well rounded up the work is to be considered.

## 2.3 LIST OF PUBLICATIONS

*Our overarching theme is to gain understanding of structural properties – or structure at all - in deep learning models.*

Below you can find a list of twelve publications associated with this work. In total we have conducted 18 theses around topics of structures of neural networks of which six theses continued into an extended phase of formal improvements, further experiments and went through a peer-reviewed publication cycle.

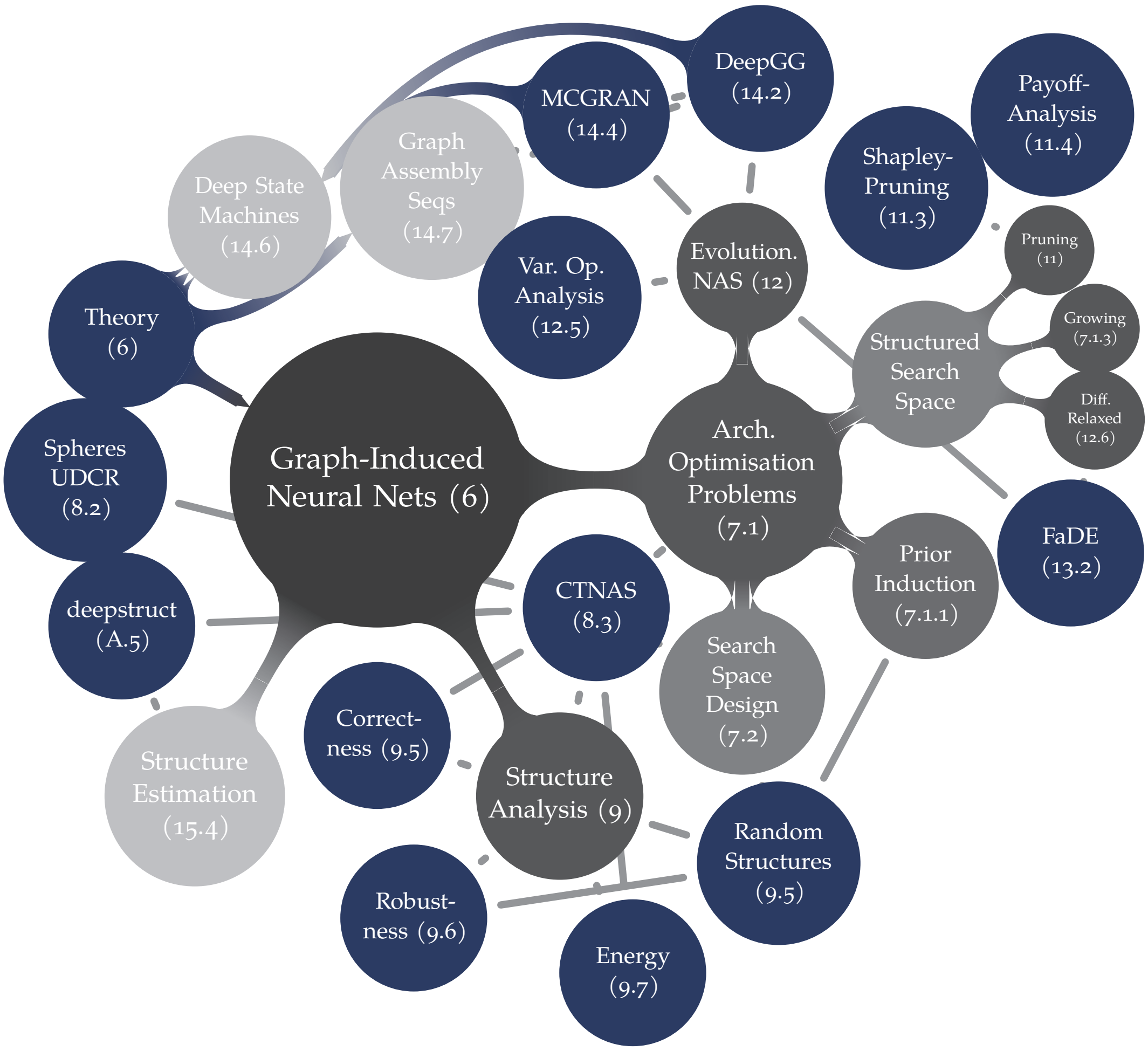

**Figure 2.1:** In addition to table 2.1, this visualisation provides an overview of the topics and contributions covered by our work. The formal background of part ii leads towards the definitions on graph-induced neural networks in sections 6.1 to 6.3. Individual sections are clickable in a digital version of our work.
Gray concepts include major topics and their connection. Dark blue concepts indicate contributions and light blue concepts indicate future work or work that goes beyond the central theme of graph-induced neural networks.

| Description | Sections | Type | Definitions | Arguments | Corr. / Clust. | Vis. Analysis | Hyp. Tests | Form. Analysis | Large-Scale | Literature | Code |
|---|---|---|---|---|---|---|---|---|---|---|---|
| **Major Contributions** | | | | | | | | | | | |
| Graph-Induced Neural Networks | 6.1-6.3 | Formal | ☑ | ☑ | ☐ | ☐ | ☐ | ☐ | ☐ | ☑ | ☐ |
| CT-NAS benchmark | 8.3 | Data | ☐ | ☑ | ☐ | ☑ | ☑ | ☐ | ☑ | ☐ | ☑ |
| Structure Analysis | 9 | Emp. Analysis | ☐ | ☑ | ☑ | ☑ | ☑ | ☐ | ☑ | ☐ | ☐ |
| Shapley Value based pruning | 11.3 | Method | ☑ | ☑ | ☐ | ☐ | ☐ | ☐ | ☑ | ☐ | ☐ |
| Analysis of Var. Ops in Evol. NAS | 12.5 | Emp. Analysis | ☐ | ☑ | ☑ | ☑ | ☑ | ☐ | ☑ | ☑ | ☐ |
| FaDE for NAS as alternative to performance prediction | 13.2 | Method | ☑ | ☑ | ☐ | ☐ | ☐ | ☐ | ☐ | ☐ | ☐ |
| DeepGG & MCGRAN for Generative NAS | 14.2 & 14.4 | Method | ☑ | ☑ | ☐ | ☐ | ☐ | ☐ | ☐ | ☐ | ☑ |
| LIB DeepStruct | A.5 | Implementation | ☐ | ☐ | ☐ | ☐ | ☐ | ☐ | ☐ | ☑ | ☑ |
| **Minor Contributions** | | | | | | | | | | | |
| Shapley-Value Approximations | 4.3 | Survey | ☐ | ☐ | ☐ | ☐ | ☐ | ☐ | ☐ | ☑ | ☑ |
| Universal Approximation Theorems | 5.9 | Survey | ☐ | ☐ | ☐ | ☐ | ☐ | ☐ | ☐ | ☑ | ☐ |
| SpheresUDCR | 8.2 | Method, Data | ☑ | ☑ | ☐ | ☐ | ☐ | ☐ | ☑ | ☐ | ☑ |
| Analysis of Payoffs for Shapley-based Pruning | 11.4 | Form. Analysis | ☐ | ☑ | ☐ | ☐ | ☐ | ☑ | ☐ | ☐ | ☐ |
| Deep State Machines | 14.6 | Method | ☑ | ☑ | ☐ | ☐ | ☐ | ☐ | ☐ | ☐ | ☐ |
| Graph Assembly Representations | 14.7 | Formal | ☑ | ☑ | ☐ | ☐ | ☐ | ☐ | ☐ | ☐ | ☐ |

**Table 2.1:** This table provides a short overview of contributions and the respective fields they have been made in. The three major methodological contributions have been previously published in [6] (SV-pruning), [7] (DeepGG), and [3] & [2] (FaDE). All empirical analyses contributed here are novel and unpublished except for correlation analyses in [8].

*Note*: various implementations accompanying this work can be found online under `https://julianstier.com`, `https://github.com/JulianStier` and `https://github.com/innvariant`.

## 2.4 FREQUENTLY USED NOTATION

We orientate in notation and terminology in graph theory on Diestel [52] and Bang-Jensen & Gutin [14], in statistics on Polyanskiy & Wu [227], Murphy [200] and James et al. [117], and in Evolutionary Computing on Eiben [59].

| Notation | Meaning |
|---|---|
| $\triangleq$ | reads *defined as* |
| $a, b, x$ | Lowercase letters usually refer to scalars or vectors but not to random variables, sets or graphs |
| $\alpha, \beta, \ldots$ | Greek letters usually refer to scalars or vectors that are used as parameters or hyperparameters or are results of measurements |
| $X$ | Capital letters refer to either random variables, matrices, sets or graphs but not to scalars or vectors |

| Notation | Meaning |
|---|---|
| $G$ | Graphs and most often directed acyclic graphs (DAGs) |
| $\mathcal{G}$ | The set of all graphs and often the set of all graphs of a certain type such as connected DAGs. We usually also exclude the empty graph. |
| $Nei_G^{in}$ | The in-neighborhood of a graph $G$ |
| $[a \in A] = \mathbb{I}_A(a)$ | The indicator function $\mathbb{I}$ or the Iverson bracket $[a \in A]$ yield 1 for which the statement $a \in A$ holds true, otherwise 0 |
| $\lvert X\rvert, \lvert G\rvert$ | Overloaded symbol for the cardinality $\lvert X\rvert$ of a set $X$ or the order $\lvert G\rvert$ of a graph $G$ with $\lvert\cdot\rvert \in \mathbb{N}$ (or seldomly $\lvert\cdot\rvert \in \mathbb{N} \cup \{\infty\}$) |
| $D_{name}$ | A dataset, usually consisting of tuples $(x_i, y_i)$ with a data sample representation (or feature vector) $x_i$ and a data sample target (class / value) representation $y_i$; with $i \in \{1, \dots, \lvert D_{name}\rvert\}$ |
| $\mathbb{N}, \mathbb{N}_+, \mathbb{R}$ | The natural numbers $\mathbb{N}$, the natural numbers without zero $\mathbb{N}_+$, the real numbers $\mathbb{R}$ |
| $\sim$ | reads *is distributed as*; we use both $X \sim Dist$ to denote that a random variable $X$ follows the probability $Dist$ in distribution but also $Dist(X = x \mid a, b, c) = ..$ as to express a probability distribution if a parameterisation w.r.t. $x$ (outcomes of the random variable $X$) is necessary |
| $\mathcal{U}$ | The uniform distribution over a discrete range $\mathcal{U}(\{a, b\})$, a finite set $\mathcal{U}(\{\omega_1, \dots, \omega_n\})$ or a continuous interval $\mathcal{U}([a, b])$ with $a, b \in \mathbb{N} \wedge a < b$ and $\omega_i \in \Omega$ |
| $\mathcal{N}(\mu, \sigma)$, $\mathcal{N}(\mu, \Sigma)$ | The normal (gaussian) distribution with mean (vector) $\mu$ and (co-)variance $\sigma$ or $\Sigma$ (matrix) |
| $\mathbb{E}(X), Var(X)$ | The expectation $\mathbb{E}(X)$ [281, Def. B.4.11, p. 516] and variance $Var(X)$ [281, Sec. 2.1.1] (or second central moment) of a random variable $X$ |
| $\mathcal{K}, C, L$ | A compact domain $\mathcal{K}$, the space of continuous functions $C$ (with $C(\mathcal{K}, \mathbb{R})$ from $\mathcal{K}$ to the reals $\mathbb{R}$), the Lebesgue $L^p$-spaces of finite $p$-norm |
| i.i.d. | identically and independently distributed according to some probability distribution |
| $\mathcal{O}$ | Bachmann-Landau Big-O notation: for functions $f, g$ on $\mathbb{N}$, $\mathcal{O}(g(n))$ means that $f(n)/g(n)$ is bounded for $n \to \infty$ |
| $a \propto b$ | proportionality, i.e. $a$ is proportional to $b$ if there is a constant $c$ such that $a = c \cdot b$ |
| $\binom{n}{k} = C_k^n$ | The binomial coefficient, given as $\binom{n}{k} \triangleq \frac{n!}{k!(n-k)!}$ where $n!$ is the factorial of $n \in \mathbb{N}$ |

| Notation | Meaning |
|---|---|
| $I_n$ | An identity matrix of size $n \in \mathbb{N}$ is a square matrix $I_n \in \{0,1\}^{n \times n}$ where $I_n = (i_{uv})$ with $i_{uv} = \begin{cases} 1 & u = v \\ 0 & \text{otherwise} \end{cases}$ |
| $\sigma^{\diagup}, \sigma^{\sim}, \sigma^{\Psi}$ | rectified linear unit, sigmoid and softmax activation functions |
| $\mathcal{A}^{d_1 \to d_2}(G)$ | A neural network architecture, constructed through a transformation $\mathcal{A}$ of a graph $G$ into a set of functions |
| $\mathcal{A}^{d_1 \to d_2}(\mathcal{G})$ | A universal architecture, if $\mathcal{A}^{d_1 \to d_2}(\mathcal{G}) = \bigcup_{G \in \mathcal{G}} \mathcal{A}(G)^{d_1 \to d_2}$ has the universal approximation property |
| $\alpha(f, D_{valid})$ | The accuracy of a model $f$ evaluated on a dataset $D_{valid}$ |

Part II

# FORMAL BACKGROUND

Given the motivation and questions that arise around the structure of neural networks, the following chapters aim to provide a formal foundation for the concepts: What is a neural network and what is its structure?

# 3

# GRAPH THEORY

*The following entails:*

With the structure of neural networks being the central object of our investigations, we will almost always work with graphs as representations of this structure. From a theoretical perspective, one can look at the connection between neural networks and their structure as the study of a discrete subspace of the topological function space neural networks rely in. But we will also draw inspiration from ideas and methods of network science, a field of graph theory which deals with graphs of non-trivial properties that are emerging with large numbers of vertices.

## 3.1 GRAPHS

The graph-theoretic formalisations follow Diestel [52] in general and Bang-Jensen & Gutin [14] for further details on directed graphs (digraphs). A graph $G$ is a tuple $(V, E)$ consisting of a set of vertices $V$ and a set of edges $E \subseteq V \times V$. We overload the symbol $V$ as to obtain the set $V(G_1)$ given a specific graph $G_1$ and do so as well with $E(G_1)$ for the edge set. Equivalently to $E$, an irreflexive binary relation over $V$ given by $Adj : V \to \mathcal{P}(V)$ can specify whether a vertex is adjacent to any other vertex in the graph: for a vertex $v \in V$ the map $Adj(v)$ gives the set of all vertices adjacent to it. It holds that $(s, t) \in E \Leftrightarrow t \in Adj(s)$. For two graphs $G_1$, $G_2$ we also write $V_{G_1}$, $E_{G_1}$ and $Adj_{G_1}$ for the first and $V_{G_2}$, $E_{G_2}$ and $Adj_{G_2}$ for the second graph.

The number of vertices of $G$ is its order denoted as $|G|$ with $|G| = |V|$ and its number of edges is denoted as $||G||$, called its size with $||G|| = |E|$. Two graphs $G$ and $G'$ are isomorphic if there exists an isomorphism $\phi : V(G) \to V(G')$ that preserves edges, is bijective, and its inverse also has the homomorphic property of preserving edges [52, Chp. 1.1]. We write $G \simeq G'$ to explicitly emphasise on the isomorphism but usually also write $G = G'$ for simplicity. Graph invariants are maps assigning equal values to isomorphic graphs. We are later especially interested in graph invariants of classes of graphs representing neural network architectures.

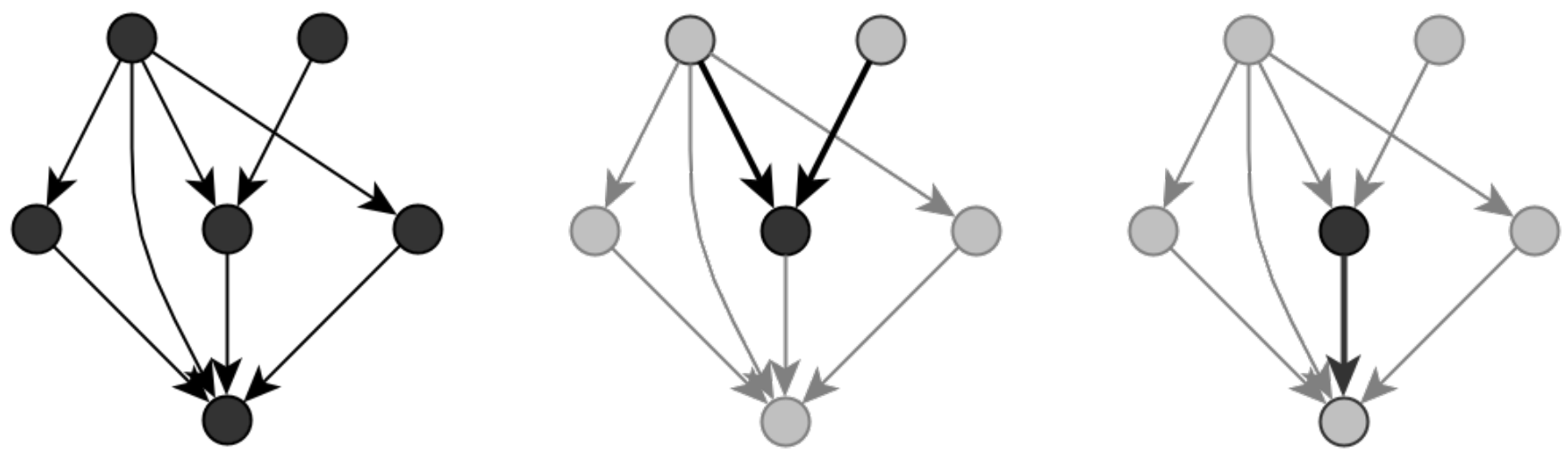

**Figure 3.1:** Exemplary directed acyclic graph $G$ with $|G| = 6$ and $||G|| = 7$. We can draw it easily in a layered manner from vertices with no in-degree at the top to vertices with no out-degree at the bottom. Degree is a property of a vertex and can be collected into a set on a per-graph basis.

For digraphs the edge set $E$ consists of ordered pairs $(s,t) \in E$ for which we call $s$ the source and $t$ the target. We write $st \in E(G)$ for edges in general and the context should make it clear if $st$ is a directed edge $(s,t)$ or an undirected edge $\{s,t\}$ in the edge set. A path is a non-empty graph $P = (V,E)$ for which all vertices $V = \{v_0, v_1, \dots, v_k\}$ are distinct and the edges follow the pattern $E = \{v_0v_1, v_1v_2, \dots, v_{k-1}v_k\}$. The length of the path is the size of $E$ and for $k \geq 3$ and $v_1 = v_k$ $P$ is a cycle. The minimum length $k$ of a cycle in a graph $G$ is the girth $girth(G)$ and if no cycle exists, the property is set to $girth(G) := \infty$. A digraph $G$ is an acyclic digraph if it contains no cycles.

The degree is a notion of connectivity of a local neighborhood of a vertex or sub-graph [14, Chp. 1.2]. For a vertex $v$ of $G$ we call

$$Nei_G^{in}(v) = \{s \in V - v : sv \in E(G)\}, Nei_G^{out}(v) = \{t \in V - v : vt \in E(G)\}$$

the in-neighborhood, out-neighborhood and $Nei_G(v) = Nei_G^{in}(v) \cup Nei_G^{out}(v)$ the neighborhood of $v$. The degree $deg_G(v) = deg(v)$ of a vertex $v$ is the sum of sizes of in- and out-neighborhood and in our definition[1] therefore the number of edges at v: $deg_G(v) = |Nei_G^{in}(v)| + |Nei_G^{out}(v)| = |\{(s,t) \mid (s,t) \in V \wedge (s = v \vee t = v)\}|$. On the right of Figure 3.1, a single vertex with an in-degree of two, out-degree of one and $deg_G(v) = 3$ is shown. With $\bar{deg}(G) := \frac{1}{|V(G)|} \sum_{v \in V(G)} d_G(v)$ the average degree of $G$ is a first example for a graph invariant providing a *global* quantity measuring local neighborhood connectivity on average.

The distance $d(s,t)$ within a graph $G$ between two vertices $s$ and $t$ is the length of a shortest path $P$ with $v_0 = s$ and $v_k = t$. A shortest path between vertices $s$ and $t$ is also called graph geodesic and there may be none or more than one. If no such path exists the distance is set to $d(s,t) := \infty$. The diameter $diam(G)$ is the greatest distance between any two vertices in $G$. Alternatively, the distance can also be defined for two sets [14, eq. 2.2] with $dist(X,Y) = max\{dist(x,y) : x \in X, y \in Y\}$ such thtat the diameter is then given as $diam(G) = dist(V(G), V(G))$.

1 We are only concerned with undirected graphs and directed acyclic graphs without multi-edge formulations or self-loops.

Finding shortest paths, diameter or other distance-based invariants in graphs are search problems for which algorithms with mostly polynomial runtime exist. Such problems are usually formulated as e.g. $P^{s \to t}_{min} = \underset{P \text{ is } s-t-path}{\arg\min} |P|$ and $diam(G) = \max_{s,t \in V(G)} \min_{P \text{ is } s-t-path} |P|$. The expression *length of maximum geodesic* for the graph diameter includes this search problem in its name. For directed acyclic graphs, shortest paths "from a fixed vertex s to all other vertices can be found in time $O(n+m)$" [14, Thm. 2.3.4] with order $n$ and size $m$ of $G$. This is insofar interesting as we will later consider graph invariants such as the average degree, average shortest paths or edge density aggregated over whole sets of graphs, either because they have a network scientific background as a random graph family or as we are interested in graph families with not sharply described or emergent graph properties.

For computing averages of shortest path lengths, we also use $d^0(s,t) = d(s,t)$ if $d(s,t) < \infty$ else 0 but otherwise keep the classical definition following both Diestel and Bang-Jensen & Gutin. The average shortest path length, also known as characteristic path length, is the average over all path lengths of the graph $G$, denoted by $\bar{d}(G) := \frac{1}{n(n-1)} \sum_{s,t \in V(G), s \neq t} d^0(s,t)$.

Diestel [52, Chp. 7] defines density as the proportion of actual edges of $G$ and its potential edges: $||G||/\binom{|G|}{2} = ||G||/(|G| \cdot (|G|-1)) = \frac{m}{n(n-1)}$ while Bang-Jensen & Gutin mention it only as the "ratio of its size and order, i.e. $m/n$" [14]. We follow the implementation of `NetworkX` and use $dens(G) = \frac{2m}{n(n-1)}$ for undirected graphs $G$ and $dens(D) = \frac{m}{n(n-1)}$ for digraphs $D$. A notion of sparsity is then given as $spars(G) = 1 - dens(G)$ for both $dens(G), spars(G) \in [0,1]$.

Let $\pi$ be a vertex permutation function over $V(G)$, i.e. with $(\pi(v_1), \dots, \pi(v_n))$ being a permutation of $(v_1, \dots, v_n)$ and $\forall i \in \{1, \dots, n\} : v_i \in V(G)$, $n = |G|$. We call $\mathbf{A}^\pi = (a_{ij}) \in \{0,1\}^{n \times n}$ an adjacency matrix representation of $G$ under permutation $\pi$ with

$$a_{ij} \triangleq \begin{cases} 1 & \text{if } v_i v_j \in E(G) \\ 0 & \text{otherwise} \end{cases}$$

Like You et al. [328], we'd like to note, that the set of all possible vertex permutations $\Pi$ contains $n!$ possible permutation functions such that the set of all adjacency matrix representations $\mathbf{A}(G) = \{\mathbf{A}^\pi \mid \pi \in \Pi\}$ grows super-linearly in $|G|$: $|\mathbf{A}(G)| \propto |G|!$. This asymptotic growth illustrates an aspect of the graph isomorphishm problem. We overload $\mathbf{A}(G)$ to refer to an adjacency matrix with identity permutation.

Another common representations of a graph $G = (V, E)$ is its Laplacian. The Laplacian uses information of the degree matrix of a graph.

The degree matrix for a graph of order $n = |V|$ is a square matrix $D^\pi \in \mathbb{N}^{n \times n} = (d_{ij})$ with

$$d_{ij} \triangleq \begin{cases} deg_G(v_i) & \text{if } i = j \\ 0 & \text{otherwise} \end{cases}$$

and the Laplacian is then a square matrix $L^\pi \in \mathbb{N}^{n \times n} = (l_{ij})$ with

$$l_{ij} \triangleq \begin{cases} deg_G(v_i) & \text{if } i = j \\ -1 & \text{if } i \neq j \wedge (v_i v_j \in E(G) \vee v_j v_i \in E(G)) \\ 0 & \text{otherwise} \end{cases}$$

which can also be expressed as $L^\pi = D^\pi - \mathbf{A}^\pi$.

A graph $G$ is (vertex-)labelled if it additionally carries a label assignment $lab_G : V(G) \rightarrow L$ for each vertex in which $L$ denotes the set of labels. Usually $L$ is finite and in contexts of neural architecture search the labels i.e. denote operation types which are considered when the graph is expanded into a deep neural network. Technically, the graph carries an additional label vector if each vertex is associated with one and only one label or an additional label matrix if each vertex can be associated with multiple labels (i.e. using a one-hot encoding).

## 3.2 NETWORK SCIENCE

Network Science is concerned with complex networks such as cellular, biological neural, social and communication networks, power grids and trade networks [17, Chp. 1.2], and studies their characteristic or emergent properties. Barabási distinguishes the field from graph theory by its empirical nature and we take inspiration from this to transfer some of its concepts and methods to the study on deep neural networks, although the latter can be seen as both mathematical structures and as physically implemented natural objects.

Along with Xie et al. [319] we have been the first [272] to integrate large random graphs as structures into deep neural networks. Properties of random graph models have been observed in neurobiological sciences for biological neural networks and such studies have always been an inspiration to foster a connection between these fields. In the following sections, however, we will see that moving from undirected graphs to directed acyclic graphs makes significant differences when it comes to formulating a model precisely or search through a hypothesis space of graphs.

A suggested mental model for the following is that we have a set of graphs $\mathcal{G}$ which might not be given explicitly and we want to conduct sophisticated draws $A$ from a probability distribution $Pr(\mathcal{G})$ over the graphs: $A \sim Pr(\mathcal{G})$. Classical network scientific models provide us with one such generative method but there has also been progress in learning models from observed graph data $D_{obs}$: $A \sim Pr(\mathcal{G} \mid D_{obs})$. With this tool we then can analyse connections between a particularly restricted hypothesis space of graphs, from which we sample candidates, and deep neural networks built from such candidates. In a similar fashion, it can be leveraged to drive automated machine learning pipelines through neural architecture search.

Random graph models emerged in the 1960s [62, 76] and have since been studied not only by graph theorists but also network scientists. Graph theorists from the mathematical side have been interested in random graph models of infinite order, studying limiting cases of $n \to \infty$ while network scientists have been relating them in a more finite sense to real-world graphs with e.g. $n \to 500$ or $n \to 1,000,000$. The influential work of Erdős did not just bring up another notion for graphs but also brought up the *probabilistic method* as a mathematical proof technique [52, Chp. 11], making it possible to proof theorems such as the existance of graphs with both girth and chromatic number being larger than $k$ for every integer $k$ [52, Thm. 11.2.2].

We look at two constructive examples of probability measures to see that they can be properly defined and exist.

The first one is taken from Diestel [52, Chp. 11.1] and introduces the Gilbert-Erdős-Rényi model $GIL(n, p)$. With a fixed vertex set $V$ of $n$ vertices, the set of graphs $\mathcal{G}$ over vertices $V$ is turned into a probability space. For every potential edge $e \in V \times V$ an "own little probability space $\Omega_e \triangleq \{0_e, 1_e\}$" is defined with probability measure $Pr_e(\{1_e\}) \triangleq p$ and $Pr_e(\{0_e\}) \triangleq 1 - p$ based on its elementary events. The product space $\prod_{e \in V \times V} \Omega_e$ is then taken as probability space for $GIL(n, p)$. An element $\omega \in \Omega$ obtained from the product measure $GIL(n, p)$ of all measures $Pr_e$ defines the presence or absence of each $e \in V \times V$ such that we can "identify $\omega$ with the graph $G$ on $V$" whose edge set is $E(G) = \{e \mid \omega(e) = 1_e\}$. Sets of graphs on $V$ can then be seen as events on $GIL(n, p)$. Notice, that this formulation only considers a probability over fixed sets of vertices such that all graphs are of order $n$.

The second example shows a probability over varying sizes of $n$. Pinelis [340] gives an example of a construction of a probability over graphs: He defines $S_n$ to be the set of all graphs over $\{1, \dots, n\}$ with $n \in \mathbb{N}$. The infinite union $S := \bigcup_{n=1}^{\infty} S_n$ then contains all graphs and he argues $S$ to be countable because each $S_n$ is finite. With $\mathcal{B}_n = \mathcal{P}(S_n)$ being the powerset of $S_n$ and $\mathcal{B} = \mathcal{P}(S)$, $Pr_n$ is the probability measure on $\mathcal{B}_n$. An example for $Pr_n$ would be the uniform distribution over $S_n$. Pinelis further defines a sequence $(p_n)_{n=1}^{\infty}$ of non-negative real numbers such that $\sum_{n=1}^{\infty} p_n = 1$. With these weights the probability measure $P$ on $\mathcal{B}$ is given as $Pr(A) := \sum_{n=1}^{\infty} p_n Pr_n(A \cap S_n)$ for $A \in \mathcal{B}$.

### 3.2.1 *Random Graph Models*

Three models, the Erdős-Rényi model $GIL(n,p)$ [76] or $ER(n,m)$ [62], the Watts-Strogatz model, and the Barabasi-Albert model, are examples of such probability distributions over graphs. Their formal definition allows for analytically well-defined properties such as average path lengths. With such emergent properties for a growing graph order, the three models are truly distinct random graph models, i.e. provably different in nature.

ERDŐS-RÉNYI MODEL Without further construction details, we now can state the two famous random graph models $GIL(n,p)$ [76] and $ER(n,m)$ [62]. The first model, $GIL(n,p)$, defines a probability measure $Pr(\{G\}) = p^k(1-p)^{\binom{n}{2}-k}\mathbb{1}_{|G|=n}$ with parameters $n \in \mathbb{N}$ and $p \in [0,1]$ and for which $k = ||G||$ with $0 \leq k \leq \binom{n}{2}$. The second model, $ER(n,m)$, uses parameters $n \in \mathbb{N}$ and $0 \leq m \leq \binom{n}{2}$ with $m \in \mathbb{N}$ and defines a probability measure $Pr(\{G\}) = 1/\binom{\binom{n}{2}}{m}\mathbb{1}_{||G||=m}$.

The $GIL(n,p)$-model defines a probability space over unlabelled graphs of order $n$ by choosing each edge independently with probability $p$. But the $ER(n,m)$-model defines a probability space over graphs of order $n$ *and* size $m$ such that each of these graphs are uniformly distributed. The models are related in the sense that the $ER(n,m)$ graph is a $GIL(n,p)$ graph if the number of edges of $GIL(n,p)$ is k. For a graph $G_0$ with $||G_0|| = k$

$$\begin{aligned}
&Pr(\{\, GIL(n,p) = G_0 \mid ||GIL(n,p)|| = k\}) \\
&\quad = \frac{Pr(\{||GIL(n,p)|| = k \mid GIL(n,p) = G_0\})Pr(||GIL(n,p)|| = k)}{Pr(GIL(n,p) = G_0)} \\
&\quad = \frac{1 \cdot (p^k(1-p)^{\binom{n}{2}-k})}{\binom{\binom{n}{2}}{k}p^k(1-p)^{\binom{n}{2}-k}} = \binom{\binom{n}{2}}{k}^{-1}
\end{aligned}$$

for which we used the fact that there are $\binom{\binom{n}{2}}{k}$ graphs of order $n$ and size $k$. This relationship between $GIL(n,p)$ and $ER(n,m)$ can be observed in Table 3.1. Notice, that with unlabelled graphs the examples in Table 3.1 contains isomorphisms. Graphs $G_2$, $G_3$ and $G_4$ with each having one edge are isomorphic as well as graphs $G_5$, $G_6$ and $G_7$ with each two edges. The advantage of modeling $GIL(n,p)$ in that way from a probabilistic point of view is that the individual edges are independent of each other when seen as an event.

The $GIL(n,p)$ model with a fixed set of vertices over $n$ controls various properties with $p \in (0,1)$. One such obvious property is that the expected density (or sparsity) increases with an increasing probability parameter $p$. This can be observed in Figure 3.2.

| $n = 3$ | | $p = 0.5$ | $p = 0.2$ | $m = 1$ | $m = 2$ |
|---|---|---|---|---|---|
| 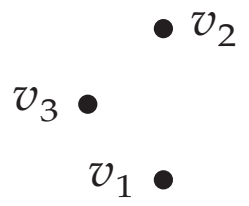 | $G_1$ | $\frac{1}{8}$ | $\frac{64}{125} = 0.512$ | 0 | 0 |
| 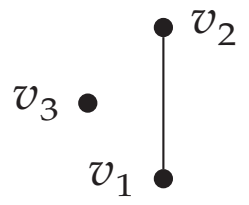 | $G_2$ | $\frac{1}{8}$ | $\frac{16}{125} = 0.128$ | $\frac{1}{3}$ | 0 |
| 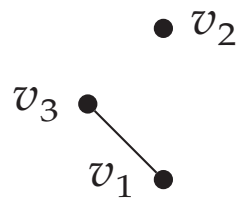 | $G_3$ | $\frac{1}{8}$ | $\frac{16}{125} = 0.128$ | $\frac{1}{3}$ | 0 |
| 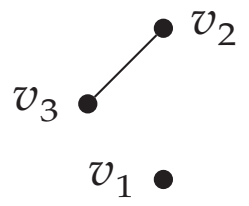 | $G_4$ | $\frac{1}{8}$ | $\frac{16}{125} = 0.128$ | $\frac{1}{3}$ | 0 |
| 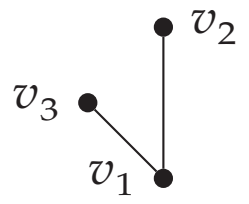 | $G_5$ | $\frac{1}{8}$ | $\frac{4}{125} = 0.032$ | 0 | $\frac{1}{3}$ |
| 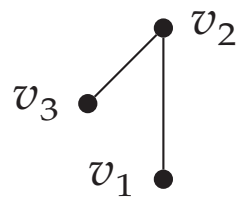 | $G_6$ | $\frac{1}{8}$ | $\frac{4}{125} = 0.032$ | 0 | $\frac{1}{3}$ |
| 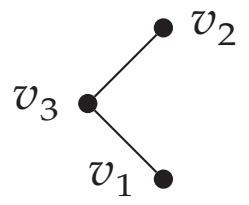 | $G_7$ | $\frac{1}{8}$ | $\frac{4}{125} = 0.032$ | 0 | $\frac{1}{3}$ |
| 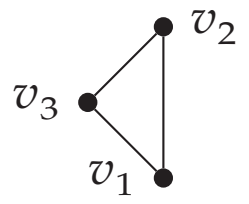 | $G_8$ | $\frac{1}{8}$ | $\frac{1}{125} = 0.008$ | 0 | 0 |

**Table 3.1:** Probabilities for graphs occuring in the random graph models $GIL(n, p)$ and $ER(n, m)$. The columns are abbreviated and refer to the probabilities that a certain graph occurs in a model parameterised with $Pr(GIL(n = 3, 0.5) = G_i)$ for $p = 0.5$, $Pr(GIL(n = 3, p = 0.2) = G_i)$ for $p = 0.2$, $Pr(ER(n = 3, m = 1) = G_i)$ for $m = 1$ and $Pr(ER(n = 3, m = 2) = G_i)$ for $m = 2$.

WATTS-STROGATZ MODEL The Watts-Strogatz model, originally proposed by Duncan Watts and Steven Strogatz [298], generates small-world graphs with distinct characteristics. Small-world graphs are sparse graphs in which a vertex can reach any other vertex within relatively few steps. A graph $G$ is sampled from $WS(n; k; p)$ in a two-step rewiring process with graph order $n$, mean degree $k$ and rewiring probability $p$; for which $n, k \in \mathbb{N}$, $k$ even, $n \gg k \gg ln(n) \gg 1$, and $p \in [0, 1]$: First, a graph $G$ is initialised as a regular graph of $n$ vertices with fixed degree $k$. That can simply be a graph with $n$ vertices in a ring on a 2d-plane with each vertex connected to $K/2$ neighbors to both sides. In a second step each edge $uv \in E(G)$ is rewired by sampling $v'$ uniformly from $V(G)\ \{u, v\}$ and replacing $uv$ with $uv'$ in $E(G)$.

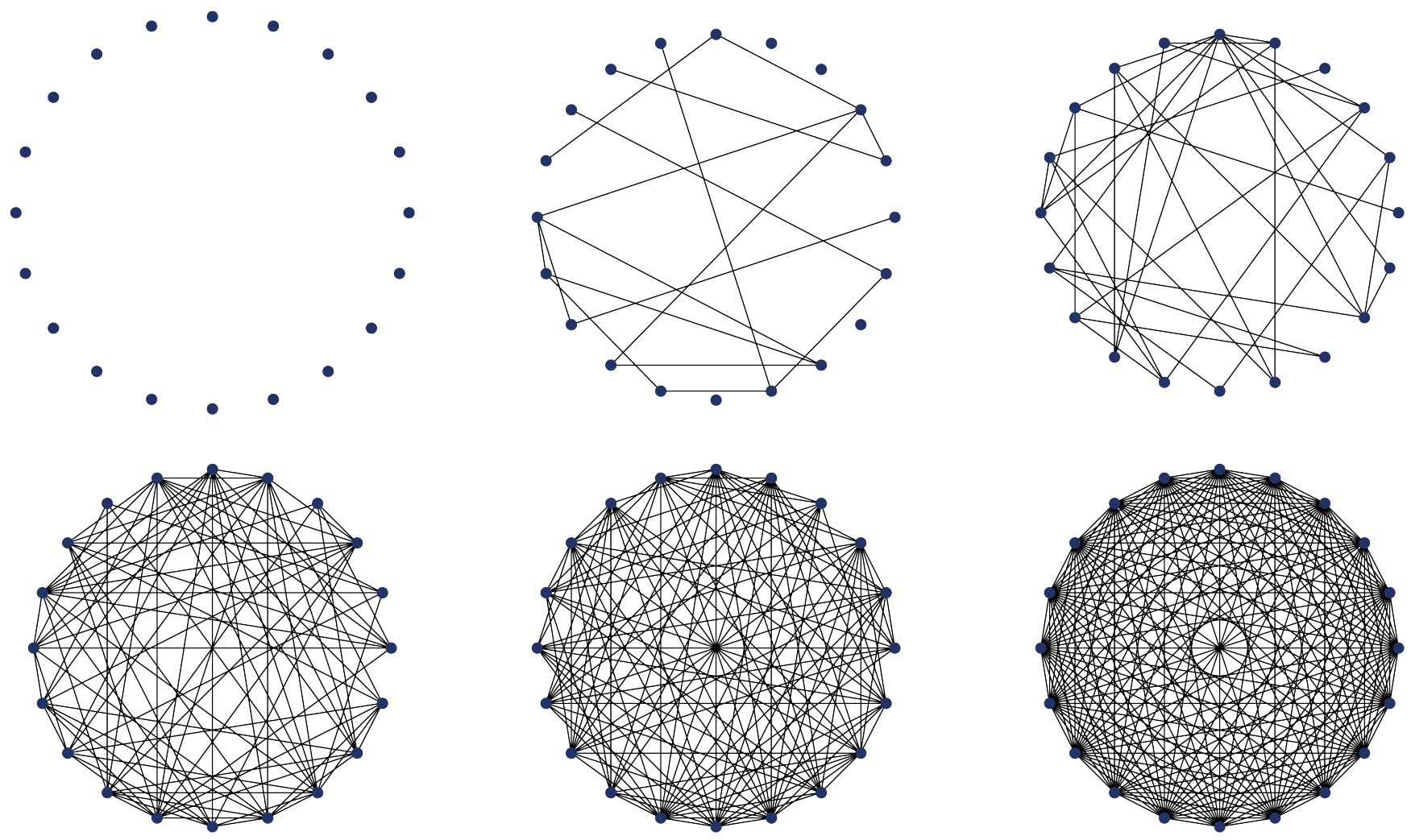

**Figure 3.2:** Exemplary graphs sampled from $GIL(20, 0.01)$, $GIL(20, 0.1)$, $GIL(20, 0.2)$ (top left to right), $GIL(20, 0.5)$, $GIL(20, 0.7)$, and $GIL(20, 0.99)$ (bottom left to right). Notice how the model is parameterised from low to high probability of sampling edges.

Examples sampled from the Watts-Strogatz model can be examined in Figure 3.3. With increasing rewiring probability $p$ from left to right, one can observe the increasing randomness in the connectivity of the graph. For $p = 1$ the Watts-Strogatz model is getting close towards a $GIL(n, p = \frac{k}{n-1})$ random graph model. The mean degree $k$ influences the average degree and the overall degree distribution is similar to the one of a random graph centered around $k$. Characteristic for such small-world graphs is the tendency of its clustering coefficient towards $\frac{3}{4}$ for increasing $k$ and an average path length of $\frac{3}{4}$.

BARABASI-ALBERT MODEL Barabasi & Albert introduced the original scale-free network model in [4, 15]. They observed, that a "common property of many large networks is that the vertex connectivities follow a scale-free power-law distribution" [15] and subsequently proposed a model based on "two ingredients", namely *growth* and *preferential attachment*.

The model starts with an initial number of $m_0 \in \mathbb{N}$ vertices which are randomly connected and for which it must hold that each vertex has at least one neighbor such that there is no vertex with degree zero. Over a course of $t \in \mathbb{N}$ timesteps, the graph undergoes a *growth* and *preferential attachment* phase per step. For growth, a vertex is added and connected to $m \in \mathbb{N}$ already existing vertices, $m \leq m_0$. The probability with which the vertex $v_t$ is connected to existing vertices is defined as $Pr(\{1_{sv_t}\}) \triangleq \frac{deg(s)}{\sum_{v \in V(G)} deg(v)}$. Note, that the evolution-like growth of the model makes it difficult to define a simple probability distribution as in the random graph model. After all timesteps, the Barabasi-Albert model has generated a graph with $|G| = t + m_0$ vertices and $||G|| = m_0 + mt$ edges. The Barabasi-Albert model is then denoted as $BA(n,m)$. We use the implementation of `networkx` which starts off with an initial star graph on (m+1) vertices.

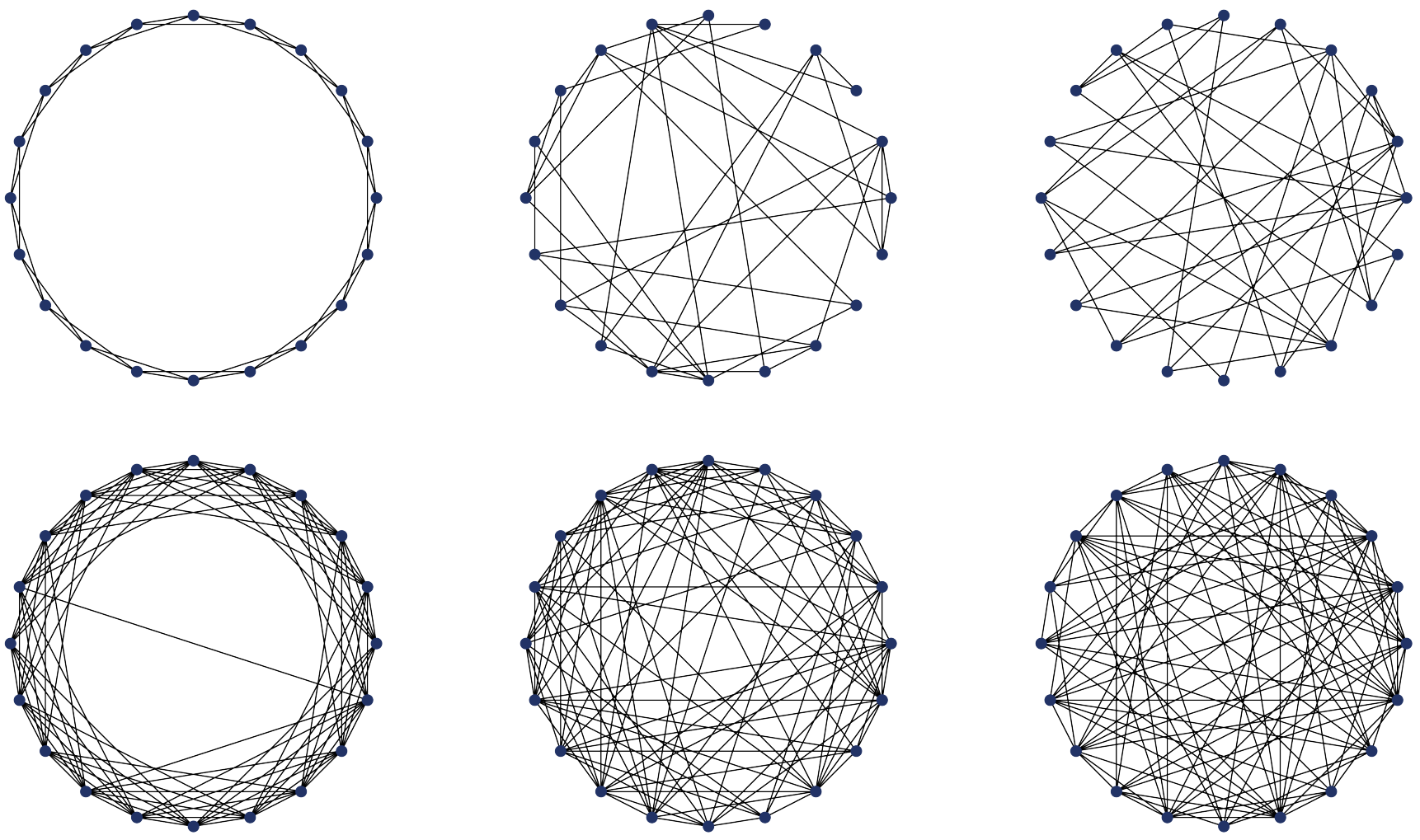

**Figure 3.3:** Exemplary graphs sampled from $WS(20;4;0.01)$, $WS(20;4;0.5)$, $WS(20;4;0.99)$ (top left to right), $WS(20;10;0.01)$, $WS(20;10;0.5)$, and $WS(20;10;0.99)$ (bottom left to right). Notice how the rewiring probability $p$ increases from left to right and turns the underlying regular graph into a more randomly one.

Figure 3.4 depicts exemplary obtained Barabasi-Albert model graphs. Note, how the connectivity first increases with higher proportion of $n/m$ and then gears towards few central vertices. The degree histograms of such graphs show a clear scale-free nature in which few vertices have high connectivity and most vertices have very low degree.

### 3.2.2 *Network Scientific Properties*

Network Science studies further properties beyond classical graph theoretic properties like neighborhood, distances or density. For example, node centralities are used to assign values to vertices to identify structurally important ones. Clustering coefficientss measure the connectedness of a vertices' neighborhood. These properties are often emergent and characteristic for some of the presented random graph models.

CENTRALITIES The most obvious concept of node centrality is the one of degree centrality. Degree centrality assigns each vertex of a graph its degree as a centrality value, expressing its number of connections in the network. The degree centrality of a vertex is given as $C_{deg}(v) \triangleq deg(v)$ and can be normalised with $\tilde{C}_{deg}(v) \triangleq \frac{deg(v)}{|G|-1}$. Figure 3.5 shows a graph of order 15 and a depiction of its degree centrality. Vertices with larger

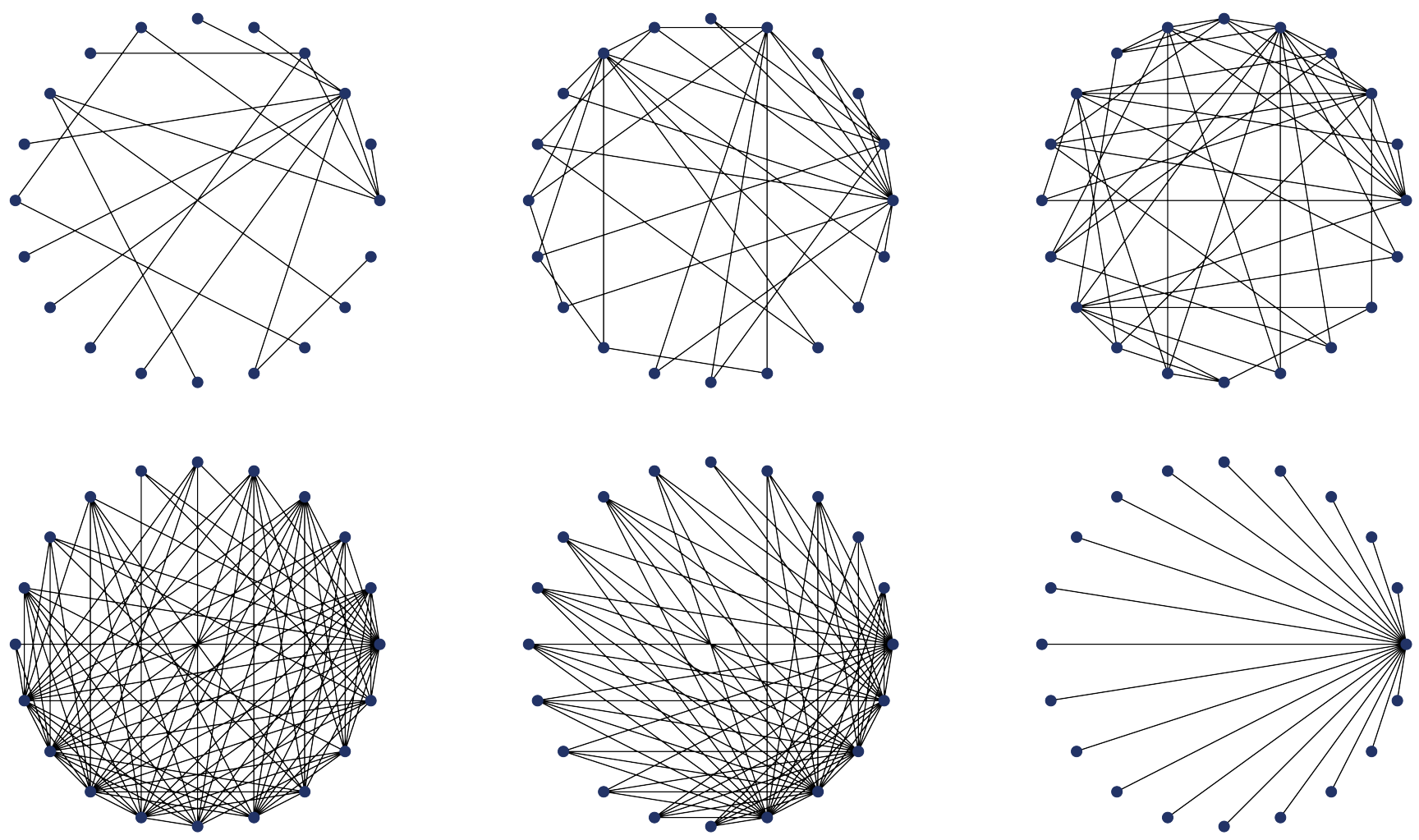

**Figure 3.4:** Depicted are six exemplary Barabasi-Albert graphs sampled from $BA(20, 1)$, $BA(20, 2)$, $BA(20, 3)$ (top left to right), and $BA(20, 10)$, $BA(20, 15)$ and $BA(20, 19)$ (bottom left to right).

degree centrality are depicted with larger dots and are thus considered as more important. Barabasi-Albert model produces graphs with few vertices of very high degree centrality and many vertices of low degree centrality.

Closeness centrality is defined as $C_{clo}(v) \triangleq \frac{1}{\sum_{s,t\in V(G),s\neq t} d(s,t)}$ with $\tilde{C}_{clo}(v) \triangleq \frac{|G|-1}{\sum_{s,t\in V(G),s\neq t} d(s,t)}$ being its normalised version. There are also furhter node centrality measures such as betweenness centrality $C_{bet}(v)$ [72], eigenvector centrality $C_{eig}(v) \triangleq \frac{1}{\lambda} \sum\limits_{v\in V(G)} \mathbf{A}[u,v]e_v$ [86, Sec. 2.1, p.11], Katz centrality, or even Shapley Value centrality [194].

CLUSTERING COEFFICIENTS Another type of network properties is the one of clusterings which expresses how *tightly clustered* the direct neighborhood of a vertex is [86].

The local clustering coefficient $C_{local,v}(G)$ of a vertex $v \in V(G)$ is defined as

$$C_{clust,v}(G) \triangleq \frac{|\{(st \in E(G) \mid s,t \in Nei_G(v)\}|}{\binom{deg_G(v)}{2}}$$

such that it puts the number of actual edges of neighbors of the vertex $v$ into proportion with the number of all possible connected edge pairs. The simplest way of aggregating the local clustering coefficient into a global one for the whole graph is to average across clustering coefficients of all vertices:

$$C_{clust}(G) \triangleq \frac{1}{|V|} \sum_{v\in V(G)} C_{clust,v}(G)$$

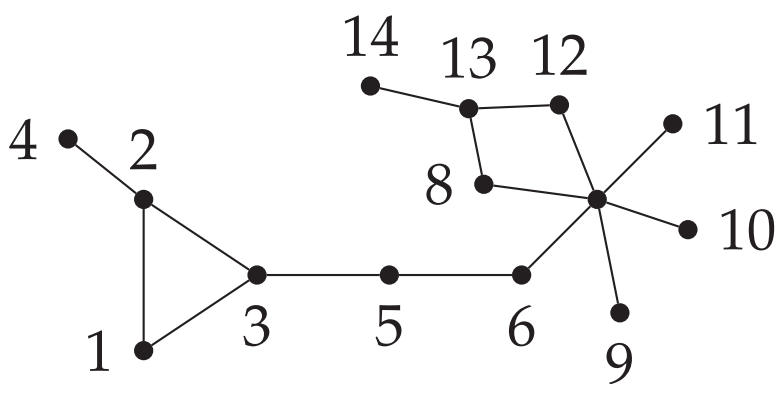

(**a**) Original graph (network) with fourteen vertices.

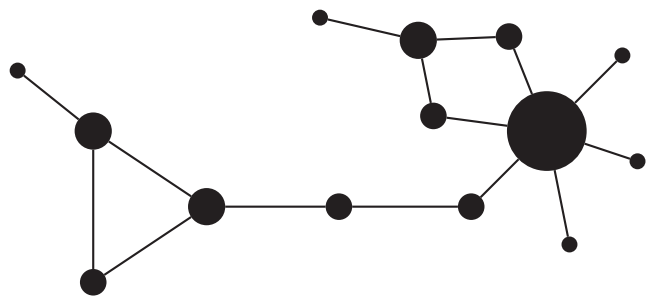

(**b**) Degree centrality of the graph.

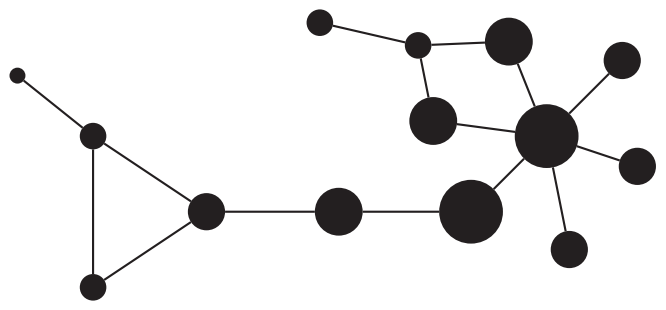

(**c**) Closeness centrality.

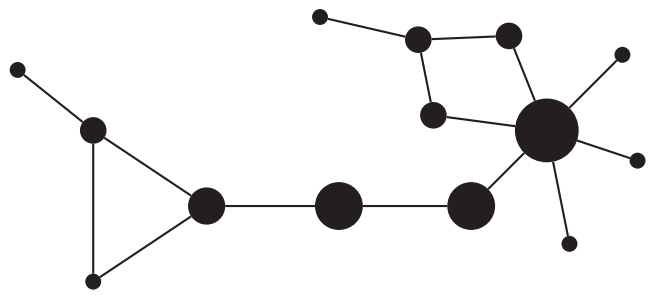

(**d**) Betweenness centrality.

**Figure 3.5:** Vertices of the original graph in Figure 3.5a can be associated with different centrality values. The centralities in all three examples are normalised and the size of their circles illustrate larger values. One can clearly see visual differences across the graphs between the different notions of centrality, e.g. how the local neighborhood influences the resulting centrality values.

## 3.3 NOTIONS OF SIMILARITY BETWEEN GRAPHS

Because this work is concerned about the underlying structure of neural networks in which structures are represented with graphs, we provide some overview of different notions of similarity between graphs to study the relationship between structures. Similarity can be looked at between two graphs $G_1$ and $G_2$, between two sets of graphs $\mathcal{G}_1$ and $\mathcal{G}_2$, as whether a graph $G$ comes from a probability distribution over graphs $Pr_{graph}$, or even how two learned distributions $Pr_1(\mathcal{G} \mid D_{obs,1})$ and $Pr_2(\mathcal{G} \mid D_{obs,2})$ relate.

Similarity between two graphs can be approached from different levels of granularity as depicted in Figure 3.6. Checking the hard equivalence of graphs falls under the term *graph isomorphism problem* and it is in general not known whether graph isomorphism is solvable in polynomial time or is NP-hard [12]. A similarity (or equivalently distance) can easy be generalised to sets of graphs and for graph isomorphism two sets of graphs can result in a pairwise binary matrix indicating isomorphism or not.

Such a pairwise binary matrix $\mathrm{M} \in \{0,1\}^{n \times n}$ can e.g. be defined as a symmetric one over one set as:

$$\forall i,j \in \{1,\dots,n\}:$$

$$\mathrm{M}_{\simeq}(G_1,\dots,G_n)_{ij} = \begin{cases} 1 & \text{if } G_i \simeq G_j \\ 0 & \text{else} \end{cases}$$

and can take the following exemplary form:

$$\mathrm{M}_{\simeq}(G_1, \dots, G_5) = \begin{array}{c} \\ G_1 \\ G_2 \\ G_3 \\ G_4 \\ G_5 \end{array} \begin{array}{c} \begin{array}{ccccc} G_1 & G_2 & G_3 & G_4 & G_5 \end{array} \\ \begin{bmatrix} 1 & & & & \\ 1 & 1 & & & \\ 1 & 1 & 1 & & \\ 0 & 0 & 0 & 1 & \\ 0 & 0 & 0 & 1 & 1 \end{bmatrix} \end{array}$$

in which it can be observed, that there are only two non-isomorphic graphs among the five graphs $G_1, \dots, G_5$. Alternatively, a similarity matrix $\mathrm{M}_{\simeq}(\mathcal{G}, \mathcal{H}) \in \{0,1\}^{n \times m}$ over two sets of graphs $\mathcal{G}$ and $\mathcal{H}$ can be formulated with $n = |\mathcal{G}|, m = |\mathcal{H}|$ as:

$$\forall i \in \{1, \dots, n\}, j \in \{1, \dots, m\} :$$

$$\mathrm{M}_{\simeq}(\{G_1, \dots, G_n\}, \{H_1, \dots, H_m\})_{ij} = \begin{cases} 1 & \text{if } G_i \simeq H_j \\ 0 & \text{else} \end{cases}$$

A binary similarity matrix can then be aggregated into sums or averages to compute a single scalar as similarity measure.

More fine-grained is the graph edit distance (graph edit distance (GED)) which is the number of operations to transform one graph into another. The GED is, however, NP-hard to compute [334] and deriving a similarity matrix for two sets of graphs can get time-consuming even for small graph orders.

Network properties of graphs provide global scalar features such as the order and size or distributions based on vertex-wise properties such as the degree, clustering coefficient or laplacian spectrum histograms. Statistical divergences (compare Section 5.3 on metrics and statistical distances) between histograms of graphs or aggregated histograms of sets of graphs have been widely used to e.g. evaluate generative models of graphs.

We used statistical divergences to compare the two graph generative models GraphRNN [328] and GRAN [160] in [282] and made similar observations as presented by Obray et al. [208]. It is, for example, a beneficial addition to compare large sets of graphs via kernel methods or other distances in a learned graph manifold space by embedding graphs into manifolds of a pre-trained graph classifier, similar to what was already employed in the image generative domain with e.g. the Fréchet Inception Distance [100]. Obray et al. compare more evaluation metrics for graph generative models and provide a good overview of the topic in the field of graph generation.

Learning similarities between graphs has become an own field of study. A taxonomy and overview of recent work is presented by Ma et al. [177] in which they categorise methods by model architecture and employed features.

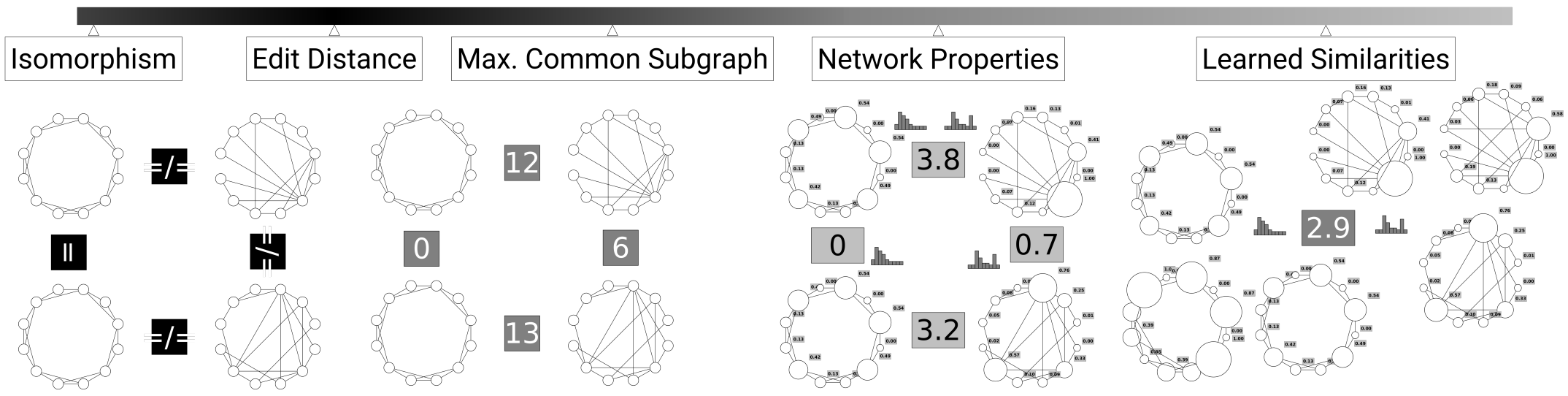

**Figure 3.6:** Computing isomorphism between graphs is likely to be computationally expensive. Interestingly "isomorphism of graphs of bounded degree can be tested in polynomial time" [12] but in general it is not known whether graph isomorphism is solvable in polynomial time. The graph edit distance provides a similarity measure between graphs but is also computationally expensive [334]. Network properties can be used to compare both graphs and sets of graphs. We've been using kernel- and learned graph embeddings [282] to compare sets of graphs and learning similarity between graphs has become its own field of study [177].

## 3.4 DIRECTED ACYCLIC GRAPHS AS NEURAL NETWORK STRUCTURES

In several experimental settings, we are interested in directed acyclic graphs (DAGs) as the building blocks of neural networks. As we will see later, this leads to a formalisation of constructing function sets based on sets of directed acyclic graphs and thus motivates thoughts on these sets of graphs.

An initial question about DAGs is: how many directed acyclic graphs are there for a given number of nodes? Knowing this number then gives us an understanding of how fast this set is growing w.r.t. the number of nodes and thus also how likely it is to draw a single graph uniformly between a lower and upper bound on the number of nodes. But it also motivates research in sampling graphs from infinite support [277].

From Robinson [236] we know a recurrence relation for the sequence of labelled directed acyclic graphs given as

$$a_n^{\texttt{A003024}} = \sum_{k=1}^{n} (-1)^{k-1} \binom{n}{k} 2^{k(n-k)} a_{n-k}$$

In Figure 3.7 we see the growth of variants of directed acyclic graphs. The sequences `A003024`, `A003025`, `A003026` and `A003026` refer to the identifiers of the online encyclopedia of integer sequences [112].

### 3.4.1 *Transforming Undirected Graphs Into Directed Acyclic Graphs*

In some experiments as in [272] we considered undirected instead of directed acyclic graphs. This is problematic but not entirely unjustified. We argue for the usage and analysis of directed acyclic graphs in section 6.4 on page 114 but it is also not uncommon to look at undirected graphs as underlying structure of deep neural networks, i.e. compare

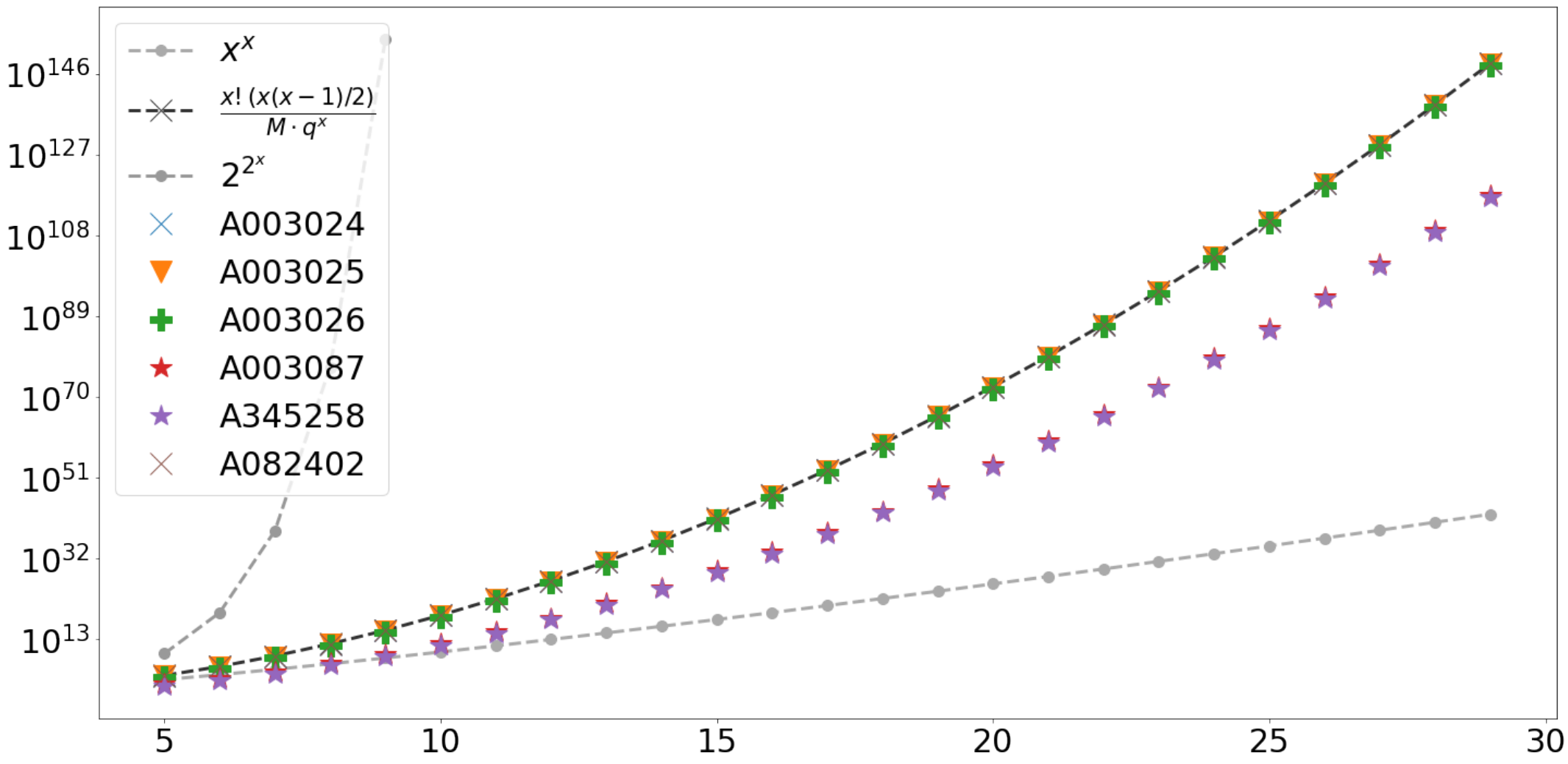

**Figure 3.7:** This plot shows analytically known sequences for the number of directed acyclic graphs with $n$ labelled vertices (A003024), of DAGs with one out-point (A003025) and two out-points (A003026) and the number of DAGs on *unlabelled* vertices with one source and one sink (A345258) as well as some exemplary functions from big-O notation to compare them visually.

[327]. The usage of undirected graphs is often motivated biologically in the sense that dentrites or axons are not (entirely) directed. Also, network properties are easier analysed for undirected graphs or defined differently.

However, the effect of changing the employed graph definition is often not explored properly. This is especially the case when undirected graphs are used for network analysis but the implementation or definition of deep neural networks requires directed acyclic graphs. A simple exemplary approach is to use undirected graphs for the analytic part but turning them into directed acyclic graphs for the technical construction of neural network realisations. It is then not always clear how network properties change. For example, while the degree distribution usually stays the same between undirected graphs and directed acyclic graphs, for directed ones it divides into in-degree and out-degree distributions and the number of paths drastically changes. This makes the definition of path lengths and their averaged distribution properties for the whole graph entirely different compared to its undirected version.

We leave deeper analysis of this relationship to future research but want to point out to the fact, that this aspect of structure analysis needs to be considered when focusing on undirected graphs and transforming them for computational forward directions into directed acyclic graphs.

### 3.4.2 *Learning Models of Directed Acyclic Graphs*

Counting and sampling directed acyclic graphs is particularly interesting to guide searches through these sets and coincides with problems of optimizing neural network architectures. Learning structures of bayesian networks is a related problem and studied e.g. by Talvitie et al. [278]. Sampling random graphs is difficult and approaches to it involve e.g. boltzmann samplers [58, 214]. Uniform sampling of directed acyclic graphs is possible in $O(n^5)$ with an algorithm of Kuipers & Moffa [140], for which we provide a detailled implementation in the appendix in appendix A.6 on page 318. These combinatoric works are important as underlying basis for the subsequent optimisation problems considered in section 7.1 on page 122. We obtain analytical arguments such as that "the average number of edges in an acyclic directed graph on $n$ vertices is $\sim n^2/4$" [193]. The work of Kuipers & Moffa [140] for example allows us to randomly and uniformly sample graphs $G_1 \sim UDAG$.

However, a large aspect of our research is also considered with learning distributions of (directed acyclic) graphs as presented in chapter 14 on page 261. If such models can properly be learned, we can obtain samples of graphs $G_1 \sim P_{DAG}$ from a distribution $P_{DAG}$ that allows more sophisticated draws than a random selection. Learning such a distribution of directed acyclic graphs from data, then is part of a bi-level (bayesian) optimisation process and can be understood as learning a structural prior of architectures where each graph $G_1, \ldots$ represents a (universal) architectural prior from which actual neural network realisations can be obtained. Also compare section 5.4 on page 60 on maximum a-posteriori estimation to this interpretation.

The graph theoretical and combinatorial work should make it clear that learning deep generative models for directed acyclic graphs is also not trivial and deserves a lot attention on its own.

# 4 GAME THEORY: THE SHAPLEY VALUE

*The following entails:*

We use the Shapley Value (SV), a game-theoretic concept, to quantify the contribution of structural elements of deep neural networks [270]. The following chapter gives the necessary background and overview on game theory with a focus on the Shapley Value as to understand our contribution in pruning and neural architecture search.

Main takeaways for the informed reader are:

- With a coalitional game on a set of players $U$ and a payoff function $v$, the Shapley Value (SV) $\phi_v$ assigns contributional values $\phi_v(i)$ to players $i \in U$.
- The SV is combinatorial in nature and requires an approximation for large $U$. This applies to our application case such that we provide an overview of approximation methods in section 4.3 on page 46 to obtain SVs $\phi_v(i)$.

## 4.1 INTRODUCTION TO GAME THEORY

Our introduction to game theory follows Martin J. Osborne [211]. According to Osborne game theory "aims to help us understand situations in which decision-makers interact" and "consists of a collection of models". These models are analyzed to "discover its implications" and the three main model types of game theory are strategic games, extensive games, and coalitional games. Here, we're just focussing on coalitional games which are "designed to model situations in which players can beneficially form groups, rather than acting individually" [211, Chp. 8]. This is interesting in that we can bring a game-theoretic perspective to NAS methods by treating structural elements as players in a coalitional game in which groups of these players contribute to the neural network's performance in a problem domain such as solving an image classification task.

COLATIONAL GAMES A coalitional game is a model focussing "on the behavior of groups of players" [211, Chp. 8]. It consists of a set $U = \{u_1, \ldots, u_n\}$ of $n \in \mathbb{N}$ players and a payoff function $v : \mathcal{P}(U) \to \mathbb{R}$ which assigns a contributional value to a coalition $S \subseteq U$. In the words of Alvin Roth "the interpretation of $v$ is that for any subset $S$ of $U$ the number $v(S)$ is the worth of the coalition, in terms of how much *utility* the members of $S$ can divide among themselves in any way that sums to no more than $v(S)$ if they all agree" [241]. Based on the set of players and the payoff function the "outcome of a coalitional game consists of a partition of the set of players into groups" [211, Chp. 8].

Two popular **examples** are the *glove game* and a game modeled on the *United Nations Security Council* (UNSC). The glove game defines a set of three player $U_{glove} = \{1, 2, 3\}$ in which players 1 and 2 produce right-hand gloves and player 3 produces left-hand gloves. A payoff function for the glove game can be defined as

$$v_{glove}(S) = \begin{cases} 1 & \text{if } |S| > 1 \wedge 3 \in S \\ 0 & \text{else} \end{cases}$$

, indicating that a pair of gloves can only be sold if both a right-hand and a left-hand glove are available through the production of the coalition $S \subseteq U$. A solution for the glove game asks for an answer to the question of which coalition should be formed or how to properly share collective income amongst participants of a coalition from the production of gloves. While both questions might seem answerable for the glove game, the game on the UNSC shows the need for a methodological approach.

The game on the United Nations Security Council consists of fifteen members. Five members are permanent and have a veto while the remaining ten seats are rotating members. The council comes to an agreement if there are nine votes and no vetos. We can define the game with $n = 15$ players $U_{unsc} = \{1, 2, \dots, n\}$ and an affirmative count $a = 9$. From the overall members, $p = 5$ are permanent $P_{unsc} = \{1, 2, \dots, p\} \subset U_{unsc}$ members. The payoff function is then

$$v_{unsc}(S) = \begin{cases} 1 & |S| \geq a \wedge \forall p \in P_{unsc} : p \in S \\ 0 & \text{else} \end{cases}$$

which states whether the coalition $S$ comes to an agreement if all group members agree. The number of coalitions to look at grows to $|\mathcal{P}(U_{unsc})| = 2^{15} = 32,768$. Obviously, the question of how much each member contributes to the overall outcome gets way more difficult. A sophisticated answer can be found with the following argumentation which uses properties and arguments of a solution concept called the Shapley Value.

Roth [241] derives fair attributed values $\phi_{v_{unsc}}(i)$ for each member $i \in U_{unsc}$: all rotating members $r \in U_{unsc} \backslash P_{unsc}$ should have the same value $\phi^r_{v_{unsc}}$ and also all permanent members $p \in P_{unsc}$ should have the same value $\phi^p_{v_{unsc}}$ and they need to add up to one. Then one obtains: $(n-$

$p) \cdot \phi^r_{v_{unsc}} + p \cdot \phi^p_{v_{unsc}} = 1$. Roth further argues on marginal contributions of rotating members: "to make a positive marginal contribution in a random order, all five permanent members and exactly three of the other nine rotating members must precede her" [241, p. 7]. With $9!/3!6!$ such coalitions and a coalition $S$ of size $|S| = 9$ occurring with probability $(s-1)!(n-s)!/n!$, and "the marginal contribution of the last rotating member" being $v(S) - v(S\backslash\{i\}) = 1$, the attribution value to the rotating members can be found to be $\phi^r_{v_{unsc}} = (9!/3!6!)(8!6!/15!) = 4/2145 \approx 0.001865$. Permanent members are then attributed with $\phi^p_{v_{unsc}} = (1 - (n-p)\phi^r_{v_{unsc}})/p = 421/2145 \approx 0.196270$. This derivation made use of the axiomatic properties symmetry, efficiency and additivity of the Shapley Value which make this solution concept being the unique function satisfying these properties [251].

## 4.2 SHAPLEY VALUE

The Shapley Value was introduced by Lloyd Shapley [251] as a solution concept for coalitional games. Shapley required three properties for the coalitional game and its payoff function $v$ [251, p. 294 -2-]:

1. $v(\emptyset) = 0$,
2. $v(S) \geq v(S \cap T) + v(S\backslash T)\ \forall S, T \subseteq U$
3. $N \subseteq U$ is carrier of $v \Leftrightarrow v(S) = v(N \cap S)\ \forall S \subseteq U$

and "restricted attention to games with finite carriers" [251, p. 2]. Based on these axioms, the Shapley Value $\phi_v : U \rightarrow (0, 1)$ is defined for player $i \in U$ as

$$\phi_v(i) = \sum_{S \subseteq U} \gamma_n(s)\,(v(S) - v(S\backslash\{i\}))\ \text{[251, eq. (13)]} \tag{4.1}$$

$$\text{with } \gamma_n(s) \triangleq (|S|-1)!(n-|S|)!/n!\ \text{[251, eq. (12)]}$$

$$= \frac{1}{n!} \sum_{S \subseteq U\backslash\{i\}} (|S|!(n-|S|-1)!) \cdot (v(S \cup \{i\}) - v(S)) \tag{4.2}$$

$$= \frac{1}{|U|!} \sum_{\pi \in \Pi} \left(v(P^\pi_i \cup \{i\}) - v(P^\pi_i)\right), \tag{4.3}$$

in which $\Pi$ is the set of all permutations of $U$ and $P^\pi_i = \{j \in U : \pi(j) < \pi(i)\}$. The subset-based eq. (4.2) is commonly used in literature. Note, that w.r.t. to the amount of players it takes up to $2|U-1|!$ payoff evaluations. In game settings with large amounts of players or with expensive payoff evaluations, this usually requires an approximation, e.g. by assumptions on coalitional structures or acceptance of error bounds.

Shapley proves [251, Thm 13, p. 7] that there is only one function, namely the Shapley Value $\phi$, fulfilling the three axioms *symmetry*, *carrier* and *additivity* and notes that remarkably "no further conditions are required to determine the value uniquely". The first axiom, *symmetry*, "requires that the names of the players play no role in determining the value" [241, p. 5]. The second axiom, *efficiency*, requires for carriers $N$ that $\sum_{i \in N} \phi_v(i) = v(N)$. The third axiom, *additivity* (also called *law of aggregation*), requires for two games $v$ and $w$ that $\phi_{v+w} = \phi_v + \phi_w$ [251, p. 3].

We are interested in three aspects of the Shapley Value: 1/ employing the Shapley Value as an attribution method for structural elements during pruning or growing neural nets and therefore 2/ obtaining them computational efficiently and 3/ understanding their general behaviour in settings with neural networks.

Approximating the actual values is important as the computational complexity grows quite early even for small games with player sizes of $n > 8$. We have been using distributed setups in our experiments with random subsampling of $S \subseteq U$ but even then it has to be considered that evaluating a single value $v(S)$ could require an expensive training pipeline of a neural network built with structural components defined through the coalition $S$.

We conducted a theoretical analysis in section 11.4 of Shapley Values with neural networks as non-additive payoffs. While the lower bound for any such payoff is equal to $-\frac{n-1}{n}$ we observed empirically that obtained contributional values almost never come close to this bound. We show in the theoretical analysis arguments for that observation and the most compelling argument is that practically used payoffs, even when non-additive, can have a loose property of monotonic growth such that large negative Shapley Values occur only for players $i \in U$ if they almost always hurt a coalition such that $v(S \cup \{i\}) - v(S) \to -1$. In the case of players representing structural elements of neural network, this would mean that certain added elements let the performance always drop, but that's not the case. While payoffs on structural elements of neural networks are non-superadditive or even non-monotone, they usually behave closely like it. In other words, more parameters of a neural network usually don't let it's performance decrease and one obtains mostly non-negative or very small negative contributional values.

## 4.3 APPROXIMATING SHAPLEY VALUES

Computing the exact Shapley Values of all players in a game requires evaluating the payoff of all coalitions and is NP-hard [51]. Mann & Shapley already worked on approximating the values of large games by means of Monte Carlo simulation [187]. Since Lundberg et al. started transferring Shapley Value to interprete the feature importances of deep neural networks with their seminal work on their method SHAP [174] a lot of interest has sparked in the machine learning community on improving the approximation capabilities of Shapley Values.

We listed significant contributions to the approximation of Shapley Values in Table 4.1. Only recently Kolpaczki et al. published their work on *Shapley Value Approximation without Requesting Marginals* (SVARM) in which they sample payoffs of coalitions, that are used to update a whole set of approximation values for multiple players [135]. The interested reader might refer to Chen et al. [37, p. 10, Table 1] who provide an extensive overview for approximation methods for Shapley Value with focus on attribution to deep neural networks.

A most naïve Monte Carlo approximation for the subset-based definition in Equation 4.2, as we defined it in [270, p.4], uses a randomly chosen subset $R \subset \mathcal{P}(U)$:

$$\phi_v^R(i) = \frac{1}{\sum_{S \in R} |S|!(n-|S|-1)!} \sum_{S \in R} (|S|!(n-|S|-1)!) \cdot (v(S) - v(S \backslash \{i\}))$$

and an approximation for the permutation-based definition Equation 4.3 uses $r$ random permutations $\Pi^r$ such that:

$$\phi_v^{\Pi^r}(i) = \frac{1}{r} \sum_{\pi \in \Pi^r} \left( v(P_i^\pi \cup \{i\}) - v(P_i^\pi) \right)$$

.

Maleki et al. [186] analysed error bounds of an approximation algorithm for Shapley Values based on stratified sampling. The algorithm can be inspected in algorithm 1. Stratified sampling takes groups or subpopulations (= strata) into account and makes assumptions about them having homogeneous properties. Members of coalitions in these strata are assumed to have values close to eachother. With a sampling number $m$, one can achieve the following bounds according to the proof in [186]: $|\hat{\phi} - \phi| \leq \frac{d\sqrt{-ln\frac{\delta}{2}}}{n\sqrt{m}} \sum_{k=0}^{n-1} (k+1)$ which is "significantly higher (i.e., worse) if $m > \frac{(n+1)^2}{4}$.

Kolpaczki et al. take a recent approach of dividing the original equation into signed Shapley Values with a positive Shapley Value $\phi_v^+(i)$ and a negative Shapley Value $\phi_v^-(i)$ by rearranging Equation 4.1: $\phi_v(i) = \sum_{S \subseteq U} \gamma_n(s) \cdot v(S) - \sum_{S \subseteq U} \gamma_n(s) \cdot v(S \backslash \{i\}) = \phi_v^+(i) - \phi_v^-(i)$. For these signed Shapley Values, the underlying distributions are updated based on sampled coalitions in an alternating fashion [135, Sec. 3 SVARM]. The empirical results [135, Sec. 5] are promising and based on games of sizes between fourteen and 100 players. Most applications of Shapley Value approximation algorithms are still played on games defined over feature selection for deep neural networks. Game sizes can easily increase when treating fine-grained structural elements of deep neural networks as players of the game, as intended in this work.

We provide our de-coupled implementation from [270] on github.com/JulianStier/pyshapley and refer to benedekrozemberczki/shapley for selected alternative approximation algorithms. A coalitional game can then by simply defined by the set of players (here $a$, $b$, and $c$) and the payoff value for each coalition as denoted in Listing 4.1.

**Algorithm 1:** Approximation algorithm for the Shapley Value referred to as *Stratified Random Sampling* [186, p. 12, Alg. 2]. The approach is based on the "objective to distribute $m$ samples among $k$ strata, such that the total estimation error of the Shapley Value is minimised" [186, Sec. 5].

**input** : The game $(N, v)$, number of desired samples $m, \delta$
**output**: A list of estimated Shapley Values per player $\hat{\phi}$

$\forall i \in N : \hat{\phi}(i) \leftarrow 0$ `// Initially set SV estimations to zero`
**for** $i \leftarrow N$ **do**
  $\forall k \in \{1, 2, \dots, i-1, i+1, \dots, n\} : S_k \leftarrow \mathit{CoalitionsOfSize}(k)$;
  $\forall k \in \{1, 2, \dots, i-1, i+1, \dots, n\} : m_k = \lfloor \frac{m(k+1)^{2/3}}{\sum_{j=0}^{n-1}(j+1)^{2/3}} \rfloor$;
  `//` $m_k$ `are the number of samples to be taken from stratum` $S_k$
  **if** $m - \sum m_k > 0$ **then**
    UniformlyDistributeRemainingSamplesAmongStrata()
  **end**
  $\forall k \in \{1, 2, \dots, i-1, i+1, \dots, n\} : \hat{\phi}^k(i) \leftarrow \mathit{AverageOfStratum}(k)$;
  $\hat{\phi}(i) \leftarrow \mathit{AverageOfAllStrata}()$
**end**
**return** $\hat{\phi}$

**Listing 4.1:** A coalitional game can be manually defined by denoting its player set, e.g. $\{a, b, c\}$, and specifying each payoff for all possible coalitions. Coalitions can be denoted in string-form as to not use a set notation, e.g. `"ac"` means $\{a, c\}$. The payoff needs to be fully determined to be used for calculating the solution with Shapley values. Approximations can be done by stating a payoff function that conducts an evaluation given only a single coalition. The payoff is then defined implicitly only for specific cases when needed.

```python
import pyshapley
game = pyshapley.game.ManualDictGame(
    {"a", "b", "c"},
    {
        "": 0,
        "a": 100,
        "b": 0,
        "c": 0,
        "ab": 150,
        "ac": 120,
        "bc": 0,
        "abc": 300,
    },
)
solution = pyshapley.solution.Shapley(game)
for p in game.players:
    print("phi(%s) = %.4f" % (p, solution.shapley(p)))
```

---

**Algorithm 2:** An unstratisfied version of SVARM [135], that shows good theoretical properties for large games.

---

**input** : The game $(N, v)$, $T$

**output**: A list of estimated Shapley Values per player $\hat{\phi}$

$\forall i \in N : \hat{\phi}^+(i), \hat{\phi}^-(i) \leftarrow 0$ `// Initially set signed Shapley Value estimations to zero`

$\forall i \in N : c_i^+, c_i^- \leftarrow 1$ `// Running sample number of each estimate`

$t \leftarrow 2n$;

**for** $i \in N$ **do**

    `// Warmup phase`

    $A^+ \overset{i.i.d.}{\sim} P^w; A^+ \overset{i.i.d.}{\sim} P^w$;

    $v^+ \leftarrow v(A^+ \cup \{i\}; v^- \leftarrow v(A^-)$;

**end**

**while** $t + 2 \leq T$ **do**

    $A^+ \sim P^+$ `// Sample a coalition from the positive SV distribution`

    $A^- \sim P^-$ `// Coalition from negative SV distribution`

    $v^+ \leftarrow v(A^+); v^- \leftarrow v(A^-)$;

    **for** $i \in A^+$ **do**

        $\hat{\phi}^+(i) \leftarrow \frac{c_i^+ \hat{\phi}^+(i) + v^+}{c_i^+ + 1}$;

        $c_i^+ \leftarrow c_i^+ + 1$;

    **end**

    **for** $i \in N \setminus A^-$ **do**

        $\hat{\phi}^-(i) \leftarrow \frac{c_i^- \hat{\phi}^-(i) + v^-}{c_i^- + 1}$;

        $c_i^- \leftarrow c_i^- + 1$;

    **end**

    $t \leftarrow t + 2$;

**end**

$\forall i \in N : \hat{\phi}(i) \leftarrow \hat{\phi}^+(i) - \hat{\phi}^-(i)$;

**return** $\hat{\phi}$

---

| **Title** | Year | **Ref** | **Method**(s) | **Notes** |
|---|---|---|---|---|
| Values of large games | 1962 | [187] | Monte Carlo simulation | |
| Multilinear extensions of games | 1972 | [212] | Multilinear extension | |
| A randomized method for the Shapley value for the voting game | 2007 | [64] | R-ShapleyValue | $k$-majority games |
| Polynomial calculation of the Shapley value based on sampling | 2009 | [35] | ApproShapley | |
| Computing Shapley Value in Supermodular Coalitional Games | 2012 | [161] | | Supermodular games |
| Bounding the Estimation Error of Sampling-based Shapley Value Approximation | 2013 | [186] | | |
| A unified approach to interpreting model predictions | 2017 | [174] | KernelSHAP | Aproximating weighted least squares |
| A new approximation method for the Shapley value applied to the WTC 9/11 terrorist attack | 2018 | [33] | | |
| Shapley: Discovering the Responsible Neurons | 2020 | [75] | Truncated Monte Carlo Shapley, Gradient Shapley | |
| Fastshap: Real-time shapley value estimation | 2021 | [119] | FastSHAP | |
| A Bayesian Monte Carlo method for computing the Shapley value: application to weighted voting and bin packing games | 2021 | [283] | | |
| Approximating the shapley value without marginal contributions | 2023 | [135] | SVARM, Stratified SVARM | |

**Table 4.1:** Obtaining Shapley Values of large games, e.g. with $n >>> 100$ is computationally very expensive. Although there have been early works on approximating values of large games as in [187], many new approximation methods came up during the last decade. In [270], we've been employing Monte Carlo simulations. While the setting of using Shapley Values for pruning showed to be only significant in realms when the number of structural elements get small anyways, we observed that the reliability of approximated values for very large game sizes is only satisfactory with high numbers of approximation samples. Works such as [135] seem to be promising in building games on deep neural networks with a high number of structural elements as players. Yet, it will assumably still be difficult when considering each neuron of State-of-the-art models as players because game sizes then would easily be in the millions.

# 5 DEEP NEURAL NETWORKS

*The following entails:*

The following chapter summarises formal groundings of deep neural networks (DNN), the second object of our investigation.

Main takeaways for the informed reader are, that

- most notation and terminology follows the one used in the machine learning community,
- several theoretical descriptions take shortcuts and overload symbols as of the complexity (and an appreciation thereof) of the underlying theories,
- and that the definitions of neural networks and their *structure* can vary between problem settings.

Our work includes many experiments such as analysing learned deep neural network realisations as image classification models for MNIST. This poses data as an important aspect of our work but we still deem the formal groundings important to foster new ideas or strengthen qualitative arguments about differences observed in experiments.

The formal introduction intends to pave the way for chapter 6, which describes sets of neural networks being constructed from graphs (graph-induced neural networks). With the introduced notation we emphasise on a clear distinction between a deep neural network realisation and the actual architectural space it lies in and which it represents when assessing its properties. From a theoretical perspective the work of Petersen [223] is probably closest to our formal reasoning. With universal architectures $\mathcal{A}([G])$ induced from a directed acyclic graphs $G$ we not just compare finite-parameterised architectures (i.e. finite-paramterised sets of functions) but architectures which are theoretically capable of universal approximation, i.e. having a universal approximation property. Therefore, the formal background on deep neural network in sections 5.5 to 5.7, and on universal approximation theorems in section 5.9 on page 92 are vital to understand our reasoning for introducing the notation of universal architectures in chapter 6.

## 5.1 INTRODUCTION

Graph-induced neural networks for neural architecture search, design space automation or neural structure analysis can be broadly approached from two opposing sides:

1. An experimental *top-down* side, which is more concerned about technical implementations, realisations in software and proper statistical experiment settings as to decide about choices for which strong analytical arguments will not be made in the near future,
2. and an theoretical *bottom-up* side, which is more concerned about formal rigor, unification and generalisation which sheds light on theoretical limits of methods or understandings.

We try to give both sides their space but with the compromise of not making a claim on formal soundness or rigor rather than communicating core ideas. The subsequent section gives an overview, section 5.2 introduces our used notation to describe settings with (real-world) data, on which we will measure and compute, e.g. with metrics and statistical divergences as mentioned in section 5.3 as to then conduct *machine learning*.

Machine learning has interpretations and a formal background in statistical learning, outlined in section 5.4. We then restrict our attention to (deep) neural networks for which we introduce the basic components in section 5.5 but will deepen these descriptions by contributing thoughts and formalisations in chapter 6 for graph-induced neural networks. This chapter 6 can be sufficient for the informed reader to understand the problems and ideas tackled in subsequent experimental chapters.

PROBABILITY THEORY Our notation orientates on Murphy [200], Roch [281], and partially on Polyanskiy & Wu [227]. We commonly denote a measurable space $(\mathcal{X}, \mathcal{E})$ with sample space $\mathcal{X}$ and $\sigma$-algebra $\mathcal{E}$ and assume them to be standard Borel spaces throughout this work, compare [200, Chp. 2]. By choosing a probability measure $\mathbb{P} : \mathcal{X} \to [0, 1]$ we obtain a probability space $(\mathcal{X}, \mathcal{E}, \mathbb{P})$ for which we can refer with $X : \mathcal{X} \to \mathbb{R}$ to real-valued random variables [281, Sec. B.1.12, p. 505]. The extension to a $d$-dimensional random vector $X = (X_1, \dots, X_d) : \mathcal{X} \to \mathbb{R}^d$ with $d \in \mathbb{N}^+$ follows accordingly.

## 5.2 DATA

A single data sample $\mathbf{x}$ is represented with one or multiple different features such that it can be written as $\mathbf{x} = (x_1, \dots, x_n) \in \mathbb{R}^d$ with $d \in \mathbb{N}$ being the number of features. As we saw in previous examples, the sample $\mathbf{x}$ often carries additional structure such as a dimension of discretised time steps or a finite number of dimensions containing different color channels for images. An example of such a data sample with its index notation is given in Figure 5.1. We then can e.g. write $\mathbf{x} \in \mathbb{R}^c \times \mathbb{R}^{f_1 \times f_2}$ with $c, f_1, f_2 \in \mathbb{N}$. This feature space is contained in $\mathbb{R}^d$ with $d = c + f_1 \cdot f_2$. Data samples are then stored in multi-dimensional arrays or *tensors* for which we follow the notation from Chiang et al. [39] and understand them as in Laue et al. [146] without the Ricci notation for distinguishing between covariant and contravariant components. The multi-dimensional data sample array $\mathbf{x}$ is often flattened out from

$$\overbrace{\mathbf{x} = \text{height}\overset{\text{width}}{\begin{bmatrix} x_{011} & x_{012} & \cdots & x_{01f_2} \\ x_{021} & x_{022} & \cdots & x_{02f_2} \\ \vdots & \vdots & \ddots & \vdots \\ x_{0f_11} & x_{0f_12} & \cdots & x_{0f_1f_2} \end{bmatrix}} \| \begin{bmatrix} x_{111} & x_{112} & \cdots & x_{11f_2} \\ x_{121} & x_{122} & \cdots & x_{12f_2} \\ \vdots & \vdots & \ddots & \vdots \\ x_{1f_11} & x_{1f_12} & \cdots & x_{1f_1f_2} \end{bmatrix} \| \dots}^{0 \to \text{channel} \dots}$$

$$\overbrace{\| \,\text{height}\overset{\text{width}}{\begin{bmatrix} x_{c11} & x_{c12} & \cdots & x_{c1f_2} \\ x_{c21} & x_{c22} & \cdots & x_{c2f_2} \\ \vdots & \vdots & \ddots & \vdots \\ x_{cf_11} & x_{cf_12} & \cdots & x_{cf_1f_2} \end{bmatrix}}}^{\dots \text{channel} \to c}$$

into $\mathbf{x} = \left(x_{011}, x_{012}, \dots, x_{01f_2}, x_{021}, \dots, x_{0f_1f_2}, x_{111}, \dots, x_{1f_1f_2}, \dots, x_{c11}, \dots, x_{cf_1f_2}\right) \in \mathbb{R}^c \times \mathbb{R}^{f_1 \times f_2}$.

### 5.2.1 *Distributions*

We give a brief overview of selected important probability distributions and with which notation we refer to them. Like Murphy states in [200, Sec. 2.1, p.7], "the term *probability distribution* could refer to the probability density function $Pr_X$, to the cumulative distribution function $F_X$ or even the probability measure $\mathbb{P}$".

Some examples in which we explicitly meet distributions:

- Predictions of neural networks are usually considered to be distributed according a particular probability distribution. For example, classification problems have a finite and discrete number of possible class targets and are often uniformly distributed but modeled with a parameterisable categorical distribution as to learn a function that transforms a feature representation into a parameter of this distribution.

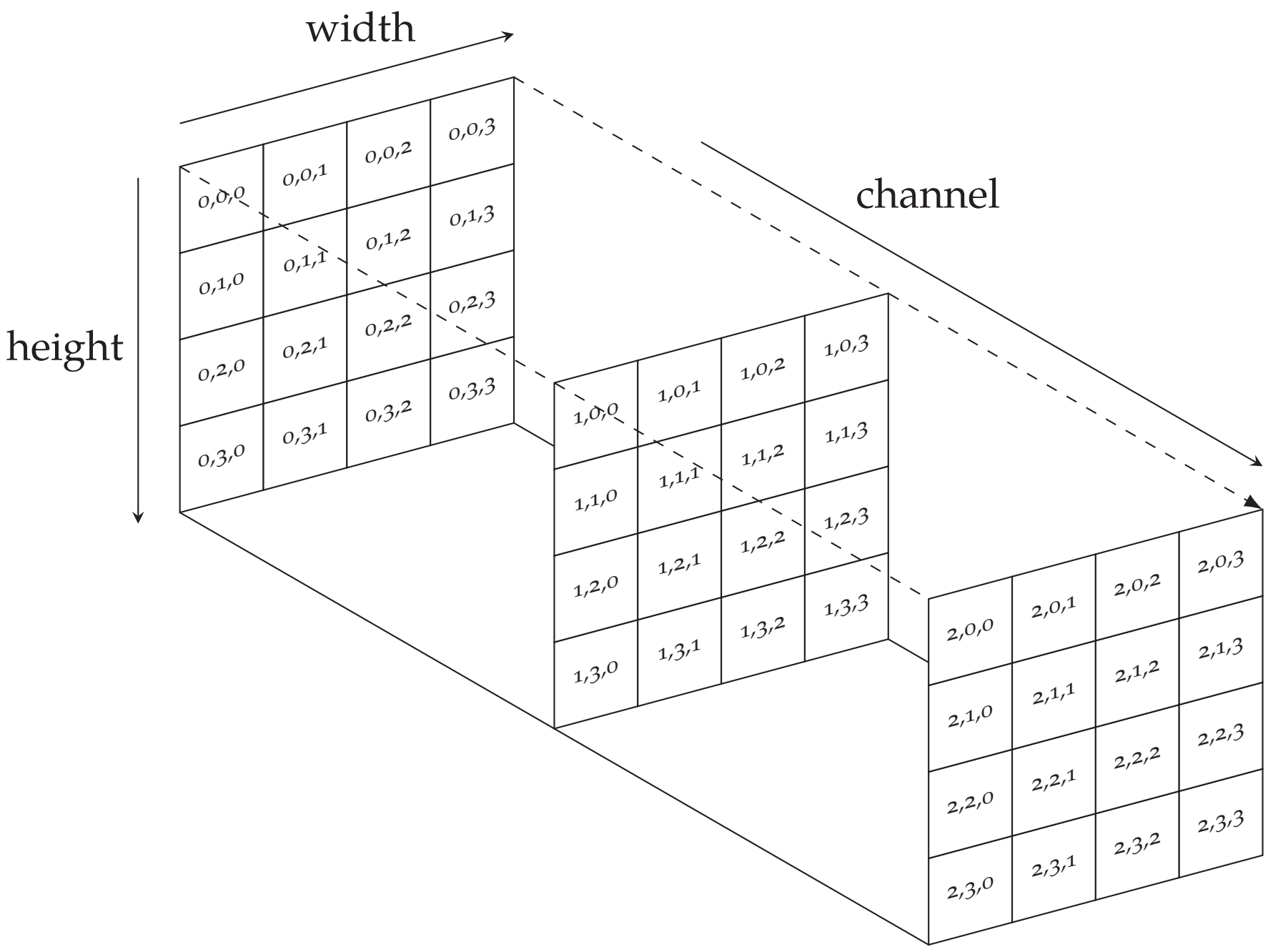

**Figure 5.1:** Depiction of a data tensor of ordered shape $(3, 4, 4)$ contained in $\mathbb{R}^{54}$ with named axes. Data samples $\mathbf{x} \in \mathbb{R}^{\text{chans}[3]\times\text{height}[4]\times\text{width}[4]}$ usually have a more complex structure than a plain vector $\mathbf{x} \in \mathbb{R}^{54}$, indicating real-world properties such as spatial correlations.

- The parameters of neural networks need to be determined through learning techniques and the associated initial value problem, i.e. the question of with which value to start learning procedures with, is crucial. This initial value is referred to as initial condition in dynamic systems. Commonly in machine learning, initial parameters are randomly initialised according to e.g. a (truncated) normal distribution.
- Experiment settings can be modelled according to probability distributions such as a bernoulli distribution and then referred to as bernoulli trial. Binary classification problems can be viewed under this perspective and interpreting experimental outcomes are then questions of hypothesis testing based on the underlying distributions fitting for the experiment.

### 5.2.2 *Discrete Distributions*

The binomial distribution Bin is defined as [200, Sec. 2.2.1]

$$\mathrm{Bin}(n,p) \triangleq \binom{n}{x} p^x (1-p)^{n-x}$$

with $\binom{n}{k} \triangleq \frac{n!}{(n-k)!k!}$ (the binomial coefficient) and for $n = 1$ it reduces to the Bernoulli distribution which is

$$\mathrm{Ber}(p) \triangleq \begin{cases} 1-p & \text{if } x = 0, \\ p & \text{if } x = 1 \end{cases}$$

with $\mathbb{E}[X] = p = Pr(X = 1)$ is the expectation of $X \sim \mathrm{Ber}(p)$. Murphy denotes it as $\mathrm{Ber}(X \mid \mu)$ as to express that the random variable $X$ is distributed according to $\mathrm{Ber}(\mu)$.

For a $k$-valued random variable $X$, the categorical distribution Cat is defined as

$$\mathrm{Cat}(X = x \mid \mathbf{p}) \triangleq \prod_{i=1}^{k} \mathbf{p}_i^{[x=i]}$$

with $x \in \{1, \dots, k\}$. For a one-hot encoding of $x$ as $\mathbf{x} \in \{0,1\}^k$, we get $\mathrm{Cat}(X = \mathbf{x} \mid \mathbf{p}) = \prod_{i=1}^{k} \mathbf{p}_i^{x_i}$ [200, Eq. 2.21, p.9].

### 5.2.3 *Continuous Distributions*

The gaussian (or normal[1]) distribution is given as

$$\mathcal{N}(\mu, \sigma) \triangleq \frac{1}{\sigma\sqrt{2\pi}} e^{-\frac{1}{2}\left(\frac{x-\mu}{\sigma}\right)^2}$$

In connection of weight initialisations for DNNs, there exist some modified versions of normal distributions made dependent on e.g. layer sizes, such as the Xavier type of initalisation [78] or the Kaiming He type of initalisation [94].

The distribution Gumbel with location parameter $\mu \in \mathbb{R}$ and scale parameter $\beta \in \mathbb{R}$ is given with density

$$\mathrm{Gumbel}(\mu, \beta) \triangleq \frac{1}{\beta} e^{(-z+e^{-z})}$$

and a sample $g$ can be drawn from $\mathrm{Gumbel}(0,1)$ by "using the inverse transform sampling" with drawing a uniform sample $u \sim \mathcal{U}(0,1)$ "and computing $g = -log(-log(u))$" [118, p. 2].

1 Karl Pearson [219] interestingly mentions to "call the Laplace-Gaussian curve the normal curve" with the "disadvantage of leading people to believe that all other distributions of frequency are [..] *abnormal*", which is "not justifiable". Nevertheless, the normal distribution is nowadays a common term for the Laplace-Gaussian distribution based on how naturally it occurs in many formulations.

### 5.2.4 *Concrete, Gumbel-Softmax and Reparameterisation*

Two major lines of developments regarding probability distributions are important for the subsequent interplay between discrete spaces in connection with graphs and continuous spaces in connection of deep neural networks:

1. The reparameterisation trick of Kingma & Welling [128] enabled the machine learning community to incorporate distributions into deep neural networks and treat them, despite their originally deterministic[2] formulation, more variational (often termed *Bayes*).
2. The Gumbel-Max trick [182] allowed a "continuous relaxation of discrete random variables" [181, p. 2] and resulted in the development of the Concrete [181] or analogously the Gumbel-Softmax distribution [118].

The Gumbel-Softmax distribution $\text{Concrete}(\alpha, \tau)$ models a continuous distribution over the simplex $\Delta^{n-1}$ with location parameter $\alpha$ and temperature parameter $\tau$ with density

$$\text{Concrete}(\alpha, \tau) \triangleq (n-1)!\tau^{n-1} \prod_{k=1}^{n} \left( \frac{\alpha_k x_k^{-\tau-1}}{\sum_{i=1}^{n} \alpha_i x_i^{-\tau}} \right)$$

with $\alpha \in (0, \infty)^n$, $\tau \in (0, \infty)$ and for $X \sim \text{Concrete}(\alpha, \tau)$ $X \in \Delta^{n-1}$ [181]. A sample can be drawn for a random variable $X \in \Delta^{n-1}$ with $X \sim \text{Concrete}(\alpha, \tau)$ by sampling $g_1, \dots, g_n \overset{i.i.d.}{\sim} \text{Gumbel}(0, 1)$ and setting $X_k = \frac{exp((log\alpha_k + g_k)/\tau)}{\sum_{i=1}^{n} exp((log\alpha_i + g_i)/\tau)}$ [181, Eq. 10] for $k \in \{1, \dots, n\}$.

## 5.3 METRICS AND STATISTICAL NOTIONS OF DISTANCES

The study on metrics and statistical distances and similarities are fundamental in statistics and we require them for the understanding of statistical learning dynamics and the quantitative evaluation of approximations and experiments. Metrics or statistical distances essentially describe the relation between two objects. The euclidean distance

$$\|p - q\| = \sqrt{(q_1 - p_1)^2 + (q_2 - p_2)^2 + \dots + (q_n - p_n)^2}$$

for vectors $p, q \in \mathbb{R}^n$ as a norm-induced metric is just an example of one of the most well-known such distances. Our two main application cases of distances are: *first*, comparing objects by comparing their distance in some space. That could be the assessment of the quality of generated objects or the result of a grouping between objects. *Second*, learning the parameters of a model by searching for a parameter combination which

2 Compare section 5.5.1 on page 65 on the basis of successive alternations of affine transformations and non-linearities yields a different distance between model-inferred predictions and expected outcomes. Murphy mentions the purposes of “determining if samples come from the same distribution” (**two-sample test**) and “how well a model approximates the data” (**goodness-of-fit**) for introducing statistical divergences thoroughly [200, Chp. 2.7].

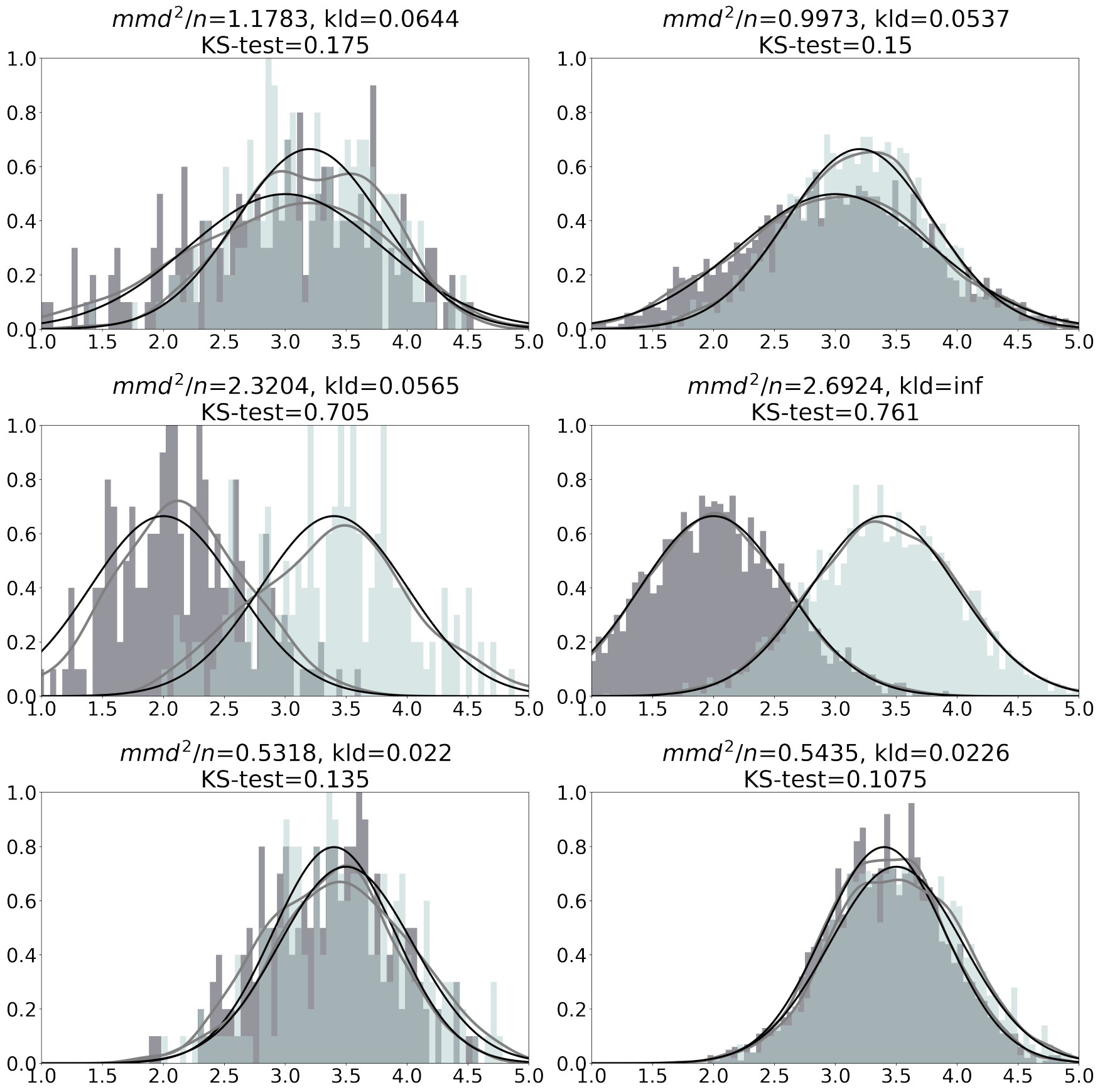

**Figure 5.2:** Distances between samples from gaussian distributions. The left column uses 200 samples, the right column 2000 samples per population. The first row shows $(\mu_{11} = 3, \sigma_{11} = 0.8)$, $(\mu_{12} = 3.2, \sigma_{12} = 0.6)$, the second row $(\mu_{21} = 2, \sigma_{21} = 0.6)$, $(\mu_{22} = 3.4, \sigma_{22} = 0.6)$ and the third row $(\mu_{31} = 3.4, \sigma_{31} = 0.5)$, $(\mu_{32} = 3.5, \sigma_{32} = 0.55)$. The gray solid lines show a kernel density estimation and the black line shows the used probability density.

The empirical maximum mean discrepancy and Kullback-Leibler divergence (KLD) are shown as exemplary distances in Figure 5.2. Each example contains two population samples drawn from gaussian distributions with specified expected value $\mu$ and standard deviation $\sigma$. All $p$-values of the provided Kolmogorov-Smirnov test statistic are below 0.03. Observe how the KL-divergence (KLD) does barely change for a shifted mean between row one and row two. The kernel density estimation is based on the population samples with a gaussian kernel.

### 5.3.1 *Metrics Induced by a Norm*

A norm (intuitively a length) is a function $\|\|: X \to [0, \infty)$ on a vector space $X$ over $\mathbb{R}$ if $\|x\| = 0 \Leftrightarrow x = 0$, $\|\alpha x\| = |\alpha|\|x\|$ for $\alpha \in \mathbb{R}, x \in X$ and $\|x + y\| \leq \|x\| + \|y\|$ [122, p. 9]. We are using the notation of a $l_p$-norm as $\|x\|_p \triangleq (\sum_{j=1}^{n} |x_j|^p)^{1/p}$ for vectors $x = (x_1, \ldots, x_n) \in \mathbb{R}^n$. Metrics are functions $d : M \times M \to \mathbb{R}$ on a set $M$ for which $\forall x, y, z \in M$ holds that $d(x, x) = 0$, $d(x, y) > 0$ for $x \neq y$, $d(x, y) = d(y, x)$ and $d(x, z) \leq d(x, y) + d(y, z)$ [122, p. 9]. The classical manhattan metric with $p = 1$, euclidean metric with $p = 2$ and maximum metric with $p = \infty$ are induced by this norm:

$$d(x, y) \triangleq \|x - y\|_p \tag{5.1}$$

### 5.3.2 *Employed f-divergences*

Polyanskiy & Wu [227] provide an insightful book grounding notation and theory on information and its measures. Also compare Murphy in [200, Chp. 2.7]. We are particularly interested in practical divergences and especially their empirical estimates but provide a quick look into their formal definition to recap some of their relationships. The divergences can be e.g. used as optimisation objectives or as a notion of how well we could reach a target distribution, e.g. how well or close a generative model could sample objects.

Divergences have information-theoretic origins based on the (Shannon) entropy $H$ of a discrete random variable $X$:

$$\begin{aligned} H(X) &= -\mathbb{E}\left[\log P_X(X)\right] = \mathbb{E}\left[\log 1 - \log P_X(X)\right] \\ &= \mathbb{E}\left[\log \frac{1}{P_X(X)}\right] = \sum_{x \in \mathcal{X}} P_X(x) \log \frac{1}{P_X(x)} \end{aligned} \tag{5.2}$$

For continuous cases, this notion of entropy needs to be generalised and Radon-Nikodym derivatives $\frac{dP}{DQ}$ are introduced [227, Chp. 2]. For probability distributions $P$ and $Q$ on a measurable space $(\Omega, \mathcal{F})$ we obtain $f$-divergences as $D_f(P \parallel Q) \triangleq \mathbb{E}_Q\left[f\left(\frac{dP}{dQ}\right)\right]$ [227, Chp. 7.1] with $f : (0, \infty) \to \mathbb{R}$ a convex function with $f(1) = 0$. For the discrete case this can be written as $D_f(P \parallel Q) = \sum_x Q(x) f\left(\frac{P(x)}{Q(x)}\right)$. $f$-divergences are often also denoted as $\phi$-divergences and called Ali-Silvey distance or Csiszár's $\phi$-divergence [266].

KULLBACK-LEIBLER DIVERGENCE The Kullback-Leibler divergence is a widely used statistical $f$-divergence [227, Chp. 7.1] and often referred to as KLD, originally introduced in [141]. The Kullback-Leibler divergence measures information loss or gain between two distributions $P$ and $Q$ of which $Q$ usually denotes the reference distribution. KLD is defined as $D_{KL}(P \parallel Q) \triangleq \mathbb{E}_{x \sim P(x)}\left[\log \frac{P(x)}{Q(x)}\right] = \int_{-\infty}^{\infty} P(x) \log \frac{P(x)}{Q(x)} dx$ which can be also written as $\mathbb{E}_{x\sim P(x)}\left[\log P(x) - \log Q(x)\right]$ and shows where alternative names such as *relative entropy* (in the sense of proportions) or *loss* (in the sense of differences) come from. We empirically estimate the divergence through

$$D_{KL}(X,Y) = \sum_{i=1}^{n} P(x_i) \log \frac{P(x_i)}{Q(x_i)} \tag{5.3}$$

JENSEN-SHANNON DIVERGENCE The Jensen-Shannon divergence can be understood as a symmetric version of the Kullback-Leibler divergence with $D_{JS}(P \parallel Q) \triangleq \frac{1}{2} D_{KL}(P \parallel M) + \frac{1}{2}(Q \parallel M)$ with $M = \frac{1}{2}(P+Q)$. Although, according to Cai et al. this understanding "hides an important distinction, namely, the Jensen-Shannon divergence may be defined on probability measures without densities" [31]. The convex function $f$ for the Jensen-Shannon divergence is $f(x) = x \log \frac{2x}{x+1} + \log \frac{2}{x+1}$. We use the empirical estimate

$$D_{JS}(X,Y) = \frac{1}{2} D_{KL}(P \parallel M) + \frac{1}{2}(Q \parallel M) \tag{5.4}$$

in practical applications.

TOTAL VARIATION DIVERGENCE For sake of completeness, we also mention the total variation divergence, defined as $D_{TV}(P \parallel Q) \triangleq \sup_{A \in \Sigma(\mathbb{R}^n)} |P(A) - Q(A)| = \inf_{\gamma \in \Gamma(P,Q)} \int_{\Omega \times \Omega} \mathbb{I}_{x \neq y}(x,y) d\gamma(x,y)$ [31, p. 8, Sec. V]. The total variation divergence is "a metric on the space of probability distributions" [227, Sec. 7.1, p. 89].

### 5.3.3 *Employed Integral Probability Metrics*

Integral probability metrics "are essentially different from $f$-divergences" [266], e.g. Sriperumbudur et al. argue that "estimators for IPMs are quite simple to implement and are not affected by the dimensionality of the data, unlike $f$-divergences". $\gamma_{\mathcal{F}}(P \parallel Q) \triangleq \sup_{f \in \mathcal{F}} \left| \int_\Omega f(dP) - \int_\Omega f(dQ) \right|$

WASSERSTEIN METRIC With a measurable space $\Omega$ and $p \in \mathbb{N}$ [31] the Wasserstein $p$-distance is $W_p(P \parallel Q) \triangleq \left[ \inf_{\gamma \in \Gamma(P,Q)} \int_{\Omega \times \Omega} \mathbb{E}_{(x,y)\sim\gamma(P,Q)} \left[ \|x-y\|_p \right] \right]^{1/p}$ for couplings $\gamma(P,Q)$[3]. However, the Wasserstein metric "has a definite drawback in that its explicit calculation is difficult" [77]. Murphy & Cuturi mention the Wassertstein distance in [200, Sec. 6.8.2.4] in a chapter on *optimal transport* "to compare two probability distributions" [200, Sec. 6.8].

3 The definition of couplings is given in e.g. [281, Chp. 4]. Roch describes the idea of couplings as "to compare two probability measures $P$ and $Q$" based on the construction of "a joint probability space with marginals $P$ and $Q$".

MAXIMUM MEAN DISCREPANCY abbreviated with MMD, is defined as $MMD(P \parallel Q; \mathcal{F}) \triangleq \sup_{f \in \mathcal{F}} (E_{x \sim P}(f(x)) - E_{y \sim Q}(f(y)))$ for random variables $x, y$ with observations $X \sim P$ and $Y \sim Q$, each independently and identically distributed, based on a metric space $(\mathcal{X}, d)$ and a suitable function class $\mathcal{F}$[4] [82]. For practical purposes, we will refer with MMD to $MMD^2$, which under suitable assumptions resolves to $\sup_{f \in \mathcal{F}} (E_{x \sim P}(f(x)) - E_{y \sim Q}(f(y))) = \|\mu_P - \mu_Q\|^2_{\mathcal{H}}$ and further becomes $\langle \mu_P - \mu_Q, \mu_P - \mu_Q \rangle = \langle \mu_P, \mu_P \rangle - 2\langle \mu_P, \mu_Q \rangle + \langle \mu_Q, \mu_Q \rangle$ which then leads to our used (biased) empirical estimate for the MMD

$$\begin{aligned} MMD^2(X, Y; \mathcal{F}) &= \frac{1}{m(m-1)} \sum_{i=1}^{m} \sum_{j \neq i}^{m} k(x_i, x_j) \\ &+ \frac{1}{n(n-1)} \sum_{i=1}^{n} \sum_{j \neq i}^{n} k(y_i, y_j) - \frac{2}{mn} \sum_{i=1}^{m} \sum_{j=1}^{n} k(x_i, y_i) \end{aligned} \tag{5.5}$$

involving the use of the kernel trick as in [200, Sec. 2.7.3.2].

### 5.3.4 *Summary on Distances*

Distances are required for comparing vectors, tensors or even distributions and are the basis for learning by minimizing them (or conversely maximizing similarity). You should be familiar with these terms and notation:

REMARKS

- The $l_p$-norm $\|x\|_p$ and the metric $\|x - y\|_p$ for $x, y \in M$ with $M$ being the used metric space,
- the empirical Kullback-Leibler divergence $D_{KL}(X, Y) = \sum_{i=1}^{n} P(x_i) \log \frac{P(x_i)}{Q(x_i)}$,
- and the empirical maximum mean discrepancy $MMD^2(X, Y; \mathcal{F})$.

## 5.4 STATISTICAL LEARNING WITH MAXIMUM LIKELIHOOD

With the prerequisite on data and distances, we'll now consider the guiding principle for statistical learning used in this work: maximum likelihood estimation (MLE). Polyanskiy & Wu list the principle of MLE as "one of the four fundamental optimisation problems in information theory" [227, Chp. 5, p. 64] and mention the estimator underlying this principle in [227, Chp. 29] on asymptotic behaviour under large samples.

4 $\mathcal{F}$ is proposed to be the "unit ball in a reproducing kernel hilbert space $\mathcal{H}$" and further detailled notation is given in [82, p.725ff].

For a sample $X = (x_1, x_2, \dots, x_n) \overset{i.i.d.}{\sim} P_{\theta^*}$ with $\theta^*$ being the true underlying parameter of the desired distribution $P_{\theta^*}$, the maximum likelihood estimator is:

$$\hat{\theta}_{MLE} \in \arg\max_{\theta \in \Theta} L_\theta(X) \tag{5.6}$$

with

$$L_\theta(X) = \sum_{i=1}^{n} \log p_\theta(x_i)$$

being the total log-likelihood in which "$p_\theta(x) = \frac{dP_\theta}{d\mu}(x)$ is the density of $P_\theta$ w.r.t. some measure $\mu$" as detailled in [227, 29.3, p. 502].

Observe how this is derived from Bayes' theorem

$$Pr(\theta \mid X) = \frac{Pr(X \mid \theta) Pr(\theta)}{Pr(X)} \propto Pr(X \mid \theta) Pr(\theta)$$

for which the posterior $Pr(\theta \mid X)$ as a function of $\theta$ is proportional to $Pr(X \mid \theta) Pr(\theta)$ if data $X$ is kept fix such that $Pr(X)$ is constant.

$$\begin{aligned} \hat{\theta}_{MLE} &\in \arg\max_{\theta \in \Theta} Pr(X \mid \theta) \\ &\propto \arg\max_{\theta \in \Theta} \log Pr(X \mid \theta) \\ &= \arg\max_{\theta \in \Theta} \log \prod_{i=1}^{n} Pr(x_i \mid \theta) \\ &= \arg\max_{\theta \in \Theta} \sum_{i=1}^{n} \log Pr(x_i \mid \theta) \end{aligned} \tag{5.7}$$

The maximum likelihood estimator seeks to maximise the likelihood function in order to find an optimal parameter setting $\hat{\theta}$ from the possible hypothesis space $\Theta$. This problem is equivalent to finding the minimum of the log-likelihood $L_\theta$. In the following we are interested in hypothesis spaces built from neural networks. Training a neural network with the principle of MLE means then to find the best parameter combination in which the log-likelihood is minimised. The parameter space $\Theta$ of a neural network will consist of e.g. its weights and biases and even further its activation functions and how these parameters are composed, i.e. structured. Maximum likelihood estimation is therefore an optimisation problem and actually minimises the Kullback-Leibler divergence.

RELATING MLE AND KLD The importance of statistical distances in learning from data gets clear when looking at the relationship between MLE and minimizing the KLD. Using the words of Arjovsky et al. [7] "asymptotically, maximum-likelihood estimation amounts to minimizing the Kullback-Leibler divergence". The KLD between the desired distribution $Pr(X \mid \theta^*)$ and a guess for the parameter $\theta$ can be expressed as $D_{KL}\left[Pr(X \mid \theta^*) \parallel Pr(X \mid \theta)\right]$ such that the estimator becomes

$$\begin{aligned} \hat{\theta}_{\text{KL}} &\in \underset{\theta \in \Theta}{\arg\min}\, D_{KL}\left[Pr(X \mid \theta^*) \parallel Pr(X \mid \theta)\right] \\ &= \underset{\theta \in \Theta}{\arg\min}\, \mathbb{E}_{X \sim P_{\theta^*}}\left[\log \frac{Pr(X \mid \theta^*)}{Pr(X \mid \theta)}\right] \\ &= \underset{\theta \in \Theta}{\arg\min}\, \mathbb{E}_{X \sim P_{\theta^*}}\left[\log Pr(X \mid \theta^*) - \log Pr(X \mid \theta)\right] \end{aligned} \tag{5.8}$$

Note, that this becomes equivalent to finding the minimum of the right term of the difference in Equation 5.8 as $\log Pr(X \mid \theta^*)$ can be again treated as constant such that we obtain

$$\begin{aligned} \hat{\theta}_{\text{KL}} &\in \underset{\theta \in \Theta}{\arg\min}\, \mathbb{E}_{x \sim P_{\theta^*}}\left[-\log Pr(X \mid \theta)\right] \\ &= \underset{\theta \in \Theta}{\arg\max}\, \mathbb{E}_{x \sim P_{\theta^*}}\left[\log Pr(X \mid \theta)\right] \\ &\approx \underset{\theta \in \Theta}{\arg\max} \sum_{i=1}^{n} \log Pr(x_i \mid \theta) \end{aligned}$$

such that the KLD-estimator is equivalent, if not the same, as the maximum likelihood estimator.

RELATING MLE AND MAP The estimator in Equation 5.7 has actually only considered to maximise the likelihood term $Pr(X \mid \theta)$. Taking the prior probability $Pr(\theta)$ of the parameters into perspective, a maximum a-posteriori estimation can be formulated:

$$\begin{aligned} \hat{\theta}_{MAP} &\in \underset{\theta \in \Theta}{\arg\max}\, Pr(X \mid \theta) Pr(\theta) \\ &= \underset{\theta \in \Theta}{\arg\max}\left[\log Pr(X \mid \theta) + \log Pr(\theta)\right] \\ &= \underset{\theta \in \Theta}{\arg\max}\left[\log \prod_{i=1}^{n} Pr(x_i \mid \theta) + \log Pr(\theta)\right] \\ &= \underset{\theta \in \Theta}{\arg\max}\left[\sum_{i=1}^{n} \log Pr(x_i \mid \theta) + \log Pr(\theta)\right] \end{aligned} \tag{5.9}$$

and we can observe that this introduces an additive term $\log Pr(\theta)$ which needs to be maximised. If all parameters $\theta \in \Theta$ are equally likely, e.g. $Pr(\theta)$ behaves like a uniform distribution, the term is constant and can be neglected. Maximum likelihood estimation is therefore a special case of maximum a-posteriori estimation with uniform prior probability over the parameter space. That is insofar interesting for this work in that certain parameterisations of neural networks can be favored over others. In fact, that would be those neural networks with a favorable structure. But this requires a non-uniform distribution over the parameter space from which proper sampling needs to be possible. That can be e.g. achieved through known properties through analysis of common structures or through a learned parameter distribution which we will tackle with generative models for graphs.

There are two popular examples for the prior distribution in maximum a-posteriori estimation: a gaussian (normal) prior and a laplacian prior. Both give an intuition of how the hypothesis space can be influenced during learning by imposing a regularisation constraint.

The first example of imposing a prior is done by assuming a gaussian (normal) prior distribution. For the normal distribution $\mathcal{N}(\mu, \sigma^2)$ with mean $\mu$ and variance $\sigma^2$ on $\mathbb{R}$ or its multi-variate companion $\mathcal{N}(\mu, \Sigma)$ on $\mathbb{R}^d$, the density is given as $\rho(x) = \frac{1}{\sigma\sqrt{2\pi}} e^{-\frac{1}{2}\left(\frac{x-\mu}{\sigma}\right)^2}$. The second term of maximum a-posteriori estimation in eq. (5.9) on the facing page becomes

$$Pr(\theta) = \prod_{i=0}^{|\theta|} \frac{1}{\sigma\sqrt{2\pi}} e^{-\frac{1}{2}\left(\frac{\theta_i-\mu}{\sigma}\right)^2}$$

and with $\mu = 0$ we can rewrite it into

$$Pr(\theta) \approx -\sum_{i=0}^{|\theta|} \frac{\theta_i^2}{2\sigma^2}$$

such that during maximum a-posteriori estimation the second term can be turned into

$$\begin{aligned} \hat{\theta}_{MAP} &= \underset{\theta\in\Theta}{\arg\max} \left[ \sum_{i=1}^{n} \log Pr(x_i \mid \theta) + \log Pr(\theta) \right] \\ &= \underset{\theta\in\Theta}{\arg\min} \Big[ -\sum_{i=1}^{n} \log Pr(x_i \mid \theta) + \underbrace{\lambda \sum_{i=0}^{|\theta|} \theta_i^2}_{l_2-\text{regularisation term}} \Big] \end{aligned}$$

which means, that the gaussian prior for $Pr(\theta)$ can be incorporated by applying $l_2$-regularisation on the model weights. For the variance-dependent scaling term, a regularisation rate $\lambda$ is introduced. Murphy notes in that context, that "the value of $\tau$" (here $\sigma$, i.e. the variance) "plays a role similar to the strength of an $l_2$-regularisation term in ridge regression" [200, Sec. 3.4.6.3, p. 100].

The second example of using a Laplace prior probability is related in effectively imposing a $l_1$-regularisation on the model weights. Hastie notes that "the lasso estimate is the Bayesian MAP (maximum aposteriori) estimator using a Laplacian prior, as opposed to the mean of the posterior distribution, which is not sparse" [93]. The density of the Laplace distribution is given as Laplace$(\mu, b) \triangleq \frac{1}{2b} e^{-\frac{|x-\mu|}{b}}$ and if we plug it in the maximum a-posteriori estimation for $P(\theta)$ we obtain

$$P(\theta) = \prod_{i=0}^{|\theta|} \frac{1}{2b} e^{-\frac{|\theta_i-\mu}{b}}$$

and again for a prior centered around $\mu = 0$ the optimisation problem of maximum a-posteriori in eq. (5.9) on page 62 becomes

$$\begin{aligned}\hat{\theta}_{MAP} &= \arg\max_{\theta\in\Theta}\left[\sum_{i=1}^{n}\log Pr(x_i \mid \theta) + \log Pr(\theta)\right]\\ &= \arg\max_{\theta\in\Theta}\left[\sum_{i=1}^{n}\log Pr(x_i \mid \theta) - \log\prod_{i=0}^{|\theta|}\frac{1}{2b}e^{-\frac{|\theta_i|}{b}}\right]\\ &= \arg\min_{\theta\in\Theta}\Big[-\sum_{i=1}^{n}\log Pr(x_i \mid \theta) + \underbrace{\lambda\sum_{i=0}^{|\theta|}|\theta_i|}_{l_1-\text{regularisation term}}\Big]\end{aligned}$$

for which we can see, that for $\lambda = \frac{1}{b}$ another reglarisation scale term[5] is used.

Like Murphy, we interprete "the field of neural architecture search (NAS) as a form of structural prior learning" [200, Sec. 17.2.5]. An alternative approach to the bayesian perspective of regularizing the structure of the model comes in form of a higher-level optimisation as we will define NAS in section 7.1 on page 122.

## 5.5 BASIC COMPONENTS OF DEEP NEURAL NETWORKS

Based on the underlying principles of statistical learning, we now introduce elements of deep neural networks to lead towards a formulation of graph-induced neural networks.

Deep neural network are combined with a probabilistic learning perspective in two ways:

1. as non-linear functions "used inside conditional distributions" [200, Chp. 16.1] such that a classifier $Pr(Y \mid X, \theta) = Cat(Y \mid \sigma^{\Psi}(f(X;\theta)$ uses a neural network $f(X;\theta)$ that transforms data $X$ and parameters $\theta$ into logits of a categorical distribution. Or, as a joint probability distribution in which each conditional probability distribution $Pr(X_i \mid Par(X_i))$ is a deep neural network.

2. a deep neural network can be used to "approximate the posterior distribution" [200, Chp. 16.1] such that $f$ is "an inference network with parameters $\theta$" that computes $q(\mathbf{z} \mid f(\mathcal{D};\theta))$.

This work mostly considers the first case and makes use of the second only in methodological situations to approach structures of neural networks. We further restrict our attention solely to predictive models in which an output $\mathbf{y}$ is predicted based on an input $x$ by means of a function $f$ that is estimated from data and can be deterministic or stochastic. Murphy considers neural networks as "parametric models", that "can overfit when N is small, and can underfit when N is large, due to their fixed capacity" [200, Chp. 18.1].

5 The regularisation term $\lambda$ can become dependent on other terms of the model if the likelihood $Pr(X \mid \theta)$ is approximated.

### 5.5.1 *Neurons, Weights and Biases*

A classical single-layer neural network in its original functional form can be defined as $f : \mathbb{R}^{d_1} \to \mathbb{R}^{d_2}$ with $f(\mathbf{x}) \triangleq \mathbf{x} \mapsto W_2\sigma(W_1\mathbf{x} + B_1) + B_2$ with $h \in \mathbb{N}$ being the size of the hidden layer, $W_1 \in \mathbb{R}^{d_1 \times h}, W_2 \in \mathbb{R}^{h \times d_2}, B_1 \in \mathbb{R}^h, B_2 \in \mathbb{R}^{d_2}$ being the weights and biases[6] and $d_1, d_2 \in \mathbb{N}$ being the input and output dimensions, respectively. This functional form will be generalised in Part iii by a graph induced structure.

The parameters are collectively referred to with $\theta = \{W_1, W_2, B_1, B_2\}$ or in a parameterised notation with $\theta(f)$ for a neural network $f$. Parameters are subject to learning through an optimisation procedure of minimizing the loss in context of MLE. Multi-layer neural networks (also called multi-layer perceptrons) are then a generalisation of this form in that it has as many weight matrices as to express multiple consecutive layers, separated with $\sigma$, the activation function, also called non-linearity or simply *activation*. With proper context, the resulting feed-forward computation after passing through an activation function can also be called activation, i.e. the resulting vector.

The underlying building blocks of deep neural networks are therefore neurons and they pass aggregated and squished (activated) information through connections to other neurons. An examplary illustration can be seen in Figure 5.3 and bears some relation to BNNs in that dendrites and synapses are represented with connections and axons with aggregations and non-linear activation functions.

Bias information can be considered as part of the nucleus, carrying inherent information of the neuron. Often, it is depicted as an own vertex carrying the bias value and being connected to the neuron without having own incoming connections (compare boxed rectangle in gray in Figure 5.3). This depiction of the bias of a neuron as separate incoming value from another vertex is also rooted in suitable mathematical formulations in which the bias term can be included into the weights of incoming connections by an additional column of constant ones in the input $\mathbf{x}$ before multiplicating the weight matrix $W$.

### 5.5.2 *Layers*

Successive applications of transformations result in $L \in \mathbb{N}$ neural network layers – until the desired output dimension $d_2$ is reached. Neurons of one layer receive the output of neurons from preceeding layers as input. The input to neurons of the first layer is the input $\mathbf{x} \in \mathbb{R}^{d_1}$ to the overall neural network. Sometimes elements of the input $\mathbf{x}$ are therefore also depicted as neurons, connected to the first hidden layer. The term

6 In machine learning and deep learning, the coefficients and intercepts are usually referred to as weights and biases. For one layer, all parameters can also be written in one matrix form by moving the bias into an additional column of the weight matrix of that layer.

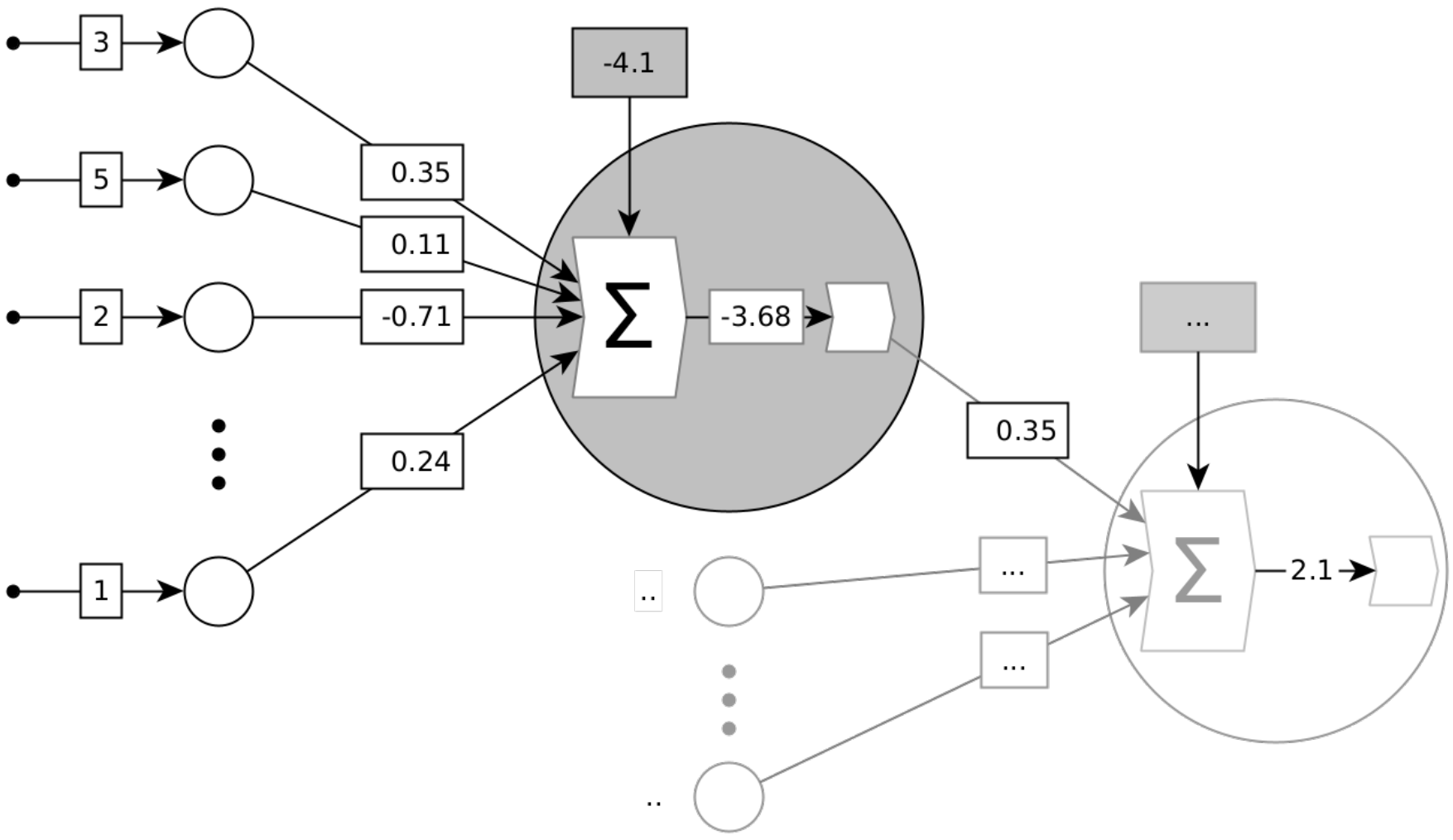

**Figure 5.3:** An illustration of an isolated artifcial neuron (center gray). Circles on the left depict inputs with values 3, 5, 2, .., and 1. Weights in white rectangles along incoming edges scale the incoming signal values before passing into the neuron. The $\sum$-symbol depicts an aggregation step to which a bias term is added (here gray rectangle with value -4.1). An intermediate value (here -3.68) is fed through an activation (white broad arrow) before passing on to the next dendritic connection. Compare the dendritic structures of biological neural network (BNN)s as in Figure 1.2 from which the original artificial neural network models are inspired.

*hidden* refers to the fact, that available data does not provide immediate information about how parameters in that layer distribute.

With multiple layers $l \in \{1, \dots, L\}$ and the initial input $\mathbf{z}^{(0)} = \mathbf{x}, \mathbf{h}^{(0)} = \mathbf{z}^{(0)}$, information is passed through these layers via $\mathbf{z}^{(l)} \mapsto W^{(l)}\mathbf{h}^{l-1} + B^{(l)}$ with $\mathbf{h}^{(l)} = \sigma(\mathbf{z}^{(l)})$. The layer-wise parameters then become $\theta^l(f) = \{W^{(l)}, B^{(l)}\}, \theta^0 = \{W^0, B^0\}$. This a common recursive definition of (multi-layered) feed-forward deep neural networks, e.g. compare [79] or [200, Sec. 16.3.1].

Vector representations $\mathbf{h}^l$ are usually referred to as hidden vectors and the learned $H_l$-dimensional manifold $\mathbb{R}^{H_l}$ which they rely in is often called *latent space*. The sub-module (or function composition) $f^{(d_1)\to(l)} = f^{(l-1)\to(l)} \circ \dots \circ f^{(1)\to(2)} \circ f^{(d_1)\to(1)}$ which returns $z^{(l)}$ for an input $\mathbf{x}$ is then an embedding of $\mathbf{x}$ into this latent space. The sub-module $f^{(l-1)\to(l)}$ for an input $\mathbf{z} \in \mathbb{R}^{H_{l-1}}$ is also sometimes used separately as an embedding of a $H_{l-1}$-dimensional into a $H_l$-dimensional manifold. The modularity or decomposability of the overall network model $f$ illustrates already the various possibilities for taking control on different *architectural* or *structural* aspects of $f$.

Disregarding all further details, functional forms (or architectures) of multi-layer neural networks can be quickly denoted as ordered tuples $(H_0, H_1, \dots, H_L)$ with dimensionalities $H_l \in \mathbb{N}$ for layers $l \in \{1, \dots, L\}$ and a network depth of $L \in \mathbb{N}$. Jointly with input- and output di-mensionalities this tuple representation becomes $(d_1, H_0, H_1, \ldots, H_l, d_2)$. Also compare the matrix-vector tuple representations for neural networks of Petersen [223]. We used this ordered tuple representation in LIB DeepStruct to quickly construct multi-layerered neural networks as outlined in technical details in appendix A.5.

### 5.5.3 *Activation Functions*

The point-wise function $\sigma$ is commonly referred to as activation function and was originally motivated by the Heaviside[7] step function $\sigma_{step}$ : $\mathbb{R} \to \mathbb{R}$ with

$$\sigma_{step}(x) \triangleq x \mapsto \begin{cases} 1, & x > 0 \\ 0, & x \leq 0 \end{cases}$$

. Idea of the activation is to be able to express non-linear relations between input and output of a neural network while at the same time restricting the overal non-linearity of the model locally. Its origins are based in perceptron theory and neuroscience to resemble a biological neuron that builds up an action potential through signals from incoming dendrites until it reaches a threshold and the biological neuron fires (a signal through subsequent dendrites). Heaviside's step function $\sigma_{step}$ resembles this idea by a threshold located at $x = 0$.

The developments on alternative activation functions were crucial for nowadays deep learning practices. Early generalisations have been sigmoid- or S-shaped curves which shared the property of being two-sided bounded on [0,1] – representing a binary signal and also having a threshold around which the output changes from 0 to 1. However, the differentiability of e.g. $\sigma^{\sim} \triangleq x \mapsto \frac{1}{1+e^{-x}} = 1 - \sigma^{\sim}(-x)$ made it analytically more plausible to work with gradients and gradient-based updates during learning (more on that in the subsequent section 5.6). Further, the phase transition of $\sigma^{\sim}$ between 0 to 1 was more smooth and allowed to express uncertainty. The hyperbolic tangent $\sigma^{tanh} \triangleq x \mapsto \frac{sinh(x)}{cosh(x)} = \frac{e^x - e^{-x}}{e^x + e^{-x}}$ also falls into this category and was long considered an alternative to $\sigma^{\sim}$ with reasons lying in its co-domain being [-1,1] instead of [0,1].

The properties of activation functions have various technical and theoretical implications. One is, that locally almost linear activations such as the rectified linear unit (ReLU) $\sigma^{\diagup} : x \mapsto max(0, x)$ have almost everywhere constant derivatives such that back-propagating errors have more chance of not vanishing or exploding over deep architectures. This line of analytical research on accumulating errors goes back to a diploma of Sepp Hochreiter [101] and resulted for example in the development of Long Short-Term Memory Networks [102] and the usage of ReLUs.

7 The step function is named after Oliver Heaviside (1850-1925), known for first using it in electromagnet theory.

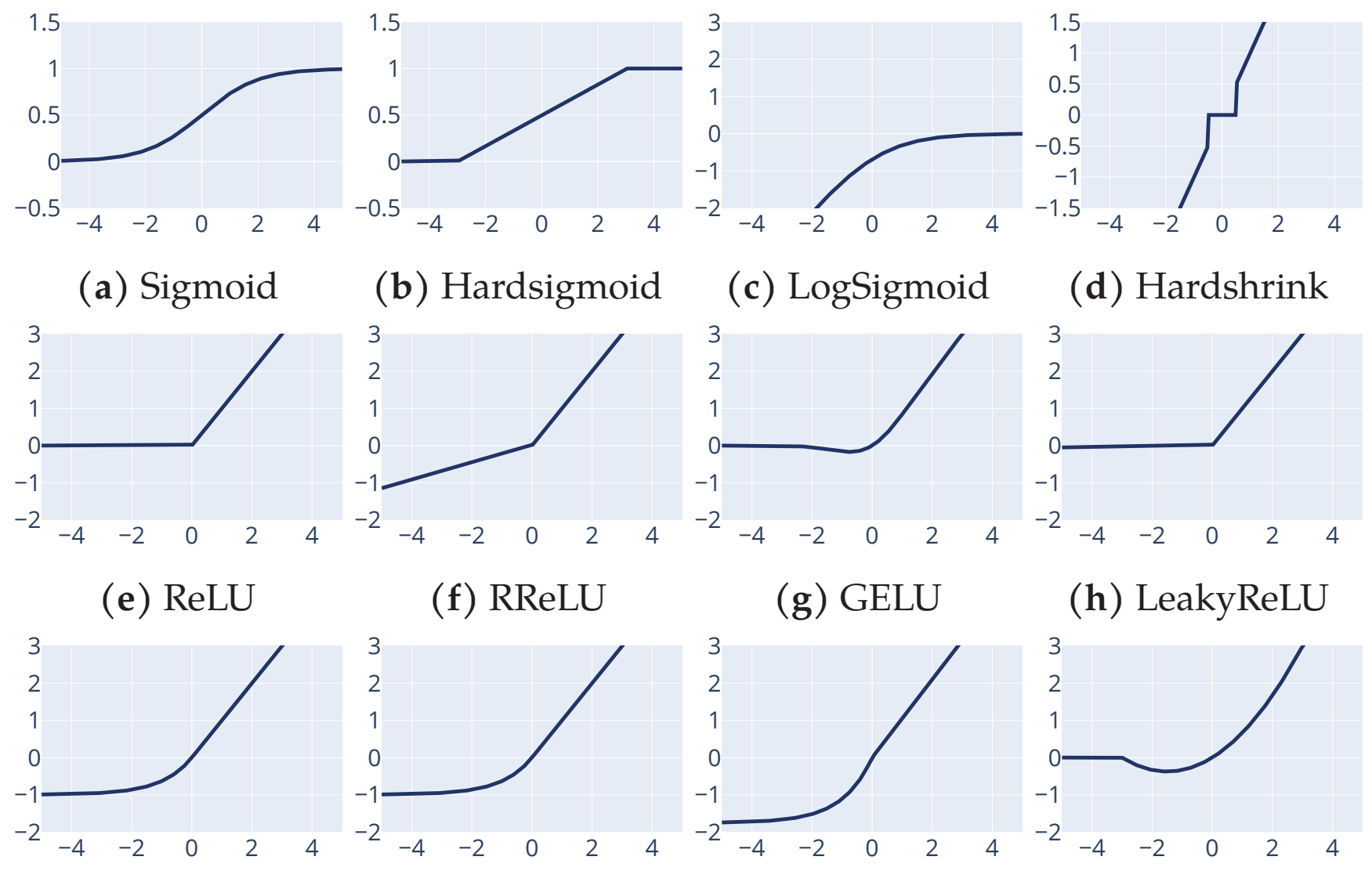

(a) Sigmoid (b) Hardsigmoid (c) LogSigmoid (d) Hardshrink
(e) ReLU (f) RReLU (g) GELU (h) LeakyReLU
(i) CELU (j) ELU (k) SELU (l) Hardswish

**Figure 5.4:** Many activation functions $\sigma$ have been proposed, systemised and compared empirically within and across various application domains [57]. Dubey et al. list properties such as parametricity, monotonicity, smoothness and boundedness for comparison and categorise them into logistic and TanH-based, rectified-based, exponential, learned- or adaptive, and other (miscellaneous) activation functions.

Further activation functions are Hardshrink

$$\sigma_{Hardshrink,\lambda} \triangleq x \mapsto \begin{cases} x & \text{if } x > \lambda, \\ x & \text{if } x < -\lambda, \\ 0 & \text{otherwise} \end{cases}$$

, and Hardswish [108, Sec. 5.2]

$$\sigma_{Hardswish,\lambda} \triangleq x \mapsto \begin{cases} 0 & \text{if } x \leq -3, \\ x & \text{if } x \geq 3, \\ x\frac{(x+3)}{6} & \text{otherwise} \end{cases}$$

, a leaky version of ReLU called LeakyReLU as

$$\sigma_{LeakyReLU,\lambda} \triangleq x \mapsto \begin{cases} x & \text{if } x \geq 0, \\ \beta \cdot x & \text{otherwise} \end{cases}$$

with $\beta \in \mathbb{R}, \beta < 0$ and default value $\beta = 0.01$, and a "Gaussian error linear unit" GeLU [97] with the continuous distribution function $\Phi(\mu, \sigma) \triangleq x \mapsto \frac{1}{\sigma\sqrt{2\pi}} \int_{-\infty}^{x} exp\left(-\frac{(t-\mu)^2}{2\sigma^2} dt\right)$ of the Gaussian distribution (see section 5.2.3 on page 55) such that $\sigma_{GeLU,\lambda} \triangleq x \mapsto x\Phi(x; \mu =$ $0, \sigma = 1)$, and a LogSigmoid with $\sigma_{LogSigmoid,\lambda} \triangleq x \mapsto log(\sigma^{\sim}(x)) = log\left(\frac{1}{1+e^{-x}}\right)$.

Practically, we have mostly been employing rectified linear unit over alternative activation functions because of its broad usage in deep learning applications and its major impact on overcoming issues with vanishing or exploding gradients. Occasionally, we investigated on the choice of activation function as a hyperparameter but there is a whole branch of research on this on its own, e.g. by Dubey et al. [57]. Some remarks of their work which we found valuable are that "three major problems of ReLU" are "under-utilisation of negative values, limited nonlinearity and unbounded output" while "exponential based" activation functions "focus on better utilisation of negative values and" might "avoid saturation for important features" and adaptive activations try "to find the better base function and number of trainable parameters". Importantly, "both negative & positive values should be used to ensure near zero mean" [57]. We note at this point, that this zero mean can also be accomplished by employing batch- or layerwise normalisation techniques (more on that later). Dubey et al. recommend that "logistic sigmoid and tanh-" activations "should be avoided for convolutional neural networks as it leads to poor convergence" but "this type of" activations "is commonly used as gates in recurrent neural networks". Their statement that "ReLU, LReLU, ELU, GELU, CELU, and PDELU functions are better for the networks having residual connections for image classification" confirms our employment of locally linear activation functions in subsequent experiments. We leave further and deeper analysis of the influence of this hyperparameter to other studies.

Another practical aspect are technical choices of activation functions that allow to turn a deep neural network into a classifier. This can be achieved by using the normalised exponential or *softmax* function $\sigma^{\Psi} : \mathbb{R}^d \to (0,1)^d$ with $x_i \mapsto \frac{e^{x_i}}{\sum_{j=1}^{d} e^{z_j}}$ and $i \in \{1, \dots, d\}$. The derivation of it in conjunction with a cross-entropy loss comes quite naturally from a probabilistic perspective in which a multi-class classification $P(Y = j \mid x)$ is approached with maximum likelihood estimation and the last output of the neural network is interpreted as the logarithm of the odds. In other words, feeding the last activated layer output of a deep neural network through a softmax activation $\sigma^{\Psi}$ results in class-wise estimated probabilities for a multi-class classification task.

Importantly from a theoretical perspective, activation functions have to satisfy certain properties to comply with universal approximation theorems, that we require later on for a formalisation of graph-induced neural networks to study their structural properties. This can be summarised into the requirement, that an activation function needs to be **one-sided bounded** and **differentiably** almost everywhere. Local boundedness allows successive layers of transformations to become non-linear models, and this distinguishes deep neural networks from linear models.

## 5.6 TRAINING DEEP NEURAL NETWORKS WITH BACKPROPAGATION FROM DATA

For a given neural network $f$ with parameters $\theta$, how can appropriate values for $\theta$ be obtained? Here, we are only interested in finding values of parameters $\theta$ itself, not how the parameters structurally relate. Lillicrap et al. note in that regard, that "it should be acknowledged that brains undoubtedly have prior knowledge optimised by evolution, e.g. in the form of neural architectures and default connectivity strengths" [162]. This means that there may be learnable information in how the parameters are structurally related, but here we're interested in training a structurally fixed neural network first.

So, how are deep neural networks trained? While the bayesian answer with MLE to the stated question suggests to minimise a loss function (loss) (such as the KLD) between a batch of predictions and expected outcomes naturally leads to the use of an optimisation algorithm that makes use of the differentiability of the neural network $f$, the general question of *credit assignment* was not only theoretically investigated but also inspired by the human brain [305].

The biological inspiration, theories on learning and unsolved questions about biological plausible learning led to various approaches of learning, usually including interchangable components that can be argued about. One of these components is the learning rule. Methods such as Hebbian learning [96] without feedback, perturbation learning with scalar feedback and many variants of vector-feedback such as backpropagation and backpropagation-like learning with feedback networks [162, Fig. 1] have been heavily and are actively investigated [305].

As an example, Hebbian learning summarises methods that follow the principle stated by the psychologist Hebb in 1949: "When an axon of cell A is near enough to excite a cell B and repeatedly or persistently takes part in firing it, some growth process or metabolic change takes place in one or both cells such that A's efficiency, as one of the cells firing B, is increased" [96, p. 62]. "When one cell repeatedly assists in firing another, the axon of the first cell develops synaptic knobs (or enlarges them if they already exist) in contact with the soma of the second cell." [96, p. 63]. The derived learning rule is often summarised as $\partial\theta = \eta f(x, \theta)x$ (compare e.g. [5]) in which the change of each parameter $\partial\theta$ is obtained by a learning rate $\eta \in \mathbb{R}^+$ and connection between network response $f(x, \theta)$ and its original input $x$.

As we use stochastic gradient descent (SGD) in our experiments and as SGD is employed in most successful applications of deep learning, we restrict our attention to the MLE framework with this principle as driving force for the learning process. To successively update parameters with gradient-based optimisation, partial derivatives with respect to different variables of a function are required. This is the reason why backpropagation (or the chain-rule of derivation) is required: gradients of the neural network $f$ are computed with automatic differentiation for composable parts of the network such that each parameter can then be updated with respect to gradients of an employed loss function (based on data).

In theory, any optimisation method can be used to improve a neural network during a learning process. Because of the availability of gradient information from a loss function based on data, it makes more sense to leverage the available information in gradient-based optimisation methods instead of e.g. using gradient-free methods such as evolutionary searches.

### 5.6.1 *Obtaining Gradients of Deep Neural Networks*

To employ gradient-based optimisation during the learning process of neural networks, partial derivatives w.r.t. each trainable parameter are required. Three major concepts might come into mind to obtain gradients of a function: finite differences, symbolic differentiation and automatic differentiation.

Finite differences approximate the gradient of a differentiable function by a difference quotient at an input vector with a small perturbation. In the words of Speelpenning "this method approximates a derivative by sampling the function in nearby points, computing the slope of a secant" [265]. For the one-sided approximation of the gradient it has the form $\frac{\partial f(x)}{\partial x_i} \approx \frac{f(x+he_i)-f(x)}{h}$ [263, Sec 6.3 p. 157], while the "two-sided approximations involve measurements of the form" $f(x \pm h)$ such that the quotient $\frac{f(x+he_i)-f(x-h)}{2h}$ is used for gradient approximation. The value $h$ is a small perturbation scalar (or vector). This method, however, requires $O(n)$ in complexity for $n$-dimensional gradients and includes numerical errors. Speelpenning calls the approach "crude and primitive" [265].

Symbolic differentiation has major benefits over finite differences such as not being affected by numerical instabilities. However, it requires the availability of a closed-form expression and has theoretical differences to automatic differentiation. Baydin et al. interestingly note, that "confusion can arise because automatic differentiation does in fact provide numerical values of derivatives and it does so by using symbolic rules of differentiation, giving it a two-sided nature that is partly symbolic and partly numerical" [22] Although James Spall notes in 2005 that "in software, automatic differentiation methods provide means for calculating gradients. However, these methods require extensive knowledge of the *inner workings* of the software" and "while automatic differentiation provides a systematic means for gradient calculation, the term *automatic* is somewhat of a misnomer" [263, Sec. 6.2, p. 153], with nowadays knowledge and proof of concepts like with PyTorch, the inner workings are well worked out to choose automatic differentiation over alternative approaches.

In fact, Sören Laue [145] gives a good overview on the relationship between symbolic differentiation and reverse mode automatic differentiation and claims that several myths on symbolic differentiation such as the *expression swell* do not apply as it is merely a theoretical consideration. This latter reverse mode automatic differentiation is algorithmically also described extensively by Roy Frostig in Murphy et al. [200, Alg. 6.4]. Applying reverse mode automatic differentiation on computational graphs as we have them with deep neural networks is "precisely the backpropagation algorithm, a term introduced in the 1980s" [200, Sec 6.2.2.2 p. 264].

In summary, backpropagation is the application of reverse mode automatic differentiation to obtain partial derivatives of differentiable programs based on their computational directed acyclic graph. Automatic differentiation is preferred over finite differences for numerical reasons and over symbolic differentiation for software engineering reasons [265]. For $f : \mathbb{R}^n \to \mathbb{R}^m$ with $n > m$ the reverse mode is computationally to prefer while with $m > n$ the forward mode is to prefer [22, p. 10], as the reverse mode has $O(m)$ in complexity and the forward mode $O(n)$ – the one is dependent on the output domain, the other on the input domain. The reverse mode is preferred in deep learning over the forward mode automatic differentiation as usually $n > m$ applies.

We employ PyTorch, which "supports reverse-mode automatic differentiation of vector-Jacobian products of functions with multiple outputs" [217], compare the algorithm 3 which is described in [200, Alg. 6.4].

### 5.6.2 *Backpropagation in Context of Automatic Differentiation*

Automatic differentiation is concerned with obtaining (partial) derivatives of differentiable functions. These functions are composed of basic operations of which symbolic or automatic derivatives are available such that the partial derivatives of the complex composed function can be automatically obtained.

Backpropagation is the reverse mode of automatic differentiation in which the derivative over an output perturbation w.r.t. to parameter variables in the computation graph are computed. The backpropagation algorithm consists of two passes: a forward pass and a backward pass. Lines 2-5 of algorithm 3 sketch the forward-pass in the notation used by Murphy et al. [200] and lines 7-11 sketch the backward pass.

The function to differentiate is not the neural network $f(\mathbf{x}) \triangleq x \mapsto W_2\sigma(W_1\mathbf{x} + B_1) + B_2$ alone but an error between the neural network

**Algorithm 3:** An algorithmic sketch of reverse-mode automatic differentiation resulting in the vector-Jacobian product (VJP) as provided by [200, p. 264, Alg. 6.4], also known as backpropagation.

```
input  : f : ℝ^n → ℝ^m composing f_1, …, f_T in topological order,
         where f_1, f_T are identity
input  : linearisation point x ∈ ℝ^n and perturbation u ∈ ℝ^m

x_1 ← x;
for t ← 2 to T do
    [q_1, …, q_r] ← Pa(t);
    x_t ← f_t(x_{q_1}, …, x_{q_r});
end
u_{(T−1)→T} ← u;
for t ← T − 1 to 2 do
    [q_1, …, q_r] ← Pa(t);
    u'_t ← ∑_{c∈Ch(t)} u_{t→c};
    u_{q_i→t} ← ∂_i f_t(x_{q_1}, …, x_{q_r})^T u'_t for i = 1, …, r;
end
output : x_T, equal to f(x)
output : u_{1→2}, equal to ∂f(x)^T u
```

prediction $\hat{\mathbf{y}} = f(\mathbf{x})$ and data $(\mathbf{x}, \mathbf{y})$ obtained from a loss (of an interchangable functional form). Such a loss $\mathcal{L}$ can be for example a *mean squared error* of the form $\frac{1}{n}\sum_{i=1}^{n}(y_i - \hat{y}_i)^2$ or a *binary cross-entropy* of the form $-\frac{1}{n}\sum_{i=1}^{n}(y_i \log(\hat{y}_i) + (1-y_i)\log(1-\hat{y}_i))$. We elaborate more on losses in Section 5.6.3. Here, we simplify the notation to an error in differentiable form based on the original network and use the symbol $E$ such that we are now interested in partial derivatives $\frac{\partial E_{\mathcal{L},f,\mathbf{y}}(\mathbf{x})}{\partial \theta_i}$ and for fixed network, data and loss have $\frac{\partial E(\mathbf{x})}{\partial \theta_i}$ in which $\theta_i$ is a single parameter of the network $f$.

The input to backpropagation as sketched in algorithm 3 is therefore not just the neural network $f$ but the error $E$, although this detail is usually not important as soon as the derivative of the loss is clarified. Automatic differentiation is only concerned with partial derivatives of parameters of the model and usually not w.r.t. data.

Because backpropagation – as the *reverse mode* of automatic differentiation – starts from the output side, we'll first have a look at the loss function and its derivative. Our example will make use of the mean squared error. Following Gareth et al. in [117, Eq. 10.28 p. 429], the mean squared error is a sum over data samples such that "its gradient is also a sum over $n$ observations" [117, p. 429] and only a single term can be followed. A single part of the considered squared error is then $(y_i - \hat{y}_i)^2$. This reduction already makes use of the *chain rule* of differentiation.

Backpropagation can be theoretically reduced to just the repeated application of the *chain rule*. We follow Euler's notation used by Murphy et al. [200, Sec. 6.2.1] for derivatives and for a function over two variables $f(x_1, x_2) = x_1^3 x_2$ we have partial derivatives $\frac{\partial f}{\partial x_1} = 3x_1^2 x_2$ and $\frac{\partial f}{\partial x_2} = x_1^3$. According to the chain rule, derivatives of chain compositions of linear maps are chain compositions of the derivatives of the individual maps. For $f_1 : \mathbb{R}^{d_{in}} \to \mathbb{R}^n$ over variables $(x_1, x_2, \dots, x_{d_{in}})$ and $f_2 : \mathbb{R}^n \to \mathbb{R}^{d_{out}}$ over variables $(a_1, a_2, \dots, a_{d_{out}})$, the partial derivatives of the chained function $g : \mathbb{R}^{d_{in}} \to \mathbb{R}^{d_{out}} \triangleq f_2 \circ f_1$ at point $p \in \mathbb{R}^{d_{in}}$ are given as $\frac{\partial g}{\partial x_i} = \sum_{k=1}^{n} \frac{\partial f_2}{\partial a_k}(f(p)) \frac{\partial f_k}{\partial x_i}$. The chain rule is used to break down a given composed function into elementary functions of which the derivatives are explicitly known. Autograd frameworks as contained in PyTorch carry a list of such elementary operations [217] and their derivatives.

A SIMPLIFIED EXAMPLE Consider the function in Figure 5.5, respresented with a directed acyclic graph[8], and given as $(sin(log(x)) - cos(log(x)))^2$ composed of basic operations $f_1, \dots, f_5$.

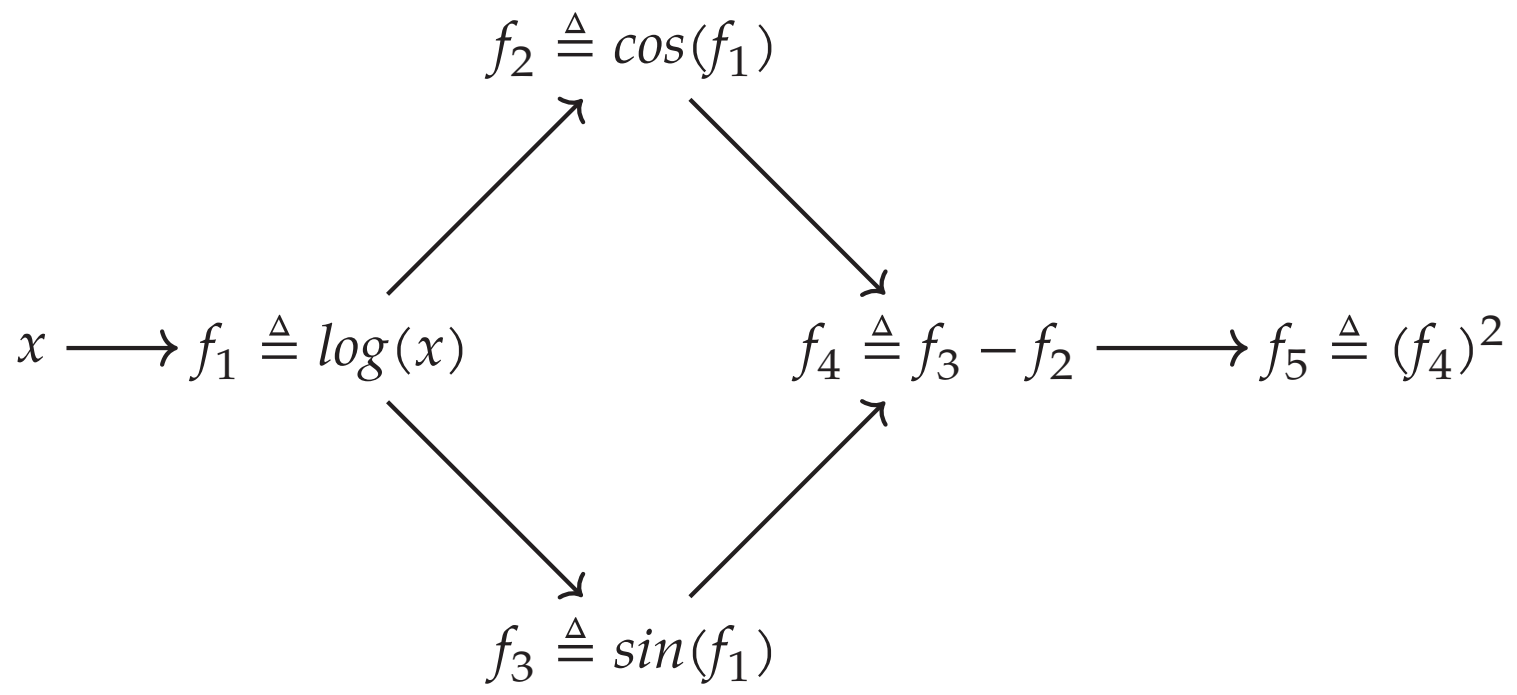

**Figure 5.5:** The function $(sin(log(x)) - cos(log(x)))^2$ composed of $f_5 = f_4^2$, $f_4 = f_3 - f_2, f_3 = sin(f_1), f_2 = cos(f_1), f_1 = log(x)$. Symbolic differentiation gives $\frac{df_5}{dx} = -\frac{2(cos^2(log(x)) - sin^2(log(x)))}{x}$.

The *adjoint* is a support variable denoted with a bar as in $\bar{x} = \frac{\partial f_5}{\partial x}$ and denotes the perturbation (compare [200, Sec. 6.2.1 Eq 6.8]) or the sensitivity (compare [22, Sec 3.2 p. 12]) of a considered output $f_5$ w.r.t. changes in $x$. A vectorised evaluation of the adjoint is what PyTorch provides us with when querying the *x.grad* property of a tensor variable after conducting a backward pass.

8 Neural networks quite naturally induce a DAG as computational graph consisting of basic operations. This makes the idea of the reverse way quite interesting but in the same way as decomposing functions into elementary operations can turn out as quite difficult, inducing function(space)s from directed acyclic graphs is even more difficult as they need to be equipped with various properties, compare Part iii.

Observe, that we know the derivatives of the basic functions:

$$\frac{\partial f_5}{\partial f_4} = 2f_4, \quad \frac{\partial f_4}{\partial f_3} = 1, \quad \frac{\partial f_4}{\partial f_2} = -1,$$
$$\frac{\partial f_3}{\partial f_1} = cos(f_1), \quad \frac{\partial f_2}{\partial f_1} = -sin(f_1), \quad \frac{\partial f_1}{\partial x} = \frac{1}{x},$$

for example by symbolic derivation and hard-coded into software.

FORWARD PASS We go into the forward pass, set e.g. $x := 0.5$, and by following the dependencies we observe by means of the *chain rule* $\hat{x} = \frac{\partial f_5}{\partial x} = \frac{\partial f_5}{\partial f_1}\frac{\partial f_1}{\partial x} = \bar{f}_1 \frac{1}{x}$. Following the pattern we obtain in the next step $\bar{f}_1 = \frac{\partial f_5}{\partial f_1} = \frac{\partial f_5}{\partial f_2}\frac{\partial f_2}{\partial f_1} + \frac{\partial f_5}{\partial f_3}\frac{\partial f_3}{\partial f_1} = \bar{f}_2(-sin(f_1)) + \bar{f}_3 cos(f_1) = \bar{f}_3 cos(f_1) - \bar{f}_2 sin(f_1)$. The function evaluates at this point with $f_1(0.5) \approx -0.6931$.

Note, that the adjoints are still functions and we only need to keep track of the modified input vector and the dependencies in the graph. For each adjoint, we substitute parts of the chain rule such as $\frac{\partial f_2}{\partial f_1}$ with the known derivative – a function that can be evaluated as soon as the output was reached and the errors are getting backpropagated.

The algorithm continues with $\bar{f}_2 = \frac{\partial f_5}{\partial f_2} = \frac{\partial f_5}{\partial f_4}\frac{\partial f_4}{\partial f_2} = -\bar{f}_4$ and $\bar{f}_3 = \frac{\partial f_5}{\partial f_2} = \frac{\partial f_5}{\partial f_4}\frac{\partial f_4}{\partial f_2} = \bar{f}_4$ and the last adjoint gets to be $\bar{f}_4 = \frac{\partial f_5}{\partial f_5}\frac{\partial f_5}{\partial f_4} = \frac{\partial f_5}{\partial f_4} = 2f_4$ with $\bar{f}_5 = \frac{\partial f_5}{\partial f_5} = 1$ in our example.

The function evaluations at each node are stored such that one successively obtains the following values while traversing the computational graph:

$$x := 0.5, \quad f_1(0.5) = -0.6931, \quad f_2(0.5) = 0.7692,$$
$$f_3(0.5) = -0.63896, \quad f_4(0.5) = -1.4082, \quad f_5(0.5) = 1.9830$$

which are shown in Figure 5.6 above each node of the function graph (below for $f_3$, respectively).

BACKWARD PASS After the forward pass has reached the output node, the reverse pass can start. Compare Figure 5.6 to visually follow along the computation. During the forward pass the value $f_4(0.5) \approx -1.4082$ for the initial input $x := 0.5$ was obtained and can now be used to obtain a value for the adjoint under perturbation $u \in \mathbb{R}^1 := f(0.5) = 1.9830$:

- $\bar{f}_5[u] = 1$ evaluates to a constant as it is linear under an output perturbation.

With $\bar{f}_4 = 2f_4$ we obtain the first interesting evaluated adjoint and can further traverse the DAG back along the opposite direction of its edges.

- $\bar{f}_4[u] \approx 2 \cdot (-1.4082) = -2.8164$

We obtain

- $\bar{f_3}[u] \approx \bar{f_4}[u] = -2.8164$, and
- $\bar{f_2}[u] \approx -\bar{f_4}[u] = 2.8164$

and require both gradients to calculate $\bar{f_1}[u] = \bar{f_3} cos(f_1) - \bar{f_2} sin(f_1)$

- $\bar{f_1}[u] \approx= -2.8164 \cdot cos(-0.6931) - 2.8164 \cdot sin(-0.6931) = -0.36691$

such that we finally get $\bar{x} = \frac{\bar{f_1}}{x}$

- $\bar{x}[u] \approx \frac{-0.36691}{0.5} = -0.73382$

and the backward pass finished.

Picking up the functional form of an exemplary loss function, e.g. $\frac{1}{n}\sum_{i=1}^{n}(y_i - \hat{y}_i)^2$, and plugging in the neural network for the estimation $\hat{y}_i$ one obtains for example a composed function $\frac{1}{n}\sum_{i=1}^{n}(y_i - (W_2\sigma(W_1\mathbf{x} + B_1) + B_2))^2$ and can now use backpropagation with $n$ data sample points to obtain gradient information $\mathbf{g}_t$ w.r.t. each network parameter $\theta_t$ at a learning time step $t \in \mathbb{N}$.

CHECKING BACK WITH PYTORCH The example can be checked with few lines of python code. We construct an input tensor variable x for which we request the gradient later on. Gradients for intermediate computations (values of the adjoint) can be obtained when explicitly stating `f_1.retain_grad()` on desired variables. With calling `.backward()` on the final computation variable, PyTorch executes the reverse mode of automatic differentiation via vector-Jacobian products and returns $-0.7338$:

```
import torch
torch.__version__
>>> '1.10.2'

x = torch.tensor([0.5], requires_grad=True)
f_1 = torch.log(x)
f_2 = torch.cos(f_1)
f_3 = torch.sin(f_1)
f_4 = f_3-f_2
f_5 = torch.pow(f_4, 2)

f_5.backward()
x.grad
>>> tensor([-0.7338])
```

We can also use the result of symbolic differentiation and run it through a python interpreter:

```
df5 = lambda x: -(2*(math.cos(math.log(x))**2-math.sin(math.log(x
    ))**2))/x
df5(0.5)
>>> -0.733827898973207
```

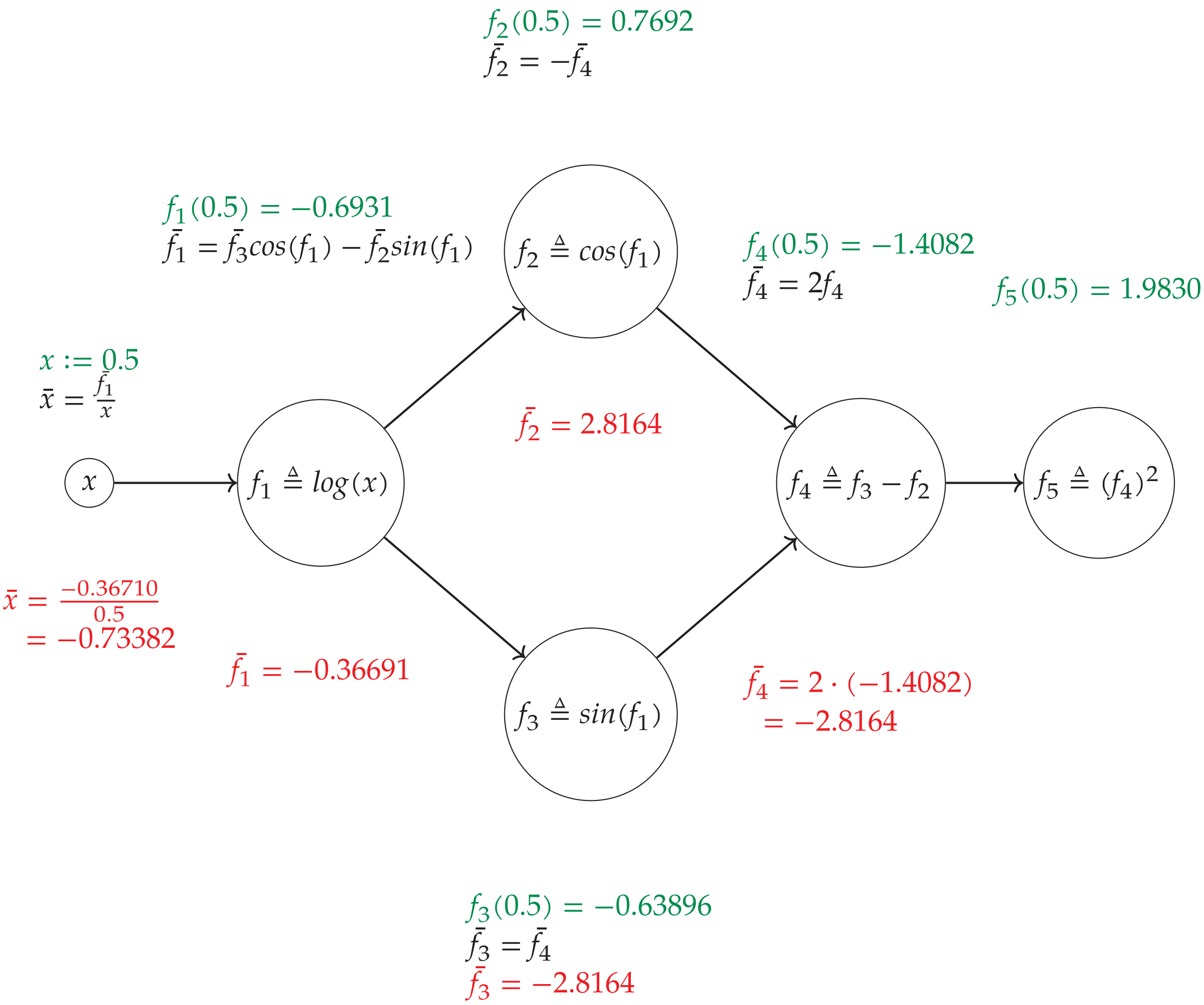

**Figure 5.6:** Both intermediate and final results during a run of the backpropagation algorithm are depicted within the computational graph given in Figure 5.5. The forward-pass equips each node with an intermediate function evaluation for an input (e.g. $x := 0.5$) on which the adjoints such as $\bar{f_4} = 2f_4$ depends. In a neural network, the basic functions are often composed of weighted sums as in matrix-vector multiplications. The automatic differentiation provides gradient information given an input vector and the computational graph. With this gradient information, sophisticated steps to decrease the final output, a data- and loss-dependent error, can be undertaken – the heart of deep learning.

### 5.6.3 *Loss Functions & Risk*

A probabilistic perspective on learning was motivated in section 5.4. While learning with MLE has a nice interpretation when approximating it with the cross-entropy loss (CE-loss) or an $l_2$ based loss, we will also make use of other loss functions which empirically work well. This goes along with a pure predictive setting as presented in [200, Chp. 14.1]. The form of maximum likelihood estimation in eq. (5.7) on page 61 can then be loosened towards minimizing a loss for the purpose of conducting empirical risk minimisation (ERM). Losses can be divided into classification and regression, i.e. discrete class and continuous valued targets. Often, however, if the form is analytically motivated as in the Kullback-Leibler divergence, there can be strong relations between both groups. For the discrete multi-class case, an approximation of the KLD turns into the form of a CE-loss.

We already mentioned losses in section 5.6.2 in that the neural network $f$ is not differentiated on its own, but in conjunction with a loss objective based on data. More properly looked at this aspect, we consider error values for single predictions and theoretically intend to conduct empirical risk minimisation (ERM) [200, Sec. 6.3.2] in which the risk refers to not just an aggregated form of errors on a portion of available data but the true underlying risk if all information of the data distribution would be available. Risk is the theoretical concept description, while we define *loss* based on an error-function and over a batch (or portion) of data. The loss is used to calculate an approximation of risk and even the loss is usually not computed over all available data samples as of its size but over a mini-batch.

We consider data $D_{train} = (X, Y) \subset (\mathcal{X} \times \mathcal{Y})^n$ with $((x_1, y_1), (x_2, y_2), \ldots, (x_n, y_n)) \overset{i.i.d.}{\sim} Pr_{\mathcal{X}\times\mathcal{Y}}$ and predictions $\hat{y}_i = f(x_i)$ of a deep neural network $f$ with $i \in \{1, \ldots, n\}$ and $\mathcal{Y}$ having dimensionality $d_{out}$, i.e. $\mathcal{Y} = \mathbb{R}^{d_{out}}$ or $\mathcal{Y} = \{0, 1\}^{d_{out}}$. An error function provides a real error value to assess how wrong a prediction is w.r.t. reference information. We treat error functions as a distance $l : \mathcal{Y} \times \mathcal{Y} \to \mathbb{R}$ with $l(\hat{y}, y) \geq 0$ for an expected value $y \in \mathcal{Y}$ and a prediction $\hat{y} \in \mathcal{Y}$. The signature can be e.g. $l : ([0,1]^{d_{out}}, [0,1]^{d_{out}}) \to \mathbb{R}$ for a discrete-output or $l : \mathbb{R}^{d_{out}}, \mathbb{R}^{d_{out}}) \to \mathbb{R}$ for a continuous-output neural network $f$.

We denote the error underlying the cross-entropy loss (CE-loss) with

$$l_{CE}(\hat{y}, y) \triangleq -\sum_{k=1}^{d_{out}} y_k \log(\hat{y}_k) \tag{5.10}$$

A loss (or cost) function can then be defined over a batch $D_{batch} \subseteq D_{train}$ of data and an error function $l$ as a real-valued function $\mathcal{L} : (f, D_{batch}, l) \to \mathbb{R}$ in which the signatures of $f$ and $l$ of course need to align. To conduct gradient-based learning, the loss has to be differentiable (at least up to the degree to conduct automatic differentiation). Otherwise, a gradient-free optimiser needs to be employed.

The CE-loss then can be defined as

$$\begin{aligned}\mathcal{L}_{CE}(f, D_{batch}) &\triangleq \frac{1}{|D_{batch}|} \sum_{(x,y)\in D_{batch}} l_{CE}(f(x), y) \\ &= -\frac{1}{|D_{batch}|} \sum_{i=1}^{|D_{batch}|} \sum_{k=1}^{d_{out}} y_{ik} \log(\hat{y}_{ik})\end{aligned} \tag{5.11}$$

and for continuous-output neural networks $f$ we can define the mean absolute error (MAE) based on the $l_1$-norm as

$$\begin{aligned}\mathcal{L}_{MAE} &\triangleq \frac{1}{|D_{batch}|} \sum_{(x,y)\in D_{batch}} \|f(x) - y\|_1 \\ &= \frac{1}{|D_{batch}|} \sum_{(x,y)\in D_{batch}} \sum_{j=1}^{d_{out}} |f(x)_j - y_j|\end{aligned} \tag{5.12}$$

and the mean squared error (MSE) based on the $l_2$-induced metric as

$$\begin{aligned}\mathcal{L}_{MSE} &\triangleq \frac{1}{|D_{batch}|} \sum_{(x,y)\in D_{batch}} \|f(x) - y\|_2^2 \\ &= \frac{1}{|D_{batch}|} \sum_{(x,y)\in D_{batch}} \left(\sqrt{\sum_{j=1}^{d_{out}} |f(x)_j - y_j|^2}\right)^2 \\ &= \frac{1}{|D_{batch}|} \sum_{(x,y)\in D_{batch}} \sum_{j=1}^{d_{out}} (f(x)_j - y_j)^2\end{aligned} \tag{5.13}$$

as similarly outlined by James et al. [117, Eq. 10.9] or Murphy [200, Sec. 15.1.4].

### 5.6.4 *Gradient-Based Learning for Deep Neural Networks*

At the heart of training deep neural networks relies a learning rule. With stochastic gradient descent (SGD) in its simplest form, the change of parameters $\boldsymbol{\theta}$ of a neural network can be expressed as:

$$\boldsymbol{\theta}_{t+1} = \boldsymbol{\theta}_t - \eta_t \mathbf{g}_t \tag{5.14}$$

with the gradient $\mathbf{g}_t$ of the loss, a learning rate (or step size) $\eta_t$ and the state of the parameters $\boldsymbol{\theta}_t$ at time step $t \in \mathbb{N}$ and after their update $\boldsymbol{\theta}_{t+1}$ at time step $t + 1$ (compare [200, Eq. 6.43, p.265]).

## 5.7 FURTHER CONCEPTS OF DEEP NEURAL NETWORKS

We shortly mention further concepts and terminology for deep neural networks which we require and refer later on in our experiments. These include convolutional neural networks, *weight sharing*, batch normalisation, layer normalisation and graph neural networks (GNNs).

### 5.7.1 *Convolutional Neural Networks*

Convolutional neural networks are deep neural networks with convolutional layers and the term CNN might refer to the underlying principle or to a whole convolution-based architecture.

Convolutional layers make an implicit or explicit assumption about the relationship of the input data they are applied to. With this assumption, they can reduce the number of parameters and additionally employ weight sharing techniques. This combination makes CNN-based network designs significantly more powerful over standard fully-connected networks.

The example in fig. 5.1 depicts very small images of $4 \times 4$ width and height and additionally three channels such that a data tensor $\mathbf{x} \in \mathbb{R}^{\text{chans}[3] \times \text{height}[4] \times \text{width}[4]}$ is of shape $(3, 4, 4)$. Fully-connected neural networks as presented with multiple hidden layers in section 5.5.1 consume this input by flattening the data tensor into $\mathbf{x} \in \mathbb{R}^{54}$. The signature then can e.g. become $\mathbb{R}^{54} \to \mathbb{R}^{\text{hidden}_1} \to \mathbb{R}^{\text{hidden}_2} \to \cdots \to \mathbb{R}^{\text{hidden}_L} \to \{0, \dots, C\}$ with $L \in \mathbb{N}$ being the networks depth and $C \in \mathbb{N}$ being the number of classification targets.

Le Cun et al. noted that this fully connected layers "would have far too many parameters" [151, Sec. 4] and they proposed that "a restricted connection-scheme must be devised, guided by our prior knowledge about shape recognition" [151].

Convolutions preserve the spatial tensor shape and use small parametric *filters* (or kernels) to compute local dot-products. A filter is a parametric matrix $F \in \mathbb{R}^{k_1 \times k_2}$ with $k_1, k_2 \in \mathbb{N}$ being its filter dimensions with usually $k_1 = k_2$. Multiple filters can be stacked into e.g. three channels $F \in \mathbb{R}^{\text{chans}[3] \times k_1 \times k_2}$. These convolutional filters are re-used in calculating dot products as they slide across the width- and height-dimensions of the input and multiple filter dimensions are used to sustain the channel dimension.

While convolutions are well-defined in functional analysis, deep learning misuses the term and often refers to a discrete variant. Many technical libraries "implement a related function called the *cross-correlation*, which is the same as convolution but without flipping the kernel" [79, Sec. 9.1, eq. 9.6]. For example, LIB PyTorch uses "discrete convolution which is defined for a 2D image as the scalar product of a weight matrix, the kernel, with every neighborhood in the input" [268, Sec. 8.1.1, p. 195]. Stevens, Antiga & Viehmann also note, that "there is a subtle difference between PyTorch's convolution and mathematics' convolution: one argument's sign is flipped" and suggest that "one could call PyTorch's convolutions discrete cross-correlations" [268, Sec. 8.1.1, p. 195].

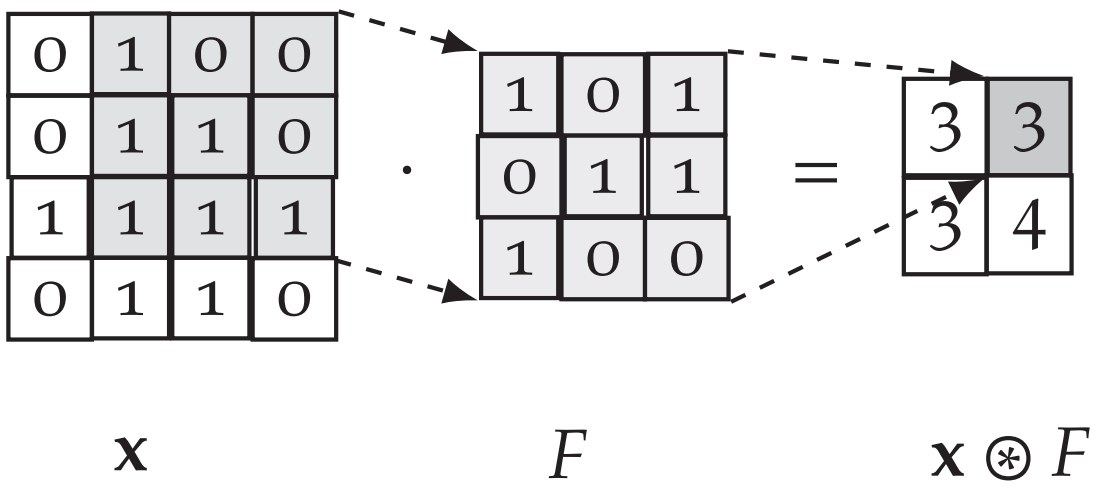

**Figure 5.7:** Depiction of the convolutional operation $\mathbf{z} = \mathbf{x} \circledast F$ as discrete cross-correlation between input $X \in \mathbb{R}^{\text{height}[4] \times \text{width}[4]}$ and filter $F \in \mathbb{R}^{k_1[2] \times k_2[2]}$. Visualisation adapted from tikz.net.

We follow Murphy in denoting the cross-correlation operation with $\mathbf{z} = \mathbf{x} \circledast F$ [200, Sec. 16.2.3, eq. 16.4].

$$\mathbf{z}_{i,j} = \sum_{u=0}^{k_1-1} \sum_{v=0}^{k_2-1} \mathbf{x}_{i+u,j+v} F_{u,v} \tag{5.15}$$

The operation is visualised in fig. 5.7. Images as depicted in fig. 5.1 often have multiple channels used to carry color information, i.e. three channels of red-green-blue for RGB-coded images. A convolutional layer then generally has $\text{outputChans} \in \mathbb{N}$ filters $F \in \mathbb{R}^{k_1 \times k_2 \times \text{chans}}$ – each of a shape that respects the number of input channels $\text{chans} \in \mathbb{N}$. The (2d) CNN-layer then computes [200, Sec. 16.2.3, eq. 16.5]

$$\mathbf{z}_{i,j,d} = \sum_{u=0}^{k_1-1} \sum_{v=0}^{k_2-1} \sum_{c=0}^{C-1} \mathbf{x}_{i+u,j+v,c} F_{u,v,c,d} \tag{5.16}$$

in which multiple filters take a $\text{height} \times \text{width}$-dimensional input with chans channels and map it to a reduced $\text{height}-2 \times \text{width}-2$-dimensional output of outputChans channels.

There exist lots of variations to this formulation in which input edge cases with no zero *padding*, the sliding of the filters with *strides* [158], or a scattering of the filters through *dilation* [329] are considered.

Keep in mind, that while the filter reduces the parametric space $\theta(f)$ of a model $f$ drastically, a fully-connected neural network is still capable of representing the same function $f$. Given a convolutional layer $f_c$ one can easily construct a non-convolutional fully-connected neural network $f_{mlp}$. This makes the relationship of these two types of architectures especially interesting in context of analyses of structures. Convolutions impose a structural prior on the connectivity between two consecutive layers. Le Cun notes already 1989 that "weight sharing can be interpreted as imposing equality constraints among the connection strengths" [151, Sec. 2.1, p.3]. Filters, carrying the weight parameters of convolutional layers, are *shared* across different connections of an equivalent fully-connected neural network. Or, in other words, weights of filters $F$ are re-used multiple times when sliding across the input $X$.

### 5.7.2 *Normalisation Techniques*

When, a decade ago, deep neural networks became significantly deeper, such as with 8-layered AlexNet[9] [139] or 100-layered HighwayNets [267], one of the major issues which needed to be solved was that of vanishing or exploiding gradients when backpropagated across many layers – similarly as it was previously analysed for deep recurrent neural networks when unfolded through a time dimension [101, 102].

The vanishing gradient effect is a "decreasing gradient norm as the networks grow in depth" [45]. Daneshmand et al. note, that "randomly initialised neural networks are known to become harder to train with increasing depth, unless architectural enhancements like residual connections and batch normalisation are used" [45]. Amongst other techniques such as residual connections, model-dependent weight initialisations (compare section 5.2.3), and data normalisation during pre-processing, normalisation techniques within the architecture of deep neural networks became a driving factor (or critical component) for the successive training of very large neural network models [253].

Le Cun noted on normalizing the inputs that "in general, any shift of the average input away from zero will bias the updates in a particular direction and thus slowdown learning" [149, Sec. 1.4.3] and they empirically found that "convergence is faster not only if the inputs are shifted as described above but also if they are scaled so that all have about the same covariance" [149, Sec. 1.4.3].

Early and notably techniques are batch normalisation [113], weight- [242] and layer normalisation [11]. Goodfellow et al. called batch normalisation the "most exciting recent innovations in optimizing deep neural networks" [79, Sec. 8.7.1, p. 317]. There exist many more approaches and Lubana et al. [173] "analyze ten normalisation layers [which] were chosen to cover a broad range of ideas: e.g., activations-based layers [113], parametric layers [242], hand-engineered layers [316], AutoML designed layers [166], and layers [11, 45] that form building blocks of recent techniques [175]".

BATCH NORMALISATION Ioffe et al. introduced batch normalisation for which they "seeked to reduce the internal covariate shift" which they define as "change in the distribution of network activations due to the change in network parameters during training" [113, Sec. 2, p.2]. Further, they "expected to improve the training speed" [113, Sec. 2, p.2] and found batch normalisation to "also act as a regulariser" [113].

9 Alex Krizhevsky describes the AlexNet architecture over the IMAGENET input of $\mathbb{R}^{\text{height}[224]\times\text{width}[224]\times\text{chans}[3]}$ with [253 440, 186 624, 64 896, 64 896, 43 264, 4096, 4096, 1000] [139, Fig. 2, p. 5] – although the flattened out dimensions hide details of the underlying convolutional filter sizes.

Let $D_{batch} = \{(\mathbf{x}_1, y_1), \dots, (\mathbf{x}_{n_{batch}}, y_{n_{batch}})\} \subset D_{train}$ be a training batch with $n_{batch} \in \mathbb{N}$. Batch normalisation acts as an own layer which computes $\forall i \in \{1, \dots, n_{batch}\}$ [113, Sec. 3, p.3]:

$$\hat{\mathbf{x}}_i = \frac{\mathbf{x}_i - \mathbb{E}(\mathbf{x})}{\sqrt{Var(\mathbf{x}_i + \epsilon)}}\gamma + \beta \tag{5.17}$$

in which $\gamma$ and $\beta$ are learnable parameters of the layer and $\mathbb{E}(\mathbf{x})$ is estimated with the sample mean

$$\mu_{batch} = \frac{1}{n_{batch}} \sum_{i=0}^{n_{batch}} \mathbf{x}_i$$

and $Var(\mathbf{x})$ is estimated with the (biased) sample variance

$$\sigma^2_{batch} = \frac{1}{n_{batch}} \sum_{i=0}^{n_{batch}} (\mathbf{x}_i - \mu_{batch})^2$$

The technical detail of $\epsilon > 0$ is to ensure non-zero values and is usually kept constant at e.g. $10^{-5}$. For an input shape $(n_{batch}, \text{chans}, \text{height}, \text{width})$ the parameters $\gamma$ and $\beta$ are of shape $(\text{chans}, \text{height}, \text{width})$ and initialised usually with $\gamma = (1, \dots, 1) = \mathbf{1}$ and $\beta = (0, \dots, 0) = \mathbf{0}$ [10]. The parameter $\gamma$ is referred to as *scale* and the parameter $\beta$ as *shift* of the batch normalisation layer.

Using our neural network notation from section 5.5, batch normalisation can be integrated directly on the unsquished output $z^{(l)}$ or on the activated output $h^{(l)}$ of a preceding layer $l \in \{1, \dots, L\}$. $\mathbf{x}$ in eq. (5.17) then refers to either $z^{(l)}$ and can be additionally activated such that $h^{(l+1)} = \sigma(\hat{\mathbf{x}})$ or it can stay unactivated such that $h^{(l+1)} = \hat{\mathbf{x}}$ or $\mathbf{x}$ can refer to the activated output $h^{(l)}$ such that $h^{(l+1)} = \hat{\mathbf{x}}$. There exist different opinions on whether to apply normalisations before or after activations.

We originally considered batch normalisation in some experiments but switched mostly to layer normalisation for the following reasons:

- Wu et al. note that the "unique property" of batch normalisation "of operating on *batches* instead of individual samples introduces significantly different behaviors from most other operations in deep learning. As a result, it leads to many hidden caveats that can negatively impact model's performance in subtle ways" [317]. This can be seen in eq. (5.17) by the fact, that the output $\hat{\mathbf{x}}$ is dependent on not just $\mathbf{x}_i$ but the expectation over a mini-batch – which is sample-dependent.

- Santurkar et al. consider the effectiveness of batch normalisation as "still poorly understood" but claim that it "makes the optimisation landscape significantly smoother" [245]. On the other hand, Santurkar et al. point out that batch normalisation "might not even be reducing internal covariate shift" [245].

10 Compare the PyTorch documentation on BatchNorm2d for more details.

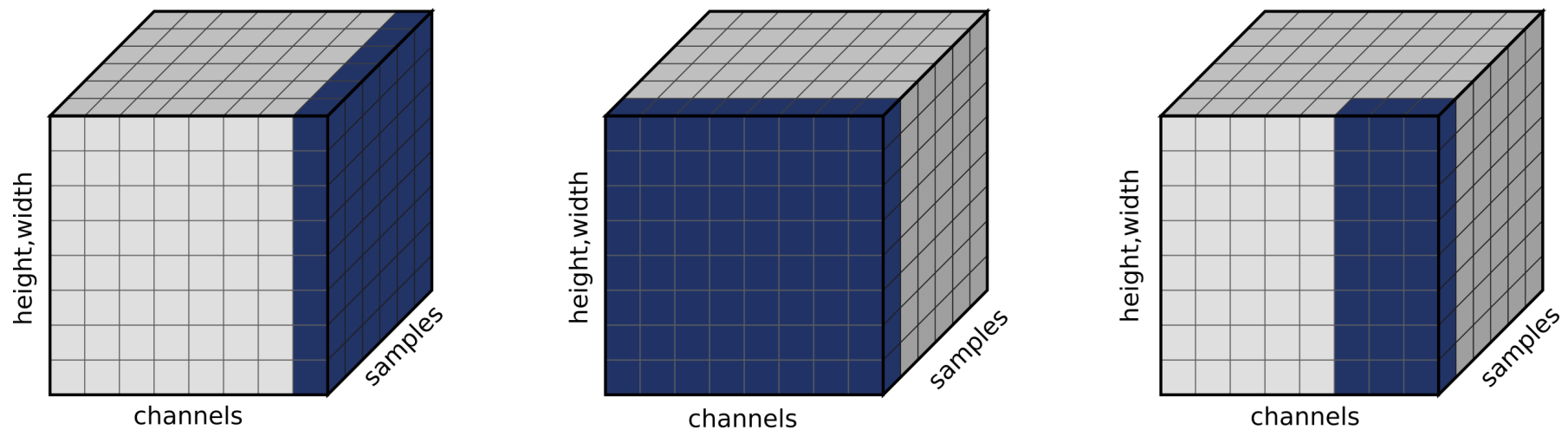

**Figure 5.8:** Differences between normalisation techniques. From left to right: Batch normalisation [113], layer normalisation [11] and group normalisation [316]. Visualisation inspired by [316, Fig. 2].

- As Ba et al. note that "batch normalisation requires running averages of the summed input statistics" and it "cannot be applied to online learning tasks or to extremely large distributed models where the minibatches have to be small" [11].
- Bronstein et al. also write that "a general theory that can provide guiding principles is still missing at the time of writing" [28, Sec. 5.1, p. 73].

LAYER NORMALISATION Ba et al. propose layer normalisation [11] as an alternative normalisation technique in which they estimate the following sample mean and sample standard deviation for a layer $l \in \{1, \dots, L\}$ of size $h \in \mathbb{N}$:

$$\mu^{(l)} = \frac{1}{h}\sum_{i=1}^{h} z_i^{(l)} \quad \sigma^{(l)} = \sqrt{\frac{1}{h}\sum_{i=1}^{h}(z_i^{(l)} - \mu^{(l)})^2} \tag{5.18}$$

Like in other normalisation layers, a scale (or weight) and shift (or bias) are learned in shape of the input shape excluding the sample dimension. Therefore, layer normalisation can be seen as a map $\mathbb{R}^{n_{batch}\times\text{chans}\times\text{heightwidth}} \to \mathbb{R}^{n_{batch}\times\text{chans}\times\text{height}\times\text{width}}$ computing the normalisation in dimensions chans × height × width.

REMARKS

- Normalisation techniques have become part of the architecture as own layers instead of treating normalisation outside the training loop
- We mostly employ layer normalisation for very deep neural networks
- DeepStruct provides a switch to enable or disable normalisation layers and for graph-based constructions of neural network realisations we almost always used layer normalisations

### 5.7.3 *Graph Neural Networks*

Graph neural networks (GNNs) are end-to-end differentiable neural network models that have been transferred to the domain of graphs. While we do not further investigate on the structure of GNNs itself, we employ GNNs to investigate on the structure of classical deep neural networks. The reason for not considering GNNs will be discussed later more thoroughly but can be summarised in that permutation invariances make it difficult to impose a structure on DNNs as we propose it in chapter 6 on page 101. Such formulations occur not only with GNNs but also with deep convolutional neural networks in which pooling operations are employed. GNNs are still very important in our work as to use them e.g. to learn graph classification models as to analyse structures of DNNs or to learn graph generative models.

Graph neural networks are part of endeavours in the field of geometric deep learning which deals with broadening deep neural networks to data of more general non-euclidean topologies. Bronstein et al. summarises these topologies as the "5Gs of geometric deep learning: grids, groups, graphs, geodesics, and gauges" [28]. They suggest "using the lens of invariances and symmetries" to study them and propose a blueprint of geometric deep learning [28, Sec. 3.5]. We refer the interested reader to Bronstein et al. [28] and Hamilton et al. [86] for more technical depth on GNNs or geometric deep learning and focus here only on selected and for our work necessary GNN-techniques.

GRAPH CONVOLUTIONAL NETWORKS An early special case, graph convolutional networks (GCNs), was introduced by Kipf & Welling [129]. They proposed a layer-wise linear model over graphs with repeated applications. Consider the adjacency matrix $\mathbf{A}^{\pi}(G)$ of a graph $G = (V, E)$ of order $n = |V|$ under a vertex permutations $\pi$. $\tilde{\mathbf{A}}$ is the adjacency matrix with self connections [129, Eq. 2, p.2] by adding the identity matrix $I_n$. Each vertex is associated with a feature vector $\mathbf{x} \in \mathbb{R}^d$ such that we obtain a stacked feature matrix $X \in \mathbb{R}^{n \times d}$ (according to $\pi$).

With a degree matrix $D$ of $G$ and parameters $W^{(l)} \in \mathbb{R}^{d \times H_l}$ of dimension $H_l \in \mathbb{N}$ the graph convolutional layer of number $l \in \{1, \dots, L\}$ is given as $f_{conv} : (\mathbb{R}^{n \times H_{l-1}}, \mathbb{R}^{n \times n}) \to \mathbb{R}^{n \times H_l}$ [129, Eq. 2]:

$$f_{conv}(\mathbf{h}^{(l-1)}, \mathbf{A}) \triangleq \mathbf{z}^{(l)} \mapsto D^{-\frac{1}{2}} \tilde{\mathbf{A}} D^{-\frac{1}{2}} \mathbf{h}^{(l-1)} W^{(l)}$$

for which the initial input is $\mathbf{h}^{(0)} = X$. Analogously to section 5.5.2 on page 65, $H_i \in \mathbb{N}$ for layers $i \in \{1, \dots, L\}$ denote the dimensionalities of the feature representation per layer. The pre-activation $\mathbf{z}^{(l)}$ is further passed through an activation function $\sigma$ such that $\mathbf{h}^{(l)} = \sigma(\mathbf{z}^{(l)})$.

Kipf & Welling managed to repeat such graph convolutional layer applications by introducing a *renormalisation trick* [129, Sec. 2.2, p.3]

$I_n + D^{-\frac{1}{2}}\mathbf{A}D^{-\frac{1}{2}} \rightarrow D^{-\frac{1}{2}}\tilde{\mathbf{A}}D^{-\frac{1}{2}}$ with $\tilde{\mathbf{A}} = \mathbf{A} + I_N$, i.e. they considered a normalised version of the Laplacian of the underlying graph.

## 5.8 MACHINE LEARNING EVALUATION

Evaluating machine learning models is a large topic on its own and we restrict our attention here to the following intentions:

1. assessing whether a model learned something (at all),
2. comparing deep learning models under performance scores,
3. analysing the correlation between experimental variables,
4. comparing the quality of generative models with statistical divergences w.r.t. expected distributions,
5. and conducting some more advanced hypothesis tests such as to compare classifiers in a Friedman-Nemeny-test.

All of these intentions are tied to forms of hypothesis testing. In many manchine learning applications, the term *hypothesis testing* is often not explicitly mentioned but implicitly applied. For example, computing evaluation scores of a binary classifier from predictions on test data can be looked at from a binary hypothesis testing perspective. The major difference is that machine learning is often concerned with evaluating the quality of a *prediction* of a model, while hypothesis testing is a more rigorous framework "to decide whether or not some hypothesis that has been formulated" (through means of a model) "is correct" [153, Sec. 3.1].

Conducting hypothesis tests is a "topic on which entire books can be (and have been) written" [117, Chp. 13, p. 557] such as in *Statistical Hypothesis Testing* by Lehman & Romano [153] or in *Information Theory*: *Binary hypothesis testing* by Polyanskiy & Wu [227, Pt. III] on which we would like to refer more interested readers.

The underlying hypothesis tests to choose differ heavily for the particular experimental setting. We organise our employed methods into evaluation of classification models in section 5.8.1 and further methods of evaluation such as assessing regression or generative models (especially for graphs) in section 5.8.2 on page 89.

Practically, the intentions are roughly approached in the following way:

1. Assessing whether a classifier learned at all can be easily determined by knowing the underlying target class distribution and if it is uniformly distributed, a simple check of accuracy, precision, recall or $F_1$ score suffice to understand whether the classifier is better than e.g. choosing any target class randomly. For regression models, a correlation coefficient such as Pearson's $\rho$ or the coefficient of determination, coefficient of determination, can be used in conjunction with an contextual understanding of the mean absolute error or mean squared error.

2. Comparing deep learning models under performance scores such as precision, recall, $F_1$ score or Pearson's $\rho$, mean absolute error (MAE), mean squared error (MSE) then also involves evaluation strategies with at least employing holdout data sets for validation and testing besides a training subset of the overall data set. Hypothesis tests checking for the *goodness of fit* [153, Chp. 14] such as the Kolmogorov-Smirnov test [153, Sec. 14.2] are also used.
3. For analyses between structural properties (of graphs that are used to construct deep neural networks) and above mentioned performance scores such as the accuracy, we used correlation analysis and visual tools such as *scatter plots*, *histograms* and kernel-density estimations, e.g. with pair plotting from PLOTLY [226].
4. Comparing the quality of generative models of graphs is conducted by comparing statistical divergences such as the Kullback-Leibler divergence or maximum mean discrepancy between graph property distributions of a generated and a reference test set of graphs. Further, embedding-based measures are employed which are further explained and elaborated in Chapter 14.
5. In some experimental settings such as for the comparison of deep neural networks based on over 800 graph-induced architectures, a rank comparison based on the Friedman-Nemeny test [48] have been employed.

### 5.8.1 *Classifier Evaluation*

Classification models can be seen as a function $f_c : D \rightarrow C$, namely a mapping from a domain $D$ into $n_c \in \mathbb{N}$ different classes $C = \{1, \dots, n_c\}$. The simplest case with $n_c = 2$ is the one of binary classification. We introduce the idea of a confusion matrix and evaluation measures such as precision, recall, accuracy, and $F_1$ score based on this case.

For a set of $n_{eval} \in \mathbb{N}$ samples $D_{eval} = \{(\mathbf{x}_1, y_1), \dots, (\mathbf{x}_{n_{eval}}, y_{n_{eval}})\}$, we obtain a set of predictions $\{\hat{y}_i \mid \forall i \in \{1, \dots, n_{eval}\} : \hat{y}_i = f_c(\mathbf{x}_1)\}$. Further, let $c_{pos} \in C$ be the **positive class**. In the binary setting, the negative class w.r.t. $c_{pos}$ then has only one element, and in general $C_{neg} = C \cap \{c_{pos}\}$.

A confusion matrix $\mathbf{C} = (c_{ij}) \in \mathbb{N}^{n_c \times n_c}$ for a classifier $f_c$ and data $D_{eval}$ is given as

$$c_{ij} \triangleq \sum_{k=1}^{n_{eval}} \mathbb{I}_{y_k = i \wedge \hat{y}_k = j}$$

and its diagonal entries $c_{ii}$ for $i \in C$ contain exactly the predictions of $f_c$ that are **true** w.r.t. to the provided data. All other entries contain **false** predictions, while they are commonly categorised into type I and type II errors, as shown in section 5.8.1.

| Predicted → ↓ Actual | Positive | Negative | |
|---|---|---|---|
| Positive | True Positive (TP) | False Negative (FN) **Type II Error** | **Recall** sensitivity true positive rate $\frac{TP}{TP+FN}$ |
| Negative | False Positive (FP) **Type I Error** | True Negative (TN) | **Specificity** selectivity true negative rate $\frac{TN}{TN+FP}$ |
| | **Precision** $\frac{TP}{TP+FP}$ | **Negative Predictive Value** $\frac{TN}{TN+FN}$ | **Accuracy** $\frac{TP+TN}{TP+TN+FP+FN}$ |

**Table 5.1:** Confusion matrix **C** of a binary classification setting with equations for common evaluation measures derived from it such as the accuracy $\alpha = \frac{TP+TN}{TP+TN+FP+FN}$. The semantics of rows and columns are also often swapped.

The naming **positive** and **negative** in section 5.8.1 refers to in which class the classifier $f_c$ put a data sample $\mathbf{x}$ into. Four terms are common to derive further evaluation measures from the confusion matrix:

- *True positives* **TP** refers to the number of samples in $D_{eval}$ which the classifier has assigned to the class $c_{pos}$ and did so correctly according to their class labels, i.e. $c_{c_{pos}c_{pos}}$.
- Analogously, *true negatives* **TN** refers to the number of samples in $D_{eval}$ which have not been assigned to the positive class and did so correctly according to their class labels, i.e. $\sum_{i,j \in C_{neg} \times C_{neg}} c_{ij}$.
- *False positives* **FP** refers to the number of samples in $D_{eval}$ which were put into the positive class $\hat{y}_i = c_{pos}$ by $f_c$ but actually do not belong in it according to $y_i$, i.e. $\sum_{k \in C_{neg}} c_{kc_{pos}}$.
- *False negatives* **FN** refers to the number of samples in $D_{eval}$ which were not put into the positive class by prediction $\hat{y}_i$ of $f_c$ but should have been assigned to it, i.e. $\sum_{k \in C_{neg}} c_{c_{pos}k}$.

Observe, that for the binary case the terms TP, TN, FP, and FN simply refer to single entries of the confusion matrix, while for multi-class cases it is more important to focus on the fact that the negative classes are viewed under the perspective of a positive class such that there are multiple of them. Except for the true positives TP, sums to count the number of samples of a certain type run over multiple cells of the confusion table.

We define accuracy as

$$\alpha(f_c; D_{eval}) \triangleq \frac{TP + TN}{TP + TN + FP + FN} \tag{5.19}$$

and precision as

$$\pi(f_c; D_{eval}) \triangleq \frac{TP}{TP + FP} \tag{5.20}$$

and recall as

$$\rho(f_c; D_{eval}) \triangleq \frac{TP}{TP + FN} \tag{5.21}$$

and the *true negative rate* as $\frac{TN}{TN+FP}$ such that the $F_\beta$-score as a importance-scaled mean between precision and recall becomes $F_\beta = (1 + \beta^2) \cdot \frac{\pi \cdot \rho}{(\beta^2 \cdot \pi) + \rho}$ which for $\beta = 1$ becomes the harmonic mean between precision and recall, the $F_1$ score:

$$F_1(f_c; D_{eval}) \triangleq 2 \frac{\pi(f_c; D_{eval}) \cdot \rho(f_c; D_{eval})}{\pi(f_c; D_{eval}) + \rho(f_c; D_{eval})}$$

We will usually just refer to $\alpha(f_c), \dots, F_1(f_c)$ for a neural network classifier realisation $f_c$ whose domain and co-domain fit an implicitly defined evaluation data set $D_{eval}$.

### 5.8.2 *Further Tools for Evaluation*

A regression model has a continuous output variable $y_i \in \mathbb{R}$ in contrast to classification models. For an evaluation set $D_{eval} = \{(\mathbf{x}_1, y_1), \dots, (\mathbf{x}_{n_{eval}}, y_{n_{eval}})\}$ in which the response $y_i \in \mathbb{R}$ to an input $\mathbf{x}_i$ is known for $i \in \{1, \dots, n_{eval}\}, n_{eval} \in \mathbb{N}$, the point-wise error $d(\mathbf{x}_i, y_i)$ in terms of a distance $d$ is a typical first approach to assess how close a regression model predicts an expected target. This makes errors or losses as we discussed them in section 5.6.3 not only an objective for learning through empirical risk minimisation but also a tool for evaluation. The mean absolute error, for example, is a valuable tool to assess how far predictions on the evaluation set have been off from expectations.

We use the following further tools for evaluation:

- The coefficient of determination to measure the *goodness of fit* of a regression model,
- Pearson's $\rho$, a coefficient of correlation to assess the (linear) relation between two sets of predictions,
- rank correlation coefficients such as Spearman's $\rho$ to assess e.g. a predicted and an expected ranking,
- tests such as the Shapiro-Wilk test to assess whether samples are normally distributed,

- and the Kolmogorov-Smirnov test [153, Eq. 6.58, p. 256] to compare continuous distributions which are represented by samples.

James et al. define the coefficient of determination $R^2$ [117, Sec. 3.1.3, p. 77] based on the residuals sum squared $RSS$ and the total sum of squares $TSS$ [117, Sec. 3.1.3, p. 77]. The residuals sum squared $RSS(Y, \hat{Y}) \triangleq \sum_{i=1}^{n} (y_i - \hat{y}_i)^2$ is the non-averaged MSE $RSS(Y, \hat{Y}) = n \cdot \mathcal{L}_{MSE}(Y, \hat{Y})$ with predictions $\hat{Y} = f(Y)$ from a model $f$. The total sum of squares $TSS$ $TSS(Y, \hat{Y}) \triangleq \sum (y_i - \hat{y}_i)^2$ "measures the total variance in the response $Y$" [117, p. 79]. The coefficient of determination $R^2$ is then given as [117, Eq. 3.17]

$$R^2(Y, \hat{Y}) \triangleq \frac{TSS - RSS}{TSS} = 1 - \frac{RSS}{TSS} \tag{5.22}$$

and "measures the proportion of variability in $Y$ that can be explained using $X$". A number close to 1 "indicates that a large proportion of the variability in the response is explained by the regression" while a number closer to 0 "indicates that the regression does not explain much" of it [117, p. 79]. Further note, that "the $R^2$ statistic has an interpretational advantage over the residual standard error (RSE), since unlike the RSE, it always lies between 0 and 1"[117, Sec. 3.1.3, p. 77].

The sample correlation coefficient of Pearson's $\rho$ is given in Lehmann et al. [153, Sec. 5.13, p. 190] or James et al. [117, Eq. 3.18, p. 79] as

$$\begin{aligned} \rho_p(X, Y) &\triangleq \frac{cov(X, Y)}{\sigma_X \sigma_Y} \\ &= \frac{\sum \left( (X_i - \bar{X})(Y_i - \bar{Y}) \right)}{\sqrt{\sum (X_i - \bar{X})^2 \sum (Y_i - \bar{Y})^2}} \end{aligned} \tag{5.23}$$

and we refer to it as Pearson's $\rho$. The method is also referred to as "that of *product moment*" [264, p. 77]. Note that for simple *linear* regression "the squared correlation and the $R^2$ statistic are identical" [117, p. 80].

For assessing two rankings, we employ rank correlation coefficients such as Spearman's $\rho$ [264, Sec. 4 d), p. 86]. The Spearman's $\rho$ is based on Pearson's $\rho$ and is given as

$$\begin{aligned} \rho_s(X, Y) &\triangleq \frac{cov(R(X), R(Y))}{\sigma(R(X))\sigma(R(Y))} \\ &= \rho_p(R(X), R(Y)) \end{aligned} \tag{5.24}$$

with $R(Z)$ denotes the ranking of a random variable (or its samples). Underlying the ranking of $Z$ are order statistics where $Z_{(i)}$ is the $i$-th smallest random variable. Spearman's $\rho$ is restricted by -1 and 1 and assesses how similar two rankings are.

The Shapiro-Wilk test is an examplary hypothesis test and its purpose is to assess whether samples are normally distributed. The "W test statistic for normality is defined by" Shapiro & Wilk over $n$ samples $y = (y_1, \dots, y_n)$ ("a vector of ordered random observations" [250, Sec. 2]):

$$W \triangleq \frac{\left(\sum_{i=1}^{n} a_i y_i\right)^2}{\sum_{i=1}^{n} (y_i - \bar{y})^2} \tag{5.25}$$

with "$m' = (m_1, m_2, \dots, m_n)$ denoting the vector of expected values of standard normal order statistics, and let $V = (v_{ij})$ be the corresponding $n \times n$ covariance matrix" [250, Sec. 2] and the coefficients $a_i$ be given as $(a_1, \dots, a_n) = \frac{m' V^{-1}}{(m' V^{-1} V^{-1} m)^{\frac{1}{2}}}$. The statistics of a standard normal distribution are usually randomly sampled.

The mentioned tools are common ways for statistically assessing experiments and we use them extensively in our chapters on structure analysis and subsequent experiments on NAS methods.

*Evaluation of Learned Generative Models*

We consider two types for evaluation of (graph) generative models.

The first type considers statistical divergences between distributions of properties from a set of generated samples $D_{gen}$ and a set of expectations $D_{eval}$. This generically evaluates a given generative model in a supervised fashion based on an evaluation set of examples. A downside of this approach can be costly operations to obtain the properties. For example, obtaining properties from a set of graphs can be computationally expensive for many types such as based on on path lengths, centralities or clustering coefficients.

The second type of evaluating generative models employs a classification model $f_{eval}$ which carries an embedding $f_{eval}^{\downarrow}$ into a $d$-dimensional manifold on which the evaluation between generated samples $D_{gen}$ and expectations $D_{eval}$ is carried out. The embedding $f_{eval}^{\downarrow}$ yields representations $\mathbf{h}_{gen} = f_{eval}^{\downarrow}(D_{gen})$ and $\mathbf{h}_{eval} = f_{eval}^{\downarrow}(D_{eval})$. In the manifold of $f_{eval}$, metrics such as *precision*, *recall*, *density*, and *coverage* [142] can be defined (other than the classical precision etc). For $f_{eval}$ a learned DNN classifier or a randomly initialised DNN can be employed.

We found both types of evaluation approaches useful in context of graph generative models [282].

### 5.8.3 *Validation with Resampling Methods*

The simplest approach to estimate the performance of a DNN is to compare predictions of it with reference data from a test sample set, which is is disjoint from the samples used for training. This is called holdout validation and for available real-world data it can be simply referred to as $D_{full} = D_{train} \cup D_{test}$ where usually $D_{train} \cap D_{test} = \emptyset$.

Because we are often conducting hyperparameter studies in which the influence of a hyperparameter is investigated by repeatedly estimating its influence on a resulting performance, overfitting on the test set would at some point occur. Therefore, cross-validations is further used in which data is not only split into a training and testing set but further a validation set on which the performance of a model after training is measured to estimate the influence of a hyperparameter before conducting a final evaluation on the test set after the hyperparameter study (or optimisation loop). Figure 5.9 illustrates this idea of holding out two sets $D_{valid}$ and $D_{test}$ besides a larger training set $D_{train}$.

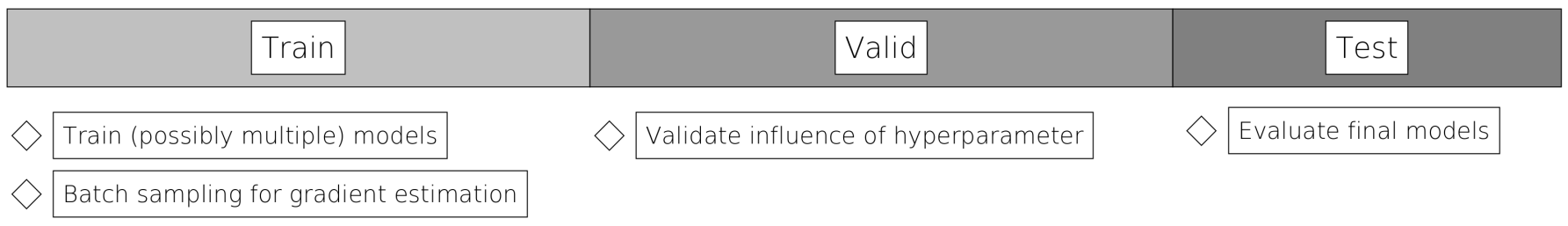

**Figure 5.9:** A sketch of a holdout strategy which splits all available data samples into separate training $D_{train}$, validation $D_{valid}$, and testing $D_{test}$ subsets. The subsets have different purposes such as finding a good neural network realisation through learning on the train set, validating the influence of second-order optimisation problems such as assessing hyperparameter influences on the validation set or to assess the overall risk of a model through an estimation of an evaluation score on the test set.

There exist further resampling strategies on the available data as to avoid overfitting or reducing variance. Examples include *leave p out*, repeated random subsampling validation, stratified $k$-fold cross-validation, repeated $k$-folds, and nested $k$-folds.

## 5.9 UNIVERSAL APPROXIMATION THEOREMS

The following section provides an overview of research on universal approximation capabilities of neural networks. Universal approximation refers to the properties of sets of neural networks to closely approximate functions from well-understood function spaces such as the continuous functions. Usually, this means that such sets of neural networks uniformly approximate continuous functions on compact spaces. We follow researchers working in that field like Kidger et al. in their notation and language [126].

Importantly, we are interested in UAPs for two reasons:

1. **Comparable Potential**: By engineering deep neural networks in a way that they posess the same *minimum learning capabilities* we improve reliability of empirical research between their performance on application domains and their underlying structure. Despite the complexity of engineering of deep neural networks with high amounts of data and often in distributed settings with heterogeneous hardware environments, we enforce awareness of the many influences of observed estimated properties of such deep neural nets. A quite obvious example of common capabilities would be to not simply include any operational layer in neural network design without an understanding of the implications in learning capabilities. Pooling layers are an example which we excluded for this reason in a lot experiments as the implications of their usage on expressivity is not properly covered, yet, as to our knowledge.

2. **Theoretical Alignment**: Second, following the findings of neural network theory on theoretical capabilities allows to more *selectively design experiments*. An exemplary selectively designed experiment by this theory is to check whether defined neural network sets and a chosen training scheme actually show signs of convergence from low to higher parameterisation sizes.

There exist a broad variety of UAPs. Two major distinctintions are to consider either *finite-depth infinite-width* or *infinite-depth finite-width* networks. Finite and infinite then refers to a formulation in which sets of finite-parameterised neural networks grow with respect to their width or depth such that close approximation under a specified mode of convergence can be achieved. For both approaches it is known that neural networks approximate relevant function classes. Concretely, it is well known that an ever-growing single hidden layered neural network suffices to approximate functions from $\mathbb{R}^{d_{in}}$ to $\mathbb{R}^{d_{out}}$. Further, we know that the minimum width $w_{min}$ of an infinite-depth neural network for $L^p$ approximation from $\mathbb{R}^{d_{in}}$ to $\mathbb{R}^{d_{out}}$ is $max\{d_{in}, d_{out}\}$ [30, 126, 216]. These results are mainly based on two approaches: *upper bounds* are usually provided by constructive techniques in which properties of the neural networks in the defined set are used to show that a given target function can be closely approximated. *Lower bounds* require tougher proof techniques and construct counter-examples of targets that can not be closely approximated under given constraints.

During the last years, minimal widths for infinite-depth networks have been found [30] such that research focus now turns towards understanding trade-offs between width and depth under various cases. Our motivation of determining relationships between structure and neural network performances or building methods to automatically construct structures highly aligns with these question of structural trade-offs and supposedly will require restrictions on topological properties of the input domain to derive further statements.

Studies on universal approximation theorems also give a better distinction between neural network realisations and neural architectures than empirical works on e.g. neural architecture search. For example, Yarotsky defines architectures as neural network realisations with unspecified weights. They "say that a network architecture is capable of expressing any function" from a particular function space "with error $\epsilon > 0$, meaning that this can be achieved by some weight assignment" [324]. It is, however, not always well-defined whether the neural network architecture is then defined based on a functional form, equipped with a norm, proven to be a function space or whether there exist transformations between the architecture and other function spaces.

REMARKS

- The field of universal approximation theorems is an important backbone for empirical considerations of this work.
- We deem it necessary to ground shared minimum or maximum learning capabilities as to make comparisons between structures of deep neural networks and therefore partially rely on UAPs.
- More rigorous formulations of structure, architecture and neural network realisation allow to better explain where NAS methods act upon

### 5.9.1 *Universal Approximation Through Infinite Width*

Early work dates back to the 1990s in which single-hidden-layered neural networks with sigmoidal activation functions were shown to approximate continuous functions with support in the unit hypercube $[0, 1]^{d_{in}}$ [43]. There were slightly different formulations for such UAPs based on the particular used functional form of a neural network as investigated by Hornik [106, 107], Barron [18, 19] or Cybenko [43].

Barron's work used "an average of the norm of the frequency vector weighted by the Fourier magnitude distribution [..] to measure the extent to which the function oscillates" [18, 133] which later led to deeper studies of Barron spaces [299]. Barron spaces are based on two-layered neural network forms [299, Eq. 2.1] and the underlying Barron norm $\|f\|_{\mathcal{B}_p} = \inf_{\rho}(\mathbb{E}_\rho[|a|^p(\|\mathbf{b}\|_1 + |c|)^p])^{1/p}$ [299, Eq. 3, p.4] allowed to bound the Rademacher complexity [21] for a set of functions with bounded Barron norm [299, Thm. 6, p.6] which led to the interpretation that "Barron spaces can be learned efficiently" [299, p.6].

Note, that Kratsios et al. showed "that if an architecture is capable of universal approximation, then modifying its final layer to produce binary values creates a new architecture capable of deterministically approximating any classifier" [137]. Results are therefore usually not directly studied for classification models.

The universal approximation theorems on classical feed-forward formulations of two-layered neural networks can be simplified to the following: if the functional form of a neural network with two layers is given and the parameterisation can be increased in width, the approximation error for a target function from many relevant function spaces can be bound with a negligible error $\epsilon > 0$. We then say that this functional form or neural network architecture has a universal approximation property. The property only guarantees the existence of a neural network realisation but not how it can be (efficiently) found.

### 5.9.2 *Universal Approximation and Minimum Width Through Infinite Depth*

As summarised in section 5.9.2, the upper and lower bounds for neural networks of finite width but infinite depth have been extensively studied in recent years. The difference to previously studied infinite-width neural networks is that the error bounds are now reduced with respect to a growing depth instead of a growing width. Vardi et al. conclude from [215] and own work [291] that "deep networks can memorise $N$ samples using roughly $\sqrt{N}$ parameters, while shallow networks require $N$ parameters" [292].

| Function Class | Activation | Upper / lower bounds | Source |
|---|---|---|---|
| $L^p(\mathcal{K}, \mathbb{R}^{d_{out}})$ | arbitrary | $w_{min} \geq max\{d_{in}, d_{out}\}$ | Cai [30] |
| $L^p(\mathcal{K}, \mathbb{R}^{d_{out}})$ | cont. nonpoly | $w_{min} \leq max\{d_{in}+2, d_{out}+1\}$ | Park et al. [216] |
| $L^p(\mathcal{K}, \mathbb{R}^{d_{out}})$ | LeakyReLU | $w_{min} = max\{d_{in}, d_{out}, 2\}$ | Cai [30] |
| $L^p(\mathcal{K}, \mathbb{R}^{d_{out}})$ | LeakyReLU+ABS | $w_{min} = max\{d_{in}, d_{out}\}$ | Cai [30] |
| $L^p(\mathbb{R}^{d_{in}}, \mathbb{R}^{d_{out}})$ | ReLU | $w_{min} = max\{d_{in}+1, d_{out}\}$ | Park et al. [216] |
| $L^p(\mathbb{R}^{d_{in}}, \mathbb{R}^{d_{out}})$ | ReLU | $w_{min} \leq d_{in} + d_{out}$ | Kidger & Lyons [126] |
| $L^1(\mathbb{R}^{d_{in}}, \mathbb{R})$ | ReLU | $d_{in}+1 \leq w_{min} \leq d_{in}+4$ | Lu et al. [172] |
| $L^1(\mathcal{K}, \mathbb{R})$ | ReLU | $w_{min} > d_{in}$ | Lu et al. [172] |
| $C^p(\mathcal{K}, \mathbb{R}^{d_{out}})$ | arbitrary | $w_{min} \geq max\{d_{in}, d_{out}\}$ | Cai [30] |
| $C^p(\mathcal{K}, \mathbb{R}^{d_{out}})$ | ReLU+FLOOR | $w_{min} = max\{d_{in}, d_{out}, 2\}$ | Cai [30] |
| $C^p(\mathcal{K}, \mathbb{R}^{d_{out}})$ | UOE+FLOOR | $w_{min} = max\{d_{in}, d_{out}\}$ | Cai [30] |
| $C(\mathcal{K}, \mathbb{R}^{d_{out}})$ | nonaffine poly. | $w_{min} \leq d_{in} + d_{out} + 2$ | Kidger & Lyons [126] |
| $C(\mathcal{K}, \mathbb{R}^{d_{out}})$ | cont. nonpoly. | $w_{min} \leq d_{in} + d_{out} + 1$ | Kidger & Lyons [126] |
| $C(\mathcal{K}, \mathbb{R}^{d_{out}})$ | ReLU | $d_{in}+1 \leq w_{min} \leq d_{in}+d_{out}$ | Hanin & Sellke [90] |
| $C(\mathcal{K}, \mathbb{R}^{d_{out}})$ | ReLU+Step | $w_{min} = max\{d_{in}+1, d_{out}\}$ | Park et al. [216] |
| $C(\mathcal{K}, \mathbb{R}^{d_{out}})$ | unif. cont. | $w_{min} \geq d_{out} + 1_{d_{in}<d_{out}<2d_{in}}$ | Kim et al. [127] |
| $C(\mathcal{K}, \mathbb{R}^{d_{out}})$ | cont. nonpoly. | $w_{min} \leq max(2d_{in}+1, d_{out}) + 2$ | Hwang [110] |
| $C(\mathcal{K}, \mathbb{R})$ | unif. cont. | $w_{min} > d_{in} + 1$ | Johnson [121] |
| $C([0,1], \mathbb{R}^{d_{out}})$ | UOE | $w_{min} = d_{out}$ | Cai [30] |
| $C([0,1], \mathbb{R}^2)$ | ReLU | $w_{min} = 3 > max\{d_{in}+1, d_{out}\}$ | Park et al. [216] |

**Table 5.2:** Summary of Park et al. and Cai [30, 216] on universal approximation properties for various function classes with neural networks of finite width but infinite depth.

These studies on infinite-depth networks also provided statements about minimum widths. If networks fall below such widths in certain parts, these parts can be considered as *information bottlenecks*. According to Lu et al., all functions $f$ from $R^{d_{in}}$ to $\mathbb{R}^{d_{out}}$ "(except for a negligible set) cannot be approximated by any ReLU network whose width is no more than $d_{in}$" [172].

### 5.9.3 *Universal Property for Convolutional Neural Networks*

Similar results of UAPs have been also established for architectures with layers such as convolutional neural networks. For example, Hwang & Kang study CNNs which "maintain the original shape of the input data throughout the output data" and prove universal approximation theorems for such cases [111] and w.r.t. an input dimension $d \in \mathbb{N}$. According to Hwang & Kang "CNNs require a much deeper minimum depth" as compared to multi-layered perceptrons "which only needs a two-layered network to get universal property" and argues that "this is because the receptive field, the range of the input component which affects the specific output component, is restricted by the convolution using the kernel" [111, Sec. 4.3]. They conclude that "for a CNN with kernel size three to have universal property, at least $d - 1$ layers are required." [111, Sec. 4.3] if the

There also exist alternative works such as of Liu et al. who also studied Topelitz matrices underlying the constraints of CNN-based matrix-multiplications (compare e.g. Ye et al. [325]). Liu et al. found "a consequence of [their] Toeplitz result is a fixed-width universal approximation theorem for convolutional neural networks, which so far have only arbitrary width versions" [168]

With the results on universal approximation theorems, we can state

REMARKS

- Many architectural designs of deep neural networks allow for universal approximation if sufficient parametric capacity w.r.t. at least the input and output dimensionalities are provided.
- The results on UAPs for convolutional neural networks show that the structure, i.e. a constraint connectivity between consecutive layers, has an impact on whether universal approximation on common function classes is theoretically possible.
- We subsequently will use the observation of sufficient parametric capacity for universal approximation as a prerequisite of a construction for architectures that assume to have a minimal approximation capability by providing this capacity. By this prerequisite we enforce to search for a reason of differences between architectures in structure instead of finding a reason for the difference in non-sufficient approximation capability.

### 5.9.4 *A Note on Other Notions of Architectural Expressiveness*

A universal approximation property can be considered as a notion of architectural expressiveness. The property guarantees the theoretical existence of a neural network realisation which should yield an arbitrarily desirable small training error. Under this consideration, different functional forms based on different structures should all contain equivalently good solutions for a given problem. The notion of a universal approximation property is therefore a lower bound for the architectural expressiveness.

Notions of architectural expressiveness serve two ideas of bounding the targeted structure analysis of deep neural networks:

Lower bounds of architectural expressiveness guarantee that the **reason** for consistent empirical differences between different architectures are not explained by the employed restriction of the analysed architectures but **might** be explainable by structural differences or other nuances in defining different sets of functions with that expressiveness.

Upper bounds of architectural expressiveness separate stronger in the sense that there exists a definite difference between architectures of different expressiveness. We suppose that computability poses an example for such a bound if arguments are shown that deep neural networks with our used definitions of being functions in $\mathbb{R}^{d_{in}} \to \mathbb{R}^{d_{out}}$ are not Turing complete but other neural architectures – such as recurrent neural networks [136] or attentional neural networks [221] or transformer neural networks [25, 222] – are. According to Siegelmann & Sontag certain forms of *rational* neural networks can simulate any Turing machine [256] but the definitions diverge from our notion of deep neural networks.

We mention this notion of architectural expressiveness for the sake of completeness but do not further study them. Instead, the universal approximation theorems serve as argument for our subsequent work and as base for theoretical known limits.

Part III

# GRAPH INDUCED NEURAL NETWORKS

The first research complex in section 1.1.4 is concerned with the formalism of neural networks, search problems that can be formulated with it and advantages & disadvantages of this kind of formalism. This part contains answers to this complex by contributing a construction of neural networks from directed acyclic graphs, and uncovers assumptions and ideas connected to our definitions.

**Research Complex I: Formalism**

From section 1.1.4:

- How can the structure of deep neural networks be defined such that structural optimisation problems can be formulated?
- Are graphs a suitable representation for neural network structure?
- What are the difficulties, advantages and disadvantages of the proposed formalism?

# 6 GRAPH-INDUCED NEURAL NETWORKS

*The following entails:*

We present a formal framework to induce sets of deep neural networks based on directed acyclic graphs. The intention and benefits of this construction are threefold:

> *First*, the structure of deep neural networks (DNN) is made explicit. We present a detailed construction of a neural network based on a graph, and of a neural architecture based on sets of graphs. Many of these constructions even provide universal architectures which are provably a superset of universal approximators.

> *Second*, a unified language for neural architecture search techniques such as pruning or evolutionary searches reveals similarities and differences between these techniques more easily.

> *Third*, theoretical properties can be properly derived for different families of techniques and a constructive framework advances future formulations of a more unified and powerful theory for deep learning architectures.

In a constructive way, the framework formulates sets of neural networks built from graphs. This formulation allows to treat the search space of a neural architecture search problem as a set of graphs. Multiple separate components that lead to an analytically useful formulation of neural networks are unveiled through such a formulation. It is well established to emphasise on search space design, the search method and the performance estimation technique as the three major pillars in neural architecture search [61, p. 2, Fig. 1]. This framework follows these three pillars but further sheds light on two formal necessities: First, the expressivity of sets of neural networks is restricted to adhere to theoretical properties such as universal approximation properties. Second, the search space design is restricted to an established mathematical structure that allows for rich expressivity but also adheres to formality. The constructive aspect of the framework can be considered as a ***synthesis*** *of structural priors of deep neural networks*.

In a forgetful way, the framework allows for a reduction of neural networks into their underlying structure. Such an analysis has been investigated over recent years and has both biological and technical motivation. The biological motivation boils down to the empirical observation that biological neural networks, as well as other complex networks, follow distinct structural patterns. The technical motivation is, that if such structural patterns are identified, the design of a deep neural network or the search for it within a larger search space can be guided in an informed way. Examples for that are general structural design patterns, trade-offs between different objectives such as inference time, energy consumption, performance or robustness of models, and design space optimisations for faster searches. These analytical approaches are summed up under the term ***analyses*** *of structural priors of deep neural networks*.

## 6.1 TRANSFORMING A DIRECTED ACYCLIC GRAPH INTO A DEEP NEURAL NETWORK

We describe the construction $\mathcal{A}$ of a deep neural network $f \in \mathcal{A}^{d_{in} \to d_{out}}(G)$ based on a directed acyclic graph (DAG) $G$ and input and output dimensions $d_{in}, d_{out} \in \mathbb{N}$.

The guiding theme of this framework is the observation with universal approximation theorems, that sets of neural networks need to be constructed with a manner of infinite growth in order to be able to approximate a given target function sufficiently close. By this and some further constructive details, it can be usually guaranteed that there always exists another neural network realisation that ensures to be close to a given target function with respect to some difference – a convergence criterium can be formulated. Originally [43], this growth has been implemented by an ever growing single layer. This means that a set of neural networks contains fixed-width realisations with at least a minimum width of neurons but for an ever smaller difference $\epsilon > 0$, there is always a realisation of a neural network found with a fixed width of at least $O(\frac{1}{\epsilon})$ neurons.

The constructed set of neural networks is thus always infinite in size and, analytically optimal, we would even favor a construction such that the set is a function space. The latter, however, is assumably not the case, compare i.e. Petersen et al. [223, 224].

To transform a set of directed acyclic graphs (DAGs) into a set of neural networks, a transformational mapping is required. This mapping takes a directed acyclic graph and maps it to a neural network architecture and can already yield an infinite set of architectures in its own but does not necessarily need to. A neural network architecture is a finite-parameterised model of a distinct structure and will now be properly defined.

Consider a directed acyclic graph $G$, and input and output dimensions $d_{in}, d_{out} \in \mathbb{N}$. An architecture $\mathcal{A}^{d_{in} \to d_{out}}(G) : \mathbb{R}^{d_{in}} \to \mathbb{R}^{d_{out}}$ is induced from $G$ by treating vertices of the graph as neurons, and treating edges of the graph as connections with weight parameters in the neural network. To feed input $\mathbf{x}$ into a realisation $f \in \mathcal{A}^{d_{in} \to d_{out}}(G)$ each computational step requires a constructive explanation. We provide this construction by following the layering of the DAG $G$ and outline how $G$ transforms into a layered neural network.

### *Layering of a DAG*

A topological sorting of $G$ is required to determine the source and sink vertices of the graph: $source(G) = \{v \in V(G) \mid |Nei_G^{in}(v)| = 0\}$ and $sink(G) = \{v \in V(G) \mid |Nei_G^{out}(v)| = 0\}$. The topological sorting of $G$ further provides us with a *layering* such that each vertex and its corresponding neuron only precede vertices or neurons from a larger layer. A vertices' layer $layer_G : V(G) \to \mathbb{N}$ is given as $layer_G(v) = max(\{layer_G(s) \mid s \in Nei_G^{in}(v)\} \cup \{0\}) + 1$. Follow fig. 6.1 on the following page for an illustration of this layering.

For the DAG in step one on the left of Figure 6.1, multiple topological sortings can be found. Some of them are denoted below the DAG with e.g. "4 2 5 1 6 3 7" and are implicitly obtained if $layer_G(v)$ is recursively computed, e.g. with a breadth-first-search through the graph. In the second step, the source vertices "4" and "2" are assigned to layer zero as the set $\{layer_G(s) \mid s \in Nei_G^{in}(v)\} \cup \{0\}$ collapses to $\{0\}$ because $Nei_G^{in}(v)$ is empty, e.g. there are no incoming sources to these source vertices. The breadth-first-search reaches vertices "5", "6" and "3" and for "5" and "6" the layers of vertices from incoming edges are already obtained, such that the expression of $layer_G(5)$ for example gets to be $max(\{layer_G(2), layer_G(4)\} \cup \{0\}) = max(\{1, 0\}) + 1 = 2$. A layer number for "3" is not known, yet, up until to this step, such that the algorithm continues with children of vertices with known layers. The algorithm recursively fills up all layer indices for preceding vertices until the last layer index can be computed. One topological sorting is automatically obtained during the process and is a selection of a random order of vertices from each successive layer.

We write $\mathcal{L}(G) = \{1, \dots, L\}$ for the ordered layer of $G$ with $L$ being the last layer. The layer $layer_G$ can omit the symbol $G$ if the context makes it clear and due to its analogy we will overload $layer(\cdot)$ as to refer to the layer of a neuron.

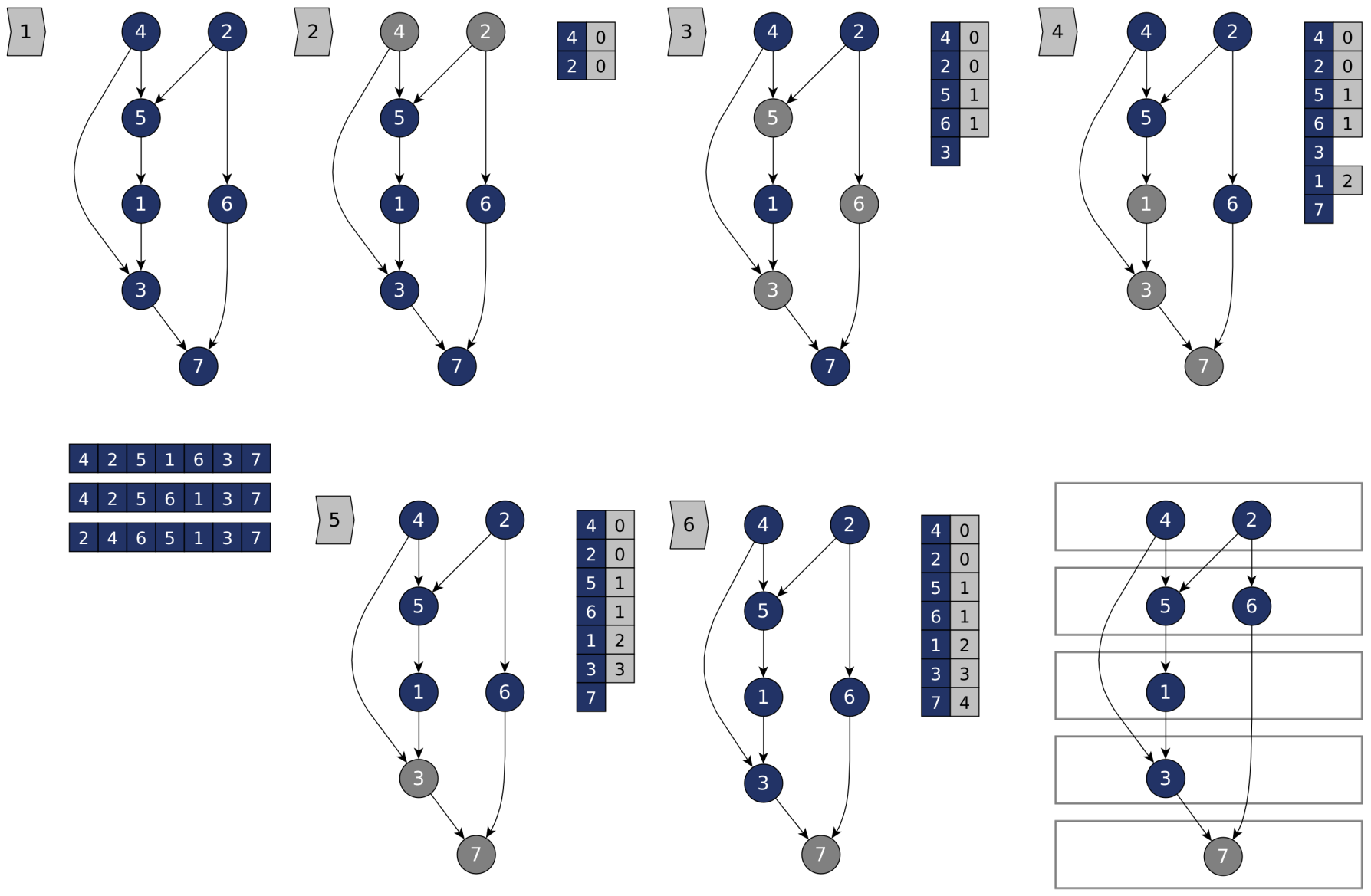

**Figure 6.1:** The layering of a directed acyclic graph can be obtained while traversing the graph for one topological sorting. Different topological sortings are possible as indicated below step one with "4 2 5 1 6 3 7", "4 2 5 6 1 3 7", and also "2 4 6 5 1 3 7".

*Feeding Input Through a Graph-Induced Neural Network*

An input vector $\mathbf{x} \in \mathbb{R}^{d_{in}}$ is mapped to the first layer in an at most affine transformation. We refer to this step as the *entrance* of the construction: $entr : \mathbb{R}^{d_{in}} \to \mathbb{R}^{|source(G)|}$. The restriction to an affine transformation is chosen as to pass the information of $\mathbf{x}$ into the architecture construction without adding or removing too much expression capabilities.

As common in literature, the subsequent processing core is defined in a recursive manner. Compare the section on architecture design of Goodfellow et al. in [79, Chp. 6.4] in which they define the output of a first layer as $h^{(1)} = g^{(1)}\left(W^{(1)\top}x + b^{(1)}\right)$ and let the reader infer the subsequent definitions [79, p. 197]. Peterson [223, p. 2, eq. 2.2] provides a proper recursive definition in the following form:

$$\begin{aligned} x_0 &:= x, \\ x_l &:= \sigma(A_l x_{l-1} + b_l) \quad \text{for } l = 1, \dots, L-1 \\ x_L &:= A_L x_{L-1} + b_L \end{aligned} \tag{6.1}$$

The referenced equations are to be taken decoupled of the notation in this work but should show parallels to our formulation.

For the recursive definition, layer-wise transformations from all preceeding layers are required, capturing not only the output of the previous layer in a transformation but outputs of all previous layers. A layer $l \in \mathcal{L}(G)$ contains the vertices $V^l(G) = \{v \mid v \in V(G) \wedge layer_G(v) = l\}$. The vertices in a layer of $G$ can be brought into a similar layer-wise affine transformation as in eq. (6.1). By means of a permutation $\pi : V(G) \rightarrow \{1, \ldots, |V(G)|\}$, we can equip each vertex with a neuron $j \in \{1, \ldots, |V(G)|\}$. The permutation induces a layer-wise order $\pi_l : V^l(G) \rightarrow \{1, \ldots, |V^l(G)|\}$ of vertices such that each neuron can be equipped with a bias value which are summarised into a layer-wise bias vector $B^l \in \mathbb{R}^{|V^l(G)|}$.

For each preceeding layer $s \in \{1, \ldots, l-1\}$ of $l$, we define a weight matrix $W^{s \rightarrow l} \in \mathbb{R}^{|V^s(G)| \times |V^l(G)|}$. Each weight $w_{ij} \in W^{s \rightarrow l}$ is set to 0 iff $(\pi_s^{-1}(i), \pi_l^{-1}(j)) \notin E(G)$ and otherwise left open to be a learnable parameter of the deep neural network. We refer with $\mathcal{D}$ to a distribution[1] from which the weight and bias values are initially sampled. This could be a joint-distribution per layer or an individual (independent) distribution per weight or also be different between weights $w_{ij}$ and biases $B_i^l$. We leave studies on the choice of such distributions to research on initial value problems and focus on enforcing the connectivity of $G$ through explicit zeros in the parameters of the DNN here.

$$\begin{aligned} w_{ij} &\triangleq \begin{cases} 0 & \text{iff } (\pi_s^{-1}(i), \pi_l^{-1}(j)) \notin E(G), \\ \sim \mathcal{D} & \text{otherwise} \end{cases}, \\ B_i^l &\triangleq \begin{cases} 0 & \text{iff } |Nei_G^{in}(\pi_s^{-1}(i))| = 0, \\ \sim \mathcal{D} & \text{otherwise} \end{cases} \end{aligned} \tag{6.2}$$

Importantly, eq. (6.2) defines exact zeros in both the weights & biases such that the detailled connectivity of $G$ is reflected in the computation of the resulting deep neural network.

We can now recursively define the core inference mechanism of the architecture with

$$\begin{aligned} \mathbf{z}^{(0)} &\triangleq entr(x), \ \mathbf{h}^{(0)} \triangleq \mathbf{z}^{(0)} \\ \mathbf{z}^{(l)} &\triangleq \sum_{s=0}^{l-1} W^{s \rightarrow l} \mathbf{h}^{(s)} + B^{(l)}, \\ &\text{with } \mathbf{h}^{(l)} = \sigma(\mathbf{z}^{(l)}) \qquad \text{for } l \in \mathcal{L} \end{aligned} \tag{6.3}$$

The symbol $\mathbf{z}$ refers to layer-wise pre- and the symbol $\mathbf{h}$ to layer-wise post-activations. Post-activations are subsequently often referred with $\mathbf{h}$ and contain vector representations of the learned non-linear transformation up until to a certain point in the computation of a deep neural network. The last post-activation $\mathbf{h}^{(L)} \in \mathbb{R}^{|sink(G)|}$ is passed through an at most affine transformation $exit : \mathbb{R}^{|sink(G)|} \rightarrow \mathbb{R}^{d_{out}}$ such that the architecture induced by $G$ computes $\mathcal{A}^{d_{in} \rightarrow d_{out}}(G)(\mathbf{x}) \triangleq \mathbf{x} \mapsto exit(\mathbf{h}^{(L)})$.

Observe the similarity between eq. (6.1) on the facing page and eq. (6.3) in which the deep neural network is defined recursively over finitely many layers $\mathcal{L}$. The major difference lies in $l-1$ many matrix-vector multiplications from all previous layer outputs instead of a single

1 Compare section 5.2.1 on page 53 on exemplary distributions for defining initial values before training the DNN i.e. with maximum likelihood estimation and backpropagation.

transformation. By applying a non-linear activation function, different layers are properly separated and could be otherwise combined into a single transformation. The bare definition of eq. (6.3) is a densely connected network[2] and structurally restricted by $G$. Alternatively, the operations could be directly defined based on vertices and edges of $G$ but would less resemble the parallelisation capabilities of matrix multiplications.

Similar to the DenseNet architecture of Huang et al. [109], eq. (6.3) carries up to $\frac{L(L+1)}{2}$ many weight matrices $W^{s \to l}$ of varying dimensionalities depending on layer-wise cardinalities $V^l(G)$. For many $G$, this can be optimised by e.g. dropping certain matrices $W^{s \to l}$ based on whether there exists a connectivity between layers $s$ and $l$ or by only keeping dense matrices by having the same number of vertices for layers $s$ and $l$. Therefore, the structure of $G$ already has direct implications for the parallisability or memory footprint of resulting DNN realisations.

An alternative and algorithmic description of the outlined construction is provided in algorithm 4 on the next page. The main difference of both definitions is that eq. (6.3) on the preceding page recursively defines the forward transformation of the described deep neural network while algorithm 4 on the next page describes a functional construction of the DNN itself. The inferencing type of eq. (6.3) is significantly more compact to define and also used by Peterson [223, p. 2, eq. 2.2] or Goodfellow et al. [79, p. 197] while algorithm 4 more closely aligns with a technical implementation as e.g. in LIB DeepStruct based on LIB PyTorch.

Note the following minor construction details:

1. The non-linearity $\sigma$ is the important element to discretise a deep neural network into layers and even a graph structure. It also plays an important role for the graph-to-network direction. Without non-linear elements, structural components in the same layer would sucessively join to linear transformations and layers would collapse into linear functions.

2. Equation 6.3 allows for skip-layer connections if the directed acyclic graph contains edges across layers. Such connections are often called residual connections. Although not considered much in neural network theory, they have been very successfully employed empirically in image recognition with ResNets [95] and U-Nets [238].

3. Details of non-linearities $\sigma$ are left out although their choice makes a crucial difference in practice. Non-linearities could be chosen

2 Densely connecting networks have technically been previously proposed e.g. by Huang et al. [109] for convolution-based architectures. These studies were more focussed on technical issues such as realizing matrix multiplications of the growing same-sized layers efficiently rather than questions on structural connectivity.

**Algorithm 4:** Construction of a graph-induced deep neural network $\mathcal{A}(G)^{d_{in} \to d_{out}}$ given a single directed acyclic graph $G$ and input and output dimensions $d_{in}, d_{out} \in \mathbb{N}$. Intermediate layer-wise functions $f_l$ in e.g. line 6 demonstrate the layer-wise recursive nature of computing a resulting representation based on all previous hidden representations. The bare structure resembles a densely connected network across all layers in which it is important that the weights & biases in $\theta(f)$ are initialised according to eq. (6.2) on page 105 such that a detailled sparse structure of $G$ is reflected.

**input** : A DAG $G$, dimensions $d_{in}, d_{out} \in \mathbb{N}$, activation function $\sigma$
**output**: An (untrained) neural network $f \in \mathbb{R}^{d_{in}} \to \mathbb{R}^{d_{out}}$ with parameters in $\theta(f)$

`// Compute layering of` $G$

1 $layer_G : V(G) \to \mathbb{N}; L(G) \leftarrow \{1, \dots, L\}$ with $L := \max_{i \in V(G)} layer_G(i)$;

`// Gathering transformation for entrance`

2 $\left(f_0 : \mathbb{R}^{d_{in}} \to \mathbb{R}^{|source(G)|}\right) \leftarrow (\mathbf{x} \mapsto W_{entr}\mathbf{x} + B_{entr})$;

3 $\left(p_0 : \mathbb{R}^{d_{in}} \to \mathbb{R}^{|source(G)|}\right) \leftarrow$ identity;

4 $\left(g_0 : \mathbb{R}^{d_{in}} \to \mathbb{R}^{|source(G)|}\right) \leftarrow$ identity;

`// Compose Layer-wise Functions`

5 **for** $l \in L$ **do**

  `//` $f_l$ `computes pre-activations` $\mathbf{z}^{(l)}$

6   $\left(f_l : (\mathbb{R}^{|source(G)|}, \mathbb{R}^{|V^1(G)|}, \dots, \mathbb{R}^{|V^{l-1}(G)|}) \to \mathbb{R}^{|V^l(G)|}\right) \leftarrow \left(\mathbf{h}^0, \dots, \mathbf{h}^{(l-1)} \mapsto \sum_{s=0}^{l-1} W^{s \to l}\mathbf{h}^{(l-1)} + B^{(l)}\right)$;

  `// Overload` $\theta(\cdot)$ `as to capture layer parameters`

7   $\theta(f_l) \leftarrow \{W^{s \to l} \mid \forall s \in \{0, \dots, l-1\}\} \cup \{B^{(l)}\}$;

  `//` $p_l$ `computes post-activations` $\mathbf{h}^{(l)}$

8   $\left(p_l : \mathbb{R}^{|V^l(G)|} \to \mathbb{R}^{|V^l(G)|}\right) \leftarrow (\mathbf{z} \mapsto \sigma(\mathbf{z}))$;

  `// Gathering layer-wise sequence of post-activations as inputs for next layer`

9   $\left(g_l : \mathbb{R}^{d_{in}} \to (\mathbb{R}^{|source(G)|}, \mathbb{R}^{|V^1(G)|}, \dots, \mathbb{R}^{|V^{l-1}(G)|})\right) \leftarrow (\mathbf{x} \mapsto (p_s(f_s(g_s(\mathbf{x}))))_{s \in \{1, \dots, l-1\}})$;

10 **end**

11 $\theta(f) \leftarrow \{W_{entr}, B_{entr}, W_{exit}, B_{exit}\} \cup \{\theta \in \theta(f_l) \mid \forall l \in L\}$;

12 $(f : \mathbb{R}^{d_{in}} \to \mathbb{R}^{d_{out}}) \leftarrow (\mathbf{x} \mapsto W_{exit} f_L(g_L(\mathbf{x})) + B_{exit})$;

13 **return** $f, \theta(f)$

uniformly for all architectures, layer-wise or even neuron-wise. As we are more interested in properties of $G$ for structure analysis, we leave out further notation on $\sigma$ for now.

4. We implicitly set the entrance and exit to be an identity mapping as to not loose information. The number of source vertices later will be required to be at least of size of the input dimensionality as for the resulting neural networks to be capable of universal approximation. For more source vertices, the linear mapping could pass the information in redudant ways or even be learned. However, we consider the choice of the entrance and exit mappings to be a minor technical detail.

We write $f \in \mathcal{A}^{d_1 \to d_2}(G)$ with $f : \mathbb{R}^{d_1} \to \mathbb{R}^{d_2}$ for a realised function of that architecture. The architecture has finite parameters and one advantage is that several properties of the deep neural network are now characterised by the directed acyclic graph $G$.

We further overload the symbol $\mathcal{A}$ for a set of DAGs $\mathcal{G}$ and obtain $\mathcal{A}^{d_1 \to d_2}(\mathcal{G}) = \bigcup_{G \in \mathcal{G}} \mathcal{A}(G)^{d_1 \to d_2}$. This has practical implications of the set $\mathcal{A}$ towards fulfilling properties of universal approximation theorems by considering infinite sets of finitely parameterised neural networks.

**Definition 1** *A directed acyclic graph $G$ induces an architecture $\mathcal{A}^{d_{in} \to d_{out}}(G)$ through an architectural induction transformation $\mathcal{A}^{d_{in} \to d_{out}}$ by means of the outlined construction in algorithm 4 on the preceding page. The symbol $\mathcal{A}$ is overloaded as to induce an architecture $\mathcal{A}^{d_1 \to d_2}(\mathcal{G}) = \bigcup_{G \in \mathcal{G}} \mathcal{A}(G)^{d_1 \to d_2}$ for a set of DAGs $\mathcal{G}$.*

An architecture can contain finitely or infinitely many neural networks with input and output dimensions $d_1$ and $d_2$. Each neural network in itself has a finite amount of parameters. An architecture therefore contains many functions with finite or infinitely many graph structures $G$ while a neural network realisation $f$ is of finite structure and can theoretically take any parameterisation for its weights & biases. This is one of the reasons why the terms *neural architecture* and *neural network* are often used interchangeably.

We will subsequently motivate and define the term of a universal architecture which necessarily contain infinitely many neural network realisations as to align with universal approximation theorems. Furher, constructing a neural architecture can then be controlled by specifying the set of DAGs $\mathcal{G}$ implicitly or explicitly by e.g. defining an equivalence class over graphs, constructing graphs based on basic operations or rules, or defining a probability distribution over sets of graphs.

## 6.2 OPERATIONS IN GRAPH SPACE

Operations in graph space affect the induced architectures. Graph operations are insofar interesting that **1st/** genetic algorithms make use of them in neural architecture search, and **2nd/** formal constructions of more complex architectures make us of them.

In the proposed framework, the genetic encoding of such a neural architecture search is a graph. The assumption in a genetic algorithm, especially over an evolutionary algorithm, is that the operations in the genetic encoding space improve the overall search procedure. That would require understanding of which operations in graph space change the induced architecture such that the neural architecture search is improved with respect to a desired objective (e.g. improving accuracy). While evolutionary algorithms only make use of populations from which they conduct selection, re-sampling and possibly mutation, genetic algorithms apply n-ary variation operators (more on that in section 12.3 on page 223). Graph operations include e.g.

- **unary operations**: random or sophisticated edge or vertex mutations, vertex collapsing, edge collapsing,
- **binary operations**: concatenation, parallelisation, sub-graph insertion,
- **n-ary operations**: merging of multiple graphs, conseuctive concatenation, and many more.

An interesting and promising open research field is to learn operations in graph space implicitly during a genetic neural architecture search.

Two simple operations, concatenation and parallelisation, for architectural graphs can be defined and are depicted in Figure 6.2 and Figure 6.3, respectively. These have analogy to the definitions of Petersen et al. [223] with operations for matrix-vector tuple neural networks.

Two directed acyclic graphs $G_1$ and $G_2$ can be concatenated by fully connecting all sink vertices of $G_2$ with all source vertices of $G_1$. The concatenation for graphs can be expressed with $G_1 \bullet G_2$. For $f_1 \in \mathcal{A}^{d_1 \to d_2}(G_1)$ and $f_2 \in \mathcal{A}^{d_2 \to d_3}(G_1)$, we can easily find a $g \in \mathcal{A}^{d_1 \to d_3}(G_1 \bullet G_2)$ such that $f_1(f_2(\mathbf{x})) = g(\mathbf{x})$.

Parallelisation allows to work with unconnected directed acyclic graphs: two directed acyclic graphs $G_1$ and $G_2$ can be parallelised into the only two connected directed acyclic graph components of $G := G_1 \parallel G_2$. For $f_1 \in \mathcal{A}^{d_1 \to d_2}(G_1)$ and $f_2 \in \mathcal{A}^{d_1 \to d_3}(G_1)$, we can let a $g \in \mathcal{A}^{d_1 \to d_2 + d_3}(G_1 \parallel G_2)$ such that $f_1(x) \oplus f_2(x) = g(x)$.

Both operations, concatenation and parallelisation, are motivated by theoretical research [223, 224] as they can be explicitly defined and used in analytical and constructive approaches to the expressiveness of deep neural networks.

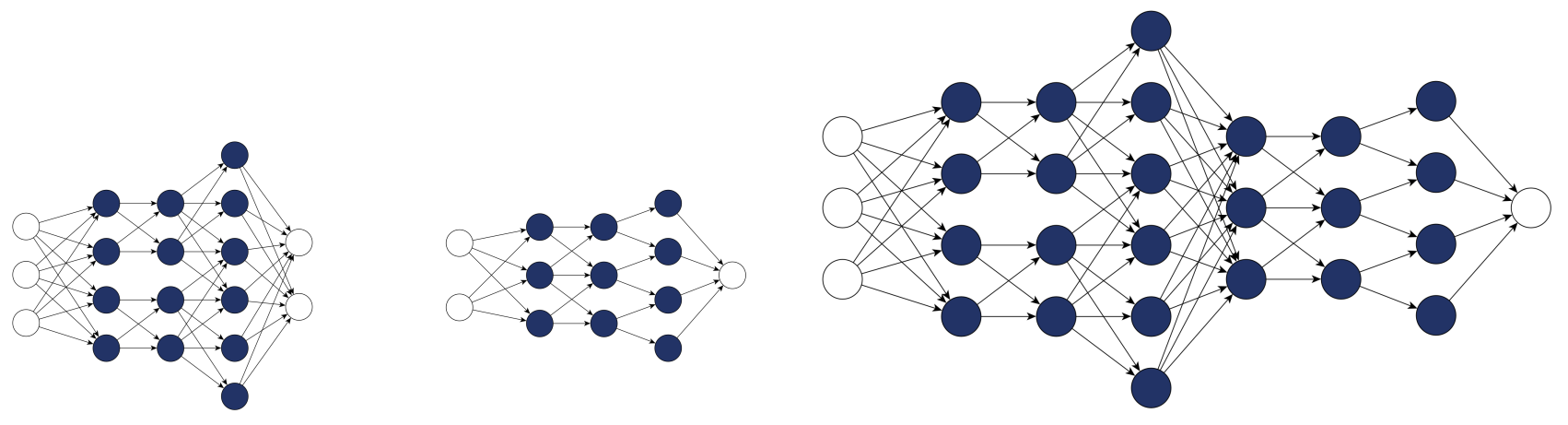

(a) Two DAGs integrated into neural network architectures $\mathcal{A}^{d_1 \to d_2}$ and $\mathcal{A}^{d_3 \to d_4}$.

(b) Concatenation of DAGs from (a) into an architecture $\mathcal{A}^{d_1 \to d_4}$.

**Figure 6.2:** The concatenation, e.g. expressed as $G_1 \bullet G_2$, is visualised based on DAGs from (a) and an architecture $\mathcal{A}^{d_1 \to d_4}$ is obtained. Note, that the input dimension of the first architecture and the output dimension of the second dimension are retained.

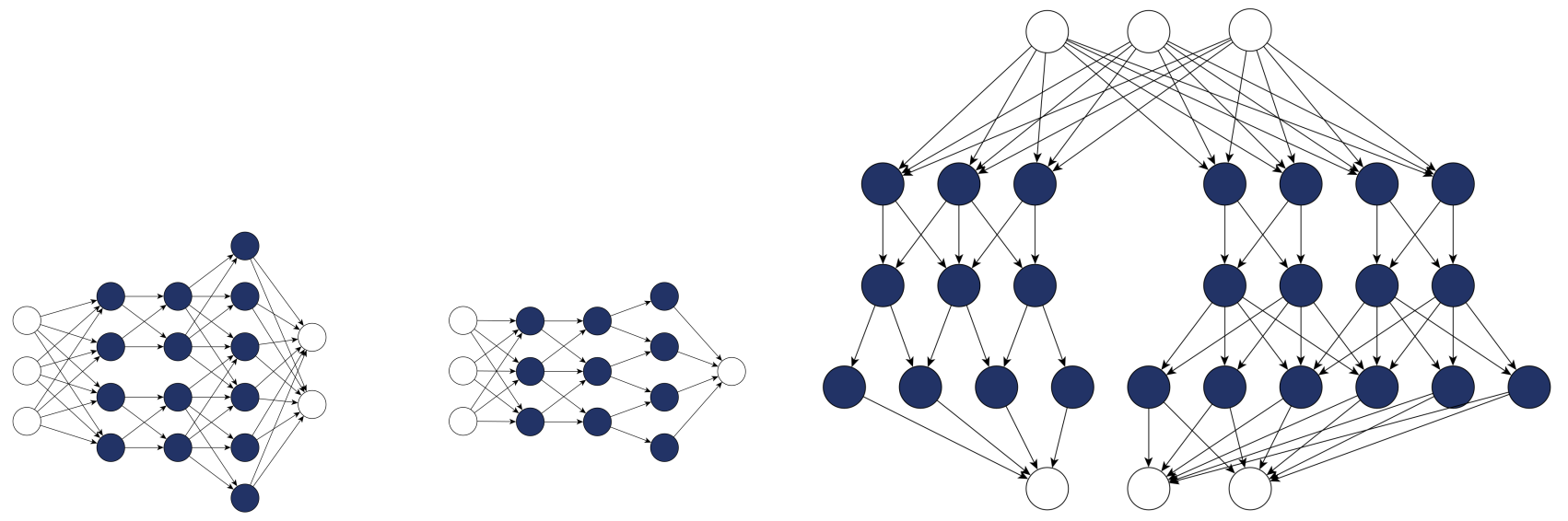

(a) Two DAGs integrated into neural network architectures $\mathcal{A}^{d_5 \to d_6}$ and $\mathcal{A}^{d_5 \to d_7}$.

(b) Parallelisation of DAGs from (b) into an architecture $\mathcal{A}^{d_5 \to d_6 + d_7}$.

**Figure 6.3:** Parallelisation $G_1 \parallel G_2$ (right) induces a new architecture from two DAGs (left) for which the input dimensions are retained but the output dimensions are added up.

## 6.3 GRAPH-INDUCED NETWORKS WITH UNIVERSAL APPROXIMATION PROPERTIES

We are now interested in unions of sets of neural network architectures which provide UAPs. This enables us to argue about two or more architectures such as $\mathcal{A}^{d_1 \to d_2}(\mathcal{G}_1)$ and $\mathcal{A}^{d_1 \to d_2}(\mathcal{G}_2)$ over infinite sets of graphs $\mathcal{G}_1$ and $\mathcal{G}_2$ for which both architectures guarantee universal approximation capabilities. In addition, we can formulate a question such as "how do these *universal* architectures differ (structurally)?".

### 6.3.1 *Representations of Graph Equivalence Classes*

To this end, we need to look at architectures of infinitely many neural networks. We take an abstraction level of DAGs to specify how the underlying set $\mathcal{A}$ grows. Similar to representations of equivalence classes in general, we reduce the representation of an infinite set of directed acyclic graphs to a single symbolic representation. The equivalence class can be considered a search space when finding a good approximation of a target function and the space of all equivalence class definitions as a higher-order search space design exploration. Directed acyclic graphs are therefore chosen for both the construction of finite-parameterised realisations (concrete model) of an architecture as well as for the representation or description of architectures.

Consider the DAG $G_1 :=$ "$\cdot \to \square \to \cdot$" of order three and think of the sources and sinks as a placeholder for input and output dimensions of an architecture $\mathcal{A}^{d_1 \to d_2}$. The square $\square$ should now depict that it can be replaced by any number of at least $d_1 \in \mathbb{N}$ vertices such that the graph becomes a representation of infinitely many graphs in the style of $\{\cdot \to \cdot \to \cdot, \cdot \rightrightarrows\!:\!\rightrightarrows \cdot, \dots\}$, e.g. $\{G \in \mathcal{DAG} \mid G \sim G_1\}$ containing all DAGs with one hidden layer and arbitrary but at least $d_1$ vertices in that layer.

As we treated the inner vertices different from source and sink vertices by letting only inner vertices grow, we actually used an equivalence class definition over labelled directed acyclic graphs to specify an infinite set of DAGs and can continue to do so. This approach allows very open definitions of sets of graphs through an equivalence relation $\sim$. The equivalence class is then given as $[G_1]$.

**Corollary 1** *A directed acyclic graph G is an equivalence class representation of a set of DAGs if it is one of its elements with smallest order. The equivalence class is denoted as $[G]$ and implies the definition of an equivalence relation $\sim_{[G]}$.*

The equivalence class $[G]$ can also contain labelled information. Labels could e.g. include kernel sizes of convolutional layers which then translate into particular sparsity patterns in sets of directed acyclic graphs. Note, that $\mathcal{A}([G])$ is an architecture induced from an infinitely large set of DAGs.

The construction of sets of graphs is particular tricky as it decides a lot about which neural network functions are grouped into an architecture. Figure 6.4 shows a simplified example. In the top row of Figure 6.4, single-layered neural networks with theoretically infinite width are depicted. The bottom row shows feed-forward networks with three layers and skip-layer connections between the first and last hidden layer. An implicit equivalence class definition transforms each representation on the left into the set of graphs of which some examples are shown on the right.

One could also use vertex labels for the graph equivalence class representations to make more complicated constructions. Each vertex of the equivalence class representation could unfold a pattern such as representing always two vertices in the graph in the underlying set. The possibilities are limitless but concrete definitions are necessary when comparing e.g. two or more representations and the resulting sets of graphs from which architectures are induced.

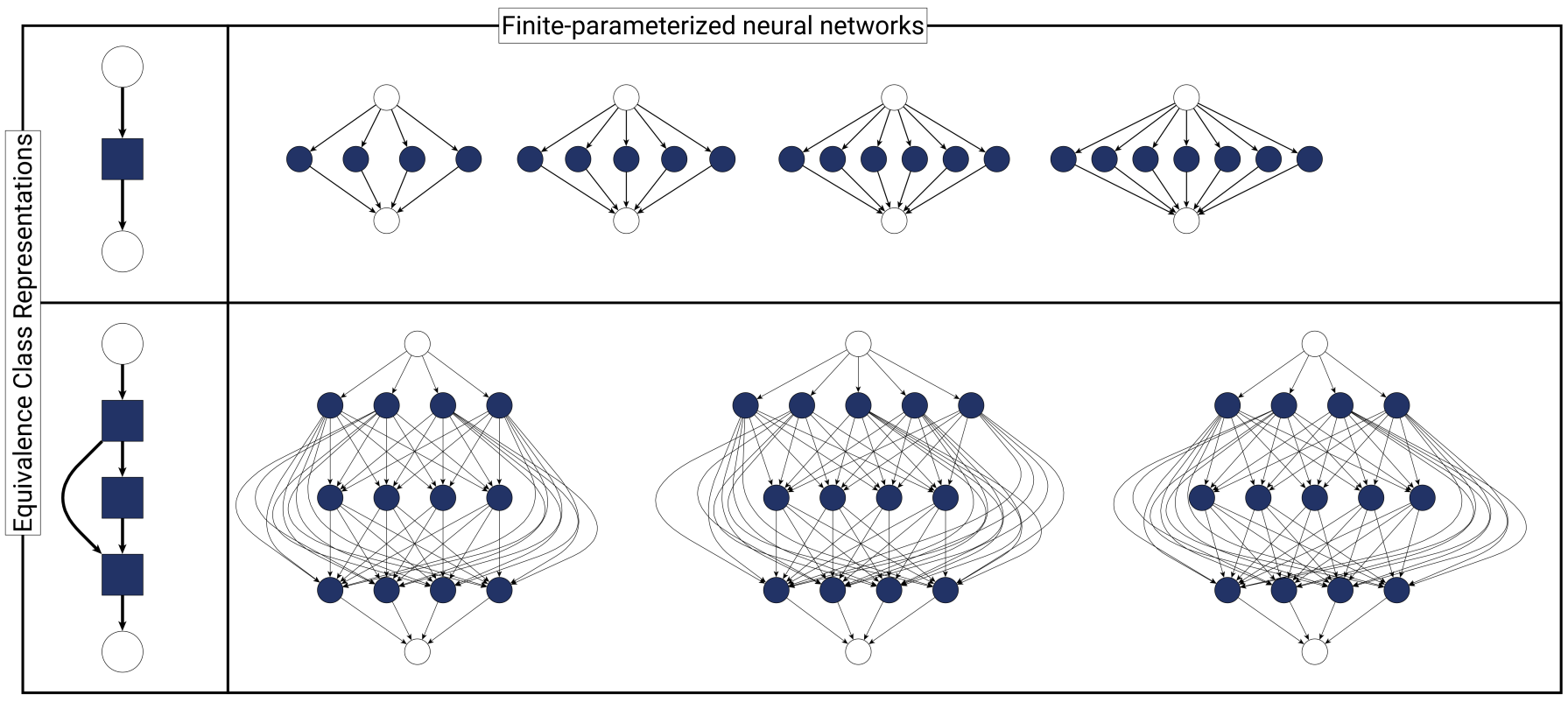

**Figure 6.4:** Two examples for equivalence class representations (left DAGs with squared interior vertices) and some examples of realisations of finite-parameterized neural networks sitting inside the respective graph-induced architectures.

Observe that Figure 6.4 is an example for an architecture from dimensions $d_1 = 1$ to dimensions $d_2 = 1$ and can naturally be chosen differently (although usually with $d_1 > d_2$ with an information-theoretic perspective). The input and output dimensionality are moved to the background on purpose as to explicitly emphasise the emerging existence of hidden structure of deep neural networks.

### 6.3.2 *Graph-Induced Networks with UAP*

Induced architectures are then further restricted to the ones that provide UAPs. An overview of universal approximation theorems and their properties is given in Section 5.9. Architectures with UAP separate comparably expressive spaces through the graphs they are induced by in different ways. In other words, these are the architectures $\mathcal{A}([G_1]), \mathcal{A}([G_2]), \ldots$ induced from representative directed acyclic graphs $G_1, G_2, \ldots$ we are interested in because the equivalence classes over $G_1, G_2, \ldots$ differ (in some network theoretic structural way). We further will use the symbol $T$ for the equivalence class representations instead of $G$ as to denote that the induced architectures have the desired properties and can then be called *themes*.

As a first step, we'll see how the equivalence class of the just provided example with $G_1 :=$ "$\cdot \rightarrow \square \rightarrow \cdot$" collapses our neural network definitions just into the right form used by Hornik [106] or Cybenko [43].

Hornik [106] defines a neural network with one hidden layer, "only one output unit" and over "$n$ hidden units" with

$$\mathcal{H}_k^{(n)}(\Psi) = \left\{ h : \mathbb{R}^k \to \mathbb{R} \mid h(x) = \sum_{j=1}^{n} \beta_j \Psi(a_j' x - \theta_j) \right\}$$

and shows [106, Thm. 2] that "$\mathcal{H}_k(\Psi) = \bigcup_{n=1}^{\infty} \mathcal{H}_k^{(n)}(\Psi)$ is dense in $C(K; \mathbb{R}^k)$ if $\Psi$ is continuous, bounded and non-constant".

Cybenko [43] "investigates conditions under which sums of the form"

$$G(x) = \sum_{j=1}^{N} \alpha_j \sigma(y_j^T x + \theta_j)$$

are dense in $C(K; \mathbb{R}^n)$ and concludes that they are; provided that "$\sigma$ is continuous and discriminatory".

Observe the similarity of both definitions. Negating the bias term $\theta_j$ and exchanging the symbols for the non-linearity, it can be easily seen that we have an affine transformation of an affine transformed and squished input. The conditions of the non-linearity are nowadays not restricted to just sigmoidal functions as in [43, Lem. 1] anymore but allow a way larger variety of at least one-side-bounded continuous functions to act as activation functions.

With $G_1$ having only one layer and no skip-layer connections, we observe that the recursive definition of Equation 6.3 collapses into $\mathbf{z}^{(0)} = entr(\mathbf{x}), \mathbf{h}^{(0)} = \mathbf{z}^{(0)}, \mathbf{z}^{(1)} = W^{0\to 1}\mathbf{h}^{(0)} + B^{(1)}$ and $\mathbf{h}^{(1)} = \sigma(\mathbf{z}^{(1)})$ and with identities for the entrance we obtain $\mathbf{z}^{(1)} = W^{0\to 1}\mathbf{x} + B^{(1)}$ which leads to the final form $\mathbf{h}^{(L)} = \sigma(W^{0\to 1}\mathbf{x} + B^{(1)})$ of the architecture $\mathcal{A}^{d_1\to d_2}(G_1)$ and an affine transformational exit shows the relationship between $\mathcal{A}^{d_1\to d_2}([G_1])(\mathbf{x}) = W^{exit}\sigma(W^{0\to 1}\mathbf{x} + B^{(1)})$ and the definitions of Cybenko and Hornik. The one-layered neural network set $\mathcal{A}^{d_1\to d_2}([G_1])$ is therefore uniformly dense in the continuous functions on compact subsets if the non-linear activation $\sigma$ is properly chosen. Parameters of the neural network architecture are given as $\{W^{0\to 1}, W^{exit}, B^{(1)}\}$ and the sizes depend naturally on the input and output dimensions of the particularly chosen problem.

This minor proof of graph-induced networks collapsing into classical feed-forward neural networks with universal approximation capabilities reveals that the framework is a generalisation of the latter: There exist sets of graphs which induce universal architectures.

A result of above formulation is a graphical approach towards the description of neural network architectures. The single-layered architecture $\mathcal{A}^{(1)} \triangleq \mathcal{A}^{d_1\to d_2}([G_1])$ with $G_1$ :=" $\cdot \to \square \to \cdot$" can now be extended to deep multi-layered architectures $\mathcal{A}^{(l)} \triangleq \mathcal{A}^{d_1\to d_2}([G_l])$ with $G_l$ :=" $\cdot \to \overbrace{\square \to \cdots \to \square}^{l \text{ squares}} \to \cdot$" and $l \in \mathbb{N}$. Here, $G_l$ contains a fixed number of $l$ consecutive layers. The underlying set of DAGs of $[G_l]$ then contains any directed acyclic graph with one source vertex, one sink vertex and arbitrary but at least $d_1$ many vertices per layer. Each vertex between two layers is fully connected. As the weight matrices between two layers of the induced architecture are at least of size $d_1 \times d_1$, they can easily represent an identity mapping. Therefore, the set $\mathcal{A}^{(l)}$ contains $\mathcal{A}^{(1)}$ and must also have a universal approximation property.

We have the representations $G_l$ for a sequence of equivalence classes and know that their induced architectures $\mathcal{A}([G_l])$ are universal approximators if proper activation functions are chosen. The question about an underlying structure of neural architectures can now be posed by asking which architecture should practically be chosen based on how the architectures induced by a graph representation relate. In this particular example, we know that $\mathcal{A}^{(1)} \subset \mathcal{A}^{(2)} \subset \mathcal{A}^{(3)} \subset \cdots$ such that we could conclude by *Occam's razor*, the principle of parsimony (or simplicity), that $\mathcal{A}^{(1)}$ should be chosen. We'll investigate on this setting in experiments later.

In general, we have architectures that are theoretically capable of universal approximation but can be induced from different sets of graphs. Usually and in opposite to the presented example, these sets of graphs have non-trivial relationships.

**Definition 2** *We call a set of graphs $\mathcal{G}$ with architectural induction $\mathcal{A}$ a (structural) theme set of neural networks iff $\mathcal{A}(\mathcal{G})$ has a universal approximation property. $\mathcal{A}([T])$ is a universal architecture. The equivalence class representation $T$ with $[T] = \mathcal{G}$ is called the structural theme of $\mathcal{A}([T])$.*

Not every set of DAGs induces an architecture with UAP. A theme set must be infinite in size as each individual directed acyclic graph only provides the construction of a finite-parameterised architecture. With the example $\mathcal{A}^{(1)}$ such theme sets exist and the graphically depicted directed acyclic graphs $G_1, G_2, \dots, G_l$ are examples for equivalence class representations of structural theme sets, i.e. they are structural themes.

## 6.4 A COMPARISON OF FORMULATIONS OF NEURAL NETWORKS

In research complex I in section 1.1.4 on page 6, we observed gaps in proper formalisms of deep neural networks for neural architecture search. A proper formalism should capture definitions of objects with as few uncertainties as possible, allow to argue about analytical properties, and reduce complexity when acting upon it. The proposed graph-induced networks are an answer to the question in Section 1.1.4 on how structure of deep neural networks can be defined. However, alternative formulations of neural networks exist, depending on the intended usage.

The intention for different formulations of neural networks can be broadly categorised into *analytical* and *empirical* goals. Analytical goals usually require very precise and formal definitions which lead to simpli-fications and assumptions. The benefits from this line of mostly mathematical research are strong arguments and the possibility to systemise the field on a long-term basis. Disadvantages include the difficulty of unifying formalisations across different disciplines and transferring results to applied research and practical cases. And this difficulty in transferring results already applies to fields such as neural architecture search, which, being still in an infancy state, is mostly empirical research but does not compete with state-of-the-art applications in most domains. Empirical goals therefore relinquish formalizing neural networks from the ground up on the benefit of describing technically more applicable situations. State-of-the-art neural network systems such as GPT-4 (ChatGPT) [210] can barely be captured in a formal way due to their immense technical setup and usage of highly recent and actively developed methods.

Table 6.1 summarises selected formulations used across different disciplines and contextualises our framework among them. Observe, how a clear gap between formalisation and application can be seen.

| **Formulation** | **Refs** | **Form** | **UAP** | **Struct** | **NAS** | **App** |
|---|---|---|---|---|---|---|
| Infinite-Width Single-Layered NNs | [43, 106] | ++ | ++ | - - | - - | - |
| Infinite-Depth Fixed-Width NNs | | ++ | ++ | o | - - | - - |
| Multi-Layered NNs | | ++ | ++ | o | - | ++ |
| Matrix-Vector-Tuple MLPs | [223] | ++ | ++ | o | - | - |
| Relational Graph Networks | [327] | - | - | ++ | + | ? |
| NAS Search Space Definitions | [61, 326] | - - | - - | + | ++ | ++ |
| Search Space Design | [230] | - - | - - | o | ++ | + |
| DAG-Induced Networks | (ours) | ++ | + | ++ | + | - |

**Table 6.1:** Taxonomy: **Form**: Formal definition; **UAP**: Analytical Usage for e.g. Universal Approximation Theorems; **Struct**: Applicability for Structure Analysis; **NAS**: Applicability in Neural Architecture Search Methods and Research (empirical), **App**: Applicability to real-world state-of-the-art models

*Infinite-Width Single-Layered NNs are the classical analytical form to proof UAPs of neural networks.*

To capture UAPs, infinite-width single-layered ("flat wide") neural networks have been used in the 1990s (compare Section 5.9 on UAP). Linear layers, separated with activation functions, can still be considered as the standard building block within any larger neural network system. As a general rule of thumb, increasing the layer size guarantees improvement of the approximation capabilities of the model. This usually holds as long as more complex generalisation measures are not taken into account. Beyond this relationship, the formal definition of infinitely wide single-layered neural networks are of no applied use.

Neither NAS considers this form of networks, nor can structure analysis be conducted with it.

*Infinite-Depth Fixed-Width NNs are used to to proof minimum widths of neural networks in context of UAPs.*

Infinitely deep networks with finite-width have revolutionised research in UAP by providing first results on minimum widths for approximation capabilities. From a perspective of analysing their structure, they are far more interesting than flat wide networks, as they can theoretically capture varying path lengths and degree distributions throughout their hidden layers. The infinite-depth, however, makes it practically difficult to capture varying structural properties properly and leaves this analysis to a mathematical one. As to our knowledge, analysis or formal usages of network properties of structures of such formulations have not yet been conducted. Considering our empirical research in this context, it also remains questionable whether there will be analytical results on network properties influencing approximation capabilities any time soon. While structural properties can be defined upon networks with infinite-depth, it can be considered as difficult to derive approximation properties for networks with different network theoretic properties. Applied systems usually make no use of infinitely-deep formulations of networks and if they make use of parameter-free formulations they usually turn to models such as gaussian processes. Neural Architecture Search can also only make use of finitely many layers such that the formulation of infinitely-deep networks also lacks proximity to empirical research.

*Multi-layered neural networks are the most common formulation. Their simple finite-parameterised feed-forward form serves both a practical usage and intuitive technical descriptions.*

Multi-layered neural networks are the classical formulation as outlined in Section 6.1. We distinguish this formulation from the previous two analytical formulations in the sense that it describes finitely parameterised networks. However, it usually does not consider skip-layer (residual) connections. That reduces the amount of underlying structural properties drastically but makes it both useful in analysis and closeness to application. Neural architecture search is usually less interested in this form as the only architectural parameters that can be influenced are the number of layers, the individual widths of layers and with some reformulation the functional form of layers. As natural extension of linear layers, this form is, however, the most used sub-module of larger neural network systems and thus shows closeness to real-world applications.

*Matrix-Vector-Tuple MLPs are less known formulations that serve a more rigorous definition and are a good example that take the importance of operations in representation space into account.*

Matrix-Vector-Tuple MLPs from [223] are additionally mentioned here, as they emphasise a form in which the networks as mathematical objects are treated as matrix-vector tuples while making use of the functional form of multi-layered neural networks. Matrix-vector tuples then allow for operations such as concatenation and parallelisation on tuples and that allows to clearly describe effects on the underlying functional form. As genetic algorithms in Neural Architecture Search theoretically make use of such relationships between operations in encoding space and the resulting architectures (mentioned in Section 6.2), this direction of analytical research is also very important. Apart from that, the formulation shares the same advantages and disadvantages as multi-layered neural networks.

Another type of neural network formulations can be observed in the field of NAS. For a comprehensive overview, compare Chapter 12. From a formal perspective, works in that field propose varying definitions of search spaces, e.g. based on undirected graphs, directed acyclic graphs, formal grammars, binary encodings, and many more. The formulations are usually method-driven such that the goal is to empirically show that e.g. high-performing models can be found with the proposed setting of search space and search method. A clear advantage is the flexibility of architectural definitions and search methods. The downside is a large amount of different formulations which are hard to compare. With the rise of neural architecture search benchmarks [61], the field improved in comparing the available methods under common search space definitions. But results are difficult to transfer between varying search space definitions. This fact gets worsened with higher level optimisation problems in which the search space design itself is subject of optimisation [230].

*Neural architecture search deals with various definitions of search spaces with the purpose of serving more recent technical details.*

Last but not least, there exists a large body of work based on undirected graphs as underlying structure for neural networks. For example, we originally used undirected graphs to construct neural networks from different random graph generators [272]. Building from this experience, we developed the here presented understanding of neural networks induced from directed acyclic graphs. You et al. [327] propose Relational Graph Networks for the relationship between neural networks and their structure.

*Undirected graphs as representation for structure of neural networks are more common in empirical studies on structural properties of neural networks.*

The idea of using undirected graphs has interesting aspects to it: network properties of undirected graphs are often easier to capture or not defined for directed acyclic graphs. Cyclic and bi-directional communication between vertices is naturally represented by undirected graphs. These aspects are often biologically inspired from e.g. BNNs such as the human brain, which suggests cyclic structures and not necessarily strict directed communication between vertices. Also, the evolution of undirected graphs is better studied than that of digraphs. Theories and methods such as graph convolutions or message passing have been originally developed for undirected graphs and are sometimes not trivial to transfer to their directed acyclic counterparts. Furthermore, there exist models such as echo state networks or liquid state machines in fields such as reservoir computing or self-organizing maps in which graphs might be or are a more natural representation.

There are, however, major disadvantages when working with undirected graphs as representations for neural networks. Computation naturally requires a direction and is sometimes not invertible or reversible. Transforming undirected graphs into a directed and possibly acyclic version as employed by Stier et al. [272] or Ben Amor et al. [6] can conceal underlying network properties. For example, while the

*Graph-induced neural networks make the usage of DAGs in structure analysis or NAS methods more explicit as compared to other formulations. This uncovers both theoretical and practical problems on which individual contributions were made and future research continues.*

degree distribution stays the same and additional information can be gained from looking at in- and out-degrees, many property distributions based on centralities or clustering coefficients change drastically for digraphs as paths are inherently differently defined in comparison to their undirected counterparts. Looking at well established literature on undirected graphs as by Diestel [52] and comparing it to established literature on digraphs as by Bang-Jensen & Gutin [14] shows how much the two fields can diverge. The field of structure learning for bayesian neural networks also considers directed acyclic graphs as the natural underlying representation and many advancements are driven by this field such as sampling DAGs[3]. Keeping formal precision when defining structure for neural networks and deducing structural properties from that structure is a clear advantage of digraphs over undirected graphs.

Above arguments lead to the formulation of neural networks induced from directed acyclic graphs as a generalisation of multi-layered neural networks. The generalisation comes on a fixed-parameterised level by introducing skip-layer connections such that directed acyclic graphs instead of sequences become a representation of a neural network realisation. On a function set level, an equivalent representation of DAGs can be used to represent architectures with universal approximation property and requires an understanding of how equivalence classes of directed acyclic graphs compare.

Skip-layer connections make a formulation tremendously more difficult without always bringing obvious benefit to it. Having a skip-layer connection between two non-consecutive layers also often requires technical workarounds in the implementation: a matrix multiplication between two connected layers must have appropriate sizes. That either requires all layers to have equal widths, exclude certain pairs of layers from having skip-layers in between them, or to blow up layers artificially up to a maximum size and shrink them down with binary masks when needed.

The presented formulations in table 6.1 show a slight increase in representational complexity of structure. Early studies worked with no structure in flat-wide networks and more recent research takes graphs as representations of the structure of hidden parameters into account. Directed acyclic graphs might not be the last choice for this representation.

One reason for this is that the order of edges going into a vertex is not taken into account in DAGs. In recent years and especially with the successful developments in geometric deep learning, the order of in- and also out-going edges might matter to reflect the intended employed principle. String diagrams from category theory might be a future solution to that. According to [225], the "use of string diagrams to describe the network structure only cursorily appear in [69]". Fong

3 Also compare Chapter 14 on advanced NAS methods with generative models for graphs.

et al. write in their discussion that "working in this bicategorical setting gives language for relating different parametrised functions and neural network architectures. Such higher morphisms can encode ideas such as structured expansion of networks, by adding additional neurons or layers." [69, Sec. VII. Discussion] That might give rise for a more category theoretical description of structure in deep learning. An overview of category theory in machine learning for the interested reader can be found in [254].

Other alternative representations for structure that are often employed in NAS include context-free grammars, program representations, or domain-specific languages. We leave further investigations on these alternative representations for the future and deem graphs with its rich underlying theory as a proper choice to frame our proposed research questions.

# 7 WORKING WITH GRAPH-INDUCED NETWORKS

*The following entails:*

We now can very explicitly associate a structural theme $T \in \mathbb{S}$ and its properties with a neural network realisation $f$ from a graph-induced universal architecture $\mathcal{A}([T])$. This allows to formulate questions on structures of neural networks in a unified language. Recalling our research complexes, we see three major approaches to work with the proposed formulation:

1. In *structure analysis* we are concerned about graph properties of $T$ and how they influence measures such as the accuracy $\alpha(f)$ of $f$.

2. For methods of neural architecture search, we wonder how we can exploit graph properties of $T_1, T_2, \dots$ or even $\mathbb{S}$ as to find better $T'_1, T'_2, \dots$ w.r.t. evaluated accuracies $\alpha(f'_1), \alpha(f'_2), \dots$ and their architectures $\mathcal{A}([T'_1]), \mathcal{A}([T'_2]), \dots$. After describing the optimisation problem, we categorise neural architecture search methods into the five major types prior induction, pruning, growing, evolutionary-, generative-, and differentiable approaches.

3. And beyond the NAS-methods acting on training a neural network (finding a realisation for a $T$) and finding an optimal architecture (finding an optimal $T$ in $\mathbb{S}$), we wonder whether $\mathbb{S}$ can change or has an influence. This reflects a change in the definition of structural themes as a way of *search space design*.

The formulation of graph-induced neural networks allows us to turn our focus on very fine details of the overall framework.

Different optimisation problems can be formulated with graph-induced networks: First, training a neural network is restricted through an implicitly or explicitly defined prior architecture $\mathcal{A}([T])$. Second, finding a better architecture can be formulated as a second-level optimisation problem on top of neural network training. And third, the optimisation problem can be generalised to a multi-level optimisation problem across e.g. data or application domains.

By considerations of the construction of graph-induced networks it gets clear, that a performance estimation for a theme $T_1$ as a representative of an architecture $\mathcal{A}([T_1])$ needs to be analysed carefully. The performance of $T_1$ can be superior to another theme $T_2$ only after some time of approximate training or only under a certain training scheme.

The following sections describe problems on graph-induced networks on a higher level and provide an overview of common approaches to find solutions for them.

## 7.1 OPTIMISATION PROBLEMS FOR GRAPH-INDUCED NETWORKS

Training a neural network can now be posed as an optimisation problem

$$\mathcal{F} \triangleq \underset{f \in \mathcal{A}([T])^{d_1 \to d_2}}{\arg\min} \mathcal{L}_{train}(f, D_{train}) \tag{7.1}$$

which aims to minimise a training loss $\mathcal{L}_{train}$ on training data $D_{train}$ with input and output dimensions $d_1, d_2 \in \mathbb{N}$. The structural theme $T$ induces the possible finite architectures and in convenient cases of $\mathcal{A}([T])^{d_1 \to d_2}$, the optimisation problem can even collapse to a common continuous setup of maximum likelihood estimation with backpropagation (instead of possibly having a discrete optimisation problem).

The formulation in Equation 7.1 covers most common neural network trainings and puts emphasis on an initial choice of a structural theme $T$. Usually, the choice of $T$ is done implicitly and could be considered as irrelevant as due to the UAP the solution set $\mathcal{F}$ is non-empty and can be theoretically reached.

For two structural themes $T_1$ and $T_2$ it is therefore a question of **1**) how efficient the elements of $\mathcal{A}([T_1])$ and $\mathcal{A}([T_2])$ can be evaluated, and **2**) whether the universal architectures provide convenient properties for searching and sampling from their space. In other words, it is a question of how well the respective solution sets $\mathcal{F}_1$ and $\mathcal{F}_2$ can be found or approximated for structural themes to be relevant or not. There is evidence, that the choice of a structural theme is not irrelevant and can therefore be explicitly incorporated into neural architecture search procedures.

Under this assumptions that the choice of a structural theme matters, an optimisation problem for neural architecture search methods can be formulated on a second level:

$$\underset{T \in \mathbb{S}}{\arg\min} \mathcal{L}_{val}(T, \underset{f \in \mathcal{A}([T])^{d_1 \to d_2}}{\arg\min} \mathcal{L}_{train}(f, D_{train})) \tag{7.2}$$

in which we make use of a search space $\mathbb{S}$ of directed acyclic graphs that contains meaningful structural themes.

Except for cases of differentiable architecture search in which the search space $\mathbb{S}$ is relaxed into a differentiable hyper-architecture of sub-modules, evaluating Equation 7.2 usually amounts to a discrete optimisation problem. Further, sampling $T \in \mathbb{S}$ now unveils why generative models of directed acyclic graphs are of recent interest for neural architecture search: learning a distribution of graphs with good structural themes allows to improve the navigation through the search space $\mathbb{S}$. The three directions of **1**) differentiable architecture search to relax the discrete optimisation into a continuous optimisation problem, **2**) sampling sophisticated architectures $T \in \mathbb{S}$ through generative models for graphs, and **3**) using surrogate or predictive models to estimate $\mathcal{L}_{val}$ or $\mathcal{L}_{train}$ are all tackling fundamental issues in solving Equation 7.2 and will be covered in Chapter 14 with more detail. Naïve evaluation of *Equation* 7.2 is exponential in nature (compare e.g. Section 3.4 for the space complexity of graphs) and grows super-exponentially with more levels of generalisation. A next generalisation would be to optimise the search space design:

$$\underset{\mathbb{S}\in\mathcal{P}(DAG)}{\arg\min}\ \underset{T\in\mathbb{S}}{\arg\min}\ \mathcal{L}_{val}(T,\ \underset{f\in\mathcal{A}([T])^{d_1\to d_2}}{\arg\min}\ \mathcal{L}_{train}(f, D_{train})) \tag{7.3}$$

There exist only few solution methods for multi-level optimisation problems with theoretical guarantees [246]. Decoupled from our symbolic notation, Sato et al. [246] sketch multi-level optimisation problems with continuous objectives $f_i$ with $i \in \{1,\dots,n\}, 2 \le n \in \mathbb{N}$ as

$$\begin{aligned} &\min_{x_1\in S_1\subseteq\mathbb{R}^{d_1}, x_2^*,\dots,x_n^*} f_1(x_1, x_2^*, \dots, x_n^*) \text{ s.t.} \\ &\qquad x_2^* = \underset{x_2\in S_1\subseteq\mathbb{R}^{d_1}, x_3^*,\dots,x_n^*}{\arg\min} f_2(x_1, x_2, x_3^* \dots, x_n^*) \text{ s.t.} \\ &\qquad\qquad \ddots \\ &\qquad\qquad\qquad x_n^* = \underset{x_n\in\mathbb{R}^{d_n}}{\arg\min} f_n(x_1, x_2, \dots, x_n) \end{aligned}$$

A convergence analysis for an exemplary trilevel optimisation problem can be found in [249]. The complexity of generalizing to multi-level optimisation and the fact that the solution methods are still unexplored make works in search space design exploration hard to explore. Often, higher level optimisation problems like Equation 7.3 are collapsed into Equation 7.2 by implicit assumptions on $\mathcal{S}$ or the problem formulation is not explicitly elaborated. Continuous relaxation of the search space as in DARTS [167, 176] is also often used to at least get rid of the combinatorial optimisation problem and to treat NAS as a bi-level continuous optimisation problem. Compare section 12.6 on page 239 on details of this approach.

Many methods which we analysed, such as all kinds of pruning methods, are not directly concerned with optimizing the underlying theme. In the case of pruning, there exist also goals such as information compression for memory footprint reduction. But the methods can be viewed under this framework to unveil underlying assumptions of each method.

REMARKS

- Training a neural network is an optimisation problem captured with Equation 7.1. Section 5.6 on page 70 describes how this problem is usually tackled with stochastic gradient descent.
- Subsequent types of techniques, i.e. prior induction, pruning, growing, or evolutionary searches tackle the optimisation problem captured with Equation 7.2.
- Search space design and hyperparameter studies need to be taken into account and are captured in higher-level optimisation problems as in Equation 7.3.

### 7.1.1 *Prior Induction*

In many settings of deep learning, the architecture $\mathcal{A}([T])^{d_1 \to d_2}$ used in Equation 7.1 is imposed implicitly. Often, a fixed description such as "a three-layered deep neural network" is used and the parameterisation is based on expert knowledge. This is equivalent to imposing prior information into Equation 7.1 by employing Equation 7.2 implicitly or without the explicit purpose of second-order optimisation. But this approach can also be seen as applying knowledge from well-guided studies on structural information. An example of this is the use of convolutional neural network (CNN) layers or residual networks.

From a pure structural perspective, convolutional neural networks are a structural constraint in $\mathcal{A}([T])^{d_1 \to d_2}$ that allow only local and few connections between successive layers. This is also connected to the assumption of having local spatial correlations in the feature space and that certain invariances such as translation, rotation or size invariance exist. Many such properties can be validly assumed, e.g. in the image domain as illustrated in Figure 7.1. The reduction in parameter size due to the structural constraint is one of the reasons for the successful application of convolutions (amongst e.g. weight sharing, that is also apparent in the application of convolutional layers).

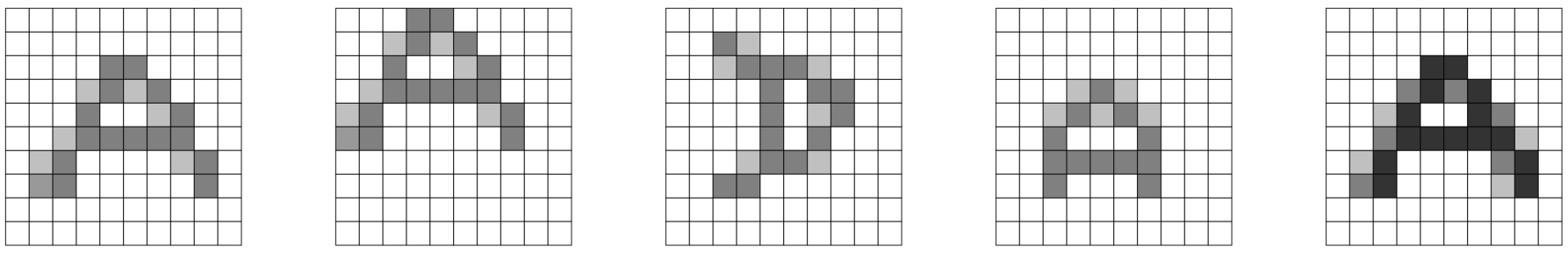

**Figure 7.1:** Examples for invariances in images: the original image of a sketch of the character "A" should still be recognisable after it is **translated** as in the second image, **rotated** as in the third/middle image, **scaled** (shrinked) as in the fourth image or darkened as in the last image.

Another example of prior induction includes whether to allow for residual (skip-layer) connections or not. Although residual connections require some technical tweaks because dimensionality across different (and not subsequent) layers need to be respected, employing residuals has empirically been shown to introduce valuable benefits to deep neural networks [95].

The presented formulation of graph-induced neural networks inherently focusses on the question whether structural priors even exists and how to characterise different priors. If all induced sets of neural networks are attributed with comparable UAPs, where does their structural difference come into play? Empirically we'll observe that variances of path lengths in structural themes seem to have a slight influence, i.e. in section 9.5 on page 167 and section 9.6 on page 174. A possible explanation might be given in approximation theoretic terms, when certain target functions might be better approximated with more variation of non-linearities of the input than with successive non-linearities. But even if a clear analytical answer in a formal way can not be made apparent, technical aspects of structural differences in themes will certainly play a rule in future applications. Considering the parametric size of state-of-the-art deep learning models make parallel computing paths interesting for accelleration or hardware optimisation. In that case, choosing themes with multiple parallel paths might be a valuable structural prior to choose over simple structural themes such as deep or consecutive ones.

As of our knowledge, we've been among the first to use complex networks (or network-theoretic probability distributions over graphs) as priors for the construction of deep neural networks in [272], around the same time as Xie et al. [319]. The search space exploration in such a case of a random graph model can be understood as sampling from a distribution of DAGs $P(G)$ such that $T \in \mathbb{S}$ becomes $T \sim P(G)$. Changing (improving) a prior during an evolutionary search can then be seen as learning the generative model $P(G)$ while conducting the architecture search – maximum a-posteriori learning with updating the underlying architectural space.

### 7.1.2 *Pruning*

Removing structural elements from a neural network model is called pruning. The term can be seen as a principle for a whole field of methods within deep learning. Usually, pruning is applied to a trained neural network model to obtain a derived model that has e.g. less parameters or other forms of reduced structural components. While focussing on the placement of pruning within our proposed formulation here, we dedicate Chapter 11 to the principle and investigate in empirical pruning experiments.

Pruning can be used on multiple levels of the posed optimisation problems on graph-induced networks. We describe three apparent interpretations:

1. **Better Theme Representative Estimation Through Pruning**: Finding a good solution for Equation 7.1 (*training a net*) is difficult and

usually results in an estimation through a training scheme. The scheme can be extended with a pruning loop in which a realisation of a neural network (with implicitly or explicitly employed prior theme) is first trained, then pruned and then often re-trained. The assumption is then, that the pruning step helps in discovering better realisations within $\mathcal{A}([T])$ w.r.t. both the resulting performance and paramterisation size of the realised neural network. Often, such a neural network has still representative performance for the theme but is compressed in the sense that it requires less parameters, memory or inferencing time.

2. **Pruning Themes**: In Equation 7.2 (*architecture search*), pruning could be applied to the outer optimisation, changing from an initial large theme and then gradually shrink down to smaller themes. Requirement for this is, however, understanding whether pruning has beneficial properties as an operation in the given search space of themes. And the pruning operation must be applicable to the theme definition, e.g. by allowing to prune vertices or edges of the DAG theme.

   An assumption in this context and for the particular application domain in which this might hold, is then at least that larger themes must have initial benefits over smaller ones. This could also hold, if the inner optimisation might be easier solvable for large hypothesis spaces $\mathcal{A}([T])$ initially but might benefit from hypothesis space restrictions through universal architectures of smaller themes.

3. **Pruning Realisations to Estimate Shrinked Themes**: The inner level of pruning during the estimation of a realisation of a theme can also be seen as an implicit change of the underlying theme such that one changes from $T_0$ to $T_1$. The density decays then in a sense that $dens(T_0) > dens(T_1)$. To figure out the change of themes, one would need to estimate the new theme based on the pruned network realisation. However, this approach requires a complex formulation for the themes $T_0, \ldots$ on the structural side: Recall, that an induced set $\mathcal{A}([T_0])$ requires a possibility for infinite growth to meet the requirements of universal approximation theorems. Pruning a part of the connections or neurons in a realised neural network, an element contained in $\mathcal{A}([T_0])$, would still be contained in $\mathcal{A}([T_0])$, but a change from $T_0$ to a less dense theme $T_1$ would need to reflect this change.

In summary, pruning is a general principle that can be applied to different levels of the optimisation problem on graph-induced neural networks. The field of pruning neural networks is mostly concerned with the first presented interpretation in which the underlying prior theme (or architecture) is fixed or implicitly applied and the actual realised neural networks are iteratively pruned. Empirically it has been shown [159] that simple pruning strategies can be applied on top of the end-to-end differentiable training of neural networks to achieve reduction in parameters, compression and beneficial structured sparsity. We confirmed some of these observations in own experiments [269].

Pruning can be also seen as an variational operator in an evolutionary search (compare section 12.3 on page 223). Applying pruning for guided optimisation of architectures is an open research topic.

### 7.1.3 *Growing*

Growth is a complementary principle to pruning in the sense that it increases existing structure, e.g. by adding neurons, layers or connections of a neural network realisation or by adding vertices, edges or sub-graphs into the underlying structural graph. Although it can be seen as the opposite to pruning, the requirements and assumptions of employing growth differ. Growing a neural network realisation has seen recent revival [103, 184, 280, 315], although it has been already explored during the 1990s as standalone methods on neural network realisations [125, 302] or within evolutionary approaches to neural network structures [323].

The following ideas of applying growth in the optimisation problems of Equation 7.1 and Equation 7.2 are apparent and go in analogy to the presented approaches with pruning:

1. **Growing Neural Network Realisations For Better Theme Estimation**: Neural network realisations contained in an architecture $\mathcal{A}([T])$ are finite-paramterised, while the architectural space requires the possibility of theoretically infinite parameterisation as to approximate target functions desirably closely. Growing the number of neurons from an architecture theme $T$ is the most evident approach in obtaining the underlying universal architecture $\mathcal{A}([T])$. Finding a solution for Equation 7.1 (*neural net training*) can therefore be usually approached by conducting a training scheme on a fixed-parameterised neural network realisation of $\mathcal{A}([T])$, test its approximation sufficiency e.g. in terms of classification $F_1$ score, and then increase the parameterisation and repeat the steps until the approximation converges. However, due to the non-linearity of the underlying problems, there is usually no guarantee for convergence. The resulting neural network realisation can therefore be seen as an estimated representative for evaluation of a theme $T$ and growing is one employable principle to guide finding a better estimation.

2. **Growing Themes**: Starting an architecture search in Equation 7.2 with a small theme $T_0$ from the search space would fit a principle of simplicity. As soon as neural network realisations are non-sufficient w.r.t. an objective, a next theme $T_1$ can be chosen by growing theme $T_0$ in the search space. Growth in the search space, like pruning, assumes the possibility of such an operation in the theme definition.

3. **Growing Realisations to Estimate Grown Themes**: Mixtures of optimisation levels can also be approached with growing principles. An example would be to grow a neural network realisation, possibly also beyond the restrictions of the architecture $\mathcal{A}([T_0])$ induced by its theme $T_0$. Obtaining a new theme $T_1$ could then be based on an estimation from the grown neural network realisation. This could be especially interesting, if the architectures $\mathcal{A}([T_0]), \mathcal{A}([T_1]), \ldots$ overlap such that during training-growing-alternations the probability of a realisation belonging to one or another universal architecture changes.

Maile et al. summarise questions associated with growing very sharp in their title: "when, where, and how to add new neurons to ANNs" [184]. The *Why* of growing can already be found in the potential of improved approximation capabilities of the trained network and thus estimations for the underlying architecture. But there might be also beneficial behaviours associated with growing, e.g. growing a formerly trained neural network might have different conditions in the optimisation landscape than directly training an untrained neural network of equivalent parametric size.

### 7.1.4 *Population-based, Evolutionary and Genetic Algorithms*

Without further requirements on the properties of a search space $\mathbb{S}$, population-based algorithms such as evolutionary or genetic algorithms can be easily employed on the proposed problems – as long as a sampling method for new candidate themes is available. A result of such searches are themes $T_1, \ldots$ that perform optimal w.r.t. evaluation performances on selected application datasets.

The search algorithms can include both pruning or growing strategies on themes as long as they respect the definition of structural themes, but are not restricted in the sense that any sampling method or variational operator in the search space can be employed. Multi-level optimisation problems can even be tackled with transferring information across levels, e.g. in lamarckian evolution, in which learned weights of a neural network realisation are carried across candidates of the outer search space to e.g. initialise realisations from another theme. The amount of variations, extensions or flavors of the employed algorithms are countless and we describe the basics in section 12.3 on page 223.

Finding optimal themes during a NAS is supposedely better achievable through evolutionary algorithms than through pruning or growing. Navigation through the search space is more flexible and can be bet-ter tailored to specific definitions of structural themes. But it remains an open research question on how these algorithms compare under a common computational budget.

### 7.1.5 *Generative Approaches to Neural Architecture Search*

When we recall the definition of an maximum a-posteriori estimation from Equation 5.9, e.g.

$$\hat{\theta}_{MAP} \in \arg\max_{\theta \in \Theta} Pr(X \mid \theta) Pr(\theta)$$

, we observe that the prior probability distribution $Pr(\theta)$ is taken into account and updated during approximating the true but not explicitly known distribution. Learning a generative model for DAGs can be understood as approximating parts of that prior distribution in which the structure of the model is taken into account.

Such a generative model for graphs (compare chapter 14 on page 261 on advanced methods with graph generators) can be summarised as a probability distribution $P(G)$ over graphs. Examples of such $P(G)$ were introduced in section 3.2 on page 30 with i.e. the Erdős-Rényi model and the Watts-Strogatz model. Recently, many new models and techniques have appeared that learn based on exemplary graph data through deep learning techniques with graph neural networks [47, 294, 313].

A distribution $P(G)$ can be used as a standalone approach in which e.g. iterative sampling or conditioning is used to obtain graph samples from $P(G)$ that perform better than randomly sampled graphs. But it can be also combined with e.g. evolutionary searches, in which e.g. a generative model $P(G)$ is used after a burn-in phase of randomly sampling graphs such that the already explored architectures can be incorporated into the information of the generative model. Another way would be to learn meta-information from different application domains as a prior distribution into such a generative model $P(G)$ and then use it to be re-fined on a required target application domain. The variety of combining these principles is endless but make it also empirically and theoretically difficult to compare in a systematic manner.

### 7.1.6 *Differentiable Approaches to Neural Architecture Search*

The outer optimisation problem of neural architecture search is often discrete and combinatorial which makes the problem hard to solve. Luo et al. & Liu et al. [167, 176] have been among the first to relax the discrete optimisation into a differentiable hyper-architecture such that only certain paths might be chosen among the architecture. The idea of making a discrete problem differentiable such as in DARTS [167, 176] has sparked many variations. Finding solutions in the hyper-architecture has the benefit of sharing weights among different solutions and additionally gradient-based optimisers can be used instead of discrete or gradient-free methods. We introduce the technical details to DARTS in section 12.6 on page 239.

The downside of this differentiable relaxation is that the search space gets to be restricted within this hyper-architecture. However, in open-ended searches, the approach can be combined with e.g. evolutionary methods such that resulting estimations from the differentiable hyper-architecture are used as a proxy for fast evaluations. The upside of differentiable methods are their computational efficiency, e.g. taking only few GPU hours in comparison to weeks of training neural networks separately.

### 7.1.7 *Summary on Optimisation of Graph-Induced Neural Networks*

In sections 7.1.1 to 7.1.6 we presented six major types of principles applicable to the optimisation problems presented in eq. (7.1) for training a neural net and in eq. (7.2) for bi-level optimisation of neural network realisation and its structural theme. The principles can be applied during training, on a level of the structural theme and often on both. For example, pruning can be used for compression during training outside the end-to-end differentiation or in the architectural search space $\mathbb{S}$ from one to another structural theme.

The optimisation methods applicable to graph-induced neural networks are not limited to the here presented approaches. For example, there also exist reinforcement-based or variational approaches to neural architecture search. Regularisation and sparsity constraints [73] are also related approaches. However, proper comparisons between methods applied to the presented optimisation problems are difficult because the methods make **1/** different assumptions about the underlying problem, **2/** apply different restrictions to the problems as to make solutions more accessible, and **3/** aim at different objectives as e.g. in finding architectures, good neural network realisations or obtain meta-information that generalises on other applications.

We present experiments on detailled aspects of selected methods such as pruning and genetic algorithms in subsequent chapters and also sketch current trends on advanced methods such as learning distributions of directed acyclic graphs.

## 7.2 SEARCH SPACE DESIGN

We now wonder about characteristics of the search space $\mathbb{S}$. From a minimal example we will build up towards how commonly understood multi-layered neural networks can be understood as different structural themes of a layer-wise search space definition. However, these search spaces appear too simple under the light of popular NAS benchmarks such that we consider arguments under which these benchmarks can be integrated into our formulation of graph-induced neural networks. The NAS benchmarks have advantages for providing fully sampled search spaces but do not fully agree with our fomulation. For example, the idea of graphs of such benchmarks resembling universal architectures is not given. We therefore argue for a more theoretically motivated search space of computational themes which we will analyse in addition to e.g. the NAS-Bench-101 benchmark in the subsequent chapters.

The possibilities in which a search space $\mathbb{S}$ in the proposed framework can be defined is – as of corollary 1 – restricted to be a set of directed acyclic graphs, represented by a graph $T$ (a structural theme), and each such set of DAGs needs to induce a universal architecture as of definition 2.

Additionally, to make sense of different themes, it must hold that for any themes $T_1, T_2 \in \mathbb{S} : [T_1] \cap [T_2] \neq \emptyset$, i.e. the sets of directed acyclic graphs used for the architecture construction of neural network realisations must be different. Otherwise the separation of the space is not meaningful.

A minimal example of such a $\mathbb{S}$ would be set of two representative themes $\mathbb{S}_1 = \{T_1, T_2\} = \{\Box, \Box \to \Box\}$, previously denoted in section 6.3 on page 110. Here, the first theme $T_1$ represents the set of all single hidden-layered neural networks and the second theme $T_2$ represents the set of all deep neural networks with two hidden layers. An uninformed uniform prior over these two architectures would then be simply a Bernoulli distribution $\mathrm{Ber}(p = 0.5)$, i.e. a distribution over a finite support of two possibilities with equal probabilities for each outcome.

This sheds some light on both the strengths and weaknesses of modeling a prior distribution on a hypothesis space based on themes of graph-induced neural networks: the hypothesis space is discretised into a set on which proper informed prior distributions can be defined which makes modeling structural priors interesting. This is, however, only possible for finite search spaces or ones on which prior distributions with infinite support can be employed.

The usefulness of a search space design can be assessed on whether meaningful distributions for it can be found. For example, if, after many successive updates of an initially uniformly concentrated dirichlet distribution as prior to a multinomial distribution over the possible themes of a finite search space, the concentration tends towards one or few themes, there should be a difference in the universal architectures. While this yields (hopefully) information about structural differences, the search space design also becomes obsolete as one theme can be chosen over the other one. We suppose, that conducting such an experiment should yield no valuable result across all application domains as to not conflict with the No Free Lunch theorems (NFLs) [310, 311], i.e. one universal architecture can not always be superior to another one.

### 7.2.1 *Layer-wise Themes*

An example, previously used in Section 6.3, was the extension to arbitrarily stacked layers in the sense that $\mathbb{S}_2 = \{\mathcal{A}^{(1)}, \mathcal{A}^{(2)}, \mathcal{A}^{(3)}, \dots\} = \{\square, \square \to \square, \square \to \square \to \square, \dots\}$ denotes the search space of universal architectures over one-hidden-layered, two-hidden-layered, three-hidden-layered neural networks, and so on. Practically, it can easily be made finite such that a categorical random draw with initially uniformly concentrated dirichlet prior would be a valid uninformed prior distribution over these themes. But a weighting towards themes with fewer layers would be also applicable as the induced sets are contained in the sets induced from themes with more layers and therefore one could lean towards smaller architectural spaces.

The example of layer-wise themes $\mathbb{S}_2$ is an example of countable themes in a search space. Together with the example $\mathbb{S}_1$ it should illustrate, that the construction of the search space has a direct influence on how experimental setups can later be constructed to gain valuable information about the search space and how it divides the hypothesis space into structurally discretised subsets. Many other search spaces are designed based on (context-free) grammars or directed acyclic graphs and are then combinatorically hard to capture, e.g. for sampling from these spaces.

### 7.2.2 *NAS-Bench-101 & other conv-like NAS-Benchmarks*

Significant contributions have been made by introducing thoroughly evaluated neural architectures and providing them publicly as benchmark datasets, as e.g. NAS-Bench-101 by Ying et al. [326]. We elaborate in more detail on them in Section 12.2 for empirical purposes.

*Many benchmarks and existing search space formulations can be translated into graph-induced neural networks, as they often work with DAGs with operational restrictions as labels.*

Strictly speaking, most of these search space descriptions do not fall under the definition of a search space as described in this work because each candidate architecture – whether consisting of staged cells or other compositions – is finite-paramterised and not capable of universal approximation. One could argue, that the candidates still serve as proper representative neural network architecture realisations and we will actually treat the valuable information of these benchmarks in that way, but some formal problems need to be considered: most benchmarks define e.g. hierarchical finite-parameterised architectures but these "architectures" could be formally equivalent or at least partially overlap when scaling them up. There are e.g. theoretical indications, that "for a CNN with kernel size three to have universal property, at least $d - 1$ layers are required" [111, Sec. 4.3].

The space of NAS-Bench-101 architectures[1] is described as " all possible directed acyclic graphs on $V$ nodes" [326, Sec 2.1. Architectures] with each vertex having one out of $L$ labels denoting the particular operation and $V$ being restricted to $|V| \leq 7$, the number of edges restricted to $|E| \leq 9$ and the three possible operations are $3 \times 3$ convolution, $1 \times 1$ convolution and $3 \times 3$ max-pooling. This design has empirical backgrounds, e.g. "to ensure that the search space still contains ResNet-like and Inception-like cells" [326] and its success and adoptions illustrate the need of the research field to proceed with more unification such as working with benchmarks.

One possibility to view NAS-Bench-101 as a search space for graph-induced neural networks is to treat the employed finite-sized labelled directed acyclic graphs defined in NAS-Bench-101 as the set of themes of the search space. The fact that the realisations of these DAGs are stacked as cells when transformed into a neural network realisation and the possibility to repeat this stacking over and over allows to conform with universal approximation theorems under the assumption that the paramterisation is wide enough to have no information-bottleneck. It remains, however, open for consideration on how architectural themes with e.g. three parallel running $3 \times 3$ convolutions then actually represent meaningful different architectural spaces. Or whether a non-linear behaviour in the relationship of different architectural themes occurs when stacking is deepened beyond the three consecutive cells.

### 7.2.3 *Computation Themes*

Recall, that universal architectures are joint sets of parameterised neural networks consisting of directed acyclic graphs, induced from again a combinatorial structure such as directed acyclic graphs. Our considerations with universal architectures led to a definition of computational themes (CTs) for neural architecture search and an associated benchmark dataset called CT-NAS.

The idea is the following: a DAG allows for the formulation of more interesting structural properties than just a sequence of layers, e.g. the path length distribution. The combinatorial nature and richness make DAGs complex to work with but at the same time DAGs are more interesting as a search space when compared to e.g. few finite elements as in previous examples.

*While many benchmarks concentrate on labelled graphs, unlabelled directed acyclic graphs pose not the best way to compare structural themes in this framework. A puristic view without diverse operations for different computational paths motivates an interesting subset of DAGs based on a vertex-collapse condition.*

With that motivation, the set of all unlabelled DAGs would be a suitable space for investigating the structure of neural networks. But with the proposed architectural induction, one can make an interesting

1 We use the term *architecture* loosely here and in alignment with the usage in the field of neural architecture search in which an architecture is usually a high-level directed acyclic graph description of a finite-parameterised neural networks with labelled vertices that describe the custom used operations such as convolutions with certain kernel sizes.

observation: having two DAGs of which one has an additional vertex with the same set of in-degree and out-degree neighborhoods as an existing vertex present in both DAGs, the underlying induced neural networks would be subsets of each other. This can be seen visually in two depicted DAGs in Figure 7.2 on the left and in the middle.

A lot of definitions would not be necessary if we search through the space of all DAGs – the direct underlying computational graph structure of each and every neural network realisation. But then differences need to be searched for in the *parametric representation* rather than in any network-theoretic structural properties. Therefore, we would better treat these two DAGs as equivalent and use the one with smaller order as representation for an equivalently induced architecture.

Another and related perspective of the same phenomenon is, that each vertex of such a computational theme represents a unique linear transformation based on an input domain. Duplicating this vertex, defined by its neighborhood, would just represent the same transformational capacity without any non-linearity in between. In other words, each vertex of a computational theme represents a unique transformational module, that is separated by local small non-linearities along the edges (through the activation function in a neural network realisation).

The third argument for reducing the set of DAGs is that one of the sole reasons of this work is being able to define reduced search spaces of universal architectures and to formulate hypothesis about their difference. And computational themes, as we will shortly see, still grow – despite being a subset of DAGs – in comparable combinatorial complexity as DAGs which still poses a problem for constructing distributions over them. This means, that more reduced search spaces would be even favorable from this point of view.

**Definition 3 (Computational Themes)** *We define a computational theme as a connected directed acyclic graph* $G = (V, E)$ *with root vertex* $v_{root} \in V, Nei_G^{in}(v_{root}) = \emptyset$ *and sink vertex* $v_{sink} \in V, Nei_G^{out}(v_{root}) = \emptyset$ *restricted to*

1. $|V \backslash \{v_{root}, v_{sink}\}| > 0$
2. $\forall v \in V \backslash \{v_{root}, v_{sink}\} : Nei_G^{in}(v) \neq \emptyset \wedge Nei_G^{out}(v) \neq \emptyset$
3. $\forall v_1, v_2 \in V : Nei_G^{in}(v_1) = Nei_G^{in}(v2) \wedge Nei_G^{out}(v_1) = Nei_G^{out}(v2) \Rightarrow v_1 = v_2$

Computational themes are a subset of directed acyclic graphs. The term is chosen for the lack of a better name and for being an instance of structural themes. Main property for the definition of computational themes is the last restriction, a *vertex-collapse condition* that captures the aforementioned property, that two vertices are considered equivalent if they have the same sets of incoming and outgoing neighbors (compare neighborhood). The first restriction just assures that there exists a vertex, that is not the root or sink vertex, and the second restriction assures that all other vertices have both in-degree and out-degree neihbors.

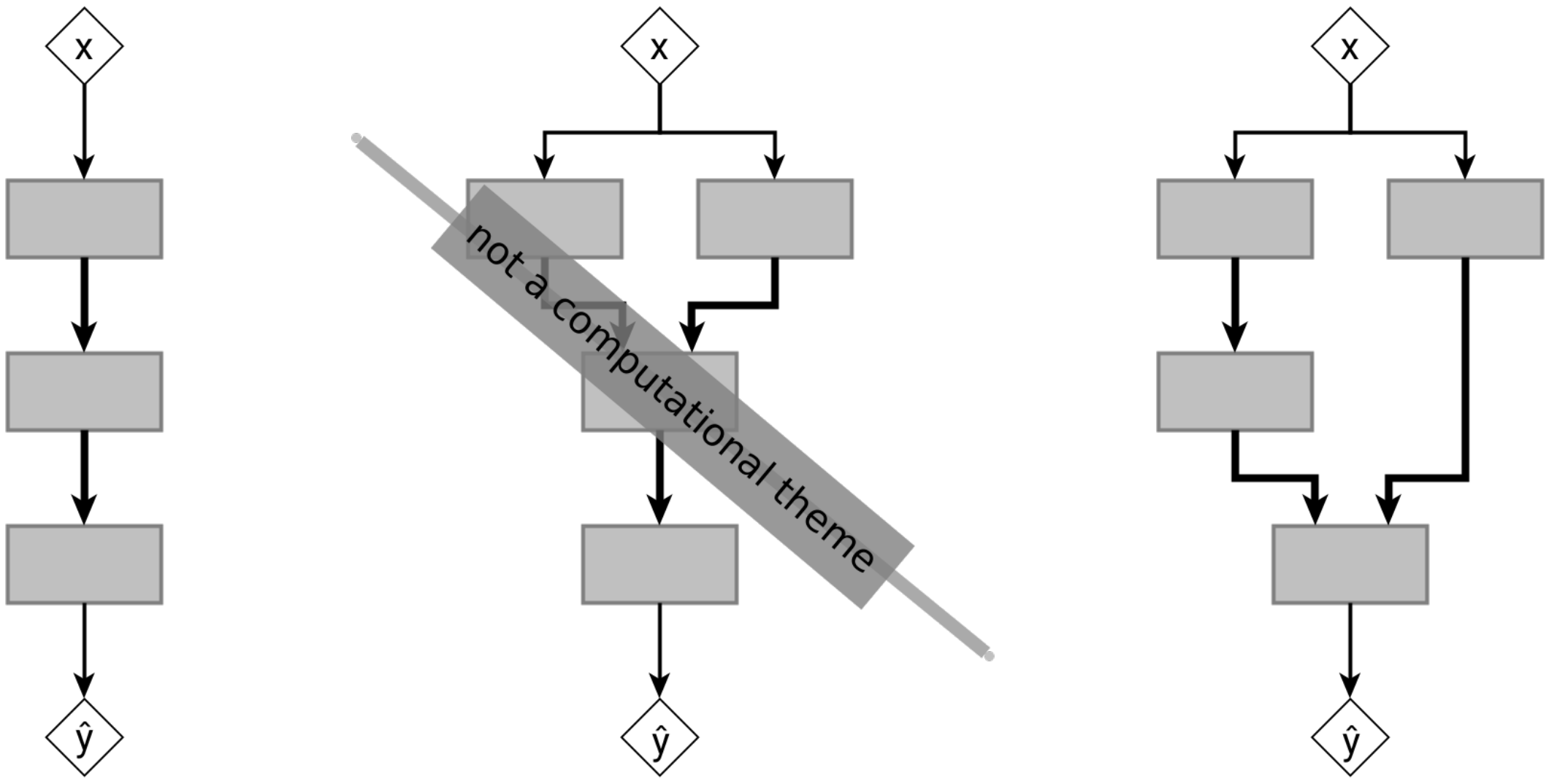

**Figure 7.2:** Visual examples for computational themes (CTs). The vertices are depicted as squares, indicating, that they represent layers of arbitrary sizes when related to the induced universal architecture of the theme. In a familiar manner of defining equivalence class representations for sets of graphs and then transforming them into neural network architectures, we now define an outer space of computational themes. Universal architectures are not restricted to that definition, as they just require a description of constructing sets of neural networks that fulfill the universal approximation property. Both definitions – computational themes and universal architectures – serve different motivation but are inherently related because of their nature based on directed acyclic graphs. Elements of the set of computational themes are equivalence class representations of sets of directed acyclic graphs, that can be used to induce universal architectures. Elements of universal architectures are finite-parameterised neural networks based on a specific input- and output domain as common for functions.

The nature of the vertex collapse condition makes computational themes (CTs) representatives of an equivalence class of directed acyclic graphs. This set can directly be used to construct universal architectures of neural networks. We employed computational themes in our subsequent experiments and explained in section 8.3 the setup for such graphs up to an order of six. Although we have not derived a counting strategy, yet[2], with sampling strategies such as rejection sampling and other existing samplers for DAGs, sufficiently exhaustive sets up to few graph orders can be obtained. This allowed the analysis of structural properties beyond benchmarks with labelled graphs as to force

2 We suppose that a generating function could be similarly obtained through exclusion-inclusion principles as for directed acyclic graphs, but we have not carried out such analysis for computational themes.

more emphasis on the pure structural aspect of the analysis and less on complex interactions between operations.

### 7.2.4 *Search Spaces of Random Graphs*

The fourth and last example for a definition of a search space $\mathbb{S}$ is based on probability distributions over graphs. We considered this approach in Stier & Granitzer [272] by using the three random graph models presented in section 3.2.1, namely the Erdős-Rényi model, the Watts-Strogatz model, and the Barabasi-Albert model, to construct deep neural networks, and will later also provide structure analyses based on this search space.

A search space based on random graphs requires a construction of a sample space over an infinite (or at least infinitely growing) graph order such that the model itself can act as the surrogate equivalence class representation. Technically[3], for taking the Erdős-Rényi model as a structural theme we sample $G_1, G_2, \cdots \sim GIL(n, p)$ of sufficiently large graph order $n \in \mathbb{N}$ from which the deep neural networks is then constructed. But we can also substitute $GIL(n, p)$ with $WS(n; k; p)$, $BA(n; m)$ or even a probability distribution learned from samples of graphs.

An important detail of using random graph models as structural themes is that undirected graphs are only related to directed acyclic graphs but are not equivalent. This was already discussed in section 3.4 and leaves open research questions in random models for directed acyclic graphs on which emergent graph properties are analytically better understood.

The exemplarily $\mathbb{S} = \{T_1, T_2, \dots\}$ becomes finite and small in which $T_1, T_2, \dots$ are not only structural themes but also graph generators. Graph generators might even overlap to a high degree in yielding particular graphs – making the original idea of equivalence classes of graphs fuzzy. Experimentally evaluating realisations of an induced architecture to assess a theme in a search space of random graph models might become more difficult because the evaluation includes a sampling strategy and the architectural induction $\mathcal{A}$ has to be carefully studied.

An advantage of random graphs as structural themes is that the underlying graph properties are not only more diverse but also certain models such as the Watts-Strogatz model or the Barabasi-Albert model exhibit very distinguishable emergent properties. Emergent in this context comes from growing numbers of vertices and this might

3 We are aware, that we formally join distributions over ever larger sample spaces and this is not simply done by taking the product space. The challenge of constructing arbitrary large edge spaces as exemplified in section 3.2 is not only difficult but also occurs as a technical challenge when modeling learnable distributions of graphs over different graph orders.

be also reflected in approximation capabilities of the induced universal architectures.

### 7.2.5 *Summary on Search Spaces*

Our formulation of graph-induced neural networks restricts search spaces of deep neural networks to ones based on directed acyclic graphs. Still, we find very distinct types of search space formulations: classical layer-wise designs as mostly used in theoretical settings, convolution-like spaces as mostly used in applied NAS settings, CTs as motivated from our formulation itself, and spaces based on random graph models instead of a deterministic definition of a set of graphs. Within these types, there are still various differences possible and the choice of a space design depends heavily on the intended use case. Theoretical work might shift towards graph-induced formulations if varying computational paths turn out to make a difference in approximation capabilities while applied NAS work might shift towards more unified search space designs as to improve comparability and repeatability of experiments.

REMARKS

- We presented four major types of search space designs: layer-wise, conv-like, CTs, and based on random graph models.
- The search space design types not only have different origins but also implications for interpreting results or our formulation of graph-induced neural networks itself.

Part IV

# ANALYSIS OF DEEP LEARNING STRUCTURES

Graph-induced neural networks capture this works understanding of what a neural network is and how its structure can be understood. Does structure empirically make a difference?

**Research Complex II: Structure Analysis**

From section 1.1.5:

- Do neural network models differ in terms of structure or are they just different with respect to the choice of other hyperparameters such as activations, training epochs, data sampling or augmentation strategy, optimisation procedure, or the choice of loss function?
- Are well studied network theoretic properties involved in the behaviour of neural networks? What are common structures or patterns that can be observed?
- How could analytical results look like to guide further architecture development?

# 8 CONSIDERED DATA

*The following entails:*

Wolpert & Macready introduced a number of No Free Lunch theorems that resulted from very general motivating questions such as "what is the underlying mathematical *skeleton* of optimisation theory before the *flesh* of the probability distributions of a particular context and set of optimisation problems are imposed?" [310, 311]. Adam et al. summarise the No Free Lunch theorems (NFLs) as "averaged over all optimisation problems, without re-sampling, all optimisation algorithms perform equally well" [2]. The argument is nowadays summarised under the term No Free Lunch theorems and its theoretical implications are of high interest in light of this work.

The framework presented in Part iii poses quite general optimisation problems in Section 7.1 and the question of underlying structure-inducing graphs is one about *a priori* knowledge. While the NFL provides theoretical limits to the generalisability of the framework, we have eagerly worked with varying application domains to investigate on the posed research questions about the structure of neural networks. It could well be, that certain structural properties impose a very strong and generic prior in one application domain while its influence diminishes across all domains. We mostly considered the image processing domain, but also conducted some experiments in natural language processing, e.g. with `20newsgroups dataset`, in the area of graphs, took selected data sets from `UCI dataset`, and also employed artificially generated data. Employing architecture search within and across application domains is required to uncover structures dependent or independent of the domain restriction. We suppose, that the NFL as a general argument includes so many diverse hypothesis spaces of unusual properties that useful common priors within large domains such as image processing should still be possible.

## 8.1 IMAGES

The image processing domain contains an enormous amount of real-world data and traditionally serves as a standard benchmarking domain for comparing new techniques. Available datasets range from many samples of tiny images as in MNIST ($60,000 \times (28,28)$), over enormous samples of tiny images as in SVHN ($600,000 \times (32,32)$), to quite a lot samples of medium sized images as in CelebA [169] ($202,599 \times (178,218)$), or to enormous amounts of large-sized images as in ImageNet [50] (e.g. $1,281,167 \times (64,64)$ for ImageNet64).

PREPROCESSING Benchmarking image datasets are often prepared such that objects are neatly centered in the available image frame. In addition to these benchmarks, transformations such as a *z-score standardisation*, often referred to as normalisation, is applied. For this, each feature (such as a pixel value) is rescaled with the sample mean $\mu$ and standard deviation $\sigma$ of the dataset: the normalised feature value $z = \frac{x-\mu}{\sigma}$ is obtained based on the actual feature value $x$.

DATA SET OVERVIEW Most of our experiments in the image domain have been conducted with `MNIST` and `CIFAR10`. Further data sets we considered are `SVHN` [203], `IMAGENET` [50] for scalability considerations, Fashion-MNIST [318] and CelebA, a database of celebrity faces with annoated face attributes [169].

## 8.2 ARTIFICIAL SPHERESUDCR DATA

We employed an artificially generated type of classification data set, which we simply refer to as `SpheresUDCR`. The underlying mechanism is to use overlapping spheres in a $D$-dimensional space of varying radii and assign them randomly chosen class labels. This classification data set has the benefit of 1st/ generating preferable sample sizes, 2nd/ choosing the dimensions of both the input and the output space and 3rd/ influencing the difficulty of the task by modifying few parameters.

The idea of the artificial generation process is as following: each point $x \in \mathbb{R}^D$ in `SpheresUDCR` is assigned a single class label $c \in \{0, \dots, C-1\}$ with $D$ being a small input dimension and $C$ being the number of classes for the classification task. The class label can be considered a unique color in which the sphere (or ball) can be visualised with. The underlying generation principle is close to the workings of $k$-nearest neighbors [41] or k-means [179, 331].

The generation process is initialised with an empty ordered list of spheres $\mathbb{S} = []$. A sphere $S$ is a triple $(c, r, l) \in \mathbb{R} \times \mathbb{R} \times \{0, \dots, C-1\}$. Subsequently, the process can be divided into two phases.

**First**, a new point $x \sim U$ is sampled. For a this point $x$, it is checked whether a sphere $S \in \mathbb{S}$ exists for which $x$ lies in $S$. Each existing sphere is checked in first-in-first-out manner. A point $x$ lies in $S$ if $\sum_{i=0}^{D-1}(x_i - c)^2 < r^2$.

If no such sphere exists, the **second** phase of the process starts. A new sphere is generated and appended to the list of spheres. The given point $x$ is chosen as the center point $c$ of the new sphere. A pre-defined distribution $R$ is used to sample the radius $r \sim R$ of the new sphere and a (usually) uniform distribution is used to select a label $l \sim U(\{1, \dots, C\})$ for the new sphere.

The data set is extended with the new sampled point and its label – may it be coming from an existing or the new appended sphere. Until the number of desired samples $N$ is reached, the generation process repeats the steps by sampling a new point $x$. algorithm 5 provides an alternative form for each step of the algorithm.

**Algorithm 5:** The generation process for `SpheresUDCR` datasets is based on two phases: one phase needs to determine if a point already belongs to a sphere by e.g. using a direct method *GetLabel* as in algorithm 6. In a second phase, new spheres are sampled if the point is not within the radius of an existing sphere, yet. Smaller radii sampled from $R$ increase the number of spheres within the sample space $U$ and thus shatter the space more. More shattered spheres then belong to the same label (or color) but are not topologically close which makes the generated classification problem more difficult for machine learning algorithms that rely on that implicitly such as CNNs and Decision Trees or explicitly such as $k$-means or $k$-nearest neighbors.

**input :** An input dimension $D$, a number of classes $C$, a univariate distribution $R$ with positive support, a number of desired samples $N$ and a uniform-like distribution $U$ over $D$ dimensions to draw uninformed samples from

**output:** The sampled spheres $LS$ and labelled points $xs$

```
LS ← [] // the list of spheres
xs ← [] // the set of samples
for ix ← 1 to N do
    x ~ U // Draw a new sample point
    l ← GetLabel(LS, x);
    if l < 0 then
        c ← x, r ~ R, l ~ U({1, …, C});
        S ← (c, r, l) // the new created sphere
        LS ← LS ∪ {S} // appended into FIFO list
    end
    xs ← xs ∪ {(x, l)};
end
return LS, xs
```

An important aspect of this type of classification data set is the iterative and ordered generation of spheres. Early spheres overshadow later generated spheres based on the first-in-first-out (FIFO) principle. The FIFO principle guarantees a determination for which sphere a point belongs to – and therefore which class label a point belongs to. Different spheres can belong to the same or different classes. This leads to overlapping or merging spherical shapes.

**Algorithm 6:** A subroutine `GetLabel` for checking whether the point should belong to a new sphere or to an already defined one. It could be optimised by carrying a more complex data structure for spheres which e.g. hierarchically reduces the location of centers of spheres to sublists such that not all spheres or points need to be checked when looking for the closest sphere of a given point. The algorithm runs with an asymptotic quadratic complexity over the number of samples, e.g. $\mathcal{O}(|xs|^2)$ as each point needs to be sampled and in worst case for each point all spheres need to be checked – which is bound by $|LS| \leq |xs|$.

**input** : An ordered list $LS$ of spheres and a point $x \in \mathbb{R}$
**output**: The class label $l$ for point $x$ out of all spheres in $LS$

```
for S ← LS do
    (c, r, l) ← S;
    if ∑_{i=0}^{D−1} (x_i − c)^2 < r^2 then
        return l
    end
end
return -1
```

The complexity of the generation process increases obviously by more sampled data points. For each point, it can happen that an own sphere needs to be created as the point does not yet lie within the radius of an existing sphere. In a worst case, this leads to $|LS| = |xs|$, i.e. the number of spheres growing up to the number of samples. A simple data structure for checking all existing spheres to determine the class label of a given point as in algorithm 6 then leads to a quadratic complexity of the generation process. This is not insignificant for several data sets of each dozens of thousands sample points.

The generation process can be sped up by carrying additional topological information for the spheres. Having hierarchical information about which spheres are within certain subspaces of the overarching sample space $U$ allows to reduce the runtime complexity of *GetLabel*. We leave improvement and further development of these algorithmic aspects to future work.

Following the process in algorithm 5, we generated several hundred artificial classification data sets with varying parameters and stored them with unique identifiers. The data sets then can easily have $N = 20,000$ samples which can be used to split into train-, test- and validation sets. We usually have chosen the uniform distribution $R = \mathcal{U}(u_1, u_2)$ with $(u_1, u_2)$ having different settings between datasets such as e.g.

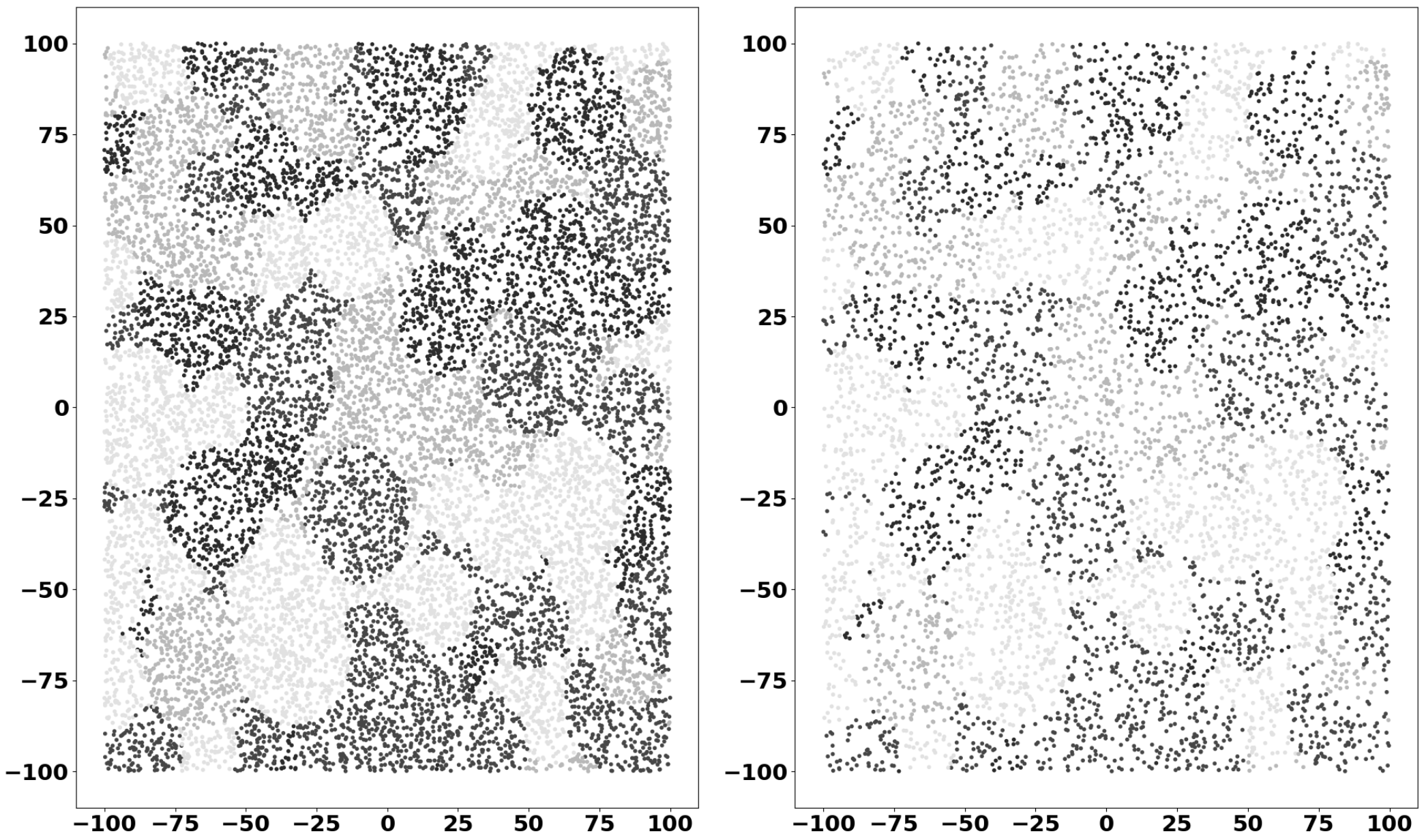

**Figure 8.1:** The two visualisations illustrate a dataset sampled through algorithm 5 for two-dimensional points. Each point is colored with one of the four ($C = 4$) possible class labels – based on the sphere it belongs to. Earlier sampled spheres overlap later sampled spheres due to the FIFO queue. The right visualisations shows testing samples that are not contained in the left visualisation such that the interpolation capabilities of a machine learning model can be easily measured. Smallest radii are ten and largest radii are set to twenty, such that $R = \mathcal{U}(10, 20)$.

$(10, 21)$. Two- and three-dimensional points, i.e. $D = 2$ or $D = 3$, were used to inspect datasets visually. Some of these examples can be seen in Figure 8.1, Figure 8.2 and Figure 8.3. Higher dimensional datasets are more difficult to visualise and usually require a projection.

Six of these data sets are called (or abbreviated based on an assigned unique identifier) `spheres-0a19afe4` (⊛), `spheres-23aeba4d` (✠), `spheres-6598864b` (⨁), `spheres-b758e9f4` (▷), `spheres-b8c16fd7` (♠), and `spheres-bee36cd9` (♥). The symbols will be restated and used to visually quickly refer to a particular dataset. All of these six datasets were generated with $C = 10$ classes as to compare them later with `MNIST` and `CIFAR10`. Further, they are bounded in $(-100, 100)$ along each dimension and the radius was sampled from $(10, 20)$. For `spheres-23aeba4d` (✠), `spheres-b8c16fd7` (♠), and `spheres-bee36cd9` (♥) we used $D = 2$ while for `spheres-0a19afe4` (⊛), `spheres-6598864b` (⨁), and `spheres-b758e9f4` (▷) we used $D = 3$ as input dimensionality. In the manner we employed the artificially generated data sets, it is less important to know the particular names of them rather have an impression of how easy they are to be approximated by a machine learning model.

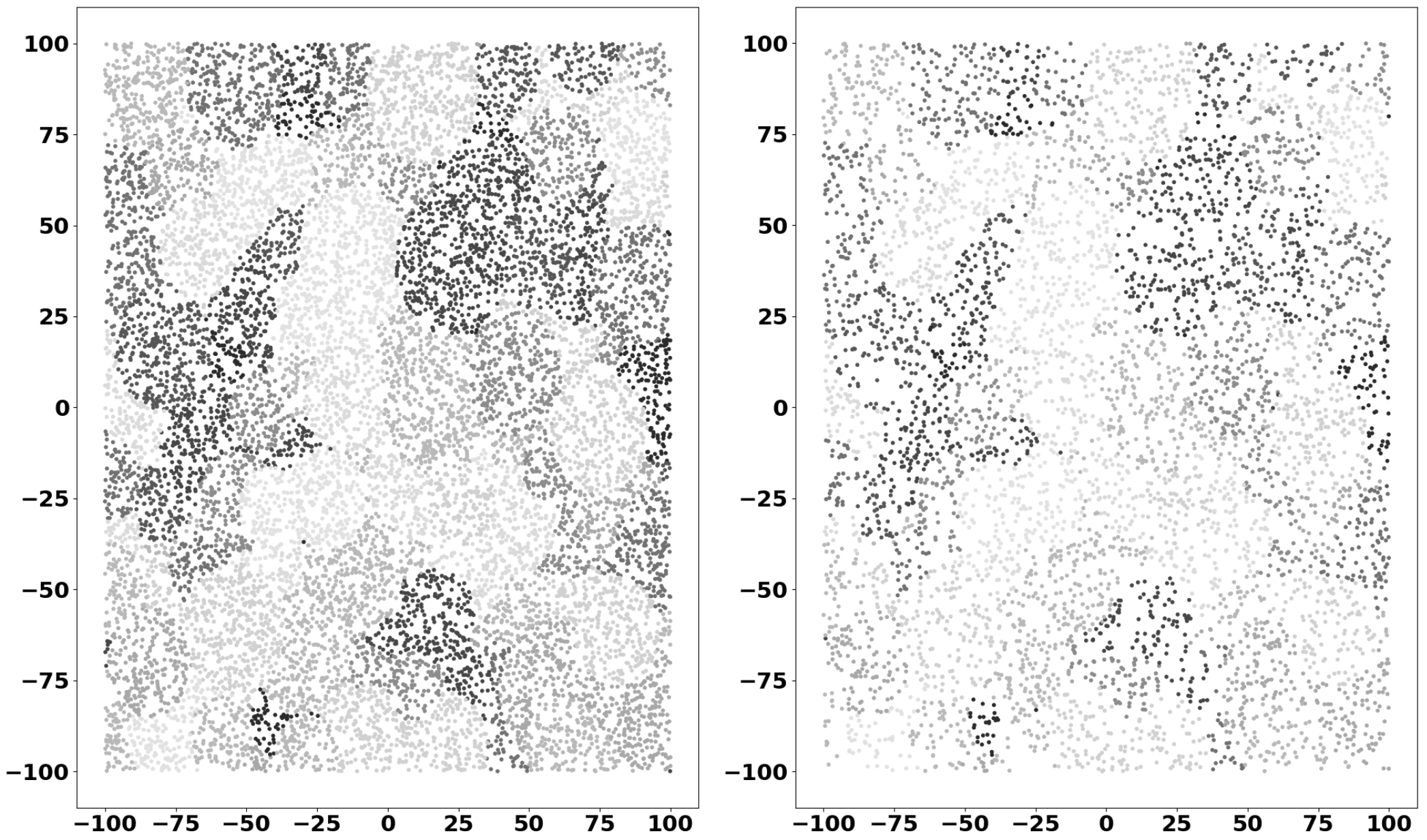

**Figure 8.2:** An equivalent illustration as Figure 8.1 but with ten ($C = 10$) instead of four classes during generation. With the same distribution for possible sphere radii, the two dimensional plane has a similar shattering as in Figure 8.1 but more possible colors to learn.

*The difficulty of the artifically generated data sets actually align across different types of machine learning models.*

We conducted a grid-search over $k$-nearest-neighbor and random forest models with varying $k$ and tree depths. The order of average scores of $k$-nearest neighbors, random forests and deep neural networks based on computational themes align: ▷ < ⨁ < ⊛ < ✠ < ♠ < ♥. This shows that classification datasets with varying difficulty across multiple machine learning models can be generated. The reason for these differences in complexity are 1) varying settings for $\mathcal{U}(u_1, u_2)$ and 2) complicated shaped spaces between spheres from iteratively created centers.

*What's the explanation for varying levels of difficulty? Number of target class labels and sizes of sphere radii.*

The varying complexity of generated datasets can be observed when comparing figures Figure 8.1 and Figure 8.3: with smaller radii the possible class labels are distributed across the same area in a more scattered way. Each class can be thought of as a color (here depicted in grayscale) such that the visual task of the dataset is to associate different circle-shaped areas with one color. An algorithm such as $k$-nearest-neighbor works very natural in such a situation as only a few examples need to be memorised for each circle. The required memory, however, is increased when having more scattered circles.

More possible class targets as illustrated in Figure 8.2 also increase complexity as it gets less likely that neighboring circles share the same label (color) and thus can not be combined by e.g. remembering representative centers.

While the underlying mechanisms of `SpheresUDCR` are not very complex or exciting in context of the learning capabilities of deep neural net-works, different artificially generated datsets with empirically varying difficulty levels support subsequent large-scale experiments. Training deep neural networks on datasets such as `MNIST` takes several minutes with currently standard hardware, depending on the employed training scheme. Training hundreds of thousands of repetitions with varying hyperparameters and repeating experiment settings to analyse the influence of different factors and obtain an impression on the stability of estimations, easily blows up the computation times to weeks or months and requires parallel or even distributed computing approaches. Using `SpheresUDCR` is one of multiple possibilities to strengthen observations and arguments of experiments by reducing the computational overhead.

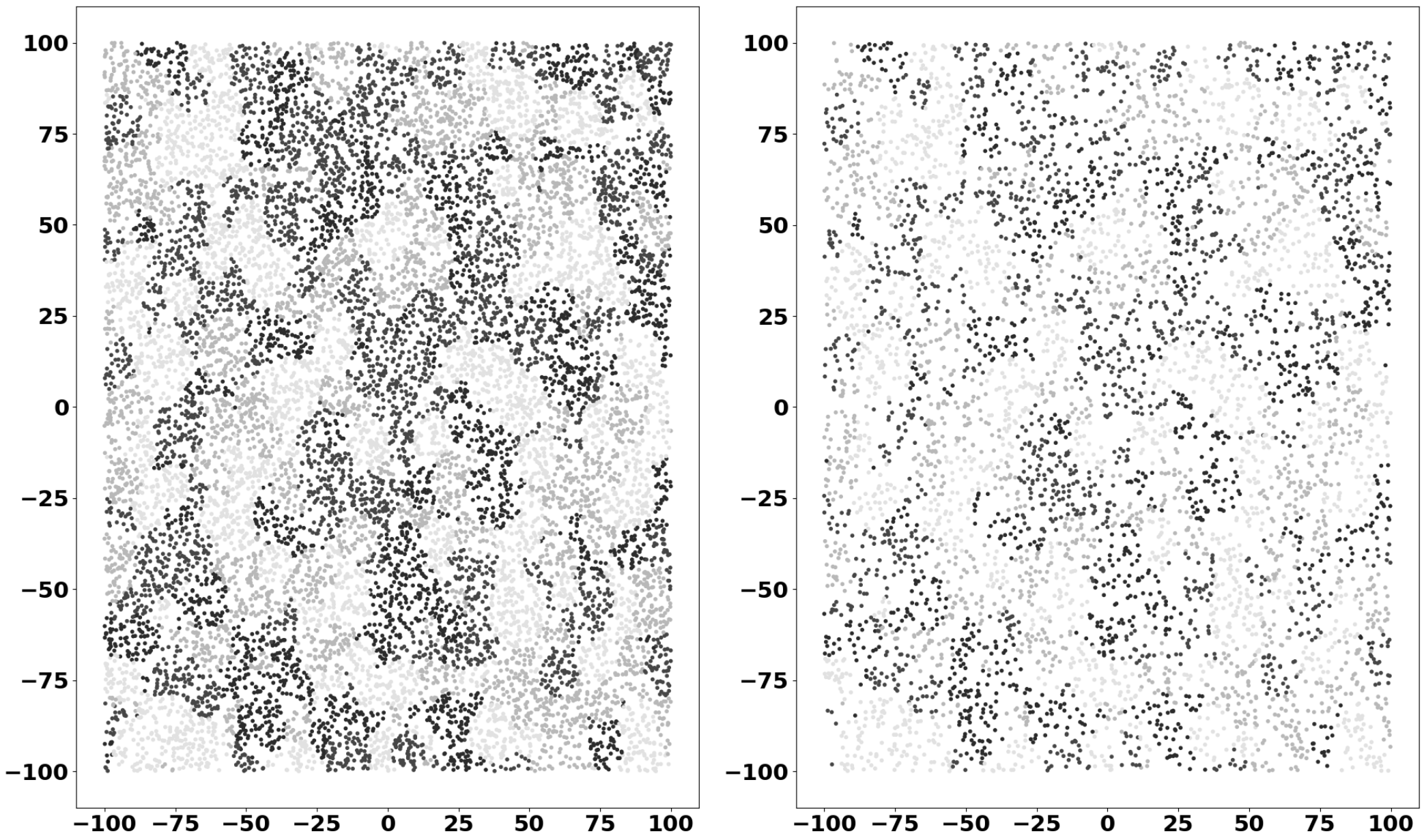

**Figure 8.3:** Smaller radii with $R = \mathcal{U}(5, 10)$. Shapes of circles (or spheres) get less apparent and in a memory-based approach one needs to remember more representative points to recover which points belong to eachother.

*Why use artificially generated data sets? Fast computations and controllable levels of diffulty.*

After conducting a data set generation based on algorithm 5, we assigned uniquely generated identifiers (*uuid*s) to each one of them and persisted them with `python pickle` in a file readable by `deepstruct.dataset.FuncDataset`[1]. Organised by the number of target classes, the input dimensions $D$, the lower and upper bound for sampled radii, the lower and upper bound for sampled $D$-dimensional points (per each dimension) are stored along in a file name such that a similarly configured data set can be recovered. However, if the exact sphere centers and sampled radii are not persisted, new points can not be

1 A minimalistic dictionary that stores the overall data set size, the original input shape and each data point is used by deepstruct.dataset.FuncDataset#L65 to save or load a generated function-based data set.

recovered for a set. With usually 20,000 samples, however, the space – depending on the bounds – is usually sufficiently covered such that the data set can be based on the stored points and still be split into a train-test validation scheme as described in section 5.8.3 on evaluation schemes based on resampling methods.

## 8.3 THE CTNAS DATABASE

The CT-NAS database consists of 863 sampled computational themes for which graph-induced neural networks have been trained and evaluated on multiple target datasets.

The subset of directed acyclic graphs, called computational themes, defined in definition 3 and depicted in fig. 7.2, was motivated by ideas of separating the search space with an infinite set of representative structural themes. We sampled a finite set of exemplary CTs of between order four to six which resulted in a set of 863 graphs in total. Note already, that the probability of choosing a graph of order six uniformly from this finite set is given as $\Pr(|V| = 6) = 0.941$.

The CTs were used to induce architectures, which we practically implemented with `deepstruct` [273]. A hyperparameters *scale* $\in \{500, 1000, 1500, 2000\}$ is used together with a proportion map that transforms each vertex of a CT directed acyclic graph into a layer of neurons. The proportion map was randomly sampled from a dirichlet distribution with a high concentration parameter $\alpha_v = 100$ for each vertex $v \in V(G)$ of the computational theme $G$. That amounts to an almost uniform distribution with a proportion value $\frac{1}{|V|}$ for each vertex. By this construction, a deep acyclic directed neural network is constructed with e.g. $scale = 1000$ neurons distributed across different layers.

We are aware, that, based on the connectivity of the theme $G$ the resulting number of parameters can vary drastically among the resulting neural network models. However, optimizing the architecture upfront such that e.g. four buckets of parameters intervals are defined and then sampling from an induced architecture until a model lying in that parameter interval is found requires a long sampling procedure (as it includes an integer programming problem). With our approach, the number of parameters for each model are counted after construction and can be used for a post-hoc selection. In few experiments we conducted a sampling of architectures restricted to parameter intervals but found that approach often not very beneficial for later analysis.

In a next step, we conducted more than[2] 287,265 model trainings in a distributed and parallel approach. We employed eight datasets, namely `MNIST`, `CIFAR10` and six `SpheresUDCR` sets (described in section 8.2), namely `spheres-0a19afe4` (⊛), `spheres-23aeba4d` (✠), `spheres-6598864b`

2 Due to the fact that we conducted distributed and parallel model training with various hyperparameter settings, some jobs have been faster completed than others such that our database resulted in only roughly uniform proportions for selected parameters.

(⨁), `spheres-b758e9f4` (▷), `spheres-b8c16fd7` (♠), and `spheres-bee36cd9` (♥). All CTs, their network properties, models and the resulting classification performance metrics were collected into a database which we refer to as CT-NAS.

We varied selected hyperparameters such as the learning rate, the batch size, the number of epochs and the scale of each model as to observe their influence and as a sanity check whether the observations align with empirical findings from literature.

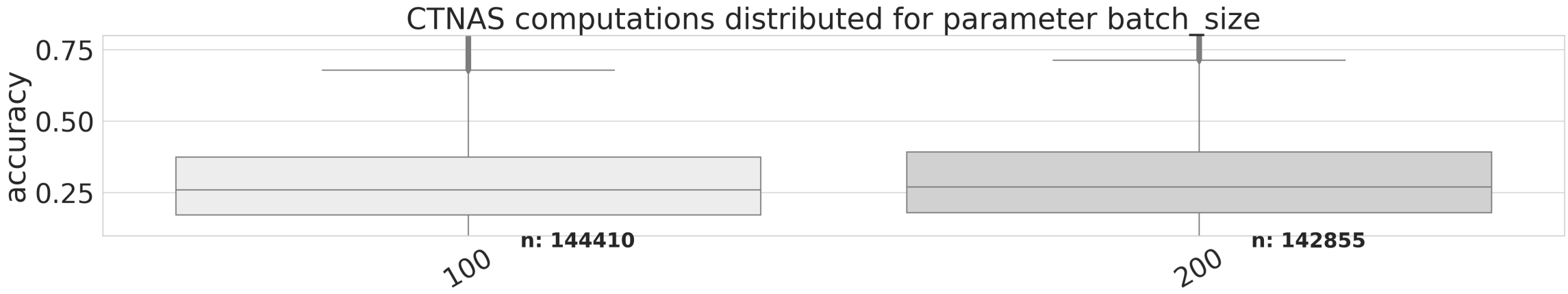

**Figure 8.4:** All models in the CTNAS database have been trained with either a batch size of 100 or 200. We did not observe a significant difference in terms of classification performance or preference o graphs between the employed mini batches.

The batch size in this particular case of the CT-NAS database was set to 100 and 200 and there are almost equally many computed models of each possible batch size value, as can be seen in Figure 8.4. Although there is usually a strong relationship between batch size and learning rate [262], we observe in our reduced setting of an ablation study only a significant influence on switching the fixed learning rate as can be observed in Figure 8.5. State-of-the art training schemes use either decaying learning rates, increasing batch sizes [262] or dynamic variations [261] with schedulers but we sticked to a simple and theoretically solid training scheme because of the complexity of the underlying research field, compare e.g. [74, 234].

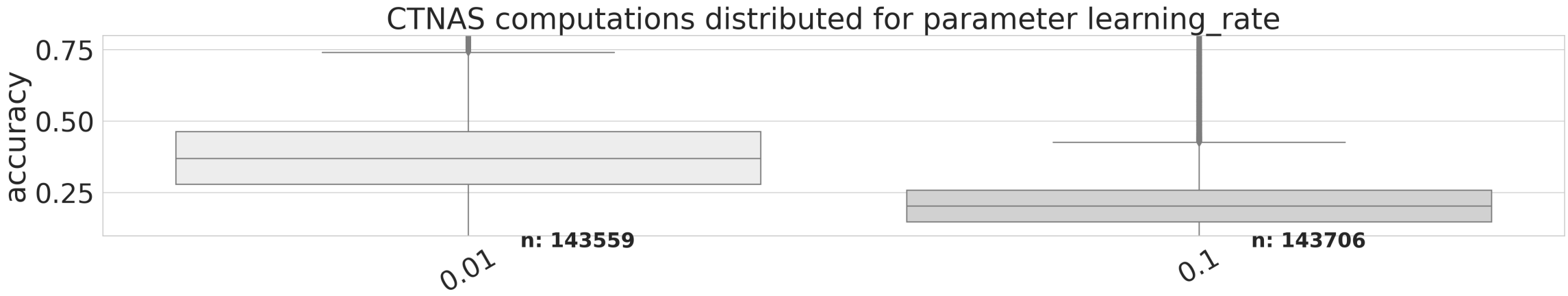

**Figure 8.5:** As often recommended in literature, we also observe a significant influence of smaller learning rates on the final performance with sufficient time to convergence. To keep the computational overhead low, we used 0.1 (which can be considered as too high) and 0.01 (which can be considered as at the recommended upper limit) to incorporate into our ablation study. The learning rate is the only significant hyperparameter we considered that could be immediately decreased to apparently reach better estimations.

It can not be guaranteed that each and every deep neural network realisation in the CT-NAS database properly converged. On the one hand, the always stated argument of non-provable convergence of non-linear models and difficult target problems are one reason. But on the other hand, it has also to be considered that while training processes of current state-of-the-art models are carefully observed and tuned, a strict experimental setting with hundreds of thousands of model trainings can barely be manually tuned. For this reason, fixed possible values for the number of epochs and the scale of each model were chosen, as well. On a macroscopic aggregated level across all datasets, we observe no significant difference for varying number of epochs, as can be observed in Figure 8.6. One might interprete this visual analysis as if all models already properly converged after $n_{epochs} = 50$ epochs but we will see in later detailled analysis on a per target dataset basis and on a graph-level analysis, that the number of epochs are sometimes important to look into or to discard outliers.

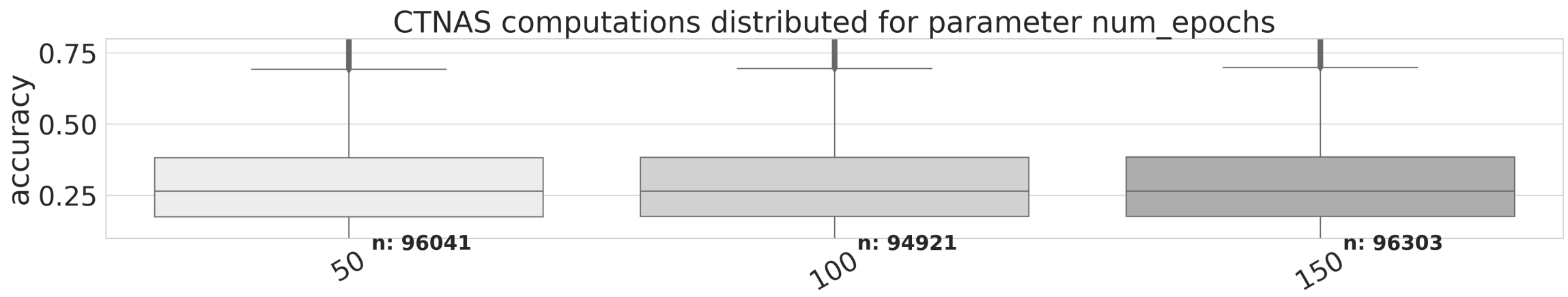

**Figure 8.6:** We used three values $n_{epochs} \in \{50, 100, 150\}$ for the number of epochs. An aggregated view on the accuracy distribution of all computations in the database across all employed datasets does not show any difference and suggests that most models converged already after as few as 50 epochs. When disassembling different architectures and datasets later on, however, there exist definitely outliers that have not reached convergence although different repetitions did so by chance.

A similar observation can be made for the chosen scale values, as illustrated in Figure 8.7. The top performing outliers actually decrease slightly when increasing the scale parameter (a) while also the overall variance of the distribution decreases with larger scale (b).

(a) A decreasing performance of upper quartile outliers are to be taken with a grain of salt: larger neural network realisations are usually more difficult to train which could be one reason that for large scales no comparing maximum outliers are observed. Another reason could be, that larger neural nets often take longer in convergence as the number of parameters is drastically higher and therefore the parametric search space is considerable larger. The phenomenon of a decreased top outlier performance could be partially blamed on the fixed training scheme and has to be understood on a per architecture basis.

(b) In contrast, the reduction in variance of the accuracy (or also $F_1$ score) distribution is a good sign as it could be argued that the estimation is getting more consistent with increased realisations of the same universal architecture.

Both observations are slightly to be seen in Figure 8.7 and are also a conclusive interpretation result of later analysis.

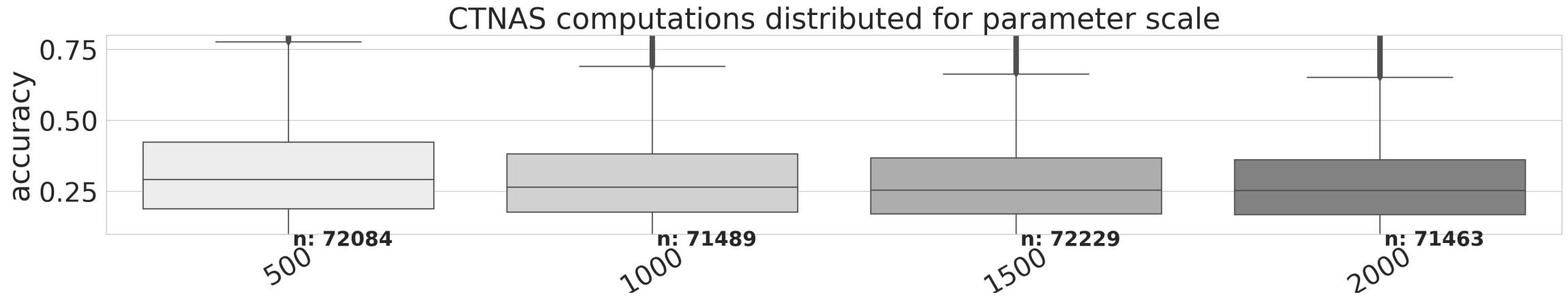

**Figure 8.7:** Similar to the number of epochs as shown in Figure 8.6, the scale seems to have no significant influence on the resulting classification performances. When taking a very detailled look, one can, however, observe that the variance of the estimated performance metric reduces with an increased scale. That is a phenomenon to be expected and when taking detailled looks on a per architecture basis, the scale is actually an important aspect as it directly influences the number of parameters of the underlying deep neural network realisations.

On average, there exist roughly 41 computations per each of the 863 CTs. The exact mean number across the graphs with an aggregated standard deviation can be seen for each dataset in Figure 8.8. Due to the same reasons of distribution, parallelisation and limited computational budget the computed models across datasets are roughly uniformly distributed like the previously stated hyperparameters. With an average standard deviation of about 16 computations, one can usually expect at least nine repeated model trainings per each graph and dataset with a probability of over 95%, assuming that the underlying is normally distributed (independent jobs trained independent models with equal hyperparmeters; $\approx 41 - 2 \cdot 16 = 9$). The $F_1$ score increases for the underlying data sets with `spheres-b758e9f4` (▷) being the most difficult task and `MNIST` being the most simple one.

The average training times for the computed models are provided in Figure 8.9. Most obviously are the higher overall training times for the real-world datasets `MNIST` and `CIFAR10` which justifies our usage of additionally artificially generated sets based on `SpheresUDCR`. But it needs to be emphasised that for `MNIST` there are around 50,000 samples for training while we only used 20,000 for the datasets based on `SpheresUDCR`. Some speedup is therefore be found in fewer data sample points but also lower dimensionality of the problem.

The overall runtime for CT-NAS cost have been 87648582.70 seconds which amounts to 1014.45 days, parallelised over several machines with GPUs and provided by the InnKube Infrastructure Group of the Faculty of Computer Science and Mathematics at the University of Passau.

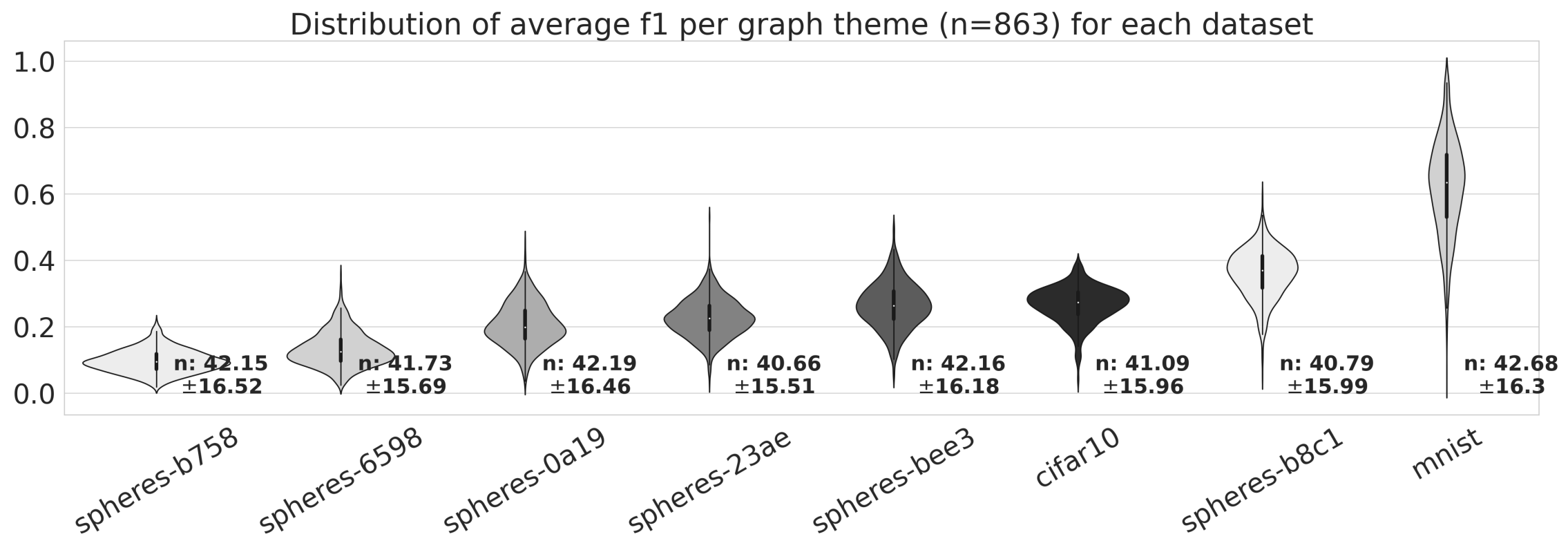

**Figure 8.8:** The 863 CTs induced 287,265 models that were trained on eight datasets. For each dataset the number of average samples per each graph within the group is denoted together with its standard deviation. This means, that there are sufficiently many repetitions for each configuration to derive order statistics of the estimated performance metrics. The violin plots are ordered based on the average $F_1$ score obtained for each dataset such that our models yield a difficulty ranking: `spheres-b758e9f4` (▷) < `spheres-6598864b` (⨁) < `spheres-0a19afe4` (⊛) < `spheres-23aeba4d` (✠) < `spheres-b8c16fd7` (♠) < `CIFAR10` < `spheres-bee36cd9` (♥) < `MNIST`

Pure training time has been 87406490.52 seconds or 1011.65 days which amounts to 99.72% of the overall computation time required. The numbers should highlight how much computational resources not only the CT-NAS database but also several other experiments in this work and the field of NAS require. Data will be made available as a `python` package and distributed through `pip` under github.com/innvariant/ctnas.

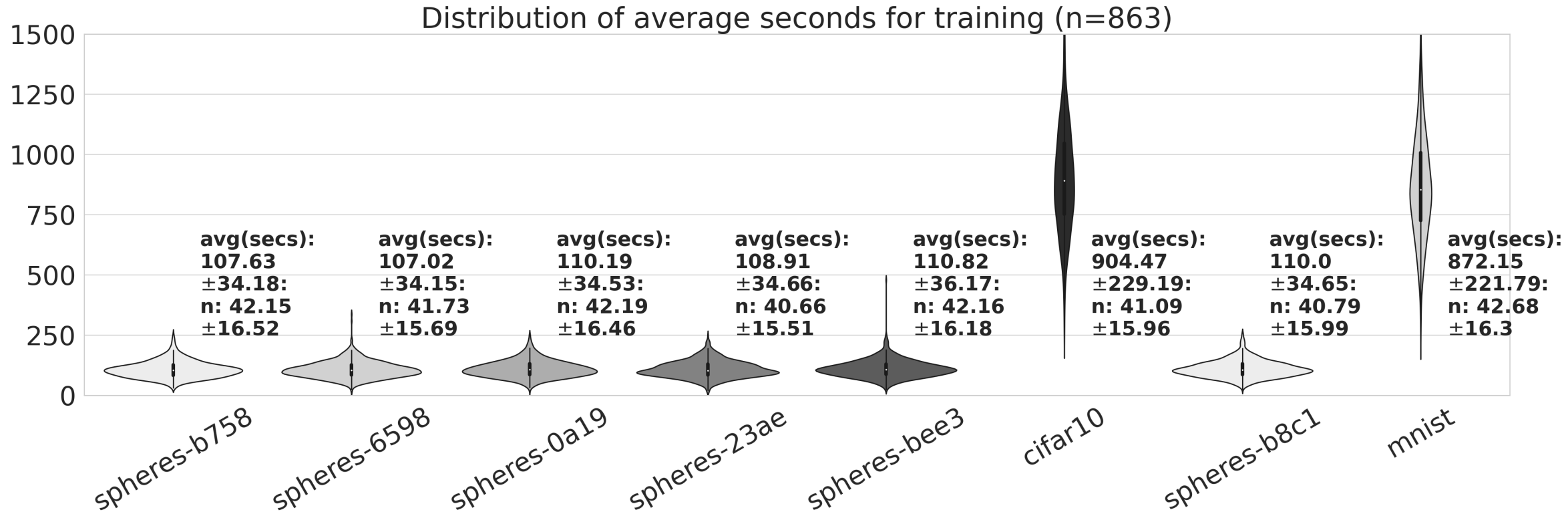

**Figure 8.9:** Aggregated statistics for the overall invested training times to obtain estimates for the 863 graphs and associated models on the employed datasets. The overall computation time amounts to ≈ 1014.45 days, parallelised over multiple machines to obtain results in reasonable time. These amounts of repetitions and insights into stabilities of training schemes are currently hardly affordable for state-of-the-art deep learning models (having more than millions of parameters).

# 9

# BASICS OF STRUCTURE ANALYSES

*The following entails:*

We're interested in ways of analysing the structure of deep neural networks and tackle the questions formulated in research complex II in section 1.1.5. Pearl [218] gives an insightful summary on studying the causal relationship between different factors. This motivates us to depict the elements of structure analysis in a causal diagram in fig. 9.1. Association analysis is on the first of three rung in Pearl's ladder of causation analysis [218]. Structure analysis means bringing the properties of the first object of our study, graph structures, into association with the second object of our study, neural networks.

We consider structure analysis as difficult because of its many influential factors and a stochastic nature of many of these factors. To obtain measurements of performances, training time or inference speed as properties of a neural network, the model has to undergo a training pipeline that involves many steps and components such as an inital seed for random number generation, a dataset, its augmentation, an optimiser with a schedule and hyperparameters, or an evaluation scheme.

The most prominent result of this analysis is that, indeed, different structures yield significantly different neural network realisations. The factor *Graph* depicted on the left side in fig. 9.1 is clearly a causal influence to properties such as performance measures or energy consumption on the right side of the causal diagram. What is this factor *Graph* assembled of? Are there certain network theoretic properties, that influence the target properties significantly?

Recall our three overarching questions on the analysis of neural network structures:

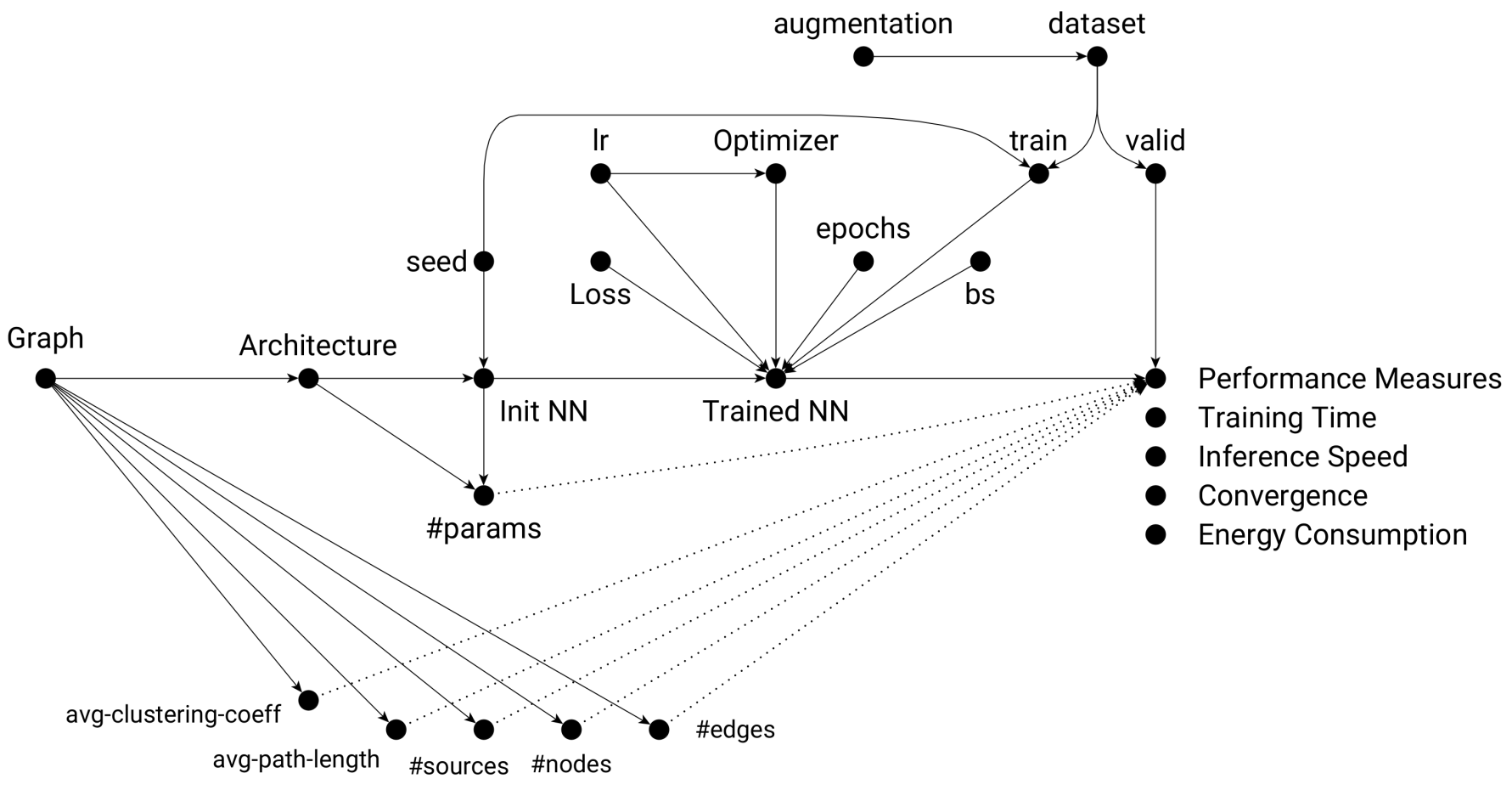

**Figure 9.1:** A causal diagram of selected factors that influence the estimation of various deep neural network properties. The graph induces an architecture from which an initial neural network model is drawn. Many factors such as the loss formulation or optimisation pipeline influence the trained model. Finally, properties such as estimated correctness measures or inference speed are based on separate data within an evaluation scheme. Note, that this diagram is far from complete but illustrates the amount of graph properties, the complexity of factors in a training pipeline and the model properties as factors of interest.

**⋄⋄ Research Complex II**

Do neural network models differ in terms of structure or are they just different with respect to the choice of other hyperparameters such as activations, training epochs, data sampling or augmentation strategy, optimisation procedure, or the choice of loss function? (see section 1.1.5)

We will see that they do differ, regardless of whether we consider a search space of layer-wise neural networks as in section 9.2, a search space of random generated graphs as in section 9.4, or a search space of fully specified CTs as in section 10.1. Differences in structure can also be observed on varying application data domains as with `MNIST`, `CIFAR10` or artificially generated data from `SpheresUDCR`. Employing popular benchmarks for neural architecture searches such as `NAS-Bench-101` also confirm this observation. However, the difference is difficult to disentangle from other causal factors and behave non-linear in relationship to e.g. the employed dataset - something, which is not surprising considering that different datasets come from different natural processes.

◇◇ Research Complex II

Are well studied network theoretic properties involved in the behaviour of neural networks? What are common structures or patterns that can be observed? (see section 1.1.5)

There exist both simple and complex network theoretic properties that influence the outcome of a neural network realisation. As the graph order is directly correlated with the parametric size of the realisation, this graph property has of course influence on the approximation quality. But we observe also interesting correlations for the average path length distribution and performance and robustness evaluations of neural networks, which we reported in [6, 272]. In-depth analysis of the CT-NAS database in chapter 10 shows that structural groups can be found but these groups are very task-dependent.

◇◇ Research Complex II

How could analytical results look like to guide further architecture development? (see section 1.1.5)

Answers to the question on network theoretic properties suggest to use automatic feature aggregations over simple selective network properties for advancing automated methods of neural network design. In the case of automatic feature aggregations, further analysis of structural properties would not be required. This approach, however, does not exclude the possibility, that there are certain combinations of network theoretic properties that could favorably support the design of neural network architectures. The chosen properties are too simple as to close analytic approaches to explain relationships between neural network realisations and their graph structure.

One further analytical result is a first observation in Chapter 10 of a phenomenon of different types of *change in estimation complexity* for the search space in which a neural architecture search is conducted. This could lead to search algorithms with dynamic performance estimation strategies chosen based on the expected type of estimation complexity.

Results from structure analysis of neural networks therefore do not solely rely on correlations between network properties and performance scores but could also yield deeper insights on the formulated search space design.

## 9.1 ELEMENTS OF NEURAL NETWORK STRUCTURE ANALYSIS

To dive into analysing the relationship of neural networks and their structure, we need to restrict our analysis to a search space design (as explained in section 7.2), come to a choice of training scheme and select a target evaluation measure. In cases of explicit structure analysis, we even need to make choices about structural properties, that we are interested in.

Experiments involving a structure analysis include estimations of candidates in the search space and then relating structural properties with the obtained evaluation measures. For extensive insights, full evaluations of the search space are necessary but are computational very expensive. Individual experiments need to balance the appropriate choice of a (small) search space, the extent to which candidate estimates are reliable in terms of the cost of training and evaluation schemes, and the number of structural properties to compare.

Evaluation measures such as the classification accuracy are given linear in the number of testing samples, while methods to estimate robustness measures might be more complex. Energy consumption requires proper and patient timing while it is given at no additional computational complexity.

While the term *architecture* commonly refers loosely to the functional form and its successive operations of a neural network (realisation), definition 2 on page 114 in section 6.3 captures the term of a *universal architecture* $\mathcal{A}([T])$ based on a structural theme $T$. The proposed definition culminated in section 7.2 on search space design with definition 3 on page 134 of computational themes as an example for an infinite set of structural themes. But CTs as a definition for $\mathbb{S}$ are not the only way of distinguishing between (universal) architectures.

For distinguishing between universal architectures, we require at least two structural themes that are different under some notion of e.g. (set-wise) graph similarity. Leading examples include layer-wise themes for neural networks (compare section 7.2.1), conv-like benchmarks (compare section 7.2.2), and computational themes (CTs) (compare section 7.2.3). With at least two elements in $\mathbb{S}$, we notice that also the search space combinatorics are an interesting study for types of architectures.

In the simplest case of $|\mathbb{S}| = 2$, two types of universal architectures can be easily kept apart - both on a formal basis as of how the individual structural theme is defined and their structural properties relate, and on an experimental basis of how their evaluation measures behave and relate.

From studies on universal approximation theorems, we can identify further types of architectures: flat, wide neural networks [43, 106] are captured by layer-wise themes in section 7.2.1 on page 132 but arbitrarily deep and thin [126, 216] are not. **Deep-and-thin** neural networks can be defined as $[T_{\mu=k_0,\omega=k_1}] = \{G_{w,l} \mid \forall l \in \mathbb{N}, \forall w \in \{\mu+1,\dots,\omega\}\}$ with $0 < k_0 < k_1 < \infty, k_0, k_1 \in \mathbb{N}$ being the lower-bound and maximum layer width allowable for all neural network realisations contained in $\mathcal{A}([T_{\mu=k_0,\omega=k_1}])$. For $\mu = \omega - 1$, we can also omit one parameter and have a $k_1$-fixed-width infinite-depth universal architecture $\mathcal{A}([T_{\omega=k_1}])$.

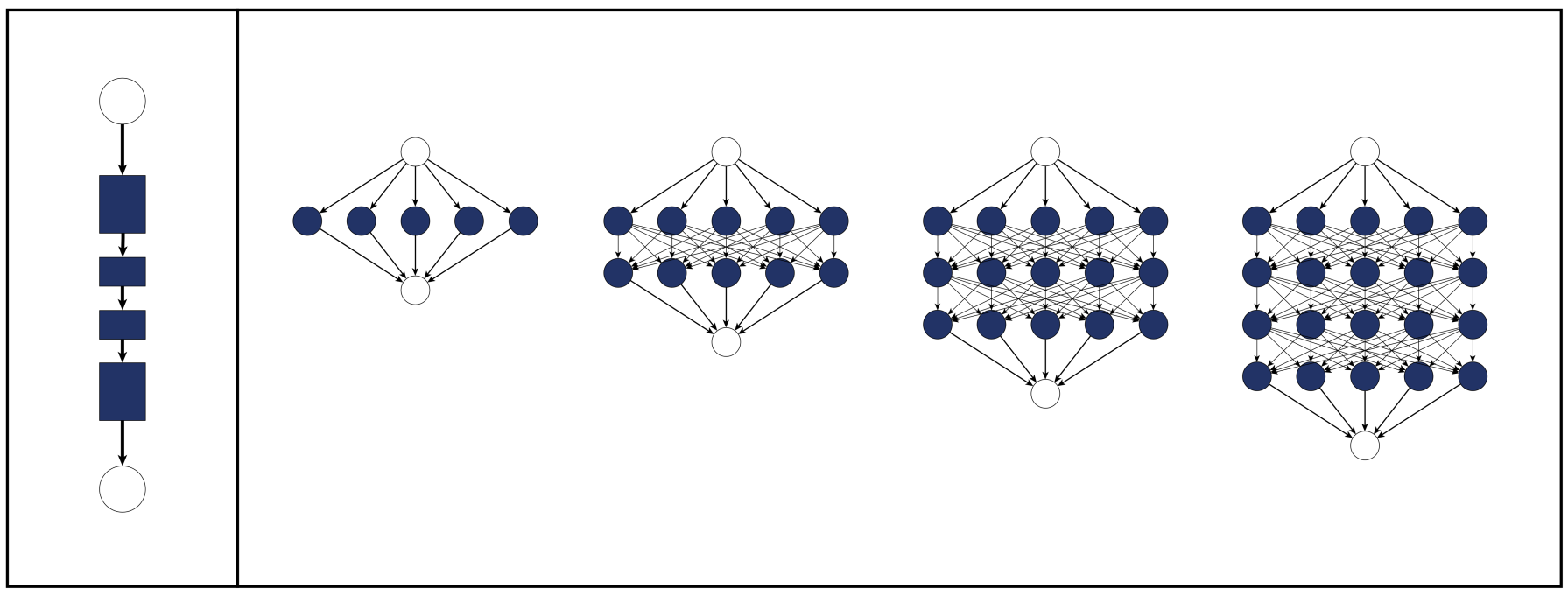

**Figure 9.2:** Visual sketch of thin but infinitely deep neural networks (left) and exemplary realisations in the set $[T_{\omega=5}]$. The realisations have a minimum and maximum width per layer as to provide UAP depending on a particular dataset or function to approximate. $T_{\omega=5}$ is seen on the left as a candidate of the minimum realisations and we associate a more depictive graph (left) with it to illustrate the definition of the equivalence relation to have arbitrary depth.

This work puts structure as a first-class citizen into the definition of neural networks but when it comes to different types of neural architectures in the commonly used understanding, higher-level operations (e.g. convolutions [147] or pooling [231, 247]) and other techniques (e.g. batch [113] or layer [11] normalisation), that can not be captured as easily in this framework, play an important role, as well. They are treated in two ways:

1. Either they are incorporated into the definition of the universal architectures such that they play a key role in how structural themes induce an architecture. A straight-forward approach to incorporate this information is to use labelled (di)graphs instead of unlabelled graphs.

2. Or the techniques are kept as part of outside factors (compare the causal diagram in Figure 9.1) of the overall training and evaluation pipeline in the same manner as hyperparameters such as optimisation algorithms or pre-processing techniques.

There are various more possibilities to design different types of deep neural architectures or whole search spaces $\mathbb{S}$. Based on graphs such designs often include labelled vertices or edges with labels containing higher-level operations such as convolutions, poolings or normalisations. Also, residual or skip-layer connections can be viewed as one type of deep neural architecture or each underlying graph as candidate of a larger $\mathbb{S}$. Under this viewpoint, one can compare ResNet-like architectures with deep-and-thin architectures or subsets of these two types. This makes a structure analysis very individual and difficult to compare across experiments.

Our proposal is to formalise the experimental setting in the light of structural themes such that each type of architecture for comparison at least satisfies the condition of being a universal architecture. However, a structure analysis then still only makes sense if the underlying structural properties are rich enough to differentiate between the two analysed types. This does often not apply, i.e. the structural properties of deep-and-thin and non-sparse neural networks are very distinguished in comparison to structural richness of ResNet-like architectures. A comparison of few types of universal architectures with low variety of structural properties therefore hardly gives a rich result for interpretation except for an argument such as *Type A is better than Type B in view of Measure C*.

*Summary of Elements of Structure Analysis*

The chosen types of deep neural architectures as implicit candidates of an analysed search space $\mathbb{S}$ have a huge impact on what results can be expected of structure analytical experiments.

REMARKS

- There are infinite ways to design a search space $\mathbb{S}$ for further structure analysis. We consider layer-wise DNNs, architectures from `NAS-Bench-101`, CTs, and randomly sampled graphs for experiments as described in section 7.2.
- Experiments need to balance out different aspects such as search space cardinality or richness of structural properties in the search space definition as to make an analysis possible.
- Small search spaces make in-depth comparisons easier but can not capture as much structural richness as search spaces that contain extensive candidates with diverse properties.

## 9.2 EXPERIMENT: LAYER-WISE NEURAL NETWORKS

We consider the initial question whether structure even makes a difference under the simple case of layer-wise structural themes as described in section 7.2.1.

Our experiment contains 80381 unique trained neural network realisations in $\mathcal{A}^{(1)}, \ldots, \mathcal{A}^{(30)}$ on multiple machines with GPUs. The experiment is based in a grid-search manner on the following properties: the dataset (`MNIST`, `CIFAR10`, and `SVHN`), the learning rate in [0.3, 0.1, 0.02, 0.01, 0.001], the number of training epochs in [10, 20, 30, 40, 50], the size of the batch in [50, 100, 150, 200, 250, 300, 350, 400, 450, 500], the number of layers in [2,...,30], the parameter initialisation distribution (one of *kaiming-uniform*, *kaiming-normal*, $\mathcal{N}(0; 0.1)$, *xavier-uniform*, and *xavier-normal*, compare section 5.2.3), the activation function in [ReLU $\sigma^{\diagup}$, TanH $\sigma^{tanh}$, Sigmoid $\sigma^{\sim}$], and possible layer sizes which start with 50 up to (and including) 1000 in a step size of 50, i.e. [50, 100, 150, ..., 1000]. Parameters were uniformly chosen from the possible configuration space.

> Does structure make a difference?
>
> **Search space $\mathbb{S}$:** Layer-wise (7.2.1)
> **Method:** We take layer-wise neural networks as structural themes, create neural network realisations with various hyperparameters, conduct training and compare number of layers with accuracy.
>
> **Data:** Employing real-world image classification datasets `MNIST`, `CIFAR10`, and `SVHN`.
>
> **Interpretation:** Structure in the sense of using different number of layers makes a difference. Disentangling the relationship between structural factors and further hyperparameters such as the used activation function, learning rate or the batch size for training is difficult.

During training we captured the total time for each computation, including its construction and training time. This resulted in a minimum time of 35.5026 seconds, a median of 199.8599 seconds, an average time of 262.3379±193.3392 seconds and the maximum time took 1526.829 seconds for one computational run. The experiment was distributed across multiple machines.

We recorded the final training and test accuracy after all epochs of a computation had been completed, and recorded the average of the last five accuracy scores during training. Further, we kept the proportions of layer widths, the bottleneck layer width, and the widest layer size. The insights of this experiments are summarised in figs. 9.3 to 9.5 and additional resources can be found in appendix A.2.

The visuals in fig. 9.3 and fig. 9.4 exemplify the strong negative relationship between the number of layers and the accuracy. The x-axis always shows the number of layers of the feed-forward neural network realisation and the y-axis a distribution of measured accuracies across all realisations with the same number of layers and with varying underlying hyperparameters. We restricted the samples to only ones with significant accuracy with a lower bound, i.e. on `SVHN` to above 65% or on `CIFAR10` to above 30%. The classifiers then did in fact learn *something* as they do not randomly choose from the ten possible classes of the particular data set, although they of course do not compare with state-of-the-art classification models on these datsets.

With the high number of computations with uniform sampling of hyperparameters we tried to reduce possible effects of conflicting influences. We analysed non-structural influences such as the learning rate, which, when increased yields better results on average. This effect is even visible when grouping it by all employed batch sizes. An increased batch size within one fixed learning rate reduces the accuracy, while the largest batch size of 500 with a higher learning rate still outperforms a small batch size of 200 with a smaller learning rate. While this relationship, however, breaks outside these studied bounds, i.e. with learning rates of or above 0.3, it can be observed across all three datasets.

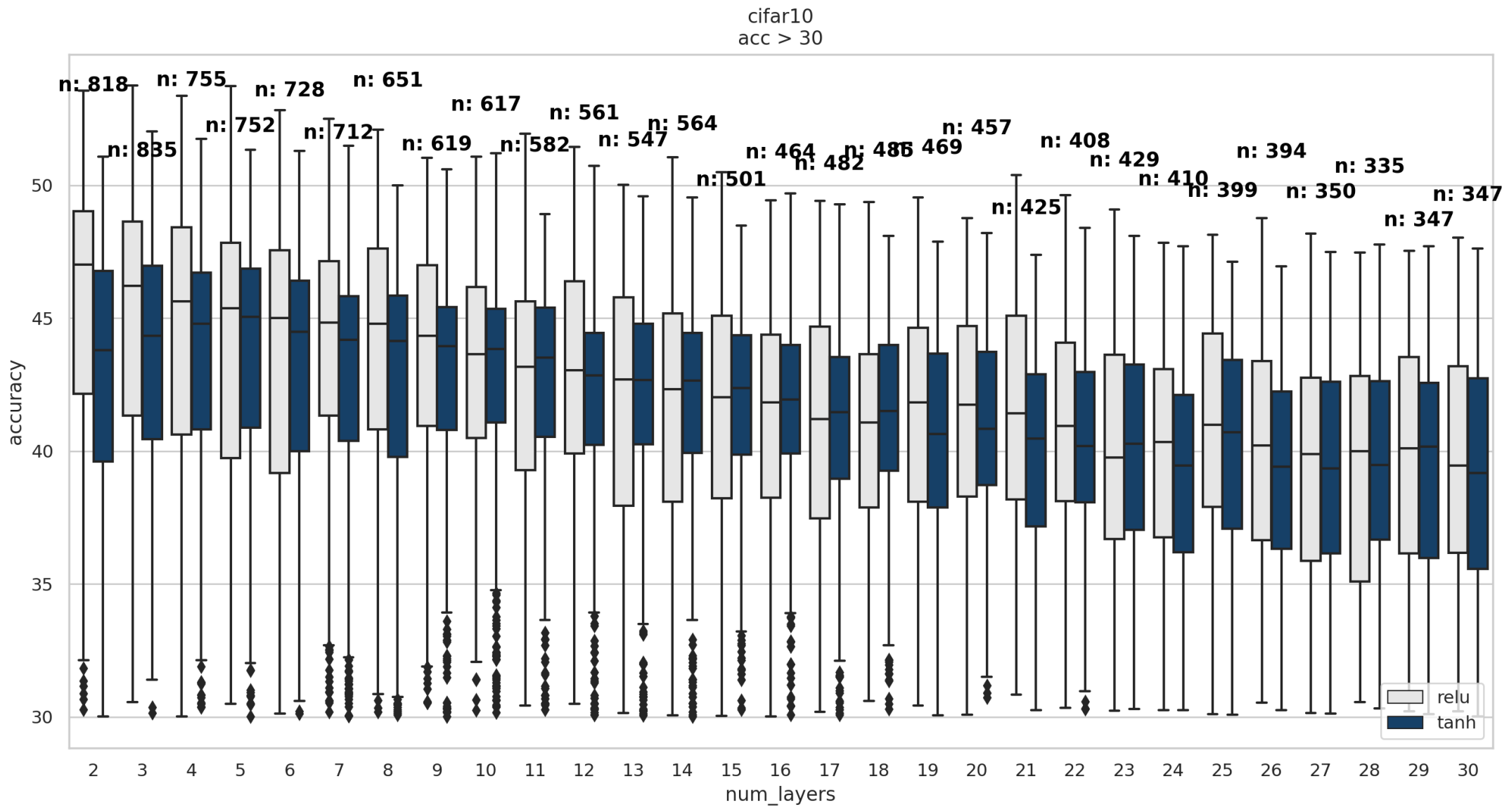

**Figure 9.3:** In comaprison to `SVHN` in fig. 9.4 or `MNIST` in fig. A.4, `CIFAR10` exhibits a stronger relationship with an increasing number of layers. While for `MNIST` the top outliers of very deep realisations can still be compared with shallow neural networks, for `CIFAR10` our training scheme and grid search could not find comparable realisations. An explanation is that the complexity of `CIFAR10` is not only higher but with a larger space of possible solutions it gets even more difficult to find high-performing realisations.

Main insight of the experiment is that the structural parameter of layer sizes has a significant influence on the classification accuracy. Other factors as described in the causal diagram of fig. 9.1 have been considered but a significant effect of the factor layer number should be present, although it varies across different datasets. We suppose that a main cause for the influence of the number of layers relies in the fact that with an increasing number of the architecture induced with such a structural theme, the space of the architecture itself grows significantly. Recall, that layer-wise structural themes as defined in section 7.2.1 have an inclusion-property, i.e. the architecture with more layers obviously contains all previous architectures.

An example for another important factor in this experiment is the case of information bottlenecks as summarised in fig. 9.5. We noticed, that a small bottleneck leads to slightly worse classification performances. This derives logically from the universal approximation theorems which suggest to have layer-widths larger than the input dimensionality to achieve low training error. But first sampling the layer size randomly and then filling it with random layer widths also leads to higher probability of having small bottlenecks for deeper themes. This is an alternative explanation for the decrease in accuracy as can be e.g. observed in fig. 9.3.

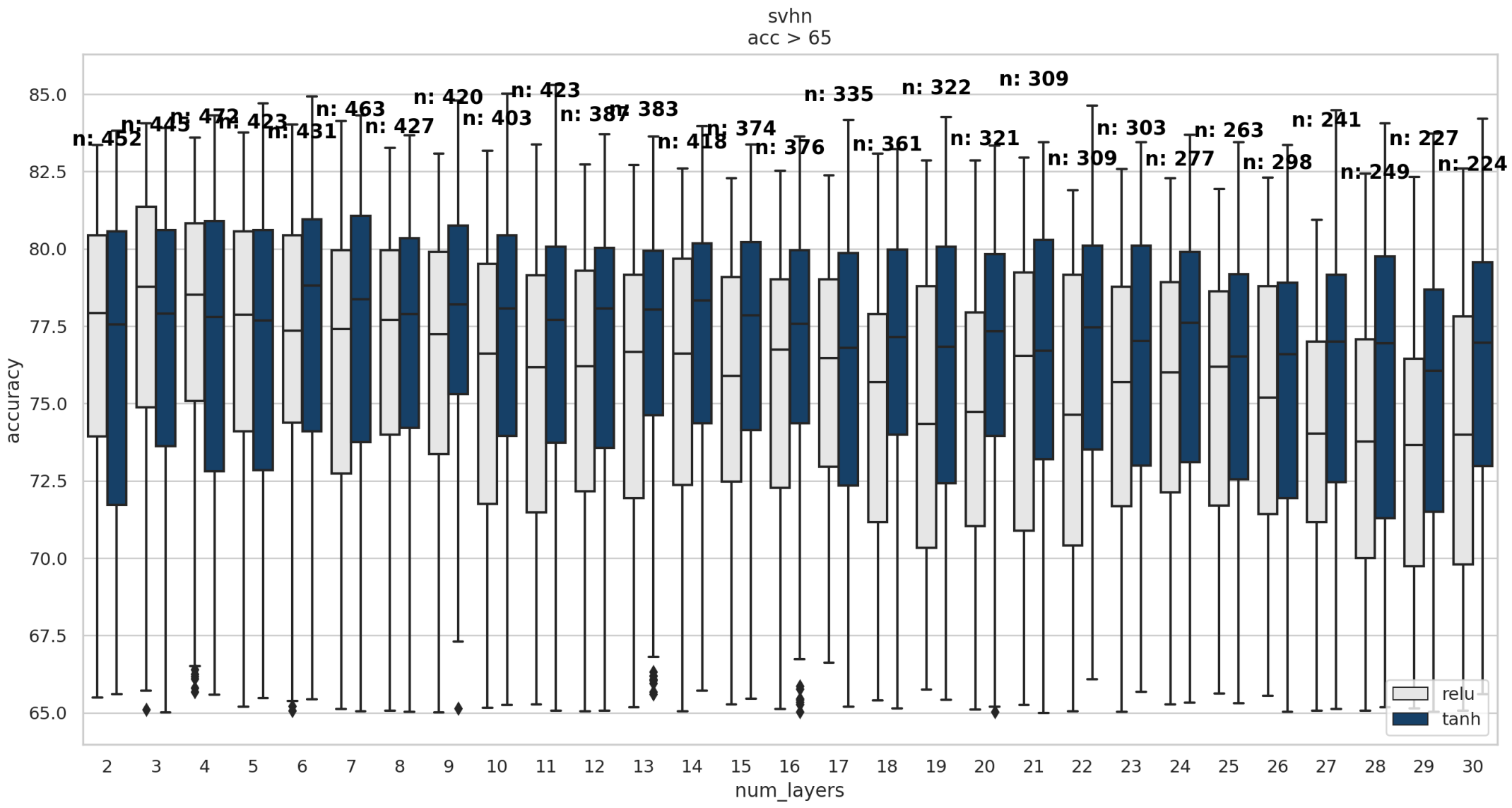

**Figure 9.4:** The choice of activation function has a significant influence on the relationship between layer depth and accuracy. For shallow networks, rectified linear unit is on average better than TanH $\sigma^{tanh}$while this trend changes with more layers such that $\sigma^{tanh}$outperforms ReLU on average for deeper layers. Note, that we did not make use of normalisation layers such as batch normalisation or layer normalisation.

We therefore also reduced the samples post-hoc to larger bottlenecks for above a width of 300 and found that the overall phenomenon of decreasing accuracy for very deep feed-forward neural networks still holds in our experiment setting. However, the decrease in accuracy does not take place as early as suggested by fig. 9.4 but exhibits a small bump of first increasing slightly with few layers before decreasing slightly with a lot layers. The decrease is not as significant as with small bottlenecks which gives an alternative explanation to the size of the architecture based on the sampling strategy of neural network realisations representing the architecture.

These conflicting influences of linear correlation analyses and unexplored structural properties with different resulting interpretations is a very important insight for subsequent analyses and a reason why the empirical results have to be studied very carefully.

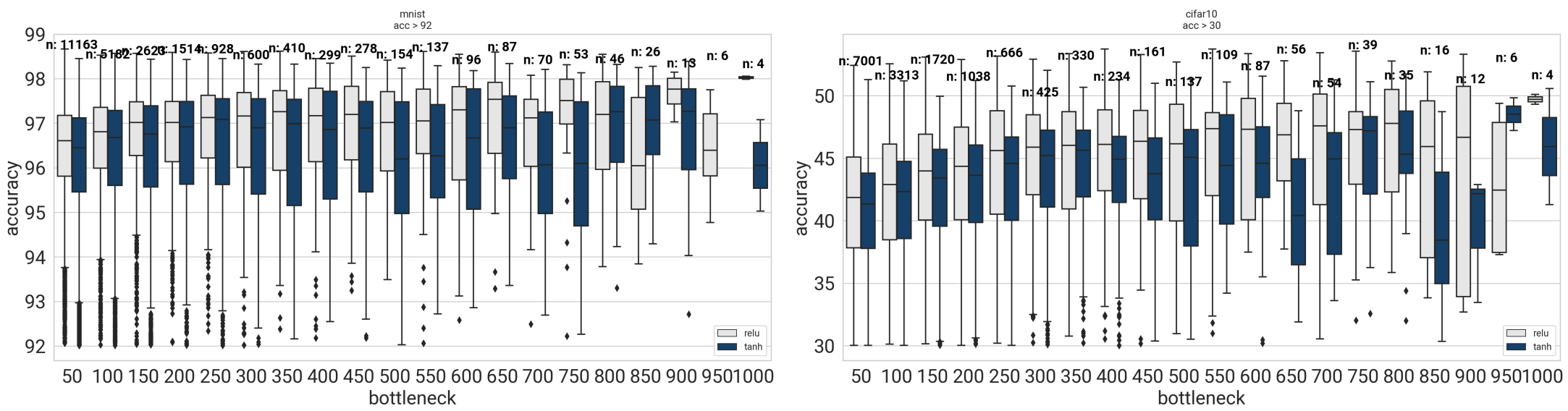

**Figure 9.5:** These two visulisations summarise additional analyses on more fine-grained relationships between structural properties and the accuracy of the trained neural networks. The *bottleneck* determines the width of the smallest layer in a neural network realisation, i.e. all information to conduct the classification task needs to pass through a compressed representation of this dimensionality. It can be observed that a certain minimum bottleneck is required to find a high-performing neural network with higher probability. Sampling layer widths uniformly after sampling the number of layers means that it is also more likely to obtain a small bottleneck – which is an additional explanation for why more layers are less likely to have high accuracies.

REMARKS

- Layer-wise deep neural networks (DNNs) as structural themes have few structural properties but already exemplify the difficulty of structural analyses: observed effects on a performance score could come from structural properties of the universal architecture or from the sampling strategy on how a performance score was estimated for taking part in the representation of a structural theme.
- A possible explanation for differences of layer-wise DNNs could be found in the growing architecture space.
- It might be of interest in structure analysis to not have strong subset-relationships between structural themes to uncover structural differences of the search space $\mathbb{S}$.

## 9.3 RELATED WORK ANALYSING STRUCTURES OF DEEP NEURAL NETS

Interest in structures of deep neural networks can historically be found under terms such as *sparsity* and in the fields of *pruning* [87, 92, 115, 150, 199, 232, 276] (compare chapter 11) or *neuroevolution* [195, 255]. However, these works majorly focussed on NAS-methods and less on the analysis of structures found by such methods. A graph- or network-theoretic analysis became only recently of interest and we summarised the literature on it in table 9.2. Whether such an analysis even makes sense is a central question of chapter 6 of this work and the results and arguments of the field can still be considered to be in its infacies.

There are many more properties than presented in table 9.1 on page 170 to bring in relation with correctness or robustness measures. For exam-ple, Waqas et al. propose *topological entropy* and *Ollivier-Ricci curvature* [163] to identify robust neural architectures [296]. Topological entropies (or indices) can be obtained in various ways as is done e.g. for molecular graphs by Manzoor et al. [188].

REMARKS

- Structure analysis of deep neural networks becomes more and more popular.
- The topological representation for a DNN heavily differs among different works.
- Some insights such as path length or degree variances seem to contribute in predicting properties of DNNs.

## 9.4 EXPERIMENT: DEEP NEURAL NETWORKS WITH RANDOM PRIOR STRUCTURES

In research complex II in section 1.1.5 we asked whether deep neural networks (DNNs) differ in terms of structure and with graph-induced networks in chapter 6 we made a construction with different structures explicit. This led us to hypothesise that there might exist universal architectures $\mathcal{A}([T_1])$ based on a random structural theme $T_1$ which are not perfectly fitted by neural network realisations of another architecture $\mathcal{A}([T_2])$ based on a different theme $T_2$.

Does structure make a difference?

**Search space $\mathbb{S}$:** Random Graphs (7.2.4)
**Method:** We take a search space definition with at least two structural themes such as CTs, create neural network realisations for these themes, conduct training with a fixed scheme and few varying hyperparameters and compare the themes under resulting evaluation measure distributions such as obtained $F_1$ score or energy consumption.

**Data:** Randomly initialised deep neural networks based on different DAG serve as approximation targets (artificial data)

**Interpretation:** All approximations reach significant low training and testing error, supporting our training scheme. Analytical arguments show that the difficulty of the objective defined by randomly initialised deep neural networks is very simple and explains why a solution can be quickly found.

For the experiment setup, we randomly sampled graphs from the Watts-Strogatz model, the Barabasi-Albert model, balanced trees and grids of a limited width and depth. Weights and biases have been ini-tialised with draws from the normal distribution $\mathcal{N}(0;0.5)$. Between consecutive layers, layer normalisation and the selected activation functions ReLU, Sigmoid, Hardshrink, Hardswish, LeakyReLU, LogSigmoid, and GELU were used. Each randomly initialised neural network realisation with a randomly chosen candidate graph of a structural theme served as either a target function for approximation or as a source model to approximate another target.

We observed, that the randomly initialised neural networks were almost always fitted by other neural networks with different structures. A visual analysis like in fig. 9.6 further led us to hypothesise about properties of randomly initialised neural network realisations serving as target functions for approximation. The analysed landscapes of randomly initialised neural networks of varying underlying graph structures appear to be surprisingly smooth and simple in two-dimensional cases. We concluded, that a randomly initialised neural network realisation resembles almost always quite simple (linear) function landscapes.

While the UAP of a universal architectures guarantees that any target could theoretically be approximated sufficiently close, we do not observe this behaviour of good approximations for real-world datasets. The original analysis of backpropagating errors for exploding or vanishing gradients of Hochreiter [101] inspired our reasoning for this phenomenon of simple function landscapes: repeated applications of an activation function over weighted sums whose parameters are drawn from $\mathcal{N}(0;0.5)$ stay in a very narrow domain on expectation.

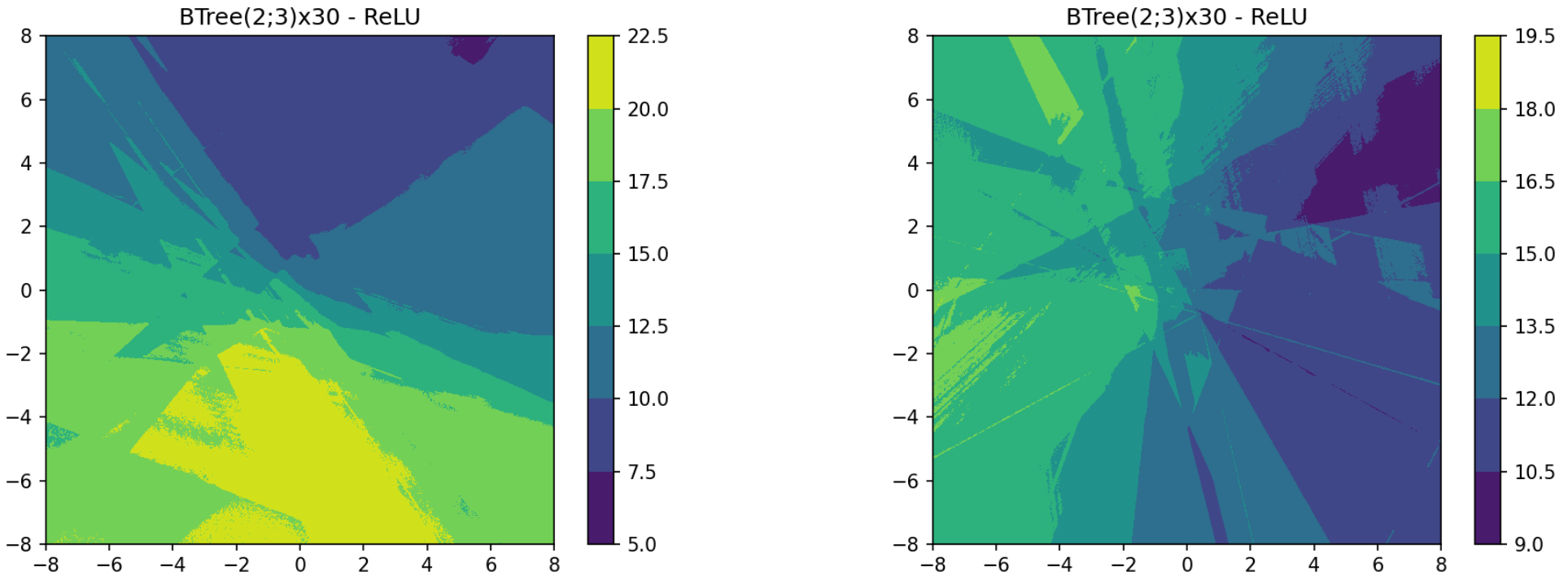

**Figure 9.6:** An aggregated landscape of a randomly initialised graph-induced neural network. The underlying structure of these two examples are balanced trees of branching factor $r = 2$ and height $h = 3$ and ReLU served as activation function. In sum, 30 repetitions of independently initialised neural network models of the same graph are aggregated into this landscape visualisation as to inspect a repeated smoothness across the landscape when re-using the same underlying structure.

Recall the inference of a graph-induced neural network in eq. (6.3) on page 105 with the mentioned activation functions:

$$\mathbf{z}^{(0)} \triangleq entr(x), \ \mathbf{h}^{(0)} \triangleq \mathbf{z}^{(0)}$$
$$\mathbf{z}^{(l)} \triangleq \sum_{s=0}^{l-1} W^{s \to l} \mathbf{h}^{(s)} + B^{(l)},$$
$$\text{with } \mathbf{h}^{(l)} = \sigma(\mathbf{z}^{(l)}) \qquad \text{for } l \in \mathcal{L}$$

For weights & biases drawn from $\mathcal{N}(0; 0.5)$ we observe for input $\mathbf{x}$ that $\mathbb{E}(\mathbf{z}^{(0)}) = \mathbb{E}(W_{entr}\mathbf{x}+B_{entr}) = 0$ and further $\mathbb{E}(\mathbf{z}^{(1)}) = \mathbb{E}(W^{0\to 1})\mathbb{E}(\mathbf{z}^{(0)})+\mathbb{E}(B^{(1)})$ which is passed through the activation function. Depending on the activation, we find for ReLU, Hardshrink ($\sigma_{Hardshrink,\lambda=0.5}$), and GeLU easily that $\mathbb{E}(\mathbf{h}^{(1)}) = \mathbb{E}(\sigma(\mathbf{z}^{(1)}) = 0$ (for the sigmoid $\sigma^{\sim}$ it would be $\mathbb{E}(\mathbf{h}^{(1)}) = \frac{1}{2}$). In other words, for randomly initialised networks the expectation and variance of the output of a graph-induced neural network is zero – like mostly found for other formulations of deep neural networks.

REMARKS

- Randomly initialised networks with weights & biases centered around zero produce very simple function landscapes.
- All considered randomly sampled graphs, used for inducing an architecture, were able to quickly approximate objectives defined by other deep neural networks induced from a second graph.
- The experiments serve as sanity check for our training scheme and the experiments align with arguments derived from analytical considerations.

## 9.5 RELATING STRUCTURE AND CORRECTNESS

A straightforward approach for the analysis of structures of deep neural networks in research complex II is to study the correlation between properties detailled in table 9.1 on page 170 and correctness or robustness measures of deep neural networks. In a series of studies we [6, 269, 272] and many others since [26, 42, 284, 319, 327] searched for approaches to relate structural and performance properties of DNNs. We did so extensively in Stier & Granitzer [272], Ben Amor et al. [6], and Stier et al. [269] and will summarise these results here. While the formulations highly vary and make comparisons and clear resulting arguments difficult, we found first correlations on which further theories could be built on.

Are network properties involved in the cause and if so which?

**Search space $\mathbb{S}$:** Random Graphs (7.2.4)
**Method:** 1) Conduct a linear correlation analysis between structural properties and evaluation measures, 2) use interpretable non-linear models such as Random Forests to relate structural properties and evaluation measures, and 3) use similar methods on clustered structural groups based on e.g. graph similarities.

**Data:** Using CTNAS database on eight datasets including `CIFAR10`, `MNIST`, and `SpheresUDCR` or independently train realisations in separate experiments.

**Interpretation:** Obvious properties such as parameter size of realisations show clear correlations and serve as sanity check. Small correlations can be found for e.g. variances of average path length distributions, which suggests that diverse path lengths through non-linearities are often beneficial for fitting to a target task.

These studies are concerned with the second research question stated in section 1.1.5:

⋄⋄ Research Complex II

Are well studied network theoretic properties involved in the behaviour of neural networks? What are common structures or patterns that can be observed?

In various ablation studies, we found variances of certain network theoretic properties to correlate with performance or robustness measures.

With 10,000 randomly sampled graphs as structure of feed-forward neural networks trained on `MNIST`, a correlation analysis with the resulting dataset `graphs10k` [272] was conducted by learning regression models to predict the test accuracies of models based on selected graph properties. The overall data set contains 5018 graphs from the Barabasi-Albert model and 4982 graphs from the Watts-Strogatz model [272, Sec. 4.1]. Ordinary least squares linear regression, support vector regression and random forests have been used as regression models.

Our results [272] are recapitulated in table 9.3 and put into context of this work. Six different sets of graph properties were analysed. The overall full feature set is denoted with $\Omega$ and contains all graph properties listed in table 9.3, starting with the number of vertices and ending with the standard deviation of the edge betweenness centrality distribution for a graph (compare section 3.2.2 for details on graph properties). The subsequent five sets are subsets of $\Omega$ and denoted $\Omega_{np}$, $\Omega_{op}$, $\Omega_{var}$, $\Omega_{small}$ and $\Omega_{min}$:

$\Omega_{np}$ contains only graph properties which do not directly relate to the parametric capacity, i.e. $\Omega_{np} = \Omega$ {#vertices, #edges, #source vertices, #sink vertices} $= \Omega \;\; \Omega_{op}$.

$\Omega_{op}$ contains **o**nly graph properties which directly relate to the **p**arametric capacity, i.e. $\Omega_{op}$ = {#vertices, #edges, #source vertices, #sink vertices}.

$\Omega_{var}$ contains only variances of properties, i.e. variance of the degree distribution etc.

$\Omega_{small}$ contains a manually selected set of properties, namely $\Omega_{small}$ = {#vertices, #sink vertices, degree variance, density, neighborhood variance, path length variance, closeness std, edge betweenness std}.

$\Omega_{min}$ contains a minimal found subset of $\Omega$ which was found to be still sufficient to achieve a high coefficient of determination.

Each prediction model was trained 20 times with different train-test splits [272, Sec. 4.2] and the coefficient of determination is used to summarise the quality of predicted test accuracies for each graph (and its underlying models).

The results exemplify similar findings of [104, 269] in which the number of vertices showed a positive correlation and several variances of e.g. node betweenness and closeness [269, Tbl. 1, p.8] showed negative correlation with the test accuracy of reccurent neural network realisations.

Overall, it can be observed that the predictability of performance metrics based solely on structural information is surprisingly high – considering that there is no information about data contained. It remains open how these properties can be better described as the analyses show a non-linear and intertwined influence among multiple properties. In chapter 13 we turn our focus on only the predictability of performance metrics as to advance NAS-methods. With graph neural network it is not even necessary to engineer concrete graph properties for good predictability [300] if an explanation for performance on a structural level is not expected.

| Graph Properties | Model Properties |
|---|---|
| Number of vertices / edges | Number of parameters |
| Densities/Sparsities | Input Shape (size) |
| Degree distribution | Output Shape (size) |
| Path Length Distributions | Non-Linearities |
| Centrality Distributions | Weights & Biases |
| Clustering Distributions | |
| Triangles / Graph-Minors | **Technical Properties** |
| Diameter | Memory footprint |
| Edge Connectivity | Training time |
| Spectral properties | Storage size |
| | Energy Consumption |
| **Evaluation Properties** | |
| **Correctness** | **Robustness Measures** |
| **Losses** | Adversarial Examples |
| MAE, MSE, RMSE | Error Rates |
| **Predictors** | CLEVER score [303] |
| Accuracy | **Pipeline Stats** |
| Precision & Recall | Required Samples |
| Confusion Matrix | Required Epochs |
| $F_1$ score | |
| r2-score | |
| **Generators** | |
| Inception Scores / Distances | |
| Embedded $\alpha$-Precision, $\beta$-Recall, Authenticity [3] | |

**Table 9.1:** Along with the causal diagram in fig. 9.1, we differentiate between four major types of properties for deep neural network architectures and their associated realisations: **graph**, **model**, **technical** and **evaluation** properties.
Search space design decides about major graph properties that shape neural network models and groups them. But the relationship between graph and model properties is not always one-way. Training determines weights & biases and can influence structural graph properties after training.
Analysing the structure is mostly concerned about relating properties on top such as graph and model properties with evaluation or technical properties.

| Title | Year | Ref |
|---|---|---|
| Neural persistence: A complexity measure for deep neural networks using algebraic topology | 2018 | [233] |
| **Structural analysis of sparse neural networks** | 2019 | [272] |
| Exploring randomly wired neural networks for image recognition | 2019 | [319] |
| Graph structure of neural networks | 2020 | [327] |
| **Correlation Analysis between the Robustness of Sparse Neural Networks and their Random Hidden Structural Priors** | 2021 | [6] |
| Convolutional neural network dynamics: A graph perspective | 2021 | [289] |
| Exploring robustness of neural networks through graph measures | 2021 | [297] |
| A Study on the Ramanujan Graph Property of Winning Lottery Tickets | 2022 | [213] |
| Leveraging the Graph Structure of Neural Network Training Dynamics | 2022 | [290] |
| Understanding the dynamics of dnns using graph modularity | 2022 | [171] |
| Exploring robust architectures for deep artificial neural networks | 2022 | [296] |
| Efficient Sparse Networks from Watts-Strogatz Network Priors | 2023 | [284] |
| Peeking inside Sparse Neural Networks using Multi-Partite Graph Representations | 2023 | [42] |
| Beyond multilayer perceptrons: Investigating complex topologies in neural networks | 2024 | [26] |

**Table 9.2:** Structure analysis of deep neural networks can be considered a quite young and recent field of study. Prior to 2018, most studies were focusses on sparsity but with the emergence of more and more complex architectures questions around structure appeared not only from a neuro-evolutionary or neural architecture search perspective but also from an empirical perspective, trying to understand the influence of discrete structures, imposed on deep learning functions, itself.

| **Model** | $\Omega$ | $\Omega_{np}$ | $\Omega_{op}$ | $\Omega_{var}$ | $\Omega_{small}$ | $\Omega_{min}$ |
|---|---|---|---|---|---|---|
| **Random forest** $R^2$ | **0.9714** ±0.0009 | 0.9314 ±0.0031 | 0.9664 ±0.0010 | 0.9283 ±0.0032 | **0.9710 ±0.0011** | **0.9706 ±0.0009** |
| OLS $R^2$ | 0.8631 ±0.0042 | 0.8476 ±0.0053 | 0.6522 ±0.0089 | 0.5968 ±0.0103 | 0.7211 ±0.0072 | 0.6907 ±0.0077 |
| $\text{SVM}_{lin}$ $R^2$ | 0.8620 ±0.0045 | 0.8463 ±0.0054 | 0.6432 ±0.0114 | 0.5551 ±0.0164 | 0.6943 ±0.0101 | 0.6676 ±0.0111 |
| $\text{SVM}_{rbf}$ $R^2$ | 0.9356 ±0.0018 | 0.9235 ±0.0028 | 0.8561 ±0.0041 | 0.7827 ±0.0092 | 0.8781 ±0.0045 | 0.8604 ±0.0033 |
| $\text{SVM}_{pol}$ $R^2$ | 0.9174 ±0.0025 | 0.8998 ±0.0032 | 0.5668 ±0.0065 | 0.6839 ±0.0117 | 0.8421 ±0.0053 | 0.7543 ±0.0083 |
| **Property** | **Feature Importance** Score of the RF-model | | | | | |
| #vertices | 0.0009 | | 0.0045 ±0.0002 | | | |
| #edges | 0.0013 | | 0.0071 ±0.0002 | | | |
| #source vertices | 0.0636 | | 0.0720 ±0.0018 | | 0.0643 ±0.0019 | 0.0659 ±0.0019 |
| #sink vertices | 0.9124 | | 0.9164 ±0.0019 | | 0.9137 ±0.0019 | 0.9139 ±0.0020 |
| degree mean | 0.0004 | 0.0083 ±0.0054 | | | | |
| degree variance | 0.0009 | **0.4582** ±0.0973 | | 0.3685 ±0.0866 | 0.0024 ±0.0002 | 0.0069 ±0.0002 |
| diameter | 0.0003 | 0.0008 ±0.0001 | | | | |
| density | 0.0007 | 0.0030 ±0.0007 | | 0.0072 ±0.0007 | 0.0023 ±0.0002 | |
| eccentricity (mean) | 0.0015 | 0.0047 ±0.0006 | | | | |
| eccentricity (var) | 0.0022 | **0.3025** ±0.1006 | | 0.3401 ±0.0994 | | |
| eccentricity (max) | 0.0003 | 0.0009 ±0.0002 | | | | |
| neighborhood (mean) | 0.0004 | 0.0050 ±0.0046 | | | | |
| neighborhood (var) | 0.0011 | **0.1417** ±0.0646 | | 0.2434 ±0.0883 | 0.0025 ±0.0001 | |
| neighborhood (min) | 0.0005 | 0.0272 ±0.0104 | | | | |
| neighborhood (max) | 0.0017 | 0.0071 ±0.0035 | | | | |
| path-length (mean) | 0.0011 | 0.0045 ±0.0013 | | | | |

| **Model** | $\Omega$ | $\Omega_{np}$ | $\Omega_{op}$ | $\Omega_{var}$ | $\Omega_{small}$ | $\Omega_{min}$ |
|---|---|---|---|---|---|---|
| path-length (var) | 0.0013 | 0.0052 ±0.0009 | | 0.0133 ±0.0018 | 0.0034 ±0.0001 | 0.0067 ±0.0002 |
| closeness (min) | 0.0014 | 0.0067 ±0.0028 | | | | |
| closeness (mean) | 0.0010 | 0.0030 ±0.0003 | | | | |
| closeness (max) | 0.0013 | 0.0034 ±0.0003 | | | | |
| closeness (std) | 0.0021 | 0.0064 ±0.0012 | | 0.0166 ±0.0017 | 0.0039 ±0.0002 | 0.0066 ±0.0002 |
| edge betweenness (min) | 0.0001 | 0.0008 ±0.0003 | | | | |
| edge betweenness (mean) | 0.0011 | 0.0041 ±0.0005 | | | | |
| edge betweenness (max) | 0.0015 | 0.0038 ±0.0004 | | | | |
| edge betweenness (std) | 0.0011 | 0.0027 ±0.0003 | | 0.0109 ±0.0011 | 0.0036 ±0.0001 | |

**Table 9.3:** Selected subsets of structural properties have been used to predict resulting test accuracies of feed-forward neural networks based on different random graphs. The data `graphs10k` is presented by Stier et al. in [272] and reflects one of several independent experiments to determine whether and which structural properties are capable of predicting performances. The coefficient of determination $R^2$ is provided for each used prediction model and a standard deviation over 20 repetitions is provided. Feature importance scores for each structural property are obtained from the employed random forest model to provide insights into the influence of each property under varying property subsets for prediction.

It can be clearly observed that the number of vertices and especially the number of sink vertices of the directed acyclic graph are the strongest predictors – because they are directly related to the number of parameters. If parameter information is omitted as with the subset $\Omega_{np}$ the degree distribution, the eccentricity (compare diameter of a graph), and the neighborhood connecitvity become the most important predictive features. We interpreted this as indicators for high performing deep neural networks making use of different paths for information flow. Varying path lengths and fanning neurons seem to provide benefits over fixed path lengths and constant-sized connectivity patterns.

REMARKS

- Structural properties that directly relate to the number of parameters of a model need to be thoroughly considered or excluded from analysis as to focus on the structural aspect of an analysis instead of the parametric capacity – the latter easily overshadows more detailled structural properties.
- Random forests and other interpretable predictors have the advantage of giving insights on the (non-linear) relationship between explicit structural properties and performance metrics of DNNs – we do not only know that the performance can be predicted but also learn about which properties contribute to it.
- Varying degrees and path lengths seem to be beneficial for feed-forward neural networks.

## 9.6 STRUCTURAL PROPERTIES OF ROBUST NETWORKS

Switching from correctness to robustness measures might influence the importance of certain structural properties. We conducted experiments with random graphs and their structural properties on the robustness of DNNs in Ben Amor et al. [6], similar to the correlation analyses in [104] and [269]. The properties and associated results which showed strongest correlations with robustness objectives of the trained DNNs hinted us to variances of path lengths, again, but also gave mixing results overall [6].

Interestingly Waqas et al. found in similar experiments that the "robustness of trained DNNs highly correlated with their graph measures of entropy and curvature" [296, p. 8]. They noted, however, that this relation "has a strong dependence on the complexity of task and model" [296, p. 8]. We agree on the observation of strong dependencies on task and model and still consider the hypothesis that most structure is supposedly dependent on individual application domains and its data.

There is future work in structure analytical studies on the robustness of DNNs but due to the additional computational complexity of measuring robustness under varying perspectives we majorly sticked to correctness and error scores.

## 9.7 EXPERIMENT: ENERGY CONSUMPTION OF CTNAS-MODELS

Neural networks undergo multiple phases from research & development until their deployment in production-ready systems. Many state-of-the-art neural network systems require such large technical environments that even medium-sized research institutes with several hundred GPUs and terabytes of storage barely manage to train them from scratch. The energy and cost involved in obtaining a fully trained neural network model of billions to trillions of parameters is difficult to estimate. Because of these scales, energy consumption of neural networks during training and even more during inference is an important factor.

Are network properties involved in energy consumption and if so which?

**Search space $\mathbb{S}$:** Computational themes (7.2.3)
**Method:** We train neural network realisations for CTs, capture their energy consumption during training and evaluate their classification performance in terms of accuracy, precision, recall and $F_1$ score.

**Data:** Models are trained on real-world image classification datasets `MNIST` and `CIFAR10`.

**Interpretation:** Clear correlations between accuracy, energy consumption and number of parameters can be observed, i.e. more parameters seems to imply more energy consumption. Other strong relationships include variances of layer sizes and variance of the degree.

Over a period of several months, we captured the energy consumption of a single machine with a VOLTCRAFT SEM6000 energy cost and consumption monitoring device. The machine HP EliteDesk 800 G3 TWR run a Intel Core i7-7700 CPU @ 3.60GHz x 8 with a NVIDIA GeForce GTX 1080/PCIe/SSE2 on a Ubuntu 22.04.2 LTS 64-bit. The monitoring device captured the energy consumption of only the main eletrical input of the machine (not attached monitors), e.g. the main board, disks, CPU, and GPU. Energy was captured hourwise and models have only been trained within hourwise windows that can be clearly associated with the repeated trainings or inferences during that window.

With manual observation, a power consumption between 0.028 and 0.029 kWh could be observed for the idling machine. To exclude time windows which contained actual computations, a restriction of the consumption between 0.01 and 0.04 kWh yielded a median of 0.031 kWh, a first quartile 0.029 kWh and a third quartile of 0.035 kWh. This observation confirms a stable idle power consumption. For further analysis the overall captured power consumption instead of any subtraction with the idle base consumption is used because we can safely assume that the real consumption is linearly related.

We created a set $\mathcal{G}^{energy}$ of randomly sampled directed acyclic graphs (DAGs) for analysing the energy consumption of more than 254 DAGs of orders between nine and eleven:

- $\mathcal{G}^{energy} = \{G_1, \dots, G_n\}, \forall i \in \{1, \dots, n\} : G_i \sim \texttt{UDAG}$
- $num_vertices \in \{9, 10, 11\}$

For each graph, a *scale* $\in \{100, 200, 300, 400, 800\}$ was initially uniformly chosen but later extended to be sampled also from $\{1000, 1200\}$ after a first analysis showed possibly meaningful correlations between structure and energy consumption for larger scales. The *scale* parameter was used to sample a neural network instance with `deepstruct`[1] [273] for which each vertex was assigned a proportion of the available scale as a number of parameters. For a three-layered DAG and a scale of 300 that would simply result in almost all cases in assigning each layer 100 neurons .

The power consumption across different scales for all trainings and tests of all monitored neural network models based on these DAGs can be observed in Figure 9.7. There is a clear correlation between model scale and energy consumption. The associated numbers denote the underlying number of models (samples) used for visualizing the energy distribution per scale. A comparative correlation can be found in Figure 9.8 for models of only scale 800 between the number of model parameters and the energy consumption, e.g. either row-4 column-5 or row-5 column-1. This should be not surprising and we consider this clear correlation as a sanity-check for our experiments in that there exists a clear dependence between more computational parameters and required energy for the computation.

Aggregation functions *min*, *max*, *median*, *mean* and *variance* have been used to bring non-scalar properties from distributions into a scalar feature format. The final features are called

*graph_num_vertices* The overall number of vertices of the graph, i.e. its order.

*graph_num_edges* The size of the graph.

*num_sources* The number of vertices with no incoming edges.

*num_sinks* The number of vertices with no outgoing edges.

*num_hidden* The number of vertices which are not source or sink vertices. This number does *not* contain the number of neurons of the resulting neural network based on the graph.

*num_layers* Given a topological sort of the DAG, this number represents the number of layers of the layered DAG.

*num_paths* The number of shortest paths of the graph.

*density* Simply the density of the graph.

*degree* The *degree_min*, *degree_max*, *degree_median*, *degree_mean*, and *degree_var* based on aggregations of the degree distribution of the graph.

1 We used ScalableDAN.build() from `deepstruct` 0.10.0 for this experiment.

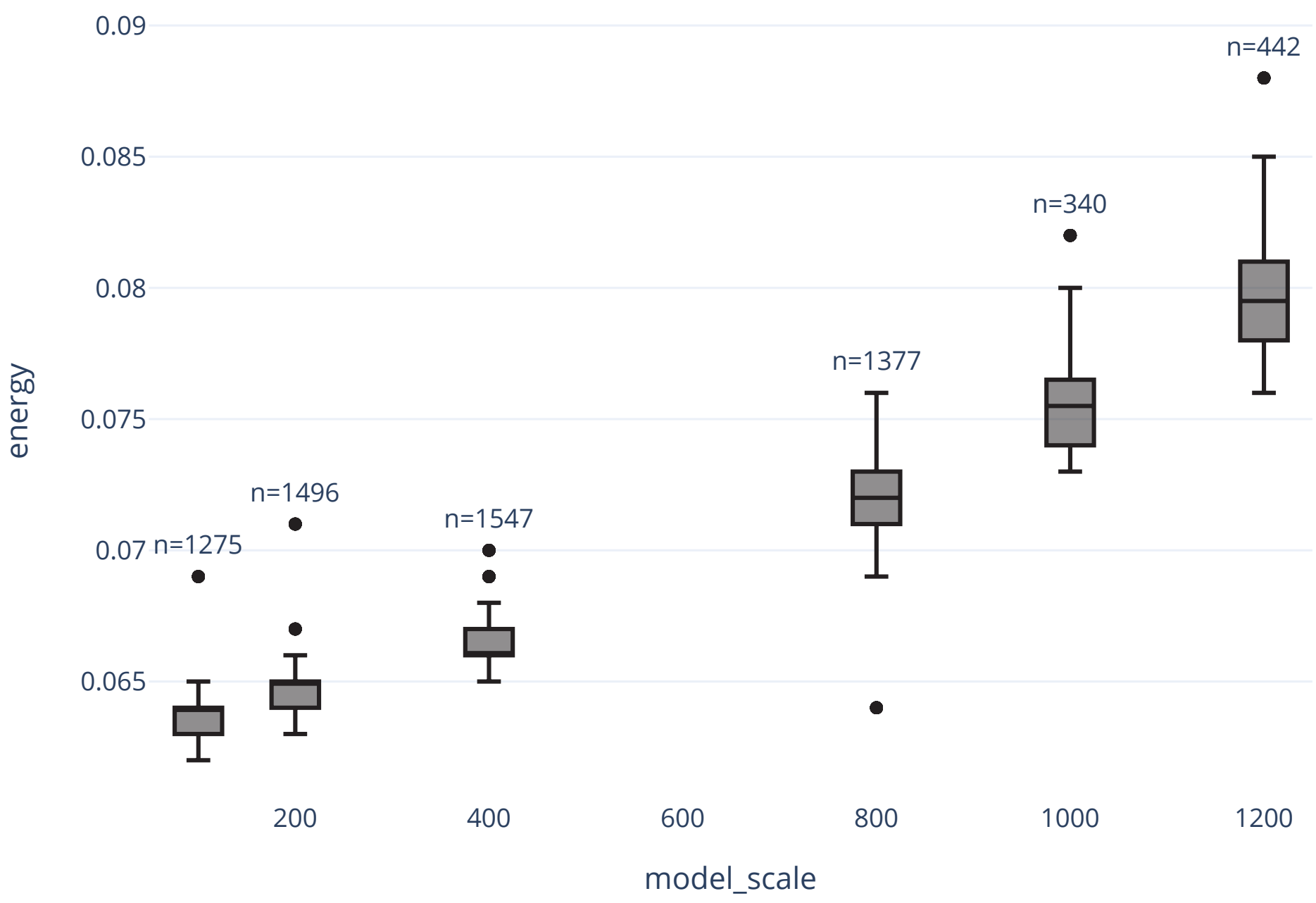

**Figure 9.7:** We sampled different scales for deep neural networks based on the underlying directed acyclic graphs (CTs). The number of model instances is denoted for each scale. A clear increase in energy consumption with larger model scales can be seen. Further, the variance in energy consumption appears to increase with larger scales. But that could be explained partially due to the fact, that with larger model scales the variance in number of model parameters also increases.

*eccentricity* The *undir_ecc_min*, *undir_ecc_max*, *undir_ecc_median*, *undir_ecc_mean*, and *undir_ecc_var* of the eccentricity distribution (compare diameter and paths of a graph).

*shortestpaths* The *shortestpaths_min*, *shortestpaths_max*, *shortestpaths_median*, *shortestpaths_mean*, and *shortestpaths_var* of the distribution of shortest paths in the graph.

*layersize* The *layersize_min*, *layersize_max*, *layersize_median*, *layersize_mean*, and *layersize_var* of the distribution of layer sizes given the topological sort of the DAG.

We conducted correlation analyses between the listed structural properties and the energy consumption of deep neural networks induced from computational themes as exemplified in fig. 9.8. The three properties test accuracy, overall number of model parameters, and energy consumption displayed very clear correlations as can be observed in the three rightmost correlation scatter plots of fig. 9.8. Further, variances of layer size and degree also showed some correlation with the energy con-sumption. For the layer size, we suppose that it is a good predictor for the maximum layer size and, indeed, a larger layer size also correlates with a larger parameter size. A large maximum layer size implies at least two matrix multiplications of large size which might be the reason for significantly larger energy consumption. The relationship of degree variance and energy consumption, however, is not as obvious. We suppose that a larger variance in degree is beneficial for neural network realisations of high accuracy as we found in previous experiments such as in section 9.5 on page 167. A larger degree variance, however, does not automatically imply a larger energy consumption as we can see in fig. 9.8 on the lower left. We suppose that we often can increase the degree variance of a CT as to increase accuracy but without affecting the energy consumption.

REMARKS

- Energy consumption poses an alternative objective to performance or robustness metrics of deep neural networks.
- Structural properties have an influence on energy consumption. The obvious relationship is a clear correlation with the number of operations and the energy consumption, but there are also signs of more complex structural properties: variance and maximum of layer sizes or degrees of vertices.

## 9.8 SUMMARY ON BASIC STRUCTURE ANALYSES

The basic experiments on layer-wise neural nets (section 9.2), random graphs (section 9.4 & section 9.5), and CTs (section 9.7) show clearly that structural properties make a difference. Pointing the influence out to a single structural property, however, is not possible. In fact, structure analyses need to be taken very carefully as layer bottlenecks or parametric scalings of the neural network realisations can create an impression of structural effects. There is a good chance that structural effects take only place on scales we can currently computationally not capture. The search space definition of $\mathbb{S}$ also considerably changes analyses depending on its cardinality. Layer-wise neural nets are structurally not very interesting while a search space of random graphs is hardly fully sampled. We therefore continue a more in-depth analysis with CTs of which we can capture the full search space at least up to a small graph order.

REMARKS

- We conducted basic structure analyses conducted on layer-wise, random graphs, and CTs.
- Correctness, robustness, or energy consumption serve as common objectives for analysis.
- At first glance, strong structural effects seem to exist for architectures with different structures, but deeper data analysis often unveils a hidden bias such that the structural effect is hard to capture or only slightly present.
- Besides properties directly related to the parametric capacity some structural properties appear to have an impact. Variances of the degree or path length distributions seem to be beneficial for correctness. Energy consumption is influenced by large layer sizes which is an indication for large and computationally costly matrix multiplications.

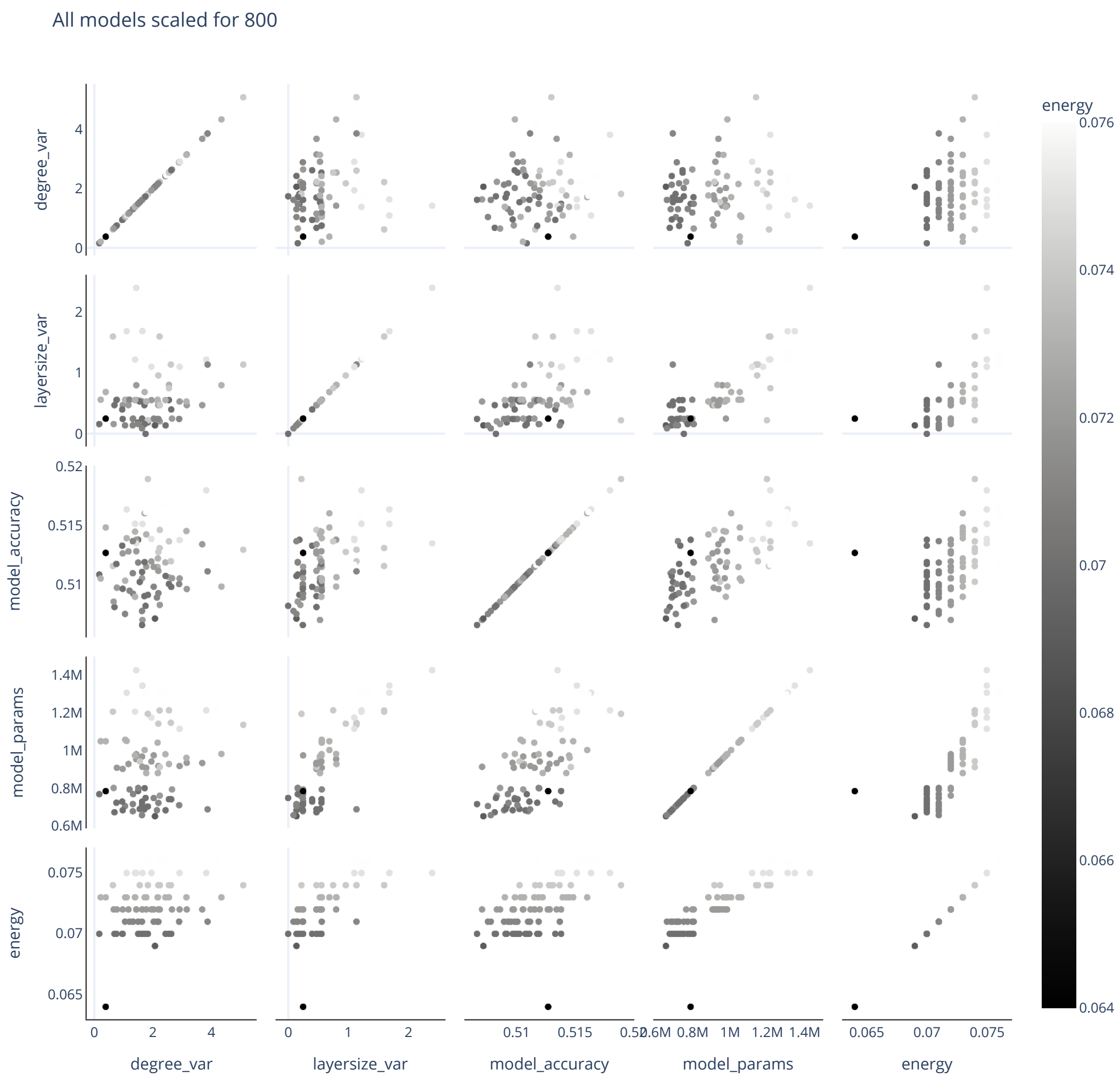

**Figure 9.8:** The scatter matrix shows the energy consumption of computational themes with a scale of 800 in relation to variances of the layer size and the degree of the CTs. Not only the number of parameters but also the accuracy show a strong correlation with energy consumption. We further found in analyses across all properties the maximum of layer sizes to be a strong influence on the energy consumption. One very large layer size is an indicator of at least two very large matrix multiplication operations in the model such that this might be a reason for the predictability.

# 10

# ANALYSIS OF COMPUTATIONAL THEMES

Working with various search space definitions for neural architecture search and formalizing graph-induced neural networks motivate the definition 3 of computational themes (CTs) in section 7.2 on search space design. From the motivation of CTs, we turn to practical considerations, construct a subset of computational themes and evaluate them on different datasets. Refer to section 8.3 for more details on the CT-NAS database. The CT-NAS database serves several experiments to further tackle our second research complex:

> ⋄⋄ Research Complex II
>
> Do neural network models differ in terms of structure or are they just different with respect to the choice of other hyperparameters such as activations, training epochs, data sampling or augmentation strategy, optimisation procedure, or the choice of loss function?

Indeed, we later observe a difference in classification performances for different CTs on the employed datasets. We elaborate on these findings which are based on 863 graphs used for the induction of neural network architectures.

These kind of experiments stand in contrast to approaches in which only few architectures are compared. For an analysis on only few architectures, one can find answers to which architecture overrules other architectures in a small setting. But as soon as the number of structural themes gets large, a pairwise comparison gets difficult. On the other hand, the opportunity of a large-scale comparison is to find factors or arguments that could *explain* why certain types of architectures overrule others. Such an analysis is achieved by e.g. re-organizing architectures into groups based on the performance or by using methods such as random forests to investigate on the importance of features derived from the architecture to explain the resulting performance.

## 10.1 EXPERIMENT: PERFORMANCE DISTRIBUTION ACROSS COMPUTATIONAL THEMES

The CT-NAS database, introduced in section 8.3, is in the following analysed on the guiding question: do the themes differ in terms of classification performance such as accuracy or $F_1$ score?

The first analytic step for the CT-NAS data is to turn the spotlight on the individual datasets. We aggregate all repeated computations per computational theme (CTs) and obtain a distribution of average performance values per theme for each of the eight datasets used as targets in the CT-NAS database. For each dataset, we then take the averaged mean performance to derive an ordering across all 831 employed graphs in the database. Each distribution of average performances then naturally increases along an x-axis and gives an impression of differences between worst and best performing architectures.

How could analytical results look like to guide further architecture development?

**Search space $\mathbb{S}$:** Computational themes (7.2.3)
**Method:** Analysis of the CT-NAS benchmark.

**Data:** `MNIST`, `CIFAR10` and six artificially generated `SpheresUDCR` classification datasets.

**Interpretation:** The search space itself exhibits three characteristic types depending on the application dataset and when looking at a large sampled number of the space.

The visualisation of box plots is reduced according to Tufte's data-ink maximised version [285, Chp. 6 Redesign of the Box Plot] to focus on essential trends, i.e. the median and other quantiles are presented but the wide boxes are left out.

### 10.1.1 *Performance of Computational Themes on CIFAR10 and MNIST*

For `CIFAR10`, we provide an impression of such an analysis plot in fig. 10.1. Except for the `MNIST` data, we scale all visualisations along a performance value between 0 and 0.7 as to visually compare the distributions across different datasets, if needed. Recall, that all datasets in CT-NAS are 10-class classification tasks such that random guessing a solution is a naïve lower bound with an accuracy of $\frac{1}{10}$. Outliers, that drop outside this lower bound, are due to computational issues or models that simply perform even worse than random sampling during evaluation, which could happen if they are strongly biased or if the hold-out evaluation data is not perfectly uniform w.r.t. the target class distribution.

If there would be no difference across themes or other hidden factors, we should see a random order of reduced tufte plots. The fact, that a clear increase in mean, median and even minimum and maximum averaged performance scores can be observed across all datasets, tells that there is a significant difference between the computed models. A stronger argument such as *structure makes a difference* can be even already made, considering the training scheme details as outlined in section 8.3 and the amount of data points underlying this observation.

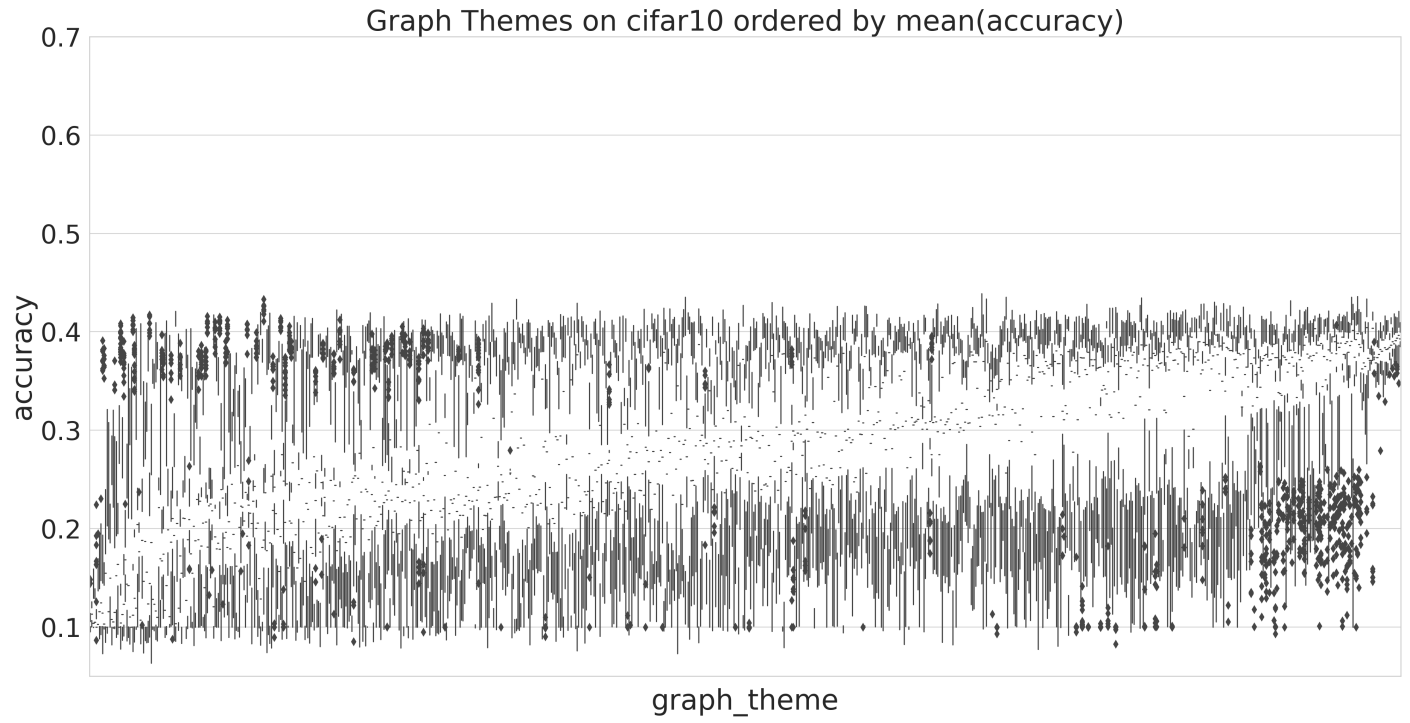

**Figure 10.1:** Each graph theme is trained on `CIFAR10` with different hyperparameters (details in section 8.3). An order of CTs is obtained by taking the overall mean of a target performance score (here *accuracy*) and then ranking each graph in asecending average score. For each CT, its distribution of performance scores is depicted with its median (center tick of each tufte-style box plot) and, quantiles in the whiskers and additional outliers outside these quantiles. Observe the low standard deviation of themes to the left (low performers) and also the low standard deviation with only few outliers on the right (top performers). Another interesting group follows the top performers on the right with high performance scores but some few but very bad outliers. There seems to exist architectures that always perform quite decent on expectation.

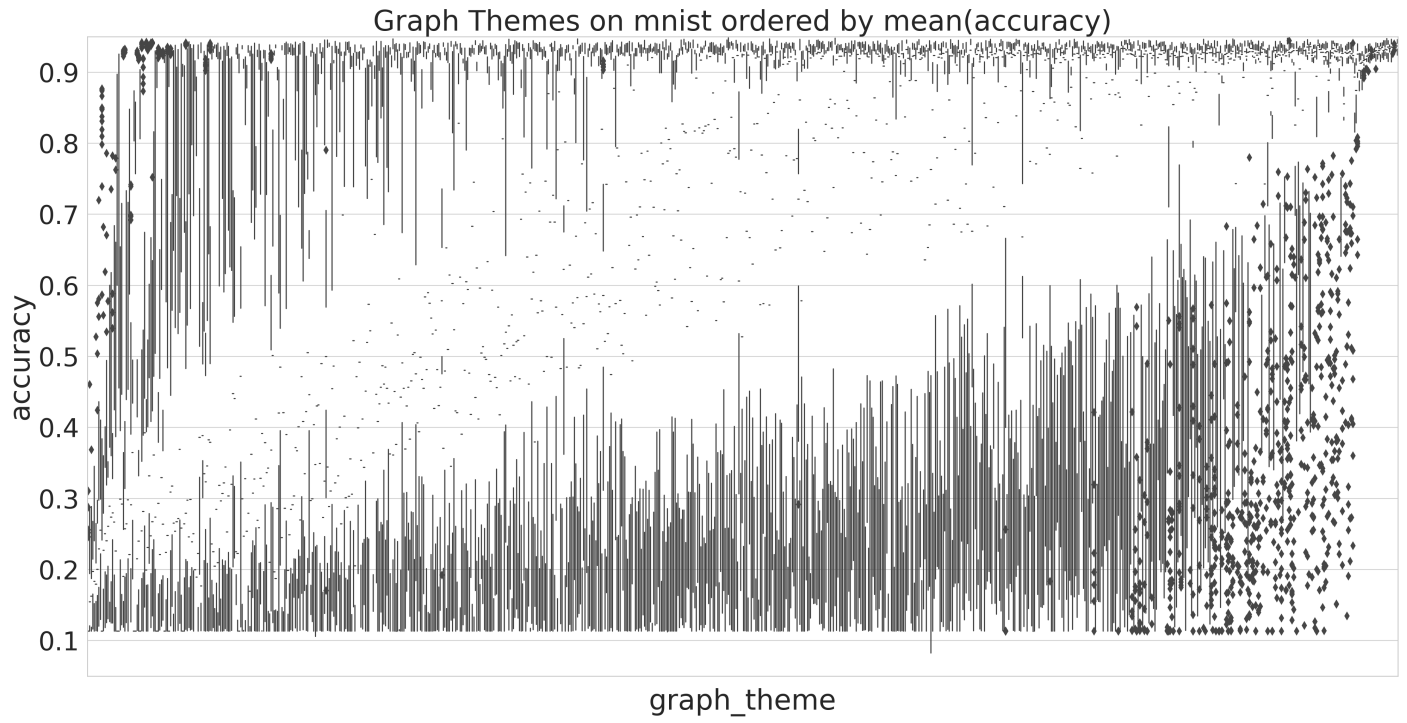

**Figure 10.2:** The `MNIST` is the only dataset that is shown in the range between 0.1 and 1.0 as almost all graph induced architectures achieve performance scores way above 90% in classification accuracy. Still, the phenomenon of reliable themes with low standard deviation and high average accuracy scores can be found on the right of the plot. Here, the graph themes are ordered according to the averaged accuracy within the `MNIST` data.

However, there are quite a lot of interesting observations to be made in the details. Both on the lower and upper 15% of the ranking, depicted on the 15% left and right in Figure 10.1, respectively, outliers occasionally reach performance levels of the best (or worst) performing computational themes. After all, each of the architectures should be theoretically capabable of universal approximation. But this also invites to look into the stability or empirical guarantee of a CT. How certain are individual or aggregated performance scores and how much do they deviate overall? We analyse the averaged standard deviations of the performance scores subsequently. The median shall for now suffice to show that there is a significant different across themes.

Another observational detail for CTs on `CIFAR10` is a surprisingly strong estimation for few themes in the top five percent, seen on the right end of fig. 10.1. These graphs not only induce architectures with high performing models under the given training scheme, but also repeatedly show unusual stability when compared with competing computational themes. This is a highly dersirable property: these themes are not only on average very good performers, but can be also easily found or reproduced. We call them reliable (computational) themes in context of `CIFAR10`.

The downside of these reliable themes seems to be, that they also exhibit fewer or worse top outliers. An interpretation could be, that top performing models are then hardly found among these themes, even when investing a lot time in exploring their potential space extensively. Reliability seems to not come automatically with high performance.

The same phenomenon can be observed in the `MNIST` data, provided in Figure 10.2. A small group of CTs on the right are reliable in the sense of being both high in expected accuracy and low in standard deviation of it. We will have a more closer look on this phenomenon of reliable themes further down.

Another group, following in the top 15% range on the right of Figure 10.2 has high performance on average, but shows some few very bad outliers. This could be an instability in the training scheme where these particular architectures sometimes fail in convergence when not treated carefully. But the group could also contain architectures for which convergence is generally more difficult, i.e. its parametric search space is more complex.

It is characteristic of themes on `MNIST` that the overall task is easily solved with an accuracy of over 95% using any such a CT induced architecture. We made this observation over various other experiments in the past, already, and therefore consider `MNIST` only as a complementary tool to understand structural differences in NAS.

Nevertheless, for the low performance regime, a small group of themes can be observed on the left that both do not reach solid top ac-curacies and even show low standard deviations, which makes themes of this group evidently fail in the task. Failing themes do not contain sufficient top performers as outliers and have a very low average performance, perhaps even close to random guessing.

### 10.1.2 *Performance of Computational Themes on Artificial Datasets*

Next, we take a look on three selected artificial datasets of `SpheresUDCR`. The underlying observations repeat for the remaining `SpheresUDCR` tasks, except that the distributions shift along the complexity of the particular dataset. We leave the detailled visualisations out for the sake of clarity. In contrast to `MNIST`, we observe more stable orders in Figure 10.3, Figure 10.4, and Figure 10.5, which we deem beneficial to explore on structural differences of the themes.

While in Figure 10.5 we can observe a similar phenomenon of reliable themes as in `MNIST` and `CIFAR10`, the standard deviation of the best performing 15% of themes in Figure 10.3 (`spheres-0a19afe4` (⊛)) and Figure 10.4 (`spheres-23aeba4d` (✠)) is comparatively high. A reason could be, that the underlying task assumably consists more of memorisation of points in space for the `SpheresUDCR`-based datasets than breaking down patterns as in the image classification tasks `MNIST` and `CIFAR10`.

However, on the lower end of graph themes for `SpheresUDCR`-based datasets, we usually can observe many failing themes in the sense that they exhibit both low expected performance, bad top-outliers, and even sometimes low standard deviations. This can be seen on all three provided figures on the left ≈5% of the x-axis. At least within the context of a fixed dataset, there seem to be structural patterns that do not work out well.

### 10.1.3 *Variance Analysis of Computational Themes*

Interestingly, we observe **three types** of very varying behaviours on the `SpheresUDCR`-based datasets when it comes to standard deviation of performance scores, ordered asending by the mean performance: some exhibit a quite similar or constant standard deviation (such as `spheres-0a19afe4` (⊛) and `spheres-23aeba4d` (✠)), while others increase (such as `spheres-6598864b` (⨁) and `spheres-b758e9f4` (▷)) and others even decrease (such as `spheres-b8c16fd7` (♠) and very slightly `spheres-bee36cd9` (♥)).

Our interpretation of the three types of change of standard deviation along the distribution of performance scores is as following:

1. **Type "—"** A constant (randomly fluctuating) standard deviation across the themes could be seen as a uniform complexity of finding neural network realisations within the structurally re-stricted universal architecture. For low performing themes it is comparably difficult to search their parametric space than for high performing themes. Structure then can make a difference in terms of performance on the underlying task, but there seems to be no benefits gained in obtaining a suitable trained model when switching the CT.

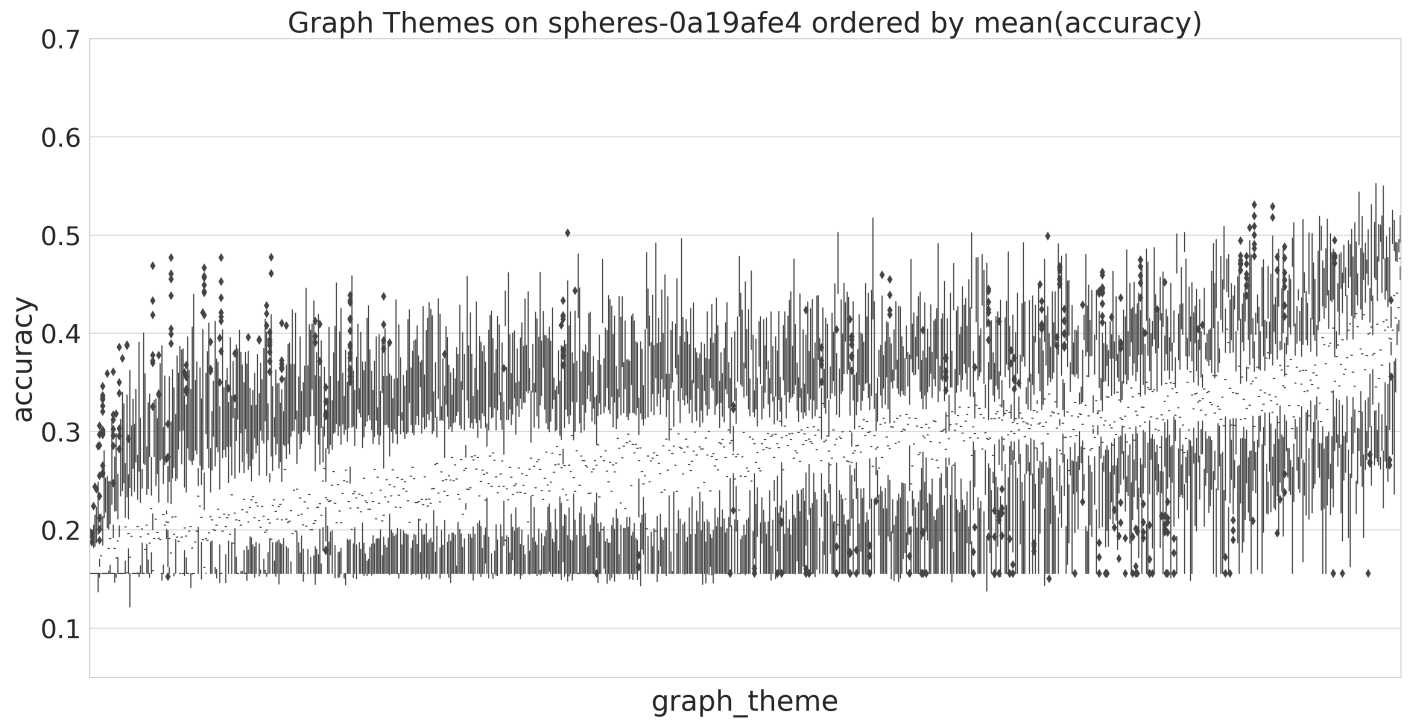

**Figure 10.3:** With artificially generated taks such as the accuracy estimated on this depicted `spheres-0a19afe4` (⊛) set, we gain additional insights into the observed phenomenons of reliable or failing themes. Suprisingly, the right end of the performance ranking does not show as reliable themes in `spheres-0a19afe4` (⊛) as compared to `MNIST` and `CIFAR10`. We suppose, that the `SpheresUDCR`-based tasks are difficult in the sense that require a certain memorisation capacity while image-based tasks could be broken down into pattern recognition, as often concluded for deep learning on images [209].

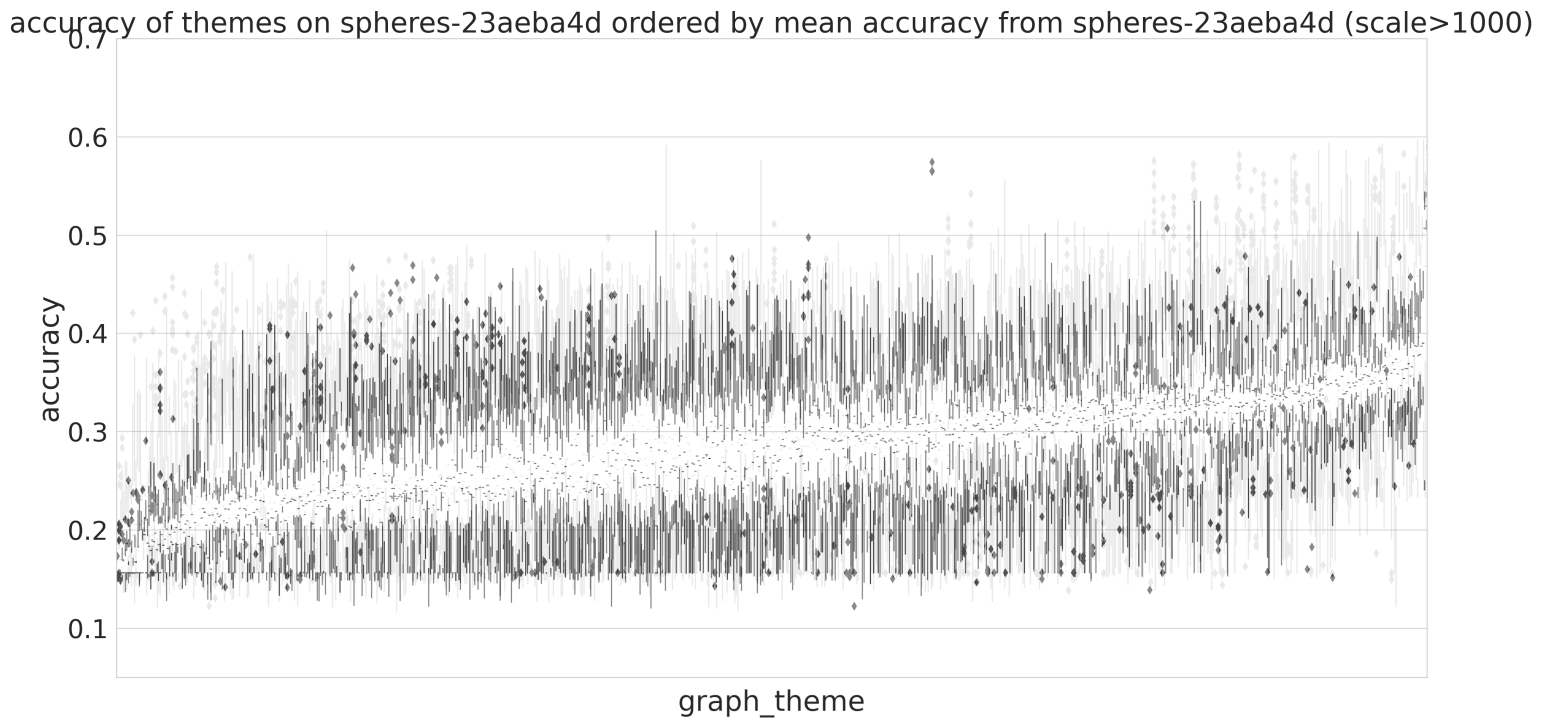

**Figure 10.4:** The generated task `spheres-23aeba4d` (✠) is an example for which we observe very similar behaviour as for `spheres-0a19afe4` (⊛) in Figure 10.3. This is an example for an equivalent tufte plot, but restricted to CTs for which only neural network realisations of a scale of at least 1,000 are taken into account. We left the original distribution and order without the scale restriction in the background. The estimations get more stable and also start to reveal less fluctuations on the lower left and upper right ends.

2. **Type "_/-"** Increasing standard deviation along improved expected performance indicates that higher performing architectures also are more difficult to train. This is an undesirable property but would be not too surprising for difficult application domains: very bad models fail early and reliable, but very good models could be both structurally different and significantly hard to train such that they sometimes also exhibit bad performance.

3. **Type "-_"** A decreasing standard deviation, on the other hand, is beneficial: better trained models in those cases also tend to always perform similarly stable. In such a case, the structure does not only influence a performance difference but also guarantees reliable results. Such application domains should be usually easily searchable for neural architecture search as high performing architectures immediately give positive and reliable feedback to the search strategy.

If our observation also holds beyond the presented instances, we refer to the three types as types of *changes in the estimation complexity* of the search space. We observe that the three types depend on the dataset, and they also depend on the search space with its contained structural themes.

Search strategies for NAS could leverage this information either by pre-defining search spaces with a favorable type of complexity change or by trying to detect the type of estimation complexity change. As an example, if one knows that the search space guarantees more stable estimates for higher performing structural themes (**type "-_"**), the search space could transition into computational less expensive training schemes or repeated estimations and focus more on search space exploration to find even better themes. On the other hand, for **type "_/-"** a search strategy can accelerate during an early search phase as the search space reveals to get more complex. During the early search phase the space can be quickly explored for better themes and eventually the strategy can invest more computational budget into training schemes of later search phases. We suppose that the second type (**type "_/-"**) is more common for complex application domains.

To further explore our observation of these three types of changes in the estimation complexity of the search space, we take a closer look into the distribution of standard deviations of the $F_1$ score. With a visualisation such as fig. 10.5 on the next page, we would get a point approximation for the standard deviation per theme and not a distribution because it is a second moment statistic. Therefore, we additionally group the samples for each CT additionally by learning rate and scale. That means, for each theme we have a distinct learning rate and scale with multiple repeated computations. The standard deviation is obtained per aggregated group and visualised on a y-axis and ordered by theme along the x-axis with asencding averaged $F_1$ score. We chose these two hyperparameters because they have a profound influence on the final performance estimation. Because this aggregated grouping reduces the overall standard deviation, the actual standard deviation obtained per CT is additionally provided in scattered points with crosses above the reduced box plots.

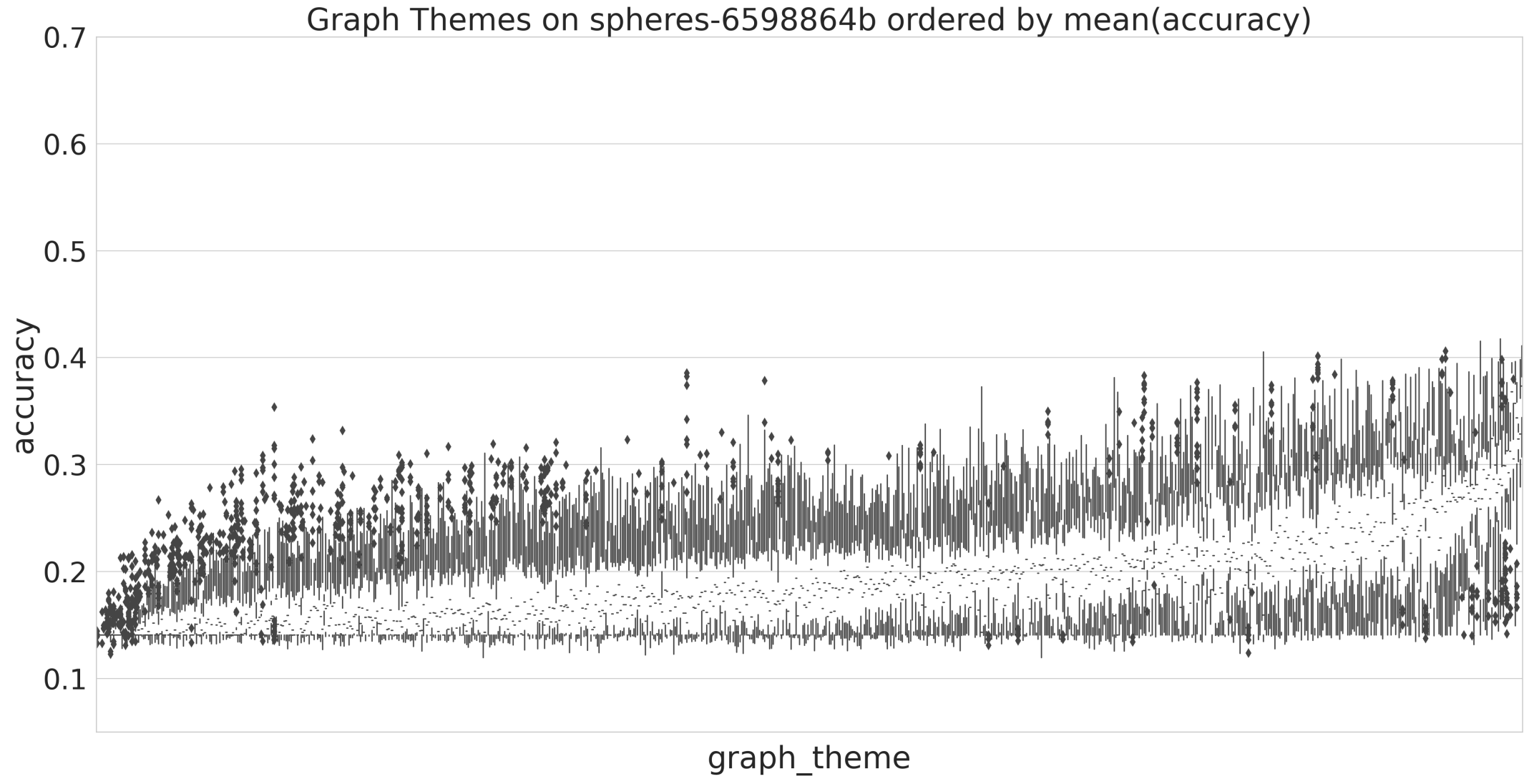

**Figure 10.5:** For `spheres-6598864b` (⊕), repeatedely failing themes can be clearly observed on the left. These themes not only seem to not learn the underlying task on average, but also show an example for an architecture for which no working realisation can be found, even with repeated training and varying hyperparameters. A visualisation for the standard deviations of the $F_1$ score of themes on the same task can be found in Figure 10.8. Note, that the standard deviation is interestingly increasing. Further, the instances of `spheres-6598864b` (⊕) are an example for a very late but then steep increase in performance for good (and reliable) themes.

Figure 10.6 provides an example for a descending **type "-_"** of deviations along improving themes. This can not only seen by smaller and less distributed standard deviations on the right but also roughly with overall high standard deviation scores on the left. The first 20-30% themes on the left easily exceed a standard deviation of 0.2 while on the right the upper 5% stay above 0.15 (with one single outlier).

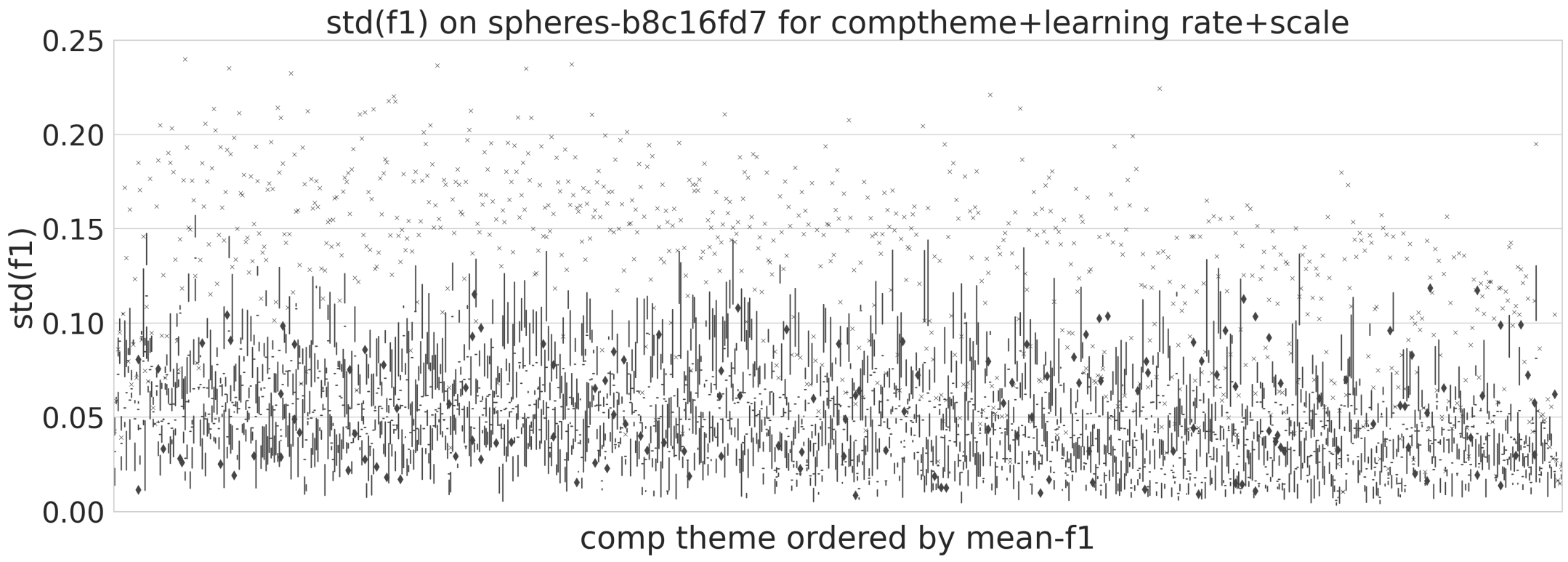

**Figure 10.6:** The standard deviation of $F_1$ score of themes on the `spheres-b8c16fd7` (♠) task is shown. With an apparent decrease of std($F_1$ score) along improving themes w.r.t. the averaged mean $F_1$-score, we associate the search space of CTs on `spheres-b8c16fd7` (♠) with the descending **type "-_"** of change in estimation complexity. Note, that the above scattered points provide the actual standard deviation while the tufte plots are obtained by grouping all scores by learning rate and scale such that we obtain an impression of the stability of the standard deviation among these categories.

For `CIFAR10`, also the descending **type "-_"** applies according to Figure 10.7. The ≈ 10% top-performers are close to or below of 0.1 in standard deviation of their $F_1$ score (right side of the plot). The remaining themes with worse performance exhibit many themes way above a standard deviation of 0.1, some even being above 0.15. However, the difference is not as significant as for Figure 10.6 and also note, that there exist themes in between this ranking which also exhibit lower standard deviations of their $F_1$ score. We interprete the three types as a phenomenon of the overall search space of structural themes rather than a property of the individual graph-induced architecture.

An example for the ascending **type "_/-"** of standard deviation of performance scores along improving themes is provided in Figure 10.8. The low standard deviation on the left shows how the worst performing themes actually repeatedely fail, even when trained from different seed starts and varying hyperparameters. While the first ≈ 20% of themes show no standard deviation of above 0.1, better themes on the right side show slightly larger standard deviations in $F_1$ score with values going even close to 0.15. Similar to the results on `CIFAR10` in Figure 10.7, this has to be taken with a grain of salt as the ordering is still with respect to the average mean $F_1$ score and the data shows also instances in between that have lower standard deviation. The data on the far right could also be interpreted in a way that the standard deviation quite suddenly drops for the top $\approx 2-5\%$ of themes, which exhibit slightly more reliability than the remaining main body of themes.

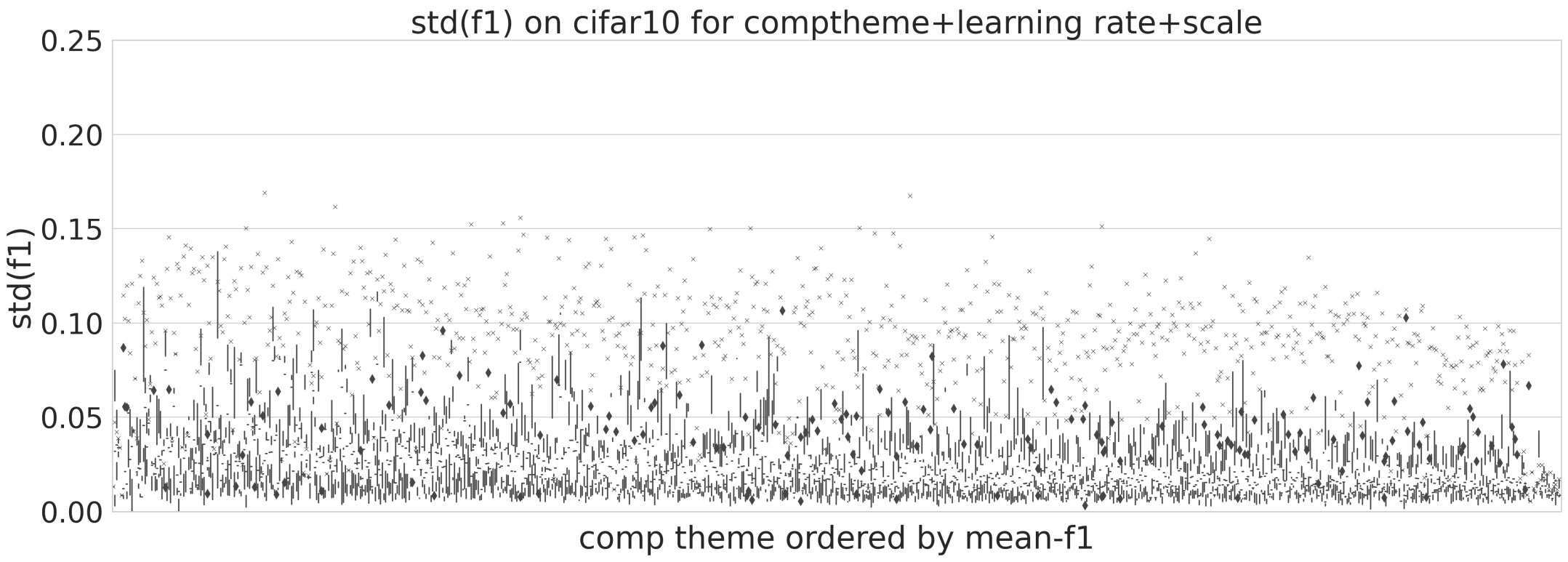

**Figure 10.7:** This visualisation provides the distribution of standard deviations of $F_1$ scores on `CIFAR10` in the same manner as in Figure 10.6. We again observe a descending **type "-_"** of change in estimation complexity. The data goes along with Figure 10.1 which showed the distribution of accuracy instead of the $F_1$-score. Observe the sudden drop at the far right of the plot, which we interprete as a sign of reliability of the top performing graphs.

### 10.1.4 *Summary of Observations*

This section was guided by the initial question whether CTs differ in terms of classification performance in the CTNAS database. For this, we looked into the distributions of performance scores and made five major observations:

1. Indeed, the employed themes empirically make a significant difference when it comes to classification performance. This leads to followup questions concerned with the reasons for this apparent difference.
2. We observed a phenomenon of reliable themes, which we defined as themes with high expected performance and low variance (or standard deviation) of the performance estimation – both relative w.r.t. the overall distribution of performance within a selected task. Reliable themes are a good sign as not only good realisations can be found within their parametric space but also it can be reliably found and few trials might suffice for a good estimation.
3. Contrary to reliable themes, we also found themes that certainly failed on the given task and we defined them based on low expected performance (even close to random guessing) and bad top outliers on the performance as failing themes. The likelihood of finding a good realisation in failing themes appears to be low despite that it should theoretically exist.
4. With further analysing the standard deviation along an axis of themes increasing in their performance score, we observed three

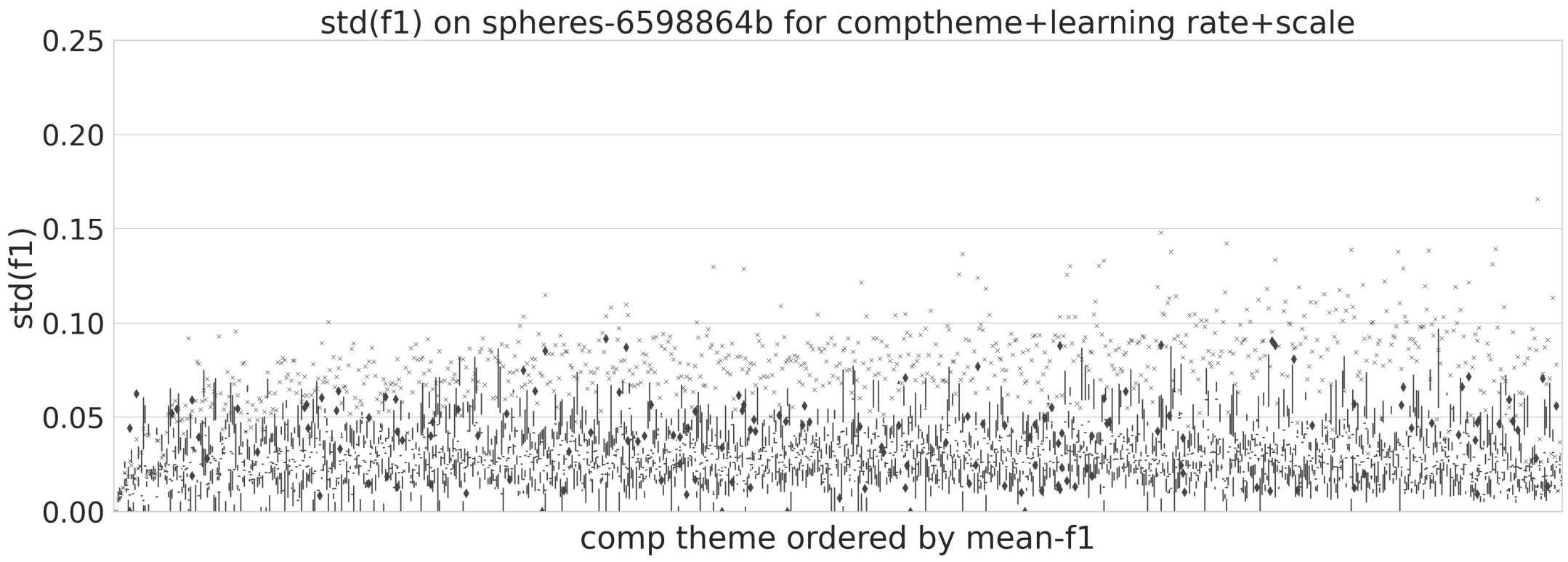

**Figure 10.8:** While having overall mostly bad performances, we took the results of `spheres-6598864b` (⊕) as an example for a space of ascending **type "_/-"** of change in estimation complexity. Themes on the left and even up to the middle have mostly standard deviations below 0.1, while the actually better performing themes show increasing values of above 0.1 to about 0.15. The task on `spheres-6598864b` (⊕) can, however, not be considered as solved or successfully trained, as the top performing themes only show an $F_1$ score of up to 0.4 on a classification task with ten classes.

types of change in the estimation complexity. The three types of change can be a uniform or constant **type "—"** of uncertainty across all themes, an increasing **type "_/-"**, and a decreasing **type "-_"** of standard deviation. We interprete them as a phenomenon of an interplay of search space and task and suppose that it could be used to guide improved search strategies during NAS.

5. The performance distributions exhibit different s-shaped curves which sometimes shows that a high performing regime can be reached either quite early or quite late, depending on the task.

REMARKS
- Computational themes as a proxy for the structure of neural networks show a significant difference.
- We can observe interesting groups of themes such as *reliable themes* and *failing themes* which can be distinctively differentiated from other themes.
- Additionally, three different types of convergence complexitiy change on the combination of search space definition + dataset could be observed. This phenomenon of convergence complexity change opens interesting new research perspectives.

## 10.2 EXPERIMENT: SIMILARITY OF GROUPS OF COMPUTATIONAL THEMES

We observed interesting phenomenons on a per-task basis in CTNAS. Structure does make a significant difference within each task. Can we now transfer certain properties from one domain to the other as to induce priors through structure? This requires to have shared structural properties within groups of computational themes (CTs).

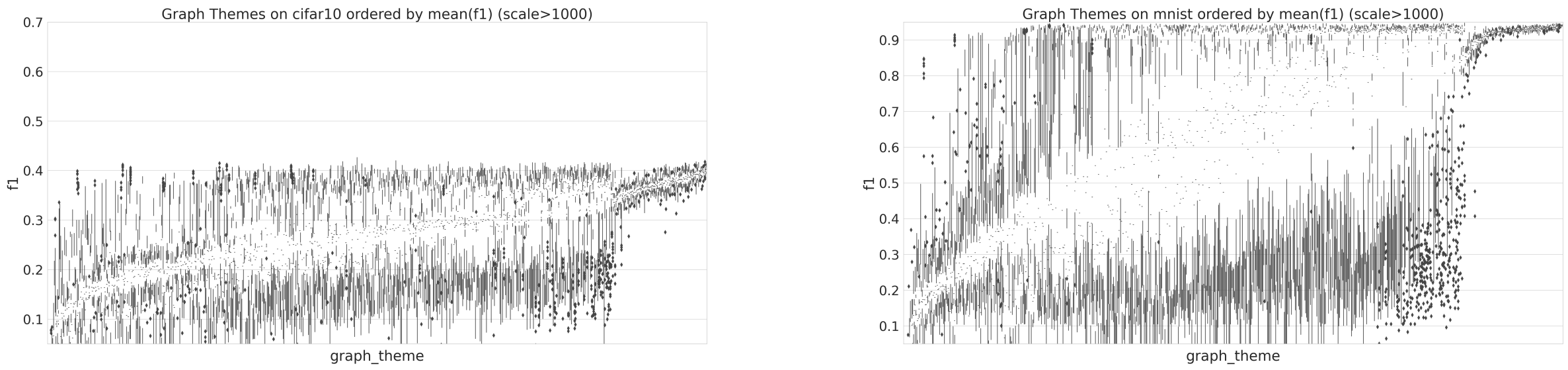

**Figure 10.9:** Both insights into MNIST and CIFAR10 are repeated here with only themes of scale above 1,000 and with respect to the $F_1$ score instead of accuracy as performance target. Visual comparison with Figure 10.2 and Figure 10.1 directly shows us how the reliability of the estimation increases. However, in both cases, this happens mostly on the lower and upper end of the performance ranking. Note, that the order of CTs here now is according to the $F_1$ score and only considering computations with more scale such that the order might be different to the previously seen plots.

> Are network properties involved in the cause and if so which?
>
> **Search space $\mathbb{S}$:** Computational themes (7.2.3)
> **Method:** Analysis of the CT-NAS benchmark on clustered CTs.
>
> **Data:** MNIST, CIFAR10 and six artificially generated SpheresUDCR classification datasets.
>
> **Interpretation:** There are signs of groups among CTs which might be related to failing or succeeding themes. Grouping search space candidates based on similarities can serve as an alternative for explicit network properties.

### 10.2.1 *Comparisons on a per-identity basis*

With taking each the best and worst 10% of themes w.r.t. the individual performance ranking on a particular task, we obtain two groups per dataset. We used a filter on the overall set of themes as outlined in Listing 10.1.

**Listing 10.1:** Example for a post-hoc filtering on the CTNAS database to obtain a ranking on themes per dataset

```python
def upper_mean(x):
    lower_quantile = np.quantile(x, q=0.25)
    res = np.mean(x[x > lower_quantile])
    return res if not np.isnan(res) else np.mean(res)

df_filtered = df_comps[(df_comps["dataset"] == ds) & (df_comps["
    scale"] > 500) & (df_comps["f1"] > (1/df_comps["classes"
    ]+0.1))]
df_sorted_agg_themes[ds] = df_filtered.groupby("graph_theme")[["
    f1"]].agg(upper_mean).dropna().sort_values("f1")
```

In other words, we took the sample average of only $F_1$ scores larger than the first quartile, i.e. the upper mean, and only considered themes with a scale strictly larger than 500 (to reduce the variance, as shown in Figure 10.9). This results in an overall ranking per dataset – similarly as used for the ordering of the visualisations in e.g. Figure 10.5 for the distribution of accuracy scores or in Figure 10.8 for the distribution of standard deviation of $F_1$ scores. For each dataset, a group of each best and worst 81 ($\approx 10\%$) themes is then selected from this ranking.

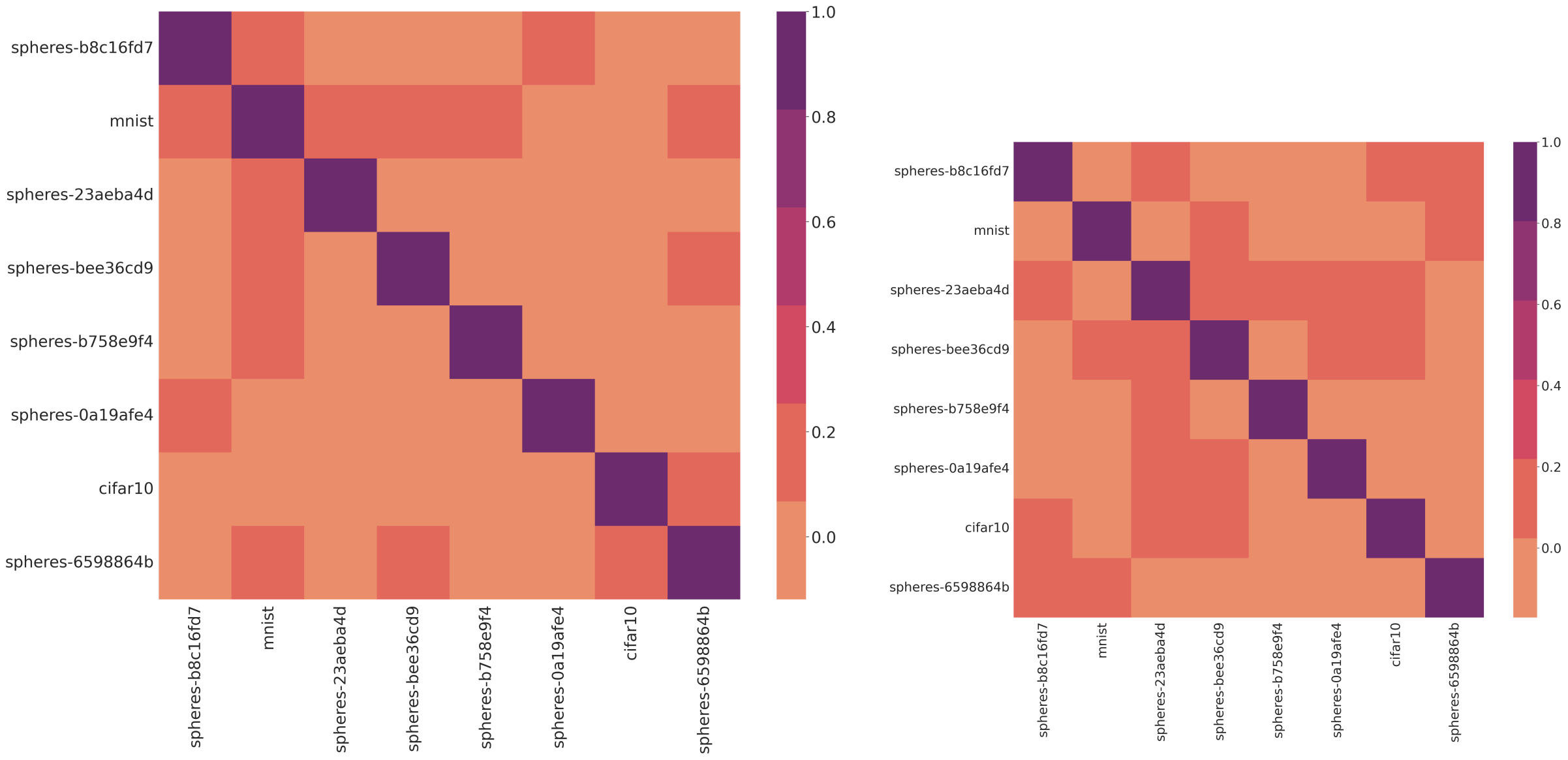

**Figure 10.10:** Correlation of each best (left) and worst (right) 10% themes per dataset. The 10% best/worst graphs are taken on a per-dataset basis and a spearman correlation matrix is computed amongst all pairs. We see a perfect alignment on the diagonal as a sanity check. There are slightly more weak correlations to be found among worst performing themes. An interpretation might be, that there are structures that usually don't work out while for best performers it depends more on individual datasets. Note, that the correlation is based on an individual correlation of computational themes, not on underlying properties that could explain commonalities.

With each dataset having a set of best and worst themes, we looked into the correlation of these sets across datasets. The result of this spearman correlation matrix can be observed in Figure 10.10. For both groups, we can see pairwise correlations between datasets. The diagonal gives a full rank correlation of 1.0 which serves as sanity check.

For the correlation matrix of best themes per dataset, most pairs show no or only very slight correlation. This means, that the underlying groups do not have any overlaps. Note, that this comparison is based on ranks and identities and does not tell us anything about common structural patterns. However, we see clearly, that on a per-theme basis we see no significant commonalities across data sets. Nevertheless we will take a look into the small correlations later.

The correlation matrix on the right side shows the equivalent plot but for the worst themes per dataset. A similar phenomenon of almost no correlations can be observed except for some small correlations. The overall amount of small correlations is significantly higher than for the group of best themes per dataset. That suggests an interpretation that failing themes might have a little bit more in common across datasets than successful or reliable themes. And the commonality might be not even a hidden or complex structural property but the simple identity of failing themes.

The order clearly differs among different datasets and to illustrate this property, we take another look on the performance distribution of themes of a single dataset but now order it according to the ranking from another dataset. With Figure 10.10 showing us, that the best and worst themes across datasets do not correlate, we would expect a pretty wild ordering of tufte plots.

First, we have a look on `CIFAR10` again, but this time ordered according to the ranking obtained through `MNIST`. The result is shown in Figure 10.11 and as expected, the visual now shows a pretty wild ordering along the x-axis. On a closer look, the leftmost themes exhibit a slight non-random standard deviation when compared with the rest of the visualisation. Also, we see a break in standard deviation between the 15% and 5% rightmost themes. Assumably, both details could be random fluctuations. For the second observation, the rightmost themes, however, we could at least ponder why a whole bunch of reliable themes on `CIFAR10` group into one place in the upper part of the ranking of `MNIST`.

According to our correlation analysis in Figure 10.10, `MNIST` and `CIFAR10` exhibited no correlation among the best and worst ten percent of themes w.r.t. their own ranking. However, `CIFAR10` and `spheres-6598864b` (⨁) show a small correlation, such that we selected it for a second presentation in Figure 10.12.

Again and most notably, the overall visual appears very random in comparison to the baseline presentation for `CIFAR10` in Figure 10.1. On a closer look, however, it can be seen that the accuracy distribution according to the median in the box plots appears to be lower on the left of Figure 10.1 than on its right. A rolling median across themes on the right shows to be around 0.3 while on the left it is significantly below 0.3. Also, the standard deviation shows to slightly decrease.

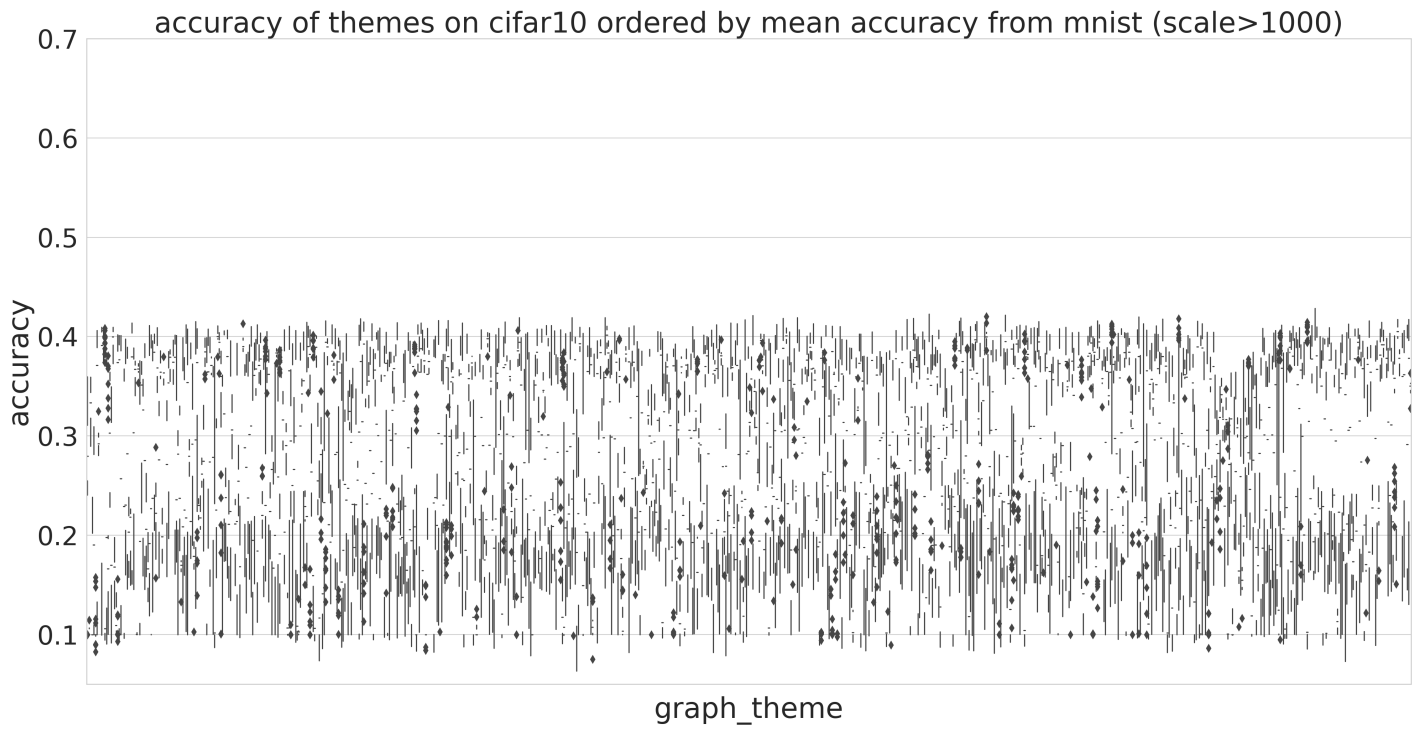

**Figure 10.11:** The visual shows the accuracy distribution on CIFAR10, ordered by the ranking from the upper mean accuracy distribution of themes on MNIST (compare Listing 10.1 for the ordering). As expected from the correlation plot of Figure 10.10, we observe a quite random pattern across the x-axis, although some minor observations and interpretations on some grouping of themes with common standard deviation or higher means in accuracy can be made.

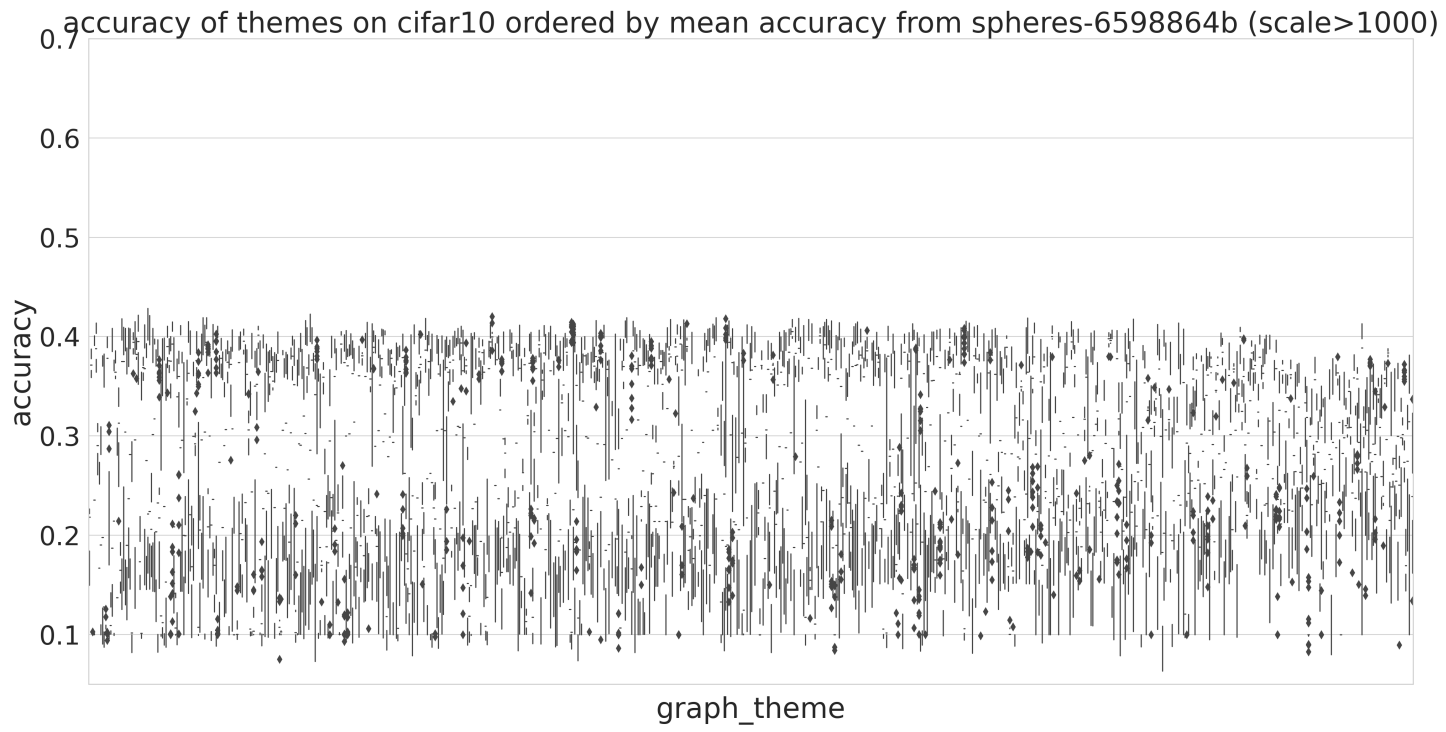

**Figure 10.12:** In analogy to Figure 10.11, this visual shows the distribution of accuracy scores on CIFAR10 along an x-axis of themes, ordered along the ranking obtained by the upper mean accuracy from spheres-6598864b (⨁). Note, that CIFAR10 and spheres-6598864b (⨁) show a weak correlation for the group of best themes per dataset and we interprete the rightmost part of this visual as a confirmation of that observation.

We keep these observations in mind: identities of CTs do not correlate strongly across datasets. Except for `MNIST`, no pair of dataset shows weak correlations in both best and worst themes, according to Figure 10.10. Weak correlations, however, are interesting and can indicate simple relationships for further analysis.

### 10.2.2 *Comparisons on a per-similarity basis*

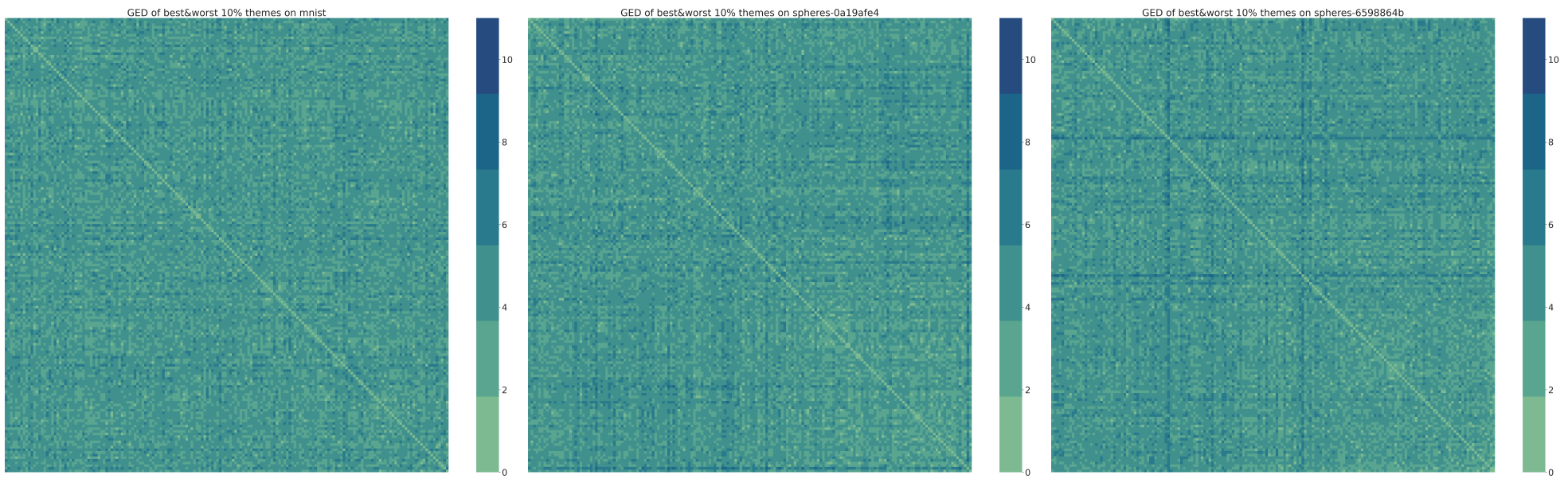

**Figure 10.13:** Three exemplary distance matrices for the datasets `MNIST` on the left, `spheres-0a19afe4` (⊛) in the middle, and `spheres-6598864b` (⨁) on the right. The first half of each matrix contains the best, the other half the worst themes along the axes. Each plot shows the GED value ranging from 0 up to 9

With this first insights, we took the graph edit distance to further look into whether a well-defined distance metric gives us hints about commonalities within or between the groups. With each group of 81 candidate themes we computed a pairwise GED. Compare Section 3.3 for details on graph similarities. Exemplary resulting distance matrices are shown in Figure 10.13.

The heat map of a distance matrix including the best 81 themes and the worst 81 themes shows a pairwise comparison based on graph edit distance. With the diagonal being therefore zero, it can be observed as a light line, showing, that each theme has zero edit distance to itself. Brighter therefore refers to closer, darker refers to more dissimilar themes. If the structural cohesion is very strong within a group of themes, we observe a bright square. The rightmost distance matrix for `spheres-6598864b` (⨁) shows the most significant cohesion, even though it is relatively weak in absolute terms.

Such a cohesion pattern can only marginally spotted visually for each task and a deeper statistical analysis in Table 10.1 confirms our observation: considering the best 81 themes and worst 81 themes individually as groups, we observe only small deviations of the expected GED within a group. The overall expected GED of all themes is 4.0556 with a median of four and a standard deviation of 1.1418. This means that most themes are between 2.5 and 5.5 edit apart while the maximum edit distance is 11. For each task, Table 10.1 provides the average, standard deviation and the difference to the overall average GED.

Interestingly, the applied tasks `MNIST` (denoted with M) and `CIFAR10` (denoted with C) are the only tasks in which there is a positive difference (column Δ) to the overall GED mean. As the difference is computed as "overall mean - groupwise mean", a positive difference means that the group-wise cohesion is high. Under the perspective of the graph edit distance as a meausre of dissimilarity between themes, the groups of best and worst themes on the applied tasks are therefore more close to eachother.

| | **Grp** | **⌀± std(GED)** | **Δ** | **max** | | **Grp** | **⌀± std(GED)** | **Δ** | **max** |
|---|---|---|---|---|---|---|---|---|---|
| all | | 4.0556 ± 1.1418 | | 11 | | | | | |
| ▷ | ↗ | 4.075 ± 1.1697 | -0.0194 | 10 | ♠ | ↗ | 4.0271 ± 1.1421 | 0.0284 | 11 |
| ▷ | ↙ | 4.0824 ± 1.1284 | -0.0268 | 11 | ♠ | ↙ | 4.0666 ± 1.1568 | -0.0111 | 11 |
| ⊕ | ↗ | 4.0835 ± 1.1565 | -0.028 | 10 | C | ↗ | 4.004 ± 1.1397 | 0.0516 | 11 |
| ⊕ | ↙ | 4.0711 ± 1.1412 | -0.0155 | 11 | C | ↙ | 4.0505 ± 1.1263 | 0.0051 | 11 |
| ⊛ | ↗ | 4.0907 ± 1.1593 | -0.0352 | 10 | ♥ | ↗ | 3.9814 ± 1.1206 | 0.0741 | 10 |
| ⊛ | ↙ | 4.1225 ± 1.1484 | -0.067 | 11 | ♥ | ↙ | 4.135 ± 1.1594 | -0.0795 | 11 |
| ✠ | ↗ | 4.0608 ± 1.1418 | -0.0053 | 11 | M | ↗ | 4.0532 ± 1.1341 | 0.0023 | 11 |
| ✠ | ↙ | 4.0628 ± 1.1258 | -0.0073 | 11 | M | ↙ | 4.0231 ± 1.1282 | 0.0325 | 11 |

**Table 10.1:** This table summarises core statistics for the graph edit distance among pairs of themes within groups by dataset and performance. The column **Grp** denotes whether the underlying set contains the best ↗ or the worst ↙ themes within a dataset (according to their $F_1$ score). The ⌀± std(GED) denotes the sample average of graph edit distances and its standard deviation. In the first row, we can observe that the average graph edit distance among all themes is ≈ 4.0556± 1.1418 with a median of four. The Δ columns provides the difference between overall averaged GED and the group-wise averaged mean. If it is negative, the inner-group distance is closer on average than expected across all themes. Surprisingly, only the real-world data sets `MNIST` (M) and `CIFAR10` (C) show both non-negative difference, which tells that w.r.t. to this similarity measure, there exists a tiny structural grouping.

### 10.2.3 *Summary of Observations*

We only found small correlations of rankings of themes between tasks, as shown in Figure 10.10. Re-ordering the performance of themes of one task with rankings from another task as exemplified in Figure 10.11 underlined this result. Although minor observations of commonalities can be made, i.e. Figure 10.11 does not show completely random behaviour, it can be clearly seen, that single themes can not be used to transfer prior information to other tasks. Extending the comparison to use a similary measure such as the graph edit distance shows first signs of structural commonalities: groups of themes exhibit lower distances than others. Dividing the groups by 10% best or 10% worst $F_1$ score-performers at least shows interesting intra-group cohesion for the real-world tasks `MNIST` and `CIFAR10`. This motivates to search for more general structural properties or groups and discards analytical methods which are based on a per-case comparison.

REMARKS

- Computational themes as a proxy for the structure of neural networks show a significant difference on selected datasets.
- Single themes can not be used to transfer prior knowledge to other domains. Either a different search space definition has to be made or common structural properties need to be further investigated on.
- Based on the graph edit distance, themes can be clustered into groups but the commonality based on this notion of similarity is not as high as one would expect initially.

## 10.3 EXPERIMENT: CRITICAL DIFFERENCE OF COMPUTATIONAL THEMES

Without aggregation, comparing 863 representative computational themes (CTs) is difficult. We manually reduce the representative graphs to thirteen examples of which we visualise six in fig. 10.14. Independent of CT-NAS, we train 11194 DNN models based on these thirteen CTs and use an automated pipeline of LIB AUTORANK [98] to conduct a Friedman-Nemenyi test [48] across these thirteen non-normal populations. The normality is checked with a Shapiro-Wilk test and rejected for all conducted tests such that significant differences are found based on median values.

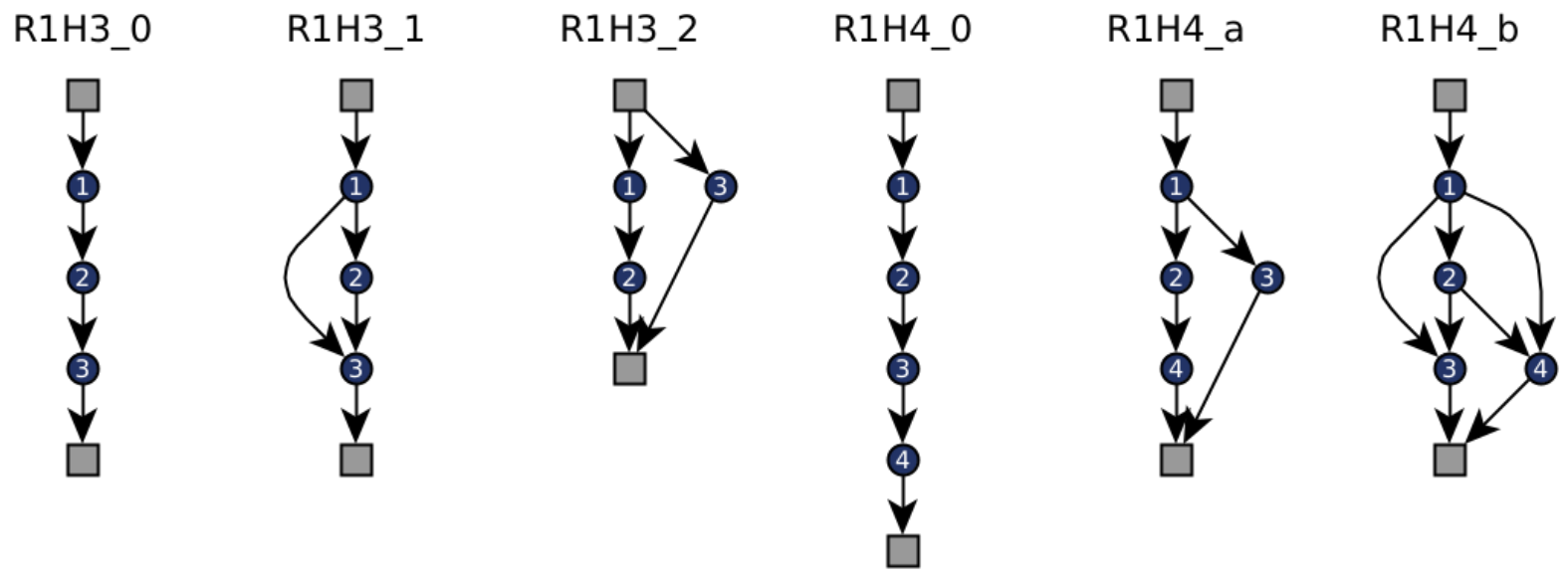

**Figure 10.14:** Six exemplary CTs used to investigate on differences in rankings with LIB AUTORANK [98]. We explicitly selected small representations which still exhibit different specific graph properties such as being just "deep" or having skip-layer connections.

Instead of generated unique identifiers to keep graphs apart like for the CTNAS database, we encode some graph properties into their naming scheme and obtain the following names: R1H3_2, R1H3_1, R1H4_b, R1H4_3_1, R1H4_2_3, R1H4_2_2, R1H4_a, R1H3_0, R1H4_1_3, R1H4_0, R1H4_2_1, R1H4_1_1, and R1H4_1_2. Besides the prefix *R1*, the naming simply denotes that there are *i* vertices with non-zero in- or out-degree such that these vertices are directly transformed into **hidden neurons** of the resulting DNN model. The second value after an underscore denotes the number of **source vertices**, i.e. two for R1H3_2. The optional third value after an underscore denotes the number of **skip-connections**, i.e. one for R1H4_1_1. More complex connectivity variations that we investigated on are additionally labelled with *_a* or *_b*. This naming scheme supports the comparison when no visual depiction is available.

In accordance with the CT-NAS database, we again take the following eight datasets and obtain accuracy, precision, recall and therefore $F_1$ score metrics. This results in the following complexity ranking: `spheres-b758e9f4` (▷) < `CIFAR10` < `spheres-6598864b` (⨁) < `spheres-0a19afe4` (⊛) < `spheres-23aeba4d` (✠) < `spheres-bee36cd9` (♥) < `spheres-b8c16fd7` (♠) < `MNIST` based on the ordering of the mean $F_1$ score across all repetitions. The 11194 computations distribute across the datasets with 1443 (`spheres-b758e9f4` (▷)), 1358 (`CIFAR10`), 1414 (`spheres-6598864b` (⨁)), 1365 (`spheres-0a19afe4` (⊛)), 1503 (`spheres-23aeba4d` (✠)), 1351 (`spheres-bee36cd9` (♥)), 1377 (`spheres-b8c16fd7` (♠)), and 1383 (`MNIST`) samples. Note, that this ranking aligns with the over 200,000 independently trained instances of CT-NAS data.

Herbold notes as an example that "the dataframe" used with LIB AUTORANK "could contain the accuracy of classifiers trained on different data sets" [98] and we apply it to this exact case. We use the mean accuracy of a CT-graph per dataset which resulted in a minimum of 77 per CT, an average of more than 106±21.2 and a maximum of 147 samples.

The result of the automated hypothesis tests are exemplarily shown for `MNIST` in table 10.2. All further statistics are to be found in the appendix in tables A.1 to A.3.

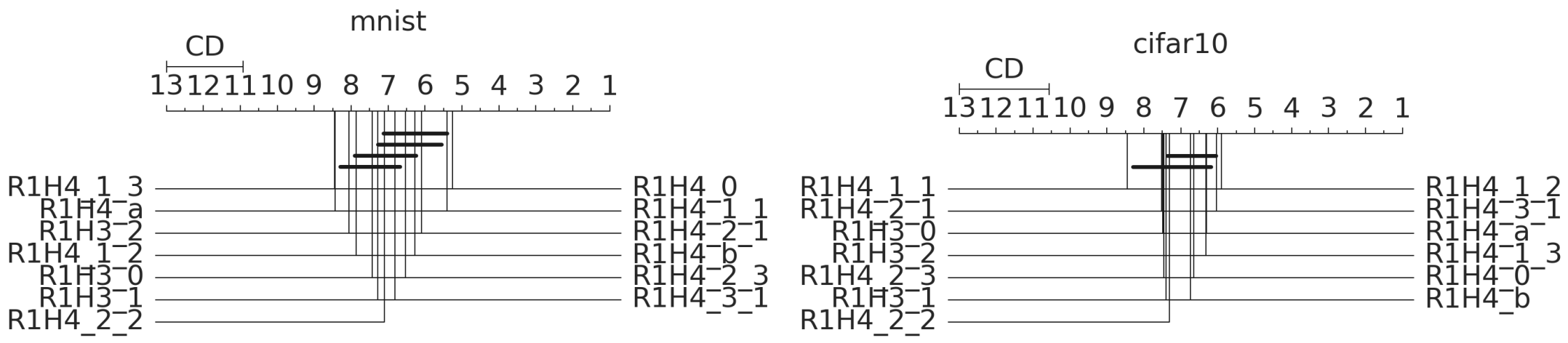

**Figure 10.15:** Critical difference plot for thirteen computational themes evaluated on `MNIST` (left) and `CIFAR10` (right)

Similar to previous experiments, we observe a significant difference between deep neural networks induced from computational themes.

| | MR | MED | MAD | CI | $\gamma$ | Magnitude |
|---|---|---|---|---|---|---|
| R1H4_0 | 8.455 | 0.962 | 0.015 | [0.844, 0.974] | 0.000 | negligible |
| R1H4_1_1 | 8.429 | 0.965 | 0.007 | [0.961, 0.972] | -0.183 | negligible |
| R1H4_2_1 | 8.052 | 0.968 | 0.008 | [0.959, 0.974] | -0.327 | small |
| R1H4_b | 7.864 | 0.968 | 0.008 | [0.955, 0.975] | -0.322 | small |
| R1H4_2_3 | 7.435 | 0.965 | 0.009 | [0.961, 0.975] | -0.181 | negligible |
| R1H4_3_1 | 7.286 | 0.969 | 0.006 | [0.962, 0.975] | -0.426 | small |
| R1H4_2_2 | 7.104 | 0.969 | 0.006 | [0.963, 0.975] | -0.411 | small |
| R1H3_1 | 6.812 | 0.970 | 0.006 | [0.964, 0.976] | -0.499 | small |
| R1H3_0 | 6.532 | 0.971 | 0.005 | [0.966, 0.976] | -0.517 | medium |
| R1H4_1_2 | 6.279 | 0.972 | 0.005 | [0.966, 0.975] | -0.591 | medium |
| R1H3_2 | 6.091 | 0.971 | 0.006 | [0.965, 0.976] | -0.512 | medium |
| R1H4_a | 5.409 | 0.974 | 0.005 | [0.965, 0.977] | -0.723 | medium |
| R1H4_1_3 | 5.253 | 0.975 | 0.002 | [0.969, 0.976] | -0.795 | medium |

**Table 10.2:** Automated summary of the AUTORANK hypothesis tests for the thirdteen classifiers on `MNIST` based on computational themes. MR stands for mean rank, MED/MD for median, MAD for median absolute deviation, and CI for confidence interval. We provide the according critical difference (CD) plot for `MNIST` in fig. 10.15 on the previous page on the left and the CD-plot for `CIFAR10` on the right.

The rankings across datasets, however, differ heavily such that we can not find a common property or structure that is universal across all domains.

REMARKS

- Independent experiments with manually selected computational themes confirm the overall difficult ranking across the datasets of the CT-NAS database.
- Automated hypothesis tests on the ranking per dataset shows that there exist critical differences between the studied computational themes.
- We could not find a pattern based on basic structural properties that is consistent for the critical differences across different datasets (also compare CD-plots in figs. A.1 to A.3).
- The rankings across different datasets differ such that the observation of no common ranking in the CT-NAS data is confirmed.

## 10.4 SUMMARY AND IMPLICATIONS OF STRUCTURE ANALYSES

We presented analyses of structures of graph-induced deep neural networks. These experiments were concerned about research complex II on analysing neural networks. Central research questions have been whether neural networks differ in terms of structure, whether and which network theoretic properties contribute to objectives such as performance, robustness or energy consumption, and how analytical results could look like to guide further developments.

Basic insights were summarised in section 9.8 on page 178 such as challenges with the definition of the search space $\mathbb{S}$ and its size or challenges with conflicting influences of structural properties. We tackled both challenges again for a more complete sampled space of CTs and by not only considering single structural properties but also a CT as surrogate for structure and similarities between CTs for groupings:

- Turning to CTs for deeper analysis of the search space, we empirically found that three types of changing standard deviations along the distribution of performance scores can be observed. The analysis suggests that every search space could be classified into one of these three types of a constant "—", increasing "_/-", or decreasing "-_" standard deviation in performance estimation. This suggests that there exist search spaces for which the search becomes more uncertain while for other search spaces the search becomes more reliable towards good solutions.

- Both the analyses of chapter 9 and of chapter 10 confirm that the influence of structural properties is very domain dependent. Many phenomenons are strongly tight to the underlying task to be solved. Manual extraction of structural properties and assessing its influence is tedious and the influence at most marginal. Grouping and cluster analysis as conducted in section 10.2 suggests that deep learning with automatic feature learning is a potential path forward for conducting neural architecture searches.

Part V

# METHODS FOR NEURAL ARCHITECTURE SEARCH

Various methods can be employed to automatically find structures of neural networks or define them a priori. How do these search methods relate? Which method should be chosen?

**Research Complex III: Automation**

From section 1.1.6:

- Which methods for neural architecture search exist, how can they be compared and what are their differences?
- With knowledge on structure from research complex II, how can neural architecture search methods be improved or guided?

# 11

# PRUNING

*The following entails:*

Pruning is an umbrella term for approaches and methods to reduce structural elements of neural networks. The research on pruning aims to compress model size, increase inference speed, find high-performing subnetworks, reduce energy cost, or find trade-offs between these goals. The term can summarised as a principle (for optimisation) which yields a new solution candidate by reducing or *pruning* a previous candidate.

PRUNING AS OPTIMISATION STRATEGY Importanty in light of the theoretical considerations of graph-induced neural networks, pruning can serve on different optimisation levels and its interpretation has to be well distinguished:

1. Pruning within one architecture is the classical interpretation in the sense that it is used as an optimisation step (towards e.g. less parameters) with subsequent optional re-training. These classical approaches usually do not change the underlying architecture.

2. Pruning such that the architectural theme changes is a second-level optimisation method and historically not as common as the two types have not been distinguished strictly. Two examples of changes of a theme $T$ are **1/** when pruning is applied in the search space $\mathbb{S}$ and **2/** if pruning is applied on a neural network realisation $f \in \mathcal{A}([T])^{d_1 \to d_2}$ and this results in a change from theme $T_1$ in search space $\mathbb{S}$ if a back-estimation yields a (better applicable) theme $T_2$ for $f$.

Classically, pruning can be interpreted as a variant of optimisation techniques to find a suitable realisation of a neural network when performing a training-pruning-pipeline. For this recall eq. (7.1):

$$\mathcal{F} \triangleq \underset{f \in \mathcal{A}([T])^{d_1 \to d_2}}{\arg\min} \mathcal{L}_{train}(f, D_{train})$$

in which pruning can be used to find $\mathcal{F}$ in a sequential process $t \in [1, \ldots, t^{end}], t^{end} \in \mathbb{N}^+$ of the right-hand optimisation side and estimate a closer solution $f^{(t+1)} \in \mathcal{A}([T])^{d_1 \to d_2}$ from $f^{(t)} \in \mathcal{A}([T])^{d_1 \to d_2}$ by removing structural elements from within $f^{(t)}$ and derive $f^{(t+1)}$. In this case, both realisations are elements of the same architecture $\mathcal{A}([T])^{d_1 \to d_2}$ but $f^{(t+1)}$ represents a potentially better estimate.

As a higher-level optimisation technique, pruning can take place in the second level of eq. (7.2):

$$\arg\min_{T \in \mathbb{S}} \mathcal{L}_{val}(T, \arg\min_{f \in \mathcal{A}([T])^{d_1 \to d_2}} \mathcal{L}_{train}(f, D_{train}))$$

One approach could be that pruning ecplicitly removes elements from $T^{(t)}$ to obtain $T^{(t+1)}$ in a sequential process $t \in [1, \dots, t^{end}], t^{end} \in \mathbb{N}^+$ such that it directly acts as a search strategy. Alternatively, a trained neural network $f^{1,T^{(t)}} \in \mathcal{A}([T^{(t)}])^{d_1 \to d_2}$ could be pruned into $f^2$ and a re-estimation provides $T^{(t+1)} = \arg\max_{T \in \mathbb{S}} Pr(T \mid f^2)$.

CONTRIBUTIONS IN PRUNING Our *first* contribution [270] are technical investigations on a pruning method, that is also motivated by analyses of the contribution of structural elements of neural networks. Based on work of Florin Leon [154], we transferred the game-theoretic Shapley Value to a performance-based criteria and compared it to popular methods at that time, namely random pruning, variants of magnitude-based pruning [89, 258], and Optimal Brain Damage [150]. We carved out, that a deep neural network does not trivially provide a super-additive payoff. The unmet requirement motivated further formal research in semi-monotone payoffs of the Shapley Value in which games are played with structural elements of neural networks. Although a required super-additive payoff would formally be required to derive some desirable properties, the Shapley Value nowadays finds application not only in pruning, but also in explainable methods such as SHAP [174].

A *second* contribution are further experiments on recurrent networks [269] and robustness of sparse networks [6] which fostered the implementation of a round-trip transformation tool between graphs and sparse neural networks, called DeepStruct/ LIB DeepStruct [273]. The subsequent systematisation of pruning methods & approaches and an extensive literature collection[1] is a further contribution of that line of work.

## 11.1 TAXONOMY OF PRUNING

Pruning approaches are iterative pipelines in which a model runs through multiple steps that include a pruning method in a loop until a stopping criterion is reached. Examples of such steps include the construction of a deep neural network model based on a prior architecture,

1 A public version of collected papers up until 2019 can be found at https://github.com/JulianStier/nn-pruning/ and is now outdated with newly emerged sub-fields in pruning which fostered literature surveys focussing only on specific aspects of pruning, e.g. pruning at initialisation as proposed by https://github.com/mingsun-tse/awesome-pruning-at-initialization.

the training of the model based on data, pruning of structural elements based on a specific criterion applied on a specified model scope with some pruning amount, possibly a re-training or fine-tuning based on data, repetition of multiple of these steps, possibly model inferences and even repetitions of parameterised outer-loops. Flow diagrams often accompany such pipelines to illustrate the experimental setting of a particular method or analysis.

PRUNING GRANULARITY To describe a pruning pipeline, discrete structural units need to be defined. These units then define the pruning granularity and can be also called *order of sparsity* as it defines a structural level on which sparsity is induced into the model or architecture. The granularity can be as fine as to consider weights or connections for removal, structured blocks of such weights, whole neurons or as coarse-grained as to consider structured blocks of neurons, cells, groups of cells, layers or even whole submodules. Hybrid granularity choices are also possible but less common and more complex to analyse.

In context of pruning, we use a bijective map $\xi$ to obtain the discrete units on which pruning can be executed. The discretisation provides a candidate set of $n \in \mathbb{N}^+$ structural elements $C = \{c_1, \dots, c_n\}$ of which one or some are chosen for pruning.

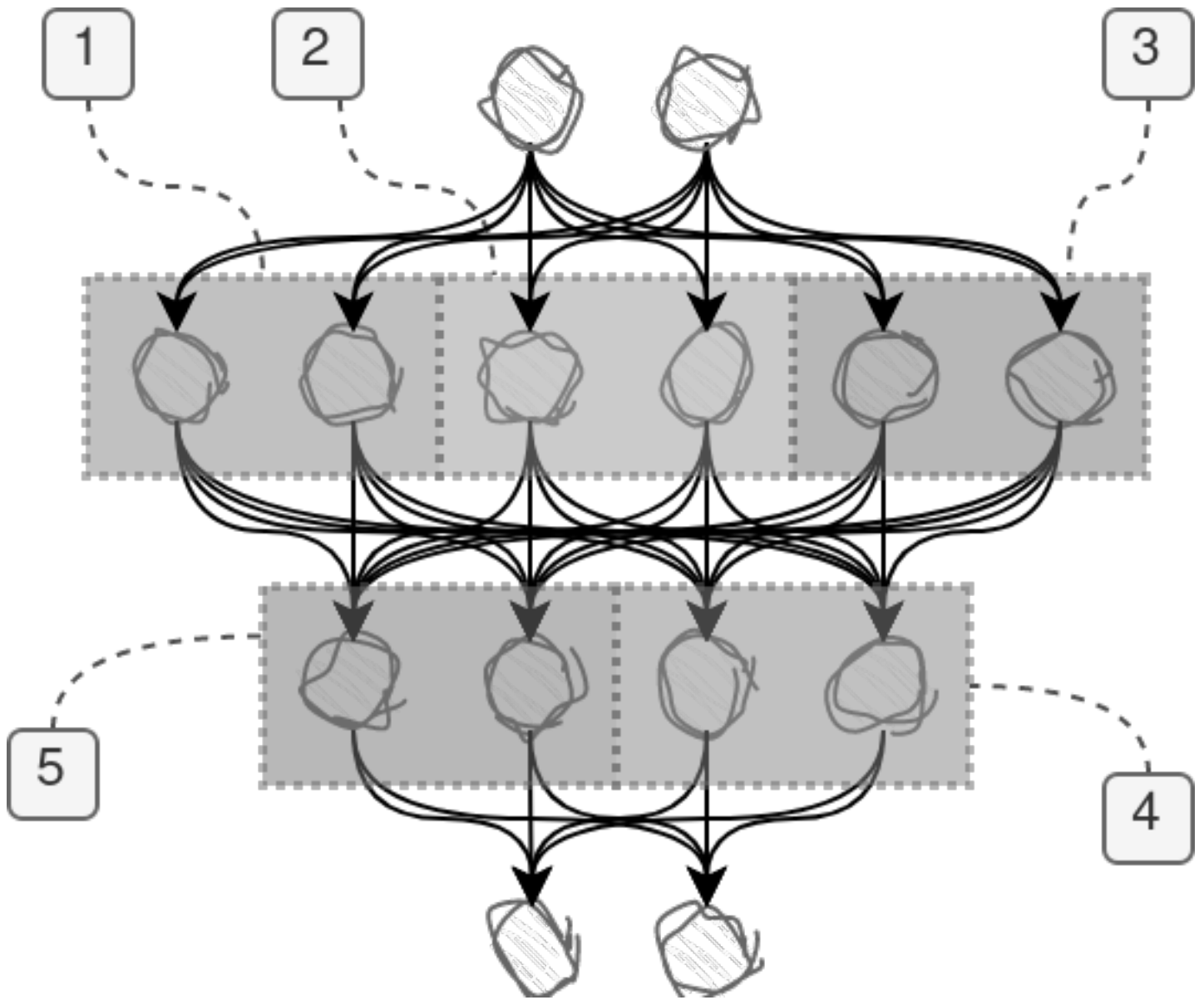

**Figure 11.1:** The pruning granularity determines structural units that belong together when considered for removal. Here, every two neurons are grouped into one structural unit such that the exemplary two layers are organised in five units. The pruning criteria and its scope control the selection process for units to be removed amongst all canidates. Such a selection process can for example either choose a single unit amongst all five units or two units independently within each layer.

PRUNING CRITERIA CATEGORIES The pruning criteria is a decisioning rule that specifies which of the structural units are to be pruned. Pruning criteria can be classified by four broad categories: pruning by *chance* (randomness), pruning based on *weight* statistics, based on *loss* statistics, and based on *performance* statistics such as the change in accuracy, out-of-distribution performance, robustness or energy consumption. The pruning criteria can then be seen as a function of the model and information of the particular category to return a reduced model. A *criteria scope* can further specify whether to prune uniformly within, across or agnostic to sub-structures, i.e. whether to prune uniformly amongst weights of each layer or uniformly amongst all weights of all layers.

The pruning criteria includes an attribution method $\chi : C \to \mathbb{R}$. For the category of weight-statistic based pruning criteria, an example with an $l_1$-norm attribution could be defined as $\chi^{(weight)}(c) \mapsto \sum_{\theta \in \xi^{-1}(c)} |\theta|$. Here, all parameters of the structural element are aggregated into their sum of absolute values.

The decisioning rule $\delta : (\xi, \chi) \to \mathcal{P}(C)$ then could simply take the one candidate with minimum $l_1$-norm, but the exactly chosen candidates are not just dependent on the attribution method, but also on how the amount of structural elements for pruning is determined.

PRUNING AMOUNT Like in recently re-investigated growing methods [63, 184, 280, 314, 315], an interesting question to compare pruning pipelines on is: How many structural elements should be pruned? The *amount* of structural elements to be pruned can be obtained in a *fixed*, a *percentage*-based or a *bucket*-value-filling manner. Parameterizing this aspect of calculating a pruning amount for each step in the pipeline can result in non-trivial hyperparameter studies [270].

A small fixed pruning amount of e.g. one ($k = 1$) element, alternating with re-training cycles, is very expensive but is very close to a non-differentiable regularisation technique. With DNNs having not seldomly billions of parameters, often large $k >>> 1,000$ are chosen. The higher $k$, the faster the pruning process reaches a moment of severe underparameterisation in which evaluation measures drop significantly in comparison to almost no changes in early stages of the process. We observed in multiple studies [6, 269, 270], that even the choice of the pruning criteria with its attribution method makes no significant difference in early stages of the pruning process. Different fixed-value amounts $k_1, \ldots$ can also be chosen for different components of a model, i.e. for different layers.

Relative or percentage-based pruning amounts use a proportion $p \in (0, 1)$ of elements to be pruned per step. As the resulting amount of structural elements to be pruned increases over time with fewer parameters, one can also choose to parameterise it, i.e. with $p = \frac{1}{c \cdot log(t+1)}$ and $c \in \mathbb{N}^+$ and $t \in \mathbb{N}$ being the pruning steps until a stopping criteria is reached. The pipeline then starts off by pruning e.g. $\approx 36\%$ of elements early for $c = 4$ and less than $\approx 10\%$ later during the process (for $t > 13$).

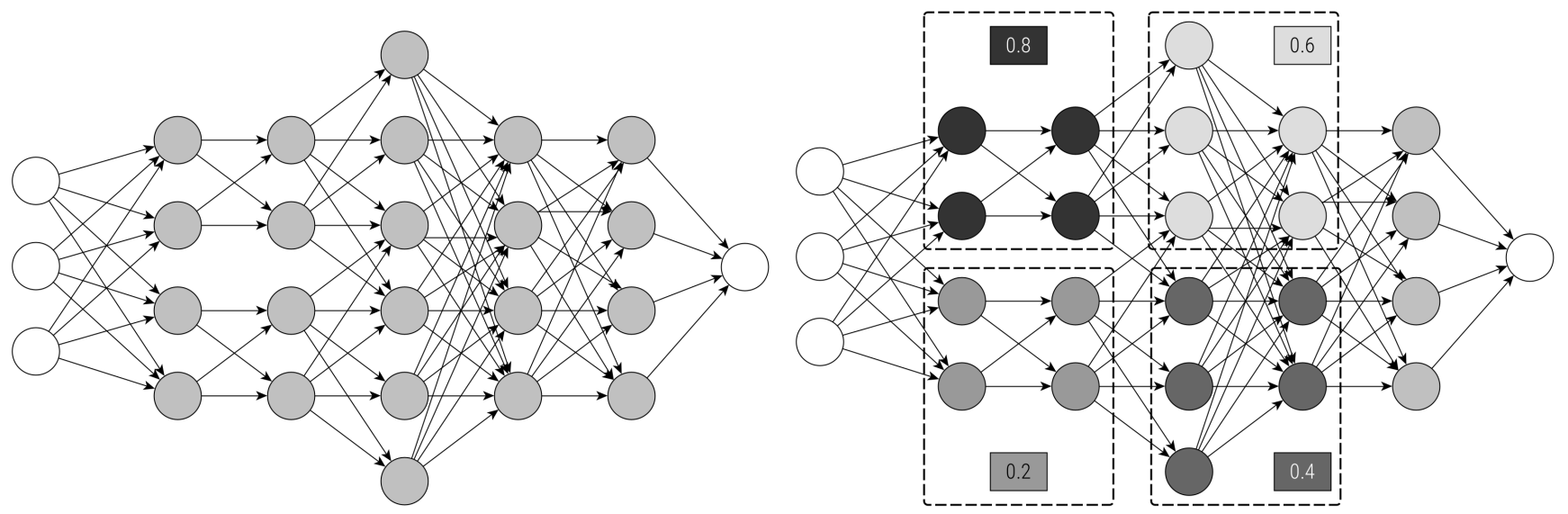

**Figure 11.2:** Sketch of a baseline network on an input dimension of three and an output dimension of one (left). The network is organised into four structural elements (dashed rectangles) and each element is attributed a value (right). Here, the values 0.8, 0.6, 0.2, and 0.4 are exemplarily chosen.

The third considered approach to obtain an amount per step is by having a bucket threshold value $\theta_b \in \mathbb{R}$ to be filled. This bucket value is heavily depdendent on the attribution method and can be e.g. thought of as an bucket amount of weight magnitudes to be removed. In this exemplary case, all structural elements are attributed with their weight magnitude, ranked accordingly and then added to the list of candidates for removal until the bucket threshold value $\theta_b$ is reached.

PRUNING PHASE Pruning can be conducted at various phases of a models lifecycle such as at *initialisation*, *before* training, *during* training, *after* training, during *fine-tuning*, or even at *inference*.

STOPPING CRITERIA A pruning pipeline usually contains multiple rounds and ends when a stopping criteria is reached. The stopping criteria can be a target *sparsity* level, a *fixed* number of pruning steps, an *empty* candidate list, a *full bucket* of attribution values, or a *threshold* on e.g. performance degradation.

The number of rounds are therefore often dynamic and determined by the stopping criteria, the chosen decisioning rule and the pruning amount.

SUMMARY ON PRUNING TERMINOLOGY A pruning pipeline for deep neural networks consists of many interchangable components. Empirical research mostly encompasses a large variety of different methods in which some of the methods' details such as the attribution method or stopping criteria are modified. The field of pruning is far from following a common terminology and experimental results remain difficult to compare. Terms like the *pruning granularity*, types of *pruning criteria*, *at-*

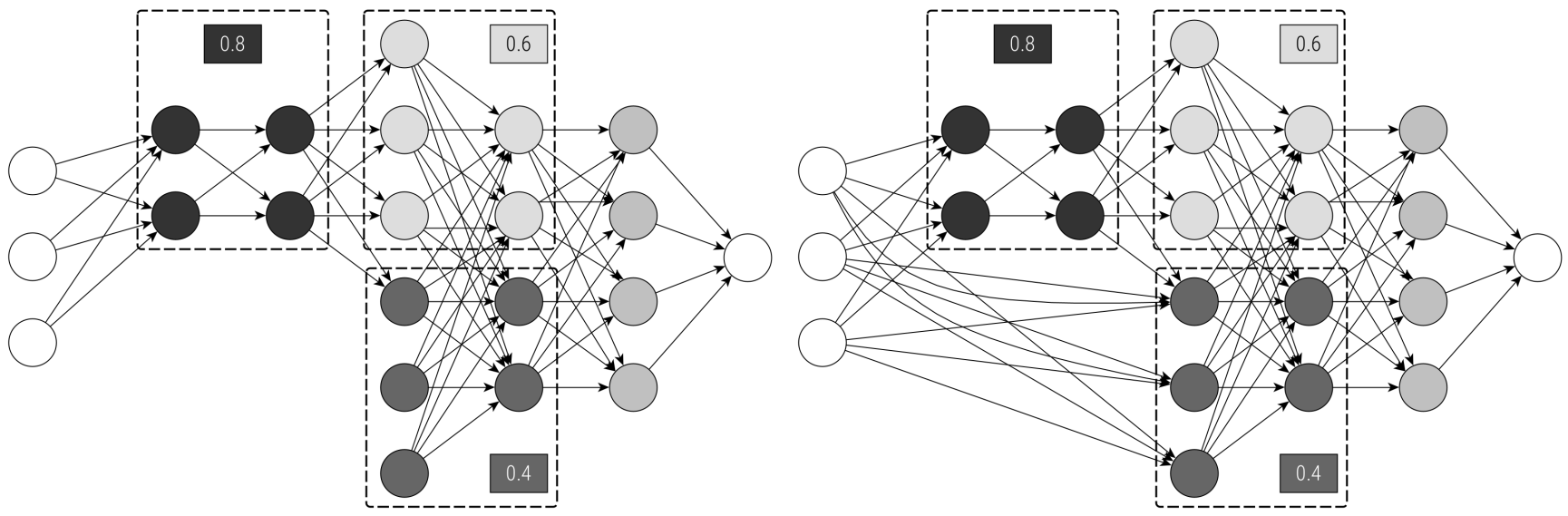

**Figure 11.3:** The network in fig. 11.3 is pruned based on the attributed values. Because the attributed value 0.2 in the lower left was lowest, the decisioning rule results in a network with this particular structural element removed. In a subsequent *rewiring step*, input and subsequent neurons of the removed elements are connected as to still allow for an adequate forward computation.

*tribution methods*, *pruning amount*, *phases* for pruning, and *stopping criteria* should be technically clarified to foster comparability and repeatability.

## 11.2 RELATED WORK ON PRUNING

Pruning dates back to at least the late 1980s and was already summarised 1993 in [232]. During the 1990s, a lot of focus has been put on varying attribution methods and ideas for deciding how to prune, compare [91, 99, 150, 199, 244, 258]. When the field of deep learning gained new traction during the mid 2010s, a lot of experimental comparisons have been conducted on state-of-the-art networks [9, 10, 191, 255] and new attribution techniques have been explored [88]. A significant methodological change in the field was the proposal of the Lottery Ticket Hypothesis [71] which was formulated based on observations that pruning could be used to find sub-structures (sparse masks) – which, when retrained with fewer epochs starting from the same original weight initialisation, achieved equivalent performances to fully trained networks. This observation sparked a lot of interest in lottery tickets.

Pruning as a methodology has become more mature w.r.t. compression and sparsity constraints when compared to earlier decades. The understanding, however, is limited to the facts that pruning can be successfully employed, the choice of element for removal makes a difference, or how the complexity of certain attribution methods relate. As far as we can judge, many unspoken questions in pruning remain unsettled:

- How can obtained structures in pruning be compared?
- Do different methods yield different structures?
- Can obtained structures be transferred across domains?

We provide an overview of selected related work in table 11.1 and refer to our collection of pruning articles up to 2019 on github and more recent surveys [288] for more related publications. As mentioned in [269], the field of pruning has been exploding on a quantitative level such that listing all published attribution methods is not meaningful anymore as compared to in e.g. Reed et al. [232]. More recent surveys can be found in [159, 288].

| Name | Year | Ref | Notes |
|---|---|---|---|
| Optimal Brain Damage | 1990 | [150] | |
| Optimal Brain Surgeon | 1993 | [92] | |
| Skeletonization | 1989 | [199] | |
| **Shapley Value - based Pruning** | 2014 | [154, 270] | |
| Snip: Single-shot network pruning | 2018 | [152] | |
| Movement Pruning | 2020 | [243] | |
| SupSup | 2020 | [312] | |
| SynFlow | 2020 | [279] | |
| SparCL: Sparse Continual Learning | 2022 | [295] | |

**Table 11.1:** A selected list of related work in the field of pruning with a focus on contributed methods. The list also serves a contextualisation of our own work [270] which is based on [154]. Using the Shapley Value for attribution of structural (hidden) elements of a deep neural network is an addition to the existing attribution methods such as magnitude-based [88] or change-in-performance-based [92, 150] methods

## 11.3 SHAPLEY VALUE BASED PRUNING

The Shapley Value as introduced in section 4.2 was first transferred to neural networks by Florin Leon [154] and further elaborated by our work in [270]. Central idea of applying SV to neural networks is to treat structural elements of a deep neural networks as players in a coalitional game. The structural elements collaborate in a game on the performance of the deep neural network as measured by e.g. its loss, accuracy or $R^2$ score. Each player and thus each structural unit is then assigned a value, the Shapley Value, with respect to the game settings and the estimated performance. The Shapley Value for each structural unit can then be interpreted as a value of contribution towards the estimated performance and with respect to all other structural units. A destructive approach such as pruning is then straightforward as simply the least contributing structural elements need to be removed. But also constructive or hybrid methods can be approached by using SVs of a structural unit.

In [270] we applied SV-based pruning to reduce neural network realisations in between sets of training batch iterations. Recall the training problem from eq. (7.1) on page 122:

$$\mathcal{F} \triangleq \underset{f \in \mathcal{A}([T])^{d_1 \to d_2}}{\arg\min} \mathcal{L}_{train}(f, D_{train}) \quad (7.1)$$

in which initially several gradient-based training rounds as in section 5.6.4 take place. The approach implicitly requires an a-priori defined architecture $\mathcal{A}([T])^{d_1 \to d_2}$. For neural network realisations in the intermediate result set $\mathcal{F}^*$ obtained through training, the Shapley Value method is used to assign attributions to structural elements of the realisations. Based on the attribution and other pruning decisions, structural elements with low Shapley Values, i.e. with low contributional value, are removed and a new set of neural network realisations $\mathcal{F}^{**}$ is obtained.

The removal is reflected technically by setting weights to zero such that gradient-based training won't affect parameters of these elements or the removal is imposed by explicitly setting a structural constraint, i.e. by multiplication with a mask or even making the pruning explicit in the used structural theme $T$. With the resulting set $\mathcal{F}^{**}$ end-to-end differentiable training for eq. (7.1) is continued.

For $f \in \mathcal{F}^* \subset \mathcal{A}([T])^{d_1 \to d_2}$ we can define a set of players by taking all neurons as player set $U$. Because $f$ is a neural network realisation, this set is finite and properly defined. The payoff can further be defined as $v(S)_f : \mathcal{P}(U) \to \mathbb{R}$ with $S \mapsto \alpha(f_S, D_{eval})$ for $f$ being a classifier or with $S \mapsto \mathcal{L}_{MSE}(f_S, D_{eval})$ for $f$ being a regression model with evaluation data $D_{eval}$. Here, $f_U$ restricts $f$ to have parameters of neurons not being in $S$ set to zero, i.e. weights & biases or other parameters related to neurons of $U \setminus S$.

The work in [270] compared this method with competing pruning methods which used weight magnitudes or random selection for attribution. We observed that both magnitude- and SV-based pruning make a significant difference over randomly pruning structural elements. However, the alternation between re-training and pruning during optimizing eq. (7.1) also shows that in over-parameterised regimes it is almost insignificant which method for attribution in pruning is chosen [270]. Because random- and weight-magnitude-based pruning attribution is considerably faster, it can be preferred over SV-based pruning in high-parameterised regimes or when the pruning granularity is considerably fine, i.e. the game size $|U|$ of the coalitional game for obtaining the Shapley Values is large. The attribution power for SV-based approaches might unfold for higher-level or coarse-grained structured pruning and when recent approximation techniques as outlined in section 4.3 are employed.

Beyond the training-pruning-setting in [270], the Shapley Value can also be employed to optimise on the second level of graph-induced neural networks. General possibilities to prune during architecture search was discussed in section 7.1.2 on page 125. Recall eq. (7.2) on page 122 for conducting architecture search for graph-induced neural networks:

$$\arg\min_{T\in\mathbb{S}} \mathcal{L}_{val}(T, \arg\min_{f\in\mathcal{A}([T])^{d_1\to d_2}} \mathcal{L}_{train}(f, D_{train})) \quad (7.2)$$

For a structural theme $T \in \mathbb{S}$ consider structural elements $u_1, \dots, u_n$ such as vertices of $T$. The player set consists of $U = \{u_1, \dots, u_n\}$ and we assume that $T\backslash S$ with $S \subset U$ is a valid structural theme of $\mathbb{S}$. Let $v(S) : \mathcal{P}(U) \to \mathbb{R}$ with $S \mapsto \min_{f\in\mathcal{A}([S])^{d_1\to d_2}} \mathcal{L}_{train}(f, D_{train})$ be the payoff of the game. Alternatively, consider the accuracy or $\mathcal{L}_{MSE}$ on a test set $D_{test}$. A solution based on the Shapley Value gives a pruned structural theme $T_{pruned} \in \mathbb{S}$ based on structural elements $S \subset U$ for an original structural theme $T \in \mathbb{S}$.

An alternative formulation of the coalitional game is to consider a player set $U = \{T_1, \dots, T_n\} \subset \mathbb{S}$ of structural themes. Let $\hat{v}(S) : \mathcal{P}(U) \to \mathbb{R}^{|S|}$ with $S \mapsto \{\min^{d_1\to d_2}_{f\in\mathcal{A}([T])} \mathcal{L}_{train}(f, D_{train})\}_{T\in S}$. An estimation of $\hat{v}(S)$ consists of minimum losses of neural network realisations, representing the structural themes contained in the coalition $S$. The payoff can then be defined with an aggregation such as the mean or minimum from $\hat{v}(S)$. A solution based on the Shapley Value yields a new architectural space $T_{new} = \bigcup_{T\in S} \mathcal{A}([T])$. For a payoff $v(S) : \mathcal{P}(U) \to \mathbb{R}, S \mapsto \frac{1}{|S|}\sum_{T\in S} \hat{v}(S)$ based on mean aggregation, we suppose that $T_{new}$ has a good and findable solution on average compared to other coalitions in $S$. More interestingly we interprete the payoff based on minimum aggregation, i.e. $v(S) : \mathcal{P}(U) \to \mathbb{R}, S \mapsto \min \hat{v}(S)$, as finding a $T_{new}$ which contains the largest smallest loss. This shrinks down the search space over all architectures induced from $U$ to a more interesting subset $S$ in which the training scheme appears to guarantee better estimates.

The first formulation is closer to pruning as in Leon [154] or Stier et al. [270] while the second formulation might act as a more general search principle where a structural relationship between the candidates of $\mathbb{S}$ are not assumed. Similar to other Shapley Value-based constructions as in [154, 270], the defined payoff might conflict with assumptions such as condition (2) $v(S) \geq v(S \cap T) + v(S\backslash T)$ of a payoff, defined in section 4.2 on page 45. We leave the comparison of both SV-based approaches to neural architecture search for future work.

REMARKS

- The Shapley Value as a solution concept can be applied during training a neural network with pruning and then compares to alternative pruning approaches such as randomly pruning or based on parameter magnitudes.
- We also transferred the game theoretic concept to neural architecture search and found an interesting interpretation for it.
- Applying the Shapley Value is very expensive as it requires estimates for many coalitions of $U$. Therefore approximations not only for the optimisation of eq. (7.1) on page 122 or eq. (7.2) on page 122 but also for Shapley Values need to be considered (compare section 4.3 on page 46).

## 11.4 AN ANALYSIS OF NON-ADDITIVE PAYOFFS

We noted in [270], that the marginal contributions $v(S) - v(S\backslash\{i\})$ "can produce negative values" when transferring the Shapley Value to settings in which non-additive payoffs are used. This is quite often the case: in [154] Florin Leon uses the loss $\mathcal{L}(\cdot) \in \mathbb{R}^+$ of a DNN and in [270] we use the accuracy $\alpha(\cdot) \in [0,1]$. Adding a structural element as a player to obtain $v(S)$ from $v(S\backslash\{i\})$ does not necessarily imply that the particular value of $v$ is increasing. The estimated loss or accuracy can also decrease such that the marginal contribution based on such a payoff becomes slightly negative.

Often, however, $v$ can be chosen such that it is bounded to $[0,1]$ in its codomain. This is e.g. the case for performance scores of neural networks as we have used them in [270]. The set-function $v$ then can be associated with non-superadditive measures and as shown in [34, Prop. 3.1, p. 7], "for non-monotonic measures $v : \mathcal{P}(U) \to [0,1]$, the Shapley Values $\phi_v(i)$ of each" player are "in between $-1$ and $+1$":

$$\begin{aligned}
\phi_v(i) &= \frac{1}{n!} \sum_{S\subseteq U\backslash\{i\}} (|S|!(n-|S|-1)!) \cdot (v(S\cup\{i\}) - v(S)) \\
&= \frac{1}{n} \sum_{S\subseteq U\backslash\{i\}} \frac{((n-1)-|S|)!|S|!}{(n-1)!} \cdot (v(S\cup\{i\}) - v(S)) \quad [34\text{, p. } 8] \\
&= \frac{1}{n} \sum_{t=0}^{n-1} \left( \frac{1}{k(|S|)} \sum_{j=1}^{k(|S|)} (v(S\cup\{i\}) - v(S)) \right) \qquad (11.1) \\
&\qquad \text{with } k(s) \triangleq \frac{(n-1-s)s!}{(n-1)!}
\end{aligned}$$

Cardin & Giove [34] derive this reordering in Equation 11.1 in which $k(s)$ is "the number of subsets of $U\backslash\{i\}$ with cardinality $s = |S|$" when "combinations of $n$ objects taken $s = |S|$ at a time. They further argue, that "such a value is minimum as soon as $v(S \cup \{i\}) = 0, v(S) = 1$, that is if"

$$v(S) = 0 \; \forall S \subseteq U \backslash \{i\}$$
$$v(T) = 1 \; \forall T \subseteq U \wedge i \in T \tag{11.2}$$

as then $v(S \cup \{i\}) - v(S) = -1, \forall S \in U \wedge i \notin S$. From this, Cardin & Giove derive $\min_{i \in U} \phi_v(i) = -1$ and note that "the upper bound can be computed in the same way". We look closer into a similar proof and even derive a tighter bound than $\phi_v(i) < 1$.

**Proposition 1** *For a non-superadditive payoff $v : \mathcal{P}(U) \rightarrow [0,1]$ of a game $U$ over $n$ players it holds for Shapley Values: $\phi_v(i) \geq -\frac{n-1}{n}$ for $i \in U$.*

We first see for **n=2**:
$\phi_v(i) = \sum_{S \subseteq U \backslash \{i\}} \frac{|S|!(n-|S|-1)!}{n!} (v(S \cup \{i\}) - v(S)) = \frac{0!(n-1)!}{n!}(v(\{i\}) - v(\emptyset)) + \frac{1!(n-1-1)!}{n!}(v(U) - v(\{i\})) = \frac{1}{2} v(\{i\}) \geq -\frac{1}{2} = -\frac{n-1}{n}$.

In the same manner, we observe that for **n=3**:
$\phi_v(i) = \sum_{S \subseteq U \backslash \{i\}} \frac{|S|!(n-|S|-1)!}{n!} (v(S \cup \{i\}) - v(S)) = \sum_{s=0}^{n-1} \sum_{j=1}^{C_s^{n-1}} \frac{(n-s-1)!s!}{n!} (v(S \cup \{i\}) - v(S)) = \sum_{s=1}^{n-1} \sum_{j=1}^{C_s^2} \frac{(n-s-1)!s!}{3!} (v(S \cup \{i\}) - v(S)) + \frac{(n-1)!0!}{n(n-1)!} (v(\{i\}) - v(\emptyset)) = \frac{(n-2)!}{n!} \Delta_1^1 + \frac{(n-2)!}{n!} \Delta_1^2 + \frac{(n-3)! \cdot 2}{n!} \Delta_2^1 + \frac{1}{n} \cdot v(\{i\}) \geq -\frac{1}{6} - \frac{1}{6} - \frac{2}{6} = -\frac{2}{3} = -\frac{n-1}{n}$.

We use a notation in which $\Delta_s^k$ denotes the marginal contribution for player $i$ for the $k$-th subset of size $s$ – for which we do not know its explicit value but can set it to $-1$ when looking for a lower bound. We further used the binomial coefficient $C_s^n = \frac{n!}{s!(n-s)!}$ to re-arrange the subset equation into a sum over all possible group sizes between $s$ and $n-1$ as Cardin & Giove did in Equation 11.1. In the limit $\frac{n-1}{n}$ tends to the same bound $-1$ but we observe two interesting points: first, the game size has a (minor) influence on the lower bound. Second, the payoff defined in Equation 11.2 has a unique structure and is actually the only payoff that reaches $\frac{n-1}{n}$.

We finalise the proof with the induction step from the argument over $n$ to $n+1$, assuming that it holds for $n \in \mathbb{N} : \sum_{S \subseteq U \backslash \{i\}} \frac{|S|!(n-|S|-1)!}{n!} \cdot (v(S \cup \{i\}) - v(S)) \geq -\frac{n-1}{n}$. For a game with player set $U^{n+1}$ of size $n+1$ we then get:

$$\begin{aligned}
\phi_v(j) &= \sum_{T \subseteq U^{n+1} \backslash \{j\}} \frac{|T|!(|U^{n+1}| - |T| - 1)!}{|U^{n+1}|!} \cdot (v(T \cup \{j\}) - v(T)) \\
&= \sum_{t=0}^{n} \sum_{k=1}^{C_t^n} \frac{(n-t)!t!}{(n+1)!} \Delta_t^k \\
&= \left( \sum_{t=0}^{n-1} \sum_{k=1}^{C_t^n} \frac{(n-t)!t!}{(n+1)!} \Delta_t^k \right) + \sum_{k=1}^{C_n^n} \frac{(n-n)!n!}{(n+1)!} \Delta_k^n \\
&= \left( \sum_{t=0}^{n-1} \sum_{k=1}^{C_t^n} \frac{(n-t)!t!}{(n+1)!} \Delta_t^k \right) + \frac{1}{n+1} \Delta_n^1
\end{aligned}$$

$$
\begin{aligned}
&= \left( \sum_{t=0}^{n-1} \sum_{k=1}^{C_t^{n-1}} \frac{(n-t)!t!}{(n+1)!} \Delta_t^k + \sum_{k=1}^{C_t^n - C_t^{n-1}} \frac{(n-t)!t!}{(n+1)!} \Delta_t^k \right) + \frac{1}{n+1} \Delta_n^1 \\
&= \left( \sum_{t=0}^{n-1} \frac{n-t}{n+1} \sum_{k=1}^{C_t^{n-1}} \frac{(n-t-1)!t!}{n!} \Delta_t^k \right) + \left( \sum_{t=0}^{n-1} \sum_{k=1}^{\frac{(n-1)!}{(t-1)!(n-t)!}} \frac{(n-t)!t!}{(n+1)!} \Delta_t^k \right) + \frac{1}{n+1} \Delta_n^1 \\
&\geq \left( \sum_{t=0}^{n-1} \frac{n-t}{n+1} \sum_{k=1}^{C_t^{n-1}} \frac{(n-t-1)!t!}{n!} \Delta_t^k \right) - \left( \sum_{t=0}^{n-1} \frac{(n-1)!}{(t-1)!(n-t)!} \frac{(n-t)!t!}{(n+1)!} \right) + \frac{1}{n+1} \Delta_n^1 \\
&= \left( \sum_{t=0}^{n-1} \frac{n-t}{n+1} \sum_{k=1}^{C_t^{n-1}} \frac{(n-t-1)!t!}{n!} \Delta_t^k \right) - \left( \sum_{t=0}^{n-1} \frac{t}{n(n+1)} \right) + \frac{1}{n+1} \Delta_n^1 \\
&= \left( \sum_{t=0}^{n-1} \sum_{k=1}^{C_t^{n-1}} \frac{(n-t-1)!t!}{n!} \Delta_t^k - \sum_{t=0}^{n-1} \left(1 - \frac{n-t}{n+1}\right) \sum_{k=1}^{C_t^{n-1}} \frac{(n-t-1)!t!}{n!} \Delta_t^k \right) \\
&\qquad - \left( \sum_{t=0}^{n-1} \frac{t}{n(n+1)} \right) + \frac{1}{n+1} \Delta_n^1 \\
&\geq -\frac{n-1}{n} + \frac{n}{n+1} - \frac{n}{n(n+1)} - \frac{n}{n(n+1)} \\
&= -\frac{(n-1)(n+1)}{n(n+1)} + \frac{n^2}{n(n+1)} - \frac{n}{n(n+1)} - \frac{n}{n(n+1)} \\
&= -\frac{(n-1)(n+1)}{n(n+1)} + \frac{n^2}{n(n+1)} - \frac{n}{n(n+1)} - \frac{n}{n(n+1)} \\
&= \frac{-n^2 - n + n + 1 + n^2 - 2n}{n(n+1)} = \frac{-2n+1}{n(n+1)} \geq -\frac{n}{n+1}
\end{aligned}
$$

Some of the steps can be abbreviated but we wanted to emphasise on leveraging following information:

1. There's always only one $\Delta_0^k$ and it takes its minimum at $v(\{j\})$ with $j$ being the current player for $\phi_v(j)$. Hence it must be $v(\{j\}) = \Delta_0^k = 0$ to be a minimal summation term (recall that $v$ is bounded in $[0, 1]$).

2. The marginal contributions of $j$ stemming from coalitions below the grand coalition $\Delta_n^1$ can be easily factored out and grow linearly with growing game sizes.

3. Even when not needing it for the induction proof, each coalition of size $t \in \{1, \dots, n-1\}$ is refactored with a growing $n$ such that we observe the extra term $\frac{n-t}{n+1}$ in the left hand side of the summation over all $t$. At the same time, the additional marginal contributions per coalition size $t$ can be rearranged from this inner sum by pulling out $C_t^n - C_t^{n-1}$ terms. We slightly overloaded the notation $\Delta_t^k$ but as we only want to show the inequality it should be clear that these are all different marginal contributions and they will be larger or equal to $-1$.

■

For the subset counting of Shapleys original formulation of $\gamma_n$ in Equation 4.1 we observe for $n > 1$ and $|S| \leq n+1$:

| | Changes or Example |
|---|---|
| $\gamma_n(s) = \frac{|S|!(n-|S|-1)!}{n!}$ | $\frac{(1\cdot 2\cdot\cdots\cdot|S|)(1\cdot 2\cdot\cdots\cdot(n-|S|-1))}{1\cdot 2\cdot\cdots\cdot n}$ |
| $= \frac{\left(\prod_{i=1}^{n-|S|-1} i\right)\prod_{i=1}^{|S|} i}{\prod_{i=1}^{n} i}$ | $\frac{(1\cdot 2\cdot\cdots\cdot(n-|S|-1))(1\cdot 2\cdot\cdots\cdot|S|)}{1\cdot 2\cdot\cdots\cdot n}$ |
| $= \frac{\left(\prod_{i=1}^{n-|S|-1} i\right)\prod_{i=1}^{|S|} i}{\left(\prod_{i=1}^{n-|S|-1} i\right)\left(\prod_{i=n-|S|}^{n} i\right)}$ | $\frac{(1\cdot 2\cdot\cdots\cdot(n-|S|-1))(1\cdot 2\cdot\cdots\cdot|S|)}{(1\cdot 2\cdot\cdots\cdot(n-|S|-1))((n-|S|)\cdot(n-|S|+1)\cdot\cdots\cdot n)}$ |
| $= \frac{\prod_{i=1}^{|S|} i}{\prod_{i=n-|S|}^{n} i} = \frac{\prod_{i=1}^{|S|} i}{\left(\prod_{i=n-|S|}^{n-1} i\right)\cdot n}$ | $\frac{1\cdot 2\cdot\cdots\cdot|S|}{(n-|S|)\cdot(n-|S|+1)\cdot\cdots\cdot(n-1)\cdot n}$ |
| $< \frac{1}{n}$ | |

On a side note, observe, that we obtained this inequality through a reordering and based on the fact that $|S|! = \prod_{i=1}^{|S|} i$ has $|S|$ many factors while $\prod_{i=n-|S|}^{n} i$ has $n-(n-|S|)+1 = |S|+1$ factors such that each factor in the numerator has a larger counterpart in the denominator and gets cancelled out.

WHAT IS THIS ANALYSIS GOOD FOR? Drawing back the connection to neural networks, we empirically observe that most contributional values $\phi_v(i)$ are close to zero or positive – based on $v$ being defined on e.g. the accuracy of a neural network. Although they can take arbitrary negative values close to $-1$, we observe that neural networks behave semi-monotonically w.r.t. structural elements being removed or added. What do we mean with semi-monotonicity? Payoff functions defined on a performance score of a neural network and structural elements can not be monotonic as required for nice properties of the Shapley Value. But we can observe a slightly weaker property, that is, the average of this score across several sizes of structural elements grows monotonoically. This let's us treat a setting with neural networks similar to coalitional game settings in which monotonicity is strictly provable.

We hypothesise from the conducted analysis, that under many conditions of coalitional games with neural networks, payoffs often behave semi-monotonically, although this is not properly proven, yet. Semi-monotonicity, however, would mean, that more contributing structural elements generally benefit the performance and this is obviously highly related to the circumstance that universal approximation is usually achieved in the limit of adding sufficiently many parameters to closely resemble a target function. These dynamics are important to understand the effects of growing, pruning, and how certain universal architectures approximate functions.

### 11.5 SUMMARY ON PRUNING

We gave a brief introduction to pruning deep neural networks. Limitations of the field and our work were contextualised in section 11.2 and we concluded that quantitative experiments are currently limited in producing new insights into pruning methods. Our main contribution in pruning based on Shapley Values was newly outlined based on graph-induced neural networks in section 11.3 and we detailled how to extend SV-based pruning to problems of neural architecture search. The formal issues mentioned in Stier et al. [270] led us to analyse properties of non-additive payoffs in coalitional games as found in context of deep neural networks. We elaborated on a related proof of Cardin & Giove [34] and worked on a lower bound proof for SVs. The obtained lower bound is interesting for very small games which themselves are practically feasible in context of pruning DNNs. Details in the proof suggest that with re-formulating a payoff with a property of semi-monotonicity across sizes of coalitions, one might gain properties close to the original Shapley Value.

REMARKS

- We discussed theoretical aspects of applying Shapley Value to optimizing problems of graph-induced neural nets.
- Related and own empirical work on quantitative pruning experiments suggest that the insights are limited if the theoretical assumptions are not more clearly layed out. We see new potential results in formal analysis of properties of the search space to progress pruning methods rather than empirically conducting parameter reduction of overparameterised deep neural networks to see which selection method is superior.

# 12 NEURAL ARCHITECTURE SEARCH

*The following entails:*

Neural architecture search is concerned with the automated design of neural networks. The field on NAS can be considered as a sub-field of automated machine learning (AutoML) and has overlaps with hyperparameter optimisation (HPO) [304, Sec. 1.1]. Pruning or growing techniques can be also found in NAS but have evolved also separately because of different goals (as in compression), assumptions and employed methods.

A recent article titled *Neural architecture search: Insights from 1000 papers* [304] illustrates the extend of the field and reports an increase of published work of over 600 articles [304, p. 3 Fig. 1] per year. The scientometric data is based on a list [49] which does not take many publications from before 2015 into account in which progress was made under different terms such as *neuroevolution* [68] or *neural network design* [196]. The term neural architecture search was assumably popularised after Zoph & Le [339] presented their method "Neural Architecture Search" in which a recurrent neural network was trained through reinforcement learning to come up with a sequence of hyperparameters and labels for a convolutional neural network to achieve state-of-the-art error rates on `CIFAR10` and low test perplexity on the `Penn Treebank corpus`.

Our definition of computational themes (CTs) in section 7.2 on page 130 [274] is an interesting search space design for neural architecture search and contains both theoretical considerations and empirical outcomes. We also have experimented with evolutionary and genetic architecture searches [301] and differentiable architecture searches [205]. Further, we present a set of experiments which affirm critical arguments about genetic algorithms applied to genetic search spaces based on graphs. For example, we show for several genetic operations on `NAS-Bench-101` that the effect is similar to random sampling and we argue that this transfers to most design spaces if the particular genetic operation is not carefully designed and evaluated.

## 12.1 NEURAL ARCHITECTURE SEARCH

A strict terminology for neural architecture search is difficult and assumably in its infacy. Following Elsken et al. [61], it became common to differ between the search space, search strategy and the performance estimation strategy. Figure 12.1 illustrates the relation between the three components of a NAS: the estimation strategy depends on the search strategy (and search space) and the search strategy depends on the search space.

The search space influences which sampling methods are suited to obtain unexplored candidates and how candidates represent realisations of deep neural networks. In context of genetic algorithms, a candidate is also often called a genotype and the search space then is a genotypical space [196]. A realisation then is a phenotype or *phenotypical expression* of the genotype. Mapping a candidate of the search space to a realisation involves stochastic aspects as a candidate can represent different functional forms of a neural network realisation and further will undergo a training scheme such that certain realisations are more likely than others, even though they belong to the same (or sometimes multiple) candidates.

A search strategy involves one or multiple sampling methods to obtain new candidates from the search space and the strategy further involves rules on how to learn based on previous evaluated candidates. The strategy therefore combines aspects of search & optimisation on a higher level optimisation problem as e.g. described in Section 7.1. Search strategies often suffer from non-differentiable objectives or containing problems of integer programming or generally tackling discrete (outer) optimisation problems.

The performance estimation strategy depends on the broad complexity of the NAS problem in the sense, that with limited computational budget the performance estimation needs to be reliable, fast or satisfy conflicting objectives such as reflecting both accuracy and robustness (or other targeted goals for the underlying DNN). Training pipelines include many hyperparameters that could influence candidates differently such that fair and reliable performance estimations among candidates are already difficult or computationally expensive. Often, NAS applications then involve multiple training schemes for faster but more fuzzy, and slower and more reliable estimations – called e.g. estimation and final training. For example, the fast estimation can then be used during search and a final training can be used to evaluate results of a NAS.

Analogously to pruning, terminology in NAS varies considerably because of following reasons:

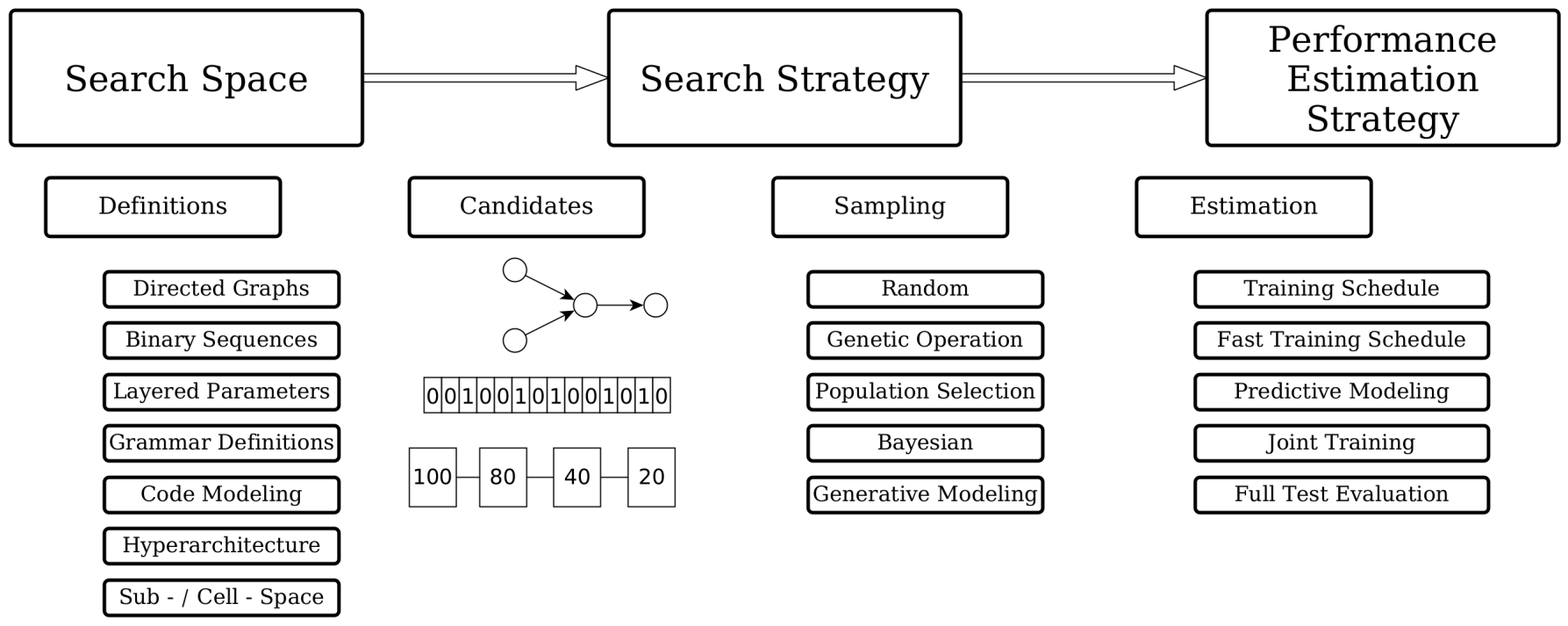

**Figure 12.1:** An established terminology in neural architecture search is the separation of *search space*, *search strategy* and *performance estimation strategy* [61]. The search space involves the expressivity of all potentially findable realisations of neural network models and combines related realisations in one candidate representation. A NAS manoeuvres through this search space by means of a search strategy. Most search spaces do not provide gradient information such that other sampling strategies need to be considered. Candidates are assessed by an performance estimation strategy that e.g. ranks candidates by their test accuracy. Different strategies can be employed, e.g. fast and extensive schedules, as to speed up the overall search under the tradeoff of having less accurate estimations for the performance quality of a candidate.

1. The importance of each of the three terms within neural architecture search can differ heavily. Often the optimisation landscape provides no gradient information such that gradient-free optimisations are chosen. Comparisons then focus on search methods under umbrella terms such as *swarm*, *population* or *evolutionary* NAS. Other works focus on comparing different estimation strategies to even obtain a good approximate optimisation landscape and then barely have resources to compare varying search methods.

2. Results are hardly transferrable across different search spaces, especially when they are of empirical nature.

3. Performance estimation strategy are mostly based on expert experience, difficult to formalise, and they usually exhibit strong differences for candidates of the same search space.

4. The search space can be relaxed into hyperarchitectures as done in DARTS such that its expressivity is heavily restricted but at the same time the importance of its definition also shifts while providing a basis for gradient-based search strategies.

The field of NAS lacks formality and contains an overwhelming body of empirical work [304]. From a formal perspective, our problem formulations based on graph-induced neural networks as proposed in Section 7.1 can be seen as a unification of problems in neural architecture search. But this has to be taken with the same grain of salt as in above stated difficulties in NAS terminology and formal unifications in general: unusual approaches and edge cases are often hardly covered in unified formulations and even if they are, the formulation needs to bring benefit to the proposed problem. One alternative solution approach to more formality was to bring more convention to the field as by Lindauer & Hutter [165] and to advance transferability of results through neural architecture search benchmarks.

## 12.2 NEURAL ARCHITECTURE SEARCH BENCHMARKS

While `NAS-Bench-101` [326] is widely considered the first extensive benchmark dataset on neural architecture search, there were data and results on neural architecture search runs before, e.g. in [272]. Still, `NAS-Bench-101` can be considered as the first large benchmark with major impact as it provided both a proper search space definition, an extensive evaluation of contained candidates and easily accessible data through a python API.

Lopes et al. [170] broadly categorise benchmarks into two types, **tabular** ones and **surrogate**-based ones. With tabular, they mean extensively evaluated architecture candidates of a defined search space with a described training and evaluation scheme for which samples are available in a data table with direct access to its results. We also considered such benchmarks in [272, 274, 301] and our CTNAS-data falls into this type of benchmark. With surrogate-based benchmarks, they refer to benchmark data in which a predictive model provides performance estimates of search space candidates. This either is used for predictive surrogate evaluations (compare chapter 13 on page 249 on advanced predictive models for NAS) or for compressing or aggregating data into a predictive model instead of providing (or even sampling) the full search space. Lopes et al. [170, Tbl. 1, p.2] provide an overview of both types of benchmarks of which selected ones overlap with our overview in table 12.1 on the facing page. Since 2020 the availability of NAS benchmarks has significantly exploded such that we only provide an overview on the literature here.

| **Title** | Year | **Ref** |
|---|---|---|
| Structural Analysis of Sparse Neural Networks - `graphs10k` | 2019 | [272] |
| NAS-Bench-101: Towards Reproducible Neural Architecture Search | 2019 | [326] |
| Nas-bench-201: Extending the scope of reproducible neural architecture search | 2020 | [56] |

| Title | Year | Ref |
|---|---|---|
| Nas-bench-301 and the case for surrogate benchmarks for neural architecture search | 2020 | [257] |
| Nas-bench-1shot1: Benchmarking and dissecting one-shot neural architecture search | 2020 | [333] |
| Nats-bench: Benchmarking nas algorithms for architecture topology and size | 2021 | [55] |
| NAS-Bench-x11 and the Power of Learning Curves | 2021 | [322] |
| Nas-bench-asr: Reproducible neural architecture search for speech recognition | 2021 | [192] |
| Hw-nas-bench: Hardware-aware neural architecture search benchmark | 2021 | [155] |
| Nas-bench-nlp: neural architecture search benchmark for natural language processing | 2022 | [134] |
| Surrogate NAS benchmarks: Going beyond the limited search spaces of tabular NAS benchmarks | 2022 | [332] |
| CTNAS-database | 2024 | [274] |

**Table 12.1:** A selection of neural architecture search benchmark data sets in literature. The growth of publications since 2019 shows that the NAS community tries to unify search space definitions to provide faster and repeatable evaluations of NAS-runs while ultimately comparing search strategy such as evolutionary searches, DARTS, bayesian optimisation, Covariance Matrix Adaption Evolutionary Strategy (CMA-ES) and many other alternative search methods. We are, however, far from a good understanding of trade-offs between good search space design, diversity of applied domains and the richness of arguments that can be drawn from NAS-runs on these benchmarks.

## 12.3 EVOLUTIONARY ARCHITECTURE SEARCH

One of the simplest approaches to conducting automated neural architecture searches is based on evolutionary principles. These evolutionary principles include using populations of candidates from a search space, conducting selection among these candidates and optionally also employing mutation such that an overall fitness objective – the performance estimation in NAS – is maximised. The same principles are also the basis for genetic algorithms.

An evolutionary algorithm holds a population $P$, a set of candidates, per each time step $t \in \{1, \dots, T\}, T \in \mathbb{N}$. The initial population is obtained through a sampling method, usually an uninformed random sample. For a search space $\mathbb{S}$ one can write $\forall i \in \{1, \dots, k\} : P_i \sim \mathbb{S}$ such that $k \in \mathbb{N}$ initial candidates are drawn.

Eiben & Smith [59] name two main forces that drive evolutionary searches:

- "Variation operators" such as recombination and mutation "create the necessary diversity within the population, and thereby facilitate novelty"
- and "selection acts as a force increasing the mean quality of solutions in the population" [59, Sec 3.1 p. 26].

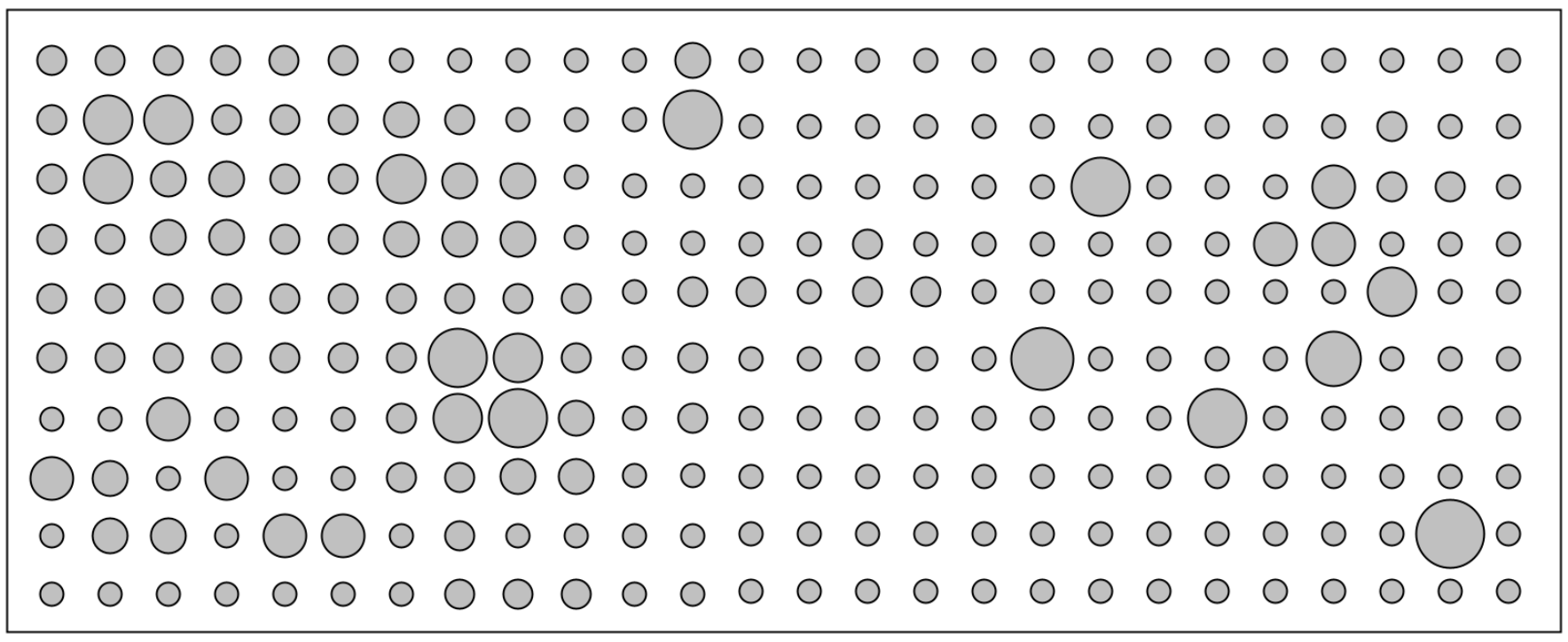

**Figure 12.2:** Visual sketch of a search space $\mathbb{S}$ on a grid that resembles locality information. Each candidate, visualised as a circle, has a true but usually unknown performance, e.g. a generalisation test set accuracy. The size of each circle reflects the true performance. The space is usually so huge, that it can not be exhaustively sampled. Benchmark datasets often provide exhaustive sampled and evaluated spaces as to compare NAS methods such as evolutionary searches.

During each time step, unevaluated candidates of the population are evaluated based on an performance estimation strategy. With predefined rules, e.g. based on a threshold or a proportion of candidates, the worst candidates are discarded from the population – or in other words, certain candidates are selected *for survival*.

New candidates are either sampled randomly or based on information of the evaluated candidates. Optionally or additionally a mutation of existing candidates can be employed. The proportions of which sampling method to choose is the origin of the *exploration vs. exploitation trade off*. Random sampling possibly is best in uninformed exploration of the search space $\mathbb{S}$ while sampling with informed information from evaluated candidates or mutation of selected candidates leads to exploitation of parts of $\mathbb{S}$ in which already better candidates than in previous steps have been drawn.

"Roughly speaking, *exploration* is the generation of new individuals in as-yet untested regions of the search space, while *exploitation* means the concentration of the search space in the vicinity of known good solutions" [59, Sec. 3.5]

Figure 12.2 shows a sketch of a search space on a two-dimensional grid. Larger circles illustrate higher performances of phenotypes obtained from a single genotype in the search space. The figure can be interpreted as a top-down look on a hill landscape, similar to analogies used in gradient ascent methods in which the highest point in a function landscape (the optimisation objective) is to be searched for. A main difference is that an evolutionary algorithm does in general not require gradient-information and can be applied to discrete spaces, as well.

Figure 12.3 illustrates a random draw that provides initial candidates of the population. Often, the performance or fitness is an estimation and therefore the information obtained for a single candidate is not as reliable as the fixed size of a circle in the figure suggests. In such a case, the fitness landscape would be more fuzzy.

A second draw, as depicted in the right of Figure 12.3, requires the evaluation of new candidates. In large search spaces it is often unlikely to draw the same candidates again. But one could also explicitly avoid evaluating the fitness of candidates twice by keeping previously chosen candidates or the current population in memory.

The selection process reduces the population based on their estimated performances and by this increases the mean of the overall population [59, Sec 3.1 p. 26]. Iteratively applying a random sampling method in such a process can then be seen as an evolutionary random search in which the algorithm ultimately returns the best performing candidate of the last population generation at time step $T$.

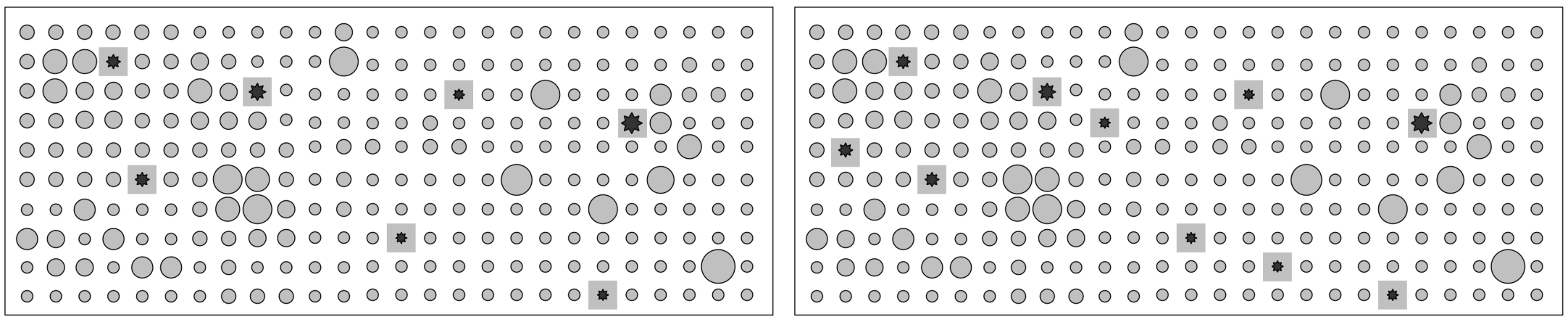

**Figure 12.3:** Candidates have been sampled (left) from the underlying search space of Figure 12.2, denoted as dark stars instead of circles. Seven candidates are randomly drawn but the ones with very large performance are missed with high probability. A random search would consecutively continue with new random samples (as depicted on the right). Such an approach is a common baseline for an evolutionary search in an empirical comparison of both NAS methods.

The sketches in figures 12.2, 12.3, and 12.4 are created on purpose in a way that the left half of the depicted space has slightly different properties than the right half. For the left half, the probability of finding a better candidate than on average is increased if the current candidate is already better than on average. In other words, large circles accumulate locally and the fitness landscape can be considered more smooth. In such a case, it would make sense to sample new candidates nearby if one already found a better candidate than on average. Locality is a property that many mutation operation use to sample better candidates.

## 12.4 GENETIC ARCHITECTURE SEARCH

The collective of methods to derive new candidates within the candidate representation search space is referred to as *variation operator* [59, Chp 4]. Mutation is then a subset of these variation operators which uses one parent candidate from the population to derive one child by applying random changes on the parent. The mutation operation is a *unary* operation on a selected parent candidate.

Generalizing this principle leads to *binary* operations, known as *crossover*, and with the arity $a \in \mathbb{N}^+$ of a variation operator they can be categorised for $a > 2$ into "multiparent combination operators" [59, Chp 4 p.50]. At the latest with the use of crossover, the evolutionary search is called a *genetic search* and the search space is considered a genetic search space with genotypical representations in it.

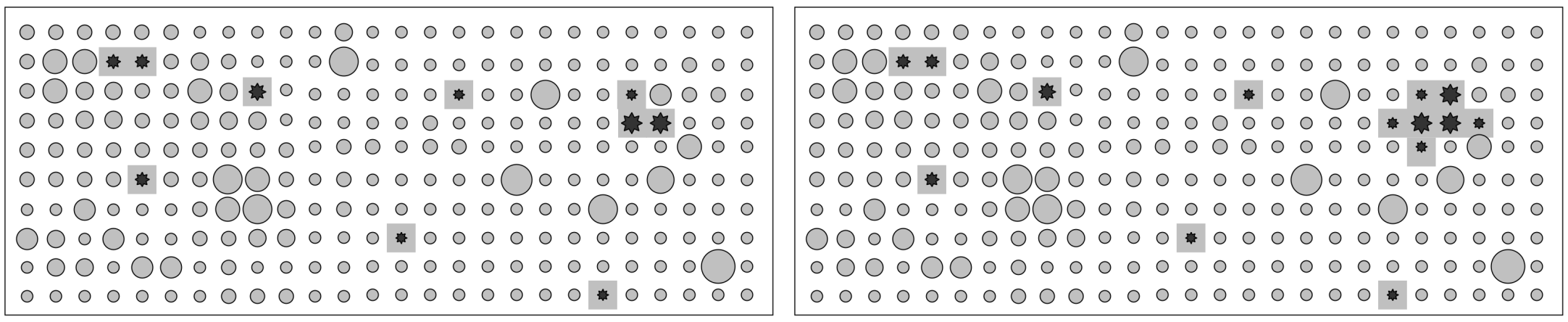

**Figure 12.4:** While a random search as in Figure 12.3 stays uninformed about the search space, the mutation operation samples *nearby* candidates by mutating properties of previously chosen and selected candidates. Nearby then refers to topological properties of the search space $S$. The assumption would be, that the probability for finding better candidates in a close region of already well performing ones is higher than on the overall space. This can be true in some spaces as in the depicted one: larger circles are locally surrounded with larger circles than on average – at least on the left half of the picture. Left: generation one with mutated candidates after the first population was randomly sampled. Right: generation two after another round of adding locally mutated candidates.

To employ variation operators such as a mutation as sketched in Figure 12.4 on the right, suitable representations need to be chosen. In fact, Eiben & Smith note that "the first stage of building any evolutionary algorithm is to decide on a genetic representation of a candidate solution to the problem" [59, p. 49]. While the illustrations in 12.2, 12.3, and 12.4 suggest a two-dimensional coordinate representation, we practically work with DAGs as representation space to tackle problems of graph-induced deep neural networks as outlined in Section 7.1. But there are various alternative representations such as binary representations [59, Sec. 4.2], integer [59, Sec. 4.3], real-valued or floating-point [59, Sec. 4.4], permutation [59, Sec. 4.5], or tree representations [59, Sec. 4.6].

Recent empirical comparisons of graph-related representations in evolutionary computing such as by Françoso et al. [70] concentrate on trees as in standard literature [59, Sec. 4.6], linear genetic programming [27], Cartesian Genetic Programming (CGP) [197, 198], and evolving graphs by graph programming (EGGP) [8]. This suggests that employing directed acyclic graphs as representation space for evolutionary algorithms has only recently been explored empirically and theoretical results are scarce. We suppose that the complexity of directed acyclic graphs as presented in Section 3.4 on DAGs is one reason for it.

From a theoretical point of view, it was just recently proven by Dang et al. & Doerr et al. that crossover as variation operator during an evolutionary multi-objective optimisation can actually guarentee exponential speed-ups [46, 54] compared to not using crossover. Whether the results with the employed binary representation transfer to other representations such as with graphs and whether our multi-objective problems with graph-induced neural networks might fall in the proposed (or related) class of problems as presented by [46] remains unsettled. However, these theoretical guarantees are a strong argument for investigating on $a$-ary genetic variation operators for multi-objective NAS.

Evolutionary and genetic algorithms can be employed to tackle the problem of finding architectures in the framework of graph-induced neural networks. Recall Equation 7.2 from Section 7.1 on optimisation problems:

$$\arg\min_{T \in \mathbb{S}} \mathcal{L}_{val}(T, \arg\min_{f \in \mathcal{A}([T])^{d_1 \to d_2}} \mathcal{L}_{train}(f, D_{train}))$$

Here, $\mathbb{S}$ denotes the search space as formulated in section 7.2 of valid directed acyclic graphs with structural themes $T$ being candidates of $\mathbb{S}$. The optimisation problem over graph-induced is tackled by genetic neural architecture searches with an initial population $P^{(0)}$ and the population changes from a generation of time step $t = 0$ with $P^{(0)}$ to $P^{(n_{gen})}$ of generation $n_{gen} \in \mathbb{N}$ through application of variation operators and selection on parts of these populations.

Figure 12.4 illustrates variation operators in the form of mutations as obtaining a topological nearby candidate in the search space. Variation operators can therefore be seen as $a$-parent functions $\mathbb{S}^a \to \mathbb{S}$ which yield a new candidate by recombination of $a \in \mathbb{N}^+$ parents in the search space $\mathbb{S}$.

Genetic algorithms contain various hyperparameters such as probabilities for mutation or crossover [59, Sec 4.1, p. 50], or population sizes [59, Sec 7.1, p. 119]. A powerful aspect of genetic algorithms is to be able to encode such parameters alongside with solution candidates such that genetic algorithms can easily operate on varying levels of the multi-level optimisation problems of graph-induced neural networks.

Taking the two most essential components of genetic algorithms applicable to NAS over graph-induced neural networks, how do *selection* and variation operators relate? We will give examples of variation operators in graph space and wonder: can any operator be chosen or are there differences among them?

## 12.5 EXPERIMENT: ANALYSIS OF OPERATIONS ON NAS-BENCH-101

Many NAS benchmarks employ variation operators in graph space. This strengthens our perspective of graph-induced neural networks and making graph structure a first-class citizen of neural structure analysis and neural architecture search. But it also raises a question on whether variation operators in graph space are even meaningful.

We conduct an experiment to compare $k = 4$ variation operators $v_k$ in graph space and compare resulting neural networks of `NAS-Bench-101` based on performance estimations of the benchmark. More formally, we randomly sample [1] $\{g_1, \dots, g_{n_{sources}}\} \sim UDAG([2, 7])$ and denote the estimated test accuracy provided through `NAS-Bench-101` with $\alpha(g_i) \forall i \in \{1, \dots, n_{sources}\}$. The test accuracy is estimated after transforming a graph into a neural network and training it repeatedly as of the training scheme of `NAS-Bench-101`.

Each variation operator $v_k : \mathcal{G} \to \mathcal{G}$ is analysed independently on the basis of $n_{sources} = 100$ randomly sampled source architectures. Up to $n_{target} = 20$ graphs are obtained by applying the variation operator $g_{i,j} = v_k(g_i), j \in \{1, \dots, n_{target}\}$ independently. This theoretically makes up for 2000 samples per operation, although the actual number is less as each variation operator has to be performed multiple times as to obtain a graph that conforms with the space restriction of `NAS-Bench-101`.

The result is an insight into the change of expectation in estimated performance on `NAS-Bench-101` and gives us an answer on how the operations relate to random sampling in the search space and amongst each other.

1 Compare random DAG sampling in appendix A.6 on page 318 [140]

Can we find variation operators in graph space that improve NAS?

**Search space $\mathbb{S}$:** `NAS-Bench-101` (7.2.2)
**Method:** Independent random draws in the graph space provide a baseline to which the operations are compared to. We inspect the expected mean performance, expected variance and change to the source graphs for each operation.

**Data:** Test accuracies of `NAS-Bench-101` serve as performance estimates and graphs are sampled independently and uniformly between three and seven vertices from UDAG [140].

**Interpretation:** Variation operators are mostly a tool for exploration but the results show that they significantly differ when compared for steering towards optimal solutions. Results are very likely to be dependent on the underlying application domain. Due to its connection to parametric growth, variation operators that increase the parametric size can improve the search. Main goal, however, is a explorative and exploitative nature of variation operators that supports the selection process in navigating through the search space.

### 12.5.1 *Experiment Setup And Selected Variation Operators*

Random samples in a search space $\mathbb{S}$ of a neural architecture search are a baseline for every strategy. A random draw from `NAS-Bench-101` results in a distribution similar to the one depicted in fig. 12.5. Here, the distribution of performances are, however, separated by the number of vertices. We can observe, that the average expected performance test accuracy $\bar{\alpha}$ is approximately 0.8948. This observation is an important baseline on which to compare variation operators. Although the result will slightly differ from draw to draw, we need to consider both single draws and many draws over a long running search strategy as to understand the dynamics of variation operators. The reported accuracies of the authors of `NAS-Bench-101` "is above 90% for a majority of models" and "the best architecture [..] achieved a mean test accuracy of 94.32%" [326, p. 3.1].

We chose four common variation operators applied on graphs in the search space, namely *relabeling*, *rewiring*, *contraction* and *distraction* and we will stick to that order in resulting data visualisations. An example of variation can be seen in fig. 12.6 in which fig. 12.6a depicts an exemplary source graph on which the other operations are applied and visually depicted. In the following experiments we use the operations depicted in figs. 12.6b to 12.6e but also note at this point with fig. 12.6f that there exist many more alternative operations.

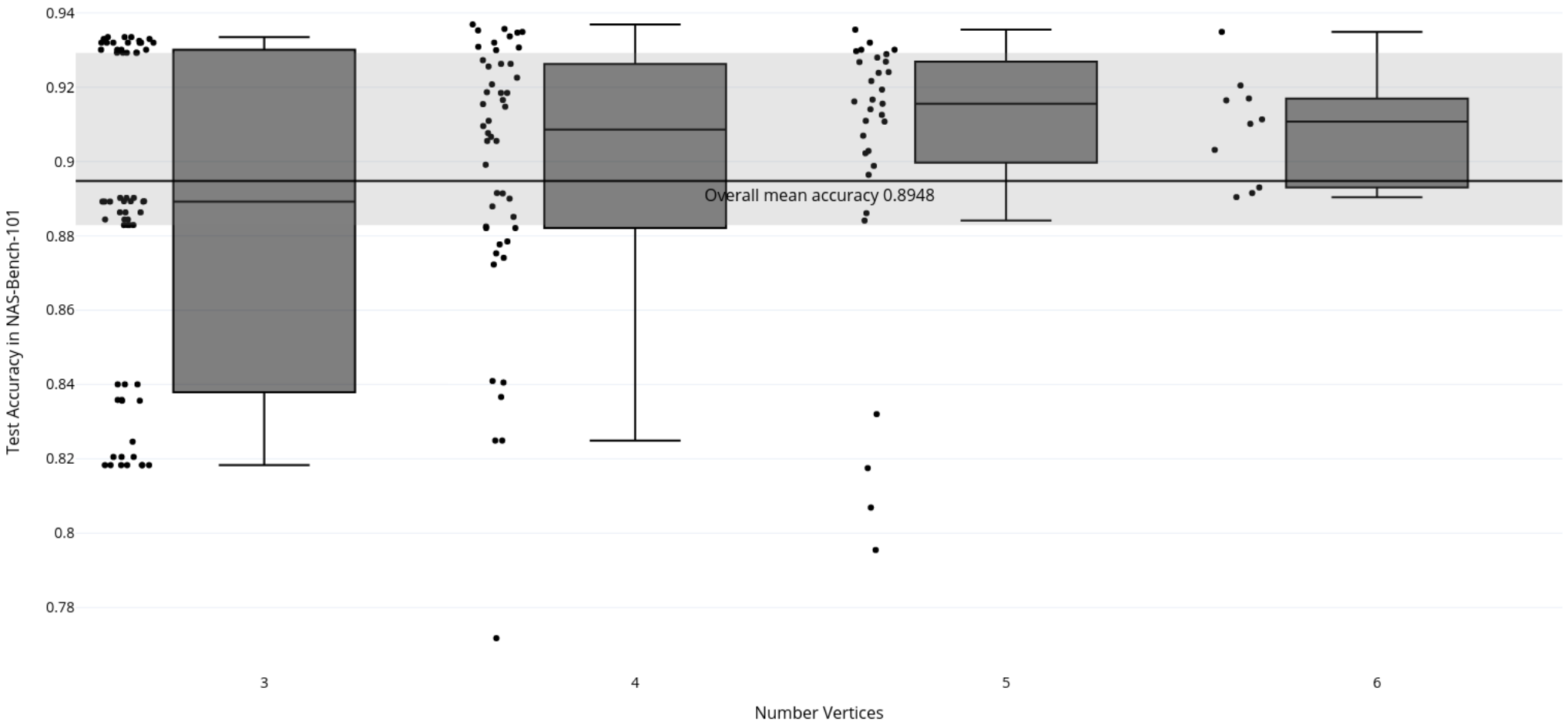

**Figure 12.5:** Drawing a random directed acyclic graph of at most seven vertices, labelling its at most five inner vertices randomly with a possible operation of `NAS-Bench-101` (namely *CONV1X1*, *CONV3X3*, *MAXPOOL3X3*) and querying the `NAS-Bench-101` database results in the depicted distributions of test accuracies. The gray shaded area depicts the lower .25- and upper .75-quantiles of the overall distribution of test accuracies of the neural architectures. It can easily be seen that more vertices increase the average test accuracy, although all groups of at least four vertices contain architectures with maximal test accuracy performance of above 0.93%. Drawing an architecture uninformed uniformly would give roughly above 0.9% test accuracy on this database, considering that it is more likely to draw architectures of larger vertices as there exist more of them (the order of graphs is not uniformly distributed).

The possible (neural) operation labels in `NAS-Bench-101` are *CONV1X1*, *CONV3X3*, and *MAXPOOL3X3*. Further, the input vertex and the output vertex need to be labelled as *INPUT* and *OUTPUT*, respectively. In case, an operation modifies the input or output vertex, the resulting graph is relabelled as to fit with this convention. For example, if an output vertex is duplicated, only one vertex will be labelled as output (the actual last vertex by topological sort) and the other one is assigned a proper (neural) operation label.

RELABELING takes an input graph and relabels a randomly chosen vertex with a randomly chosen label. For both the random vertex and the random label we chose a discrete uniform distribution. Figure 12.6b exemplifies the operation by turning the label $A$ of vertex $v2$ of the source graph in fig. 12.6a into label $B$. This can be thought of as probing a new architecture candidate with one operation such as *CONV1X1* turned into *CONV3X3*. The new graph is of same order but with one different label.

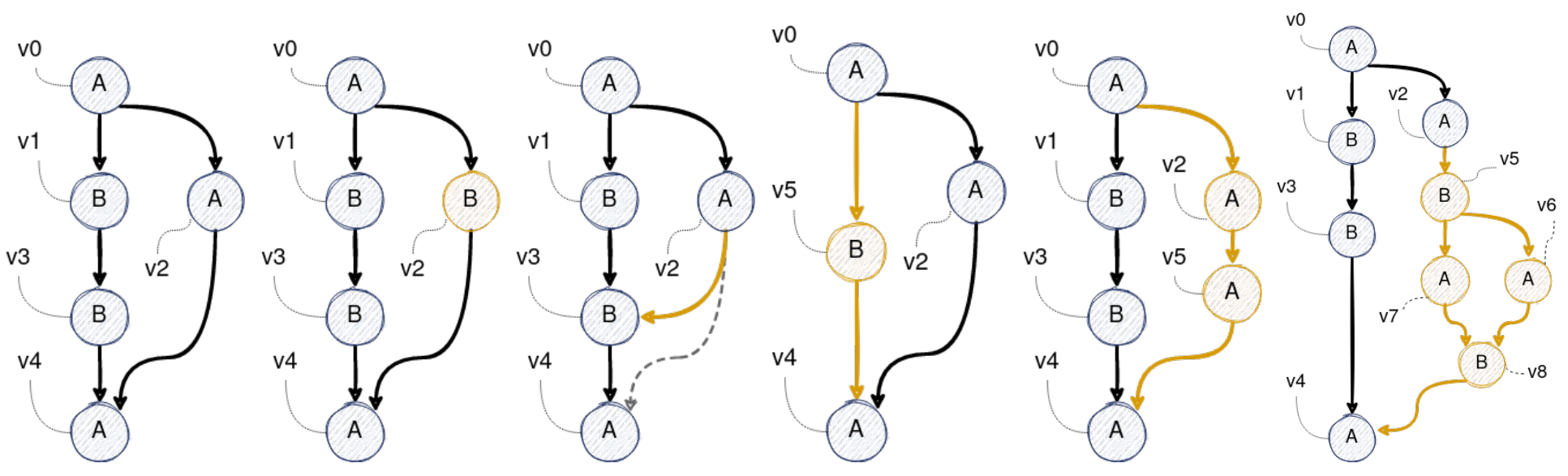

(a) Source (b) Relabeling (c) Rewiring (d) Contraction (e) Distraction (f) Subgraph Insertion

**Figure 12.6:** Examples of variation operators in graph space. Figures 12.6b to 12.6e can be understood as unary variational operations, hence mutations, while fig. 12.6f can be used as mutation (with random small sub-graphs) or as crossover operation with a second graph to be inserted into the first one.

REWIRING considers all possible, but non-existing edges from one randomly chosen vertex of the given (directed acyclic) graph and chooses randomly amongst these edges. The chosen edge is added into the graph, while the existing outgoing edge is removed. The idea of the operation depicted in fig. 12.6c is to rewire an edge to another target vertex.

An alternative form might be to randomly add an edge while removing another edge in the sense of independent random wiring. We did, however, neither explore on the relationship of familiar operations nor on properties such as runtime complexity of them.

CONTRACTION The third operation, vertex contraction, samples two neighboring vertices and contracts them into one. Figure 12.6d depicts this operation by collapsing vertex $v1$ and $v3$ of fig. 12.6a into a new vertex $v5$. A label is randomly assigned but alternative forms could use a non-uniform draw based on the two original vertices.

DISTRACTION The last considered operation, vertex distraction, considers a random vertex of the given graph and splits it into two. In the example in fig. 12.6e, vertex $v2$ is split into two vertices $v2$ and $v5$ and all outgoing edges of the former $v2$ are now outgoing edges of $v5$ while vertex $v2$ only connects directly to $v5$. The label is randomly drawn (but could also be duplicated from the existing vertex as an alternative operation).

### 12.5.2 *Results of Variation Operators on* `NAS-Bench-101`

All subsequent results of the four variation operators on the i.i.d. drawn graphs are provided in figs. 12.7 to 12.10.

Starting from one source graph $g_0 \sim UDAG(3,7)$, we obtain $n_{target} = 50$ resulting graphs $\forall j \in \{1, \dots, n_{target}\} : g_{0,j} = v_k(g_0)$. Resulting test accuracies are visualised in fig. 12.7 for *relabeling*, *rewiring*, *contraction* and *distraction* from left to right. The test accuracy $\alpha(g_0)$ is provided with a horizontal line and keep in mind that this reflects a single draw from the `NAS-Bench-101` distribution, as exemplified in fig. 12.5. A higher median of the four box plots therefore does not imply a better operation because it has to be seen relative to its source accuracy. However, the visualisation is still important as to obtain an impression of a single draw and the possibilities of neural network performances after a graph variation operator $v_k$ was applied.

While for the case of *relabeling* and *rewiring* we can see a significant relative increase, we observe a significant relative drop in test accuracy for *contraction* and *distraction*. In all cases, it can happen that the obtained graph is significantly better or worse compared to its source graph. This difference can be, however, considered as quite local in comparison to the overall distribution which also contains architectures with a test accuracy way below 0.8. Ying et al. investigated locality based on graph edit distance and measured both the "impact of replacing one [neural] operation with another one", which is related to the *relabeling* variation operator, and "a random-walk autocorrelation" [326, Sec. 3.3 Locality]. From their results they argued that "locality-based search may be a good choice for" the `NAS-Bench-101` space.

We took a further look into the operations with figs. 12.8 and 12.9 by now visualizing the accuracy distribution of $n_{sources} = 100$ source architectures and comparing it to a resulting accuracy distribution of architectures obtained after applying each variation operator $v_k$. The results of *relabelling* and *rewiring* are given in fig. 12.8, the results of *contraction* and *distraction* in fig. 12.9. The four box plots depict the distribution of test accuracies $\alpha(g_{i,j})$ from `NAS-Bench-101` of $i \in \{1, \dots, n_{sources} = 100\}$ on the left (light gray) and up to $j \in \{1, \dots, n_{target} = 20\}$ operated graphs $v_k(g_i)$ per source graph on the right (dark gray). A light dashed horizontal line shows the median test accuracy of the $n_{sources} = 100$ i.i.d. drawn source architectures, while the continuous horizontal line in light gray shows their mean accuracy. For comparison, the dark gray horizontal lines each depict the mean of architectures obtained from applying the variation operator on the source architectures.

Observe, that the *contraction* operation is the only one for which the median falls below the mean in operated accuracy in fig. 12.9 on page 235. *Contraction* is also the only operation that explicitly removes a structural element resulting usually in fewer parameters of the neural network realisation. Further, except for the *distraction* operation, the median of all original source accuracies are above the ones obtained through an operation. Therefore, there exist operations such as *distraction* that also seem to push in the direction of better solutions, although it could be explained by more parameters. The other three operations serve more as a tool for exploration while selection remains the main source of force towards optimal solutions in genetic algorithms with such variation operators.

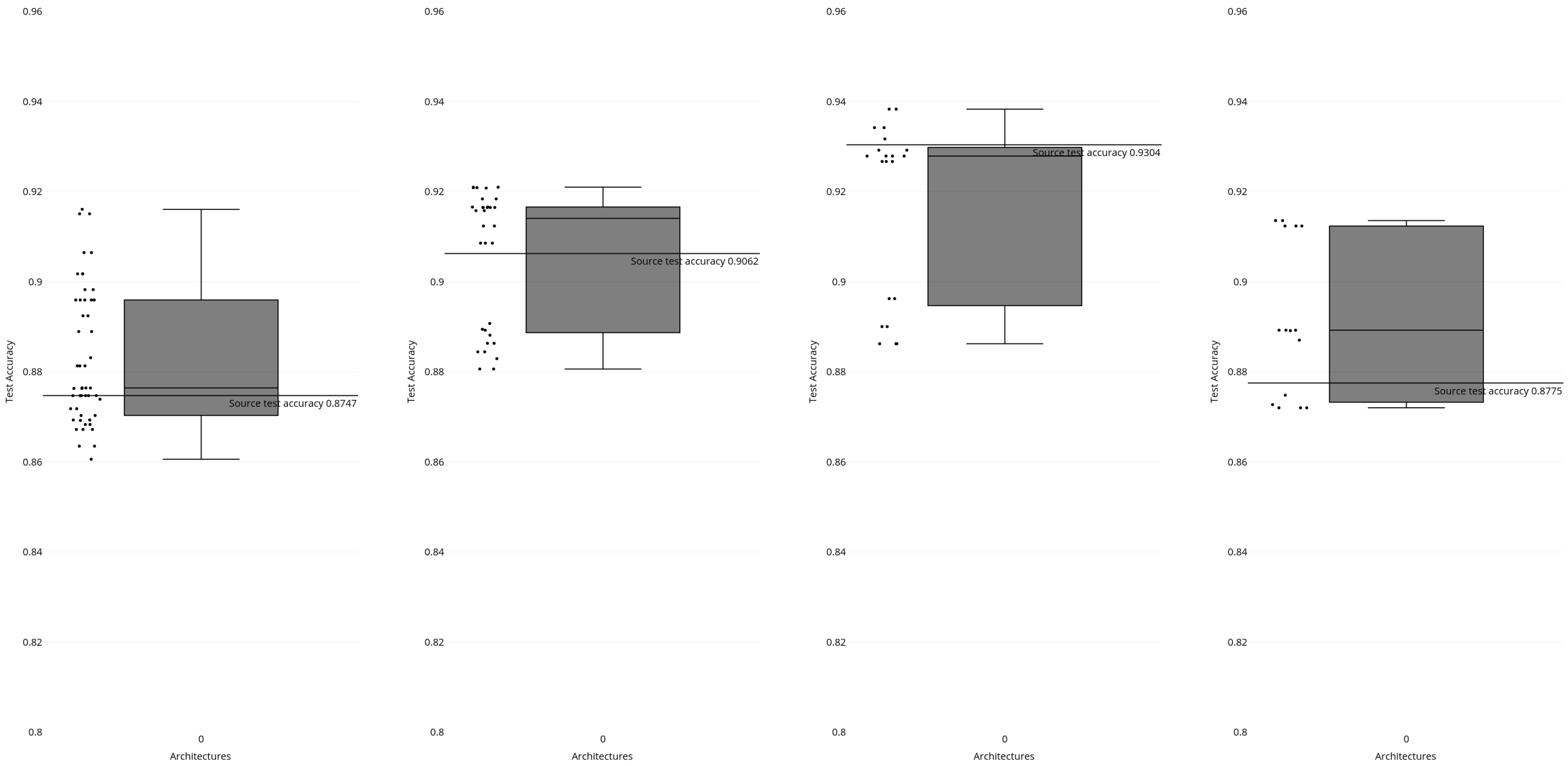

**Figure 12.7:** A single draw $g_0$ with its test accuracy $\alpha(g_0)$ depicted in a horizontal line and up to $n_{target} = 50$ graphs that are obtained by independent applications of $v_k(g_0)$. Note, that test accuracies `NAS-Bench-101` "is above 90% for a majority of models" and "the best architecture [..] achieved a mean test accuracy of 94.32%" [326, p. 3.1]. From left to right: *relabel*, *rewire*, *contract*, *distract*.

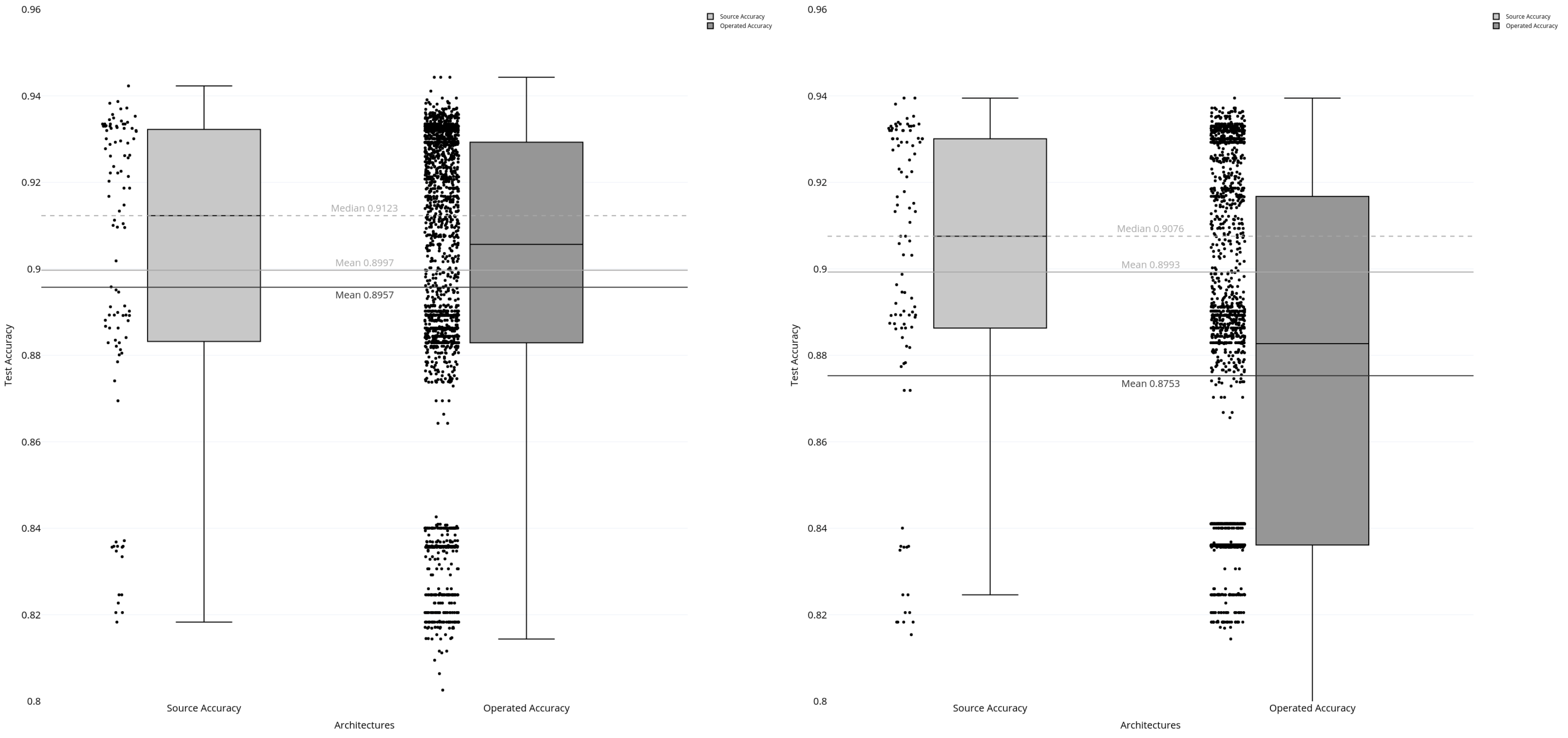

**Figure 12.8:** Operations: **relabeling** and **rewiring**.
Shown are test accuracies $\alpha(g_{i,j})$ from NAS-Bench-101 of $i \in \{1, \dots, n_{sources} = 100\}$ randomly sampled source graphs plus up to $j \in \{1, \dots, n_{target} = 20\}$ operated graphs $v_k(g_i)$ per source graph. The resulting number of samples are therefore upper bounded by 2000 per variation operator.

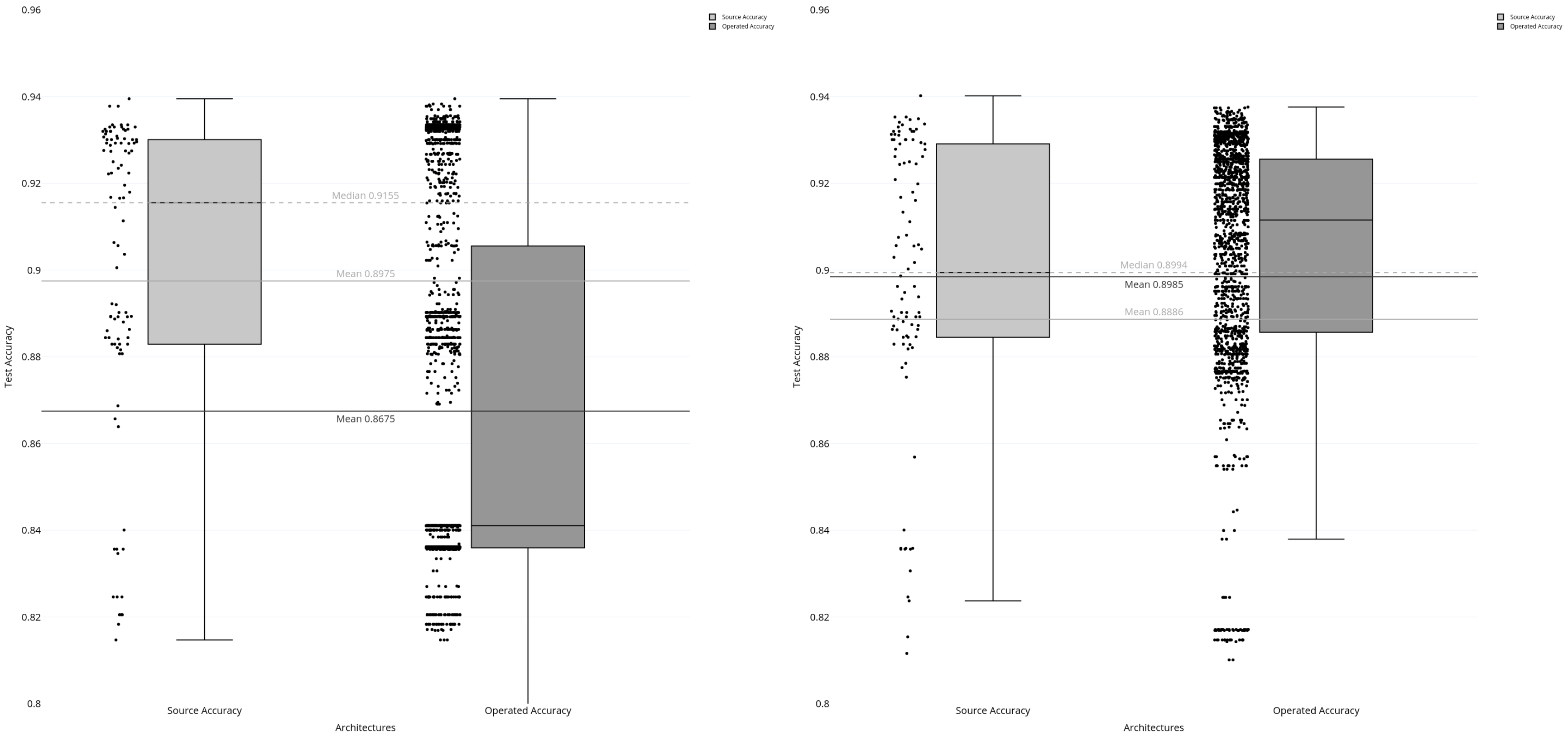

**Figure 12.9:** Operations: **contraction** and **distraction**.
Shown are test accuracies $\alpha(g_{i,j})$ from NAS-Bench-101 of $i \in \{1, \dots, n_{sources} = 100\}$ randomly sampled source graphs plus up to $j \in \{1, \dots, n_{target} = 20\}$ operated graphs $v_k(g_i)$ per source graph. The resulting number of samples are therefore upper bounded by 2000 per variation operator.

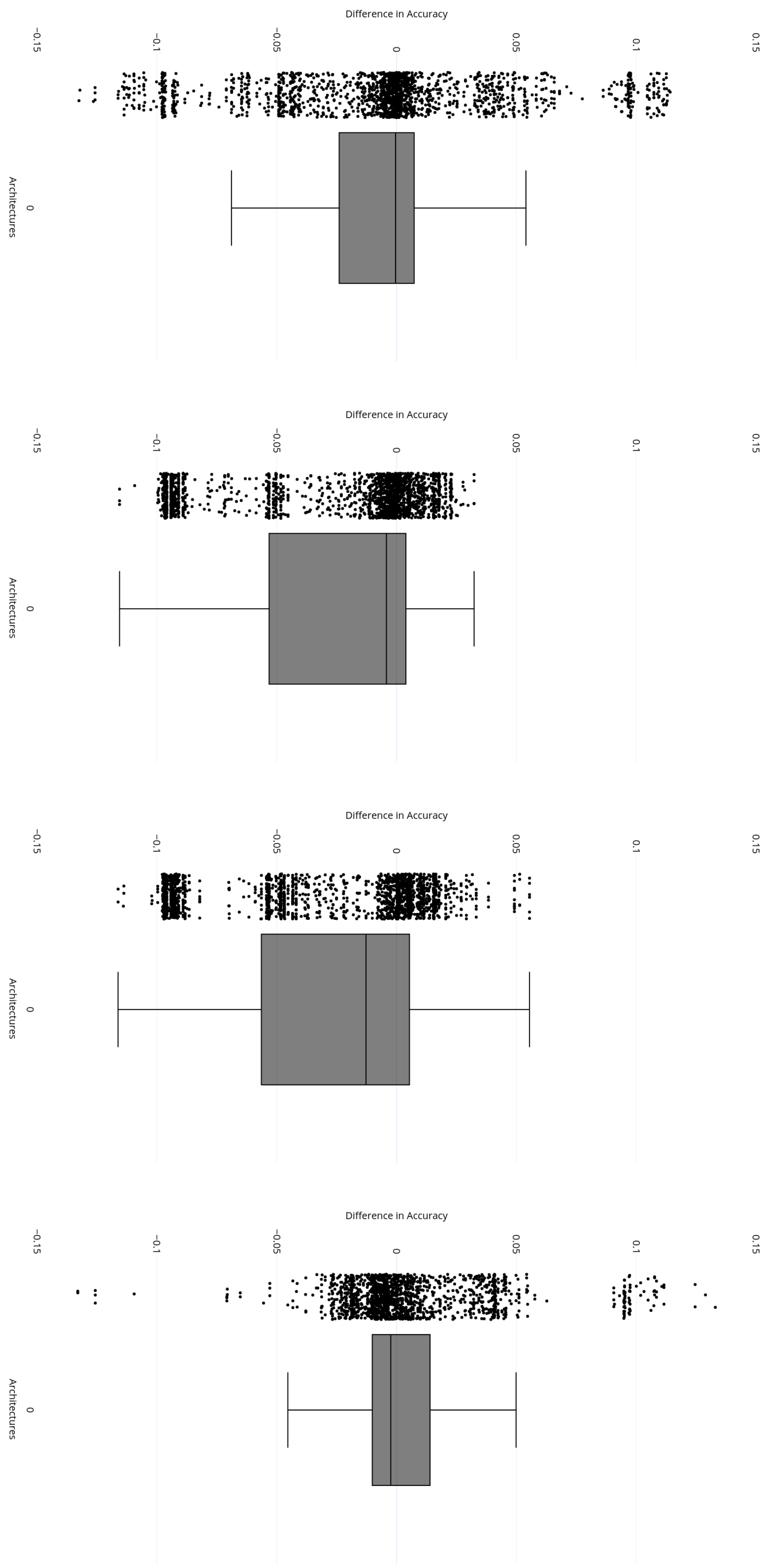

**Figure 12.10:** To better assess the overall performance of each variation operator, we looked into the distribution of differences $\Delta\alpha_{i,j} = \alpha(g_{i,j}) - \alpha(g_i)$. From left to right: *relabel*, *rewire*, *contract*, *distract*.

Further observe, how a single source test accuracy with repeated operations in fig. 12.7 can tell a different story about the behaviour of an operation compared to many source test accuracies as used in figs. 12.8 and 12.9. The aggregated results in figs. 12.8 and 12.9 suggest that operations which employ growth are beneficial for increasing the resulting performance. However, this would suggest that starting with a very large initial architecture could directly provide good solutions. Both, pruning and genetic NAS, however, provide many cases which highlight that exploration and not only size is necessary for solutions that provide good performance and simultaneously other properties such as small memory footprints. The question therefore remains not whether to grow, to prune or to vary neural operations, but where these operational directions in the search space trade off for beneficial multiple objectives.

Figure 12.10 on the preceding page further provides the distribution of differences in accuracies and provides an estimation of the expected change in performance per variation operator. The difference for each target graph $g_{i,j} = v_k(g_i)$ is computed and denoted as $\Delta\alpha_{i,j} = \alpha(g_{i,j}) - \alpha(g_i)$.

*Relabeling* and *distraction* seem to work more local in comparison to *rewiring* and *contraction* as their overall distributions in accuracy differences are significantly smaller. An interpretation could be, that an already good neural network which is relabelled or distracted is very likely to stay pretty performant while if it is rewired or vertices are even contracted, it is more likely to drop in performance. The two operations *relabeling* and *distraction* seem to be more local - something which we did not observe with the previous visual evaluations. However, the observation that *contraction* likely also drops in performance aligns with the previous interpretation – fewer structural elements and therefore fewer parameters usually is not beneficial. Although, the maximum gain in performance for *contraction* is above 0.05 and on par with the other two distributions for *relabeling* and *distraction*. *Rewiring*, on the other hand, does not yield as strong jumps as the other three operations. The amount of possible connections is comparatively high and indicates that it is usually difficult to find a good structural wiring. But it could be also argued that, after such a visual analysis, the *rewiring* operation is not as suited for a genetic NAS as the other opations if exploration is not as favorable. We assume, however, that a good mix of all broader types of variation operators are beneficial as to be able to actually find all candidates in the search space: growing and shrinking vertices and edges of the graph and interchanging labels attached to it.

### 12.5.3 *Summary of Variation Operators on* `NAS-Bench-101`

Variation operators sample new candidates from a search space based on existing candidates in a population of the current generation. While statistical differences between the analysed operations can be observed, the experiments on `NAS-Bench-101` suggest also that nearly any operation can be chosen as long as novelty or exploration is fostered.

There also appears to be a bias towards growth of architectures for which it needs to be recalled from the growth behaviour of graphs in section 3.4, that the number of potential candidates is also growing for larger graphs. Recall, that it is difficult to sample uniformly from this growing space of graphs to de-bias this effect properly. Also, `NAS-Bench-101` is limited to very small graph orders and also other benchmarks such as `graphs10k` [272] show us that the performance distribution across the sample space is very sparse in most parts and very dense in others. Therefore, growing towards larger architectures is an obvious direction in the search space.

Variation operators are difficult to analyse and in the presented case also difficult to distinguish. This suggests to learn graph generative models as replacement sampling strategy for variation operators. It also suggests for the idea of learning operations in graph space in an end-to-end manner as to optimise the overall neural architecture search instead of caring about manually engineered variation operations.

REMARKS

- Distributions of resulting accuracies when variation operators are applied show that the operations provide local explorative properties, i.e. the maximum jumps in performance change are $\pm 0.05$.
- Some variation operators such as *relabeling* or *distraction* are more local than *rewiring*.
- Based on experimental results, it appears very likely that variation operators can be quite easily exchanged and the true power of genetic algorithms still stems from the pressure of selection from generation to generation.
- The variety of possible variation operators and the trend of automation invite to investigate on learned variation operators in graph embedding space instead of analysing them manually on a per application case basis.

## 12.6 DIFFERENTIABLE ARCHITECTURE SEARCH

During the last decade, an alternative to previously mentioned evolutionary approaches to neural architecture search emerged, called differentiable architecture search [167, 176]. Differentiable architecture search (DARTS) employs a so called hyper-architecture which contains sub-paths for which discrete choices can be made. This means, that different sub-architectures are contained within the hyper-architecture. Because the hyper-architecture is a finite-paramterised neural network, the discretisable sub-paths are also finitely many. DARTS learns both the shared neural network weights and the architectural parameters in a bi-level differentiable fashion.

DARTS as the abbreviation for differentiable architecture search became the general term for methods in which neural architectures are learned with the idea of a differentiable hyper-architecture, although the literature also speaks of one-shot architecture search and the hyper-architecture is then referred to as the one-shot model [333]. We stick, however, to the terms DARTS and hyper-architecture here as we deem them more appropriate to carry the meaning for the underlying idea.

Architectural parameters of the hyper-architecture control which sub-paths are chosen later to e.g. be part of a single architecture resulting from the neural architecture search. This is often realised by using a softmax $\sigma^{\Psi}$ activation function or by sampling from Gumbel-Softmax together with a reparameterisation trick (see section 5.2.4). A very simple example is to think of two possible paths to choose from and the logits of architectural parameters could look like $[3.5, 1.2]$ – which, when passed through $\sigma^{\Psi}$, become $[0.9089, 0.0911]$. The activations $[0.9089, 0.0911]$ then reflect the strength or importance of the path and are usually used as multiplicative factors for the inferencing results of the neural network at this parting of the ways.

The same principle with different paths is nowadays used for so called Mixture of Expert (MoE) models for Large Language Models (LLMs), or to achieve modularity or learned routing in neural networks in general [85, 130, 239]. An advantage is that, while all of the experts are trained jointly, during inference only sub-paths or experts in the sense of joint sub-modules of a larger neural network are used to compute an output. This allows for parallelisation, specialisation, distribution and scalability of large models.

To learn the architectural parameters in fig. 12.11, Liu et al. sketch a bi-level optimisation problem in which gradient-based learning is used for both the learning of weights & biases and the architectural parameters on the second level. This idea was already investigated by Bengio [24], Maclaurin et al. [180], Pedregosa [220], and more [144] but in light of NAS popularised through DARTS [167].

Bengio notes that “hyperparameters are usually chosen by trial and error with a model selection criterion” [24] and also Larsen mentioned that the "standard approach for estimation of regularisation parameters [..] is by more and less systematic search and evaluation of the validation set error" [144, p. 66]. Back then, "architecture selection schemes [were] based on either network pruning or network growth" [144, p. 66]. Using gradient-based learning for second- or multi-level optimisation is therefore not new, but nonetheless difficult. For DNNs it applies that the "training criterion is not quadratic in terms of the parameters" such that "it will in general be necessary to apply an iterative numerical optimisation algorithm to minimise the training criterion" [24]. What is new is that the focus with NAS shifted from other hyperparameters such as weight decay for regularisation [144, p. 69] to new forms of architectural parameters.

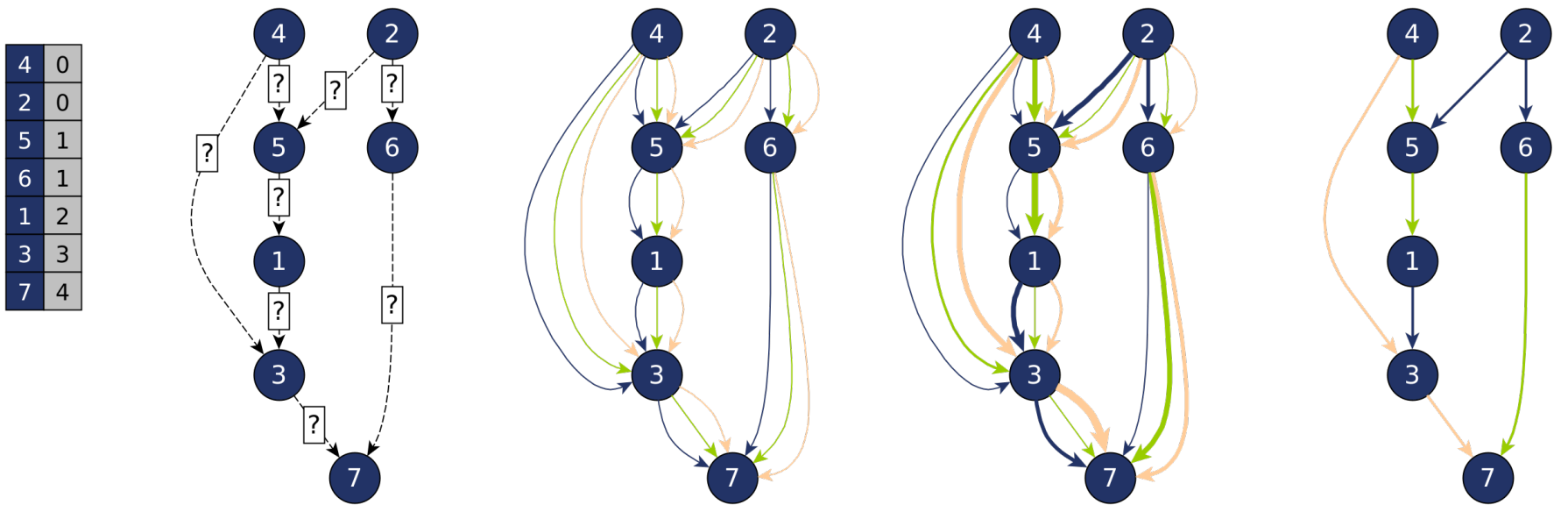

**Figure 12.11:** A layered DAG as used in fig. 6.1 visualises the idea of DARTS as originally presented by Liu et al. [167, Fig. 1]. Here, a choice for the operations on the edges of the graph has to be made (left) and are initially unknown. Three possibilities are available (second from left), each operation has its own lightness. Making these choices explicit is the "continuous relaxation of the search space" [167, Fig. 1]. Learning the architectural parameters for choosing an operation for each edge is done in a bi-level optimisation and results in a weighting which is depicted as line widths (second from right). The final architecture is defined through a discrete choice based on the architectural parameters (right).

Liu et al. sketch the bi-level optimisation problem as the following [167, Eq. 3 & 4]:

$$\begin{aligned} \min_{\alpha} \ & \mathcal{L}_{val}(w^*(\alpha), \alpha) \\ \text{s.t.} \ & w^*(\alpha) = \arg\min_{w} \ \mathcal{L}_{train}(w, \alpha) \end{aligned} \tag{12.1}$$

for which they propose approximation techniques for the gradient of the architectural parameters $\alpha$.

Observe the familiarity to the problem formulation in eq. (7.2) for neural architecture search of graph-induced neural networks:

$$\arg\min_{T \in \mathbb{S}} \mathcal{L}_{val}(T, \arg\min_{f \in \mathbb{A}([T])^{d_1 \to d_2}} \mathcal{L}_{train}(f, D_{train}))$$

NAS can be understood as a kind of hyperparameter optimisation if the architecture can be formulated as a parameter of the overall model. DARTS makes some restrictions to the optimisation problem: the search space $\mathbb{S}$ consists of finitely many continuous parameters instead of not specifying it closer or having discrete candidates or infinitely many. While genetic variation operations seek topological relationships between candidates, for continuous architectural parameters they are given in a form that gradient-based learning can be employed.

Another advantage is, that DARTS learns both parameters of a neural network realisation and the architectural parameters $\alpha$ such that parameters are shared across different architectures. This sharing of common information is referred to as *Lamarckian evolution* [60] in context of evolutionary NAS. DARTS therefore combines weight sharing between genotypical and phenotypical representations and gradient-based learning in tackling the problem of eq. (7.2).

## 12.7 summary on methods for nas

Chapter 12 gave an introduction of selected neural architecture search methods and put our contributions into its context:

- NAS-methods can be captured in the multi-level optimisation problems based on graph-induced neural networks as presented in section 7.1,
- benchmarks as presented with `graphs10k` [272] contribute in running repeatable experiments to understand structural properties of neural architectures or how NAS-methods relate to eachother,
- and benchmarks such as `NAS-Bench-101` enable us to investigate on the meaning of variation operators as conducted in section 12.5.

NAS-methods provide solutions for problems as formulated in eq. (7.2) in section 7.1. The methods can be compared under equivalent computation budgets of $\mathcal{L}_{val}$ or formally based on their complexity. Each method requires to navigate the search space in an effective and efficient way. Taking shortcuts by faster performance estimations can be applied to all methods. Many methods such as evolutionary approaches can incorporate alternative sampling strategies, not only by genotypical variation operators but also by e.g. sampling from learned graph generative models. Differentiable architecture search can not directly incorporate new sampling strategies because of its hyper-architecture and finite space but gives rise to new ideas on faster performance estimations.

Alternative NAS-methods include (but are not restricted to):

- Bayesian Optimisation [124, 178, 337]
- Covariance Matrix Adaptation Evolutionary Searches (CMA-ES) [32, 260]
- Reinforcement Learning [38, 114, 339]

The following part vi concentrates on new challenges arising when improving established NAS-methods such as evolutionary approaches with predictive and generative models, exploiting hidden properties of the search space and its candidates.

Part VI

# ADVANCED METHODS FOR NEURAL ARCHITECTURE SEARCH

Automatically designing neural networks has a long tradition since before the 1990s and while several ideas keep re-ocurring in different ways, advanced methods beyond generic search approaches integrate new predictive and generative methods for guiding neural design. How does structure fit into this picture of automated neural design and what novel methods can be proposed that considers structure of neural networks?

**Research Complex III: Automation**

From section 1.1.6:

- Which methods for neural architecture search exist, how can they be compared and what are their differences?
- With knowledge on structure from research complex II, how can neural architecture search methods be improved or guided?

This part is concerned around research complex III which we initially posed in section 1.1.6 on *automating neural networks and neural structure*. Our main research question of complex III is:

**⋄ ⋄ ⋄ Research Complex III**

With knowledge on structure from research complex II, how can neural architecture search methods be improved or guided?

And the gist of our answer to it is:

**How can NAS methods be improved or guided?**

- By faster, less expensive performance estimations through e.g. predictive models based on structure,
- By qualitative better sampling with e.g. generative models that leverage topological properties of the search space and thus requires less expensive performance estimations,
- By constructing better search space designs with useful topological properties or a good balance between search space dimensionality and size and expressiveness,
- And formalism and theory can be improved to better guide NAS, e.g. by finding suitable reasons for changing between architectural spaces through growing, pruning or other dynamic optimisation procedures.

Entrance points for advancing methods for neural architecture search (or more generally neural design) can be summarised in one big picture and take place on different levels of a multi-level optimisation problem for finding neural network realisations – compare eq. (7.3) on page 123 on search space design:

$$\arg\min_{\mathbb{S}\in\mathcal{P}(DAG)} \arg\min_{T\in\mathbb{S}} \mathcal{L}_{val}(T, \arg\min_{f\in\mathcal{A}([T])^{d_1\to d_2}} \mathcal{L}_{train}(f, D_{train})) \quad (7.3)$$

A search space $\mathbb{S}$, whether it will be based on a formulation of graph-induced neural networks or not, can be thought of as a discrete or continuous space in which candidate architectures are to be found. We build up on the evolutionary search visualisations in figs. 12.2 to 12.4 on pages 224–226 and illustrate in fig. 12.12 how two major aspects, **predictive** and **generative** modeling engage in the overall optimisation process of eq. (7.3) on page 123:

Predicting an evaluation measure for a (previously unknown) candidate architecture can be thought of as running a complex training pipeline including data, following a performance estimation strategy. Such a performance estimation, depicted as $f_{1,\alpha}$ on the left in fig. 12.12 on the facing page as to obtain e.g. an accuracy estimate for an architecture is usually computationally expensive. An alternative is to employ faster estimation strategies, as depicted with $f_{2,\alpha}$. This can be achieved by

- small training schemes which use less data or less expensive processing pipelines,
- employing surrogate estimations by e.g. learning predictive models only based on architectural information,
- or combining both ideas by obtaining performance estimates of groups of architecture candidates with architecture and reduced data information.

Figure 12.12 on the next page depicts the idea of determining a proper performance estimate during an overall neural architecture search with different strategies in the left box.

Coming up with a new sample in the search space during a NAS can range from randomly sampling over mutating existing candidates (ES) to sampling from a multi-variate normal (CMA-ES) or obtaining a sample from a gaussian process surrogate. These are all forms of generative modeling as depicted on the right in fig. 12.12 on the facing page.

The sampling strategy $g_{1,\alpha}$ can be thought of as a fast method based on locality properties in the search space, i.e. a cross-over strategy in an evolutionary genetic search. Alternatively, $g_{2,\alpha}$ could be a graph generative model that was trained on only high-performing candidates. Graph Neural Networks [28] can provide a bridge between rich discrete search spaces but worthy generative modeling.

We already observed in section 12.5 on page 228, how much the results of a variation operation depend on the search space and need to be questioned in comparison to more naïve methods such as random sampling. A tradeoff between complexity of generative surrogate models and their capability to support the search process has still to be investigated on.

Advancing methods in neural architecture search with generative models $g_{2,\alpha}$ include

- the study of low-dimensional or differentiable representations of architecture candidates to employ strong optimisers that rely on continuity or differentiability, e.g. CMA-ES or Bayesian Optimisation,

- integrating graph neural networks into evolutionary searches over discrete architecture representations such as graph structures as to learn variation operations in a differential (neural) manner,
- or to employ recent graph generative models to guide evolutionary searches with sophisticated samples from learned distributions of graphs.

The following chapter 13 on page 249 is concerned with predictive advancements for NAS, while the subsequent chapter 14 on page 261 is concerned with generative advancements for NAS.

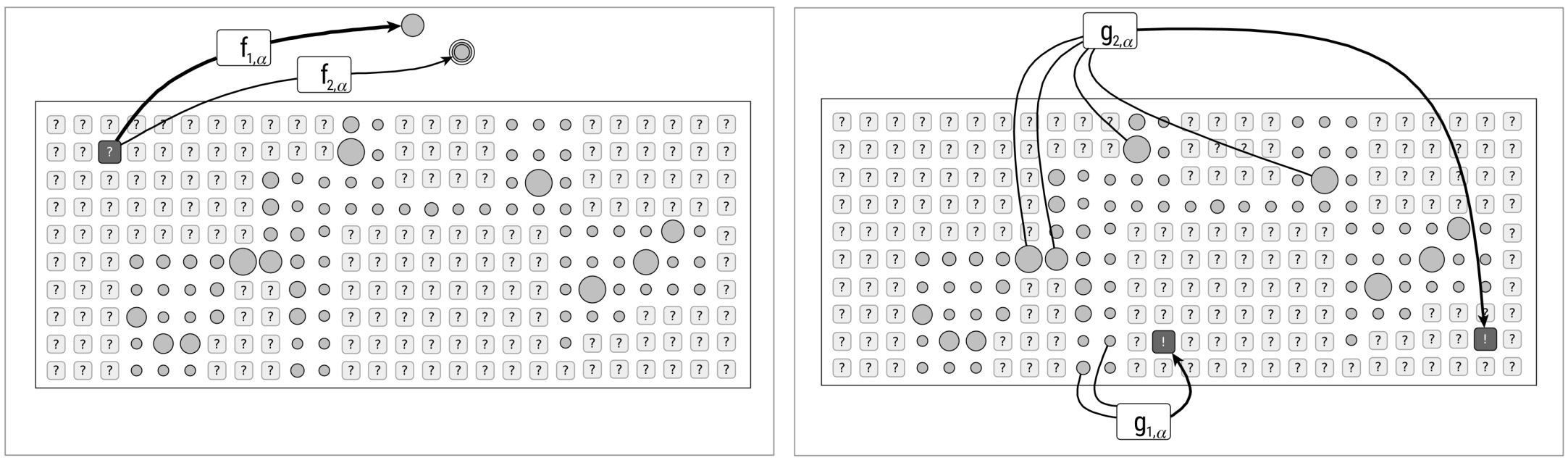

**Figure 12.12:** Dark-gray boxes depict finite or infinite search spaces with candidates, that are either unevaluated ("?") or assigned to an estimated performance score. Finding proper search space definitions influence through their topological properties, such as containing continuous or discrete candidates, choices of subsequent search methods. Obvious advances in neural architecture searches can be made by improving predictive methods (left) to quickly obtain good performance estimates or by generative models (right) to decide on how to come up with new candidates in the search space:
→ Predictive models $f_{1,\alpha}$ or $f_{2,\alpha}$ seek to give an estimate of the performance of a candidate, either by full evaluation on a hold-out dataset or by approximations.
→ Generative models $g_{1,\alpha}$ or $g_{2,\alpha}$ seek to sample new candidates that are promising.

# 13 PREDICTIVE MODELS FOR NEURAL ARCHITECTURE SEARCH

*The following entails:*

Performance estimation in neural architecture search is an expensive component of the overall search. When conducting a training scheme with the best hyperparameter settings knowledgable, the training dynamics are highly non-linear and have no guarantees for convergence to a proper minimum. Even the existance of an optimisation minimum yielding good generalisation scores and serving as performance estimation is not guaranteed. Often it is even unknown whether repeated very expensive training schemes yield stable or reproducible performance estimations. **Predicting** the performance as alternative to expensive evaluation schemes at least promises to short-cut performance estimations.

In practical NAS applications, it became common to employ

- training schemes with varying computational costs for quick and full evaluations,
- learning curve predictions [13],
- structural properties for performance prediction [6, 269, 272, 300],
- and other performance prediction tricks such as FaDE [205]

to circumvent the computational (or other technical) costs of performance estimation.

Our contribution encompasses **1st/** experimental results with structural performance predictions and **2nd/** the idea and methodological description of using FaDE-scores as surrogate for performance scores.

We used structural properties for performance prediction before [272]. The results in [272] have been reproduced and confirmed by Holze et al. [104] and partially by Ben Amor et al. [6]. The gist insight from these studies is that some few noticable structural properties – which are not directly related to the number of parameters of a model – can make up for solid predictions of performances. Some of these notable properties have been the variance of path lengths, variance of eccentricities and variance of the degree distribution of the underlying used graphs. While we interprete the overall influence of single graph properties as not sufficient for the sole explainability of differences between structural priors, the influence of these three properties fostered further research and aligns with theoretical observations. All three properties are concerned with the fact that varying long and short paths and diverse branching across non-linearities suggest to be beneficial for generalisation properties.

Integrating graph structures into a differentiable hyper-architecture in the spirit of DARTS [167, 176], further led us methodologically to derive locally valid FaDE-scores from the hyper-architecture as a surrogate for performance estimations [205]. We investigated on neural architecture searches in which computationally cheaper scores are used to guide a graph-focussed NAS. A search strategy in an open-ended search space, mixing different training schemes, might be a fruitful advancement of this method.

## 13.1 STRUCTURAL PERFORMANCE PREDICTION

We call a prediction of model performance estimations based on only model *structure* but no data a structural performance prediction. The targeted performance metric can e.g. be the loss, the test accuracy or even the robustness as outlined in table 9.1. Structural information can be represented with vectorised features such as the number of vertices or the density or in a more structured form as a graph or a tree from a grammar.

In section 9.5, we conducted a correlation analysis between the preditctions of instances of three different regressor types and expected test accuracies. The regressor types are ordinary least squares linear regression learn, support vector regression learn and random forests learn. The intention of section 9.5 is to analyse and understand the influence of structural properties on how DNNs behave. For structural performance prediction, we reduce the expectation of explainability and shift our interest to improving a NAS method only with a good predictive model. The central question of research complex II is:

| Feature | Feature Importance of RFs | | | Feature | Feature Importance of RFs | | |
|---|---|---|---|---|---|---|---|
| | Mean±Std | Max | Min | | Mean±Std | Max | Min |
| graph_num_vertices | 0.009±0.0055 | 0.0205 | 0.0019 | undir_ecc_median | 0.0005±0.0003 | 0.001 | 0.0001 |
| graph_num_edges | 0.0189±0.0092 | 0.0368 | 0.0064 | undir_ecc_mean | 0.0101±0.003 | 0.0144 | 0.006 |
| num_sinks | 0.003±0.0025 | 0.0088 | 0.0004 | undir_ecc_var | 0.0143±0.0101 | 0.0385 | 0.0034 |
| num_hidden | 0.0342±0.0181 | 0.0702 | 0.0104 | shortestpaths_max | 0.0044±0.0032 | 0.0109 | 0.0007 |
| num_layers | 0.0042±0.002 | 0.0086 | 0.0011 | shortestpaths_min | 0.0±0.0 | 0.0 | 0.0 |
| num_paths | 0.034±0.0165 | 0.0735 | 0.0171 | shortestpaths_median | 0.0001±0.0003 | 0.0012 | 0.0 |
| density | 0.0345±0.0175 | 0.0795 | 0.0127 | shortestpaths_mean | 0.0207±0.0098 | 0.0395 | 0.0078 |
| degree_max | 0.0259±0.0101 | 0.0431 | 0.0151 | shortestpaths_var | 0.0327±0.0188 | 0.067 | 0.0103 |
| degree_min | 0.0128±0.0085 | 0.0312 | 0.0065 | layersize_max | **0.2481**±0.0794 | **0.4301** | 0.1512 |
| degree_median | 0.0056±0.0023 | 0.0088 | 0.0022 | layersize_min | 0.0006±0.0006 | 0.002 | 0.0 |
| degree_mean | 0.0229±0.0123 | 0.0502 | 0.0098 | layersize_median | 0.0113±0.0072 | 0.0275 | 0.002 |
| degree_var | **0.0855**±0.0535 | **0.1638** | 0.0137 | layersize_mean | 0.0153±0.0065 | 0.0288 | 0.0076 |
| undir_ecc_max | 0.002±0.0013 | 0.0051 | 0.0006 | layersize_var | **0.3494**±0.0744 | **0.4736** | 0.2225 |
| undir_ecc_min | 0.0001±0.0001 | 0.0004 | 0.0 | | | | |

**Table 13.1:** This table provides aggregated feature importance scores of $n = 10$ random forests which were trained on 70% of structural properties of CTs and their energy consumption as presented in section 9.7 on page 174. Each repetition used a different random state for the train-test split of 70/30 and the same state was used to construct a new RF model. The analysis shows us that the layer size maximum and the variance of layer sizes of a model are significant predictors for energy consumption. Similar to other experiments, the degree variance shows to be a predictive property.

**How can NAS methods be improved or guided?**

Every NAS method needs to evaluate architectures of their search space. Reliably point estimates which are quickly obtained can help probing many architectures in both the initial and a late phase of the search. But also less reliable estimates with distributional estimates can guide methods as we observed in section 10.1 with different types of estimation complexity changes of search spaces.

In table 9.3 on page 173 we provided the coefficient of determination $R^2$ based on predictions on a test set of evaluated models. We manually split the overall set of available features $\Omega$ for the regressors into subsets $\Omega_{np}$, $\Omega_{op}$, $\Omega_{var}$, $\Omega_{small}$, and $\Omega_{min}$ to understand the effect of having no direct information on the number of parameters of models available. The predictions were surprisingly strong with $R^2$ scores of above $0.9314 \pm 0.0031$ for a RF regressor with no direct information about the number of parameters of the model. However, as noted in [272], it needs to be considered that the distribution of test accuracy of models is highly skewed and contains major bumps around very low performing models and very high performing models.

In table 13.1, we provide feature importance scores of ten random forests which have been trained on different train-test splits of the energy consumption data used in the analysis in section 9.7 on page 174. The coefficient of determination $R^2$ over these ten independent RFs yielded decent predictive capabilities on the energy consumption of $\varnothing(R^2) = 0.4806 \pm 0.1638$.

Because of these results, NAS methods can be guided by structural performance predictions as long as a sufficiently large set of evaluated models can be build up and the performance estimation scheme can be exchanged. This applies for most population-based NAS methods such as evolutionary searches. Conversely in DARTS it can not be directly used.

Structural performance prediction then advanced in two ways: **1st/** new experimental results were added e.g. on `NAS-Bench-101` and other NAS benchmarks and **2nd/** instead of manually feature-engineering performance predictors, graph neural networks have been employed to automatically learn performance predictors on structural representations such as graphs [207, 300]. For example, Ning et al. report that their "GATES-powered predictor-based NAS is 511.0× and 59.25× more sample efficient than random search and regularized evolution" [207, Fig. 5, p.13] and Wendlinger et al. report a Kendall ranking coefficient of "$\tau_k = 0.9198 \pm 0.0023$" for the predicted vs. actual test accuracy with a "mean and standard deviation across five trials with different dataset splits" [300, Sec. 5.1, p.8]. Both works use graph neural networks for automatic feature engineering of graph properties to conduct structural performance prediction.

## 13.2 FAST DIFFERENTIABLE ESTIMATION OF SCORES FOR OPEN-ENDED NAS

Performance prediction or learning curve extrapolation are two popular examples for accelerating neural architecture searches by avoiding costly evaluations of each potential architectural candidate. We investigated on a new alternative approach based on predicting relative scores obtained from a DARTS-inspired differentiable hyper-architecture [204, 205]. As of our knowledge this is the first work using a differentiable hyper-architecture and transferring information from its finite architectural search space into an open-ended NAS.

FaDE (**F**ast **D**ifferentiable **E**stimation) [205] is a method consisting of two phases:

1. The score estimation phase includes the construction, training and evaluation of a hyper-architecture in which cells are stacked into multiple stages and after optimizing the bi-level problem of DARTS from eq. (12.1) on page 240, a FaDE-score is attributed to individual cell candidates.
2. An optimisation step phase in which FaDE-scores are used to conduct a pseudo-gradient step in an open NAS search space.

We describe FaDE with a focus on a graph-based search space as this is the focus of this work and leave the more general idea of it to [205]. The core idea is the first phase of score estimation is, that a finite set (a window) of search space candidates are integrated into a chained DARTS-like one-shot model for which architecture parameters are learned. These architecture parameters are aggregated into an attribution score for a single search space candidate and subsequently used in a NAS.

### 13.2.1 *The Score-Estimation Phase*

Consider a (large) search space of structural themes $\mathbb{S} = \{T_1, T_2, \dots\}$ and a fixed number of stages $n_{stages} \in \mathbb{N}$ such that for a theme $T \in \mathbb{S}$, a realisation $f_T \in \mathcal{A}([T])$ can be structurally broken into $f_{T,1} \circ f_{T,2} \circ \cdots \circ f_{T,n_{stages}}$. Further consider an evaluation window of themes $W \subset \mathbb{S}$ of finite size $w = |W|$. We construct a hyper-architecture $H(W) : \mathbb{R}^{d_{in}} \to \mathbb{R}^{d_{out}}$ containing in total $w \cdot n_{stages}$ neural network realisations from $W$. For each $T \in W$ we have a realisation $f_{T,1} \circ f_{T,2} \circ \cdots \circ f_{T,n_{stages}}$ for each stage such that the hyper-architecture has a grid-layout as in fig. 13.1. Further, architectural parameters $\alpha_{H(W)} \in (0, \infty)^{w \times n_{stages}}$ are part of the hyper-architecture $H(W)$.

The input $\mathbf{x} \in \mathbb{R}^{d_{in}}$ to $H(W)$ is fed to the realisations in the first stage. An intermediate neural network layer such as a CNN might be used to reduce the dimensionality of $\mathbf{x}$. Resulting output tensors are of additional dimension $w$ and a tensor $\mathbf{z}_1 = \{f_{T,1}(\mathbf{x}) \mid T \in W\}$ [1] is obtained. This tensor is the intermediate result of the first stage and will be weighted by the architectural parameters of the stage through

$$\hat{\mathbf{z}}_1 = \mathbf{z}_1 \odot \alpha_{H(W),1}$$

in which $\odot$ refers to a point-wise product along the window-stacked dimension of size $w$.

The outputs $\hat{\mathbf{z}}_k$ per stage $k \in \{1, \dots, n_{stages}\}$ are pooled and for stages $k \in \{1, \dots, n_{stages} - 1\}$ fed through another convolution aggregation. This pooling and transformation is conducted to ensure that the resulting tensor $\hat{\hat{\mathbf{z}}}_k$ has proper shape to serve as input to neural network realisations $f_{T,k+1}$ of the subsequent stage.

Final output of the hyper-architecture $\hat{\hat{\mathbf{z}}}_{n_{stages}}$ is additionally fed through another linear layer as to transform the last latent representation into the desired output dimensionality $d_{out}$ of $H(W)$.

In summary, for a given window of themes $W \subset \mathbb{S}$ we obtain a hyper-architecture $H(W) : \mathbb{R}^{d_{in}} \to \mathbb{R}^{d_{out}}$ constructed with stage-wise neural network realisations of themes in $W$. The hyper-architecture carries stage-wise architecture parameters $\alpha_{H(W)} \in (0, \infty)^{w \times n_{stages}}$ which we

1 The particular tensor shape of $\mathbf{z}_1$ is dependent on the chosen (common) dimensionalities of the realisations $f_T$ and do not necessarily need to align with $\mathcal{A}([T])^{d_{in} \to d_{out}}$.

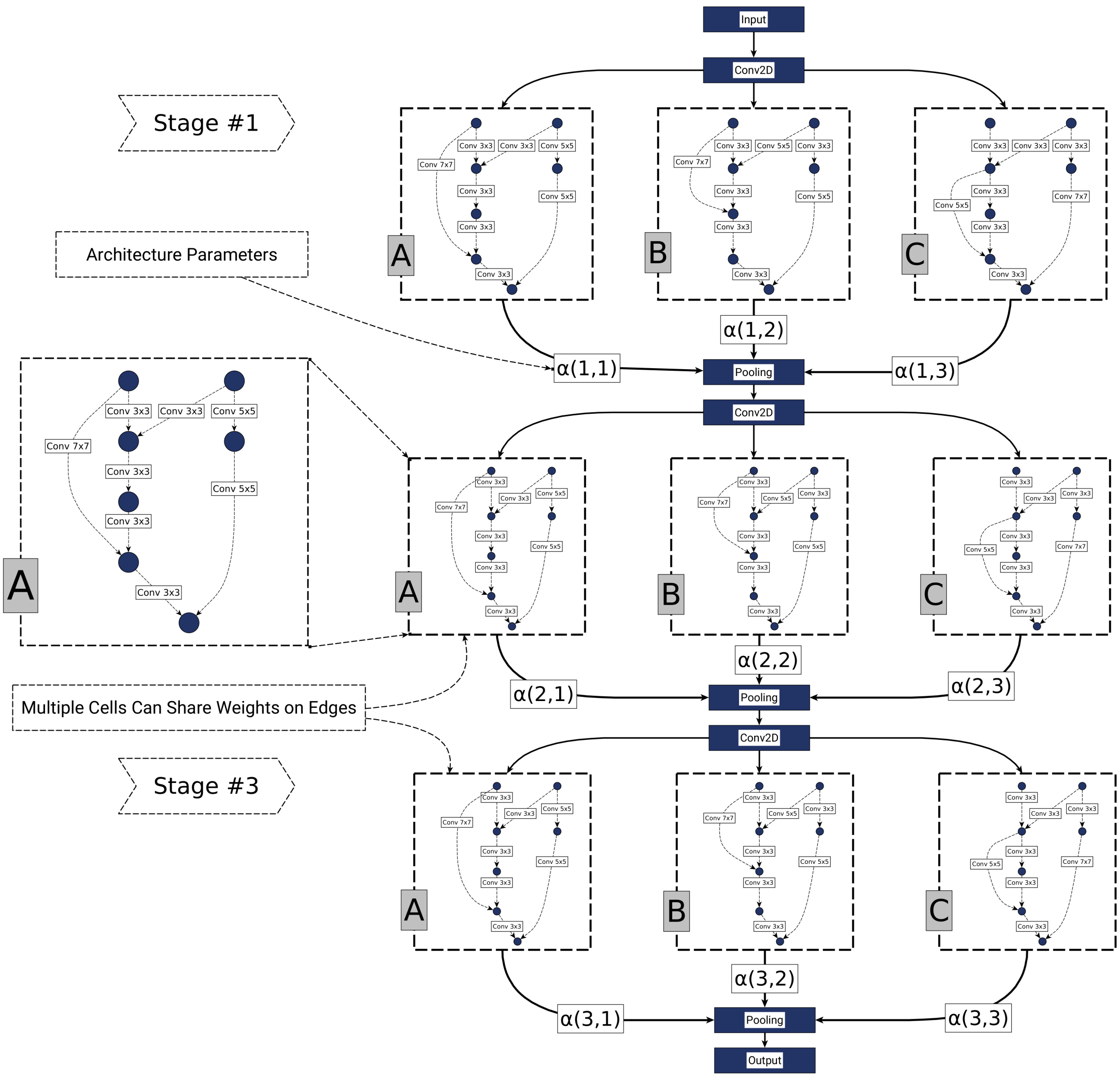

**Figure 13.1:** An example for a differentiable hyper-architecture which is composed of multiple stages with each containing multiple cells. A cell in FaDE [205] is an operation on the hyper-architecture similar to the ones used in DARTS as visualised in fig. 12.11 of section 12.6, but now structurally more complex sub-graphs are used. Architectural parameters $\alpha$ are learned in DARTS-fashion and aggregated into scores.

are interested in after training the overall model on eq. (12.1) according to DARTS-techniques.

An implicit permutation determines the order of themes integrated into the hyper-architecture such that we obtain a vector of architecture parameters $\alpha_{H(W)}(T) \in \mathbb{R}^{n_{stages}}$. From a hyper-architecture we obtain FaDE-scores based on [205, Sec. 2.4]:

$$\begin{aligned} \psi_{H(W)} &: \mathbb{S} \to \mathbb{R} \\ T &\mapsto \prod_{k=0}^{n_{stages}} \alpha_{H(W)}(T)_k \end{aligned}$$

A FaDE-score is the weighted importance of a structural theme relative to a window of other themes based on architecture parameters obtained from a DARTS-like hyper-architecture.

### 13.2.2 *Local Step with FaDE Score in NAS*

Neural architecture search is conducted by obtaining an optimisation landscape of performance scores with a performance estimation strategy. We take FaDE-scores as a surrogate for such costly evaluated performance scores because we observed experimentally in [205, Sec. 3.1, Fig.4 (b)] that FaDE-scores taken as ranks correlate with taking performance scores as ranks. Neumeyer et al. concluded from this rank correlation, that "the obtained FaDE-scores can be used to guide an open-ended search with **local information**" [205, Sec. 3.1, p.9].

However, FaDE-scores have two caveats:

1. FaDE-scores are aggregated scores from a differentiable hyper-architecture which carries more parameters and operations than just a realisation of a structural theme. The integrated neural network realisations are only related but in no means equivalent to a single neural network realisation sampled and trained for a theme-induced architecture. Our intention is that the hyper-architecture differentiates between beneficial and unfavorable structural properties of themes.

2. The usage of a window of search space candidates poses the learning of architecture parameters as a competition of integrated neural network realisations. This makes a FaDE-score relative in nature, i.e. the superiority of a theme with one FaDE-score over another theme with a lower FaDE-score holds only within the evaluated window $W$ of the search space. In comparison to performance scores, FaDE-scores are not only stochastic but also local.

The advantage of FaDE-scores is that they are quickly obtained for multiple search space candidates at once. Like in DARTS, multiple architectural decisions are taken at once within the hyper-architecture and additionally weight-sharing techniques can be employed.

The search space $\mathbb{S}$ with FaDE-scores as described here is not differentiable and also not continuous which has two implications: **1st/** we need to employ a gradient-free search strategy and **2nd/** we can not easily obtain similar graphs to the already found themes in the search space $\mathbb{S}$.

Because it is not differentiable, we investigated on a kind of pattern search [105] method, related to the possibly more popular Nelder-Mead (simplex) method [202] as search strategy. The "application of direct (*pattern*) search to a problem requires a space of points $P$ which represent possible solutions, together with a means of saying that $p_1 \in P$ is a better solution than $p_2 \in P$" [105, Sec. 2]. Gradient-free alternatives would be CMA-ES, Bayesian Optimisation or Evolutionary Searches as in section 12.3. With the idea of a low-dimensional Euclidean search space based on structural properties, we considered the type of pattern search more favorable as we can use finite differences for a pseudo-gradient search [204, Sec. 2.4].

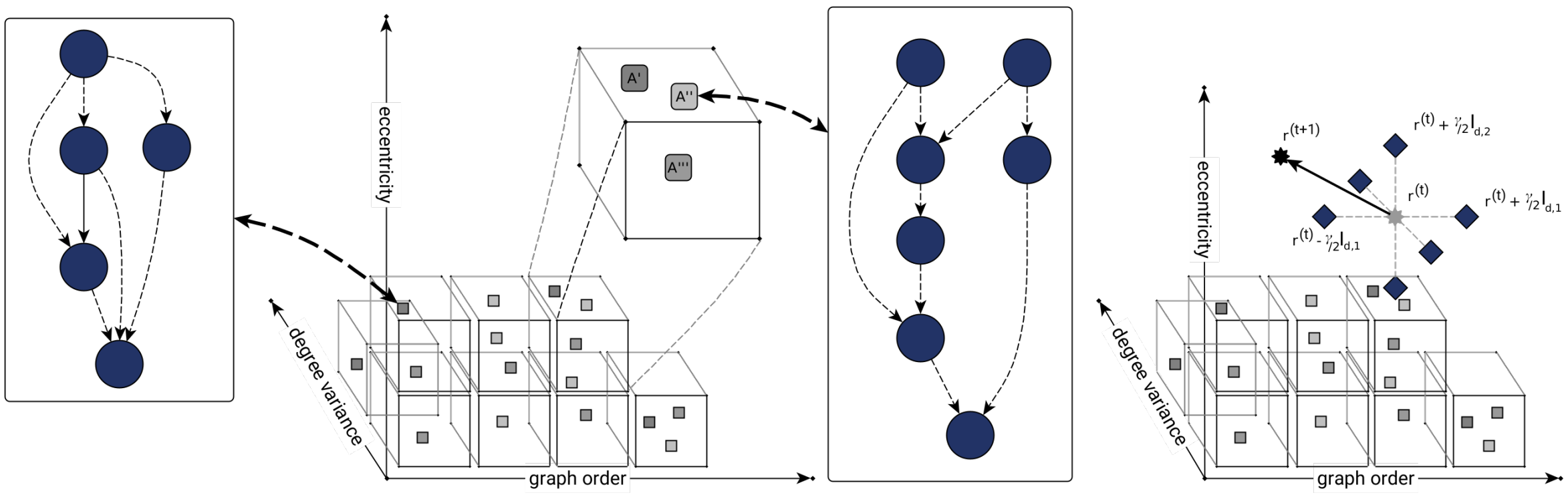

**Figure 13.2:** A surrogate search space $\hat{\mathbb{S}} \subseteq \mathbb{R}^{d=3}$ contains properties of pre-sampled graphs (structural themes). The reverse mapping $\hat{T}$ quickly provides a graph for a given point of structural properties based on a database (or sampling strategy that quickly obtains the theme for given properties). We obtain a nearby graph $\hat{T}(\mathbf{r}^{(t)} + \frac{\gamma}{2} I_d))$ by taking a point in a local environment around the reference point $\mathbf{r}^{(t)}$ with distance $\gamma > 0$. FaDE [204] originally takes two points per dimension of $\hat{\mathbb{S}}$ by coupling the structural property underlying the dimension into the window of the hyper-architecture $H(W)$.

Because $\mathbb{S}$ is not continuous, we have an issue of finding nearby graphs during the search. Recall, that a similarity between graphs is expensive (section 3.3) but we require means to sample nearby graphs to search through the space. Evolutionary or genetic searches employ variation operators for this but because of our analysis in section 12.5 we searched for an alternative for using a variation operator. This led us to using a low-dimensional Euclidean space $\hat{\mathbb{S}} \subseteq \mathbb{R}^{d=3}$ as surrogate search space in which each dimension is associated with a structural property [204].

We selected the properties graph order, *eccentricity*, and degree variances – which we previously found to be impactful as of structure analyses in chapter 9. Therefore the dimensionality of the surrogate search space becomes $d = 3$. The surrogate search space can be understood as a manually defined Euclidean embedding of graphs and enables us to have a notion of local neighboorhood around graphs as visualised in fig. 13.2. We pre-sampled the surrogate search space to obtain a database of structural properties from which we quickly could map a reference point back into a graph $\hat{T} : \mathbb{R}^{d=3} \to \mathbb{S}$. Having a refer-ence point in $\mathbf{r}^{(t)} \in \hat{\mathbb{S}} \subseteq \mathbb{R}^{d=3}$ then gives us a set of pre-sampled graphs which carry the particular structural properties defined by $\mathbf{r}^{(t)}$.

For each dimension $i \in \{1, \dots, d\}$ of $\mathbf{r}^{(t)} \in \hat{\mathbb{S}}$, we take two points along the unit vectors $\mathbf{e}_i$ to compute a pseudo-gradient in $\hat{\mathbb{S}}$. We denote them as $\mathbf{r}_{i,\uparrow}^{(t)} = \mathbf{r}_i^{(t)} + \frac{\gamma}{2} I_{d,i}$ and $\mathbf{r}_{i,\downarrow}^{(t)} = \mathbf{r}_i^{(t)} - \frac{\gamma}{2} I_{d,i}$ where $I_{d,i} \in \mathbb{R}^d$ denotes unit-vector of dimension $i$ of the identity matrix. These points are used to map back nearby themes in the original search space and evaluate FaDE-scores on $H(W)$ with $W = \{\hat{T}(\mathbf{r}^{(t)})\} \cup \{\hat{T}(\mathbf{r}_i^{(t)} \pm \frac{\gamma}{2} I_{d,i}) \mid i \in \{1, \dots, d\}\}$. Extracting the FaDE-scores along each dimension we obtain $\psi_{\mathbf{r}^{(t)},\gamma,H(W),\uparrow} = (\psi_{H(W)}(\mathbf{r}_{1,\uparrow}^{(t)}), \dots, \psi_{H(W)}(\mathbf{r}_{d,\uparrow}^{(t)}))$ and $\psi_{\mathbf{r}^{(t)},\gamma,H(W),\downarrow} = (\psi_{H(W)}(\mathbf{r}_{1,\downarrow}^{(t)}), \dots, \psi_{H(W)}(\mathbf{r}_{d,\downarrow}^{(t)}))$.

The NAS then performs a pseudo-gradient search with an initial reference point $\mathbf{r}^{(0)} \in \hat{\mathbb{S}}$ and gets updated through a finite difference:

$$\mathbf{r}^{(t+1)} = \mathbf{r}^{(t)} - \eta \left( \psi_{\mathbf{r}^{(t)},\gamma,H(W),\uparrow} - \psi_{\mathbf{r}^{(t)},\gamma,H(W),\downarrow} \right) \tag{13.1}$$

in which $\gamma > 0 \in \mathbb{R}$ is a "width of the local environment" [204, p. 7] around the reference point and $\eta > 0 \in \mathbb{R}$ is the learning rate (or step size) of the (pseudo-) gradient descent.

Recall from section 5.6.1 how this form is based on numerical approximation of the gradient [263, Sec 6.3 p. 157]. The update takes two nearby points $\mathbf{r}^{(t)} + \frac{\gamma}{2} I_{d,i}$ and $\mathbf{r}^{(t)} - \frac{\gamma}{2} I_{d,i}$ in $\hat{\mathbb{S}} \subseteq \mathbb{R}^{d=3}$ along each dimension. The points are mapped back into a graph via $\hat{T}$ and FaDE-scores are obtained by putting all related themes $T \in \mathbb{S}$ into $W$ such that $\psi_{H(W)}$ can be obtained after training the hyper-architecture $H(W)$. Simplified to just the obtained FaDE-scores eq. (13.1) performs a pseudo-gradient-style update in $\hat{\mathbb{S}}$ by $\mathbf{r}^{(t+1)} = \mathbf{r}^{(t)} - \eta(\psi_\uparrow - \psi_\downarrow)$ where structural themes for $\psi_\uparrow$ and $\psi_\downarrow$ are close under structural properties represented by $\mathbf{r}^{(t)}$ and $\mathbf{r}^{(t+1)}$. After a stopping criterion such as a maximum number of hyper-architecture trainings, the reference point $\mathbf{r}^{(t_{final})}$ can be used to obtain a final candidate $\hat{T}(\mathbf{r}^{(t_{final})}) \in \mathbb{S}$.

We compared the NAS-method in [204, Sec. 3.2] against random and Bayesian search and found the results promising to further investigate on FaDE in future, see Figure 13.3.

The here presented technical description deviates from the work in Neumeyer et al. [204, 205] in the following:

- The search space in [205] and the Euclidean surrogate search space are originally more tightly coupled and described independently of the concept of a structural theme.

- On further elaboration of the "mapping of sub-modules to their corresponding architecture parameter $\beta_i(\cdot)$" [204, Eq. 5] we found our methodological description to be unclear w.r.t. to provided implementations. In the light of graph-induced neural networks the inverse mapping $\hat{T}$ replaced this previous mapping and we found that $2d + 1$ neural network realisations for $W =$ $\{\hat{T}(\mathbf{r}^{(t)})\} \cup \{\hat{T}(\mathbf{r}_i^{(t)} \pm \frac{\gamma}{2} I_{d,i}) \mid i \in \{1, \dots, d\}\}$ are required to construct a hyper-architecture $H(W)$ to derive subsequent FaDE-scores.

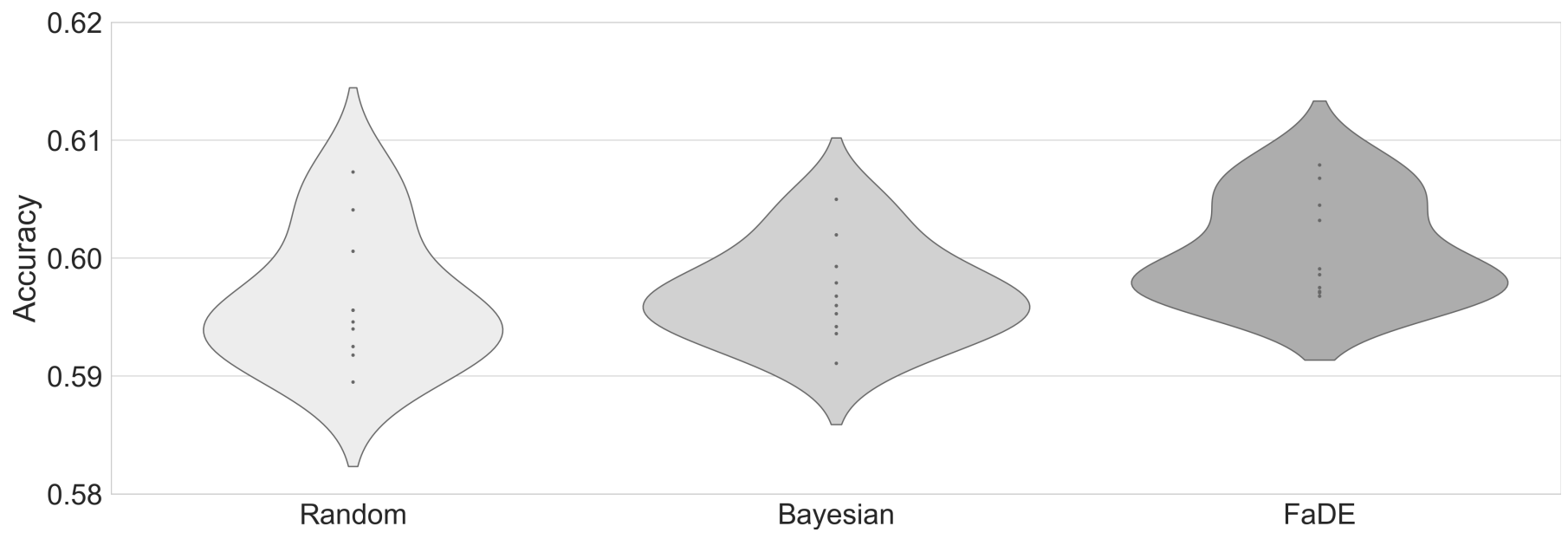

**Figure 13.3:** The result of a baseline NAS with random search (left), Bayesian search (middle) and FaDE (right) from Neumeyer et al. [204]. Reported are distributions in search space of median of five validation accuracies.

- Differences in namings: reference points $\mathbf{r}^{(t)}$ were referred to as anchor points $M^{(t)}$.

- The surrogate search space of reference points in [205] was normalised for each structural property in $(0, 1)$ but we found the formulation of eq. (13.1) then to not properly align with previous definitions.

We take these nuances as motivation to further investigate on the idea of using FaDE-scores from a hierarchical hyper-architecture as optimisation objective in neural architecture search.

REMARKS

- FaDE aggregates architecture parameters of a hierarchical **d**ifferentiable hyper-architecture into **f**ast **e**stimations of scores for candidates of a search space.
- We use FaDE-scores as substitute for performance estimations of individual neural networks with an expensive training scheme.
- FaDE-scores are relative in nature, i.e. for $T \in \mathbb{S}$ it can happen that $\exists W_1, W_2, W_3 \subset \mathbb{S}$ with $T \in W_1 \cap W_2 \cap W_3 : \psi_{H(W_1)}(T) < \psi_{H(W_2)}(T) < \psi_{H(W_3)}(T)$ while $\psi_{H(W_3)}(T) < \psi_{H(W_1)}(T)$. Therefore, for each update step we construct a new hyper-architecture $H(W)$ with window $W$ and FaDE-scores are only used for a pseudo-gradient of that particular step.
- Gradient-free methods such as evolutionary searches or pseudo-gradient methods can be used to conduct a neural architecture search with FaDE-scores instead or partially as substitute of performance scores such as the $F_1$ score.

### 13.3 SUMMARY ON PREDICTIVE MODELS FOR NAS

The performance estimation strategy is an expensive component of NAS because it takes time for training on large datasets and obtaining a performance score on a hold-out set. DARTS completely avoids iterative population-based training by searching through a differentiable hyper-architecture. The estimation strategy is integrated into the bi-level optimisation of the hyper-architecture. But DARTS is limited to a finite space and might be biased towards certain sub-structures. Performance prediction models are a promising component for other NAS-methods requiring an explicit performance estimation strategy.

Performance predictions can be based on learning curve predictions [13] but with the growing importance of the computational structures of deep neural networks, structural performance predictions also pose an interesting alternative. Manually engineered features for structural performance prediction already yield good results but based on the many findings of dependencies on the data of the application domain, automated methods might be preferable. Predictors are based on GNNs in our formulation of NAS in eq. (7.2) because the search space $\mathbb{S}$ consists of directed acyclic graph:

$$\arg\min_{T \in \mathbb{S}} \mathcal{L}_{val}(T, \arg\min_{f \in \mathbb{A}([T])^{d_1 \to d_2}} \mathcal{L}_{train}(f, D_{train})) \quad (7.2)$$

Automatic methods show that this performance prediction is indeed beneficial for NAS methods [207, 300]. Yet, the methods require some seed populations of models which are properly trained and evaluated.

With FaDE we present an alternative predictive approach: multiple neural network realisations as representatives of their architecture are jointly integrated in a DARTS-like hyper-architecture to aggregate architectural parameters into FaDE-scores. FaDE-scores then are used as a replacement estimation for the performance estimation strategies.

In a NAS, the idea of a low-dimensional surrogate search space $\hat{\mathbb{S}}$ further tackles eq. (7.2) by formulating the second level of the optimisation problem like DARTS in a differentiable manner instead of a discrete. This allows for a pseudo-gradient optimisation of FaDE-scores to conduct a neural architecture search. The method is compared to Bayesian Optimisation and Random Search on a common computational budget [204]. Results are promising and encourage to analyse the method with other NAS benchmarks in future.

# 14 GENERATIVE MODELS FOR NEURAL ARCHITECTURE SEARCH

*The following entails:*

We turn our focus on generative models for NAS and more particular to *graph generators*. Learning generative models of graphs comes in handy on multiple occations of the outlined graph-induced neural architecture search.

In Section 5.4, it was outlined, that learning the parameters of a neural network by maximizing the likelihood over data can be generalised to using maximum a-posteriori estimation. This generalisation included the prior distribution over the parameters which are often assumed or simplified to be uniform (or sometimes gaussian) such that the maximum a-posteriori formulation collapses into the commonly used maximum likelihood estimation. Capturing the graph structure of a model as part of this parameter distribution could lead towards a structural regularisation in a bayesian way. The discrete nature of graphs and the complexity of the growth of its space and certainly many more factors make this endeavour, however, currently inaccessible.

The bi-level optimisation problem in eq. (7.2) considers a search space $\mathbb{S}$ of graphs:

$$\arg\min_{T\in\mathbb{S}} \mathcal{L}_{val}(T, \arg\min_{f\in\mathbb{A}([T])^{d_1\to d_2}} \mathcal{L}_{train}(f, D_{train})) \quad (7.2)$$

Learning distributions of graphs can both navigate and accelerate this search. The scientific community on learning generative graph models has made significant steps forward in learning models that are capable of sampling both undirected and directed acyclic graphs [84, 321].

Modeling a graph generator as to learn a distribution over graphs is outlined in section 3.2 with simple but well-defined models $ER(n,m)$, $GIL(n,p)$, $WS(n,k,p)$, and $BA(n,m)$. In contrast to these well-defined models, learned models of graphs are based on exemplary graphs from a dataset or a separate (synthetic) generation process. We see the following modi of (deep) graph generators to enhance neural architecture search:

- learning a graph generator only from positive (good) examples as to resemble a distribution of desired graphs,
- learning a graph generator from positive and negative examples as to sample only from or improve on the positive class distribution,
- learning a conditional graph generator from which examples of different conditions can be drawn,
- or learning a generative model whose embedding space can be leveraged for regularisation during an optimisation problem.

Our contributions include an extension of the deep graph generative model DGMG (Deep Generative Models of Graphs) to DeepGG (Deep Graph Generators) [271], extensive experimental results comparing two successful existing models GraphRNN and GRAN [282], a multi-conditional extension of GRAN called MCGRAN (Multi-Conditional GRAN) [229], theoretical considerations for deep state machines – which is the underlying technique for DGMG [156] and DeepGG [271] –, and considerations for assemblic sequences of graphs as a non-learned substitute for deep generative models.

## 14.1 SEQUENTIAL REPRESENTATIONS FOR GRAPHS

A common way for generating graphs is to represent them in a sequential way and to learn to sample each construction step [156, 157, 271, 328]. For this, we extended on DGMG [156], investigated on debiasing the underlying model DGMG from producing scale-free graphs and generalised the model to DeepGG [271]. This also lead to further research on DGMG and DeepGG as deep state machines which we describe in section 14.6 and to representing graphs in assemblic sequences in section 14.7. Our main goal is to capture learning a distribution $P(G)$ of (possibly large) graphs for neural architecture search as to conduct efficient and effective sampling in a search space $\mathbb{S}$.

### *Representing Graphs in a Construction Sequence*

Recall, that an adjacency matrix representation of a graph $G$ is denoted with $\mathbf{A}(G)$. We now illustrate how several authors [156, 160, 328] use a sequential representation to learn graph generative models. We also call such representations *construction sequences* as to illustrate that by following the sequence step by step, a graph is created.

You et al. [328, Sec. 2.3.1] proposed to model graphs in sequences $S$ by defining a mapping $f_{\rightarrow S}$ from graphs to sequences such that for a graph $G$, a vertex permutation function $\pi \in \Pi$ the sequence is composed of $n$ items $S^\pi = (S^\pi_1, \dots, S^\pi_n)$ with $S^\pi_i \in \{0,1\}^{i-1}$ for $i \in \{1, \dots, n\}, n = |G|$:

$$S^\pi_i = (\mathbf{A}^\pi_{1,i}, \dots, \mathbf{A}^\pi_{i-1,i})^T$$

The graph $G$ is obtained from $S^\pi$ via a mapping $f_{\rightarrow G}$. You et al. then proposed to approximate the distribution $p(G)$ "as the marginal distribution of the joint distribution $p(G, S^\pi)$" with $p(G) = \sum_{\pi \in \Pi} p(S^\pi)\mathbb{I}[f_{\rightarrow G}(S^\pi) = G]$. Further, $p(S^\pi)$ can be decomposed into a product of conditional distributions over the sequence items $S^\pi_i$:

$$p(S^\pi) = \prod_{i=1}^{n} p(S^\pi_i \mid S^\pi_1, \dots, S^\pi_{i-1})$$

such that the model which learns this distribution needs to resemble the decisions whether to connect a current vertex to any vertex of the currently already created graph (with minor details as e.g. having an end-of-sequence symbol).

Technically, the data set of graphs $G_{data}$ then simply can be created by transforming the adjacency matrix of graphs into their *construction sequence* representation.

Similarly, for DeepGG [271], we represent the sequence with two decisions such that the symbol "N" reflects whether to add a new vertex and the symbol "E" reflects whether to add an edge. The second symbol, "E", is followed by a source and a target vertex, parameterizing the vertices which should be connected. The empty graph is represented by the empty sequence [ ], a graph with a single vertex with [ N ] and a graph with two connected vertices with [ N N E 0 1 ]. This is slightly different to DGMG [156] who only use one source vertex as a parameter following the symbol "E" such that [ N N E 0 ] represents a graph with two vertices connected. The advantage are shorter sequences but the disadvantage is that only the last added vertex can be connected to any other previous vertex and thus biases DGMG in our analysis towards scale-free graphs [271].

### *Block-wise Constructions in GRAN*

Liao et al. [160] proposed to resemble the adjacency matrix in blocks of construction steps instead of individual vertices. Their method, GRAN, is a more efficient model than GraphRNN, DGMG, or DeepGG, by considering larger blocks of the adjacency matrix at once and employing more recent techniques such as message-passing and attention [282].

Let $n_{max} \in \mathbb{N}$ be the maximum number of vertices over all graphs $G_{data}$ (for training and testing). $B$ describes the number of rows the model considers to generate in one block such that the $t$-th block contains rows with indices $\mathbf{b}_t = \{B(t-1)+1, \dots, Bt\}$. Depending on $B$ number of steps required to generate the graphs, we set $T = n_{max}/B$. Then, $L^{\pi}_{\mathbf{b}_i} \in \mathbb{R}^{BN}$ is a block of lower triangular matrix represented by $[L^{\pi}_{B(i-1)+1}, \dots, L^{\pi}_{Bi}]$. $[.]$ is a concatenation operator. GRAN generates the lower triangular matrix $L^{\pi}$ by factorizing $p(L^{\pi})$ into product of conditional distributions $p(L^{\pi}_{\mathbf{b}_t} \mid L^{\pi}_{\mathbf{b}_1}, \dots, L^{\pi}_{\mathbf{b}_{t-1}})$. Exemplary block-structures in time steps $t-2$, $t-1$ and $t$ are depicted in fig. 14.1.

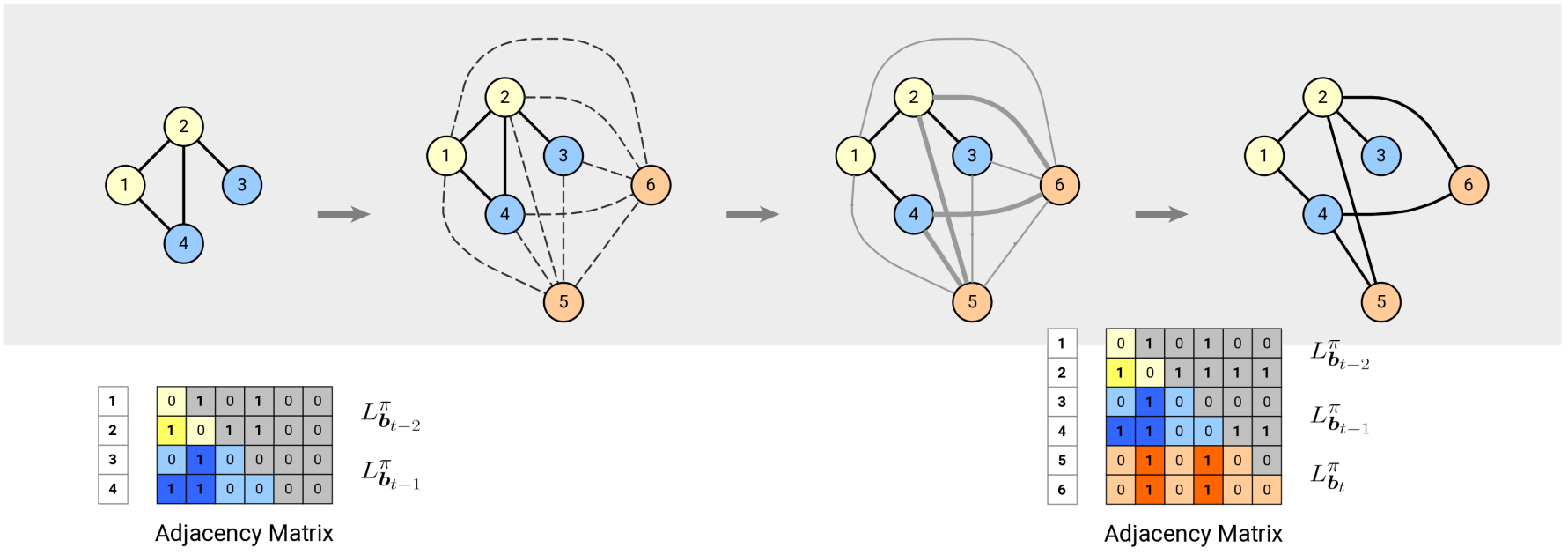

**Figure 14.1:** A visual example of GRAN [160, Fig. 1, p. 3], showing a graph at time step $t-1$ and $t$ where new edges are sampled after learning logit strengths through graph recurrent attentive message passing.

## 14.2 DEEPGG: LEARNING A DEEP STATE MACHINE FOR GRAPHS

With DeepGG [271] we modified a previous graph generative model DGMG [156] which was published shortly after GraphRNN [328]. DeepGG and DGMG are examples for state-based deep learning models, a concept which recently gains new traction as a general pattern but which we also describe independently in section 14.6 on page 273 [120].

The idea is to learn a graph based on the presented construction sequence with two operations: by adding a vertex or by adding an edge between two vertices. The decision for these operations are derived from a graph representation $\mathbf{h}_g \in \mathbb{R}^{H_g}$. We denote the vertex manifold dimension $H_v \in \mathbb{N}^+$, the edge manifold dimension $H_e \in \mathbb{N}^+$ and the graph manifold dimension $H_g \in \mathbb{N}^+$ and by default set $H_g = 2 \cdot H_v$.

DGMG starts with an empty graph $G = \emptyset$ and is represented with a zero vector $\mathbf{h}_g = \mathbf{0} = (0, \dots, 0)$. All following functions $f$ are multi-layer perceptrons. Every time, a vertex $v_{new}$ is added, a representation $\mathbf{h}_{v_{new}}$ for it is initialised with $f_{init,v} : \mathbb{R}^{H_g} \to \mathbb{R}^{H_v}$. Analogously, every time, an edge $e_{new}$ is added, a representation $\mathbf{h}_{e_{new}}$ for it is initialised with $f_{init,e} : (\mathbb{R}^{H_v}, \mathbb{R}^{H_v}) \to \mathbb{R}^{H_v}$.

We identify a state with index $\kappa \in \{1, \dots, n_{states}\}$ as $s_\kappa$ with $n_{states} \in \mathbb{N}$ being the number of employed states. States can transition into other states as visualised in fig. 14.2 on the next page. DeepGG consists of two states $s_1$ and $s_2$ for which the first state decides about adding vertices and the second state decides about adding edges. A transition decision function draws from a categorical distribution based on a vector representation consisting of a graph representation $\mathbf{h}_g$, a composition of vertex representations $\mathbf{h}_{v_1}, \mathbf{h}_{v_2}, \ldots$ or a mixture of both. Therefore, the input dimension of a transition decision function is denoted with $H_{s_\kappa}$. For each state it needs to be defined how to aggregate information from the vector representations into the particular input representation from which a state transition decision is derived [271, Sec. 4].

To get to vectorised low-dimensional representations from a graph, techniques of geometric deep learning such as graph convolutions are employed (compare section 5.7.3) within the DNN layers:

$$f_{conv}(H^{(l)}, A) = \sigma\left(D^{-\frac{1}{2}}\hat{A}D^{-\frac{1}{2}}H^{(l)}W^{(l)}\right)$$

with $\hat{A} = A + I$ and $A$ being the adjacency matrix of graph $G$.

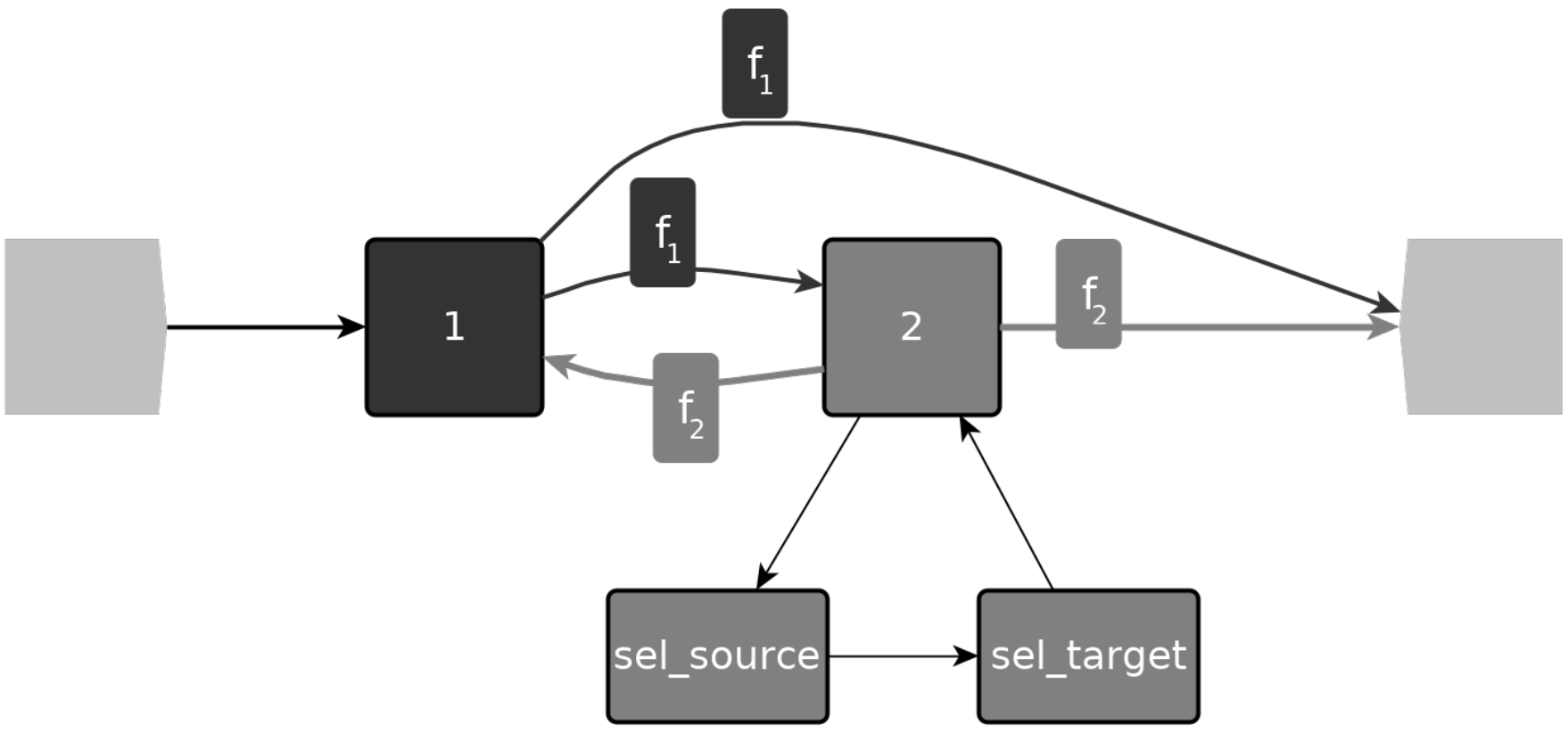

**Figure 14.2:** Two states, denoted with "1" and "2" make up the underlying state machine of both DeepGG [271] and DGMG [156]. The model becomes generative by starting in the initial state and transitioning between states until the final state is reached. Instead of randomly choosing states, the state transitions are learned from data and a graph gets steadily built up, embedded into a representation and new operations are derived from this representation.

The decision function $f_{s_\kappa}$ is defined as $f_{s_\kappa} : \mathbb{R}^{H_{s_\kappa}} \rightarrow \{1, \ldots, n_{states}\}$.

STEP-BY-STEP CONSTRUCTION Imagine a simple construction sequence such as [ N, N, E, 0, 1 ] as presented in section 14.1. The initial (and current) state of the graph $G^{(0)} = \emptyset$ is represented with $\mathbf{h}_g = \mathbf{0}$. DeepGG starts in a start state $s_{current} = s_1$ and performs its associated action, adding vertices. State one adds a vertex such that $G^{(0)}$ becomes $G^{(1)} = (\{v_1\}, \emptyset)$. A vertex representation $\mathbf{h}_{v_1} = f_{init,v}(\mathbf{h}_g)$ is associated with the freshly added vertex $v_1$. The graph representation $\mathbf{h}_g$ is updated based on the new underlying graph structure $G^{(1)}$ (now containing one vertex). State $s_1$ uses its transition decision function to decide in which state to go next: $s_{current} = f_{s_1}(s_{current})$. According to the exemplary construction sequence [ N, N, E, 0, 1 ] the result of $s_{current}$ would be expected to be state one again – to add another vertex.

During learning, the cross-entropy loss between state decisions and the observed expected states from the construction sequence add up to the overall optimisation objective of log losses. The model therefore works like most DNN models in a training and an inferencing mode. If the model conducts wrong decisions during training, it can be accounted for in the loss for learning but the state transitions are done according to the supervised construction sequence. In that way, it can be enforced that the model continues growing the correct graph.

After two vertices have been added into $G^{(2)}$, an edge needs to be added between the two vertices. For this, DeepGG transitions into state $s_2$. In this state two vertices need to be selected: first a source and second a target vertex for the edge to be added. These are categorical choices over the set of all vertices, except for the target choice excluding the source vertex. The expected vertices are parameterised in the construction sequence with its numbers used during construction. DeepGG can use this information to learn the correct choices of source and target vertices. The log loss between the choice made by the model and the expected vertex number is added up. Note, that with longer construction sequences, the possibilities of vertices to choose from as a source or target of an edge grows. To compensate for exploding losses, the cross-entropy loss is divided by the logartihm of possible choices. Again, a representation for the added edge $e_1 = (v_1, v_2)$ is initialised through $\mathbf{h}_{e_1} = f_{init,e}(\mathbf{h}_{v_1}, \mathbf{h}_{v_2})$ and $G^{(2)}$ becomes $G^{(3)} = (\{v_1, v_2\}, \{e_1\})$.

MEMORY UPDATES After each state action, modifying the underlying graph, the graph representation $\mathbf{h}_g$ is updated. For this, DeepGG uses gated vertex representations in conjunction with graph convolutions [271, Sec. 4]. The updating mechanism applies a reducing gating submodule function $f_{reduc} : \mathbb{R}^{H_v} \to \mathbb{R}^{H_r}$ with $\mathbf{h}_v \mapsto \sigma^{\sim}(W_{reduc} \cdot \mathbf{h}_v + b_{reduc})$ for which $H_r = 7$ is chosen. Each vertex representation is then used in a subsequent graph convolution whose output dimension is chosen to be $H_g$ again. The vertex representations $\mathbf{h}_{v_i}$ after message passing are then aggregated by means of an averaging such that an updated graph representation $\mathbf{h}_g$ is obtained – which should contain updated information about the modified graph $G$.

In addition, a message passing for updating the vertex representations is conducted only after adding edges in state $s_2$. For the vertex update a gated recurrent unit cell [40, 102] is used and the number of message passing propagation rounds is set to two [271, Sec. 4, p.8].

GENERATIVE MODE A trained DeepGG model can be easily put into training mode. In the same manner as for training, an empty graph is passed into the state machine and a graph representation for it gets initialised. The state transitions take place in a similar fashion as during learning, except for that the actual decisions of the differentiable sub-modules of each state are taken to choose in which state to go. Information about a correct sequence do not have to be carried along and no errors are to be computed. On the other hand, a novel con-struction sequence is built up alongside traversing the state machine. The generative mode results in a graph and a construction sequence after the DeepGG decided to transition into the final state or another stopping criterion such as a maximum number of state transitions is reached.

*Results of DeepGG*

Experimental results on DeepGG have been published in [271] and we provide only a short overview here.

Reproducing results of DGMG led us to trying to de-bias the generative model from scale-free graphs. The Barabasi-Albert model in its assemblic nature adds vertices step by step and adds edges for each new added vertex by choosing among previous vertices with a probability depending on the degree of that vertex (compare section 3.2). With the underlying state machine of DGMG, early added vertices have a high likelihood of being chosen as targets for edges. This results in a tendency of resembling graphs similar to the Barabasi-Albert model.

In DeepGG, the second state of adding edges, allows to sample any edge at any point during the generation process. This makes learning the sequence significantly more difficult as then both a source and a target vertex for an edge need to be chosen. We tried to compensate for increasing log losses by reducing the accumulating loss over time. Still, while resembling graphs of the Erdős-Rényi model, the Watts-Strogatz model, or the Barabasi-Albert model quite well, the training of DeepGG and DGMG are expensive. An assessment of the generation quality of DeepGG based on a comparison of average path length distributions is provided in fig. 14.3. Further details are to be taken from [271].

Auto-regressive (or sequential) models such as GraphRNN, DGMG or DeepGG are limited to its provided operation on graphs. Precisely, they are only capable of either adding a vertex or adding an edge. GRAN allows to sample multiple edges (or vertices) in a block-wise fashion. But this comparison also brought up the insight that other graph operations might be fruitful to investigate on. Graphs which model a natural assemblic process such as protein folding or network evolution might not just rely on adding single connections per time step. We call this generalisation of sequential representations of graphs *assemblic representations* and see future potential work in it.

The idea of DGMG and DeepGG of working on a per-state basis and carrying an additional structure which is embedded with deep neural networks as to learn it also led to an idea of generalizing the concept of deep state machines of which we will provide sketches in section 14.6. An interesting tool for analysing such models is then to compare the generative capability of state-based models with results of randomly transitioning the state machine without considering learned state transition decisions.

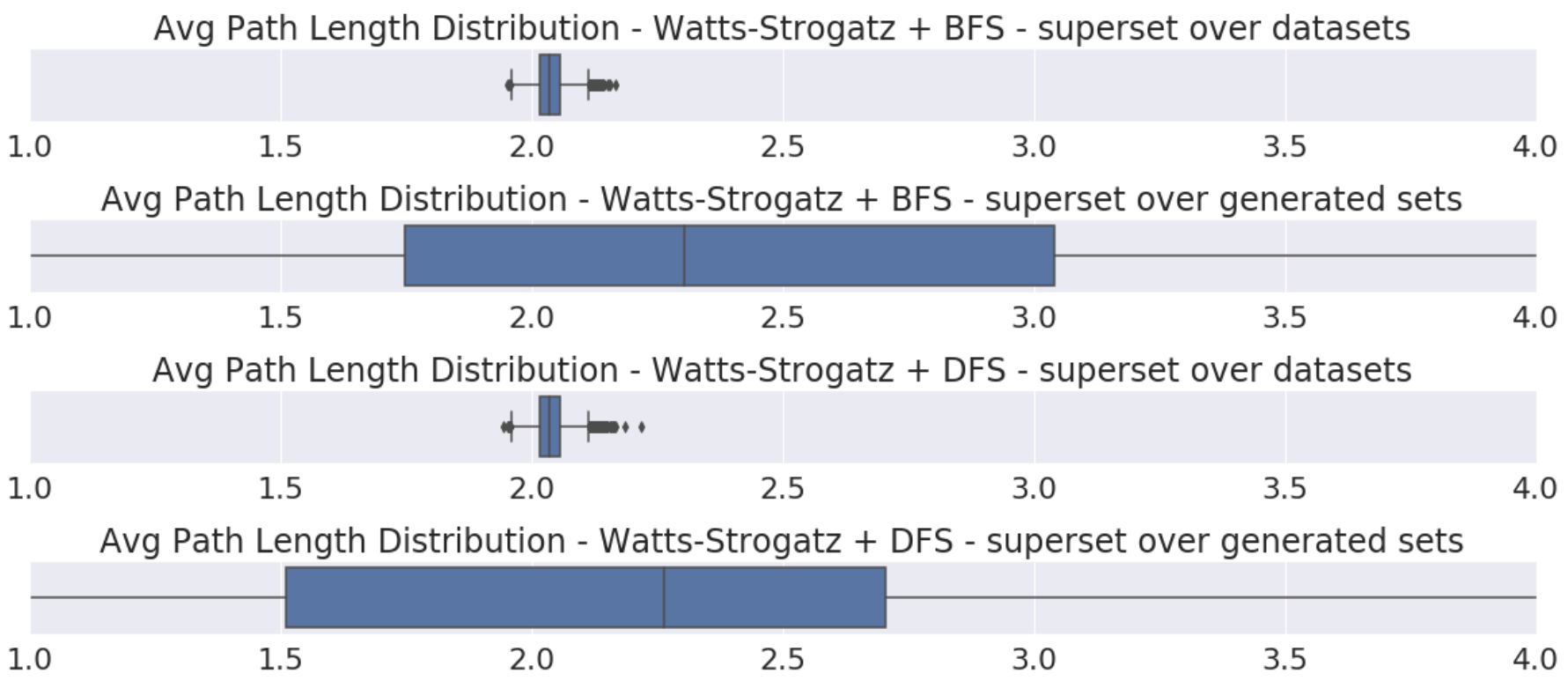

**Figure 14.3:** For multiple instances of *DeepGG* trained on Watts-Strogatz model graphs, we computed the average path length of each of the 1,000 graphs used for training for both bfs- and dfs-traversal. Equivalently, we computed the average path length of each generated graph from the model (on average 171 for the bfs-instances and 170 for the dfs-instances). Seven *DeepGG* instances have been trained on bfs-traversals, twelve on dfs-traversals. Watts-Strogatz model+BFS constitute a superset in which we generated 1,199 graphs, Watts-Strogatz model+DFS a superset in which we computed 2,034 graphs. As to no surprise, the average path length distributions for both supersets over the datasets are visually equivalent. However, we have to ascertain that the average path length distributions for generated graphs can not only be matched its dataset but also have to point out that the resulting superset distribution for different traversal choices for construction sequences differ significantly.

REMARKS

- DGMG is biased towards scale-free graphs.
- DeepGG tries to de-bias from when to add edges but suffices from more parameters in the construction sequence and also larger sequence lengths.
- The research on DeepGG further lead to an idea of random traversals for analysis of DeepGG or DGMG and assemblic representations of graphs.
- GRAN improved over previous single-step sequence-based models GraphRNN, DGMG and DeepGG by modeling whole blocks of the adjacency matrix of a graph at once [282].

## 14.3 A RECAP OF GRAN

Recall, that GRAN models the lower triangular matrix $L^{\pi}$ through conditional distributions $p(L^{\pi}_{\mathbf{b}_t} \mid L^{\pi}_{\mathbf{b}_1}, \dots, L^{\pi}_{\mathbf{b}_{t-1}})$. During training and inference, graphs smaller than $n_{max}$ are set to have $L^{\pi}_i$ padded with zeros [160, Sec. 2.2] (i.e., mask) to prevent the model from seeing the future. The initial hidden vector representation $h^0_{\mathbf{b}_i}$ of $\mathbf{b}_i$ block of vertices is obtained through a linear mapping $WL^{\pi}_{\mathbf{b}_i} + b$ for all $i < t$. $L^{\pi}_{\mathbf{b}_i}$ denotes the lower triangular matrix of the generated graph until time step $i$.

At time step $i \in \{1, \dots, T\}$ and after $r \in \{1, \dots, R\}$ rounds of message passing the vertex vector representation is denoted with $h^r_{\mathbf{b}_i} \in \mathbb{R}^{BH}$ with $H \in \mathbb{N}^+$. Further, we add $B$ additional vertices of the new block at time step $t$. Augmented edges connect all the newly added vertices and their predecessor $B(t-1)$. With the teacher-forcing technique [306, Sec. 2.3], the model learns to retain and remove some edges between the vertices. To learn the long-term dependencies, a graph neural network then updates the vertex representations of the augmented graph $G_t$ [160, Eqs. 3-6]:

$$m^r_{ij} = f(h^r_i - h^r_j) \tag{14.1}$$

$$a^r_{ij} = \text{Sigmoid}(g(\tilde{h}^r_i - \tilde{h}^r_j)) \tag{14.2}$$

$$\tilde{h}^r_i = [h^r_i, x_i] \tag{14.3}$$

$$h^{r+1}_i = \text{GRU}(h^r_i, \sum\nolimits_{j \in \mathcal{N}(i)} a^r_{ij} m^r_{ij}) \tag{14.4}$$

In eq. (14.1), $m^r_{ij}$ is a pairwise distance function $f$ between vertex representations $h^r_i$ and $h^r_j$. In eq. (14.3), vertex representations $h^r_i$ augments the $B$-dimensional binary mask $x_i$ in which $x_i$ indicates whether vertex $i$ is in the new block of $B$ vertices. Liao et al. [160] chose this setup to let GRAN "distinguish between existing vertices and vertices in the current block and to learn different attention weights for different types of edges". Additionally, in eq. (14.4), using an attention weight $a^r_{ij}$ based on a non-linearity function $g$ (Eq. 14.2) and the current vertex representations $h^r_i$, the GRU learns the vertex representation $h^{r+1}_i$ for the next round. The output distribution of generating a block $L^\pi_{\mathbf{b}_t}$ is obtained through a mixture of bernoulli distributions [160, Eqs. 7-9]:

$$p(L^\pi_{\mathbf{b}_t} \mid L^\pi_{\mathbf{b}_1}, \dots, L^\pi_{\mathbf{b}_{t-1}}) = \sum_{k=1}^{K} \alpha_k \prod_{i \in \mathbf{b}_t} \prod_{1 \le j \le i} \theta_{k,i,j} \tag{14.5}$$

$$\alpha_1, \dots, \alpha_K = \sigma^\Psi \left( \sum\nolimits_{i \in \mathbf{b}_t, 1 \le j \le i} f_\alpha(h^R_i - h^R_j) \right) \tag{14.6}$$

$$\theta_{1,i,j}, \dots, \theta_{K,i,j} = \sigma^\sim \left( f_\theta(h^R_i - h^R_j) \right) \tag{14.7}$$

The number of mixture components is determined by $K \in \mathbb{N}$. Both $f_\alpha$ and $f_\theta$ are deep neural networks with two ReLU-based hidden layers and have $K$-dimensional outputs.

## 14.4 mcgran: conditional graph generation with gran

We deeply investigated on the auto-regressive models for graph generation in Touat et. al [282] and especially compared GraphRNN [328] to GRAN [160]. For assessing the generation quality we not only used kernel-based maximum mean discrepancy but also studied the generated graphs in a learned manifold of the embedding of a Graph Isomorphism Network [320] classification model. This evaluation approach was also proposed by Obray et al. [208] around the same time. We concluded a general superiority of GRAN over GraphRNN, especially when scaling for larger graphs, and found the reasons to be based on

1st/ improved techniques such as attention mechanisms, but also

2nd/ based on a larger co-domain for sampling per step (compare the representation of blocks instead of single vertices or edges as in the sequential representation described in section 14.1).

These observations motivated us to study possibilities to learn a conditional distribution $P(G \mid C)$. In its simplest form, with two possible conditions $C \sim$ Ber, such a conditional distribution allows to be trained on two sets of graphs which could e.g. represent low-performing $C = 0$ and high-performing $C = 1$ neural architectures from a NAS benchmark such as NAS-Bench-101 or CT-NAS. During training, a conditional model is presented with an additional conditioning information $C$. In case of NAS-Bench-101, this information can be based on the estimated test accuracy. During generation, the conditioning can be used to only sample high-performing neural architectures with $C = 1$.

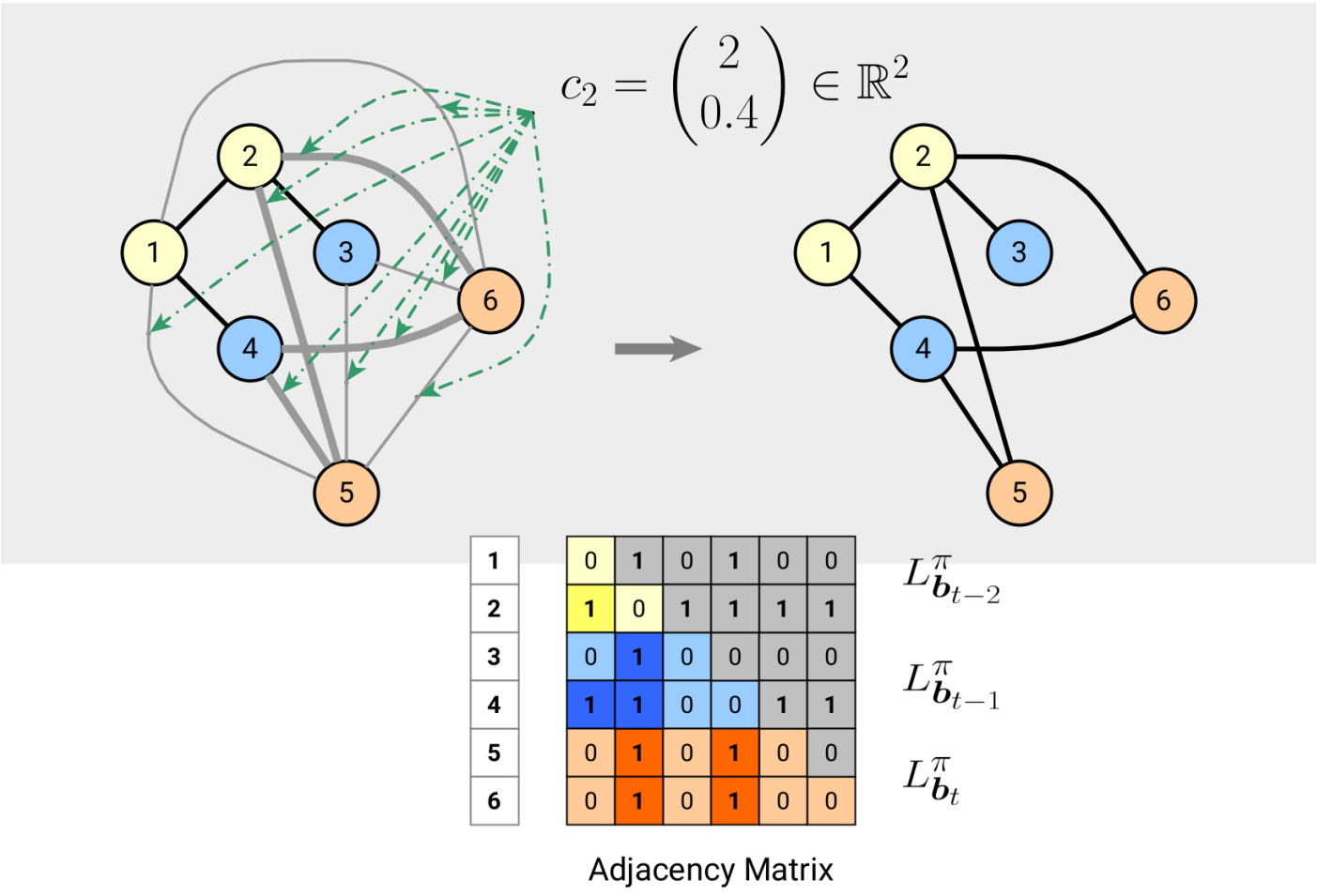

**Figure 14.4:** A visual example of MCGRAN [229] showing the transitioning process of GRAN from time step $t-1$ to $t$ with the influence of the conditioning vector on the edge sampling process for block $L^{\pi}_{\mathbf{b}_t}$. Compare fig. 14.1 and Liao et al. [160, Fig. 1, p. 3]. The graph can be directed if the upper triangular matrix of the adjacency $\mathbf{A}$ or blocks $L^{\pi}_{\mathbf{b}_i}$ are zeros. Diagonals are zero as to avoid self-loops.

The results on the comparative study on GraphRNN and GRAN and the idea to improve a NAS with conditioning led Purushothaman et al. to an extension of GRAN, namely Multi-Conditional Graph Attention Recurrent Networks (MCGRAN) [229]. MCGRAN not only introduced multi-conditioning for GRAN in application for NAS but also a component for vertex label prediction.

The label prediction component is an important detail as architectures in NAS-Bench-101 are represented with *labelled* directed acyclic graphs and this class of graphs contains significantly more non-isomorphic candidates than its unlabelled variant, compare fig. 3.7 on page 40 with sequence identifiers A345258 and A345258 [1]. Further, the generation quality of MCGRAN w.r.t. the label prediction quality becomes difficult to assess because each label assignment can be a valid graph of the desired target distribution of graphs. Depending on the employed dataset, a random label assignment could pose a valid prediction component and $F_1$ score evaluations of such a component are rendered obsolete. The study on MCGRAN contains comparisons of using affine transformations or SplineCNN [66] as alternative components for vertex label prediction [229].

MCGRAN enables multi-conditioning by introducing a conditioning vector $c \in \mathbb{R}^{n_{cond}}$ in $n_{cond} \in \mathbb{N}$ conditioning dimensions which is used to scale and shift the pre-activations in GRAN with a learned affine transformation [229, Sec. 3.1]. The transformation is applied after obtaining the log probabilities from pairwise differences of vertex representations. This turns equations eqs. (14.6) to (14.7) on page 269 into

$$z = \left(\sum\nolimits_{i\in \mathbf{b}_t, 1\leq j\leq i} f_\alpha(h_i^R - h_j^R)\right)$$

$$\alpha_1, \dots, \alpha_K = \sigma^\Psi(z \cdot \beta_{scale} + \beta_{shift}) \tag{14.8}$$

$$\theta_{1,i,j}, \dots, \theta_{K,i,j} = \sigma^\sim\left(f_\theta(h_i^R - h_j^R)\beta_{scale} + \beta_{shift}\right) \tag{14.9}$$

as outlined in [229, Sec. 3.1, eqs. 3&4] in which the conditioning scale $\beta_{scale}$ and $\beta_{shift}$ vectors are computed as

$$\beta_{scale} = f_{scale}(c) \tag{14.10}$$

$$\beta_{shift} = f_{shift}(c) \tag{14.11}$$

with new learnable non-linearities $f_{scale}$ and $f_{shift}$, usually set to be single- or multi-layered neural networks.

Experiments of Purushothaman et al. [229, Sec. 4] showed that MCGRAN successfully resembled distributions that were conditioned to specific ranges. Two independent experimental settings with once a range of high-performing and once a range of low-performing neural architectures were carried out. MCGRAN sampled graphs whose architectures yielded $0.90 \pm 0.01$ in test accuracy for high-performing training samples taken from the range of 80% to 100% in test accuracy in NAS-Bench-101 and MCGRAN yielded $0.53 \pm 0.17$ in test accuracy for low-performing training samples taken from the range of 30% to 60%

1 The sequences are identified with keys of the online encyclopedia of integer sequences [112].

in NAS-Bench-101 [229]. Detailled experiments during an iterative conditional search further demonstrated capabilities of MCGRAN to learn more fine-grained conditional distributions [229, Sec. 4.6] successfully.

## 14.5 RELATED WORK ON GRAPH GENERATORS

Without claim on completeness, we provide a simple overview of recent work in deep generative modeling for undirected and directed graphs in table 14.1.

| **Title** | Year | **Ref** |
|---|---|---|
| **Undirected Graphs** | | |
| GraphRNN | 2018 | [328] |
| DGMG | 2018 | [156] |
| Graphite | 2019 | [83] |
| DeepGG | 2021 | [271] |
| GRAN | 2019 | [160] |
| BiGG | 2020 | [44] |
| MoVAE: A Variational AutoEncoder for Molecular Graph Generation | 2023 | [164] |
| **Directed Acyclic Graphs** | | |
| Dags with no tears | 2018 | [336] |
| D-VAE var autoenc for DAGs | 2019 | [335] |
| DAG-GNN | 2019 | [330] |
| GraN-DAG | 2019 | [143] |
| LEAST | 2021 | [338] |
| Masked gradient-based causal structure learning | 2022 | [206] |
| DAG-DB | 2022 | [313] |
| Bayesian Structure Learning with Generative Flow Networks | 2022 | [47] |
| Differentiable DAG Sampling | 2021 | [36] |
| Learning Discrete Directed Acyclic Graphs via Backpropagation | 2022 | [313] |

**Table 14.1:** Related work on generative models for undirected and directed graphs. Auto-regressive models have not been the only advancements in graph generators. Diffussion-based models also gained much traction recently.

We can observe substantial progress for graph generative models in recent years and contextualised our own work in this development. The field can be considered mature enough to bring learned distributions $P(G)$ into neural architecture search applications without solving too many open questions concerned with details of graph generation. Controlling the learned distribution $P(G)$ might allow for structural regularisation of deep neural networks in future research.

## 14.6 GENERALISED CONCEPT OF DEEP STATE MACHINES

DGMG [156] and DeepGG [271] are deep state machines (DSMs). A deep state machine is a finite-state machine with learned transitions and an encoder-decoder component which translates between a structured memory and a differentiable representation space while switching between states. The model can be learned in a supervised manner by jointly minimizing predictions at different steps or by a reinforcement signal[2]. With a learned deep state machine and depending on the trained task, the starting memory can be randomly initialised or set to a specific content and the DSM subsequently applies a learned chain of operations on this memory. The two generative models DGMG and DeepGG use this principle to iteratively sample graphs starting from an empty one.

Like a finite-state machine in automata theory, a deep state machine consists of a finite set $S = \{s_1, \dots, s_{n_{states}}\}$ of $n_{states} \in \mathbb{N}$ states with an initial state $s_1 \in S$ and one or multiple final states $\{s_{\phi_1}, s_{\phi_2}, \dots\} \subset S \cap \{s_1\}$. The transition decision function $f_{s_\kappa}$ for each state $\kappa \in \{1, \dots, n_{states}\}$ is defined as $f_{s_\kappa} : \mathbb{R}^{H_{s_\kappa}} \to \{1, \dots, n_{states}\}$ for a state-specific memory representation dimension $H_{s_\kappa} \in \mathbb{N}$. Using a sequence representation $\{1, \dots, n_{states}\}$ for possible states emphasises that the decisions will be learned and drawn from a categorical distribution. This distribution is parameterised by a learnable sub-module DNN which will derive probabilities from a non-linear transformation of the used working topology.

The working topology can be a graph as for DGMG or DeepGG but it can also be text or any other structure which can be embedded into a learnable manifold such that the whole process can be integrated into an end-to-end learnable DNN model. Finite-state machines work on an input alphabet and this basically also applies for deep state machines except for that latter require means to encode the alphabet in manifolds.

Core of a deep state machine is to alternate between a working memory $M \in \mathcal{M}$ of a space $\mathcal{M}$ and its lower-dimensional representations $\mathbf{h} \in \mathbb{R}^{H_m}$. Instead of being a graph, $M$ could be a textual representation on which operations such as appending, deleting or editing are possible. For a graph these operations have previously been to add a vertex or an edge, but could also be to remove a vertex or an edge (or even triangles and more complex sub-graphs). We consider $n_{ops} \in \mathbb{N}$ operations on the working memory $O = \{o_i : \mathcal{M} \to \mathcal{M} \mid i \in \{1, \dots, n_{ops}\}\}$.

2 We did not conduct experiments with a deep state machine trained by reinforcement learning, but in theory the concept should be applicable.

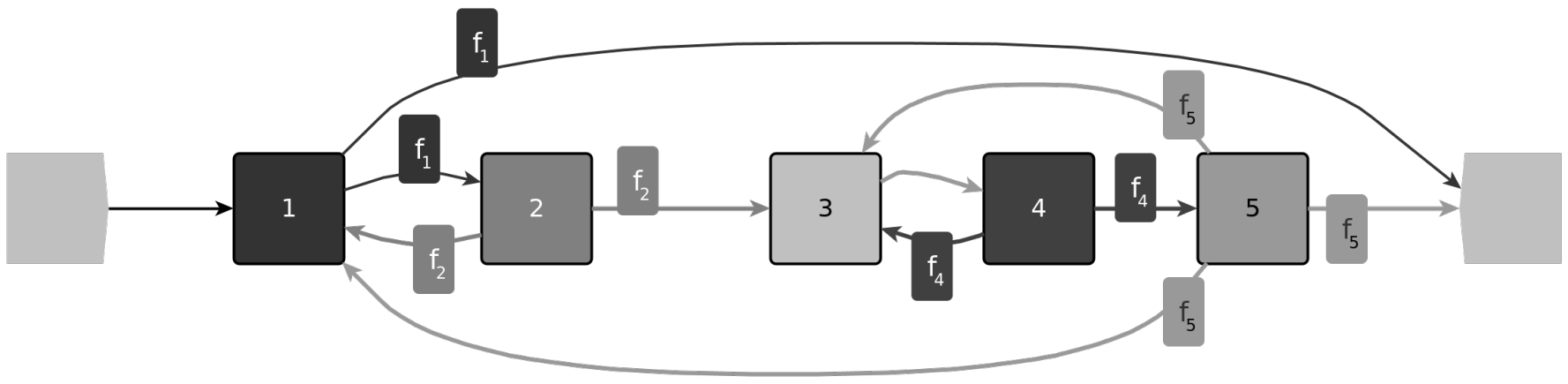

**Figure 14.5:** Example of a larger deep state machine over five states. Instead of having transitions between any two states, the control flow can be controlled by enforcing only distinct directions. The relationship to automata shows that it can be employed for tasks of various kinds. Generation is not a necessary application of the machine as it can be also used for classification tasks.

Turning working memory $M$ into memory representation $\mathbf{h} \in \mathbb{R}^{H_m}$ follows an encoder-decoder principle such that $f_{enc} : \mathcal{M} \to \mathbb{R}^{H_m}$ and $f_{dec} : \mathbb{R}^{H_m} \to \mathcal{M}$. A (variational) auto-encoder [128] is a straightforward example to use in such a case, although depending on the particular properties of the space $\mathcal{M}$, techniques such as graph neural networks may need to be considered.

In case $\mathcal{M}$ allows for fast sampling and a meaningful error $l_M$

- $f_{enc}$ and $f_{dec}$ can be efficiently pre-trained by minimzing $l_M(f_{dec}(f_{enc}(M_{sample})))$,
- and an operation $f_o : \mathbb{R}^{H_m} \to \mathbb{R}^{H_m}$ for $o \in O$ can be jointly pre-trained with $f_{enc}$ and $f_{dec}$ by computing $\hat{M}_{modified} = f_{dec}(f_o(f_{enc}(M_{sample})))$ and minimzing $l_M(M_{modified}, \hat{M}_{modified})$ for $M_{modified} = o(M_{sample})$.

An example for graphs is $M_{sample} \sim GIL(n = 30; p = 0.5)$ (using the Erdős-Rényi model) and $o$ being an operation of adding a fixed number of edges into the graph and $M_{modified}$ being its result. The encoder $f_{enc}$ and decoder $f_{dev}$ then work based on graph neural networks and $f_o$ is a deep neural network operating in the memory representation space $\mathbb{R}^{H_m}$. Using the graph edit distance as error $l_M$ is extensive such that the MMD of selected graph properties could be considered.

An example for language is a natural text $M_{modified}$ which is randomly perturbated on a character-level up to a certain percentage such that $M_{sample}$ is obtained. The encoder $f_{enc}$ and decoder $f_{dev}$ then work based on word embeddings (and possibly attention [293]) and $f_o$ is a deep neural network operating in the memory representation space $\mathbb{R}^{H_m}$, turning a perturbated text $M_{sample}$ (close) to its original version $M_{modified}$. An error $l_M$ could be based on a character-level distance.

We are aware that the details of $f_{enc}, f_{dec}$ and $f_o$ are a complex topic on its own but require their formal description as to integrate multiple operations into a deep state machine. The term *pre-training* here refers to the possibility to find optimal parameters for the sub-modules of the deep state machine by defining tasks that are not directly related to the task of the DSM. Alternatively, pre-trained models can also be incorporated instead of defining a pre-training task.

A state $s_\kappa$ performs a change on the working memory $M$, or its lower-dimensional representations $H$, or both. Subsequently the state uses $f_{s_\kappa}$ to decide about the next state to transition into. Therefore, the memory representation for the initial state $s_1$ is denoted with $\mathbf{h}^{(1)}$ and becomes $\mathbf{h}^{(t+1)}$ after each transition to a new state over in total $n_{trans} \in \mathbb{N}$ transitions.

(I) **Encode Memory**:
$\mathbf{h}_{s_\kappa}^{(t)} = f_{s_\kappa}^{read}(f_{enc}(M^{(t)})))$ or $\mathbf{h}_{s_\kappa}^{(t)} = f_{s_\kappa}^{read}(\mathbf{h}^{(t)})$
or $\mathbf{h}_{s_\kappa}^{(t)} = f_{enc}(M^{(t)})$, or $\mathbf{h}_{s_\kappa}^{(t)} = \mathbf{h}^{(t)}$ if $\nexists f_{s_\kappa}^{read}$

(II) **Apply Operation**:
$\mathbf{h}_{s_\kappa}^{(t)} = f_o(\mathbf{h}_{s_\kappa}^{(t)})$ if $o \in O_{s_\kappa} \subset O$

(III) **Update Memory**:
$\mathbf{h}^{(t+1)} = f_{s_\kappa}^{up}(\mathbf{h}_{s_\kappa}^{(t)})$ if $\exists f_{s_\kappa^{up}}$
or $\mathbf{h}^{(t+1)} = f_{up}(\mathbf{h}^{(t)}, f_{dec}(\mathbf{h}_{s_\kappa}^{(t)}))$

(IV) **State Transition**:
$\kappa = f_{s_\kappa}(\mathbf{h}_{s_\kappa}^{(t)})$

A state can carry an additional sub-module $f_{s_\kappa}^{read} : \mathbb{R}^{H_m} \to \mathbb{R}^{H_{s_\kappa}}$ if $H_m \neq H_{s_\kappa}$ to map the memory representation $\mathbf{h}^{(t)}$ of the current step $t$ into a state-wise memory representation $\mathbf{h}_\kappa^{(t)}$. Otherwise all neural networks operate in $\mathbb{R}^{H_m}$. Each state can further carry an updating sub-module $f_{s_\kappa}^{up} : \mathbb{R}^{H_{s_\kappa}} \to \mathbb{R}^{H_m}$ to modify the overall memory representation $\mathbf{h}^{(t+1)}$ after the states' $f_o$ operation has been applied. Alternatively, a state-independent updating sub-module $f^{up} : \mathbb{R}^{H_m} \to \mathbb{R}^{H_m}$ can be used if e.g. step-wise information from a memory $M^{(t)}$ is present.

Learning a deep state machines in supervised fashion is done by either minimizing the joint log loss of state transition predictions or by minimzing the loss of a final decoded memory representation. The particular choice of learning depends on available data. For DeepGG we previously described in section 14.1 on page 262 that a sequential representation of a graph $(a^{(1)}, \dots, a^{(n_{const})})$ can be used where $n_{const} \in \mathbb{N}$ and $a^{(\cdot)} \in \{0, 1\}$. The sequence then contains expected state transitions such that the learning is based on the joint log loss of $l(f_{s_\kappa}(\mathbf{h}_{s_\kappa}^{(t)}), a^{(t)})$. In that case, the number of construction steps aligned with the number of state transitions of the DSM: $n_{const} = n_{trans}$. Alternatively, for a given expected $M_{final}$ the DSM can also be learned by minimizing e.g. $l(M_{final}, f_{dec}(\mathbf{h}^{n_{trans}}))$, e.g. a resulting graph $M_{final}$ after $n_{trans}$ steps.

Deep state machines are auto-regressive models and can be considered a type of recurrent DNN. Significant characteristics in comparison to recurrent neural networks are **1st/** more state-explicit gatings and **2nd/** encoding and decoding between $M$ and $\mathbb{R}^{H_m}$ during each step. We leave further differences for future research.

The disadvantage of deep state machines is bringing manual modeling back into automatic feature learning of DNNs. This requires again expert knowledge. Further, DSMs suffer from the same problems as recurrent neural networks in controlling gradient-flow and learning long-term dependencies. An advantage of DSMs is to explicitly model possible operations in the desired working structure $\mathcal{M}$ and allowing only for certain transitions between states by excluding some transitions from a fully connected state automaton. We transferred this principle to SSH Key extraction from heap memory [65] and while it did not outperform simpler methods, the results were promising to learn models that align with program flows.

REMARKS

- Deep state machines are recurrent neural networks over discrete states of a finite-state machine.
- The principle is deduced from working models such as DeepGG and offers control over sequences of learned operations.
- The auto-regressive nature of DSMs barely fits into the formulation of graph-induced neural networks but if also recurrent models are considered, new perspectives on what constitutes structures open up [269].
- Relationships to similar models such as Neural Turing Machines [81] are yet unclear and open new research directions.

## 14.7 GRAPH ASSEMBLY SEQUENCES

Up to this point, chapter 14 is concerned with learning graph generators $P(G)$ based on deep learning methods and exemplary graphs. For this purpose of learning $P(G)$, we focussed on deep learning models that capture sequential properties and considered sequential graph representations as described in section 14.1. The analysis of both models and data representation led us to rethink the graph representation.

We observe the following issues with experiments on graph-induced neural networks and with the considered approaches to capture a distribution $P(G)$:

- The size of a set of graphs $G$ grows so fast that we can only experiment with $\mathcal{A}(G)$ of small order. Emergent properties of analytically captured models such as the Erdős-Rényi model or the Barabasi-Albert model for $g \sim GIL(n; p)$ become only apparent with e.g. $|g| \gg 100$ for which many learned models struggle.

- Properties of $G$ are difficult to analyse as we realised in structure analyses in chapter 9 on page 155 and chapter 10 on page 181.

This lets us question the nature of operation underlying the sequential representation. If we consider a graph construction sequence such as [

NNE10] as for GraphRNN, DGMG or DeepGG or such as $L^{\pi}_{\mathbf{b}_i} \in \mathbb{R}^{BN}$ in section 14.3 as for GRAN, we notice that the tensor shape per time step changes among different representations but the representation sticks to binary values. The binary values represent either whether a basic graph operation should be executed and which two vertices should be connected in GraphRNN/DGMG/DeepGG or whether an edge exists in a block of the overall adjacency matrix in GRAN. The first representation differs significantly from the second in that it is not directly related to the adjacency matrix representation.

But the first representation has a stronger relationship to the sequential nature of models such as the Barabasi-Albert model, which we related to the observed bias of the DGMG model [271]. And the sequential nature of the Barabasi-Albert model is insofar interesting in that it can be interpreted as a process which leads to a graph (or network) exhibiting the well-known nature of a scale-free degree distribution. Observing natural processes, however, does not only show elementary growth in networks but also more complex operations, even including shrinkage.

We therefore propose to consider *graph assembly representations* to study both graphs or networks and their assemblic process. A graph assembly representation is therefore not necessarily a representation of a single graph but a representation of its explicit underlying assemblic process. This assumes the existance of such an assemblic process.

Let $o : \mathcal{G} \to \mathcal{P}(\mathcal{G})$ be a *graph operation* transforming a given graph $g_1 \in \mathcal{G}$ into new graphs $G_2 \subseteq \mathcal{G}$ over a space of all graphs $\mathcal{G}$. A realisation of the operation $\hat{o} : \mathcal{G} \to \mathcal{G}$ transforms graph $g_1 \in \mathcal{G}$ into a new graph $g_2 \in G_2 \subseteq \mathcal{G}$ with $g_2 = \hat{o}(g_1)$.

An example is $o_{add}$ with $g = (V_g, E_g)$ and $v_{new} \in V_g \cap \mathbb{N}$ such that

$$o_{add} \triangleq (V_g, E_g) \mapsto \{ \; (V_g \cup \{v_{new}\} \\ E_g \cup \{sv_{new}\}) \;\mid \forall s \in V_g\}_{\simeq}$$

which results in the set of all graphs of order $|V_g| + 1$ of connected graphs based on $g$ and with one new vertex. With $\{\cdot\}_{\simeq}$ we refer to the set of only non-isomorphic graphs w.r.t. to the isomorphism refered to by $\simeq$.

Another example would be $o_{triangle}$ with $g = (V_g, E_g)$ and $v_1, v_2 \in V_g \cap \mathbb{N}$ and $v_1 \neq v_2$ such that

$$o_{triangle} \triangleq (V_g, E_g) \mapsto \{ \; (V_g \cup \{v_1, v_2\} \\ E_g \cup \{v_1 v_2, s v_1, s v_2\}) \;\mid \forall s \in V_g\}_{\simeq}$$

which results in the set of all graphs of order $|V_g| + 2$ of connected graphs with growing triangles based on $g$ (not being the empty graph).

A graph assembling sequence is a sequence $(o_i)_{i\in\mathbb{N}}$ of graph operations. Applied to an initial start value $G_0 \in \mathcal{G}$, a graph assembling sequence $(o_i)_{i\in\mathbb{N}}$ results in a sequence of sets of graphs $(o_i(G_{i-1}))_{i\in\mathbb{N}}$.

The set of all resulting graphs from a graph assembling sequence is the union over all sets, i.e. $\bigcup_{i \in \mathbb{N}} o_i(G_{i-1})$ and the set of assembled graphs up to the assemblence step $t \in \mathbb{N}$ is $G_t = o_t(G_{t-1})$.

With each recursion step $G_t$ grows super-linearly w.r.t. its cardinality. Also, the order of assembled graphs can grow super-linearly. More complex graph operations can restrict this fast growth with more complex operations or a balance between different operations of growth and shrinkage. The example of $o_{triangle}$ grows with two vertices per step but the cardinality of the assembled $G_t$ grows very fast because the number of possible vertices to which the triangle can be attached grows with each step.

Two approaches reduce this growth easily: **1st/** using global information of a graph, an operation could e.g. only attach new vertices to existing vertices of high or low degree. This would then directly control the shape of the degree distribution, similar as in the Barabasi-Albert model. **2nd/** using probabilistic methods, one could allow growth with only reduced probability. This turns the set $G_t$ into a random variable and the overall sequence into a random process and aligns even stronger with random graph models such as the Erdős-Rényi model etc.

Instead of learning from the set of graphs $G_{train}$ blindly and following a generic model, learning from a perspective of graph assembly sequences offers alternatives to learn a distribution $P(G)$. Let $O$ be a finite set of operations.

- Determining a single graph operation $o \in O$ for assemblance is the explanation for observed graphs.
- Selecting a subset $O_{solution} \subseteq O$ is the explanation for the given observed graphs.
- A vector of prior probabilities for a categorical choice $\text{Cat}(p_i)$, i.e. a sequence $(p_i)_{i \in \mathbb{N}}$ is the explanation for the given observed graphs.

Given two operations $o_1, o_2 \in O$ a graph assembly representation for a fixed number of steps could be captured as e.g. $\begin{pmatrix} 1 & 1 & 1 & 0 & 0 & 0 \\ 0 & 0 & 0 & 1 & 1 & 1 \end{pmatrix}$ with a one-hot encoding. The sequence would then assemble a graph with repeating $o_1$ three times, followed by $o_2$ three times. An issue with this representation is that it is again of fixed-length – which is good for fixed-length input to deep neural networks but it could have issues to represent assembled graphs over very different orders. The one-hot encoding of a graph assembly representation aligns perfectly with the working of a deep state machine in that a single operation needs to be chosen auto-regressively based on the currently assembled graph.

The assemblic perspective gives rise to the following questions:

1. How does the asymptotic growth of $|G_t|$ behave? This question is of interest because if there exist graph assembling sequences with sub-exponential growth behaviour, the assembled graphs might be of certain interest. Combinatorial explosion could be drastically reduced in these subsets of $\mathcal{P}(\mathcal{G})$

2. Does a probabilistic perspective on graph operations result in proper formulations in random graph theory? Taking random choices during graph assemblance makes random graph models such as the Erdős-Rényi model, the Watts-Strogatz model, or the Barabasi-Albert model instances of graph assembly sequences. As discussed in section 3.2 on page 30 it is not trivial to properly capture the sample space of random graph models. Similar to common descriptions of models such as the Barabasi-Albert model, graph assembly sequences pose a simpler notation than the probabilistic definition of a random graph model because the difficult sample space does not explicitly be modeled. Also, the resulting graphs realisations themselves are of larger interest than the probability of single edges across graphs of varying order.

In the future, we plan to carry out experiments that will provide answers to these questions. Two related fields are worthy mentioning: the similarity of (sets of) graphs and models of graphs with emerging properties.

The notion of similarity for graph assemblic sequences could be re-defined in comparison to the graph edit distance [248]. If graphs emerge from underlying assemblic operations, two graphs $g_1$ and $g_2$ might be close w.r.t. the GED but only one graph such as $g_1$ might be contained in the assembled graph set. The other graph $g_2$ might require different operations or might be less likely to appear under the assemblic sequence. A notion of similarity could be then based on the similarity of the operation set $O$ or a representation of the assembly sequence. Such a similarity is important for further learning assembly sequences with statistical distance minimisation.

Wolfram [309] proposes to study models with "rules for rewriting collections of relations" and their emerging structures for which he mentions that "one convenient way to represent such structures is as graphs" [309, Sec. 2.1, p. 3]. This is in theory very similar to graph assembly sequences although we considered the latter only as to learn generative models of graphs and in application to neural architecture search. There are significant differences between Wolframs studies and our ideas:

- Wolfram captures rules with graphical depictions and restricts them by rewriting collections of relations. Graph operations are loosely defined over the set of all graphs and different representations can be chosen. Perhaps the difference is insignificant.

- Graph assembly could contain varying operations which change over time and we have been interested in auto-regressive formulations of graph assembly sequences, i.e. one operation could act early in the assemblance and another one late.

- Graph operations can act on both local and global information. Per step, a vertex might be added only to a fraction of vertices of the existing graph with very high or low degree. This requires information about the overall graph degree distribution and can not be captured by rules used by Wolfram.

We suppose that learning representations of assembly sequences could have application beyond NAS such as in learning chemical reaction pathways [183, 259, 275, 287] like graph generative models often find application not in NAS but in learning models for molecular discovery.

REMARKS

- Our work is limited by the size of graphs we can explore and motivates the study of emergent properties of new random graph models.
- We studied sequences of graph operations to assemble graphs and motivated theoretical and practical questions associated with it.
- Reducing the problem of learning a graph generator to assemblic operations promises speed-up if assembled graphs gear towards desired emergent properties.
- Straightforward representations such as a one-hot encoding of a graph assembly sequence align with the concept of deep state machines which motivates further studies in the theory of graph assemblance and DSMs.

## 14.8 SUMMARY ON GRAPH GENERATORS FOR NAS

We have studied several deep neural networks for generating graphs and contributed in applying them in context of neural architecture search. Learning and controlling graph generators poses challenging problems in itself. Theoretical challenges can be found in the asymptotic growth complexity of graphs as mentioned in section 3.4 and the graph isomorphism problem is also related to it. Practical challenges include the formulation of proper representations that allow for sufficient flexibility of graphs of varying order, the reduction of auto-regressive steps to reduce long-term dependencies of products of conditional distributions, the parallelisability of training with many samples and many more. However, scalable and efficient graph generators are more accessible than five years ago as provided in table 14.1 such that the usage of learning $P(G)$ and sampling from it in context of neural architecture search opens new applications.

The study on graph generative models further fostered developments on concepts of deep state machines in section 14.6 and graph assembly sequences in section 14.7. Both shed light on connections to related research in material or chemical sciences, open new research directions and go beyond structure analyses of DNNs or studies on neural architecture search [252].

With a proper search space definition, $P(G)$ can be used to draw more sophisticated neural architectures. Further, a graph generator might be also integrated into multi-objective problems in which trade-offs between certain structural properties and performance objectives are desired. An obvious example is the structural property to be able to distribute an architecture effectively while not diminishing in performance. But also privacy concerns could potentially be structurally reflected in modern federated learning settings [123, 190].

# Part VII

# OVERARCHING DISCUSSIONS & CONCLUSIONS

Structure arises in different forms as a by-product of discretisation or as a first-class citizen as an imposed model prior. What role does structure play when it comes to evaluation measures of neural network realisations? For automated machine learning in the form of neural design, structure plays an immersive role in guiding optimisation across different neural architectures and holds the potential for new perspectives on neural interpretability.

# 15 OVERARCHING DISCUSSIONS & CONCLUSIONS

*The following entails:*

We formally introduced graph-induced neural networks as to pose our research questions on the structure of deep neural networks (DNN) in a unified language. In this language, neural network realisations $f \in \mathcal{A}(\mathcal{G})$ from universal architectures $\mathcal{A}(\mathcal{G})$ are induced over structural themes $T$ such that $\mathcal{G} = [T]$ or induced probabilistically with a distribution of graphs $P(G)$ through $\mathcal{G} \sim P(G)$. The search space of all considered architectures is referred to as $\mathbb{S}$. Our research is biologically motivated and originally led to random graph models as search space design [272]. But we also investigate on alternative design spaces such as in conjunction with NAS benchmarks or resulting from our formal considerations on CTs (section 7.2). With graph-induced neural networks on various design spaces, we conduct *structure analyses* (part iv) of $T$ or $\mathcal{G}$ on the one hand and *neural architecture search* (part v & part vi) on the other hand.

Structure analyses comprise questions on whether structural themes $T_1$ and $T_2$ with significantly different structural properties exhibit differences such as in classification accuracy. Observing a difference such as $\alpha(f_1) > \alpha(f_1)$ for $f_1 \in \mathcal{A}([T_1])$ and $f_2 \in \mathcal{A}([T_2])$, can this be properly explained by $T_1$ and $T_2$? We found, that it makes a difference but that the difference is currently hardly explained by structural properties except for some indications of variances of their degree or path length distributions.

Neural architecture search comprises questions on how $T \in \mathbb{S}$ or $f \in \mathcal{A}([T])$ with desirable properties can be found effectively and efficiently with varying methods, how such NAS methods differ and how they can be improved. We found, that there exist generic methods such as evolutionary searches which work out of the box but also very specialised methods which impose topological properties on $\mathbb{S}$ such as DARTS. NAS methods mostly differ in their required properties of $\mathbb{S}$ and how they exploit these properties. Advancements are accomplished by introducing techniques of predictive estimation strategies or generative models for the search space $\mathbb{S}$.

Our contributions comprise formal considerations, large-scale experiments, and methodological advancements as summarised in table 2.1 on page 20 in section 2.2. We recap these contributions in the following in section 15.2 and reflect critically on formulations and experimental results in a conclusive discussion in section 15.3. The work culminates in multiple open paths summarised in section 15.4.

## 15.1 GUIDING RESEARCH QUESTIONS

Central topic of this work is the relationship between deep neural networks and their structure. For that purpose, the work is organised in three research complexes, first on the formalisation of structure of deep neural networks in section 1.1.4, second on the analysis of such structures of trained deep neural networks in section 1.1.5, and third on the automation of discovery and construction of neural networks with desirable structure in section 1.1.6.

Concerning the first research complex, we asked:

> ◇ Question I in section 1.1.4 (Formalism)
>
> How can the structure of deep neural networks be defined such that structural optimisation problems can be formulated?

and provided a construction of graph-induced neural networks in section 6.1 on page 102. This construction results in the notation of universal architectures $\mathcal{A}(\mathcal{G})$ for structural themes $T$ with $\mathcal{G} = [T]$ or random graph priors $\mathcal{G} \sim GIL(n;p)$ and allowed to formulate the training of a neural network as

$$\mathcal{F} \triangleq \underset{f \in \mathcal{A}([T])^{d_1 \to d_2}}{\arg\min} \mathcal{L}_{train}(f, D_{train}) \quad (7.1)$$

and the discovery of an architecture i.e. as a structural theme $T$ as

$$\underset{T \in \mathbb{S}}{\arg\min}\, \mathcal{L}_{val}(T, \underset{f \in \mathcal{A}([T])^{d_1 \to d_2}}{\arg\min} \mathcal{L}_{train}(f, D_{train})) \quad (7.2)$$

An example for the existance of graph-induced neural networks is outlined with computational themes. A proof of universality of such networks for a simple case is exemplified in section 6.3.2 on page 112 by showing that there exist graph-induced neural networks which collapse to the definitions of Hornik [106] and Cybenko [43].

Recall the second question of the research complex on the formalisation of structure in section 1.1.4:

> ◇ Question II in section 1.1.4 (Formalism)
>
> Are graphs a suitable representation for neural network structure?

This question was discussed qualitatively in section 6.4 on page 114 and we found the usage of directed acyclic graphs (DAGs) to quite naturally align with the construction of graph-induced neural networks but also sufficiently suitable for considering advanced methods such as generative models for learning to generate DAGs.

The third question in section 1.1.4:

> ⋄ Question III in section 1.1.4 (Formalism)
>
> What are the difficulties, advantages and disadvantages of the proposed formalism?

was kept rather open and we mentioned several disadvantages and difficulties of graph-induced neural networks based on directed acyclic graphs: 1st/ the complexity of the combinatorial space of DAGs is difficult to capture for analysis or sampling, 2nd/ the network theoretic properties of analysed structures only showed weak to moderate correlation as to question whether there even exist differences empirically, and 3rd/ the overall formulation of graph-induced architectures is restricted to deep neural networks that hardly capture permutation invariances during computational forward passes.

In the second research complex, we continued with studying deep neural networks based on different notions of structures. Our insights are summarised in section 10.4 on page 200 and section 9.8 on page 178. We first asked:

> ⋄⋄ Question I in section 1.1.5 (Structure Analysis)
>
> Do neural network models differ in terms of structure or are they just different with respect to the choice of other hyperparameters such as activations, training epochs, data sampling or augmentation strategy, optimisation procedure, or the choice of loss function?

We found that neural networks actually differ heavily in terms of structure. This can be observed across many experiments such as in

- section 10.1 on page 181 with 831 computational themes with over 287,265 trained neural networks,
- or in section 9.2 on page 160 with over 80,381 trained neural networks of varying layer depths,
- or in section 9.5 on page 167 with over 200,000 trained neural networks based on random graphs from the Barabasi-Albert model and the Watts-Strogatz model.

A more precise question and the central question of our work was therefore whether and which structural properties affect or explain observed

differences:

◇◇ Question II in section 1.1.5 (Structure Analysis)

Are well studied network theoretic properties involved in the behaviour of neural networks? What are common structures or patterns that can be observed?

While the appearance of a structural effect appears very significant, we argued in section 6.1 on page 102 that the structure already has implications on a formal level: how a DAG transforms into a DNN can be optimised and might carry sparsely connected weight matrices. The choice on how to group computation paths in matrix multiplications or across layers has impact on memory consumption and parallel execution on hardware.

Our experiments showed correlations for properties based on the degree and the path lengths of $G$. Interestingly and in accordance with independently conducted experiments [6, 104, 269] variances of the degree or average path length distribution recurringly exhibit correlation with correctness measures such as accuracy or $F_1$ score, robustness, or energy consumption. However, these experiments did not settle the question about structural patterns independently of specific data and motivate to

- create alternative experiment designs to investigate the question,
- consider other graph theoretic properties,
- search for structural patterns within versus across certain data domains (i.e. *images* vs. *text*),
- and to analytically investigate whether the dense computation described in eq. (6.3) on page 105 resembles different approximation errors with $\mathcal{A}([T_1])$ compared to $\mathcal{A}([T_2])$ where $T_1$ and $T_2$ exhibit significantly different properties (possibly motivated by our experimental findings).

Similar to research complex I, we kept a broader question on the implications of insights of structure analytical experiments:

◇◇ Question III in section 1.1.5 (Structure Analysis)

How could analytical results look like to guide further architecture development?

Both, the formalisation of the problems with graph-induced neural networks as proposed in chapter 6 on page 101 and the empirical investigations in chapter 9 and chapter 10 led to the following insights on how results for further NAS development could look like:

- NAS-methods (not AutoML in general) only makes sense if the structure that can be differentiated by these methods makes a difference w.r.t. the proposed objectives of correctness, robustness, or energy measures. We suppose that analytical results for structural themes $T_1$ and $T_2$ with different approximation rates could strengthen the right of existance of NAS methods. A negative result as in showing an equivalence in the approximation rates across different themes would shift the question about empirical differences entirely. NAS could shift away from a focus on structure to e.g. operational diversity or composition.
- Further insights in beneficial structural properties of $G$ could be brought into biasing search methods towards that properties and by this restrict the search space induced by such structures.
- Studies on $\mathcal{A}(\cdot)$ with combinatorial results could give a tool to compare NAS-methods in how fast or effectively they can capture a search space by restricting or enhancing it to certain structures.

With both the formal and empirical questions settled, we turned our focus towards NAS-methods and asked:

> ⋄ ⋄ ⋄ Question I in section 1.1.6 (Automation)
>
> Which methods for neural architecture search exist, how can they be compared and what are their differences?

We found the variety of possible methods for NAS too exhaustive such that we focussed only on pruning, evolution, genetic algorithms, and differentiable architecture search and studied their assumptions on training a neural network as in eq. (7.1) on page 122 or finding structural themes in a search space $\mathbb{S}$ in eq. (7.2) on page 122.

The NAS methods differ in assumptions on properties of the search space $\mathbb{S}$ and how they navigate through $\mathbb{S}$. Pruning assumes an initial large structure from which single elements can be removed successively and is applicable during both training and NAS. We contribute on a Shapley Value-based pruning method and analyse its premises in section 11.4 on page 214. Empirically, we found, however, that the choice of pruning method during training appears rather insignificant for high-parameterised deep neural networks [269].

Evolutionary and genetic algorithms are very generic and simple to implement without having many restrictions to $\mathbb{S}$ except for being able to (locally) sample from it. We found the dependence of genetic algorithms on particularly chosen variation operators, however, rather small in comparison to the driving force of selection, as presented in section 12.5 on page 228.

DARTS relaxes the optimisation problem of eq. (7.2) into a differentiable form by making a sacrifice of restricting the architectural space to a finite-parameterised space but with the benefit of using gradient-based search.

With a deeper understanding of existing NAS methods we then asked:

> ⋄ ⋄ ⋄ Question II in section 1.1.6 (Automation)
>
> With knowledge on structure from research complex II, how can neural architecture search methods be improved or guided?

and we found two guiding principles to enhance NAS methods: **1st/** accelerating the performance estimation strategy by predictive models as surrogate for expensive training & evaluation schemes during neural architecture search and **2nd/** conducting informed sampling from the search space $\mathbb{S}$ by generative models for search spaces based on graphs.

For the predictive principle we present a simple method of predicting the architecture candidate performance based on structural features in section 13.1 on page 250 and found a random forest to perform best. We then investigat on a new approach called FaDE in section 13.2 on page 252. FaDE obtains scores which we use as surrogate for expensive architecture evaluations. The idea is based on the fact that DARTS restricts the search space to a finite one (w.r.t. architecture candidates) and FaDE evaluates different architecture candidates in a hyper-architecture to obtain FaDE-scores which are used for a pseudo-gradient optimisation step in an open search space. We consider the method as computationally less expensive than e.g. using evolutionary searches with full training & evaluation schemes and is compared to bayesian optimisation and random search in [204].

For the generative principle we motivate to learn generative models for graphs $P(G)$. We extensively analyse DGMG [271] and find that DGMG is biased towards scale-free degree graphs because of its underlying state machine which, when traversed randomly, directly produces graphs in a manner like the Barabasi-Albert model. In a further study we compare GraphRNN and GRAN [282] and find reasons for why GRAN is superior to GraphRNN: not only the used node-ordering for the representation has an impact as often mentioned [160, 328] but also the overall auto-regressive style of predicting one vertex and edge after another is not as beneficial as a block-wise generation of vertices. Therefore, in a first step, we extend the state-based model DGMG to a model DeepGG with a different underlying state machine in section 14.2 on page 264.

The extension of DGMG to DeepGG further leads to a concept of deep state machines in section 14.6 on page 273. However, the DeepGG-model is not comparable with scaling capabilities of GRAN. To learn an efficient generative model for NAS architecture candidates (as graphs), we extend GRAN with means to conditionally learn graphs and also label them. The resulting MCGRAN [229] successfully learns based on e.g. low-performing and high-performing architectures from the `NAS-Bench-101` and demonstrates to provide a meaningful generative models for graphs $P(G)$ in application to NAS.

The two principles, predictive and generative, for improving or guiding NAS methods can be combined in i.e. evolutionary neural architecture search.

REMARKS

- Our answers on the first research complex about formalism of structure in deep neural networks reduces the gap between applied NAS and learning theory.
- The second complex about structure analysis reveals the complexity of empirical experiments in this young research field, gives hints on first structural properties and shows current limitations of finding strong evidence about universally applicable structures.
- With new methods for the third research complex about automating NAS, we contribute examples on how to advance the field of automated methods.
- All three research complexes revealed new questions and motivate future studies on the topic of structures of deep neural networks.

## 15.2 CONTRIBUTIONS OF THIS WORK

In section 2.2 on page 17 we provide an overview of our contributions. In the following, we give a more explicit list of contributions with a brief statement:

1. In chapter 6 on page 101 we formally define graph-induced neural networks to capture how we understand the structure of deep neural networks and make structure a first-class citisen of DNNs instead of having it implicitly arise in previous definitions.

2. A constructive proof for instances of graph-induced neural networks being universal architectures is provided in section 6.3.2 on page 112, showing the existance of such architectures. The proof shows that with skip-layer connections resembling the identity function, graph-induced nets collapse into definitions of Cybenko [43] & Hornik [106], making them a super-set of the latter.

3. We qualitatively discuss in section 6.4 on page 114 alternative formulations of DNNs and assess how much they align with a NAS-perspective.

4. An algorithm for generating artificial low-dimensional data sets of varying complexity is presented in section 8.2 on page 142 and subsequently used in empirical experiments.

5. A large-scale CT-NAS database with over 287,265 trained models is presented in section 8.3 on page 148 and subsequently used for evaluating different aspects of neural architecture search methods.

6. With 80381 trained neural network realisations based on layer-wise structural themes, a structure analysis is presented in section 9.2 on page 160. The experiment is a strong example for confounders (in this case the bottleneck of a network) where a structural effect needs to be carefully analysed.

7. Random initialised graph-induced neural networks are discussed in section 9.4 on page 165 and it is empirically shown that they pose very simple (or flat) function landscapes when common and small initialisations are chosen.

8. Section 9.5 on page 167 discusses the correlation between random structures and correctness measures. In parallel to Xie et al. [319] we have been among the first to use random (or complex) networks to construct deep neural networks [272].

9. Shapley Value-based pruning is extended to NAS and statements for an analysis of non-additive payoffs are provided. We provide a new lower bound analysis for non-additive payoffs in section 11.4 on page 214.

10. The application to NAS led us to extend generative graph models to new models called DeepGG and MCGRAN which have been empirically tested and discussed in chapter 14 on page 261.

11. Base implementations for sparse neural networks and constructing DNNs based on graph structures are provided in LIB DeepStruct and have been earlier discussed in [273]. Technical details are added in appendix A.5 on page 315.

12. A new analysis of variation operators for genetic NAS is presented in section 12.5 on page 228. The results align with statements of the authors of `NAS-Bench-101` and suggest that the choice of graph-based variation operator is not very significant, although clear differences can be observed. A conclusion from the analysis is to possibly learn variation operators for NAS in future in opposite to manually design them.

13. A predictive NAS method FaDE is presented in section 13.2 on page 252 and put into context of the presented optimisation problems for NAS. FaDE is rather unusual in using scores from a hyper-architecture to evaluate an architecture candidate in an open NAS and opens new interesting research opportunities with unusual optimisation techniques.

14. The presented generative methods for NAS motivate deep state machines and graph assembly sequences for which sketches are presented in section 14.6 on page 273 and section 14.7 on page 276.

## 15.3 OVERARCHING DISCUSSION

In one blunt statement, the structure of deep neural networks shows to have a diminishing effect in comparison to other hyperparameters. Our experiments suggest, that there might exist few universal structural properties that seem always beneficial and there definitely exist domain-dependent structural effects.

With chapter 9 on page 155 in mind, it is first and foremost important to clarify which objective should be considered. The very clear effect of increasing the number of parameters and observing an increased energy consumption aligns with a correlation between the parameters and correctness measures. But for very large parameter numbers, the energy consumption will continue to increase[1] while the correctness usually converges at some point. Otherwise we would only need to increase the parameter count to achieve better performances, which empirically is not actually the case: changing operations, reformulating problem definitions and reconsidering many other parts of the whole pipeline are essential factors to achieve high performing deep neural networks. Clarifying the objective allows to determine the effect of a particular structural pattern.

Structure is an implicit resident of deep neural networks: a layering naturally arises by having non-linear activation functions. More depth then allows each layer to be locally very linear and reduces the necessity of wide layers that memorise information. Making structure beyond this deepening more explicit, however, is still not known to guarantee any benefits in approximation capabilities. The question, whether linear depth can be compensated with e.g. logarithmic depth plus skip-layer connections is as of our knowledge currently unknown. In other words, whether structural paths make a difference compared to deepening a network. Formally, however, we can see in section 6.1 on page 102, that graph-induced neural networks grow with $\frac{L(L+1)}{2}$ matrix multiplications instead of $L$ layers as in classical formulations of DNNs. Introducing skip-layer connections and thus allowing for more structural properties as for structure analysis in chapter 9 on page 155 implies a larger memory consumption, especially with the technical condition that connected layers require same-sised shapes. This either makes the technical implementation very large and sparse or forces it to compute on a low level of neurons. It is not finally settled whether

1 The energy consumption increases when the same structural pattern is used. There are certainly ways to reduce the growth of energy consumption by exploiting i.e. structured sparsity through hardware-aware design.

making structure explicit is actually beneficial or under which trade-offs it is beneficial.

*Limitations of Graph-Induced Neural Nets*

Making structure explicit leads us to the formulation of graph-induced neural networks. Alternatives to this form are possible and discussed in section 6.4 on page 114. We found the formulation based on directed acyclic graphs (DAGs) useful for considering our research questions and use the term structural themes to refer to DAGs which induce universal architectures. The structural discretisation of latent spaces of DNNs takes effect by employing activation functions between affine transformations and motivates us to consider computational themes (CTs) as structural themes: to provide sufficient parametric capacity to align with UAPs, the structural theme needs to allow for such growth. Computational themes provide this property and still present distinct equivalence classes. This definition might be helpful for future analysis and we empirically could observe clear differences of these CTs.

The formulation of graph-induced neural networks is, however, limited: it only captures classical feed-forward neural networks and the architectural construction $\mathcal{A}(\mathcal{G})$ as defined in definition 1 on page 108, based on algorithm 4, hides many details between varying definitions of how to come up with a set of DAGs $\mathcal{G}$. Further, many properties of topological input spaces might not be reflected: the formulation is hardly made invariant to input permutations. This limitation will require to re-think the construction of graph-induced networks, currently making use of fixed orderings of neurons based on permutations.

While the empirically found structural effects such as variances of the degree distribution, layer bottlenecks, or variances of average path length distributions are small, they promise some stronger results in future. There exists a good chance that the scale on which the experiments have been conducted is not sufficient to exhibit emergent structural effects, similarly as it is the case for complex networks such as the Erdős-Rényi model, the Watts-Strogatz model, or the Barabasi-Albert model. If structure on any scale would make no difference, the problems and methods of neural architecture search would be rendered obsolete or at least appear in a different light.

*Other Limitations*

Reducing the consideration of structure of graph-induced neural networks solely to directed acyclic graph motivates to study problems in geometric deep learning, network theory, topological graph theory, and combinatorics. For example sampling large graphs has a combinatorial background. Further, considering prior structures $\mathcal{G}$ by equivalence classes over DAGs or by generative models $\hat{\mathcal{G}} \sim P(\mathcal{G})$ is more explicitly studied with large graph families in graph theory.

Empirical NAS studies on the other hand are and will stay computationally expensive. Studying a single problem such i.e. in binary text classification requires modern GPU hardware and expensive training. A second level of optimisation as posed with NAS problems grows the requirements exponentially: confounding hyperparameters need to be considered and experiments require carefully observed repetitions to reach statistically significant results. DARTS presents a clever trade-off between computational costs and achievable results in NAS but also finds its limitations in architectural expressiveness or openness in search space. Either such trade-offs need to be pursued in future or new formal analytical ways need to be found to capture NAS and derive stronger statements.

## 15.4 FUTURE WORK & OUTLOOK

Beyond the presented work, we see diverse possibilities to advance either the formal analysis of graph-induced nets, the structure analyses, or especially automated methods for neural architecture search.

### *Capturing Structural Properties*

We only covered a small amount of interesting structural properties in our empirical experiments. In section 9.6 on page 174 we mentioned that Waqas et al. found the "robustness of trained DNNs highly correlated with their graph measures of entropy and curvature" [296, p. 8]. These are two properties we have not considered, yet. We suppose, that a strong factor will nonetheless difficult to be found as confounding hyperparameters make a search and analysis difficult. Also, the overall factor of structure would stronger separate the CT-NAS data but the high variance and differences across datasets suggest that trivial structural properties or simple linear relationships are unlikely to expect.

### *Structure Estimation from Deep Neural Networks*

Another perspective is to be gained by treating the problems and solution methods as a roundtrip between deep neural networks and their underlying structure. Extracting the structure of a given deep neural networks realisation is the inverse direction of transforming a graph into a neural network and can be posed as an estimation problem. To conduct such a roundtrip as illustrated in fig. 15.1 on the following page, a structure estimation has to be made from a given deep neural network realisation.

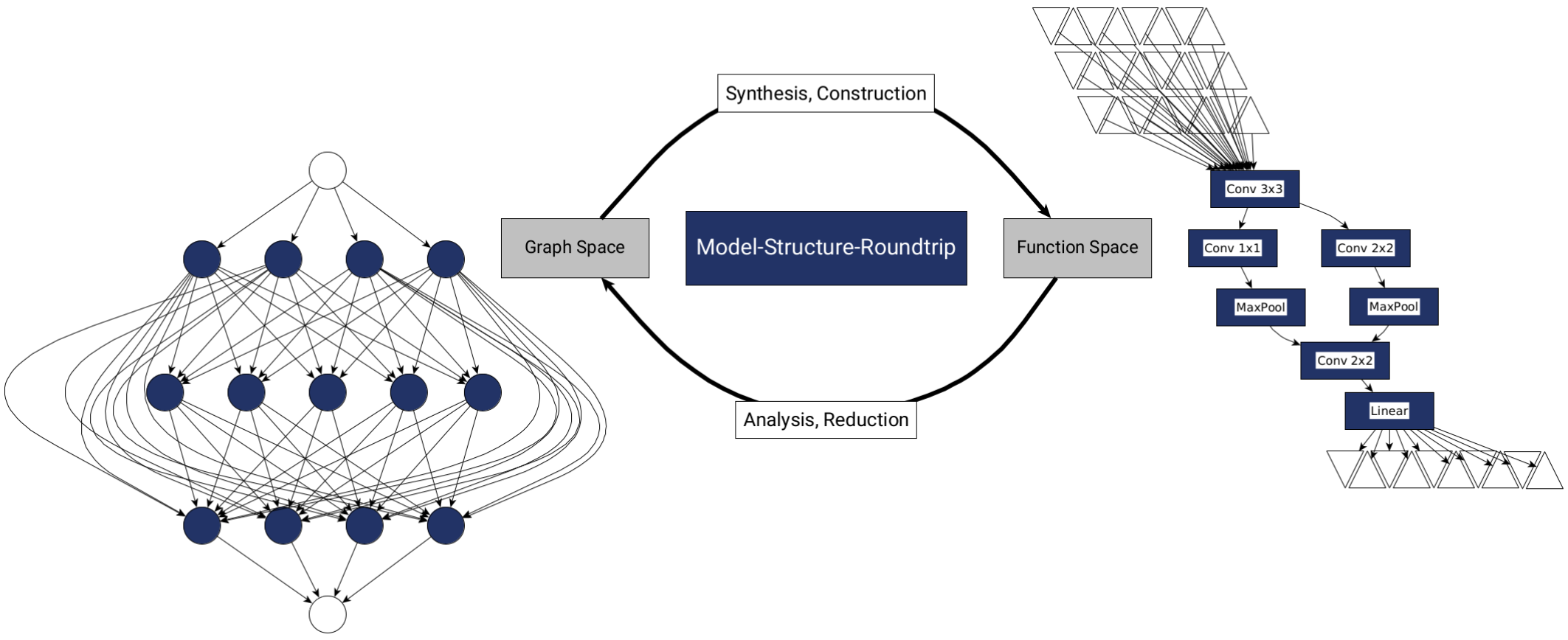

**Figure 15.1:** The problems, methods and experiments in this work can be viewed under the perspective of a model-structure roundtrip. Constructing a deep learning model from a graph is a synthetic step that imposes a structural prior on the considered space of possible solutions for a given data-driven problem. Reducing a deep neural network to its underlying structure is a reductive or analytical step with analogy to a forgetful functor in a category theoretical perspective [23, 254]. Taking the roundtrip one gains knowledge and understanding about general structural patterns.

We suppose that interesting new problems could be posed with such a roundtrip. Consider $T_2 \sim Pr(\mathbb{S} \mid f \in \mathcal{A}([T_1]))$, i.e. estimating a structural theme from a neural network realisation. The topology of search spaces could be analysed by considering spaces with stable back-estimation properties:

$$\arg\min_{\mathbb{S}} d(T_1, T_2 \sim Pr(\mathbb{S} \mid f \in \mathcal{A}([T_1])))$$

Here $d(\cdot)$ is a notion of graph distance (negative graph similarity), i.e. the graph edit distance. Back-estimation stability would be defined by a small change between construction from $T_1$ and estimation of $T_2$. In other words, this optimisation problem on the level of search space design (compare section 7.2 on page 130) would force a stronger separability over structural themes compared to other search spaces. Search space design was formulated as:

$$\arg\min_{\mathbb{S}\in\mathcal{P}(DAG)} \arg\min_{T\in\mathbb{S}} \mathcal{L}_{val}(T, \arg\min_{f\in\mathcal{A}([T])^{d_1\to d_2}} \mathcal{L}_{train}(f, D_{train})) \quad (eq.\ (7.3))$$

A search space such as layer-wise deep neural networks would be discarded over a search space which only distinguishes between very wide or very deep architectures because larger architectures of the layer-wise search space include smaller architectures.

*Optimisers with Neural Growth*

Pruning is an example for an interruption of the end-to-end differentiable learning to enforce solutions that are not always directly accessible by first-order optimisation. In the case of pruning the goal is to achieve compression. We conducted several experiments with growing deep neural networks [131] but it is currently unclear what benefit growing strategies would give over training a large DNN directly.

Some studies in the field of Neural Tangent Kernels [116] suggest that stochastic gradient descent actually stays relatively close to the initialisation and apparently does not explore the full parametric space for convergence. While this might have beneficial aspects to it, the field of universal approximation theorems brought up several constructive algorithms to trade off depth for width by employing strategies with binary coding of representations. If solutions that are found by SGD significantly differ from principles of such constructions, we suppose that it should be possible to investigate on interrupting end-to-end differentiable learning with reconstruction methods. Such an optimiser with neural growth or transformed representation coding could not only enforce a certain structure but also open gradient-based searches to start off from a new temporary solution landscape.

*Differential Pruning*

While there might be potential benefits in interrupting the end-to-end differentiable learning of deep neural networks under certain circumstances, the long standing successful history of SGD (and derivates) could also incorporate beneficial aspects of pruning by not interrupting the optimisation steps but incorporate new regularisation techniques similar to $l_1$- or $l_2$-regularisation. A goal of such a regularisation would be to incorporate structured pruning into the learning objective. We incorporated masks into LIB DeepStruct which can not only be used for pruning but in which the binary mask values itself can be used as parameters of a model. The mask then contains pre-activated information (logits) that can be used in connection with e.g. a Gumbel-Softmax sampling to achieve pruning in a similar manner as DARTS but on a very fine-grained level. Whether and how this approach relates to existing pruning methods is a question of further studies.

*Structure & Modularity*

Another potential application is structure in terms of modularity and representation. Making structure explicit allows to impose learning of representations on modular sub-structures. Klabunde et al. [132] provide an insightful survey on functional and representational measures. Such measures could not only be leveraged to analyse structure (such as discovering sub-modules) but also to impose modularity by enforcing functional or representational dissimilarity. This would fall into a different type of second-order optimisation than our proposed NAS eq. (7.2) on page 122, bearing parallels to the game theoretic setting of generative adversarial networks [80]. Two sub-modules could learn different parts of a task while minimising similarity. This could be a first step towards composability or similar benefits arising from modularity.

*Regularisation in Graph Embedding Space*

Finally, in context of maximum likelihood estimation we outlined in section 5.4 on page 60 how neural architecture search can be considered as imposing a prior probability in the general case of maximum a-posteriori estimation defined in eq. (5.9) on page 62. With proper distributions such as the Gaussian or Laplace distributions the problems can be approximated by what is known as $l_2$- and $l_1$-regularisation, respectively. A learned distribution for graphs $Pr(\mathcal{G})$ can be practically used as structural prior during learning. The work on MCGRAN showed, that this is already feasible with a conditional generative model and we suppose that this approach provides interesting applications. Structurally regularizing towards e.g. a high parallelisability in a probabilistic way could provide a tool to trade-off beneficial hardware-aware objectives and i.e. generalisation objectives.

## 15.5 FINAL REMARKS

Studying structure touches problems across different research fields: the transition or interplay between discrete and continuous problems, scalability considerations of deep learning, combinatorial problems of exponentially growing spaces over graphs, concerns of automation of (still) complex engineering tasks, questions in approximation theory on trading of depth with more structure, representing and learning graphs, and many more. We believe that contributions in these fields will also advance applications of neural architecture search. Conversely, contributions such as our work on graph generative models or deep state machines might help advancing other fields. A good example for that is DARTS, coming from differentiable relaxation of the bi-level optimisation problem, and FaDE which might advance into an optimisation method beyond NAS.

The alternation between structure analysis, applied NAS, and formal study of graph-induced neural networks reveals multiple perspectives on possibilities to advance automation of neural pipelines and thereby AutoML. Our hope is that the presented formulation of graph-induced neural networks, connected ideas and the discussed limitations contribute in advancing the understanding of structures of deep neural networks.

Part VIII

# APPENDIX

# GLOSSARY

**$F_1$ score** The $F_1$ score is the harmonic mean between precision and recall and commonly used for the evaluation of classification models. As an evaluation measure for classifiers we present it in section 5.8.1 on page 87 because of its importance in our experiments. 86, 87, 89, 125, 149, 150, 163, 168, 173, 179, 185–191, 195, 197, 256, 269, 286

**20newsgroups dataset** The 20newsgroups dataset is a text dataset of about 20,000 documents from internet newsgroups categorized into 20 classes. We used it during pruning neural classifiers based on Shapley Values [270]. 139

**accuracy** Accuracy generally determines how close a measurement is to its true value; compare eq. (5.19) on page 89 in section 5.8.1. In context of classification, it is often quantified by the proportion of correct predictions and all made predictions. The ISO 5725 "uses two terms *trueness* and *precision* to describe the accuracy of a measurement method. *Trueness* refers to the degree of agreement between the average value of a large number of test results and the true or accepted reference value. *Precision* refers to the degree of agreement between test results" [1, Sec 0.1]. 20, 86–89, 119, 159–162, 166, 167, 173, 176, 178, 182, 197, 209, 211, 212, 248–250, 268, 283, 286

**activation function** An activation function is a (pointwise) non-linear, usually one-sided (mostly lower) bounded continuous and differentiable function that is applied in deep neural networks between learnable parameterized affine transformations. Employing such non-linearities are the reason for deep neural networks being non-linear and forcing a most basic notion of structure by separating different linear layers. 4, 5, 19, 61, 64–69, 85, 105, 107, 113, 132, 158, 159, 161, 164, 165, 237, 291, 292

**acyclic digraph** An acyclic digraph is a digraph that contains no cycle. 28

**adjacency matrix** An adjacency matrix is a binary square matrix that indicates for $a_{ij}$ whether vertex $v_i \in V(G)$ of row $i$ is connected to vertex $v_j \in V(G)$ (and the other way around for undirected graphs). 29, 85, 260, 261, 263, 275

**Artificial Intelligence** Artificial Intelligence refers to the endeavour of discovering and developing methods, techniques, tools and an understanding for creating machines that rationalize and act intelligent when compared to human potential and their emergent intelligence. 9, **artificial neural network** An artificial neural network is a biologically inspired statistical model to learn from data and solve non-linear problems. The term is usually used synonymously to deep neural networks. 11

**AutoML** Automated machine learning is a field concerned with the automation of the full machine learning pipeline from data preprocessing techniques up to the automated statistic summarisation after training, validation, and evaluation of models through data. 217, 286, 296

**backpropagation** Backpropagation is the application of the chain rule for differentiation of neural nets to obtain gradients of a loss. We motivate and explain the details of backpropagation in section 5.6 on page 70 as the underlying force for learning of graph-induced neural networks. 70–72, 105

**Barabasi-Albert model** The Barabasi-Albert model is a graph model for networks with a scale-free degree distribution. 31, 34, 35, 134, 164, 166, 265, 274–277, 285, 288, 292

**batch normalisation** Batch normalisation is a parameterisable layer which normalizes data across a batch-wise dimension and can additionally shift or scale the underlying data. 79, 82–84, 161

**Bernoulli distribution** The Bernoulli distribution $\text{Ber}(p)$ models a single experiment with a 0-1-outcome (discrete) with a success probability of $p$ and a failure probability of $1-p$; i.e. for $X \sim \text{Ber}(p)$ then $Pr(X=1)=p$ and $Pr(X=0)=1-p$. 54, 129, 267

**bias** The bias in context of DNNs refers besides weights to one of two types of parameters which usually are additive terms to the aggregated signal present at a neuron. 61, 64, 65, 104, 105, 107, 108, 165, 210, 237

**binomial coefficient** The binomial coefficient $\binom{n}{k}$ is given as $\frac{n!}{k!(n-k)!}$ where $n!$ is the factorial of $n \in \mathbb{N}$. We also use $C_k^n = \binom{n}{k}$ for more compact equations. 19, 54, 213

**binomial distribution** The binomial distribution Bin (n,p) models $n$ independent experiments with a 0-1-outcome with a success probability of $p$ and a failure probability of $1-p$. 54

**biological neural network** Biological neural networks are the networks of neurons, axons, dendrites and synapses in the nervous system of animals. 3, 65, 117

**CGP** Cartesian Genetic Programming is an evolutionary computing approach that uses directed acyclic graphs restricted with vertices on a grid such that acyclicity is ensured and certain properties of the resulting graphs can be influenced. 224

**CIFAR** The Canadian Institute for Advanced Research is a well-known research organisation. Its importance in context of this work are the provided data sets on small colored images organized into different groups, see CIFAR10 and CIFAR100. 303

**CIFAR10** CIFAR10 is a database of Canadian Institute for Advanced Research (CIFAR) based on the Tiny Images dataset. 140, 143, 146, 149, 150, 154, 158–160, 166, 173, 180–184, 186–188, 190, 192–198, 217, 301, 303

**CIFAR100** CIFAR100 is a database of CIFAR based on the Tiny Images dataset containing 100 classes grouped into 20 superclasses. 303

**clustering coefficient** Clusterings expresses how tightly clustered the neighborhood of a vertex is. For the local clustering coefficient, that means that a small value close to zero means that almost no neighbors of a particular vertex are connected while a value of one means that all neighbors of a vertex are connected among each other. 35, 36, 91

**CMA-ES** Covariance Matrix Adaption Evolutionary Strategy is an evolutionary inspired optimisation method for continuous search spaces which employs a population of points for evaluation and search with support of directions obtained from the covariance of the evaluated population. 221, 239, 253

**coefficient of determination** The coefficient of determination, denoted with $R^2$, "measures the proportion of variability in $Y$ that can be explained using $X$" [117, Sec. 3.1, p. 79]. We provide its definition in eq. (5.22) on page 90. 87, 89, 90, 167, 171, 248, 249

**computational theme** Computational themes are a subset of directed acyclic graphs and by its definition each computational theme $T$ is a structural theme that induces a universal architecture $\mathcal{A}([T])$. 16, 129, 131–133, 146, 156, 173, 175, 178–180, 182, 189, 190, 196–198, 217, 284, 285, 292

**confusion matrix** A confusion matrix is spanned by the actual class targets as available from data and the predicted class from a classifier such that various evaluation scores can be easily obtained from it. 87, 88

**connection** Connections between neurons carry weights as learnable parameters and pass information from one neuron to another after a neuron passed down its activated signal. 65, 124

**convolutional neural network** Convolutional neural networks (CNN) use kernel matrices with learnable weights, that are slided across input data. The underlying operation is a convolution and often CNNs refer to a layer or module of that operation type. From a structural perspective they can be seen as sparsely connected layers which additionally employ weight sharing to speed up computation. 69, 79, 85, 95, 97, 122

**cross-entropy loss** The cross-entropy loss is a common loss function for multi-class classification problems; compare eq. (5.11) on

page 79 in section 5.6.3 on page 78 on loss functions. 78, 263, 264

**cross-validation** Cross-validation refers to splitting data for training, hyperparameter validation and performance testing into separate sets as to avoid overfitting and further comes in more advanced variants such as $k$-fold cross-validation. 92

**cycle** A cycle is a path graph $P$ with distinct vertices $\{v_1, \dots, v_k\}$ with $k \geq 3$ and $v_1 = v_k$. 28

**deep neural network** A deep neural network is a highly non-linear statistical model composed of multiple layers and usually learned through gradient-based methods. We introduce them in section 5.5. 3, 5, 10, 11, 30, 39, 40, 43, 51, 55, 64–66, 69, 79, 81, 82, 84, 85, 94, 97, 101, 102, 105–107, 114, 122, 123, 129, 134, 135, 153, 162, 163, 165, 169, 171, 175, 176, 197, 198, 207, 209, 216, 218, 256, 265, 267, 272, 276, 278, 283–285, 287, 289–291, 293–296, 304

**deep state machine** A deep state machine is a finite state machine whose transitions are learned by turning them into deep neural network as sub-modules of a larger differentiable model, the deep state machine. 17, 260, 271–274, 276, 278, 279, 288, 290, 296

**degree** The degree $deg_G(v)$ of a vertex $v$ in a graph $G$ is the sum of the sizes of its in- and out-neighborhood. 28, 40, 167, 168, 170, 171, 173, 175–178, 196, 247, 249, 254, 275, 276, 278, 283, 286, 288, 292

**degree centrality** The degree centrality is a centrality concept which assigns each vertex of a graph its degree as a centrality value, expressing its number of connections in the network. 35

**density** The density of a graph is the proportion of its size and its potential size of edges over all of its vertices. 29, 32, 124, 168, 170, 174, 248

**diameter** The diameter is the greatest distance between any two vertices in a graph $G$ 28, 168, 170, 171, 175

**differentiable architecture search** Differentiable Architecture Search is an umbrella term for NAS methods in which the optimisation problem for the architecture is relaxed into a continuous optimisation problem by learning differentiable parameters of a hyper-architecture jointly with weights shared across different integrated sub-modules or sub-architectures, refer to section 12.6 on page 239. 236, 237, 239, 287

**digraph** A digraph is a directed graph $G = (V, E)$, i.e. ordered pairs $(s, t) \in E$ make up the edge set $E$ for which we call $s$ the source and $t$ the target of an edge. 27, 29

**directed acyclic graph** A directed acyclic graph is a directed graph that contains no cyclic paths and thus has a layering. 3, 5, 18, 29, 38–41, 51, 74, 102, 107–111, 114, 117, 118, 128, 129, 131–135, 171, 173, 225, 257, 269, 284, 285, 291, 292, 304

**distance** The distance between two vertices $s$ (the source) and $t$ (the target) is the length of the shortest path between them. 28

**empirical risk minimisation** Empirical risk minimisation summarizes the principles of learning a model from data by searching a restricted hypothesis space through minimizing an empirical risk. Our notion of graph-induced neural networks and problems such as training and neural architecture search are an instance of empirical risk minimzation by posing graph-induced neural networks as the restricted hypothesis space of (structured & deep) models in which either a subset (architecture) or an instance (neural network realisation) is to be found. 78, 89

**Erdős-Rényi model** The Erdős-Rényi model is one of the two original random graph models. 31, 127, 134, 265, 272, 274, 276, 277, 292

**error** An error in context of loss functions and empirical risk minimisation refers to a pointwise measure $l : \mathcal{Y} \times \mathcal{Y} \rightarrow \mathbb{R}$ with $l(\hat{y}, y) \geq 0$ for an expected value $y$ and a prediction $\hat{y}$. 78, 89, 272

**FaDE-score** FaDE-scores are aggregated scores from architecture parameters of cells of a multi-stage differentiable hyper-architecture for one cell candidate. 247, 248, 250–257, 288

**geometric deep learning** Geometric deep learning refers to both the research field on and the research endeavour of "deriving different inductive biases and network architectures from first principles of symmetry and invariance" in deep learning [28]. 85, 118, 263, 292

**girth** The girth of a graph $G$ is the length of the smallest cycle contained in $G$ or $\infty$ if no cycle exists. 28, 31

**GPU** A graphic processing unit (GPU) is a specialized hardware that is optimized for operations such as high throughput matrix multiplications. 149, 172, 292

**graph convolutional network** A graph convolutional network is a deep neural network layer which applies a convolution over the laplacian of a graph.

**graph edit distance** The graph edit distance is a similarity measure between two graphs and is determined by the minimum number of basic operations required to transform one graph into the other (and vice versa). 38, 194, 196, 230, 272, 277, 293

**graph neural network** Graph neural networks are DNNs that have been transferred to the domain of graphs by e.g. operating on their graph Laplacian. 79, 85, 167, 249, 250, 267, 272

**Gumbel-Softmax** The continuous Gumbel-Softmax distribution Concrete $(\alpha, \tau)$ models a distribution over the simplex with location parameter $\alpha$ and temperature parameter $\tau$. 56, 237, 295

**holdout validation** Holdout validation refers to splitting available data samples $D_{full} = D_{train} \cup D_{test}$ with $D_{train} \cap D_{test} = \emptyset$ such that

$D_{test}$ is solely used for estimating the performance of a trained model. 92

**hyper-architecture** A hyper-architecture refers to a neural network architecture which contains sub-modules with varying paths, refer to differentiable architecture search. Discrete choices along these paths yield an architecture that is contained within the hyper-architecture. The term is also often referred to as one-shot architecture. 120, 127, 128, 236, 237, 239, 248, 250–257, 288, 290

**hyperparameter** A hyperparameter is a parameter that is not part of the optimisation process of a optimisation problem but still can be properly formulated as a parametric component of the process. Examples are the learning rate or step size of a gradient-based optimisation method or the hypothesis space in which a model can be found. Such hyperparameters can be incorporated into the optimisation process in multi-level abstractions of the problem and are often part of studies on their influence of finding solutions. 18, 68, 69, 92, 121, 145–147, 149, 153, 159, 181, 186, 187, 206, 225, 237, 238, 247, 290, 292, 293

**hyperparameter optimisation** Hyperparameter optimisation refers to the problem of finding an optimal hyperparameter setting. Usually, this involves testing the optimality on holdout validation data or more sophisticated resampling techniques. HPO can be understood as a higher level optimisation problem, i.e. when the first optimisation level concerns finding a suitable set of model parameters.

**identity matrix** An identity matrix of size $n \in \mathbb{N}$ is a square matrix $I_n \in \{0,1\}^{n \times n}$ where $I_n = (i_{uv})$ with $i_{uv} = \begin{cases} 1 & u = v \\ 0 & \text{otherwise} \end{cases}$. 19, 85, 254

**IMAGENET** IMAGENET is a large collection of vision datasets organized through the WordNet hierarchy. 81, 140

**in-neighborhood** The in-neighborhood $Nei_G^{in}(v)$ of a vertex $v$ of a graph $G$ contains all vertices having $v$ as a target. 18, 28

**integral probability metric** Integral probability metrics are a family of statistical divergence measures. 59

**Jensen-Shannon divergence** The Jensen-Shannon divergence is a statistical divergence and is important in context of other statistical divergences such as the Kullback-Leibler divergence. 58, 59

**Kullback-Leibler divergence** The Kullback-Leibler divergence is a measure of dissimilarity between two distributions $P$ and $Q$, also known as the relative entropy. In context of maximum likelihood estimation and multi-class classification models, it has a natural connection by being approximated with the cross-entropy loss. 56, 58, 60, 61, 78, 87

**Laplacian** The laplacian of a graph is the difference between the degree matrix $D$ and the adjacency matrix $\mathbf{A}$ of a graph $G$: $L = D - \mathbf{A}$. 29, 30, 86

**layer** A layer is a set of neurons with the same layer index from a topological sort of the computational graph of a neural network. 66, 104, 158–161, 178, 249, 292

**layer normalisation** A layer normalisation is an intermediate layer with learnable parameters similar to batch normalisation but with calculating the normalisation within the layer and one sample and not across multiple training samples. Details are outlined in section 5.7.2 on page 82. 79, 82–84, 161, 164

**learning rate** The learning rate is a hyperparameter, also called step size, which controls as a multiplicative factor on the gradient how much a parameter changes per update calculation during a learning process. 79, 147, 158, 159, 185, 187, 255

**loss** A loss (function) $\mathcal{L}$ is a function that calculates the error between a prediction of a machine learning model and a ground-truth information such as a class label or target value. 70, 73, 76–79, 89, 120, 209, 212, 248

**maximum likelihood estimation** Maximum likelihood estimation (MLE) refers to the principle of estimating the parameters of a model w.r.t. data by maximizing a likelihood function for which this work usually employs a neural network for approximation. MLE is the special case of maximum a-posteriori estimation with a uniform prior. 60–62, 69, 76, 105, 120, 259, 296

**maximum mean discrepancy** The maximum mean discrepancy is a distance between feature means. 56, 60, 87, 267

**mean absolute error** The mean absolute error is a point-wise error $|f(x) - y|$ for a target value $y$ and the associated prediction $f(x)$, based on the $l_1$-norm. 79, 87, 89

**mean squared error** The mean squared error is a point-wise error $\|f(x) - y\|$ between a target value $y$ and its associated prediction $f(x)$ based on the $l_2$-norm. 79, 87

**metric** A metric is a function $d : M \times M \to \mathbb{R}$ with distinct properties that equips $M$ to be a metric space $(M, d)$. We denote $l_p$-norm-induced metric $d$ with $\|\cdot\|_p$ as in $d(x, y) = \|x - y\|_p$ for $x, y \in M$. 56, 57, 79

**MNIST** MNIST refers to a database of handwritten images. 5, 10, 51, 139, 140, 143, 145, 146, 149, 150, 154, 158–160, 166, 173, 180–184, 190, 192–195, 197, 198, 301, 305

**NAS-Bench-101** NAS-Bench-101 is a benchmark dataset of neural architectures of order up to seven vertices evaluated on image

classification tasks. 129–131, 154, 158, 217, 220, 226–233, 235, 236, 239, 249, 268–270, 288, 290

**neighborhood** The neighborhood $Nei_G(v)$ of a vertex $v$ of a graph $G$ is the set of vertices with edges having $v$ as source or target. 28, 132

**neural architecture search** Neural architecture search denotes the research field and process of finding or designing neural networks in an automated way. Under the proposed framework of graph-induced neural networks, it can be seen as the automated search for prior architectural structures for neural networks, i.e. a search for beneficial DAG structures. 3, 17, 30, 43, 63, 93, 108, 109, 114, 116, 119, 127, 128, 131, 154, 155, 179, 185, 199, 211, 212, 216–221, 226, 236–239, 244, 245, 247, 248, 250, 252, 256, 257, 259, 260, 271, 277–279, 283, 288, 289, 292, 293, 296

**neuron** Neurons are the basic building block of a DNN and aggregate information from incoming connections and squish the aggregate through an activation function. 11, 65, 104, 124, 291

**NFL** No Free Lunch theorem 139

**No Free Lunch theorem** The No Free Lunch theorems are results on optimisation problems which roughly state, that, without resampling, all algorithms for optimisation perform equally well. The NFL theorems are related to this work in that they pose an important theoretical consideration when looking at universal architectures for which we investigate on structural differences. 129, 139

**node centrality** Node centralities are used to assign values to vertices to identify structurally important ones. 35

**norm** A $l_p$-norm is a function associating a vector with a real-value with $\|x\|_p \triangleq (\sum_{j=1}^{n} |x_j|^p)^{1/p}$ for vectors $x = (x_1, \dots, x_n) \in \mathbb{R}^n$. 56, 57, 60, 79, 206

**one-hot encoding** A one-hot encoding is a representation with multiple dimensions in which ones or zeros are used to denote whether a certain feature or label is present or not for the object to represent. We can write $\mathbf{x} \in \{0,1\}^d, d \in \mathbb{N}^+$ for a $d$-dimensional one-hot encoding. 30, 55, 276, 278

**order** The order $|G|$ of a graph $G$ is its number of vertices $|G| = |V(G)|$. 19, 27, 133, 134, 154, 174, 176, 236, 254, 274, 276–278

**out-neighborhood** The out-neighborhood $Nei_G^{out}(v)$ of a vertex $v$ of a graph $G$ contains all vertices having $v$ as a source. 28

**path** A path is a graph $P$ with distinct vertices and its edges following the pattern $\{v_0v_1, v_1v_2, \dots, v_{k-1}v_k\}$. 27, 154, 170, 171, 174, 175, 177, 283, 286, 292

**payoff** In context of coalitional games as described in section 4.1, a payoff is a (set) function assigning a contributional value to a coalition $S \subseteq U$ of a set of players $U$ (the universe). 43–46, 210–212, 216

**Pearson's** $\rho$ The sample correlation coefficient $\rho(X, Y)$ is defined in eq. (5.23) on page 90 and provides a "measure of the linear relationship between $X$ and $Y$" [117, p. 79]. 87, 89, 90

**Penn Treebank corpus** The Penn Treebank corpus is a dataset "consisting of over 4.5 million words of American English" [189] and is mostly used for Part-of-Speech tagging. 217

**performance estimation strategy** A performance estimation strategy determines how a performance score (such as a test set accuracy) is obtained for a candidate from a search space during a neural architecture search. 155, 218, 219, 221, 244, 252, 256, 257, 288

**precision** In context of classification, precision refers to the proportion of number of samples correctly associated to a class under investigation and the number of samples associated to any other class. Precision in classification is therefore measured w.r.t. a single class but can be aggregated into an overall measurement through e.g. micro- or macro-averaging.
The ISO 5725 also "uses *precision* to describe the accuracy of a measurement method" where "*precision* refers to the degree of agreement between test results" [1, Sec 0.1]. 86–89, 91, 173, 197

**pruning** Pruning refers to a set of techniques removing structural elements in a top-down fashion. 43, 123, 203, 204, 210, 216, 218, 287, 294, 295, 304–306

**random forest** A random forest is an ensemble of trees combined through the technique of bagging [117, Sec. 8.2.2]. 166, 170–172, 179, 248, 249, 288

**recall** is an evaluation measure for classification models and resembles the proportion of true positives to both true positives and false negatives. Recall is together with precision the underlying measure for the $F_1$ score. 86–89, 173, 197

**rectified linear unit** The rectified linear unit (ReLU) is an activation function defined pointwise as $x \mapsto max(x, 0)$ and therefore a one-sided bounded non-linearity commonly used in deep learning. 19, 67, 68, 161

**search space** The search space of a NAS defines the structure of the space of candidate in which a solution is searched. Typical underlying structures of such a definition are binary sequences, grammars or graphs. 176, 185, 199, 217–219, 235, 250, 253, 254, 257, 260, 283, 287, 288

**search strategy** A search strategy of a NAS determines rules on how to manoeuvre through a search space such that a solution candidate with good final performance can be found. Simple types of such strategies are evolutionary searches with random sampling or genetic algorithms. 185, 189, 204, 217, 218, 221, 253

**Shapley Value** The Shapley Value is a solution concept for a game-theoretical setting based on marginal payoff. 16, 22, 43–50, 204, 209–213, 215, 216, 287, 290

**size** The size $||G||$ of a graph $G$ is its number of edges $||G|| = |E(G)|$. 27, 174

**source** The source of an edge $(s, t) \in E(G)$ of a graph $G$ is its first vertex, often also called tail. 27

**Spearman's** $\rho$ Spearman's $\rho$ is a measure of similarity between two rankings and defined in section 5.8.2 on page 89. 89–91

**SpheresUDCR** SpheresUDCR is an artificial classification data set based on random point- and radius-sampling such that spheres of $d$-dimensional points need to be associated with a class (such as a color). 22, 140, 141, 144–146, 149, 154, 166, 180, 183, 184, 190

**stochastic gradient descent** Stochastic gradient descent is a family of gradient-based optimisation techniques. 70, 79, 121, 294

**structural theme** A structural theme is an equivalence class representation $T$ of graphs such that $[T]$ induces a universal architecture $\mathcal{A}([T])$ through a constructive step defined by $\mathcal{A}(\cdot)$. 15, 16, 114, 119, 122, 123, 126, 128, 129, 132, 134, 146, 156–160, 162–164, 179, 185, 187, 210, 211, 225, 250, 251, 253–255, 283, 284, 286, 287, 289, 291–294

**SVHN** SVHN refers to a database of images of **s**treet **v**iew **h**ouse **n**umbers. 139, 140, 158–160, 305

**target** The target of an edge $(s, t) \in E(G)$ of a digraph $G$ is its second vertex, often also called head. 27

**UCI dataset** The UCI dataset repository is a famous machine learning repository of the University of California Irvine containing over 600 datasets. 139

**universal approximation property** The term *universal approximation property* refers to a set of neural networks lying dense in a common function analytical space such as the continuous functions w.r.t. some convergence criterion and theoretically guarantees a certain capability of expressiveness of the neural network set. 20, 51, 92, 94, 95, 97, 110, 112–115, 118, 122

**universal approximation theorem** Universal approximation theorems are results on certain forms of neural networks that lie dense in a common functional analytical space. Such theorems provide necessities and proof for universal approximation properties of neural networks. 5, 51, 93–95, 97, 98, 102, 108, 112, 124, 156, 160, 295

**universal architecture** We call a set of neural network realisations $\mathcal{A}([T])$, constructed from a structural theme $T$, a universal architecture iff $\mathcal{A}([T])$ has a universal approximation property. A definition is given in definition 2 on page 114. The idea of a universal architecture is to work with sets of functions which have comparable expressiveness but originate from different structures. 15, 51, 101, 108, 113, 114, 119, 129–133, 135, 156, 157, 162–164, 283, 284, 289, 291, 304

**variation operator** Variation operator (Var. Ops.) is the collective term for methods in genetic algorithms to derive new candidates. Mutations are an example for unary variation operations and if graphs are used for search space representation, a variation operator can also be a a cross-over operator on graphs such as a subgraph insertion. If the genotypical representations are based on graphs, variation operators are operations over graphs. 16, 109, 221, 223–227, 229, 230, 232–236, 239, 253, 287, 290

**vertex permutation** A vertex permutation function $\pi$ is a permutation function over the vertex set $V(G)$ of a graph $G$ such that $(v_1, \ldots, v_n)$ with $\forall i \in \{1, \ldots, n\} : v_i \in V(G)$ are permuted into $(\pi(v_1), \ldots, \pi(v_n))$. 29, 85, 260

**Watts-Strogatz model** The Watts-Strogatz model is a graph generative model for small-world networks. 31–33, 127, 134, 164, 166, 265, 266, 277, 285, 292

**weight** The weight of a DNN is one of two main types of learnable parameters and referred to with character symbol "w" or included in the set of model parameters $\theta$. 61, 64, 104–108, 165, 210, 237

# A ADDITIONAL RESOURCES

*The following entails:*

## A.1 STATISTICAL DETAILS OF CRITICAL DIFFERENCES OF CTS

We conducted the experiments with LIB AUTORANK [98] in section 10.3 on page 198 not only on `MNIST` or `CIFAR10`, but also on `spheres-0a19afe4` (⊛), `spheres-23aeba4d` (✠), `spheres-6598864b` (⨁), `spheres-b758e9f4` (▷), `spheres-b8c16fd7` (♠), and `spheres-bee36cd9` (♥). The resulting statistics are provided in tables A.1 to A.3 for selected datasets and all remaining critical difference plots are provided in figs. A.1 to A.3.

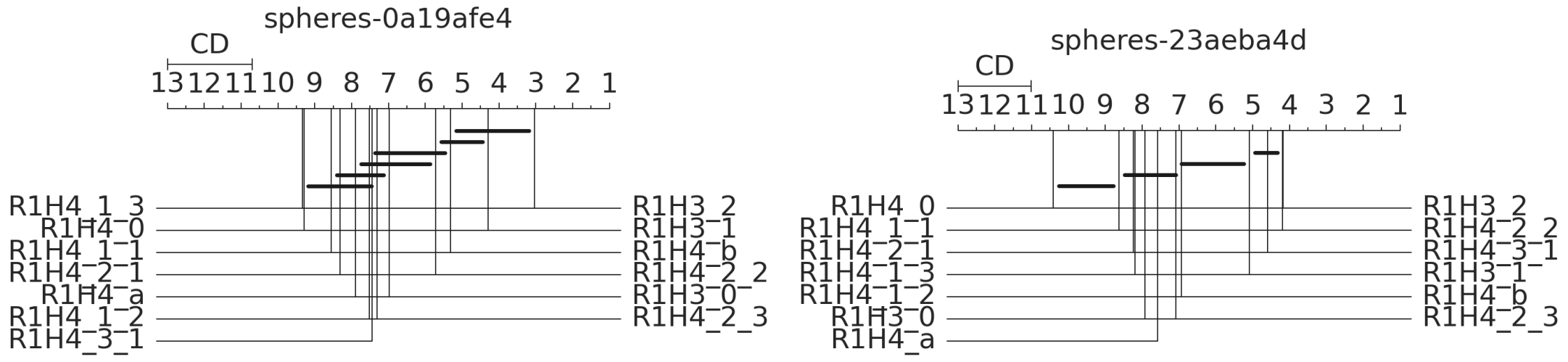

**Figure A.1:** Critical difference plot for thirteen computational themes evaluated on `spheres-0a19afe4` (⊛) (left) and `spheres-23aeba4d` (✠) (right)

## A.2 ADDITIONAL RESOURCES ON LAYER-WISE FFNS

Additional analyses as presented in fig. A.4 have been conducted.

## A.3 THE MNIST DIGIT CLASSIFICATION DATA

Many of our evaluations have been conducted with MNIST [148] which can be considered as a minimum benchmark for deep learning. With an error percentage way below 1% it can be considered as solved. The most successful classication models are using deep convolutional architectures, variants of capsule networks or ensembles thereof. Especially because its data quality of pre-processed images, the relatively small amount of image dimensions, no need for color channels and the large amount of samples make MNIST a very well suited benchmark for comparing algorithms on relatively small scale. Training a deep neural network on an NVIDIA GeForce GTX 1080 graphics card with 8192 MB memory for more than 50 epochs over the full training set usually takes less than few minutes. Thus, a performance estimate with an appropriate training schedule can be obtained in a reasonable amount of time in comparison to large-scale vision or natural language processing tasks.

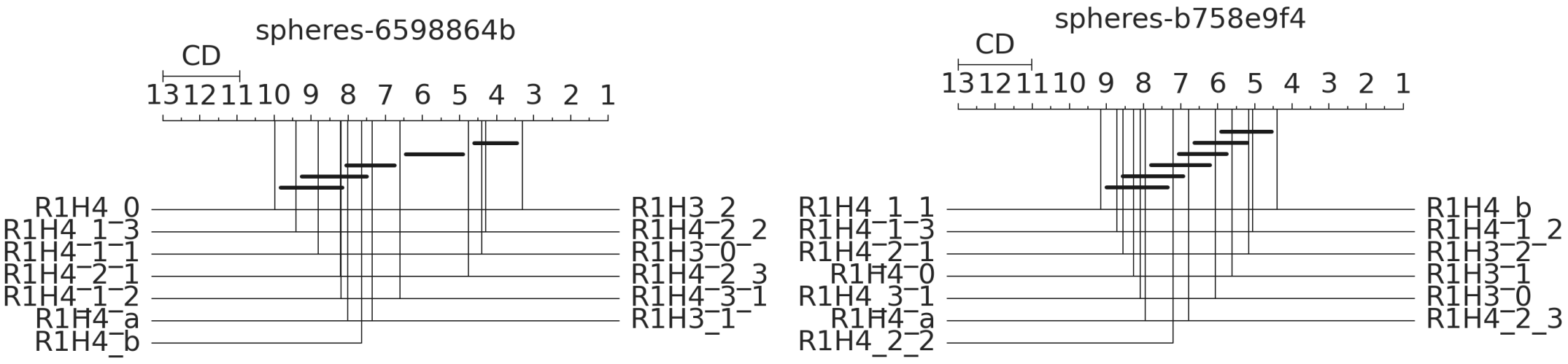

**Figure A.2:** Critical difference plot for thirteen computational themes evaluated on `spheres-6598864b` (⨁) (left) and `spheres-b758e9f4` (▷) (right)

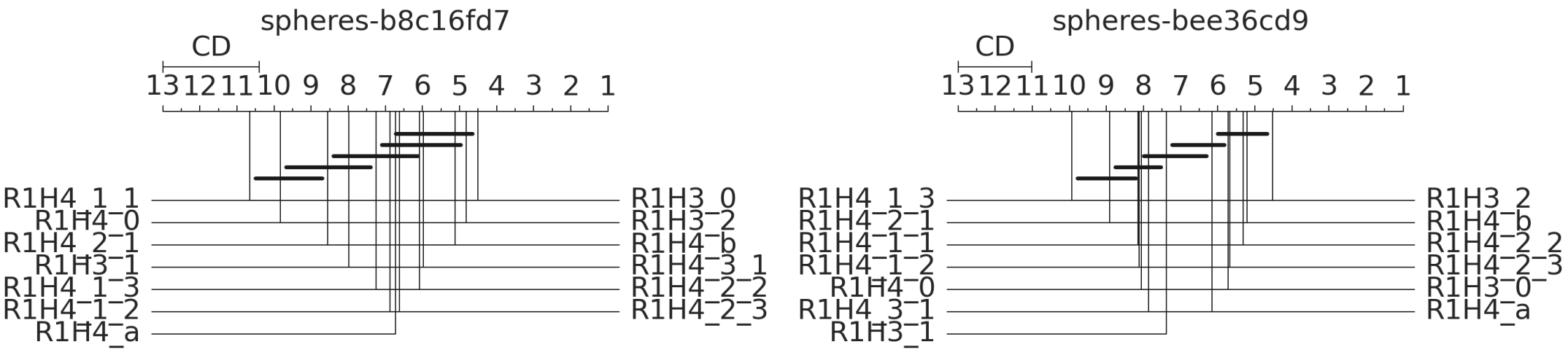

**Figure A.3:** Critical difference plot for thirteen computational themes evaluated on `spheres-b8c16fd7` (♠) (left) and `spheres-bee36cd9` (♥) (right)

We've employed the following transformation to data images as can be found on various online sources:

```
transform_mnist = torchvision.transforms.Compose([
    torchvision.transforms.ToTensor(),
    torchvision.transforms.Normalize((0.1307,), (0.3081,))])
```

The classification problem is insofar difficult that numbers "1" and "7" are often mistakenly interchanged as well as "6", "0" or "9". There exist various low-dimensional embeddings, e.g. with t-SNE or in variational auto-encoders, in which the transitions between the different classes can be observed visually. These transitions in embedding space often

| | MR | MED | MAD | CI | $\gamma$ | Magnitude |
|---|---|---|---|---|---|---|
| R1H4_b | 9.155 | 0.104 | 0.000 | [0.104, 0.159] | - | large |
| R1H4_1_2 | 8.714 | 0.104 | 0.000 | [0.104, 0.161] | - | large |
| R1H3_2 | 8.554 | 0.136 | 0.033 | [0.104, 0.161] | -0.954 | large |
| R1H3_1 | 8.262 | 0.104 | 0.000 | [0.104, 0.341] | - | large |
| R1H3_0 | 8.083 | 0.137 | 0.034 | [0.104, 0.286] | -0.954 | large |
| R1H4_2_3 | 7.952 | 0.130 | 0.026 | [0.104, 0.246] | -0.954 | large |
| R1H4_2_2 | 7.208 | 0.131 | 0.028 | [0.104, 0.402] | -0.954 | large |
| R1H4_a | 6.786 | 0.212 | 0.108 | [0.104, 0.354] | -0.954 | large |
| R1H4_3_1 | 6.060 | 0.189 | 0.085 | [0.126, 0.441] | -0.954 | large |
| R1H4_0 | 5.613 | 0.300 | 0.165 | [0.131, 0.438] | -1.138 | large |
| R1H4_2_1 | 5.161 | 0.379 | 0.104 | [0.179, 0.442] | -2.517 | large |
| R1H4_1_3 | 5.060 | 0.398 | 0.158 | [0.139, 0.495] | -1.775 | large |
| R1H4_1_1 | 4.393 | 0.411 | 0.086 | [0.320, 0.473] | -3.422 | large |

**Table A.1:** AUTORANK statistics for `spheres-b758e9f4` (▷). The associated critical difference plot is shown in fig. A.2 (right).

provide a good intuition on the separability of the particular digit classes.

## A.4 THE CIFAR-10 DATABASE OF TINY COLORED IMAGES

The CIFAR provides important databases of tiny images on which many of our experiments are built upon. We refer to them as CIFAR10, a data set of 60,000 colored images of size 32×32 over ten possible classes and to CIFAR100 which groups the same images into 100 classes [138] and therefore also provides hierarchical class information.

We've employed the following transformation in `pytorch` to samples of `CIFAR10`:

```
transform_cifar10 = torchvision.transforms.Compose([
    torchvision.transforms.ToTensor(),
    torchvision.transforms.Normalize(
        (0.5, 0.5, 0.5), (0.5, 0.5, 0.5))])
```

## A.5 TECHNICAL DETAILS OF DEEPSTRUCT

DeepStruct has the goal of providing

- tools to extract graphs from PyTorch models,
- models that can be constructed from directed acyclic graphs,
- models that represent universal architectures in the form of scalable families from which a neural network instance can be drawn,

| | MR | MED | MAD | CI | $\gamma$ | Magnitude |
|---|---|---|---|---|---|---|
| R1H3_2 | 9.929 | 0.739 | 0.110 | [0.602, 0.831] | 0.000 | negligible |
| R1H4_b | 8.917 | 0.795 | 0.101 | [0.602, 0.867] | -0.354 | small |
| R1H4_2_2 | 8.143 | 0.843 | 0.045 | [0.692, 0.879] | -0.829 | large |
| R1H4_2_3 | 8.119 | 0.826 | 0.069 | [0.661, 0.882] | -0.634 | medium |
| R1H3_0 | 8.060 | 0.835 | 0.069 | [0.514, 0.880] | -0.698 | medium |
| R1H4_a | 7.869 | 0.858 | 0.031 | [0.750, 0.879] | -0.994 | large |
| R1H3_1 | 7.387 | 0.839 | 0.051 | [0.765, 0.887] | -0.780 | medium |
| R1H4_3_1 | 6.155 | 0.872 | 0.034 | [0.822, 0.895] | -1.096 | large |
| R1H4_0 | 5.720 | 0.870 | 0.014 | [0.862, 0.884] | -1.128 | large |
| R1H4_1_2 | 5.673 | 0.879 | 0.012 | [0.868, 0.889] | -1.203 | large |
| R1H4_1_1 | 5.310 | 0.878 | 0.013 | [0.863, 0.888] | -1.197 | large |
| R1H4_2_1 | 5.208 | 0.877 | 0.020 | [0.858, 0.895] | -1.173 | large |
| R1H4_1_3 | 4.512 | 0.884 | 0.015 | [0.867, 0.895] | -1.247 | large |

**Table A.2:** AUTORANK statistics for `spheres-b8c16fd7` (♠). The associated critical difference plot is shown in fig. A.3 (left).

- and tools to schedule pruning strategies over models.

The implementation reflects technical considerations resulting out of the presented thesis such as

- graphs can be formulated in different ways upon deep neural networks such that graph extraction for the purposes of **visualisation**, **structure analysis** or model **roundtrips** is a configurable transformational process,
- a prunable model should carry an explicit mask and even pruning saliency information for the purpose of **pruning studies**, **compression** or **lottery ticket** resets,
- and the process of pruning is also configurable such that a pruning scheduling with modular strategies is a way to go forward.

An overview of the project is provided in fig. A.5.

### A.5.1 *Generalised Masked Linear Layers*

An important insight for future developments of DNN layers in context of learning or imposing structure is that it might be worth to additionally carry not only masking information but also pruning saliency information and a masking parameter structure.

Masks allow to carry information which structural elements are currently pruned in a model. Applying the mask on to the weights directly removes the learned parameter information. Pruning strategies can be switched and recompute masks if the parameter information is retained. Further, the mask or weights can be easily modified as to reset any of them, e.g. in the context of finding lottery tickets.

| | MR | MED | MAD | CI | $\gamma$ | Magnitude |
|---|---|---|---|---|---|---|
| R1H3_0 | 10.653 | 0.839 | 0.052 | [0.540, 0.884] | 0.000 | negligible |
| R1H3_2 | 9.837 | 0.848 | 0.039 | [0.736, 0.886] | -0.133 | negligible |
| R1H4_b | 8.551 | 0.868 | 0.026 | [0.789, 0.894] | -0.482 | small |
| R1H4_3_1 | 7.980 | 0.875 | 0.031 | [0.706, 0.902] | -0.577 | medium |
| R1H4_2_2 | 7.245 | 0.884 | 0.016 | [0.834, 0.899] | -0.792 | medium |
| R1H4_2_3 | 6.878 | 0.886 | 0.020 | [0.841, 0.906] | -0.807 | large |
| R1H4_a | 6.724 | 0.889 | 0.017 | [0.851, 0.907] | -0.879 | large |
| R1H4_1_2 | 6.622 | 0.889 | 0.021 | [0.630, 0.910] | -0.872 | large |
| R1H4_1_3 | 6.082 | 0.886 | 0.014 | [0.869, 0.909] | -0.844 | large |
| R1H3_1 | 5.980 | 0.893 | 0.023 | [0.848, 0.916] | -0.915 | large |
| R1H4_2_1 | 5.112 | 0.897 | 0.016 | [0.874, 0.914] | -1.030 | large |
| R1H4_0 | 4.827 | 0.900 | 0.011 | [0.884, 0.912] | -1.111 | large |
| R1H4_1_1 | 4.510 | 0.900 | 0.012 | [0.886, 0.915] | -1.101 | large |

**Table A.3:** AUTORANK statistics for `spheres-23aeba4d` (✠). The associated critical difference plot is shown in fig. A.1 (right).

Carrying additionally saliency information allows to use different pruning techniques on different structural parts of the model. Further, more expensive techniques that e.g. use first- or second-order gradients of the loss such as OBD [150] or OBS [92] over magnitude-based techniques [89] would yield saliency information that could then intermediately be stored as such.

Differentiable formulations of pruning further consider the mask as parameters. A single mask element would then not be 0 or 1 as to be an on- or off-switch but it could be a probability. Technically, this probability is then also often reflected as log-probabilities of a softmax such that the parameterisable mask would carry two parameters which would be learned.

These considerations are reflected in our `MaskedLinearLayer` implementation in LIB DeepStruct:

**Listing A.1:** Interface of a MaskedLinearLayer, extending classical Linear Layers of PyTorch

```
class MaskedLinearLayer(nn.Linear, MaskableModule):
    def __init__(self, in_feature, out_features, bias=True,
        keep_layer_input:bool=False, mask_as_params:bool=False,
        mask_param_strength:float=10.0)

    def reset_parameters(self, keep_mask=False)

```

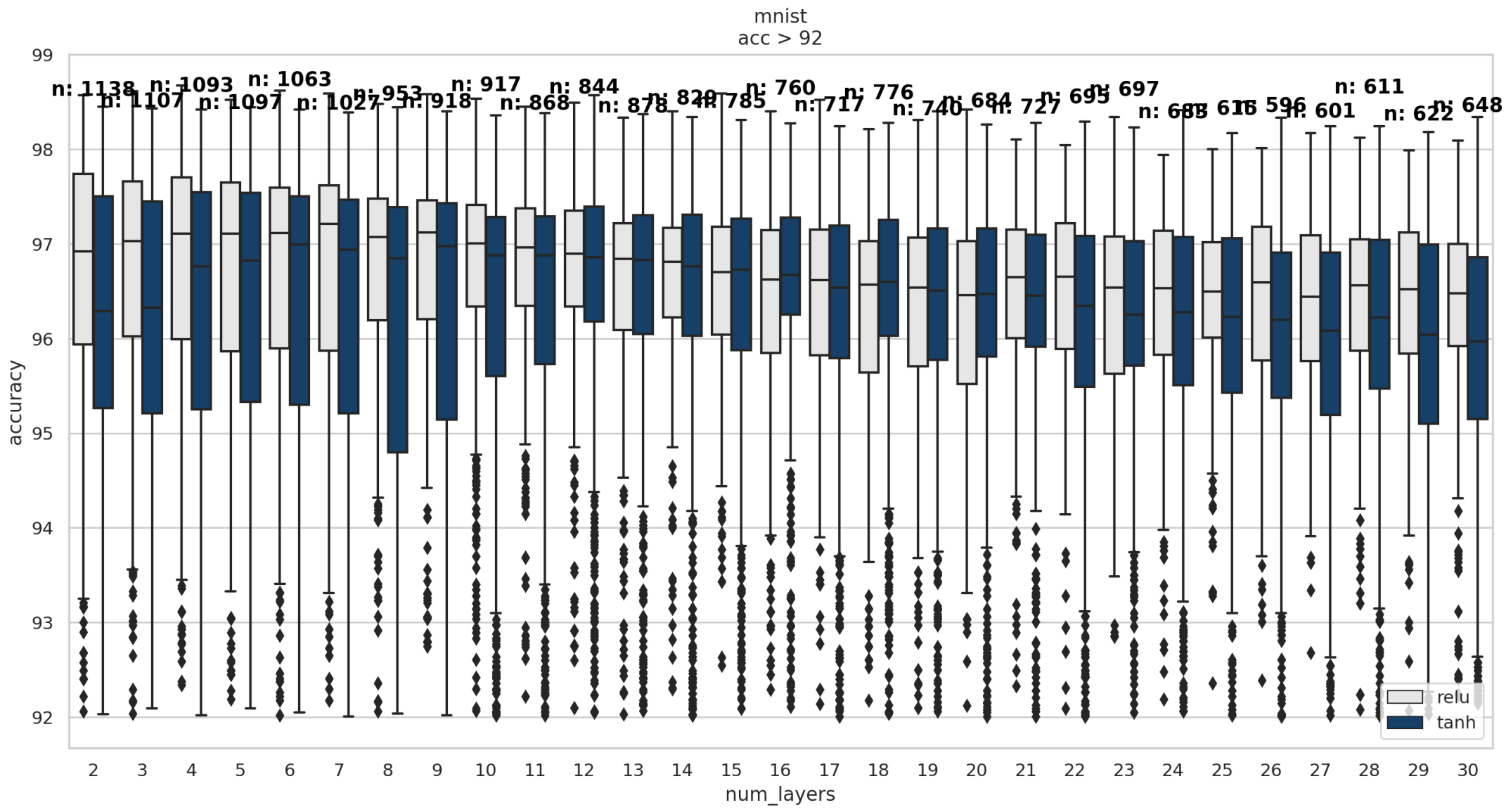

**Figure A.4:** An increasing number of layers exhibit significant signs that it becomes increasingly difficult to find as performant realisations with very deep networks as for more shallow networks. The `MNIST` is strongly related to `SVHN` except for that the latter carries a color channel and is taken from house numbers instead of handwritten digits. Compare fig. 9.4 for the same experimental setting with `SVHN`.

```
    @property
    def mask(self)

    def get_weight_count(self)

    def apply_mask(self)

    @property
    def saliency(self)

    def recompute_mask(self, theta: float = 0.001)

    def forward(self, input)
```

## A.6 UNIFORM SAMPLING DIRECTED ACYCLIC GRAPHS

We provide the algorithm of Kuipers et al. [140] in pythonic code. The code is not published due to its unconventional state.

**Listing A.2:** The algorithm presented in [140] requires some combinatorial values, e.g. the binomial coefficient for ever larger values $n$ and $k$. Caching intermediate values can speed up the calculation.

```
class DAGCombinatorics:
    _cache_a_n = {}
```

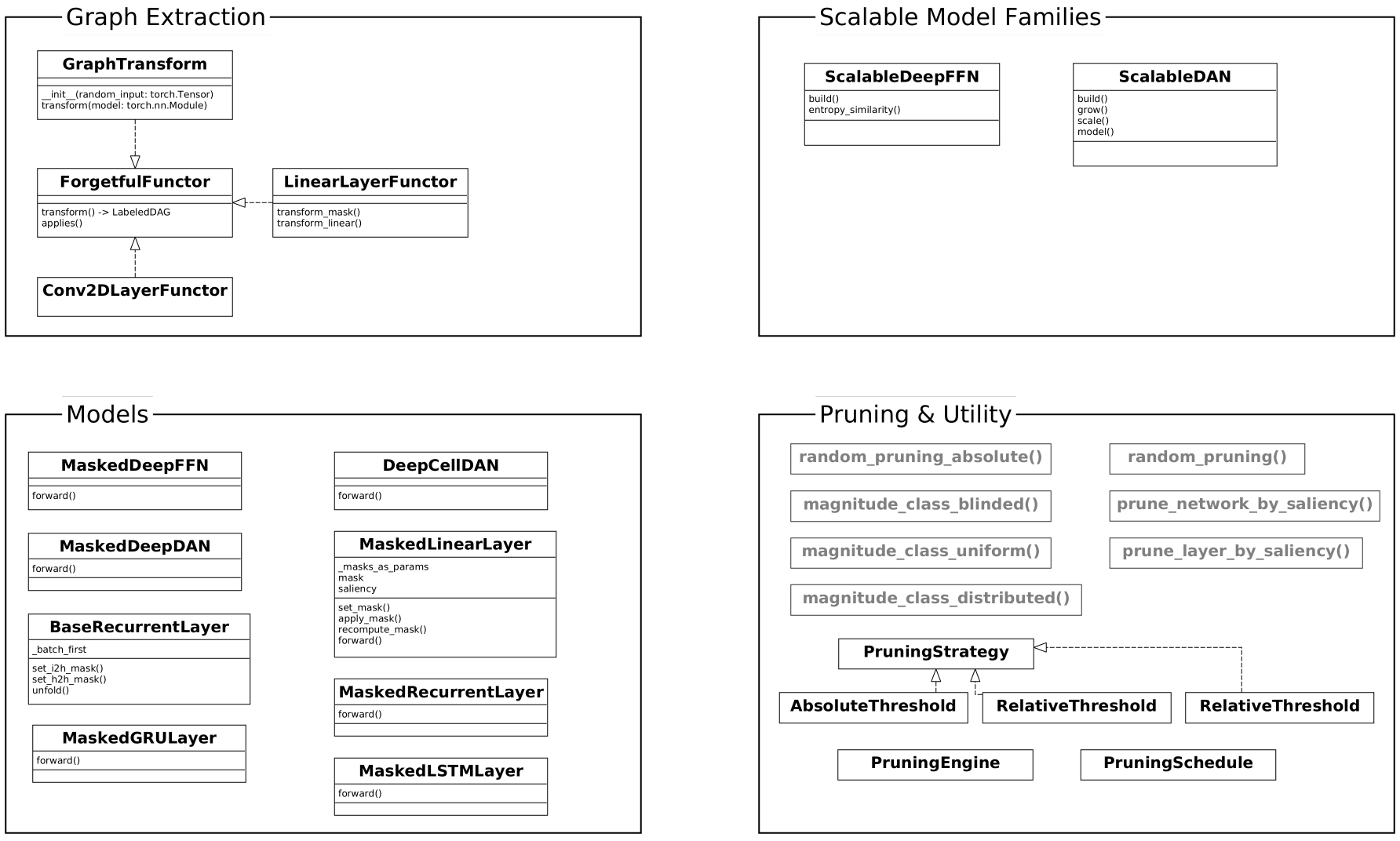

**Figure A.5:** The implementations are organised into sub-packages and stem from the mentioned purposes of *visualisation, structure analysis* or model *roundtrips*. With growing ecosystems around these purposes such as own pruning utilities within LIB PyTorch, the technical ideas reflected in DeepStruct might in future transition into separate packages.

```python
    _cache_a_nk = {}
    _cache_b_nk = {}
    _binomial_coeff_nk = {}

    def __init__(self, max_n: int = 100):
        # To avoid recursion depth, precompute caches
        for n in range(1, max_n+1):
            if n not in self._cache_a_nk:
                self._cache_a_nk[n] = {}
            if n not in self._cache_b_nk:
                self._cache_b_nk[n] = {}

            for k in range(1, n-1):
                if n == k:
                    self._cache_b_nk[n][k] = 1
                    self._cache_a_nk[n][k] = 1
                else:
                    self._cache_b_nk[n][k] = sum((2**k-1)**s*2**(k*(n-k-s))*self._cache_a_nk[n-k][s] for s in range(1, n-k+1))
                    self._cache_a_nk[n][k] = self.binom(n, k)*self._cache_b_nk[n][k]

            self._cache_a_n[n] = sum(self.a_nk(n, k) for k in range(1,n+1))

    def a_n(self, n):
        if n not in self._cache_a_n:
```

```
            self._cache_a_n[n] = sum(self.a_nk(n, k) for k in range(1,n+1))
        return self._cache_a_n[n]

    def a_nk(self, n, k):
        assert k <= n
        if n not in self._cache_a_nk:
            self._cache_a_nk[n] = {}
        if k not in self._cache_a_nk[n]:
            if n == k:
                self._cache_a_nk[n][k] = 1
            else:
                self._cache_a_nk[n][k] = self.binom(n, k)*self.b_nk(n, k)
        return self._cache_a_nk[n][k]

    def b_nk(self, n, k):
        assert k <= n
        if n not in self._cache_b_nk:
            self._cache_b_nk[n] = {}
        if k not in self._cache_b_nk[n]:
            if n == k:
                self._cache_b_nk[n][k] = 1
            else:
                self._cache_b_nk[n][k] = sum(
                    (2**k-1)**s*2**(k*(n-k-s))*self.a_nk(n-k,s) if n-k not in self._cache_a_nk or s not in self._cache_a_nk[n-k]
                    else (2**k-1)**s*2**(k*(n-k-s))*self._cache_a_nk[n-k][s]
                    for s in range(1, n-k+1)
                )
        return self._cache_b_nk[n][k]

    def binom(self, n, k):
        if n not in self._binomial_coeff_nk:
            self._binomial_coeff_nk[n] = {}
        if k not in self._binomial_coeff_nk[n]:
            self._binomial_coeff_nk[n][k] = int(scipy.special.binom(n, k))
        return self._binomial_coeff_nk[n][k]

combs = DAGCombinatorics()
```

**Listing A.3:** Helper function to sample DAGs uniformly

```
def sample_r(size_n: int):
    return random.randint(1, combs.a_n(size_n))
```

**Listing A.4:** Code for the helper function `sample_outpoints()` to sample DAGs uniformly.

```python
def sample_outpoints(n: int, r: int, cache):
    orig_n = n
    ks = {}
    k = 1
    while r > cache.a_nk(n, k):
        r = r-cache.a_nk(n, k)
        k += 1
    ix = 0
    ks[ix] = k

    # Rescaling of r in [1,a_nk] to [1, b_nk]
    # r = math.ceil(r/cache.binom(n, k))  # Results in overflow, so we use pythons integer-floor-division
    r = r // cache.binom(n, k)
    m = n-k

    while m > 0:
        s = 1
        t = (2**k-1)**s*2**(k*(m-s))*cache.a_nk(m,s)
        while r > t:
            r = r-t
            s += 1
            t = (2**k-1)**s*2**(k*(m-s))*cache.a_nk(m,s)
        r = r // (cache.binom(m,s)*(2**k-1)**s*2**(k*(m-s)))
        n = m
        k = s
        m = n-k
        ix += 1
        ks[ix] = k

    ks_arr = [ks[i] for i in ks]
    assert len(ks_arr) == ix + 1
    assert sum(ks_arr) == orig_n

    return ks_arr
```

**Listing A.5:** Code for the helper function `fill_Q()` to sample DAGs uniformly.

```python
def fill_Q(n: int, ks: list):
    Q = np.zeros((n, n))
    j = ks[-1]
    for ix in np.flip(np.arange(2, len(ks))):
        for l in range(j-ks[ix], j):
            while sum(Q[m,l] for m in range(j, j+ks[ix-1]+1)) == 0:
                for m in range(j, j+ks[ix-1]+1):
                    Q[m,l] = random.randint(0,1)
            for m in range(j+ks[ix-1]+1, n):
                Q[m,l] = random.randint(0,1)
        j = j+ks[ix-1]
    return Q
```

**Listing A.6:** Code for the helper function `permute_matrix()` to sample DAGs uniformly.

```
def permute_matrix(Q: np.array, permutation):
    n = Q.shape[0]
    R = np.zeros((n,n))
    for m in range(n):
        for l in range(n):
            R[permutation[m],permutation[l]] = Q[m,l]
    return R
```

**Listing A.7:** Function to sample (possibly unconnected) DAGs uniformly.

```
def sample_random_dag(size: int, combs=None):
    if combs is None:
        combs = DAGCombinatorics()
    r = random.randint(1, combs.a_n(size))
    ks = sample_outpoints(size, r, combs)
    Q = fill_Q(size, ks)
    perm = np.random.choice(size, size=size, replace=False)
    R = permute_matrix(Q, perm)
    graph = nx.from_numpy_array(R, create_using=nx.DiGraph)
    return graph
```

**Listing A.8:** Final wrapper function to sample connected DAGs uniformly.

```
def sample_random_connected_dag(size: int, combs=None, trials:
    int=20):
    if combs is None:
        combs = DAGCombinatorics()
    trial = 1
    result = None
    while result is None and trial < trials:
        graph = sample_random_dag(size, combs)
        if nx.is_weakly_connected(graph):
            result = graph
        trial += 1
    return result
```

# DECLARATION

The presented thesis titled "Structure of Artificial Neural Networks – Empirical Investigations" was written independently, without help, and without using other than the stated sources or tools. All adapted statements, whether verbatim or in spirit, have been identified as such.

*Passau, April 2024*

Julian J. Stier

## COLOPHON

This work is a dissertation submitted to the Faculty of Informatics & Mathematics at the University of Passau in partial fulfillment of a Ph.D.

The document was typeset using the typographical look-and-feel `classicthesis` developed by André Miede and Ivo Pletikosić, its style was inspired by Robert Bringhurst's seminal book on typography "*The Elements of Typographic Style*". Thank you for your solid groundwork.

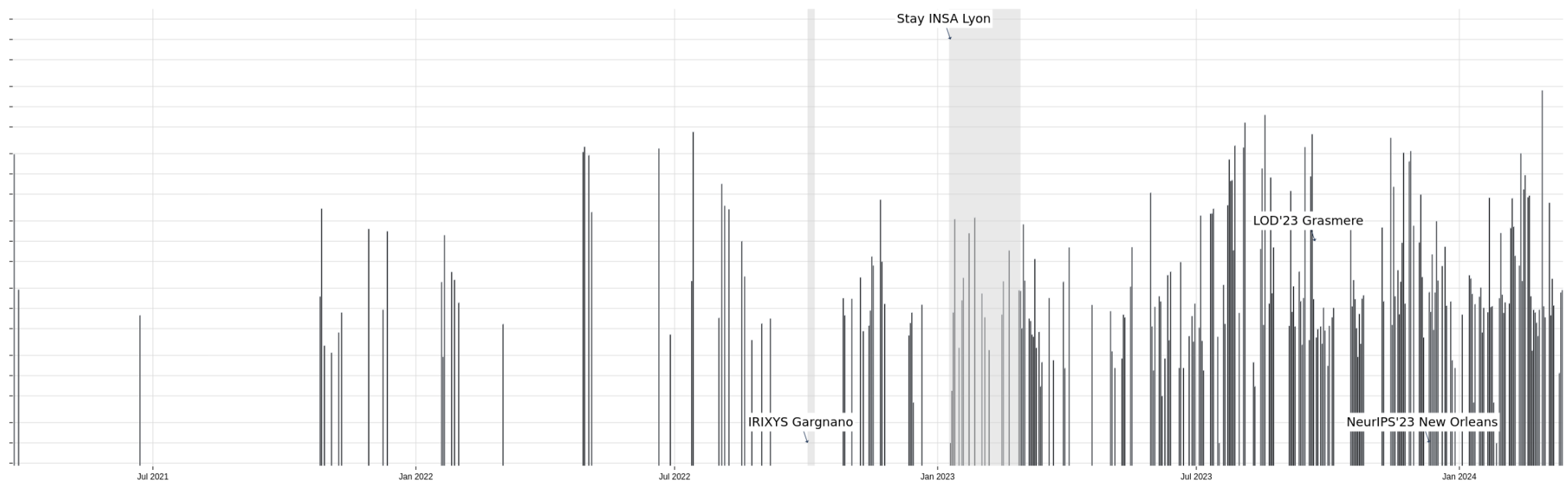

**Figure A.6:** Added lines of this document in a git repository per day on a logarithmic y-plot.